\documentclass[acmtog, ]{acmart}
\acmSubmissionID{224}

\usepackage{etoolbox}

\makeatletter
\newcommand{\suppressmainTOC}{
  \let\addcontentsline\@gobblethree
}
\newcommand{\restoreTOC}{
  \let\addcontentsline\@restored@addcontentsline
}
\let\@restored@addcontentsline\addcontentsline
\makeatother
\usepackage{hyperref}
\usepackage{booktabs} % For formal tables
\usepackage[capitalize]{cleveref}
\usepackage{booktabs}
\usepackage[ruled]{algorithm2e} % For algorithms

\SetAlFnt{\small}
\SetAlCapFnt{\small}
\SetAlCapNameFnt{\small}
\SetAlCapHSkip{0pt}

\usepackage{diagbox}
\usepackage{multirow}
\usepackage{makecell}
\usepackage{xhfill}
\usepackage{etoolbox}

\makeatletter
\DeclareRobustCommand\onedot{\futurelet\@let@token\@onedot}
\def\@onedot{\ifx\@let@token.\else.\null\fi\xspace}
\def\eg{\emph{e.g}\onedot}

\def\etal{\emph{et al}\onedot}

\makeatother

\newcommand{\rev}[1]{\textcolor{black}{#1}}
\newcommand{\methodname}{\rev{InkLayer}}  
\newcommand{\datasetname}{\rev{InkScenes}}  
\copyrightyear{2025}
\acmYear{2025}
\acmConference[SIGGRAPH Conference Papers '25]{Special Interest Group on Computer Graphics and Interactive Techniques Conference Conference Papers }{August 10--14, 2025}{Vancouver, BC, Canada}
\acmBooktitle{Special Interest Group on Computer Graphics and Interactive Techniques Conference Conference Papers (SIGGRAPH Conference Papers '25), August 10--14, 2025, Vancouver, BC, Canada}
\acmDOI{10.1145/3721238.3730606}
\acmISBN{979-8-4007-1540-2/2025/08}

\begin{document}
% Title portion
\title{Instance Segmentation of Scene Sketches Using Natural Image Priors}

% DO NOT ENTER AUTHOR INFORMATION FOR ANONYMOUS TECHNICAL PAPER SUBMISSIONS TO SIGGRAPH 2019!
\author{Mia Tang}
\email{miatang@cs.stanford.edu}
\affiliation{%
 \institution{Stanford University}
 \city{Stanford}
 \state{CA}
 \country{USA}
 }
\author{Yael Vinker}
\email{yaelvi116@gmail.com}
\affiliation{%
 \institution{MIT CSAIL}
 \city{Boston}
 \country{USA}
}
\author{Chuan Yan}
\email{chuanyan@stanford.edu}
\affiliation{%
 \institution{Stanford University}
 \city{Stanford}
 \state{CA}
 \country{USA}
 }
\author{Lvmin Zhang}
\email{lvmin@stanford.edu}
\affiliation{%
 \institution{Stanford University}
 \city{Stanford}
 \state{CA}
 \country{USA}
 }
\author{Maneesh Agrawala}
\email{maneesh@cs.stanford.edu}
\affiliation{%
 \institution{Stanford University}
 \city{Stanford}
 \state{CA}
 \country{USA}}
 
\begin{teaserfigure}
    \centering
    % \includegraphics[width=1\linewidth]{figs/teaser2-temp.pdf} 
    % \vspace{-1mm}
    \includegraphics[width=1\linewidth]{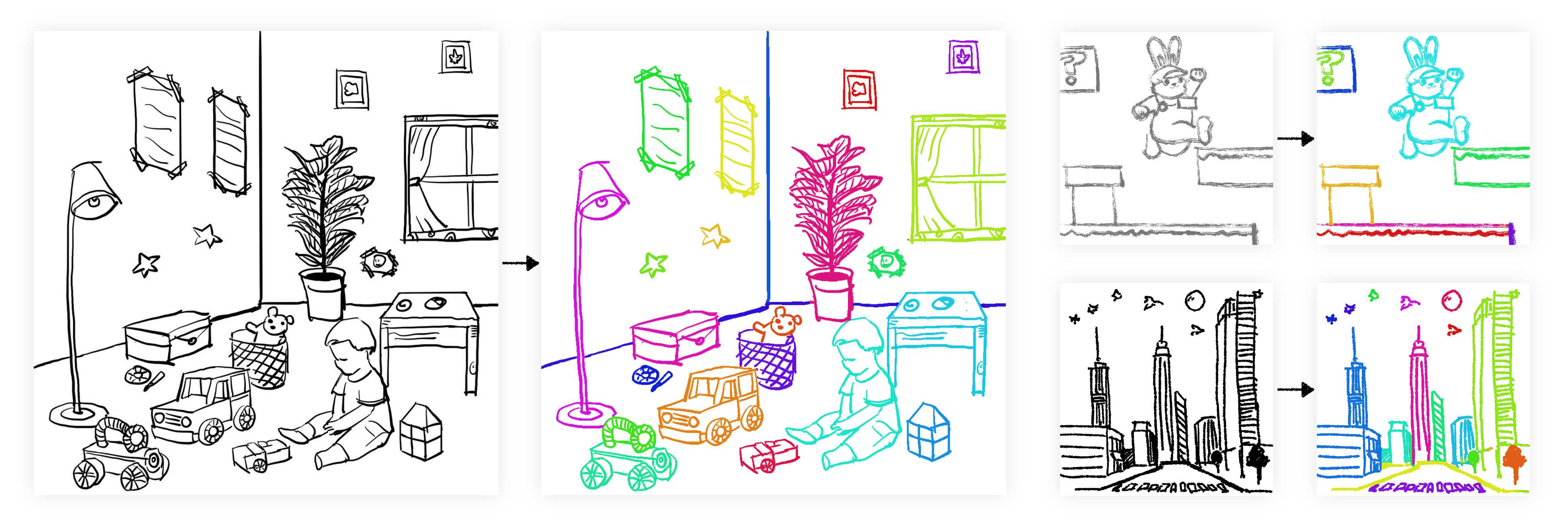} 
    % \vspace{-3mm}
    \caption{Our method performs instance segmentation of raster sketches. It effectively handles diverse types of sketches, accommodating variations in stroke style and complexity.} 
    % \caption{Our method performs instance segmentation of raster sketches and arranges them into sorted layers based on recognized object instances, enabling efficient sketch editing. It effectively handles diverse sketches, accommodating variations in stroke style and complexity.} 
    \label{fig:teaser}
\end{teaserfigure}

% \caption{Our method performs instance segmentation of raster sketches and arranges them into sorted layers based on recognized instances, enabling efficient sketch editing. It effectively handles diverse sketches, accommodating variations in stroke style and complexity.} 
%\renewcommand\shortauthors{Zhou, G. et al}
\begin{abstract}
Sketch segmentation involves grouping pixels within a sketch that belong to the same object or instance. It serves as a valuable tool for sketch editing tasks, such as moving, scaling, or removing specific components. While image segmentation models have demonstrated remarkable capabilities in recent years, sketches present unique challenges for these models due to their sparse nature and wide variation in styles.
We introduce \methodname{}, a method for instance segmentation of raster scene sketches. Our approach adapts state-of-the-art image segmentation and object detection models to the sketch domain by employing class-agnostic fine-tuning and refining segmentation masks using depth cues. Furthermore, our method organizes sketches into sorted layers, where occluded instances are inpainted, enabling advanced sketch editing applications.
As existing datasets in this domain lack variation in sketch styles, we construct a synthetic scene sketch segmentation dataset, \datasetname{}, featuring sketches with diverse brush strokes and varying levels of detail. We use this dataset to demonstrate the robustness of our approach. \rev{Code and data for this paper are released at project page: \href{https://inklayer.github.io}{\texttt{\textcolor{magenta}{https://inklayer.github.io}}}}.

% \maneesh{Contrib from slides: (1) We domain adapt object bounding box detection models from natural images to sketches using a relatively small set of class agnotic sketch data. -- can do this because domain adaptation is largely about shifting from full-color natural images to black pen strokes on white backgrounds. Exact style of sketches matters less -- but must cover some sylistic variation.}

% \maneesh{Contrib from slides: (2): We show that existing natural image segmentation models grounded with domain adapted bounding boxes perform well on sketches, but leave some ambiguity aboout strokes in the overlap regions between bounding boxes. Likely because sketches do not include all the cues (e.g. depth, texture etc.) found in natural images. -- We use priors to better resolve this issue (Not sure I follow what we are doing here.)}

% \maneesh{Should relate both of these to gestalt in the intro if possible.}
\end{abstract}




\suppressmainTOC
\maketitle

\section{Introduction}
Sketches serve as a powerful tool for visual exploration, ideation,
and planning. Traditional sketching workflows often begin with artists
working on a single canvas layer (either physical or digital) to
maintain creative momentum. As the sketch evolves and requires
refinement (e.g., adjustments to composition, perspective, or other
elements), artists face the tedious task of manually segmenting
different elements of the sketch into discrete, editable layers.
Automating the sketch segmentation process offers a promising
solution. However, this task presents unique challenges due to the
sparse and abstract nature of line drawings, as well as the inherent
variability in human sketching styles.
Existing methods for scene-level sketch segmentation typically rely on training dedicated models using annotated sketch datasets \cite{sketchydataset, Zou18SketchyScene, fscoco}. 
However, most available datasets are confined to specific sketch styles and a limited set of object categories, restricting the generalization capabilities of existing methods.

% However, constructing such datasets is challenging and requires extensive manual effort. As a result, most available datasets are confined to specific sketch styles and a limited set of categories \cite{sketchydataset, Zou18SketchyScene, fscoco}. 
% Consequently, these methods are often restricted to predefined class labels and sketch styles, limiting their generalization and applicability to real-world sketching scenarios.

% There have been active efforts to address scene-level sketch segmentation 
% \cite{bourouis2024open, Kutuk2024ClassAgnosticVS, Zou18SketchyScene}. However, existing
% works are often restricted by the limited sketch style variation across available datasets and the predefined set of class labels as shown in
% \cref{fig:existing_dataset}, limiting their generalization and applicability to sketching scenarios in the real-world. 
% Furthermore, the existing works mainly focus on semantic segmentation, 
% \mia{Only one \cite{Zou18SketchyScene} out of all the existing works is capable of instance segmentation, but is limited to a predefined set of class labels.}

In this work, we introduce \methodname{}, a method for instance segmentation of raster scene sketches that outperforms previous approaches in accommodating a wider variety of sketch styles and concepts. 
% SketchSeg produces a segmentation layer for each object instance and sorts the layers to support effective sketch editing.
We use the segmentation map to divide the sketch into sorted layers to support effective sketch editing.
% We also introduce a synthetic scene-level annotated sketch dataset that extends existing datasets by covering a broader range of object categories and a wider diversity of styles, including variations in appearance, level of detail, and stroke style.
% SketchSeg leverages the rich knowledge of large pretrained image segmentation models, adapting them to the unique domain of sketches through class-agnostic fine-tuning and the integration of depth cues.
% \begin{figure}
%   \centering
%   \includegraphics[width=0.5\linewidth]{figs/existing_datasets.png} 
%   \caption{\textcolor{red}{Mia please update}Examples of different sketch styles from existing sketch segmentation datasets: SketchyCOCO~\cite{gao2020sketchycocoimagegenerationfreehand}, SketchyScene~\cite{Zou18SketchyScene}, and CBSC~\cite{zhang2018CBSC}. \yael{could be useful to demonstrate here how a model that was trained on one dataset fails to generalize to the other.}}  
%   \label{fig:existing_dataset}
% \end{figure}

Our method builds upon Grounded SAM~\cite{ren2024grounded}, a state-of-the-art approach for open-vocabulary image segmentation, which has demonstrated remarkable capabilities in segmenting complex scenes across diverse object categories. Grounded SAM combines two models to achieve this: Grounding DINO~\cite{liu2023grounding} for object detection and Segment Anything (SAM)~\cite{kirillov2023segany} for mask generation. We analyze the performance of these models on sketch, revealing that the domain gap between real and sketched objects presents significant challenges for Grounding DINO. In contrast, SAM exhibits a surprising ability to generalize to sketches, though it still faces difficulties specific to the sketch domain.
% Image instance segmentation pipelines typically consist of two key
% stages: object detection and object mask generation. 
% In this work, we focus on two state-of-the-art models for these tasks - Grounding
% DINO~\cite{liu2023grounding} for object detection, and Segment
% Anything (SAM)~\cite{kirillov2023segany} for mask generation -- and
% analyze their domain transfer capabilities. We demonstrate that the
% domain gap between real and sketched objects poses significant
% challenges for Grounding DINO, whereas SAM exhibits a surprising
% capacity to generalize to sketches, while still struggling with
% sketch-specific challenges.
To address the gap in object detection, we fine-tune Grounding DINO on
a small subset of annotated scene sketches from the SketchyScene
dataset~\shortcite{Zou18SketchyScene}.
% , showing that fine-tuning Grounding DINO on a single sketch style generalizes effectively across diverse styles and abstraction levels. 
% By removing the dependency on predefined object categories, we aim to encourage the model to focus on structural and relational cues, drawing inspiration from Gestalt principles such as closure, continuity, and emergence.
Our fine-tuning technique achieves a substantial improvement in Grounding DINO's detection performance on sketches, with Average Precision increasing from \rev{26\%} to \rev{74\%}.

For object segmentation, we apply SAM \cite{kirillov2023segany} in the sketch domain, using detected object regions from our finetuned Grounding DINO. This is followed by a depth-based refinement stage to resolve ambiguities in overlapping regions.
Finally, we decompose the segmented sketch into sorted layers and employ a pretrained image inpainting model \shortcite{von-platen-etal-2022-diffusers} to fill in missing regions. This layered representation facilitates sketch editing, allowing users to drag or manipulate segmented objects without the need to manually sketch the affected regions, as we demonstrate in the provided video.

To evaluate our method on diverse scene sketches, we construct \datasetname{}, a synthetic annotated dataset of sketched scenes that extends existing benchmarks along three key dimensions: drawing style, stroke style, and object categories. The dataset integrates two complementary pipelines to enhance diversity.
The first pipeline builds on SketchyScene \cite{Zou18SketchyScene}, expanding its clipart-like sketches with styles ranging from high-fidelity representations to symbolic, abstract sketches, introducing challenging out-of-distribution cases. The scenes are created in vector format to allow for stroke style variations, including Calligraphic
Pen, Charcoal, and Brush Pen styles.
The second pipeline leverages the Visual Genome dataset \cite{VisualGenome2017}, which provides annotated scenes with object variety. Using the InstantStyle method \cite{Wang2024InstantStyleFL}, we generate expressive, natural-looking sketches spanning 72 categories, extending SketchyScene's original 45 categories by 53 new categories. Our dataset contains 20,542 annotated scene sketches in total, and is highly extensible.
Our evaluations demonstrate that \methodname{} generalizes well to these challenging variations, significantly advancing the state of the art. 

\begin{figure*}
  \centering
  \includegraphics[width=1\linewidth]{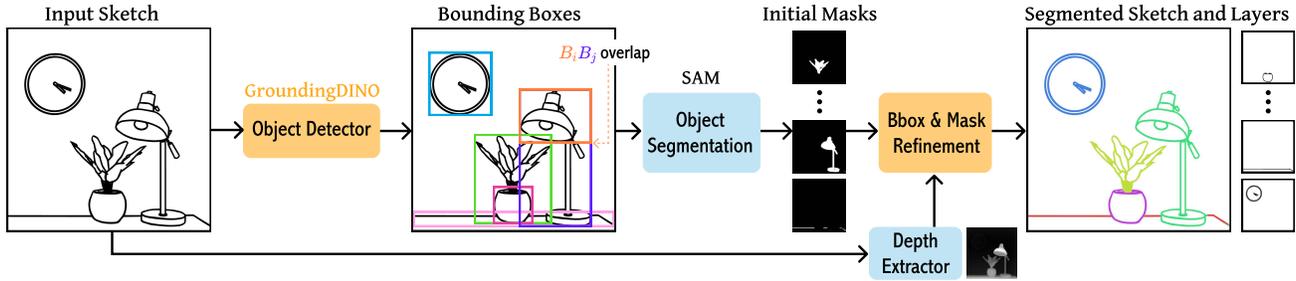}
  \vspace{-0.7cm}
  \caption{\textbf{Overview of the sketch segmentation pipeline.} Given an input sketch image, our framework first detects bounding boxes using a customized Grounding DINO to obtain region proposals, and then perform segmentation with SAM models. The localization and segmentation are refined by incorporating the depth features. The result segmentation can be viewed as a layered decomposition of object components in the original sketch.}  
  \label{fig:methods}
\end{figure*}

\section{Related Work}
\subsection{Part-Level Sketch Segmentation} 
The majority of work in the sketch segmentation domain focuses on part-level semantic segmentation, in which the goal is to assign labels to object parts (\eg, the body, wings, and head of a bird). These methods often rely on curated part-level sketch segmentation datasets
% The majority of work in the sketch segmentation domain focuses on part-level semantic segmentation. These methods label object parts (\eg, identify the body, wings, and head of a bird, \etc). They often rely on curated part-level sketch segmentation datasets
\cite{ge2020creative,li2018universalperceptualgrouping,Wu2018SketchsegnetAR,eitz2012hdhso,dataDrivenSeg2014} to train a segmentation model, and use various network architectures, including CNNs~\cite{Zhu2018PartLevelSS, wang2020multicolumnseg}, RNNs~\cite{sketchSegNet2019, Wu2018SketchsegnetAR, symbolReconSeg2020}, Graph Neural Networks~\cite{ENDE-GNN2022, sketchGNN2021}, Transformers~\cite{wang2024contextseg, segFormer2023}, and more specific techniques such as deformation networks~\cite{oneshot2021} and CRFs~\cite{CRFseg2016}. These approaches typically operate on a fixed set of object classes, and recognize a predefined set of object parts within them. Other work focuses on perceptual grouping~\cite{li2018universalperceptualgrouping, Li2019TowardDU,Li:2022:Free2CAD} to achieve class-agnostic segmentation. However these methods are designed to tackle part-level segmentation and are not suitable for scene sketches.
% \vspace{-2mm}

% which computes stroke affinity matrices in vector sketches using either hand-crafted features or implicit modeling of Gestalt principles to achieve class-agnostic grouping. 
% \vspace{-0.2cm}
\subsection{Scene-Level Sketch Segmentation}
Scene-level sketch segmentation remains largely under-explored. Qi \etal~\shortcite{edgePerceptual2015} extend the perceptual grouping approach to scene-level images, forming semantically meaningful groupings of edges, though with limited accuracy on complicated scenes.
Zou \etal\shortcite{Zou18SketchyScene} construct the SketchyScene dataset, providing annotated scene sketches with meaningful layouts of object interactions, and use it to train an instance segmentation model based on the Mask R-CNN architecture~\cite{he2018maskrcnn}. However, their method is limited to the predefined categories included in the dataset, and the proposed dataset contains sketches with clipart-like appearance which challenges the model's ability to generalize to other artistic styles. 
Building on SketchyScene, Ge \etal \shortcite{GeLocalDetailPerception} introduced SKY-Scene and TUB-Scene by replacing its object components with sketches from the Sketchy \shortcite{sketchydataset} and TU-Berlin \shortcite{eitz2012hdhso} datasets. However, their proposed fusion network is fundamentally limited to the fixed set of classes it was trained on, and the trained network weights are not publicly available.
% , as its architecture is based on DeepLabV2, a closed-vocabulary segmentation model, which restricts its ability to generalize to unseen categories.
SFSD \cite{zhang2023strokeSeg} develops a dataset featuring more complex scene sketches, and utilizes a bidirectional LSTM to produce stroke-level segmentation. Unfortunately, the dataset and model are not publicly available.
SketchSeger \cite{yang2023sceneHierTransformer} proposes a hierarchical Transformer-based model for semantic sketch segmentation. However their model is inherently restricted to the predefined set of classes used during training. Bourouis \etal \shortcite{bourouis2024open} finetune the CLIP image encoder \cite{radford2021clip} on the FS-COCO \cite{fscoco} dataset, leveraging the model's vision-language prior to enable open-vocabulary scene segmentation. However their method is designed for semantic segmentation, and it struggles to generalize to more challenging sketch styles and scene layouts.
% \yael{decide if we keep:}Kutuk et al's method ~\shortcite{Kutuk2024ClassAgnosticVS} targets vector scene sketches, leveraging a class-agnostic object detector with temporal stroke information and a pre-trained sketch classifier. They demonstrate the ability to distinguish object instances; however, their visual examples are confined to scenes with minimal or no complex overlaps between object instances, which fail to closely resemble real-world sketches. 
% The most recent scene sketch dataset, FrISS, introduced by Kutuk \etal \cite{Kutuk2024ClassAgnosticVS}, was curated by recruiting participants to draw quick sketches based on text descriptions from the MS COCO dataset. While this dataset is not publicly available, the examples shown in their paper reveal predominantly simple sketches with minimal detail.

% Other scene sketches datasets exists
% \yael{decide if to keep:}

% \yael{this one we keep and use the data for comparison, but the paper is not about segmentation}
% CBSC \cite{zhang2018CBSC}, the first scene sketch dataset to include indoor scenes, is a small-scale dataset of 332 human-drawn sketches characterized by quick, freehand designs.
% While these extensions emphasize simplified, novice-style object representations, they only expand the style variation in a single direction—towards symbolic sketches.
% \vspace{-2mm}
\subsection{Image Segmentation}
The task of image segmentation have been widely explored~\cite{he2018maskrcnn,  Bolya_2019_yolact, cheng2021mask2former, wang2021solo}. 
The advent of vision-language models \cite{radford2021clip, liu2023llava,xiao2023florence} has led to numerous object detection and segmentation methods with
impressive generalization capabilities \cite{zhang2022glipv2, ren2024grounded, kirillov2023segany, minderer2022owlvit}.
% Grounded SAM \cite{ren2024grounded} is among the leading approaches in this domain. It combines two state-of-the-art models, Grounding DINO \cite{liu2023grounding} and Segment Anything (SAM) \cite{kirillov2023segany}, for open-vocabulary image segmentation, achieving robust performance across diverse object categories. 
Grounding DINO \cite{liu2023grounding} is a state-of-the-art object detection model trained on over 10 million images. It builds on top of DINO \cite{DINOcaron2021emerging}, a strong vision encoder, with effective grounding module that fuses visual and textual information, enabling open-vocabulary detection of unseen objects. Segment Anything (SAM) \cite{kirillov2023segany} is an image segmentation model trained on over 11 million images and 1.1 billion masks, capable of producing high-quality object masks based on various forms of conditioning such as bounding boxes. Grounded SAM \cite{ren2024grounded}, which our method builds upon, combines Grounding DINO and SAM for open-vocabulary image segmentation, achieving robust performance across diverse object categories. 
Yet, despite demonstrating impressive capabilities on natural images, we show that these models struggle with segmenting sketches.  

% is among the leading approaches in this domain.

\section{Method}
Given a raster sketch, our goal is to produce a segmentation map such that pixels belonging to the same object instance are grouped together. Based on the segmentation map, we also divide the sketch into layers, sorted by depth. Our pipeline is illustrated in \Cref{fig:methods}. Given the input sketch, we first perform object detection using a fine-tuned Grounding DINO model, which produces a set of candidate object bounding boxes. These bounding boxes are then used to produce an initial set of object masks with a pre-trained Segment Anything (SAM) \cite{kirillov2023segany} model. Next, we perform a refinement stage that leverages scene depth information to assign the final segmentation. This stage also employs a pre-trained inpainting model \cite{von-platen-etal-2022-diffusers} to produce scene layers. 
%We next describe each stage in detail.

% \begin{figure}[b]
%     \centering
%     \includegraphics[trim=31cm 0cm 60cm 0cm,clip,width=0.3\linewidth]{figs/placeholder_GDINO.png}
%     \includegraphics[trim=90cm 0cm 0cm 0cm,clip,width=0.3\linewidth]{figs/placeholder_GDINO.png}
%     \includegraphics[width=0.3\linewidth]{figs/placeholder2.png}
%     \caption{Grounding DINO performance. Left: the model produces plausible results for natural images, even for such a challenging input. Right: Yet even for a less complex scene sketch \yael{mia please replace this sketch}, the model struggles with operating on sketches.\maneesh{I'd make the figs bigger so the entire column width is used.} }
%     \label{fig:groundDINO}
% \end{figure}

\subsection{Sketch-Aware Object Detection}
Grounding DINO \cite{liu2023grounding} is an object detection model which outputs bounding boxes for recognized object instances, based on a given text prompt describing the scene. While effective for natural images, the model in its original configuration demonstrates limited generalization to sketches (we show this numerically in \Cref{sec:results}). 
To address this limitation, we fine-tune Grounding DINO on sketches.  
% While standard fine-tuning approaches typically require large-scale datasets of segmented and labeled scene images, such datasets are particularly scarce in the sketch domain. 
The largest available annotated sketch dataset containing complex scenes is SketchyScene~\cite{Zou18SketchyScene}. It contains 30K segmented sketches across 45 class labels.
We find that a naive fine-tuning with the SketchyScene data leads to severe overfitting to the small set of predefined object classes.

To overcome this overfitting, we propose a class-agnostic fine-tuning strategy. Instead of relying on predefined class labels, we train the model to distinguish between instances based on their visual characteristics, aiming to push the model to rely on Gestalt properties such as closure, continuity, and emergence, to group together strokes forming a single object. Specifically, we utilize a small subset of 5000 sketches from the SketchyScene dataset, and consolidate their class labels into a single label, ``object''.  We use a Grounding DINO model initialized with a pretrained Swin Transformer~\cite{liu2021Swin} backbone and fine-tune the model's detection head for bounding box prediction. For training, we employ standard object detection losses used in the original Grounding DINO training (Focal Loss, L1 Loss, and GIoU Loss), while eliminating the class recognition loss. At inference, the model is prompted with the input image and the word ``object'' to detect all potential object instances in the scene. This results in an initial set of $k$ bounding boxes $B=\{B_i\}_{i=1}^k$ and a confidence score per bounding box. 
% We later filter this initial set using an enhanced Non-Maximum Suppression (NMS) approach (Section \maneesh{put fwd pointer to section.}).\yael{maybe we can remove "enhanced"? I dont want us to overclaim about simple/trivial components}

\vspace{-2mm}
\subsection{Mask Extraction and Bounding Boxes Refinement}
Once the bounding boxes are obtained, we use a pretrained Segment Anything (SAM) model \cite{kirillov2023segany} to extract masks for the corresponding objects directly from the sketch. This results in an initial set of $k$ masks $M=\{M_i\}_{i=1}^k$ which we refine using simple binary operations such as morphological closing and flood-fill, to eliminate small artifacts.

Next, we use the refined masks to enhance the set of generated bounding boxes. A common practice is to eliminate redundant bounding boxes often corresponding to the same object (such as $B_i, B_j$ shown in red and blue in \Cref{fig:methods}) using Non-Maximum Suppression (NMS), which filters out bounding boxes with low confidence scores that has significant intersection with others. However, IoU of the bounding boxes may not reliably reflect object overlap in cases where objects do not fully cover the pixels in their bounding boxes. This issue is especially pronounced in sketches, which are sparser than photorealistic images.
We use the initial set of masks to compute a more fine-grained IoU. Specifically, for a pair of overlapping bounding boxes $B_i$ and $B_j$, we extract the regions within the bounding boxes that intersect with the \rev{sketch $S$}: $M_i * S, M_j * S$, and compute the IoU of these regions to define an ``overlapping'' score between two objects $i, j$:
\vspace{-1pt} 
\begin{equation} 
\mathcal{O}(i, j) = \text{IoU}(M_i * S, M_j * S).
\end{equation}
% we calculate the IoU based on the regions within the bounding boxes that are part of the sketch: $M_i * S, M_j * S$. pixels in the sketch $S$, using the masks to define their respective regions:

% \begin{equation} 
% \text{IoU}_S(i, j) = \text{IoU}(M_i * S, M_j * S),
% \end{equation}

% Where $M_i$ and $M_j$ are the masks corresponding to the bounding boxes $B_i$ and $B_j$, and $*$ is the element-wise multiplication.

\vspace{-1pt} 
For an overlapping pair $B_i, B_j$ if $\mathcal{O}(i, j) > 0.5$, we consider the detections to be covering the same object and retain only the bounding box with the highest confidence score.
This results in a filtered set of bounding boxes $\hat{B}\subseteq B$ and their corresponding masks $\hat{M} \subseteq M$.

% \begin{figure}
%     \centering
%     \includegraphics[width=0.9\linewidth]{figs/placeholder_sketch_NMS.png}
%     \caption{Motivation for sketch NMS. The bonding's boxes have a low IoU value of 0.4, but their sketch content IoU value is as high as 0.8. \maneesh{Hard to see the issue at this size.}}
%     \label{fig:sketchNMS}
% \end{figure}

\begin{figure}
    \centering
    \includegraphics[width=1\linewidth]{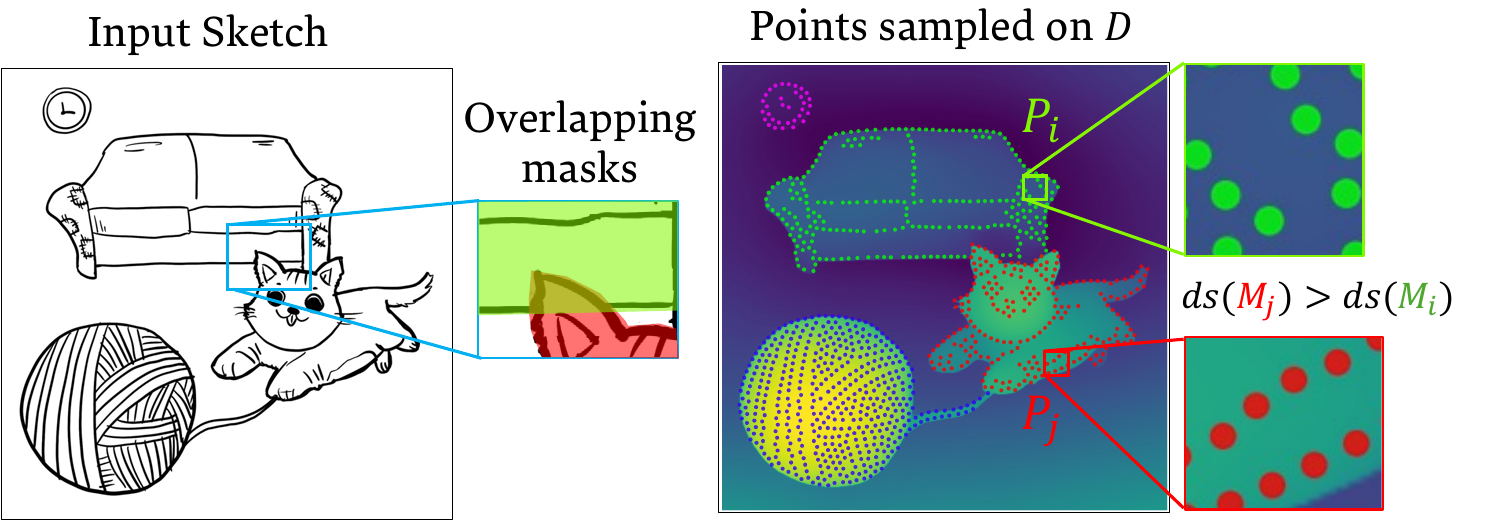}
    \vspace{-6mm} 
    \caption{\textbf{Resolving ambiguities in overlapping regions.} The depth map $D$ is sampled along sketch pixels at evenly spaced points, and the sampled points are grouped by their corresponding object (e.g., $P_i$ corresponds to the $i$'th object). Each object is assigned a depth score based on the majority of depth values from the sampled points. Ambiguous pixels are then assigned to the mask with the highest depth score, prioritizing foreground objects.}
    \label{fig:mask-sort}
    \vspace{-2mm}
\end{figure}

% The set of filtered masks requires further refinement as it contains overlapping regions, which lead to ambiguity in pixel assignments. \maneesh{What are the "filtered masks". We just did NMS on the bounding boxes, so I've lost track of what the filtered masks are.} \yael{each bbox has a corresponding mask, so filtering bbox is also filtering masks, I wrote it above "filtered bbox and their corresponding masks"}
% To resolve such ambiguous pixels, we post-process the masks into disjoint regions by assigning each pixel to the mask corresponding to the object in the foreground. 
The filtered set of masks may still include overlapping regions, as illustrated in \cref{fig:mask-sort}, where it is unclear which instance the pixels should be associated with. To resolve such overlaps and assign each pixel to a single object instance, we give priority to objects in the foreground.
We utilize DepthAnythingV2 \cite{depth_anything_v2} to extract the depth map $D$ of the input sketch. We then sample $D$ along the sketch pixels at equally spaced points $P=\{p_1,p_2,\dots ,p_n\}$, analogous to projecting rays through the sketch pixels to the scene. For each mask $M_i$, we identify the subset of points that lie within the mask: $P_i = \{p \in P | p\in M_i\}$. For example, in \Cref{fig:mask-sort}, $P_i$ represents the set of points belonging to the sofa, while $P_j$ denotes the set of points belonging to the cat. For each point $p \in P_i$ we associate a depth value $D(p)$ using the depth map. We then compute a depth score for each mask as the mode of the depth values associated with its sampled points:
\vspace{-3pt} 
\begin{equation}
    ds(M_i) = \arg \max_{D(p)} \text{count}(\{D(p)| p \in P_i\}).
\end{equation}

\vspace{-2pt} 
Based on this score, we assign ambiguous pixels to the mask with the highest depth score, ensuring that foreground objects take precedence. 
% \begin{equation}
%     label(x) = \arg \max_{i} (\{score(M_i)|x\in M_i\})
% \end{equation}
% \maneesh{How is the score used to assign a instance label to the pixel? I think a final step might be missing. Also is this depth priority score equal to the layer depth? If so we should say that. If not we should explain how we compute layer depth.} \yael{i clarified the process in the text and figure}
Lastly, to ensure complete coverage of the sketch, we employ a watershed-based~\cite{watershed1991} refinement, propagating existing mask labels to  unlabeled sketch pixels.

\subsection{Layer Inpainting}
\begin{figure}
    \centering
    \includegraphics[width=1\linewidth]{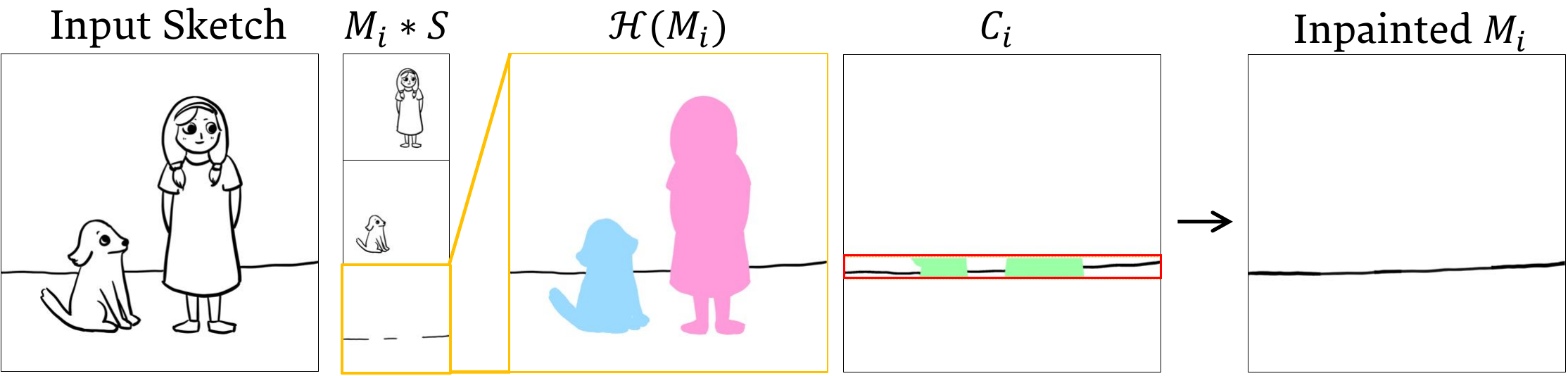}
    \vspace{-6mm} 
    \caption{\textbf{Layer completion. }Object layers are isolated and inpainted using a pretrained SDXL model. The inpainting mask for each object is defined by intersecting overlapping masks with the object's bounding box.}
    \vspace{-2mm}
    \label{fig:layers-inpaint}
\end{figure}

As a final step we extract complete layers for each object in the sketch, inpainting any occluded regions using  a pretrained SDXL inpainting model~\cite{von-platen-etal-2022-diffusers}. The goal of this stage is to support basic sketch editing operations, such as translation and scaling.
We isolate each object $i$ by intersecting the sketch with its corresponding mask $M_i * S$ (\Cref{fig:layers-inpaint}). 
We then identify the group of masks that intersect with $M_i$: $\mathcal{H}(M_i)=\{M_j | M_j \cap M_i \neq \varnothing \}$. Finally, we define the inpainting mask $C_i$ as the intersection of $\mathcal{H}(M_i)$ with the object's bounding box: $C_i = \mathcal{H}(M_i) \cap B_i$ (shown in green in \cref{fig:layers-inpaint}), and feed it into the pretrained inpainting model.
% \maneesh{Not sure I follow this. What is Ci, and what is I(Mi)?}

\subsection{Implementation Details}
Optimization is performed with the AdamW optimizer, configured with an initial learning rate of 6e-5 and a weight decay of 0.0005 to promote generalization and prevent overfitting. Training is conducted with a batch size of 4 and automatic learning rate scaling to ensure stable updates and efficient adaptation.
For the train, validation, and test sets, we sampled 5,000, 500, and 500 images, respectively, from the original SketchyScene train, val, and test splits. The experiment was conducted on a single NVIDIA 4090 GPU, with a total training time of four hours.

\begin{figure}[t]
    \centering
    \setlength{\tabcolsep}{2pt}
    % \vspace{-3mm}
    {\small
    \begin{tabular}{c c c}
        Sketch & Bounding Box & Segmentation \\
        \frame{\includegraphics[width=0.32\linewidth]{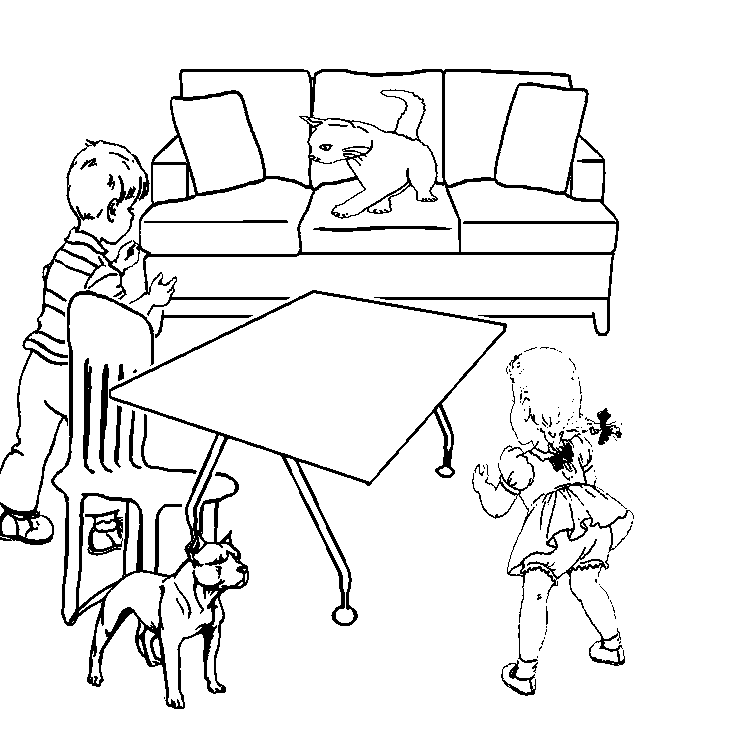}} &
        \frame{\includegraphics[width=0.32\linewidth]{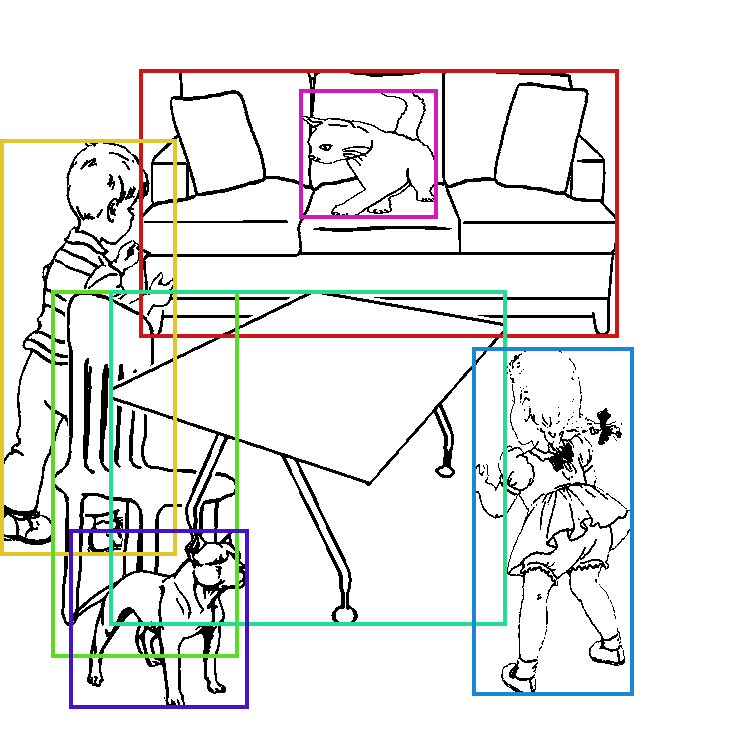}} &
        \frame{\includegraphics[width=0.32\linewidth]{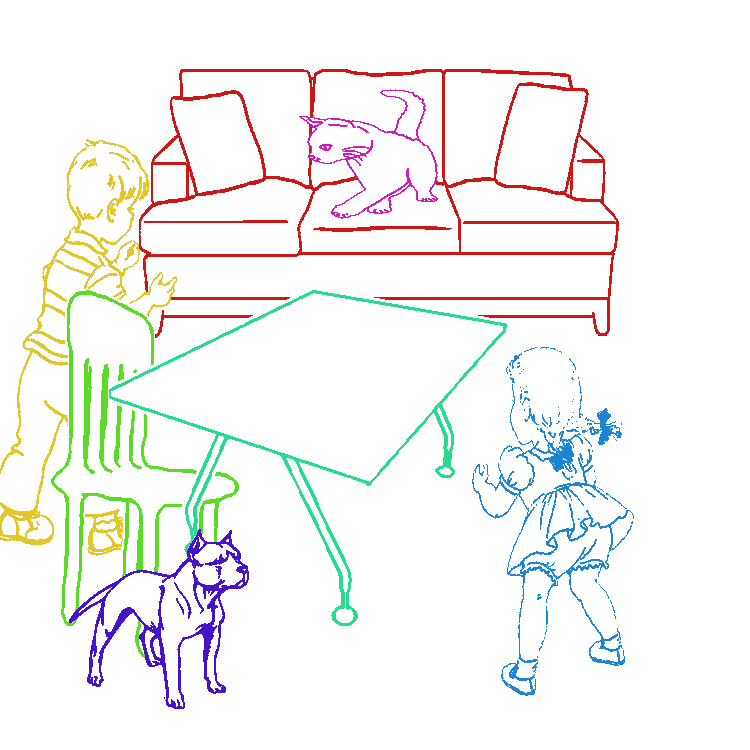}} \\
    \end{tabular}
    }
    \vspace{-0.4cm}   
     \caption{SketchyScene dataset provides ground truth object bounding boxes and pixel-level instance segmentation masks for scene layouts.}
    \vspace{-3mm}   
    \label{fig:SketchySceneLayout}
\end{figure}

\begin{figure*}
    \centering
    \setlength{\tabcolsep}{2pt}
    {
    \begin{tabular}{c @{\hspace{0.4cm}} c c c @{\hspace{0.4cm}} c}
    \toprule
        \multirow{2}{3cm}{SketchyScene Layout} & \multicolumn{3}{c}{\xrfill[0.5ex]{0.5pt}\; \textbf{Stroke Style Variation} \; \xrfill[0.5ex]{0.5pt}} & \multirow{2}{3cm}{\centering SketchAgent} \\
         & \textit{Calligraphic Pen} & \textit{Charcoal} & \textit{Brush Pen} & \\
        \midrule
        \frame{
        \includegraphics[width=0.174\linewidth]{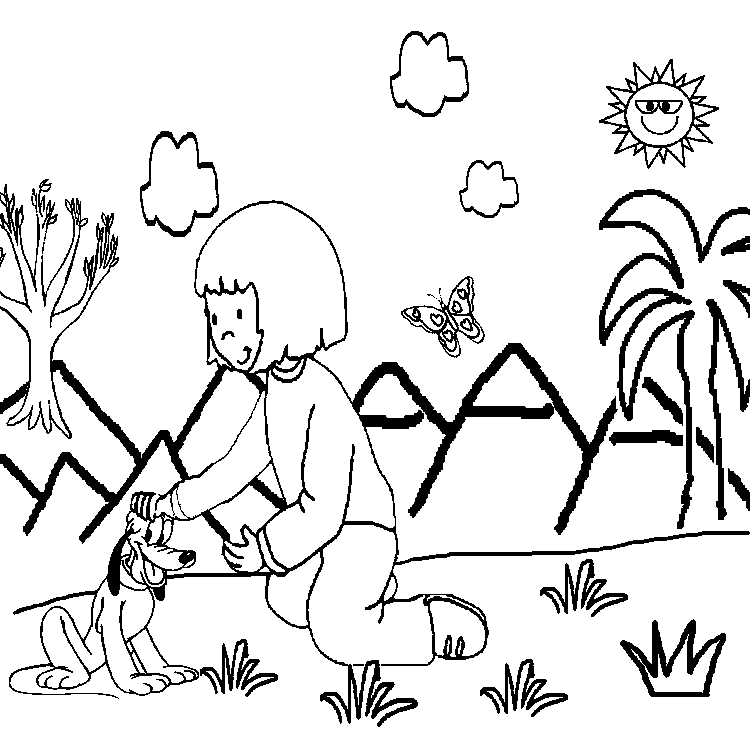}}&
        \frame{\includegraphics[width=0.174\linewidth]{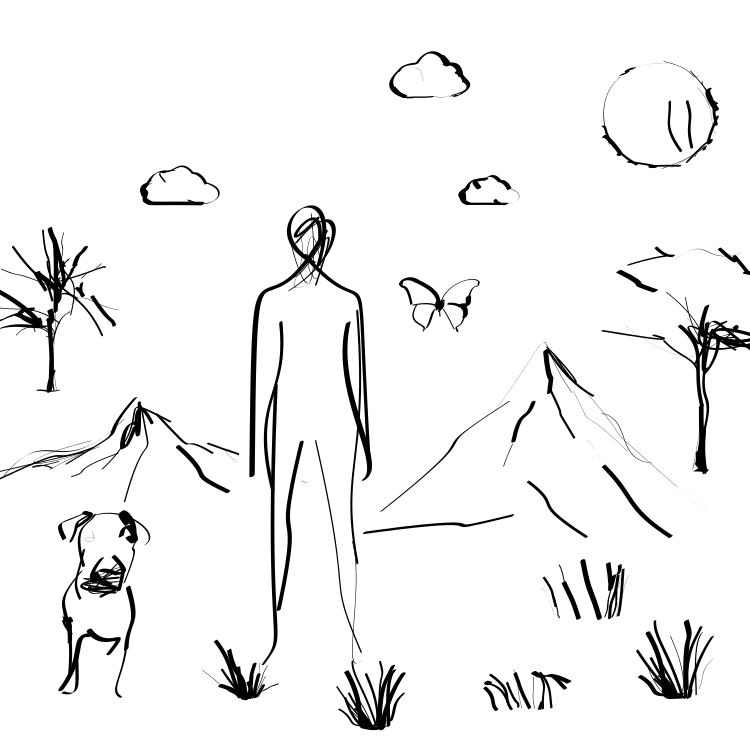}} &
        \frame{\includegraphics[width=0.174\linewidth]{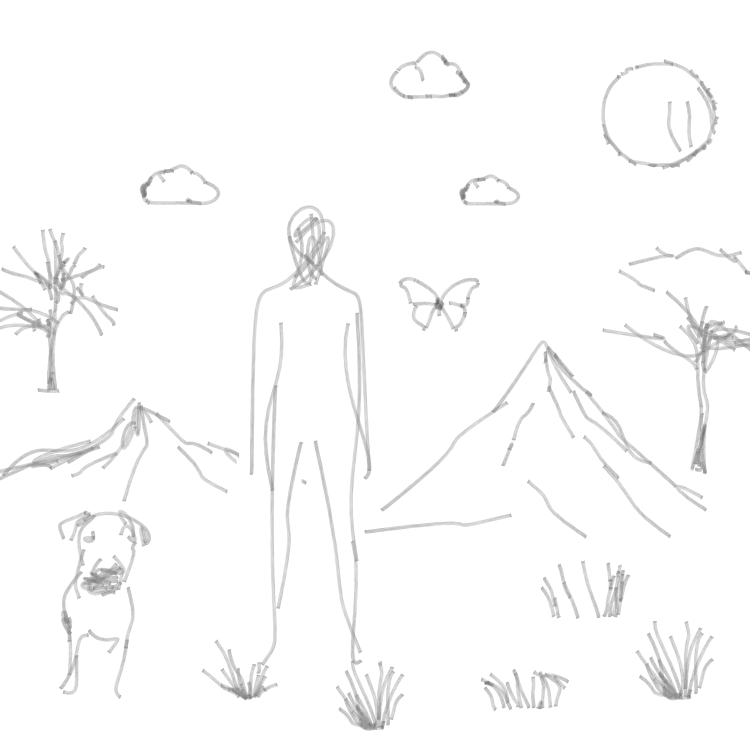}} &
        \frame{\includegraphics[width=0.174\linewidth]{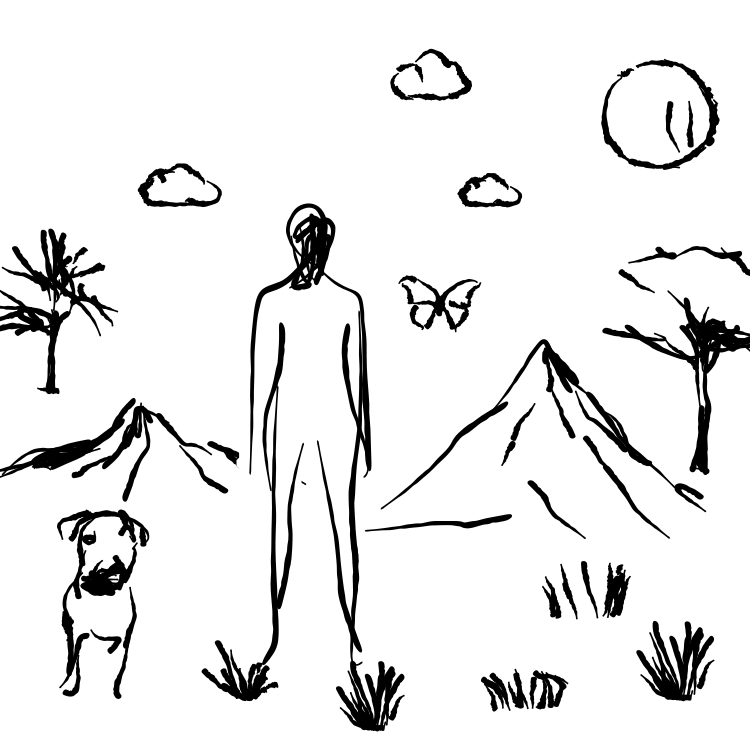}} &
        \frame{\includegraphics[width=0.174\linewidth]{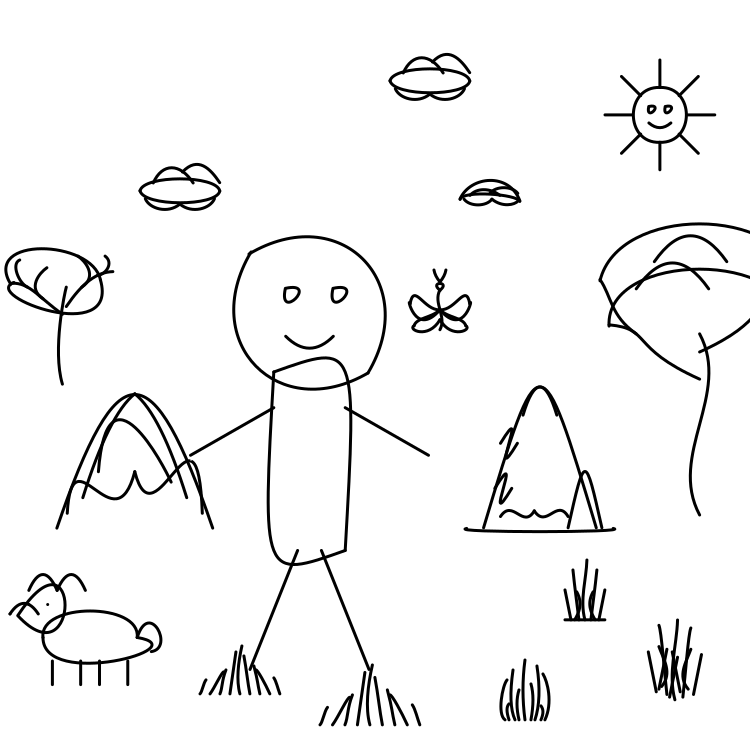}} \\

        & 
        \includegraphics[width=0.1\linewidth]{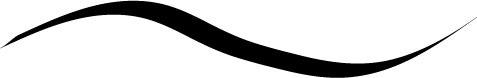} &
        \includegraphics[width=0.1\linewidth]{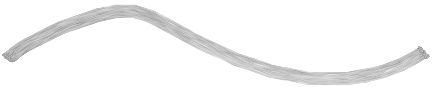} &
        \includegraphics[width=0.1\linewidth]{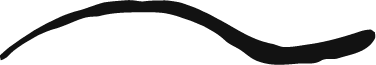} &
        \\
       
    \end{tabular}

     % \begin{tabular}{c c c c c}
     % % \midrule
     % \multicolumn{5}{c}{\xrfill[0.5ex]{0.5pt}\; Drawing Style Variation \; \xrfill[0.5ex]{0.5pt}}\\
     % \midrule
     %    \frame{\includegraphics[height=2.1cm, width=0.172\linewidth]{figs/dataset_examples/6225.png}} &
     %    \frame{\includegraphics[height=2.1cm, width=0.172\linewidth]{figs/dataset_examples/101686.png}} &
     %    \frame{\includegraphics[height=2.1cm, width=0.172\linewidth]{figs/dataset_examples/118330.png}} &
     %    \frame{\includegraphics[height=2.1cm, width=0.172\linewidth]{figs/dataset_examples/182362.png}} &
     %    \frame{\includegraphics[height=2.1cm, width=0.172\linewidth]{figs/dataset_examples/467564.png}} \\
        % \bottomrule
    % \end{tabular}
    }
    \vspace{-0.4cm}
    \caption{\textbf{Samples from our \datasetname{} dataset.} We augment the SketchyScene dataset by generating vector sketches with varied drawing styles based on SketchyScene's scene layouts. Stroke style variation is introduced by re-rendering the scenes with three different brush styles. Additionally, we create a more symbolic and challenging sketch type, resembling children's drawings, shown on the right, while maintaining the same scene layouts.}
    \label{fig:dataset}
\end{figure*}

\begin{figure}
    \centering
    \setlength{\tabcolsep}{2pt}
    \vspace{-4pt}
    {\small
    \begin{tabular}{c c c}
        Input Image & InstantStyle Sketch & Instance Segmentation \\
        \frame{\includegraphics[width=0.32\linewidth]{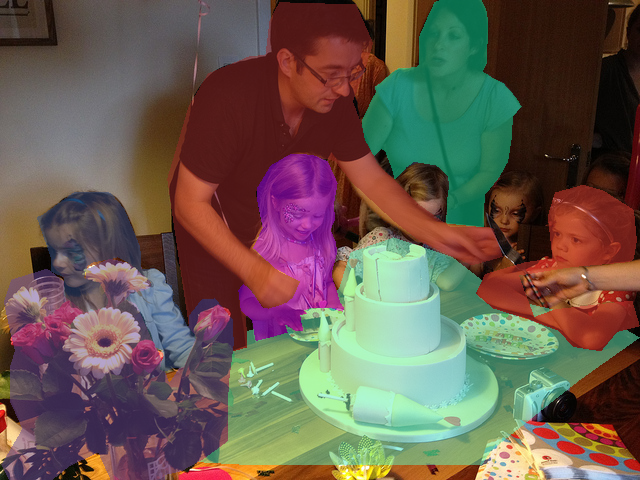}} &
        \frame{\includegraphics[width=0.32\linewidth]{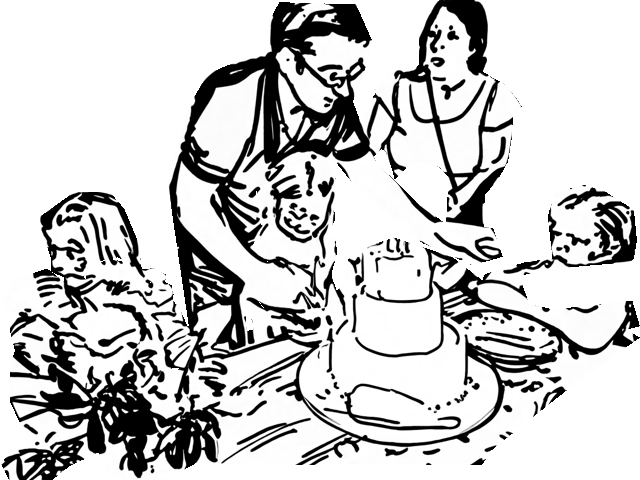}} &
        \frame{\includegraphics[width=0.32\linewidth]{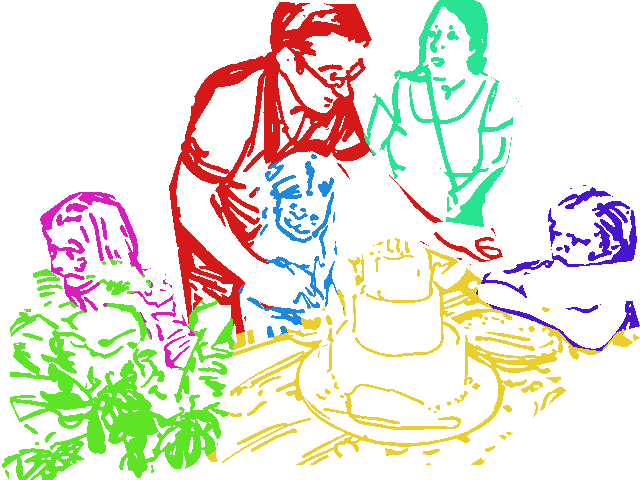}}
        \\

        \frame{\includegraphics[trim=0 1.3cm 0 0,clip,width=0.32\linewidth]{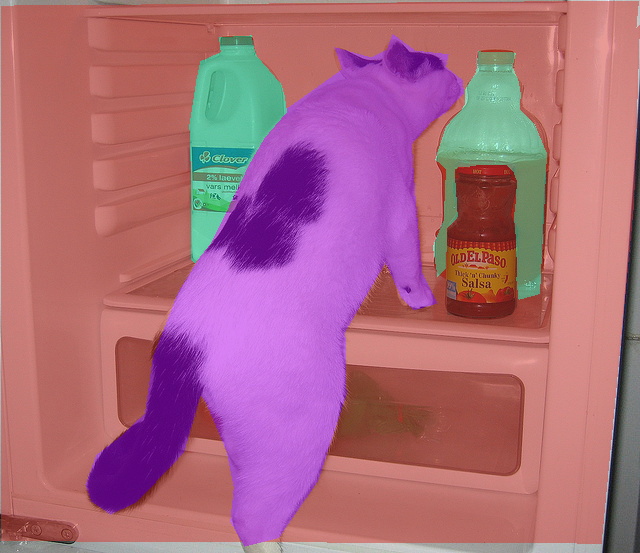}}  &
        \frame{\includegraphics[trim=0 1.3cm 0 0,clip,width=0.32\linewidth]{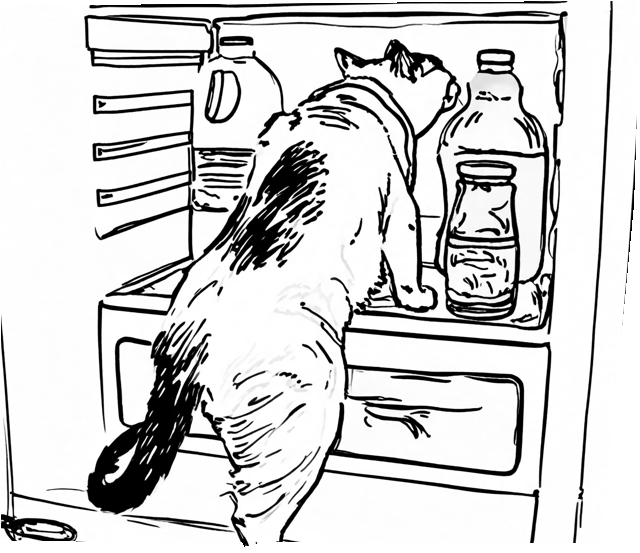}} &
        \frame{\includegraphics[trim=0 1.3cm 0 0,clip,width=0.32\linewidth]{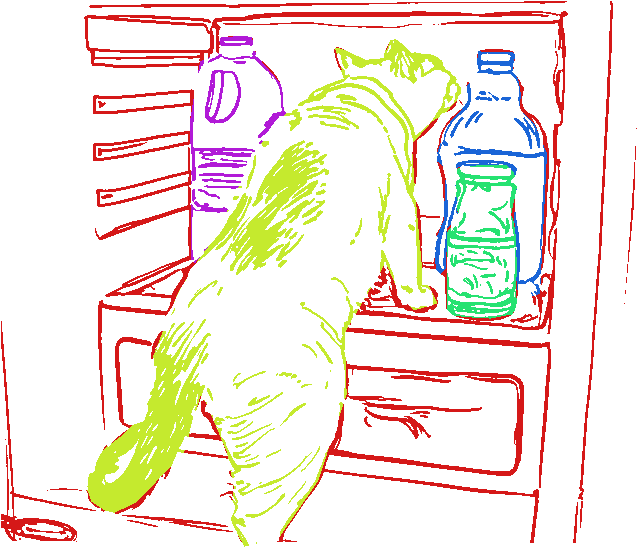}}
        
        \\

    \end{tabular}
    }
    \vspace{-0.3cm}    
    \caption{\textbf{Illustration of our \datasetname{} dataset.} The input images are sourced from the Visual Genome dataset \cite{VisualGenome2017}, which we filter to a subset of scenes containing 5 to 10 object instances. We generate the corresponding sketches with InstantStyle \cite{Wang2024InstantStyleFL}.}
    \vspace{-10pt}
    \label{fig:instantstyle}
\end{figure}

\section{Scene Sketch Segmentation Benchmark}
To evaluate our performance across a diverse set of sketches, we construct a synthetic annotated scene-sketch dataset: \datasetname{}. This dataset focuses on three key axes of variation, designed to extend existing datasets: (1) drawing style, (2) stroke style, and (3) object categories. 
% \maneesh{Not sure what drawing style is vs. stroke style. Should probably define below. Also we use different terms (sketch style and stroke variations) below which is confusing.} \mia{WIP} 
We define drawing style as a spectrum ranging from symbolic, which emphasizes abstraction and simplified representation, to realistic, which prioritizes detailed and lifelike depiction. We define stroke variation as the differences in texture, width, and flow that characterize individual strokes, similar to the variety of brush types in digital drawing software.
% By independently varying each axis, the dataset enables the isolation and analysis of how these factors influence segmentation results.
Our dataset combines two complementary pipelines to enhance diversity and object variety. 

\paragraph{SketchyScene Layouts}
% The first pipeline builds on the SketchyScene \cite{Zou18SketchyScene} dataset. 
The SketchyScene dataset \cite{Zou18SketchyScene} consists of 7,265 scene layouts containing 45 object categories. These layouts are of high quality, as they were manually constructed by humans. Each data sample includes an input sketch, object class labels, bounding boxes, and a pixel-wise segmentation map (see \cref{fig:SketchySceneLayout}).
The sketches in the dataset share a consistent clipart-like style. We extend the SketchyScene dataset to include more diverse sketch styles and stroke variations. Specifically, we incorporate recent object sketching methods that introduce significantly different sketch appearances compared to SketchyScene. These include CLIPasso \cite{vinker2022clipasso}, which transforms images of individual objects into sketches with relatively high image fidelity, and SketchAgent \cite{vinker2024sketchagent}, which generates symbolic sketches resembling children’s drawings, offering a more challenging out-of-distribution case.
Both techniques produce vector-format sketches, which we use to assemble scene sketches while avoiding artifacts caused by transformations. Each object is placed at its ground truth location and scaled to fit its bounding box while preserving its aspect ratio. \Cref{fig:dataset} demonstrate sketches produced from a given SketchyScene layout. 
% The ground truth pixel-wise segmentation maps for the new samples are defined by rasterizing the per-object sketches and stacking them according to the ordered object scene data from SketchyScene.
% We utilize the vector sketches to introduce stroke style variations, reflecting different tools that might be used in sketching like ink, marker, etc. 
% Using vector formats makes it both efficient and straightforward to introduce brush style variations, as it simply involves applying a style to the parametric representation and re-rendering the sketch.
% For the CLIPasso-style sketches, we generate three distinct stroke styles: pen, pencil, and ink using the \yael{X METHOD??}\mia{Put in chuan's writing}. 
Exploring stroke variation is crucial for testing the robustness of automatic segmentation approaches, as real-world scenarios often involve highly diverse sketch styles. 
We augment the vector sketch using three distinct brush styles through the Adobe Illustrator Scripting API - Calligraphic Pen, Charcoal, and Brush Pen. For each brush type, we manually select the stroke width that best preserved a natural and visually appealing result.
% \Cref{fig:dataset} shows selected samples from our dataset, with the original SketchyScene's input on the left and our generated sketches on the right. 

\paragraph{Extended Categories}
To extend the range of 45 object categories available in SketchyScene, we utilize the Visual Genome dataset \cite{VisualGenome2017}, a large-scale dataset containing diverse and richly annotated images containing over 33,877 distinct object categories. 
% \yael{Mia, please describe this dataset further and highlight its advantages compared to others.}\mia{done}
% The Visual Genome dataset includes many complex scenes with numerous objects (sometimes more than 40 objects per scene). The objects come from 33,877 distinct categories, offering an exceptionally wide range of diversity. 
% Compared to other natural image datasets, the Visual Genome dataset stands out for featuring images that encapsulate complex and meaningful relationships between numerous objects, as it is specifically designed to model interactions within objects in an image, making it an ideal source for generating meaningful scene sketches. 
% However, 
We use InstantStyle \cite{Wang2024InstantStyleFL}, a state-of-the-art style transfer method, to generate corresponding raster sketches from the input scene images, and segment the sketch objects based on the provided image segmentation. As sketches are typically sparse, and very small objects may disappear during the translation from image to sketch, we filtered the dataset to include 1068 images containing five to ten distinct objects per scene. Our InstantStyle subset of the \datasetname{} dataset includes a total of 72 categories, featuring 53 novel object classes not present in SketchyScene. A few examples of the resulting dataset are shown in \Cref{fig:instantstyle}.

% \begin{figure}
%   \centering
%   \includegraphics[width=1\linewidth]{figs/compare_real_sketch_seg_result.png} 
%   \caption{\yael{I think we can move this to the method with more analysis}Segmentation results for a natural image and an artist-drawn sketch of the same scene, both prompted with the label ``sheep''. The left column of each pair shows the input, while the right column displays the corresponding segmentation output produced by Grounded SAM. \mia{find an example where original groundingDINO (1) has good bbox on real image and (2) has bad bbox on sketch, which will lead to good vs bad segmentation results on real vs sketch, then this figure will make more sense}}  
%   \label{fig:compare_natural_vs_sketch}
% \end{figure}

% LocalWords:  pre XYZ GroundingDINO liu objectness kirillov segany
% LocalWords:  MagicFixup photorealistic Yael's inpainting pretrained
% LocalWords:  SketchyScene overfitting Swin GIoU DINO's NMS IoU SDXL
% LocalWords:  detections DepthAnything Chuan disocclude inpainted
% LocalWords:  cartoonish dont CLIPasso AdamW

\section{Results}
\label{sec:results}
\begin{figure}
    \centering
    \setlength{\tabcolsep}{2pt}
    
    {\small
    \begin{tabular}{c c c}
        \frame{\includegraphics[width=0.3\linewidth]{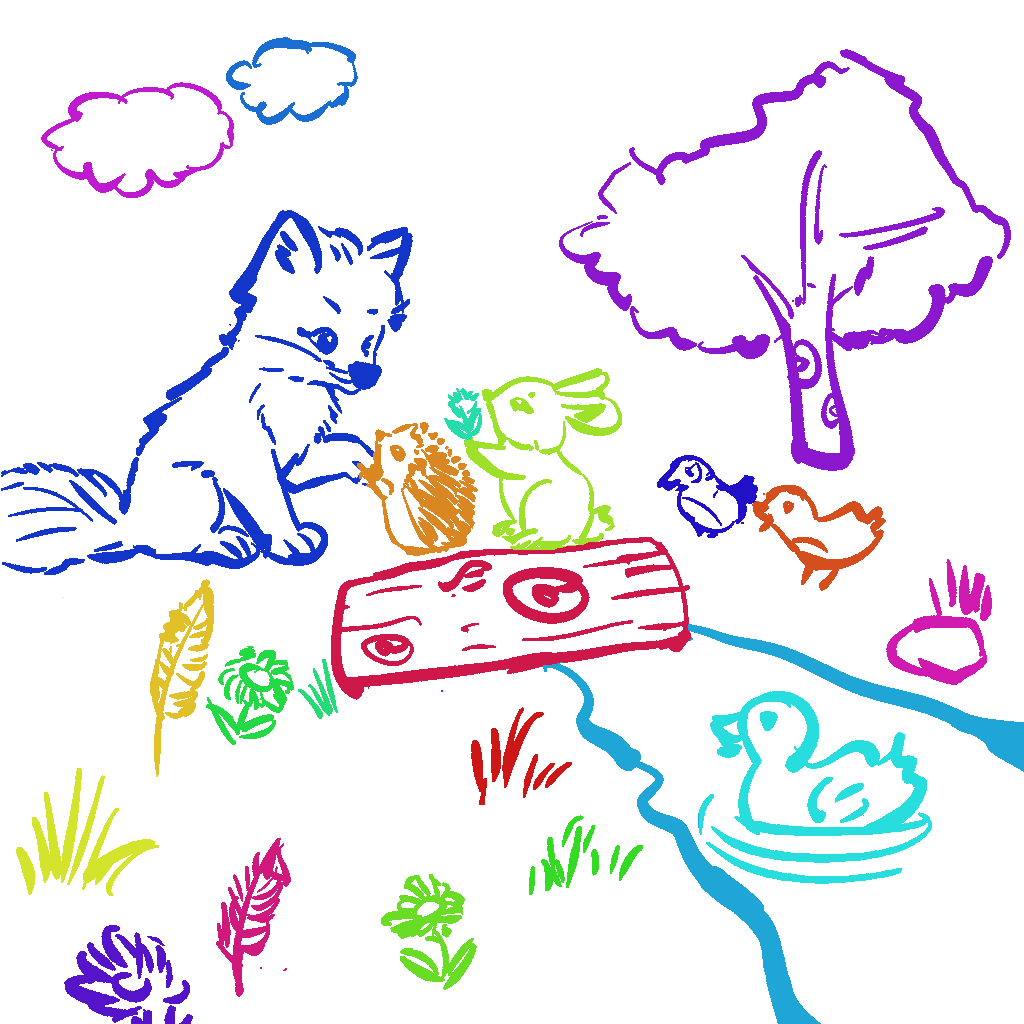}} &
        \frame{\includegraphics[width=0.3\linewidth]{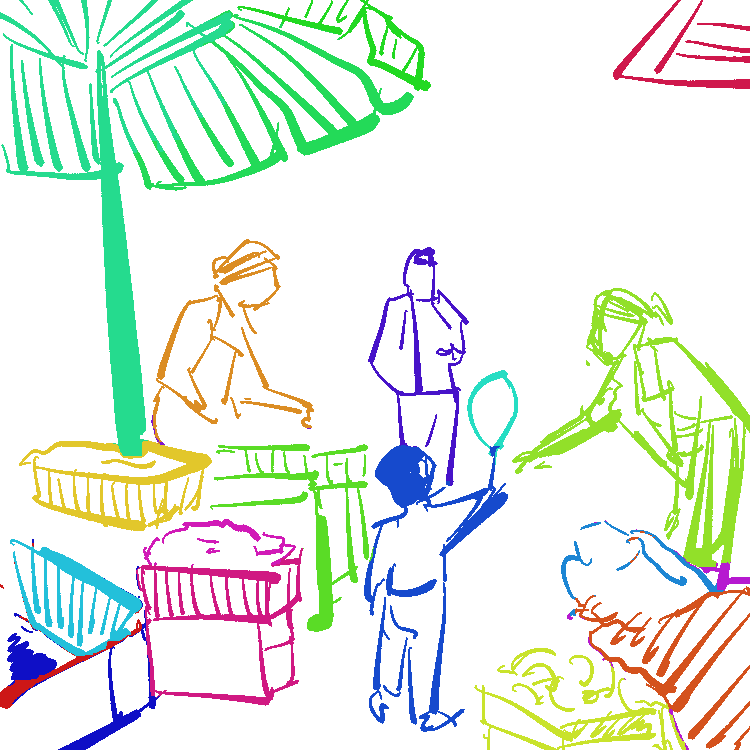}} &
        \frame{\includegraphics[width=0.3\linewidth]{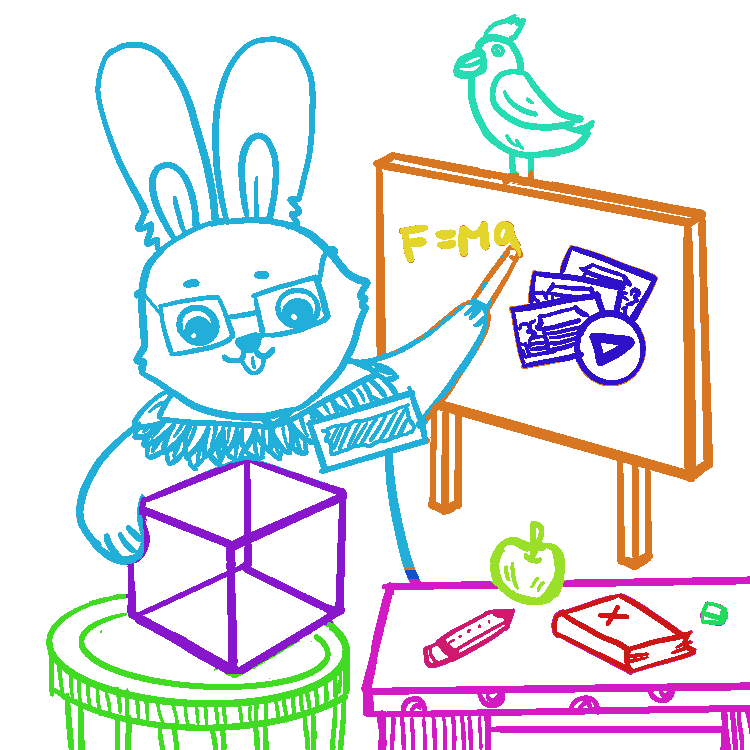}} \\
        
        \frame{\includegraphics[width=0.3\linewidth]{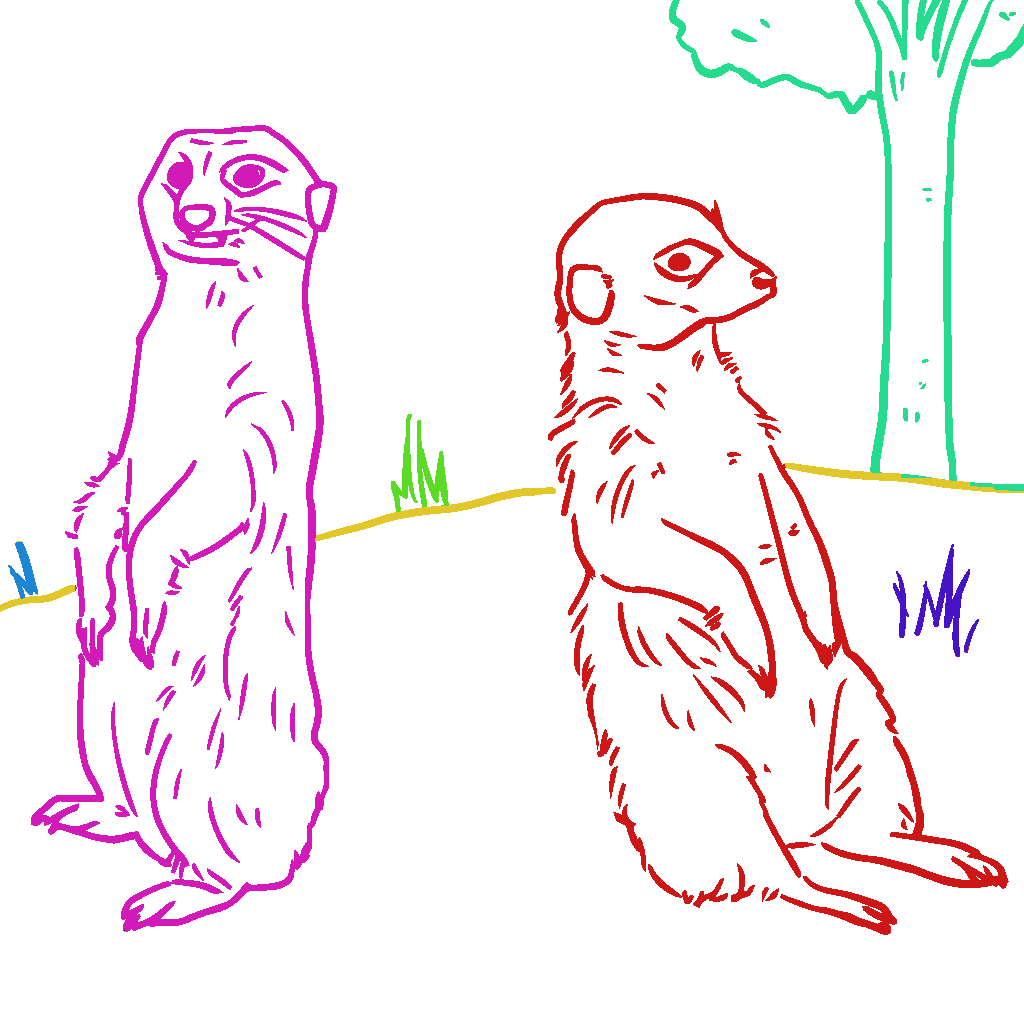}} &
        \frame{\includegraphics[width=0.3\linewidth]{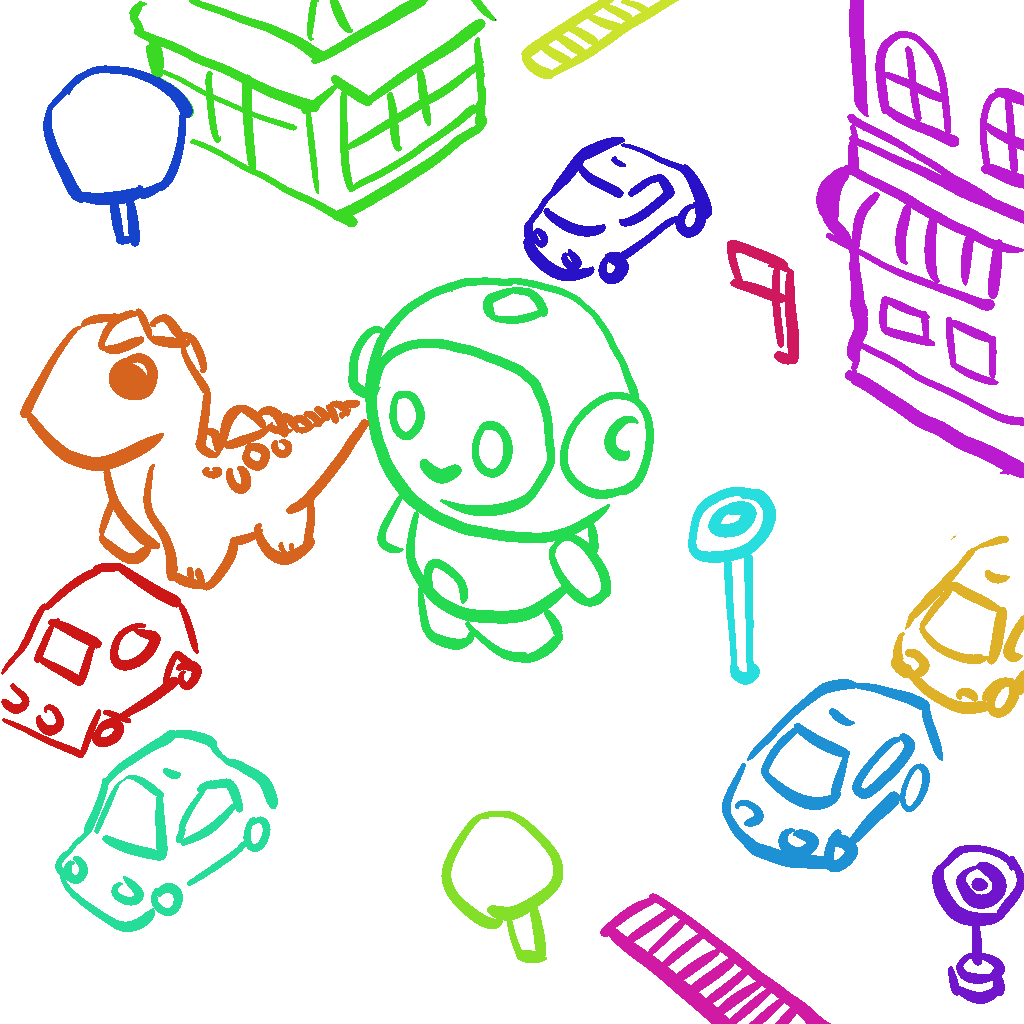}} &
        \frame{\includegraphics[width=0.3\linewidth]{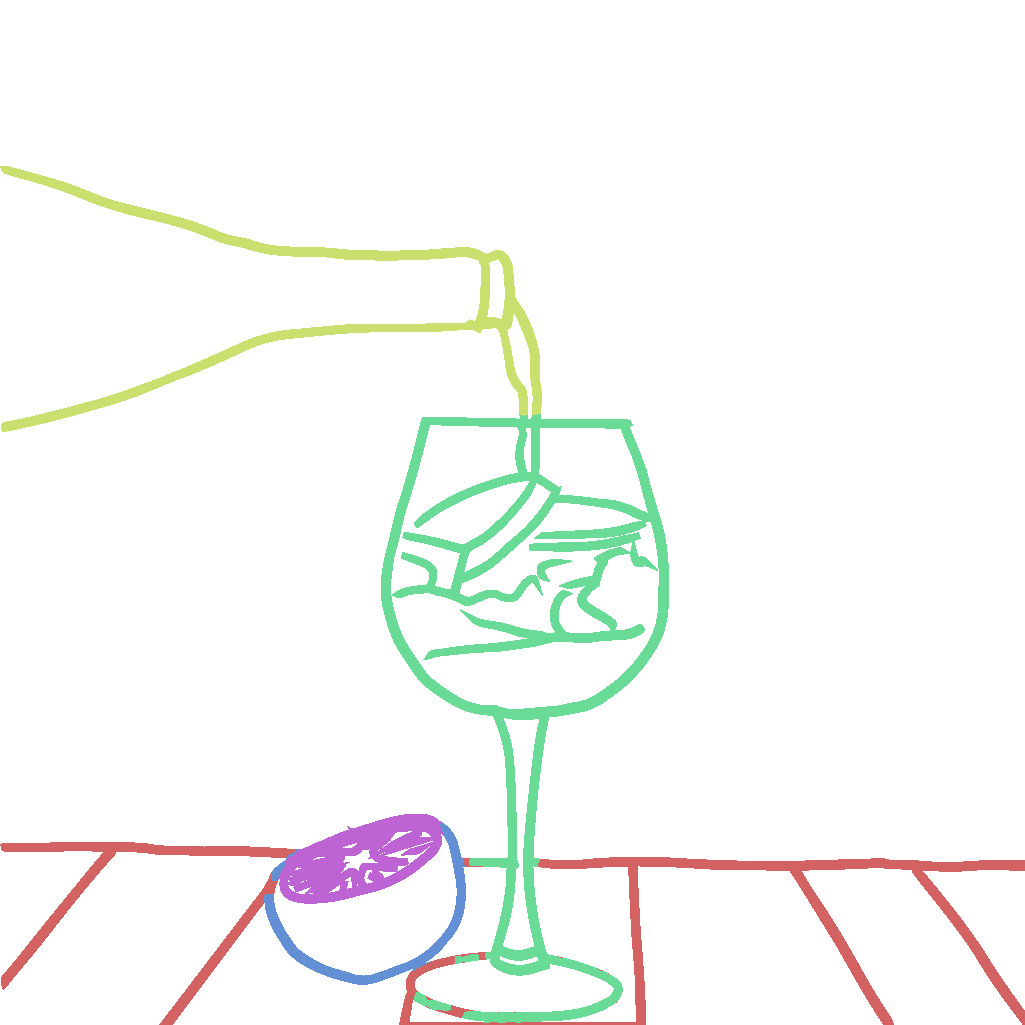}} \\
        
        \frame{\includegraphics[width=0.3\linewidth]{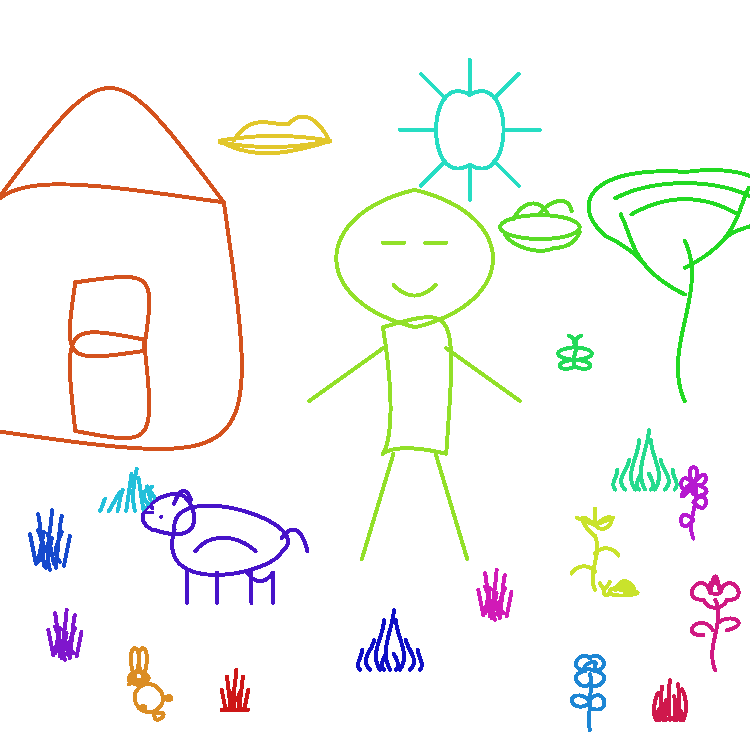}} &
        \frame{\includegraphics[width=0.3\linewidth]{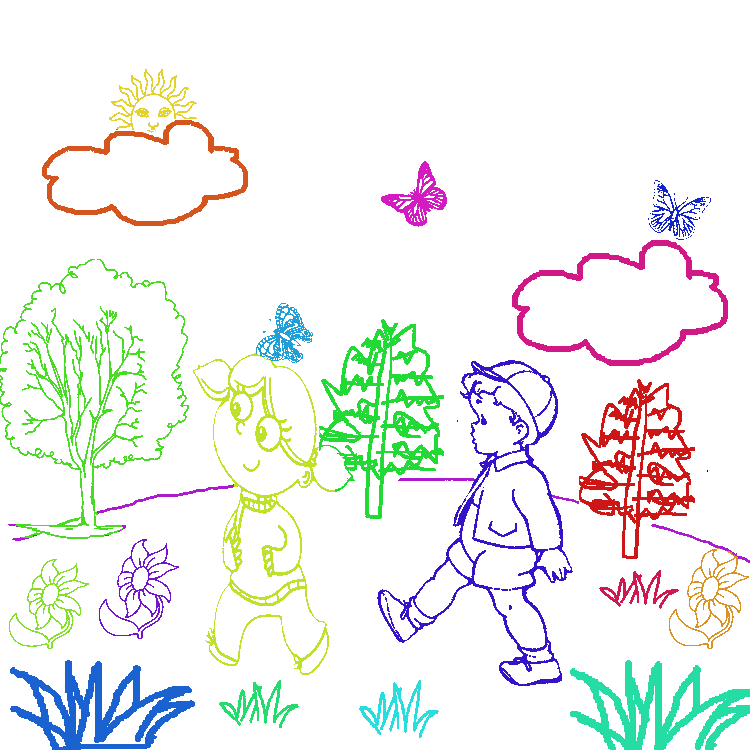}} &
        \frame{\includegraphics[width=0.3\linewidth]{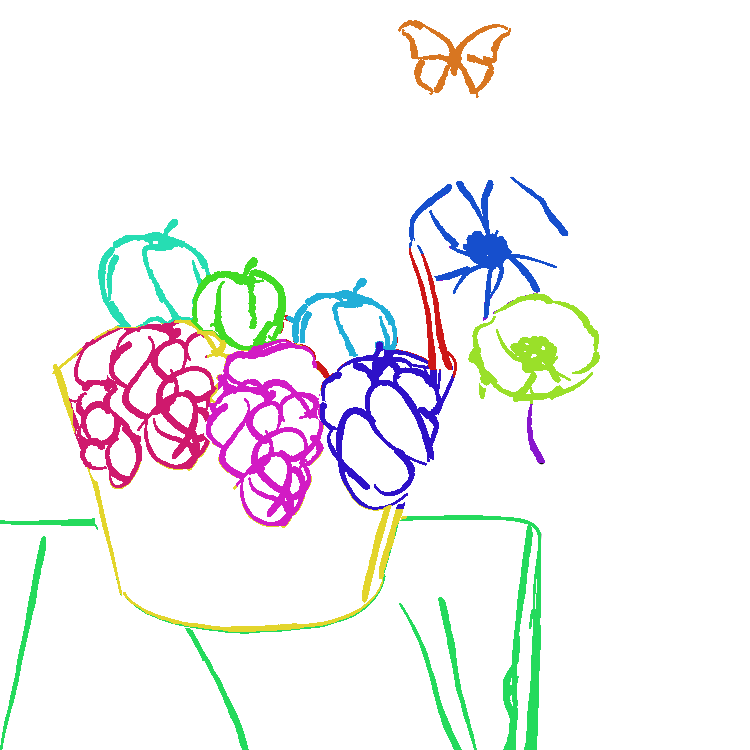}} \\
    \end{tabular}
    }
    \vspace{-0.2cm}
    \caption{\textbf{\methodname{} segmentation results. } Our method handles sketches of diverse styles and levels of complexity. The first two rows show artist-drawn sketches spanning a range of scenarios, including wildlife scenes, crowded urban markets, and cartoon characters in various settings. }
    \vspace{-4mm}
    \label{fig:qualitative}
\end{figure}

\Cref{fig:teaser,fig:qualitative,fig:qualitative_artists,fig:qualitative3} present qualitative results of our method across a diverse range of sketches. These include various object categories, both abstract and detailed scenes, different styles, and sketches from our new dataset featuring stroke variations and challenging abstractions. Our method effectively handles object categories beyond those used in our fine-tuning from the SketchyScene dataset, such as toys, furniture, and food items.
Our approach successfully addresses challenging scenarios, such as detailed scenes with numerous objects and occluded objects, as seen in \Cref{fig:teaser} and the first row of \Cref{fig:qualitative}. More results are provided in the supplementary material.
% Additionally, it demonstrates robustness to occluded objects, as illustrated in \Cref{fig:qualitative_artists}, and performs effectively on uncommon categories, such as exotic plants and abstract shapes, shown in \Cref{fig:qualitative3}. These results highlight the versatility and adaptability of our method to diverse sketching styles and complex object arrangements.

% \input{tables/object_detection_metrics}
% \multirow{2}{*}{\diagbox[height=5\line]{\raisebox{2ex}{Spalte}}{ & & \\Zeile}}
\begin{table*}
\small
\setlength{\tabcolsep}{2pt}
% \addtolength{\belowcaptionskip}{-5pt}
\centering
\caption{\textbf{Quantitative comparisons for object detection. }We report IoU, AR, and AP metrics across eight datasets, along with the mean and standard deviation for each method.  IoU measures the overlap between predicted and ground-truth bounding boxes, AR evaluates the ability to detect all relevant objects, and AP combines precision and recall across varying IoU thresholds from 50\% to 95\%. AP@50 and AP@75 indicate Average Precision at IoU thresholds of 50\% and 75\%, respectively, reflecting stricter requirements for bounding box overlap.  Our method demonstrates consistent improvements across nearly all datasets and metrics, significantly outperforming baselines, especially in detecting sketch objects with precision. } 
\vspace{-3mm} 
\label{tb:bbox_instance}

\begin{tabular}{l | c c c | c c c | c c c | c c c | c c c} 
    \toprule
    \multirow{2}{*}{\diagbox{Dataset}{Metric}} 
    &
    \multicolumn{3}{c|}{IoU $\uparrow$} & \multicolumn{3}{c|}{AR $\uparrow$} & \multicolumn{3}{c|}{AP $\uparrow$} & \multicolumn{3}{c|}{AP@50 $\uparrow$} & \multicolumn{3}{c}{AP@75 $\uparrow$} \\
    &  SketchyS & G-DINO & \textbf{Ours} & SketchyS & G-DINO & \textbf{Ours} & SketchyS & G-DINO & \textbf{Ours} & SketchyS & G-DINO & \textbf{Ours} & SketchyS & G-DINO & \textbf{Ours} \\
    \midrule
    SketchyScene & $0.55$ & $ 0.27 $ & $\mathbf{0.72}$ & $0.42$ & $0.27$ & $\mathbf{0.86}$ & $0.36$ & $0.24$ & $\mathbf{0.83}$ & $0.79$ & $0.31$ & $\mathbf{0.93}$ & $0.17$ & $0.27$ & $\mathbf{0.88}$ \\
    \rev{Zhang \etal} & $0.33$ & $0.53$ & $\mathbf{0.64}$ & $0.18$ & $0.48$ & $\mathbf{0.69}$ & $0.14$ & $0.38$ & $\mathbf{0.62}$ & $0.31$ & $0.57$ & $\mathbf{0.62}$ & $0.06$ & $0.41$ & $\mathbf{0.72}$ \\
    SketchAgent & $0.27$ & $0.16$ & $\mathbf{0.73}$ & $0.19$ & $0.16$ & $\mathbf{0.79}$ & $0.15$ & $0.12$ & $\mathbf{0.75}$  & $0.39$ & $0.18$ & $\mathbf{0.87}$ & $0.05$ & $0.13$ & $\mathbf{0.78}$ \\
    C-Base & $0.48$ & $0.31$ & $\mathbf{0.80}$ & $0.35$ & $0.30$ & $\mathbf{0.85}$ & $0.29$ & $0.26$ & $\mathbf{0.83}$ & $0.70$ & $0.37$ & $\mathbf{0.93}$ & $0.11$ & $0.30$ & $\mathbf{0.87}$\\
    C-Calligraphic & $0.48$ & $0.28$ & $\mathbf{0.72}$ & $0.35$ & $0.28$ & $\mathbf{0.84}$ & $0.29$ & $0.24$ & $\mathbf{0.81}$ & $0.71$ & $0.34$ & $\mathbf{0.93}$ & $0.11$ & $0.27$ & $\mathbf{0.86}$ \\
    C-Charcoal & $0.40$ & $0.27$ & $\mathbf{0.76}$ & $0.29$ & $0.27$ & $\mathbf{0.84}$ & $0.24$ & $0.24$ & $\mathbf{0.79}$ & $0.59$ & $0.33$ & $\mathbf{0.94}$ & $0.08$ & $0.27$ & $\mathbf{0.88}$ \\
    C-BrushPen & $0.41$ & $0.27$ & $\mathbf{0.75}$ & $0.30$ & $0.27$ & $\mathbf{0.82}$ & $0.25$ & $0.23$ & $\mathbf{0.79}$ & $0.60$ & $0.33$ & $\mathbf{0.90}$ & $0.10$ & $0.26$ & $\mathbf{0.82}$ \\
    InstantStyle & $ 0.20 $ & $\mathbf{0.49} $ & $0.45$ & $0.17 $ & $0.48 $ & $\mathbf{0.61} $ & $0.12 $ & $0.37 $ & $\mathbf{0.51} $ & $0.29$ & $0.53$ & $\mathbf{0.69}$ & $0.08$ & $0.40$ & $\mathbf{0.52}$ \\
    \midrule
     All & $	0.40 $ & $0.32 $ & $\mathbf{0.70}$ & $0.28 $ & $0.31 $ & $\mathbf{0.79}$ & $0.23 $ & $0.26 $ & $\mathbf{0.74}$ & $0.55$ & $0.37$ & $\mathbf{0.85}$ & $0.11$ & $0.29$ & $\mathbf{0.79}$ \\
    % $\text{std}^1$ 
    & \footnotesize{$\pm 0.10 $} & \footnotesize{$\pm 0.12$} & \footnotesize{$\pm 0.11$} & \footnotesize{$\pm 0.09$} & \footnotesize{$\pm 0.10$} & \footnotesize{$\pm 0.09$} & \footnotesize{$\pm 0.09$} & \footnotesize{$\pm 0.08$} & \footnotesize{$\pm 0.12$} & \footnotesize{$\pm 0.19$} &  \footnotesize{$\pm 0.13$} & \footnotesize{$\pm 0.12$}& \footnotesize{$\pm 0.04$} & \footnotesize{$\pm 0.09$} &  \footnotesize{$\pm 0.12$} \\
    \bottomrule
\end{tabular}
\vspace{-3mm}
\end{table*}

\begin{table}
\small
\setlength{\tabcolsep}{3pt}
% \addtolength{\belowcaptionskip}{-5pt}
\centering
\caption{\textbf{Quantitative comparisons from image segmentation. }We report Accuracy and IoU metrics across eight datasets, along with the mean and standard deviation for each method. Our method consistently outperforms baselines across all datasets.}
\label{tb:segmentation1}
\vspace{-2mm} 
\resizebox{\columnwidth}{!}{%
\begin{tabular}{l | c c c c | c c c c}
    \toprule
    \multirow{4}{*}{\diagbox{Dataset}{Method}}
    & \multicolumn{4}{c|}{{Acc $\uparrow$}} & \multicolumn{4}{c}{{IoU $\uparrow$}} \\ 
    &
    \makecell{Sketchy\\Scene} &
    \makecell{G-\\SAM} &
    \makecell{\rev{Auto-}\\\rev{SAM}} & 
    \textbf{Ours} & 
    \makecell{Sketchy\\Scene} &
    \makecell{G-\\SAM} &
    \makecell{\rev{Auto-}\\\rev{SAM}} & 
    \textbf{Ours} 
    \\
    \midrule
    SketchyScene & $0.79$ & $0.54$ & $0.24$ & $\mathbf{0.92}$ & $0.72$ & $0.26$ & $0.17$ & $\mathbf{0.88}$ \\
    \rev{Zhang \etal} & $0.42$ & $0.72$ & $0.09$ & $\mathbf{0.86}$ & $0.33$ & $0.51$ & $0.04$ & $\mathbf{0.74}$ \\
    SketchAgent & $0.39$ & $0.35$ & $0.12$ & $\mathbf{0.88}$ & $0.38$ & $0.16$ & $0.06$ & $\mathbf{0.84}$ \\
    C-Base & $0.70$ & $0.54$ & $0.08$ & $\mathbf{0.91}$ & $0.64$ & $0.32$ & $0.07$ & $\mathbf{0.88}$ \\
    C-Calligraphic  & $0.66$ & $0.50$ & $0.06$ & $\mathbf{0.87}$ & $0.63$ & $0.30$ & $0.05$ & $\mathbf{0.86}$ \\
    C-Charcoal  & $0.59$ & $0.47$ & $0.12$ & $\mathbf{0.85}$ & $0.43$ & $0.26$ & $0.06$ & $\mathbf{0.84}$ \\
    C-BrushPen & $0.66$ & $0.51$ & $0.05$ & $\mathbf{0.89}$ & $0.54$ & $0.29$ & $0.05$ & $\mathbf{0.85}$ \\
    InstantStyle & $0.43$ & $0.65$ & $0.25$ & $\mathbf{0.70}$ & $0.32$ & $0.44$ & $0.16$ & $ \mathbf{0.78}$ \\
    \midrule
    All & $0.58$ & $0.53$ & $0.13$ & $\mathbf{0.87}$ & $0.50$ & $0.32$ & $0.08$ & $\mathbf{0.82}$ \\
    & \footnotesize{$\pm 0.15 $} & \footnotesize{$\pm 0.11$} & \footnotesize{$\pm 0.08$} & \footnotesize{$\pm 0.04$} & \footnotesize{$\pm 0.16$} & \footnotesize{$\pm 0.11$} & \footnotesize{$\pm 0.05$} & \footnotesize{$\pm 0.07$}  \\
    \bottomrule
\end{tabular}
}
\vspace{-3mm} 
\end{table}

\begin{table}[h]
\small
\centering
\caption{\textbf{Quantitative comparisons from image segmentation on the filtered datasets.} We report Accuracy and IoU metrics across eight datasets, along with the mean and standard deviation for each method. Since Bourouis \etal performs semantic segmentation, and requires an input text prompt, we provide ground truth class labels as prompts to generate segmentations, \rev{and use 0.01 for confidence threshold to ensure all sketch pixels are segmented.}} 
\label{tb:bbox}
\vspace{-2mm} 
\resizebox{\columnwidth}{!}{%
\begin{tabular}{l|ccc|ccc}
    \toprule
    \multirow{4}{*}{\diagbox{Dataset}{Method}} &
    \multicolumn{3}{c|}{Acc $\uparrow$} & 
    \multicolumn{3}{c}{IoU $\uparrow$} \\
    & 
    \makecell{\rev{Bour-}\\ \rev{ouis}} & 
    \makecell{\rev{Sketch}\\ \rev{Seger}} & 
    \textbf{Ours} & 
    \makecell{\rev{Bour-}\\ \rev{ouis}} & 
    \makecell{\rev{Sketch}\\ \rev{Seger}} & 
    \textbf{Ours} \\
    \midrule
    SketchyScene&  $0.79$ & $\mathbf{0.96}$ & $0.93$ & $0.58$ & $\mathbf{0.90}$ & $\mathbf{0.90}$ \\
    \rev{Zhang \etal} &  $0.69$ & $0.65$ & $\mathbf{0.86}$ & $0.50$ & $0.22$ & $\mathbf{0.75}$ \\
    SketchAgent &  $0.71$ & $0.69$ & $\mathbf{0.93}$ & $0.52$ & $0.48$ & $\mathbf{0.88}$ \\
    C-Base  &  $0.83$ & $0.86$ & $\mathbf{0.94}$ & $0.66$ & $0.73$ & $\mathbf{0.92}$ \\
    C-Calligraphic &  $0.79$ & $0.76$ & $\mathbf{0.88}$ & $0.61$ & $0.61$ & $\mathbf{0.88}$ \\
    C-Charcoal  &  $0.80$ & $0.80$ & $\mathbf{0.85}$ & $0.61$ & $0.57$ & $\mathbf{0.85}$ \\
    C-BrushPen &  $0.81$ & $0.79$ & $\mathbf{0.91}$ & $0.63$ & $0.59$ & $\mathbf{0.88}$ \\
    InstantStyle &  $0.76$ &  $0.70$  & $\mathbf{0.80}$ & $0.51$ &  $0.45$  & $\mathbf{0.75}$ \\
     \midrule
    All & $0.77$ & $0.78$  & $\mathbf{0.89}$ & $0.58$ &  $0.57$ &  $\mathbf{0.85}$ \\
    & \footnotesize{$\pm 0.05 $} & \footnotesize{$\pm 0.10$}  & \footnotesize{$\pm 0.05$} & \footnotesize{$\pm 0.06$} & \footnotesize{$\pm 0.20$} &  \footnotesize{$\pm 0.07$} \\
    \bottomrule
\end{tabular}
}
\vspace{-3mm}
\end{table}

\subsection{Comparisons}
\begin{figure}
    \centering
    \vspace{-2mm}
    \setlength{\tabcolsep}{2pt}
    \resizebox{\columnwidth}{!}{%
    {\small
    \begin{tabular}{c c c c c}
        Input & SketchyScene & Grounded SAM & Ours \\
        
         \frame{\includegraphics[width=0.23\linewidth]{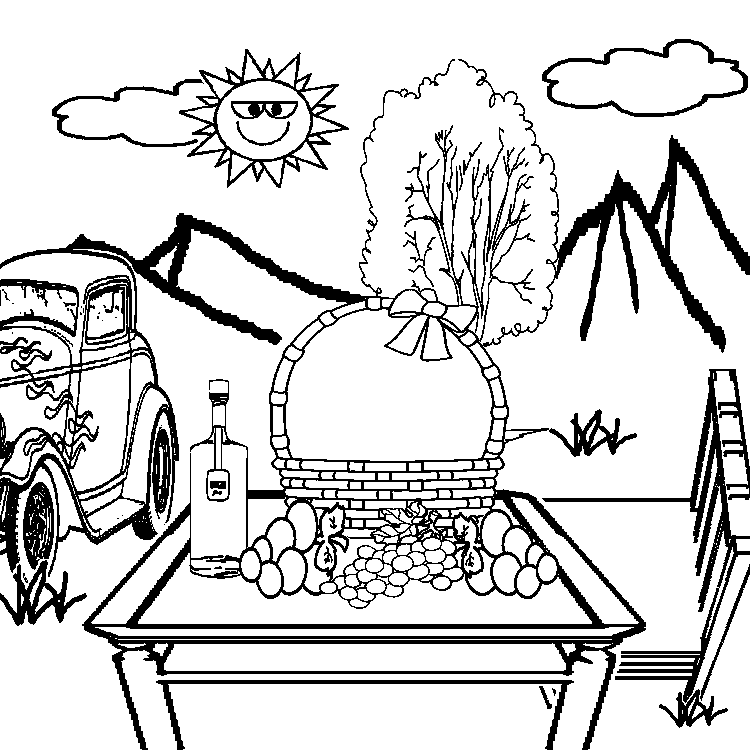}} &
        \frame{\includegraphics[width=0.23\linewidth]{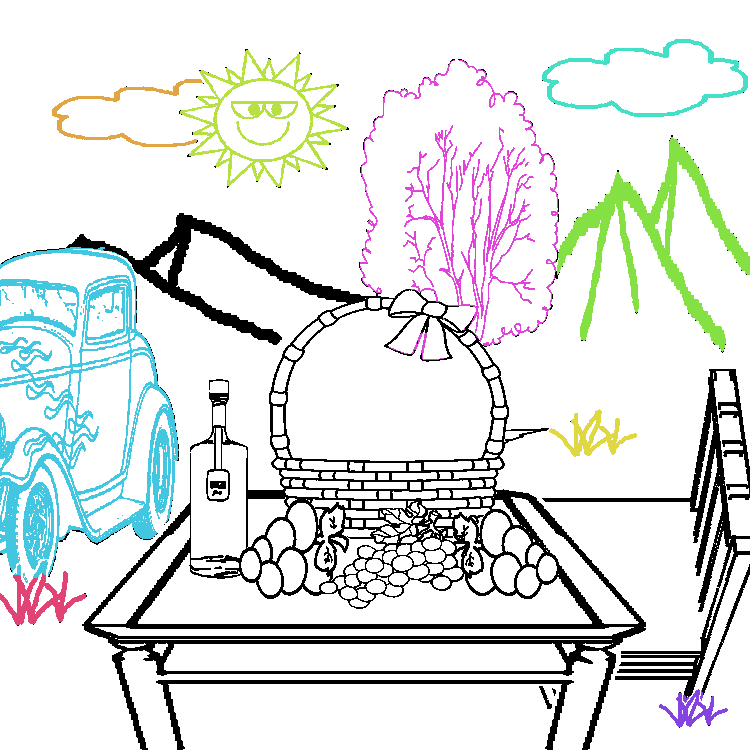}} &
        \frame{\includegraphics[width=0.23\linewidth]{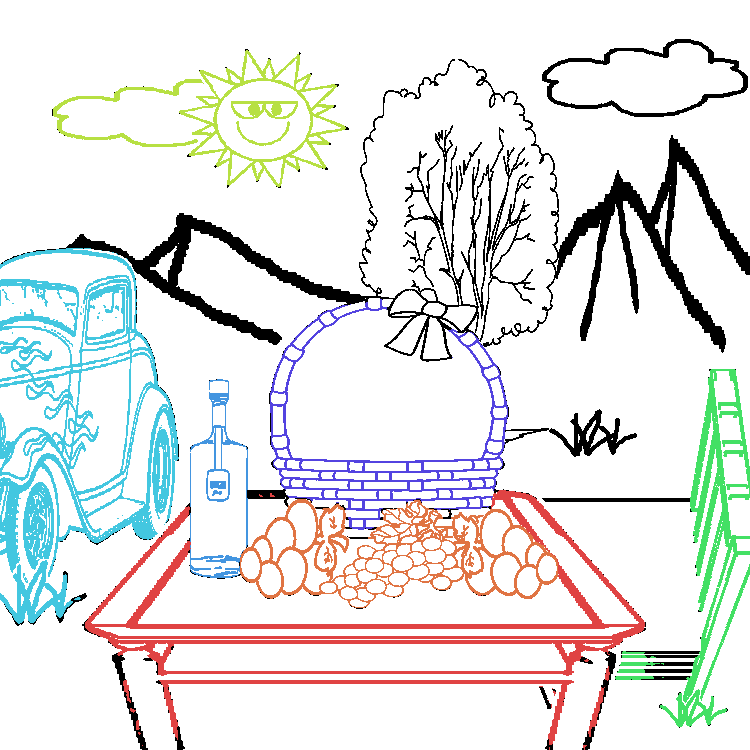}} &
        \frame{\includegraphics[width=0.23\linewidth]{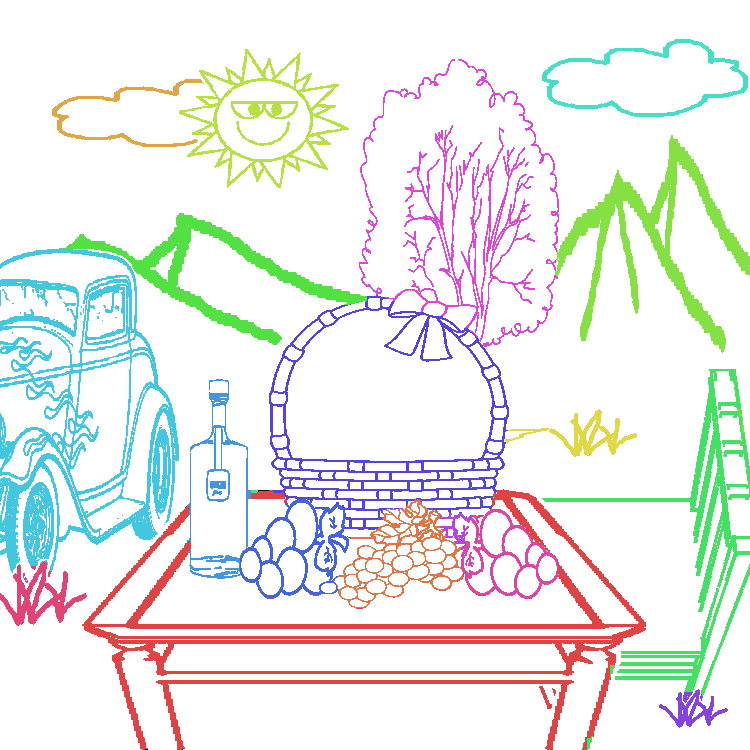}} \\

        % \frame{\includegraphics[width=0.23\linewidth]{figs/fig9_comparison_visualizations/0115004430/input.png}} &
        % \frame{\includegraphics[ width=0.23\linewidth]{figs/fig9_comparison_visualizations/0115004430/SketchyScene.png}} &
        % \frame{\includegraphics[width=0.23\linewidth]{figs/fig9_comparison_visualizations/0115004430/groundedSAM.png}} &
        % \frame{\includegraphics[width=0.23\linewidth]{figs/fig9_comparison_visualizations/0115004430/ours.png}} \\
        
        \frame{\includegraphics[trim=0 100 0 110, clip, width=0.23\linewidth]{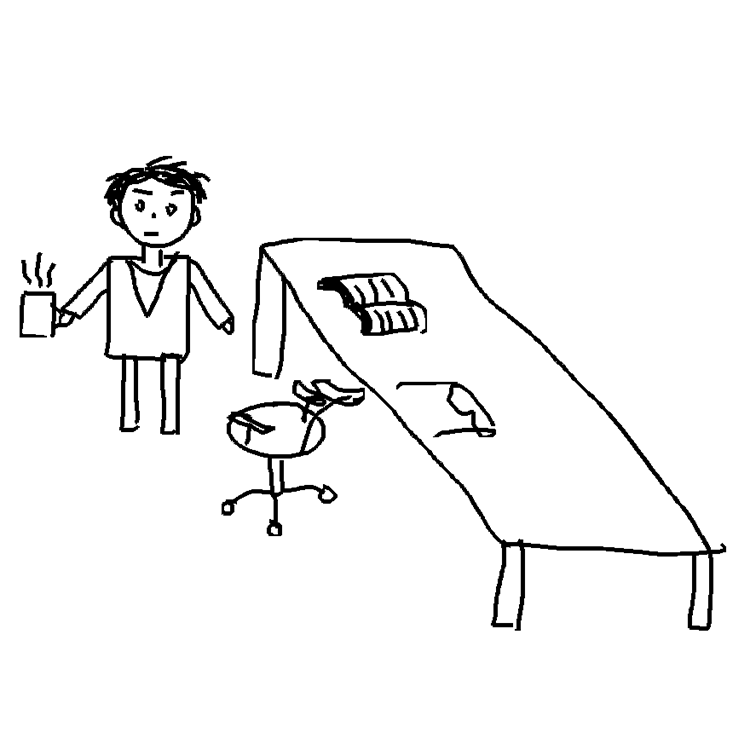}} &
        \frame{\includegraphics[trim=0 100 0 110, clip, width=0.23\linewidth]{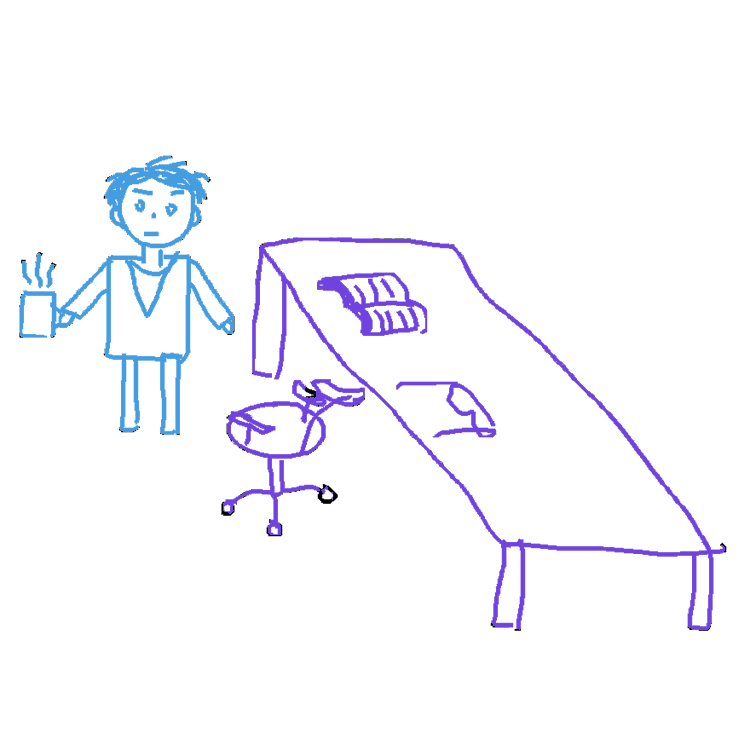}} &
        \frame{\includegraphics[trim=0 100 0 110, clip, width=0.23\linewidth]{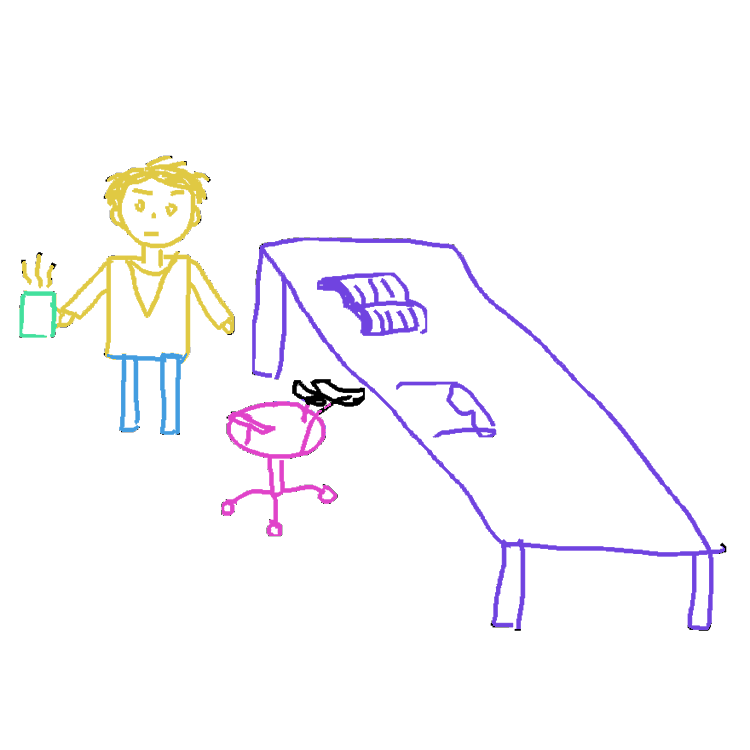}} &
        \frame{\includegraphics[trim=0 100 0 110, clip, width=0.23\linewidth]{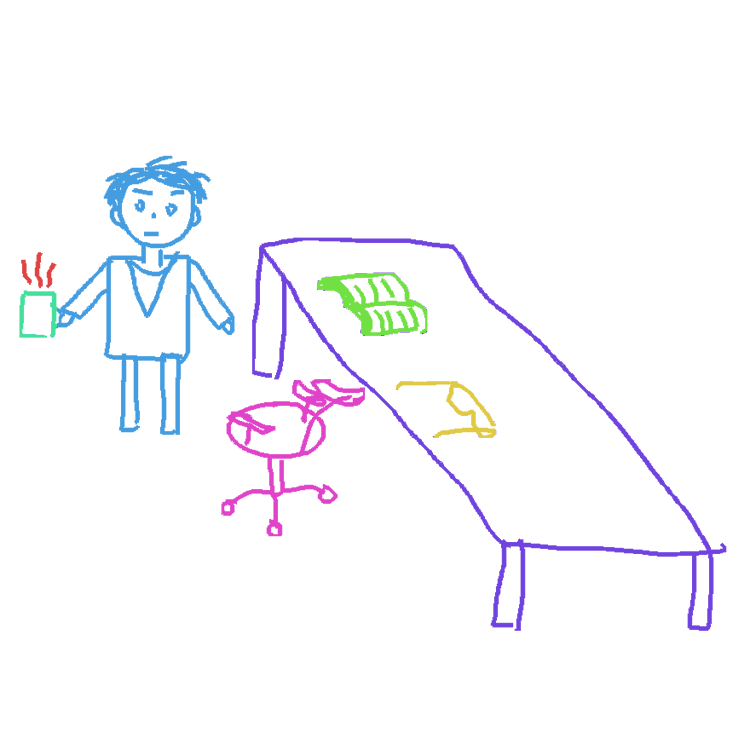}} \\

        \frame{\includegraphics[width=0.23\linewidth]{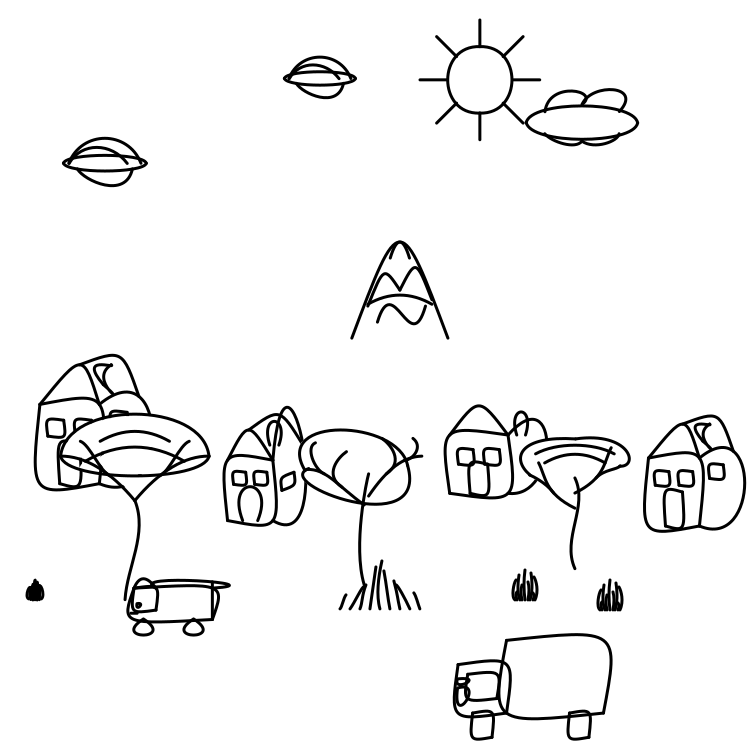}} &
        \frame{\includegraphics[width=0.23\linewidth]{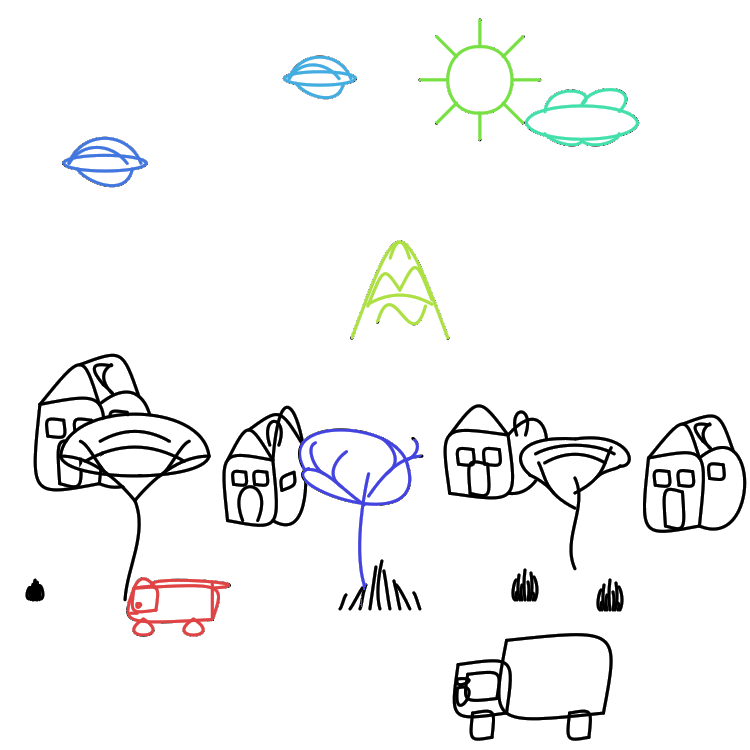}} &
        \frame{\includegraphics[width=0.23\linewidth]{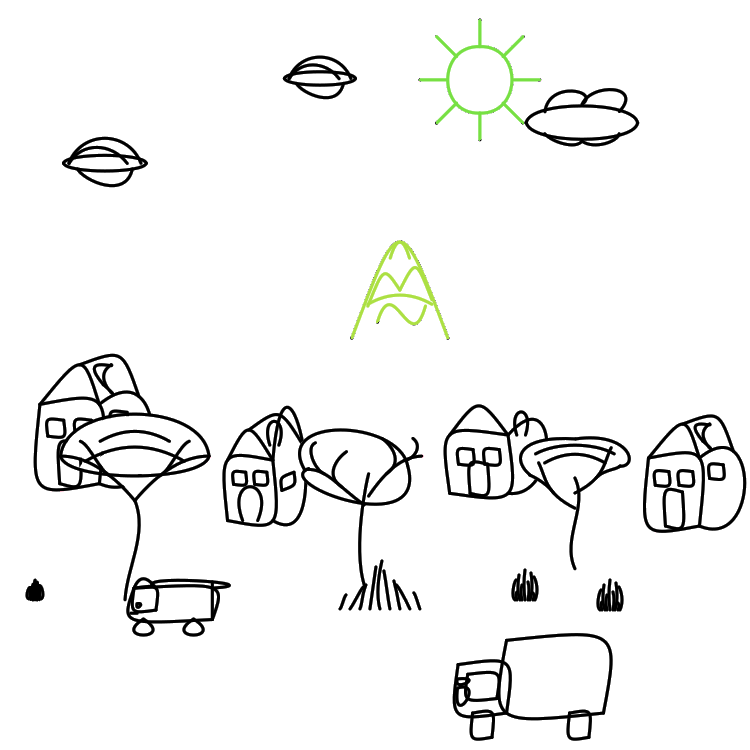}} &
        \frame{\includegraphics[width=0.23\linewidth]{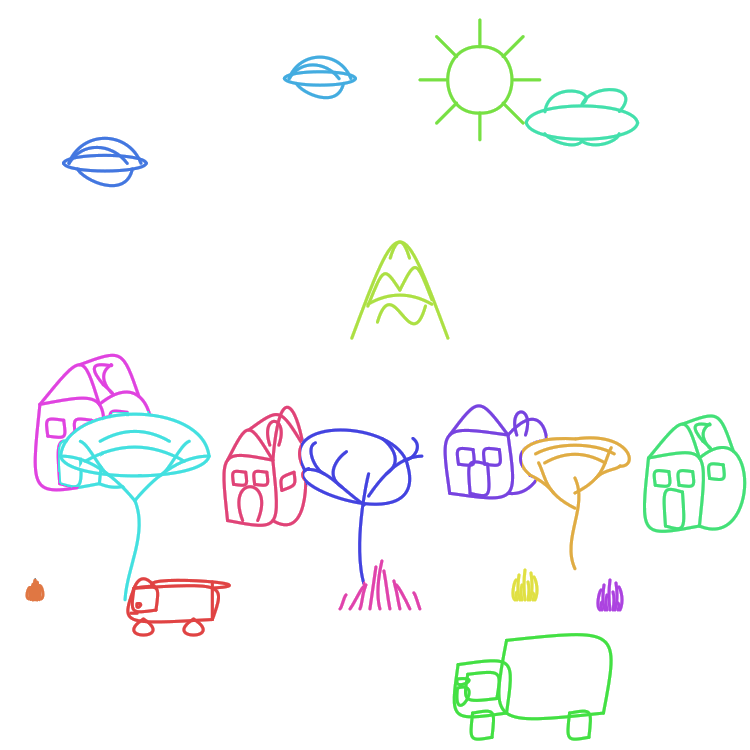}} \\

        \frame{\includegraphics[width=0.23\linewidth]{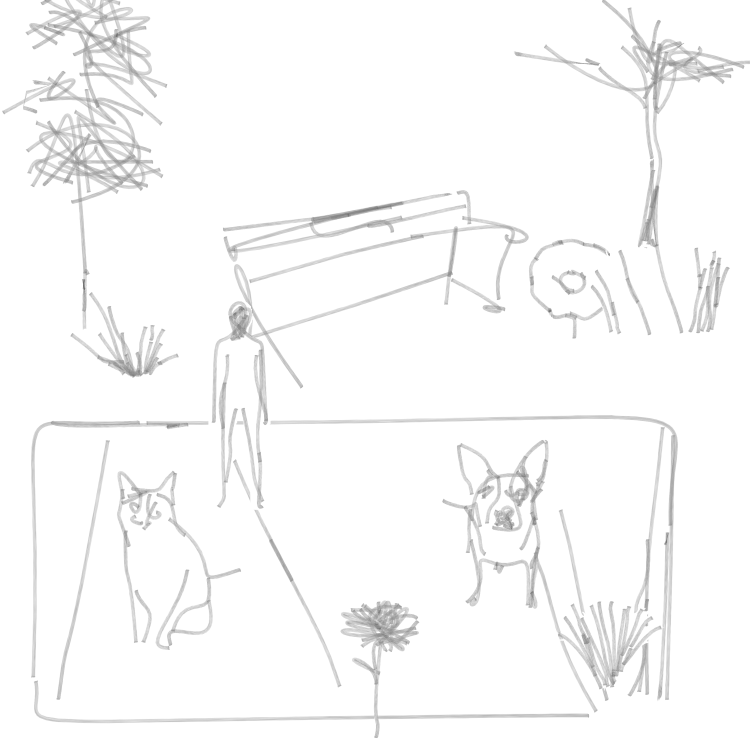}} &
        \frame{\includegraphics[width=0.23\linewidth]{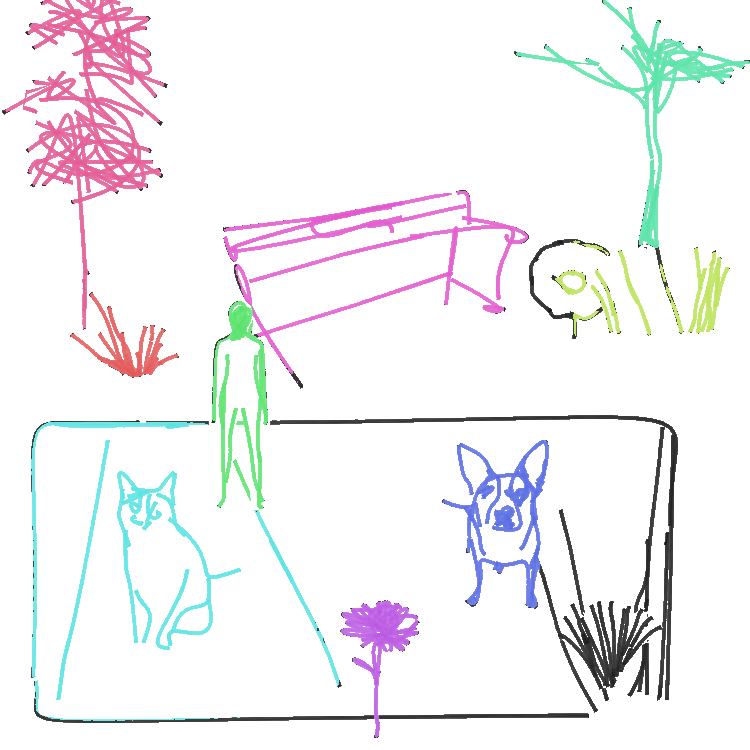}} &
        \frame{\includegraphics[width=0.23\linewidth]{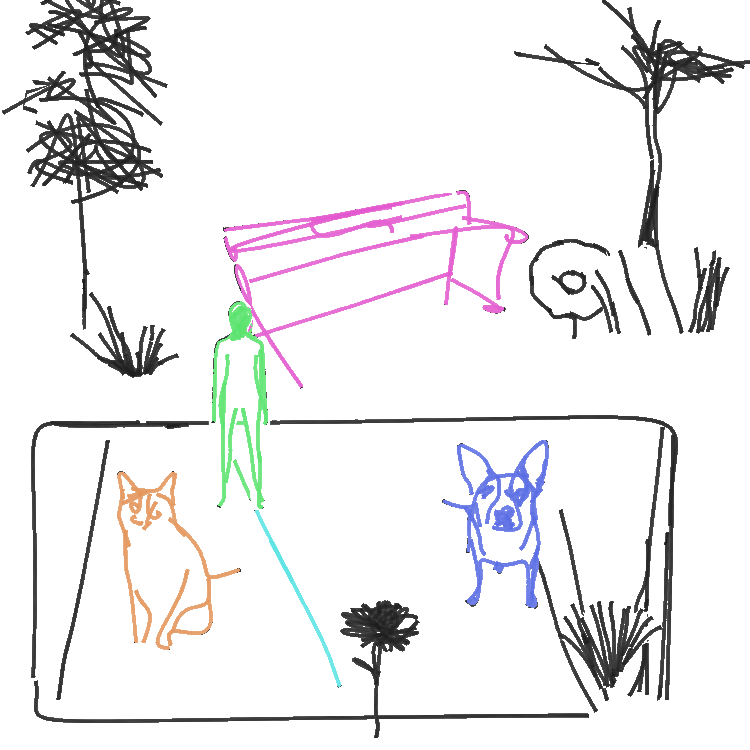}} &
        \frame{\includegraphics[width=0.23\linewidth]{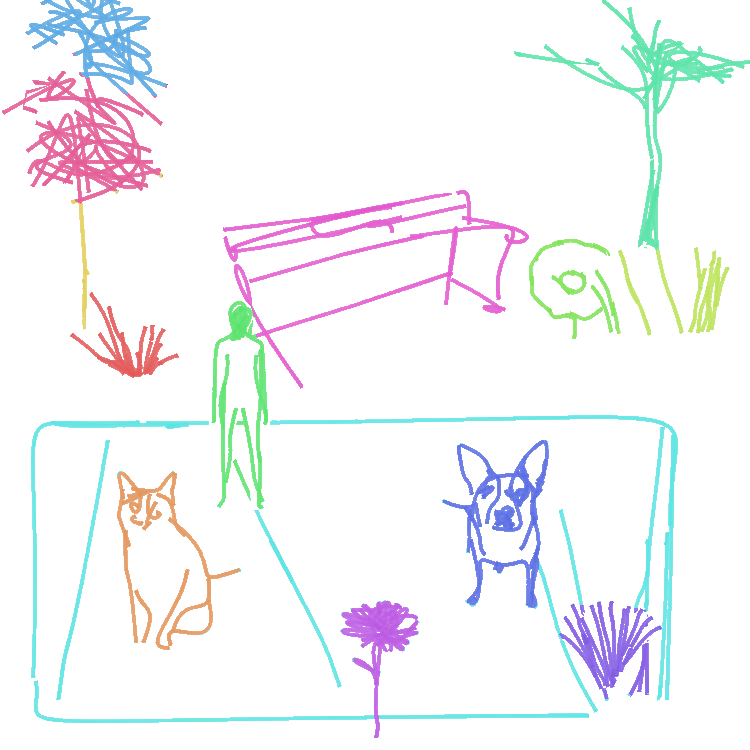}} \\

        \frame{\includegraphics[width=0.23\linewidth]{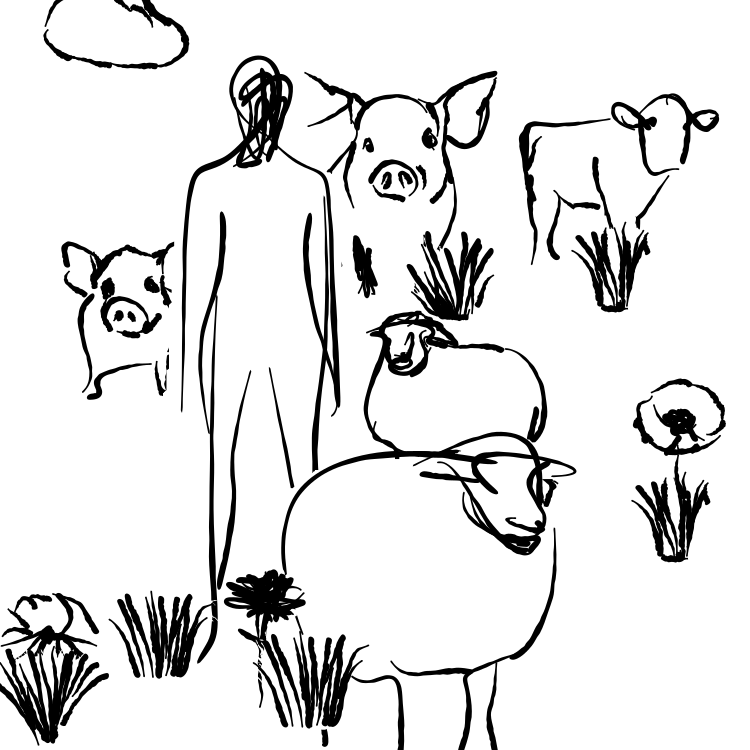}} &
        \frame{\includegraphics[width=0.23\linewidth]{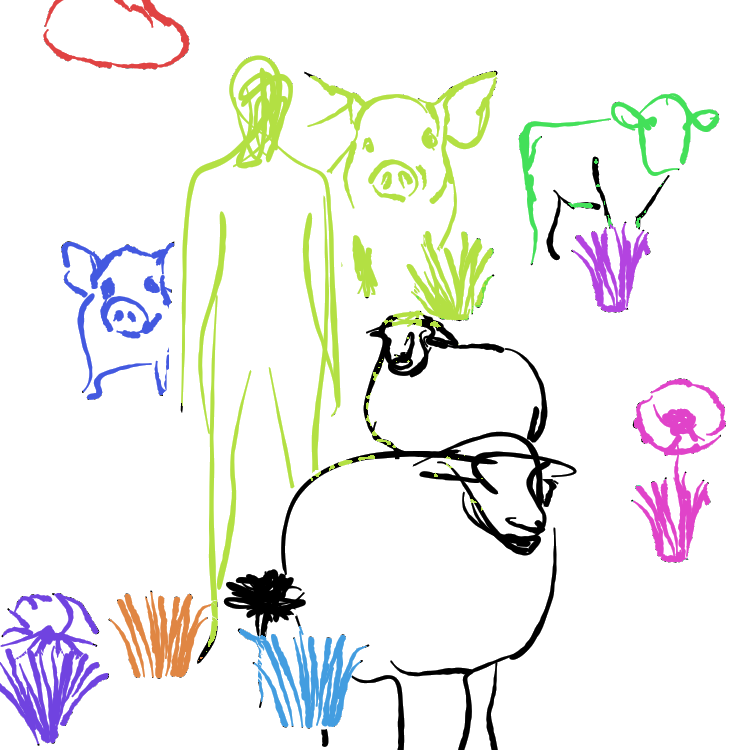}} &
        \frame{\includegraphics[width=0.23\linewidth]{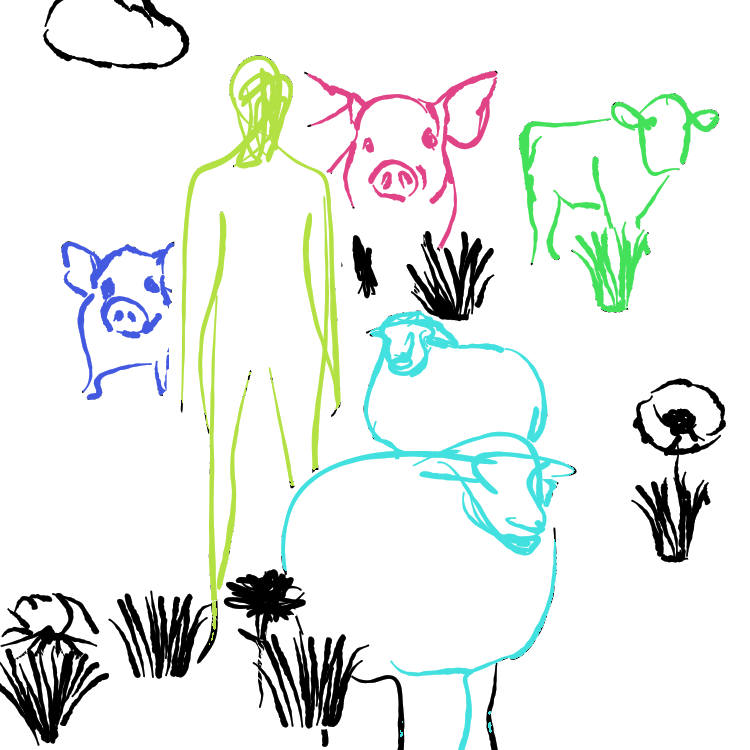}} &
        \frame{\includegraphics[width=0.23\linewidth]{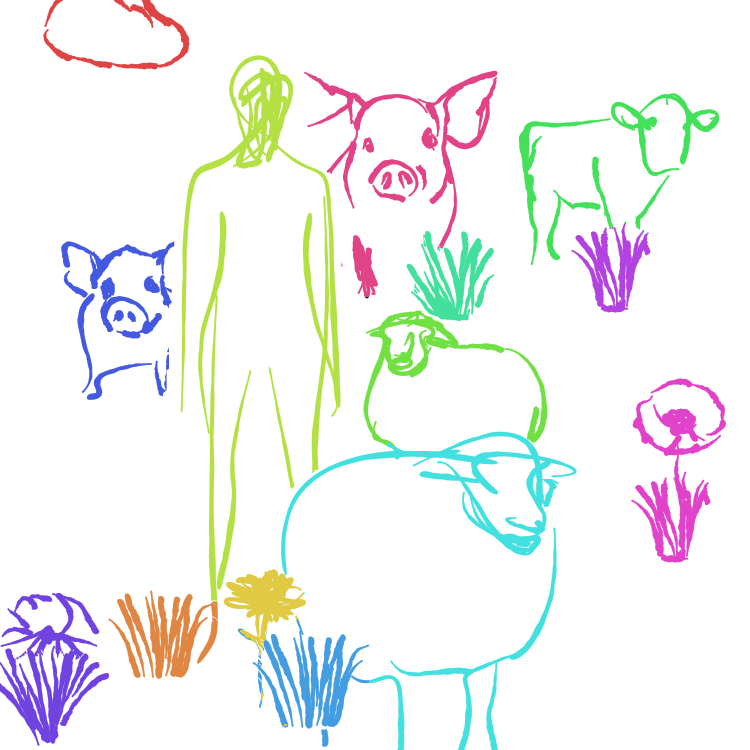}} \\

        \frame{\includegraphics[width=0.23\linewidth]{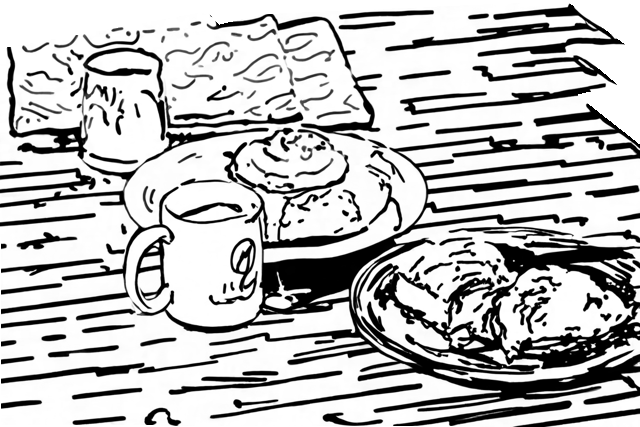}} &
        \frame{\includegraphics[width=0.23\linewidth]{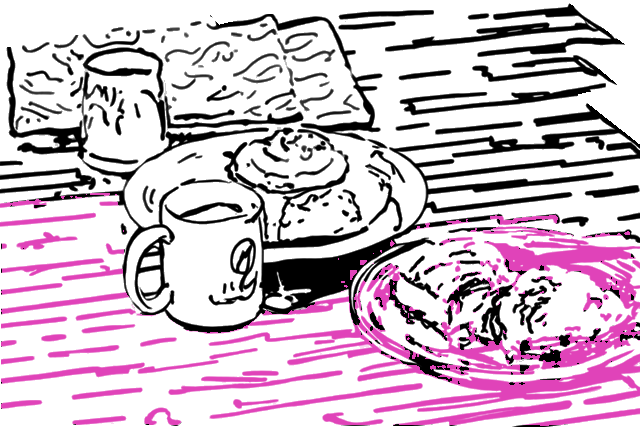}} &
        \frame{\includegraphics[width=0.23\linewidth]{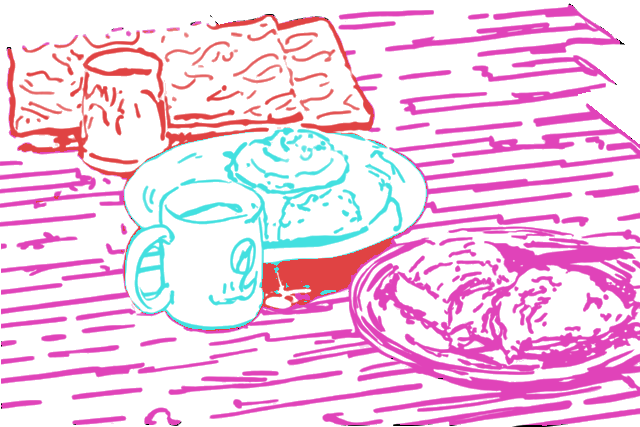}} &
        \frame{\includegraphics[width=0.23\linewidth]{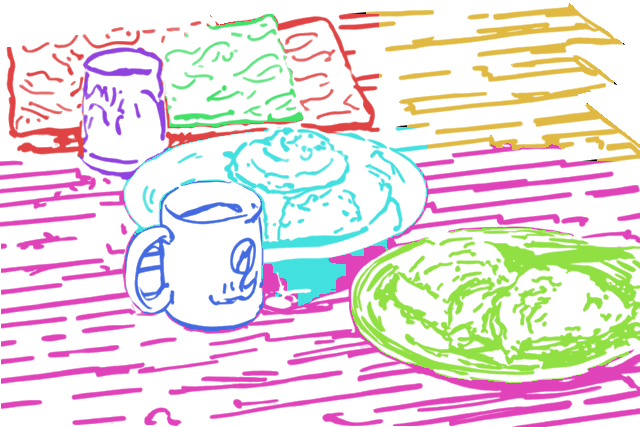}} \\
    \end{tabular}
    }
    }
    \vspace{-0.3cm}
    \caption{\textbf{Qualitative comparison of instance segmentation methods. }Each row corresponds to a different sketch dataset. Black pixels indicate regions where segmentation was not applied. Our method effectively segments sketch pixels into distinct instances, outperforming alternative approaches.}
    \label{fig:comparison}
    \vspace{-4mm}
\end{figure}

We evaluate our method alongside existing scene sketch segmentation approaches, including SketchyScene \cite{Zou18SketchyScene}, \rev{SketchSeger \cite{yang2023sceneHierTransformer}}, and the method proposed by Bourouis \etal \shortcite{bourouis2024open}. We include additional baselines: Grounding DINO \cite{liu2023grounding} applied directly to sketches, automatic labeling with the Recognize-Anything Model (RAM) \cite{zhang2023recognize}, and \rev{the Automatic SAM pipeline, which samples a 32×32 uniform grid of prompts and selects the top 50 masks to form disjoint regions.} Our evaluation dataset consists of \rev{8076} samples: $1,113$ test samples from the SketchyScene dataset, $330$ samples from the Zhang \etal dataset, $1,113$ samples from each of the CLIPasso brush styles: Base, Calligraphic Pen, Charcoal, and Brush Pen, $1,113$ samples from the SketchAgent style, and $1068$ samples from the InstantStyle sketch dataset.
Each method was applied to these datasets following their recommended best practices. Note that the SketchyScene mask generation implementation relies on legacy dependencies that are no longer executable. Therefore, we used SAM for mask generation based on their detected bounding boxes, which yielded better performance than reported in the original paper. \Cref{fig:comparison} illustrates selected segmentation results for all instance segmentation methods, with additional results provided in the supplementary material. Since the method by Bourouis \etal \shortcite{bourouis2024open} \rev{and SketchSeger \shortcite{yang2023sceneHierTransformer} }are designed for semantic segmentation rather than instance segmentation, it was evaluated separately on a filtered subset of our dataset, where each scene contained only one instance per class. 

\paragraph{Object Detection Evaluation}
For object detection, we report Intersection over Union (IoU), which measures the overlap between predicted and ground-truth bounding boxes, Average Recall (AR), which evaluates the ability to detect all relevant objects, and Average Precision (AP), which combines precision and recall across various IoU thresholds. Our goal is to assess the ability to precisely detect any object in the sketch, regardless of its class. To achieve this, we calculate the mean of these metrics across object instances rather than across classes. 
% This approach allows for flexibility in class label predictions, meaning that incorrect class labels are not penalized, and it accommodates methods that do not produce class labels.
The results for instance segmentation methods are summarized in \cref{tb:bbox_instance}. 
% The datasets are listed in the leftmost column, and each set of three columns corresponds to a specific metric (shown at the top) for each method. \maneesh{Don't need to describe things that visually obvious in the table or things taht will go in the table caption. Here I would just focus on saying the amin takeaways -- we perform better than the others. Even better if we can point to sharp increases in performance and explain why we see those increases.}
The SketchyScene method performs well on the SketchyScene data, while its performance significantly declines across all metrics when applied to other datasets, particularly for the challenging styles of the SketchAgent samples. 
In contrast, our method demonstrates consistent performance across all datasets, achieving an average AR score of \rev{0.79}, as shown in the last row of the table. Notably, our method outperforms SketchyScene even on its native dataset, demonstrating the effectiveness of leveraging priors from pretrained models on natural images and adapting them for sketch segmentation.
The scores obtained for our baseline method, Grounding DINO, support our claim that this model struggles to generalize to the domain of sketches without adaptation, despite its strong performance on natural images. This is evident from the large margin in scores between our method and Grounding DINO, seeing an increase of \rev{38\%} in IoU, \rev{48\%} increase in AR, and \rev{48\%} increase in AP.  Furthermore, while we fine-tuned Grounding DINO exclusively on the SketchyScene dataset, the results in the table confirm that this approach surprisingly generalizes well to very different types of sketches and object categories.

\paragraph{Segmentation Evaluation}
We evaluate the final segmentation results using two common metrics: Pixel Accuracy (Acc), which measures the ratio of correctly labeled pixels to the total pixel count in a sketch, and  Intersection over Union (IoU), which evaluates the overlap between the predicted and ground-truth segmentation masks. The results for instance segmentation methods are presented in \cref{tb:segmentation1}, while the results for Bourouis \etal \shortcite{bourouis2024open} and \rev{SketchSeger \shortcite{yang2023sceneHierTransformer}} are shown separately in \cref{tb:bbox} since it performs semantic segmentation and requires dataset filtering. 
As shown, our method outperforms alternative approaches across both metrics, with a particularly notable advantage over Grounded SAM. \rev{While SketchSeger performs well on SketchyScene dataset, its performance degrades significantly on the symbolic style SketchAgent and struggles to generalize to novel categories included in Zhang \etal and InstantStyle styles.} Meanwhile, our method demonstrates robustness to various styles, especially excelling on the challenging \rev{Zhang \etal and } SketchAgent style compared to other methods. 

\begin{figure}
    \centering
    \setlength{\tabcolsep}{2pt}
    \resizebox{\columnwidth}{!}{%
    {\small
    \begin{tabular}{c c c c}
        \frame{\includegraphics[width=0.3\linewidth]{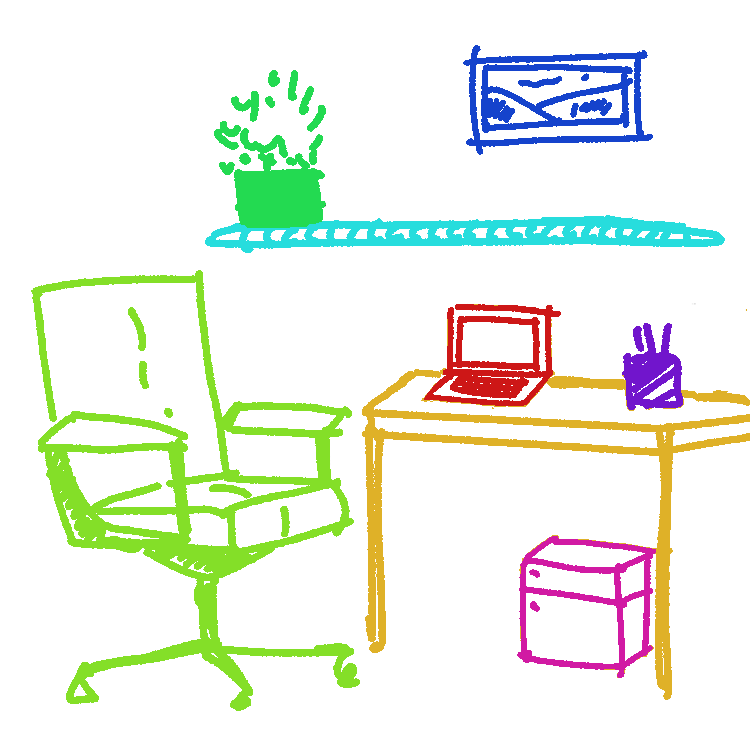}} &
        \frame{\includegraphics[width=0.3\linewidth]{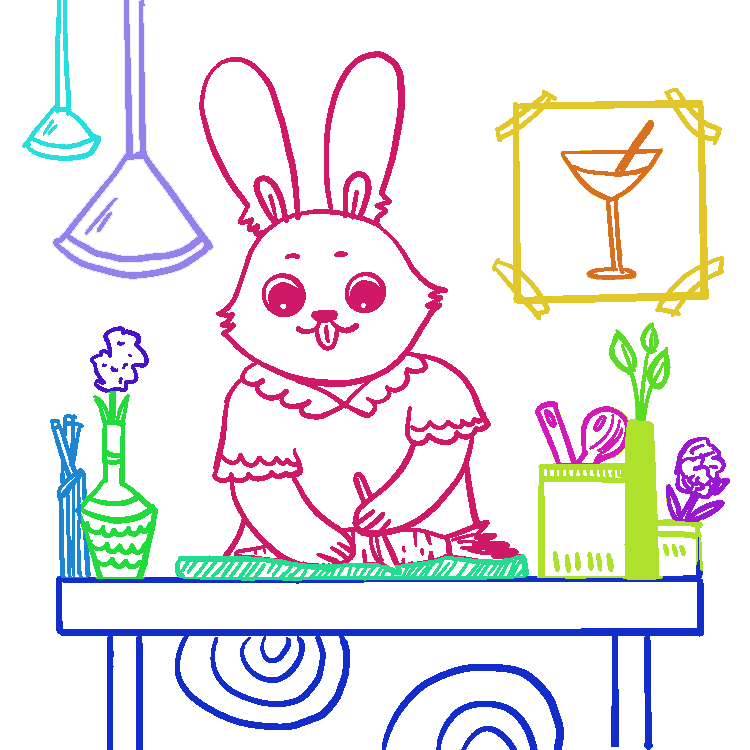}} & \frame{\includegraphics[width=0.3\linewidth]{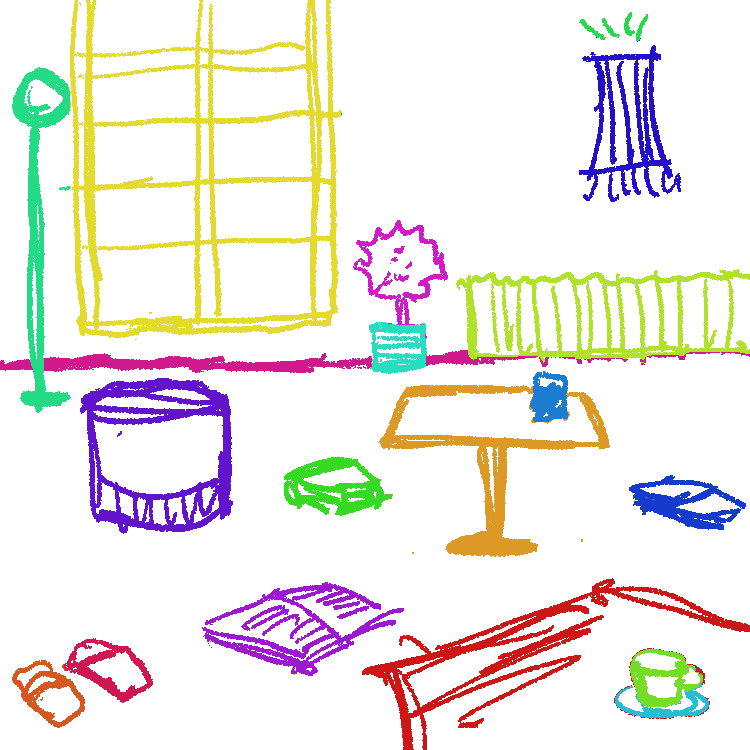}} &
        \frame{\includegraphics[width=0.3\linewidth]{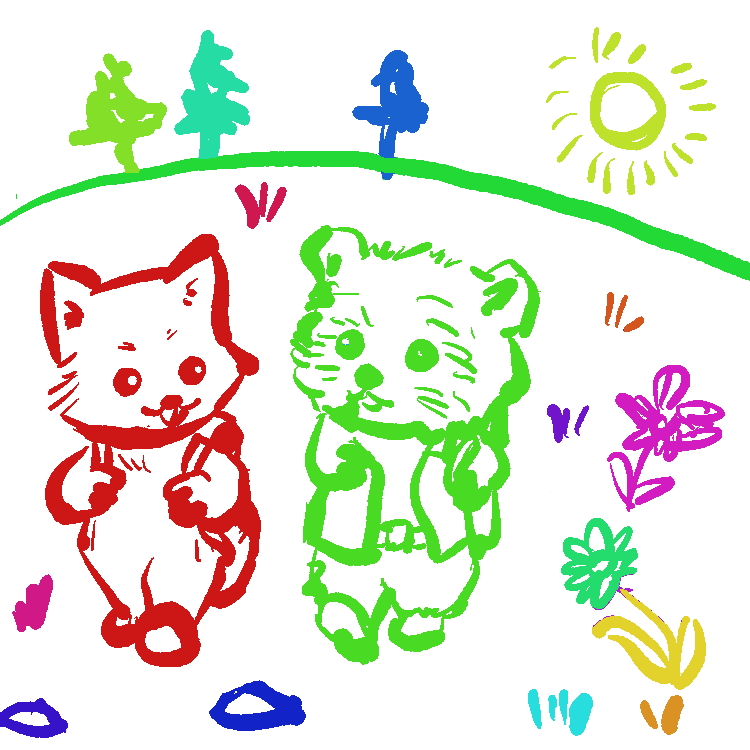}} \\
        
        \frame{\includegraphics[width=0.3\linewidth]{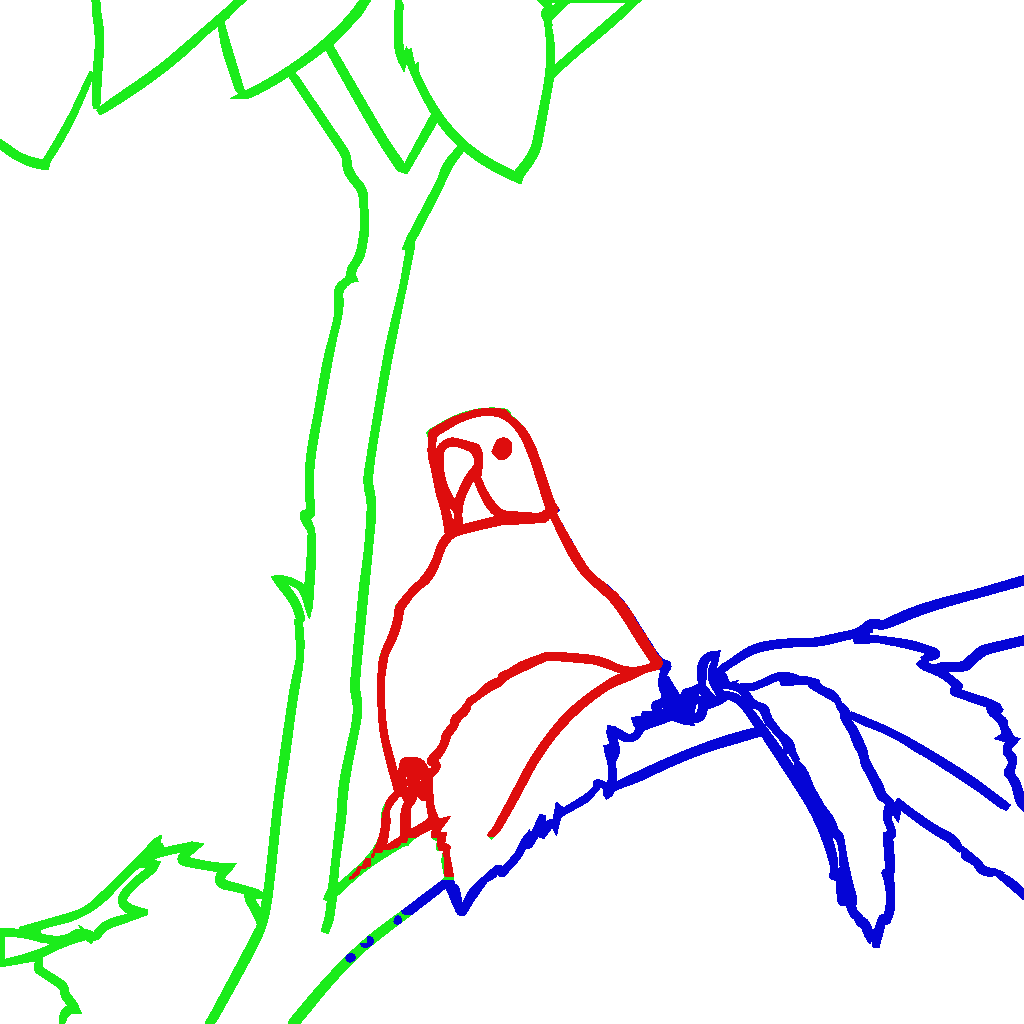}} &
        \frame{\includegraphics[width=0.3\linewidth]{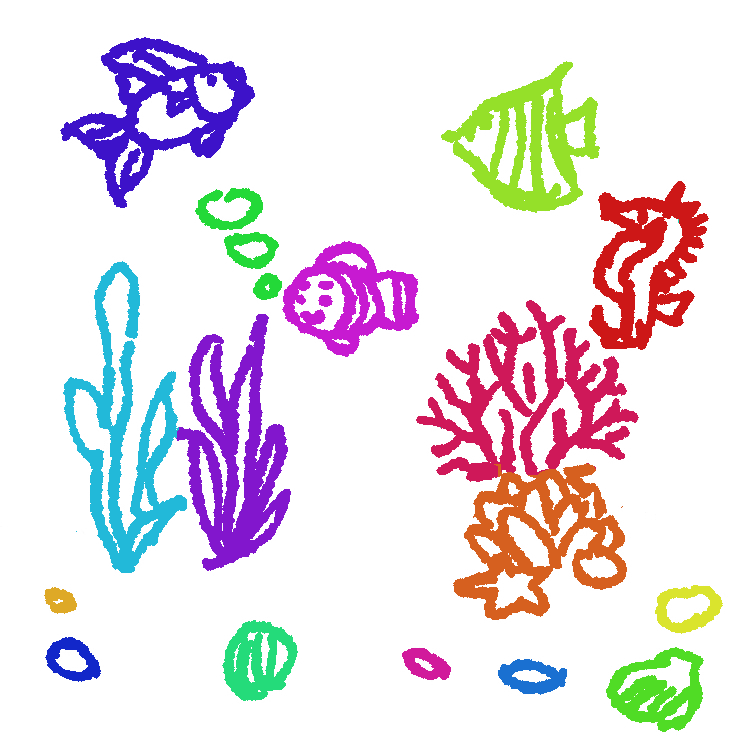}} & \frame{\includegraphics[width=0.3\linewidth]{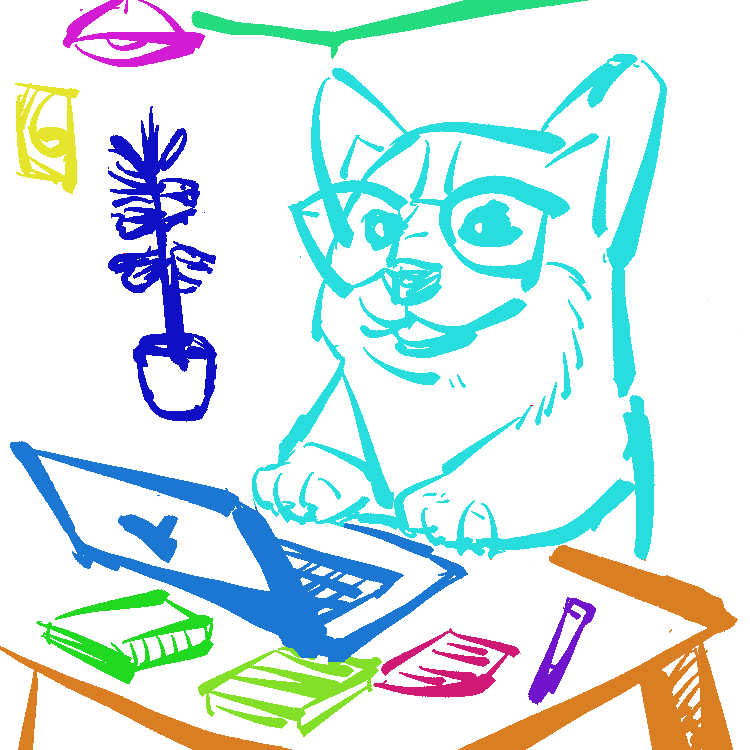}} &
        \frame{\includegraphics[width=0.3\linewidth]{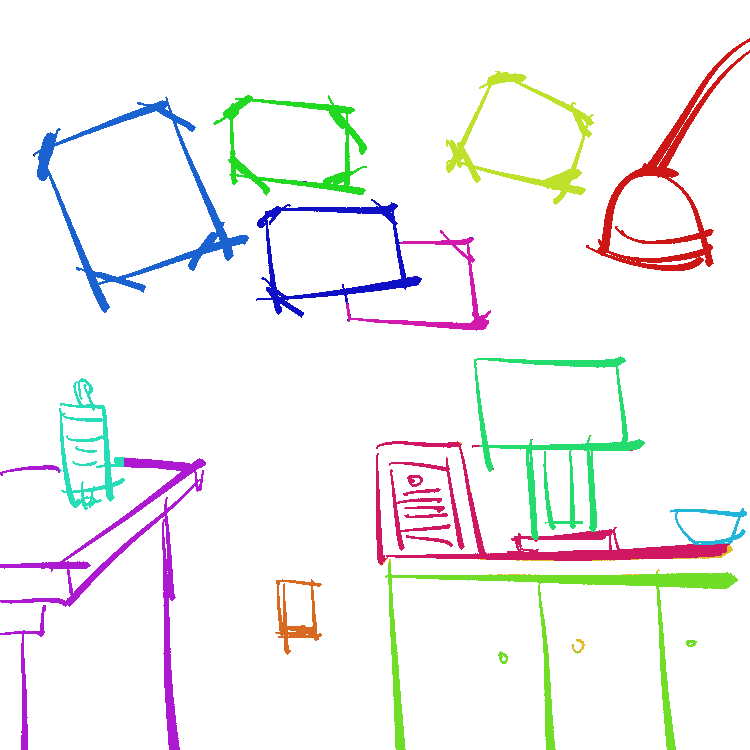}} \\
        
        \frame{\includegraphics[width=0.3\linewidth]{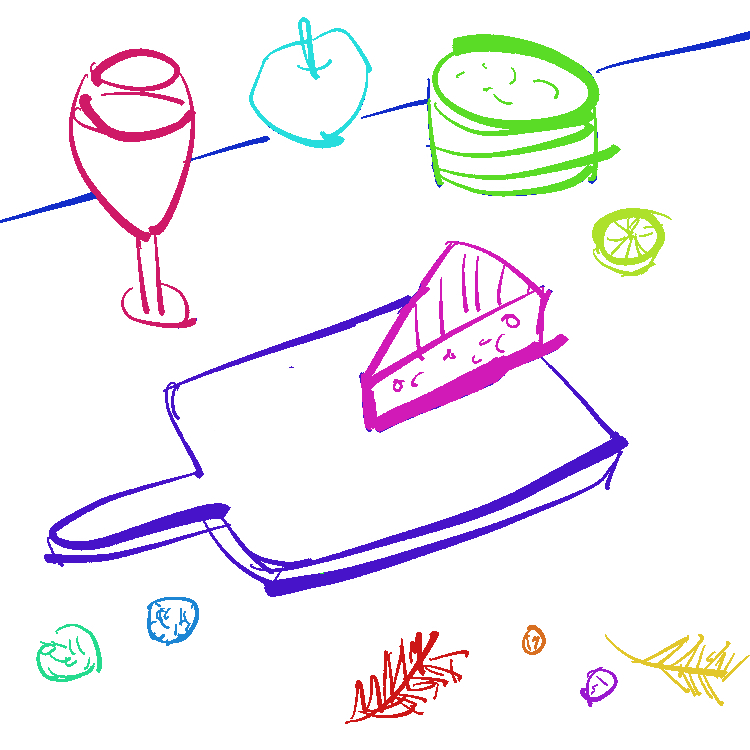}} &
        \frame{\includegraphics[width=0.3\linewidth]{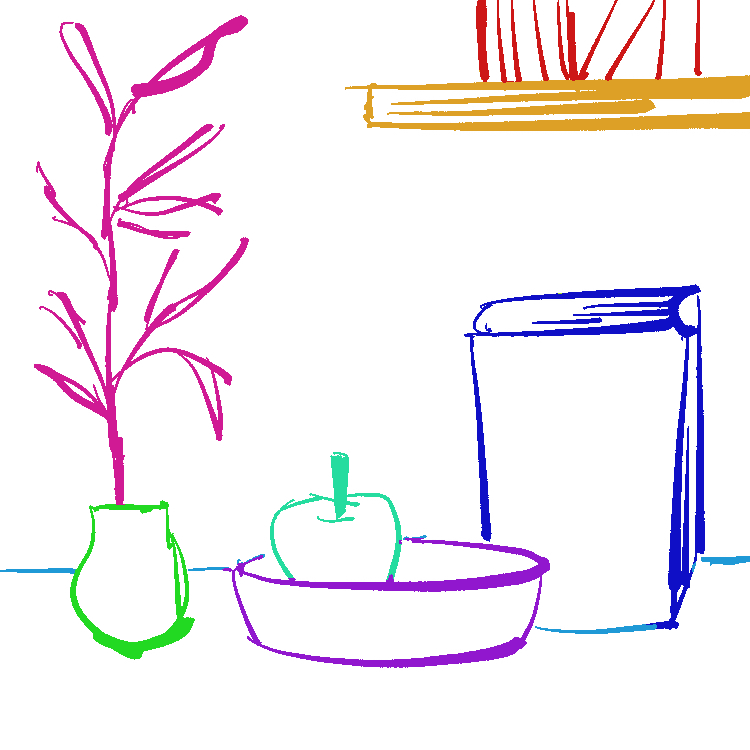}} & \frame{\includegraphics[width=0.3\linewidth]{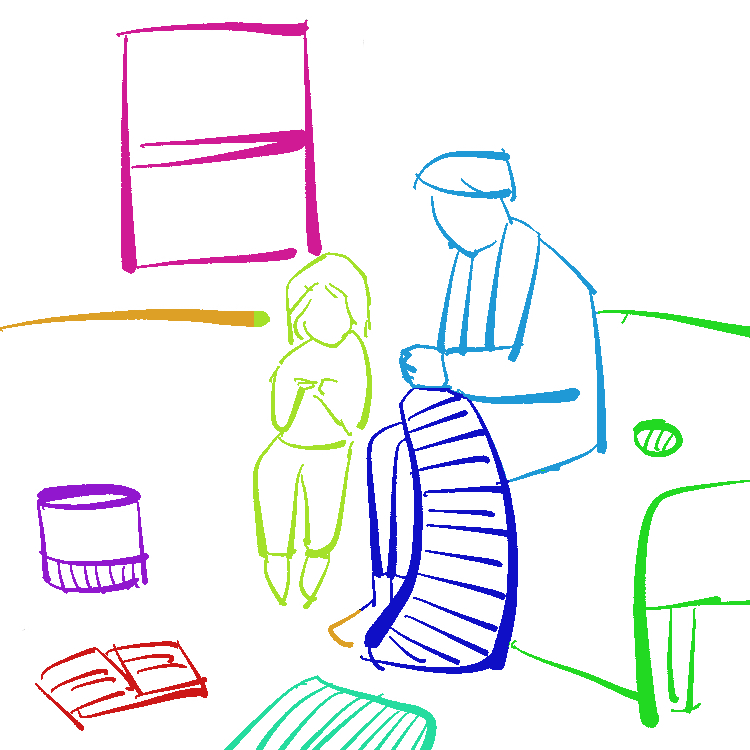}} &
        \frame{\includegraphics[width=0.3\linewidth]{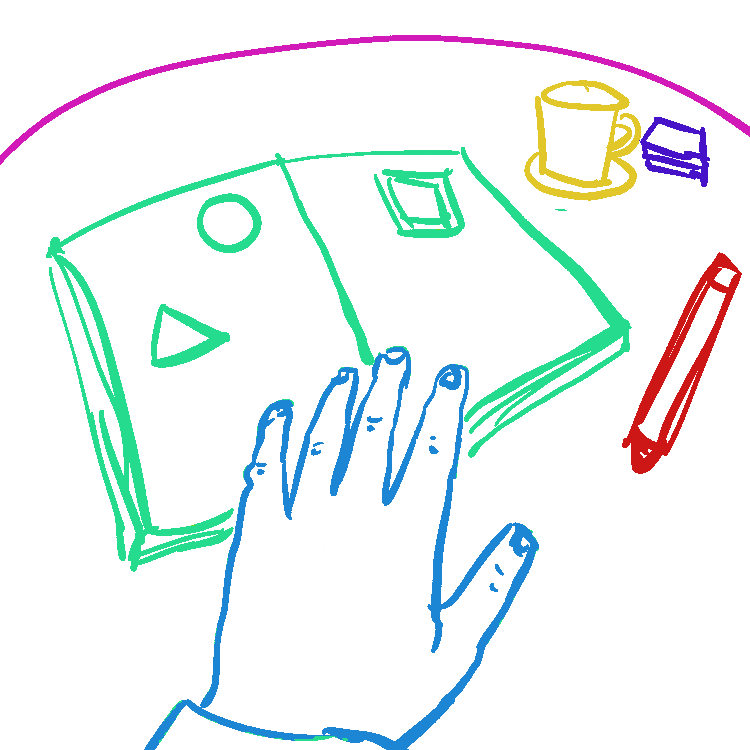}} \\

        \frame{\includegraphics[width=0.3\linewidth]{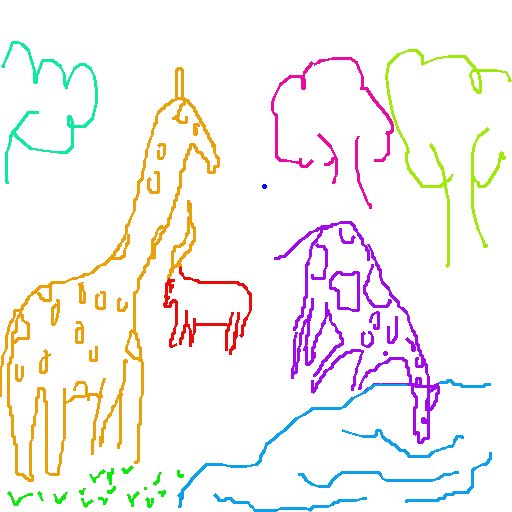}} &
        \frame{\includegraphics[width=0.3\linewidth]{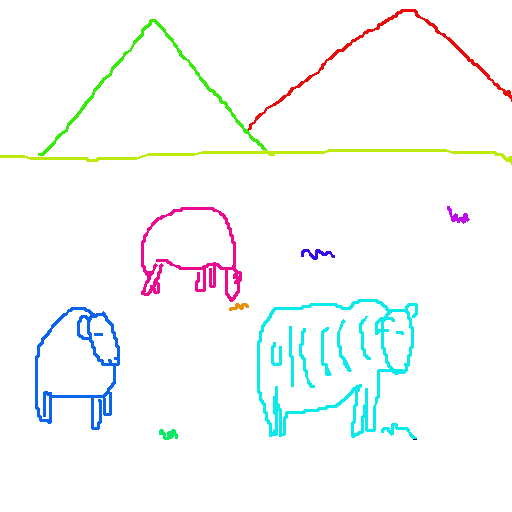}} &
        \frame{\includegraphics[width=0.3\linewidth]{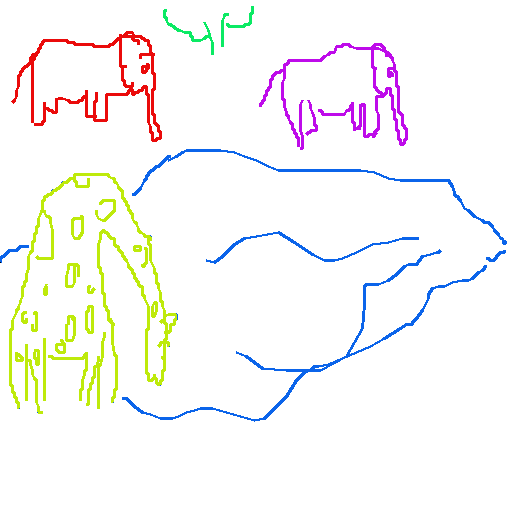}} &
        \frame{\includegraphics[width=0.3\linewidth]{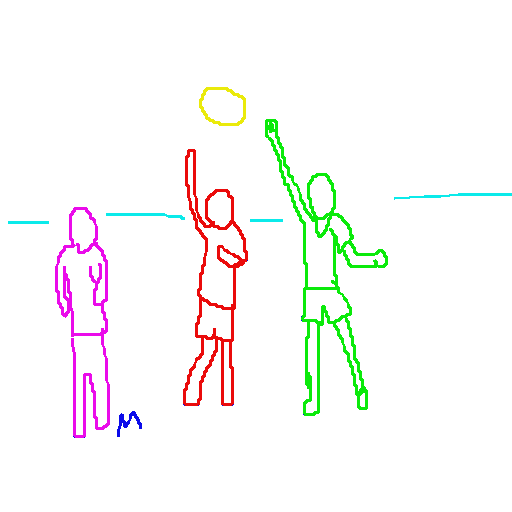}}\\

        \frame{\includegraphics[trim=0 100 0 50, clip, width=0.3\linewidth]{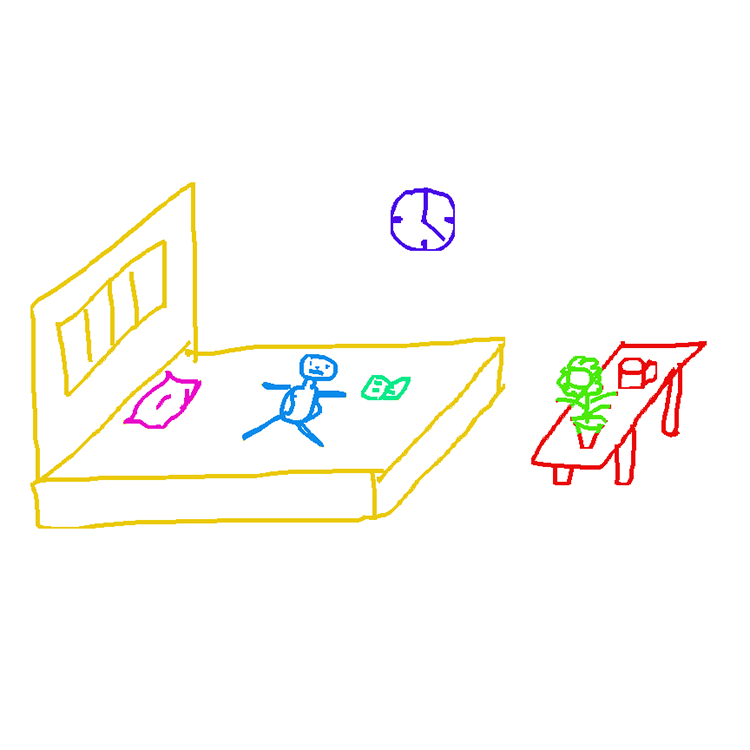}} &
        \frame{\includegraphics[trim=0 100 0 50, clip, width=0.3\linewidth]{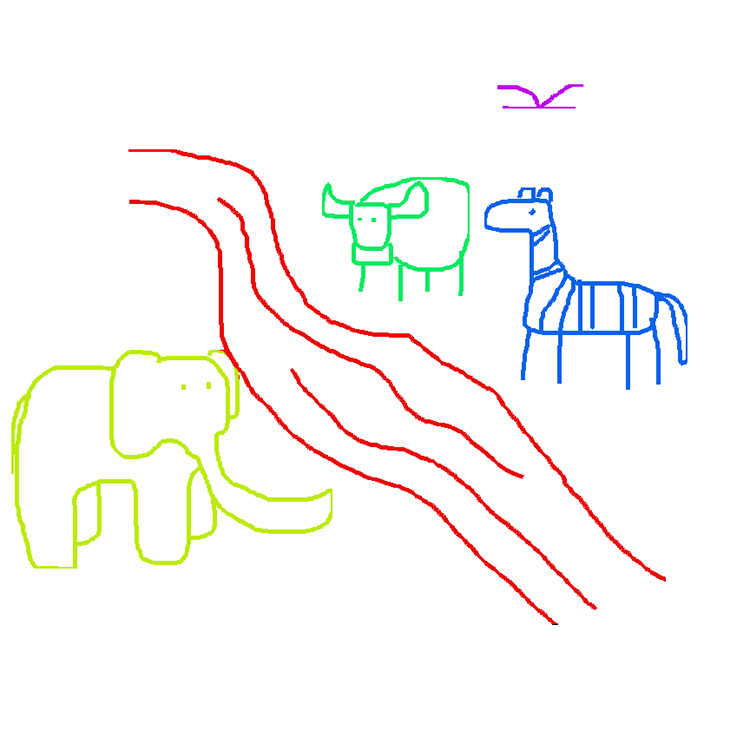}} &
        \frame{\includegraphics[trim=0 100 0 50, clip, width=0.3\linewidth]{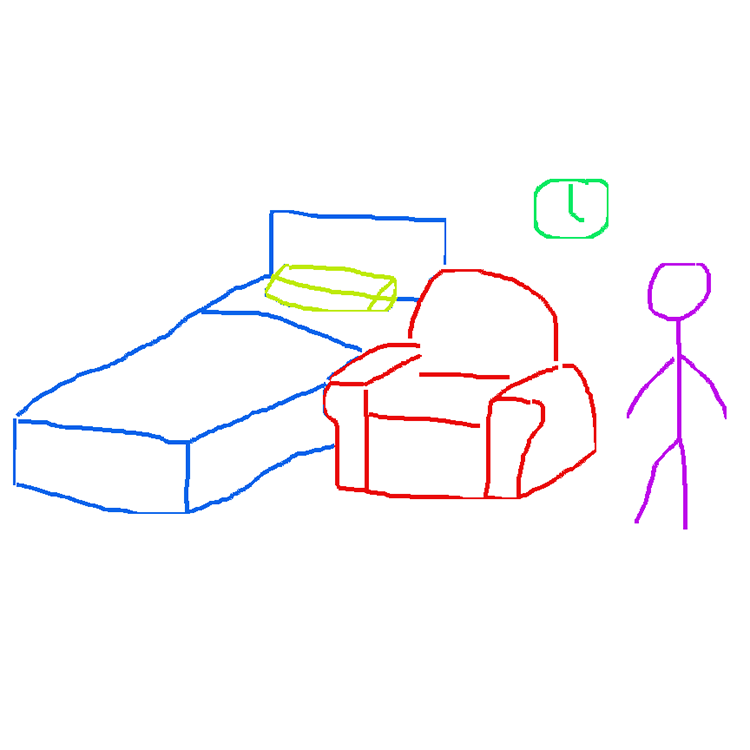}} &
        \frame{\includegraphics[trim=0 100 0 50, clip, width=0.3\linewidth]{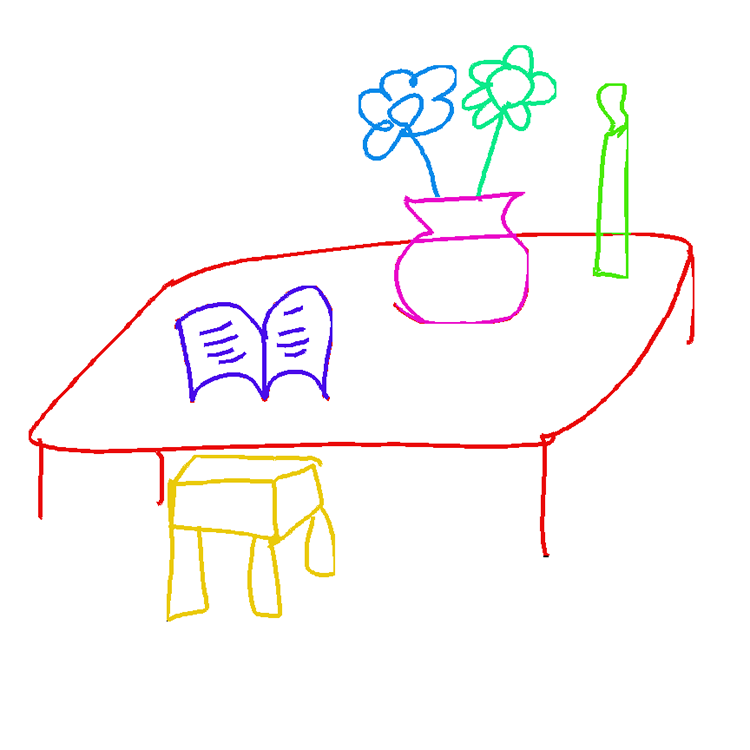}} \\
    \end{tabular}
    }}
    \vspace{-0.2cm}
    \caption{\textbf{\methodname{} results on artist-drawn and freehand sketches. }Our method accurately segments artist-drawn sketches in the first three rows and effectively handles quick, novice freehand sketches from the Zhang \etal \shortcite{zhang2018CBSC} and FSCOCO-Seg \shortcite{bourouis2024open} datasets in the bottom two rows.}
    \vspace{-2mm}
    \label{fig:qualitative_artists}
\end{figure}

\begin{figure*}
    \centering
    \setlength{\tabcolsep}{2pt}
    \renewcommand{\arraystretch}{2}
    % \addtolength{\belowcaptionskip}{-7.5pt}
    \resizebox{0.85\textwidth}{!}{%
    {\small
    \begin{tabular}{c c c c c c c}
        \frame{\includegraphics[width=0.13\linewidth]{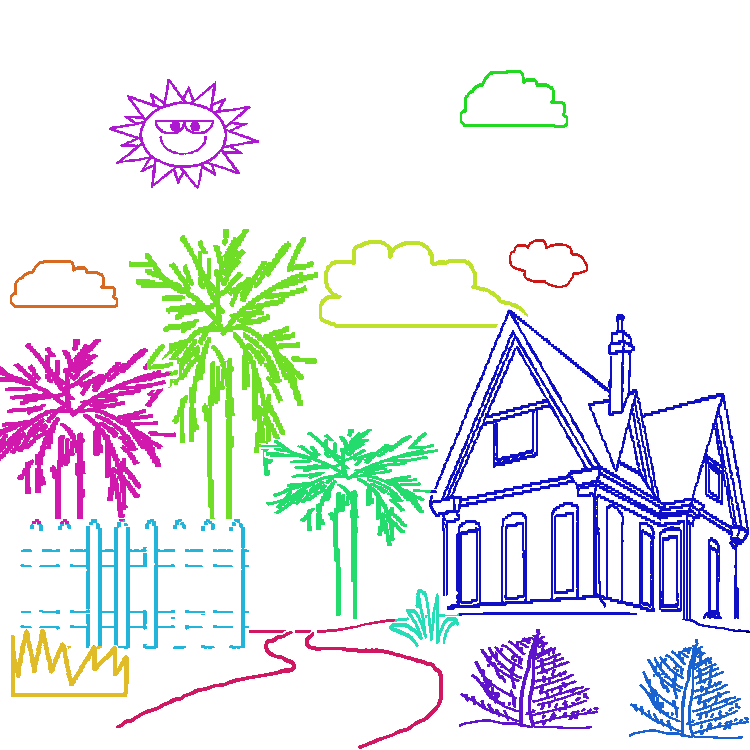}} &
        \frame{\includegraphics[width=0.13\linewidth]{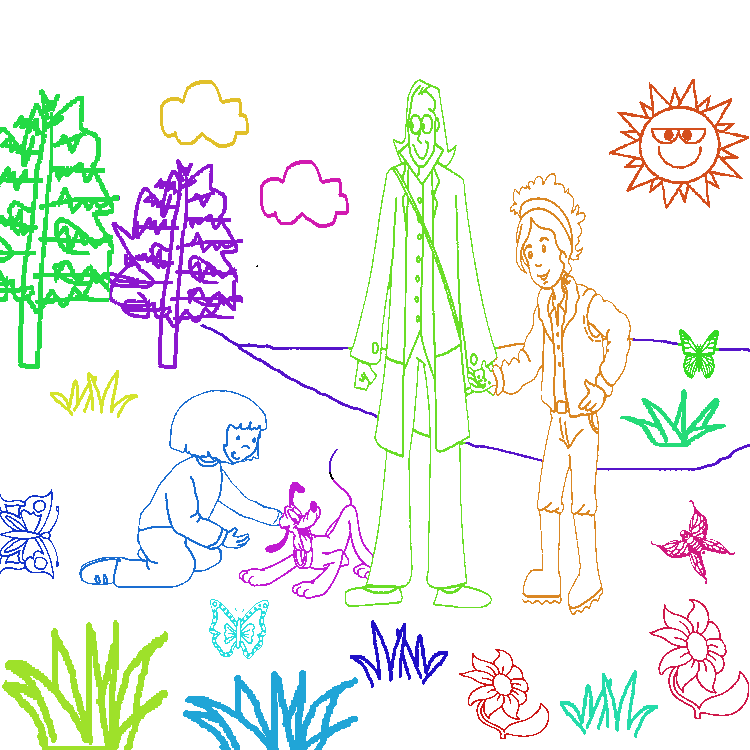}} &
        \frame{\includegraphics[width=0.13\linewidth]{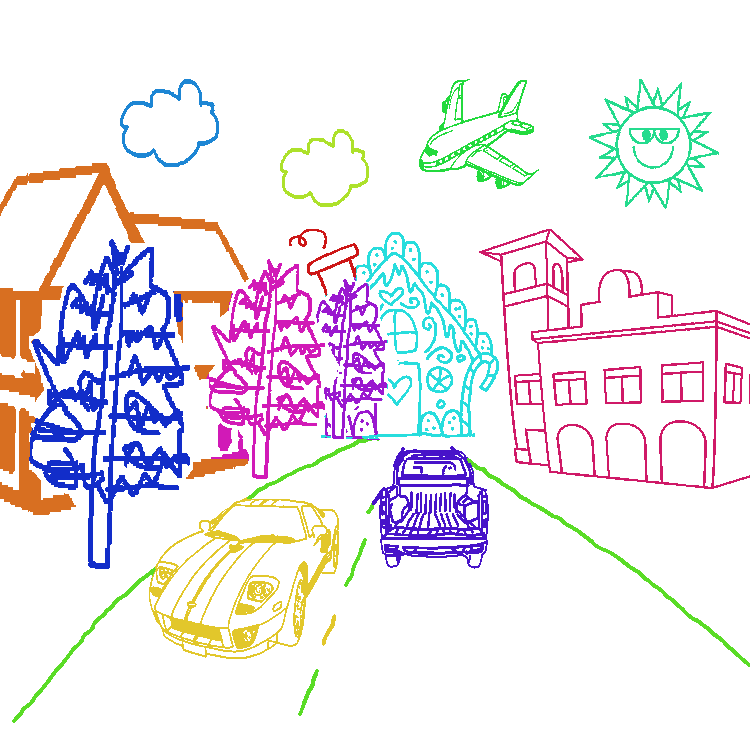}} &
        \frame{\includegraphics[width=0.13\linewidth]{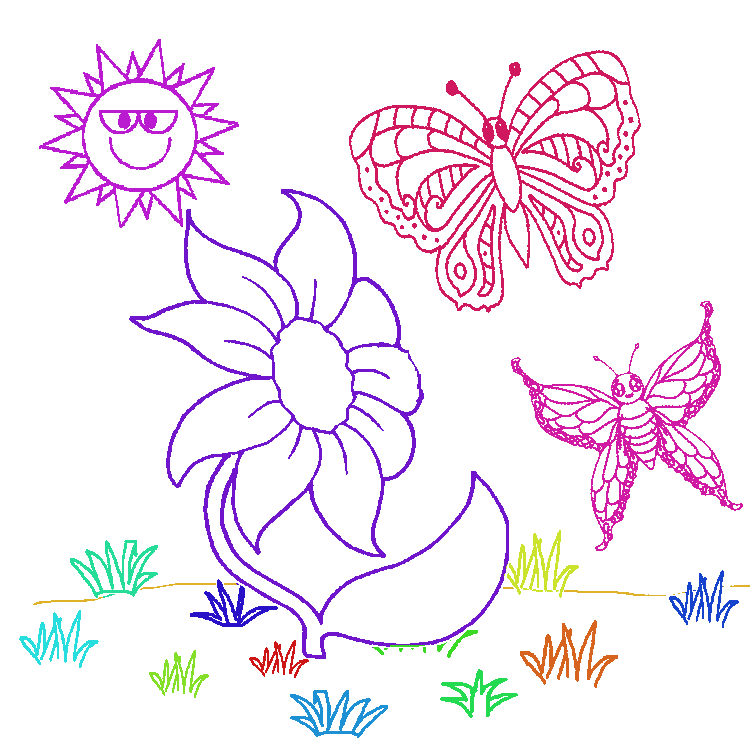}} &
        \frame{\includegraphics[width=0.13\linewidth]{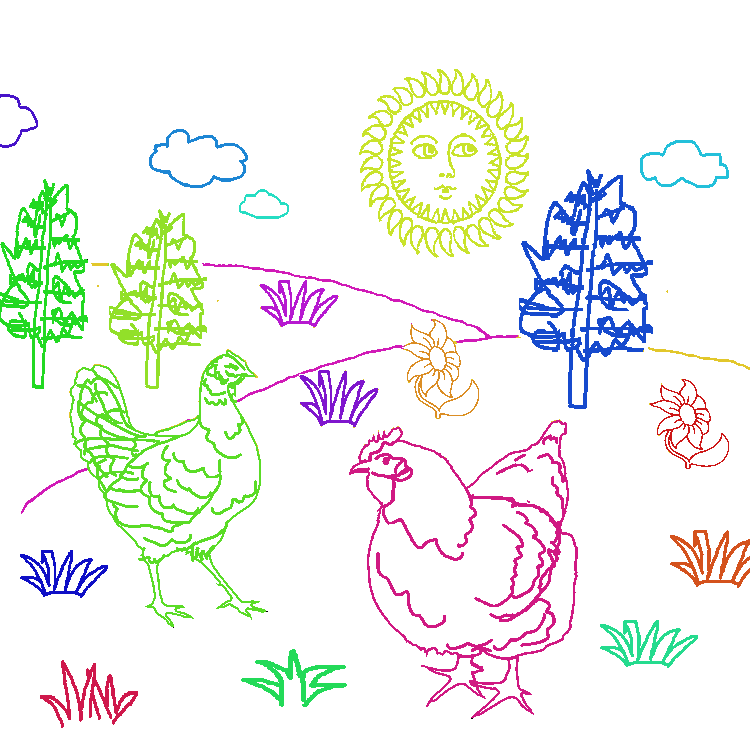}} &
        \frame{\includegraphics[width=0.13\linewidth]{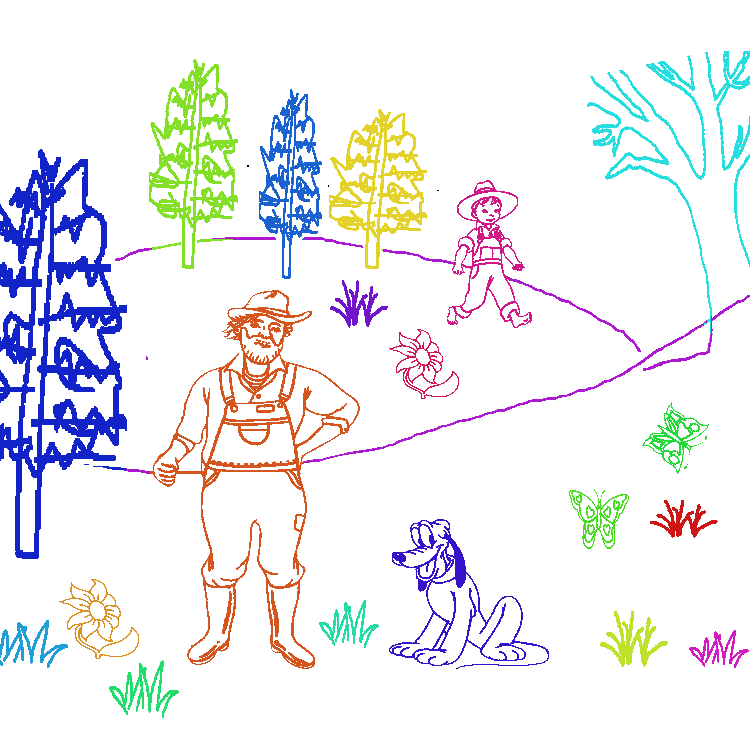}} &
        \frame{\includegraphics[width=0.13\linewidth]{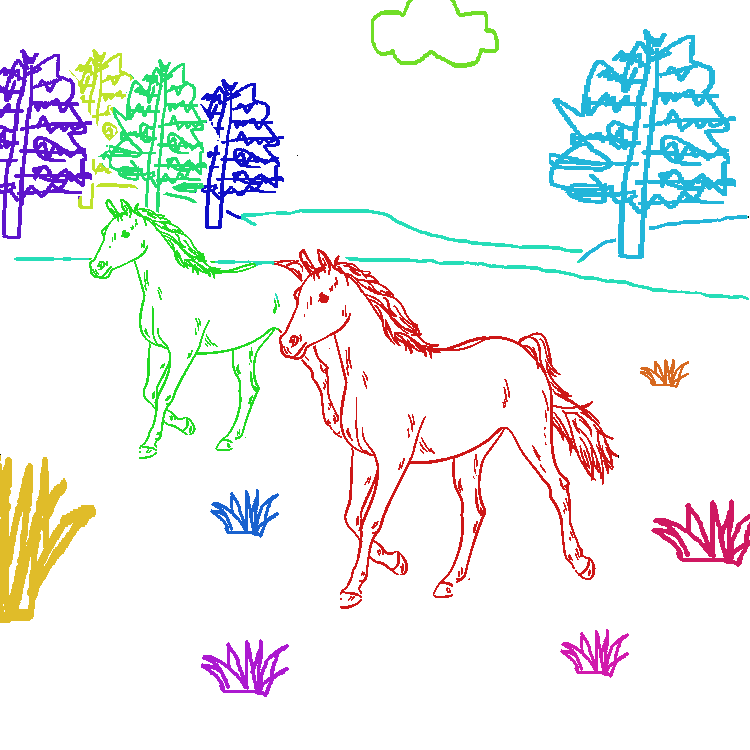}} \\

        \frame{\includegraphics[width=0.13\linewidth]{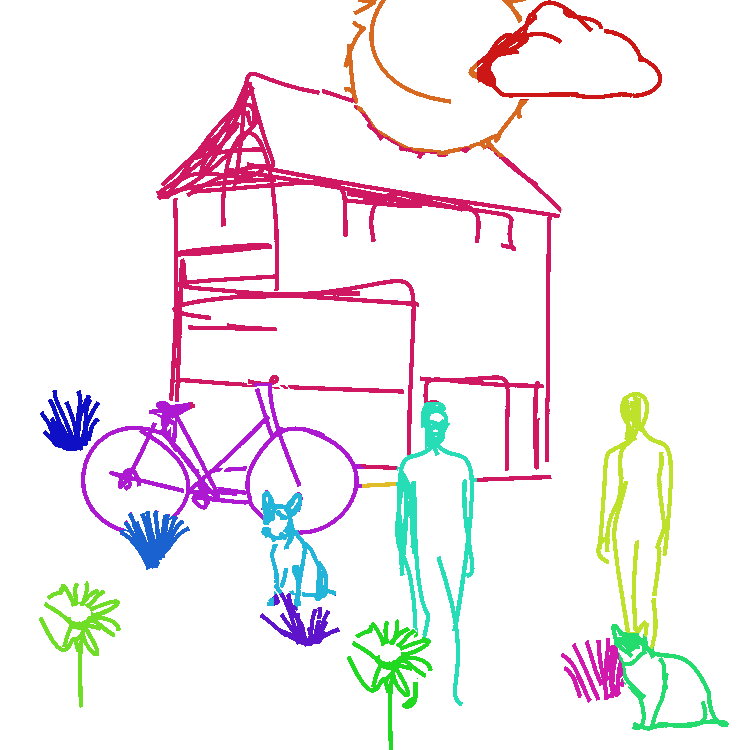}} &
        \frame{\includegraphics[width=0.13\linewidth]{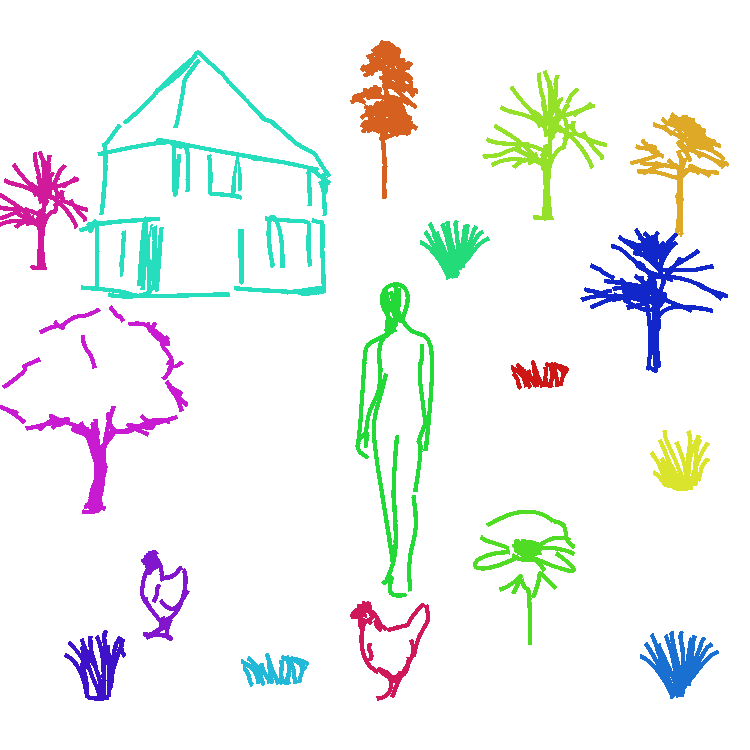}} &
        \frame{\includegraphics[width=0.13\linewidth]{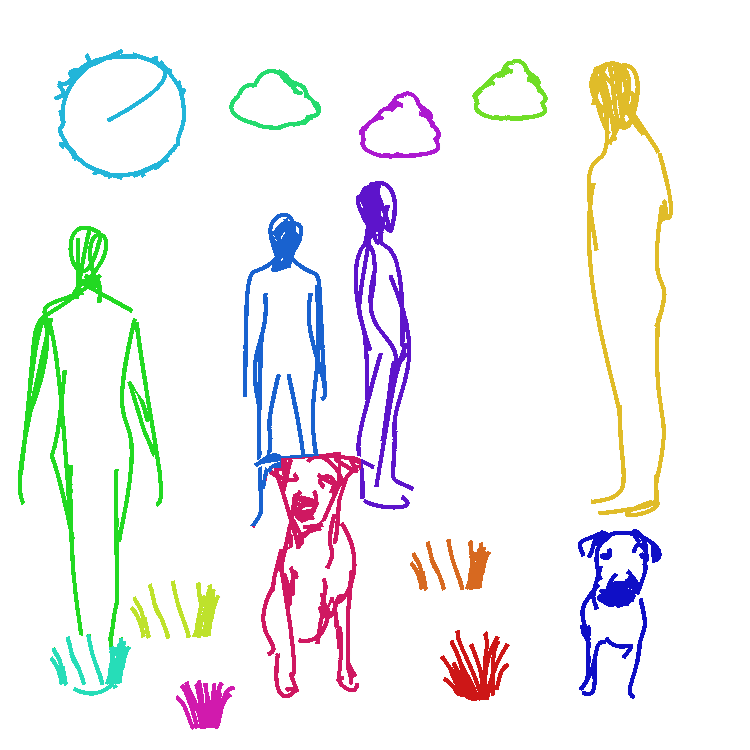}} &
        \frame{\includegraphics[width=0.13\linewidth]{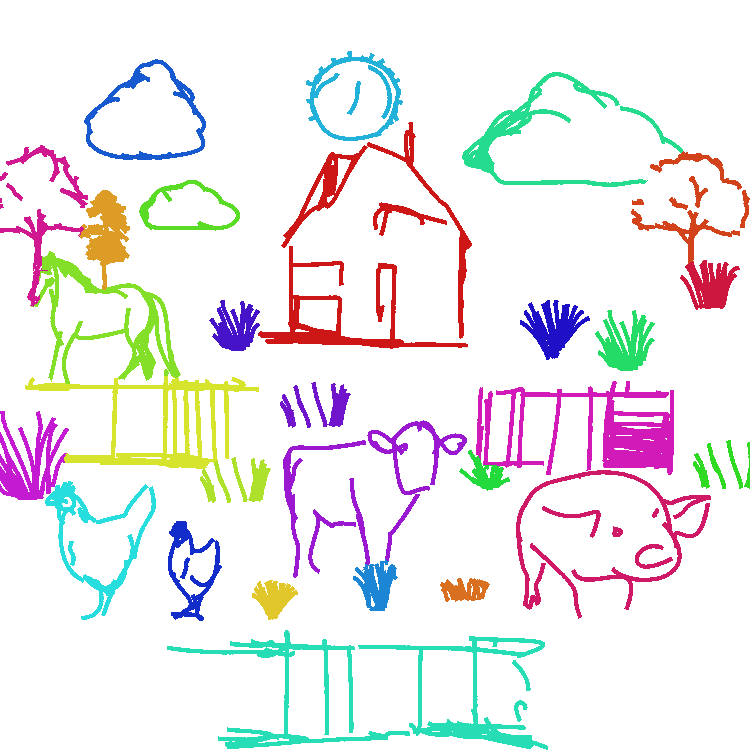}} &
        \frame{\includegraphics[width=0.13\linewidth]{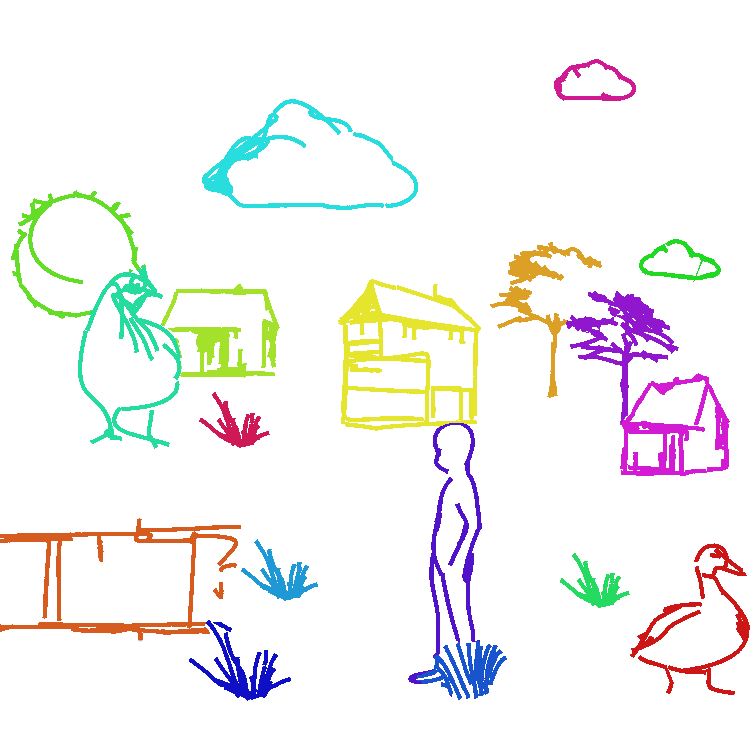}} &
        \frame{\includegraphics[width=0.13\linewidth]{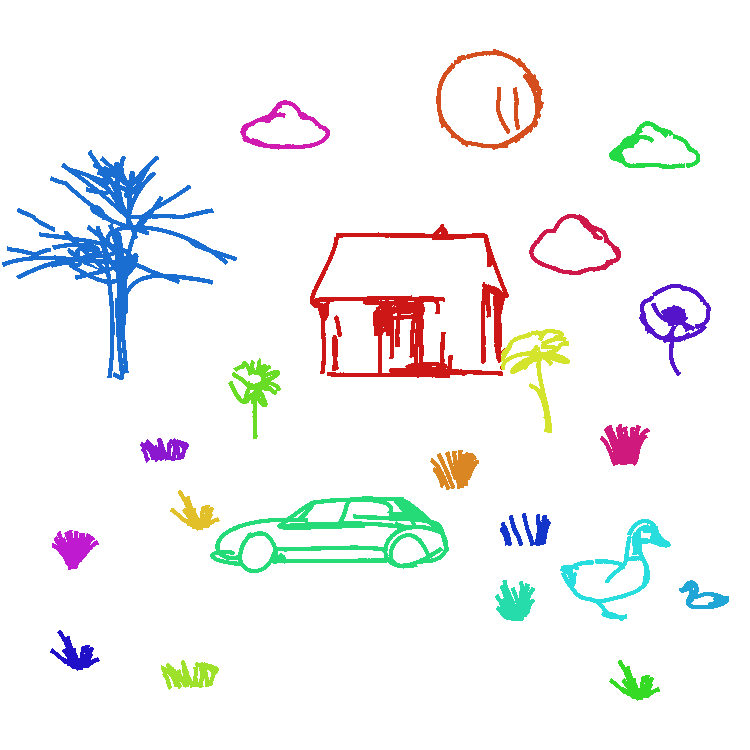}} &
        \frame{\includegraphics[width=0.13\linewidth]{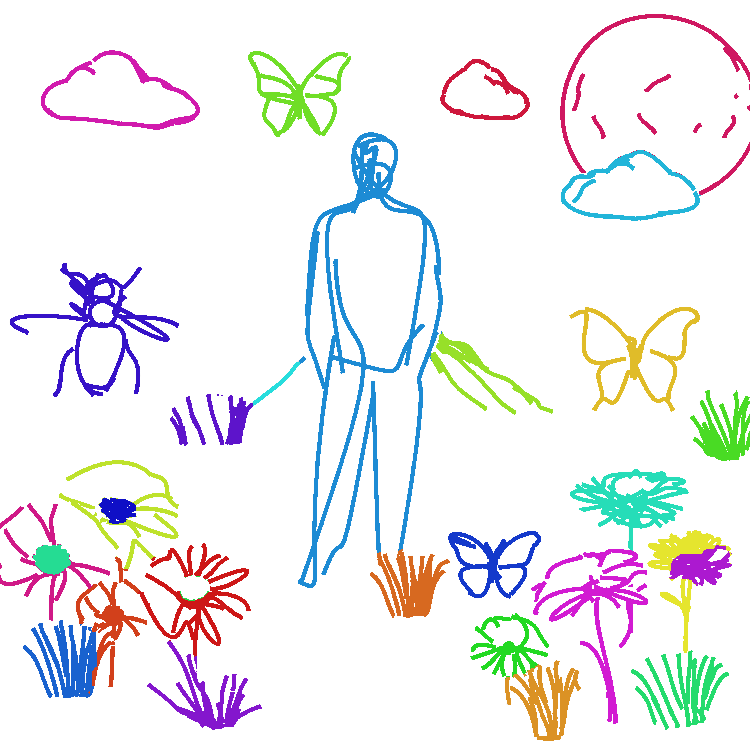}} \\
        
        \frame{\includegraphics[width=0.13\linewidth]{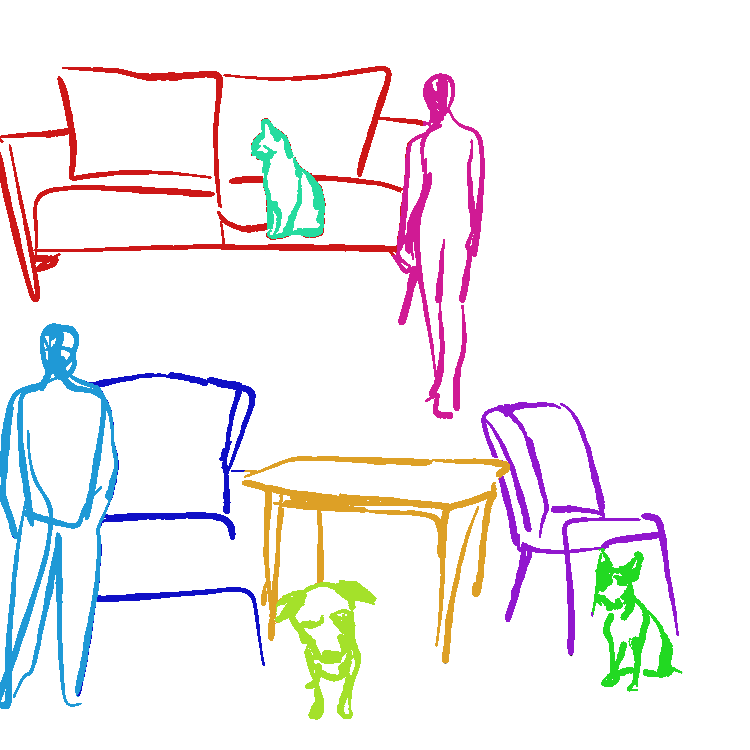}} &
        \frame{\includegraphics[width=0.13\linewidth]{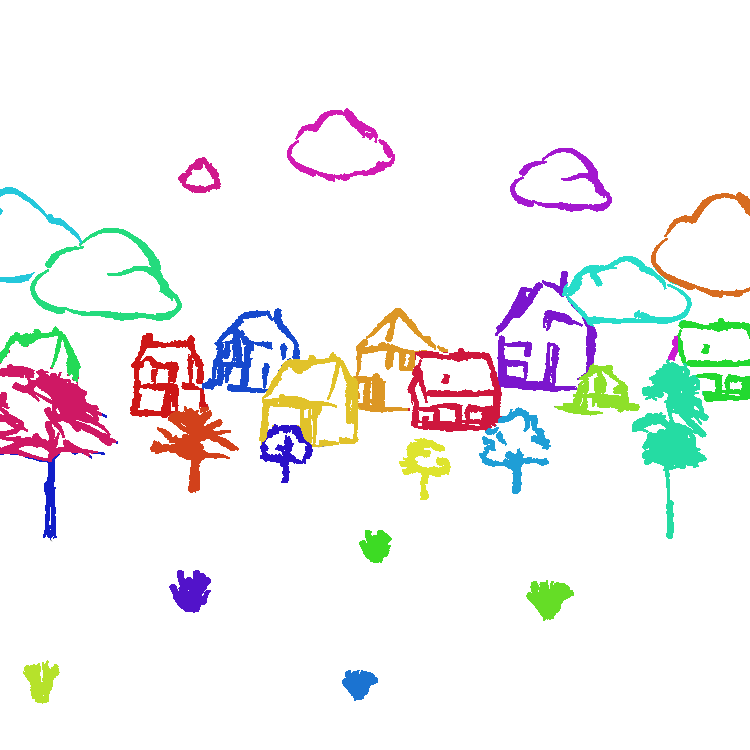}} &
        \frame{\includegraphics[width=0.13\linewidth]{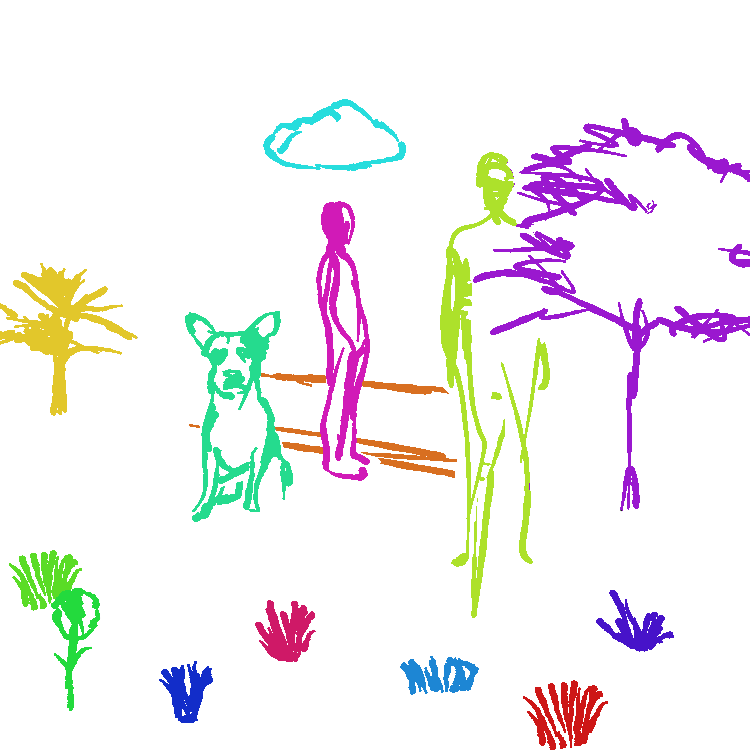}} &
        \frame{\includegraphics[width=0.13\linewidth]{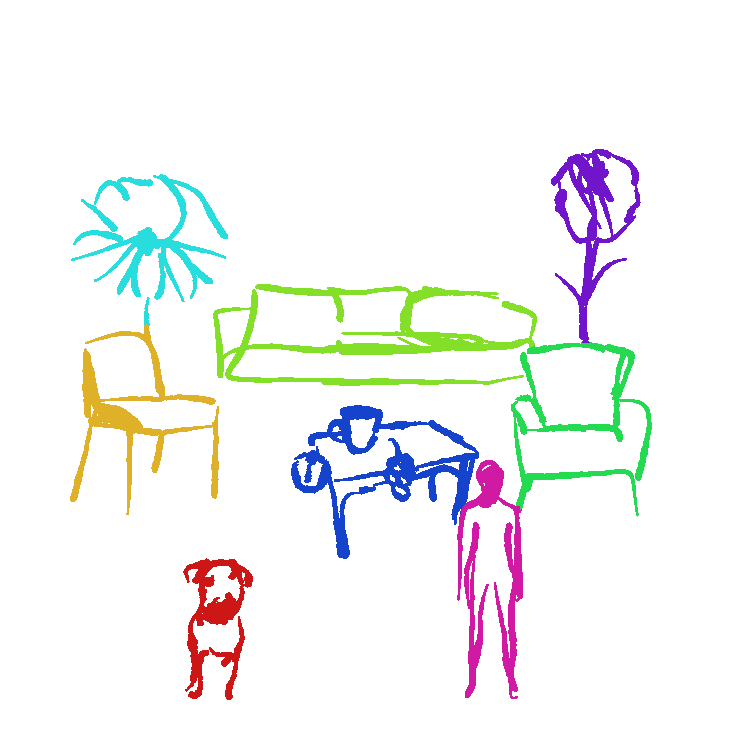}} &
        \frame{\includegraphics[width=0.13\linewidth]{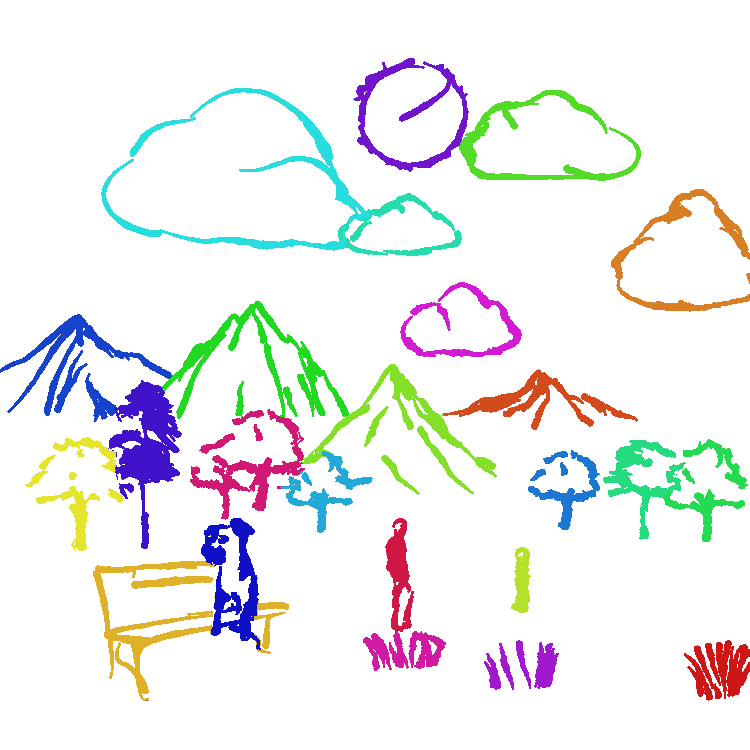}} &
        \frame{\includegraphics[width=0.13\linewidth]{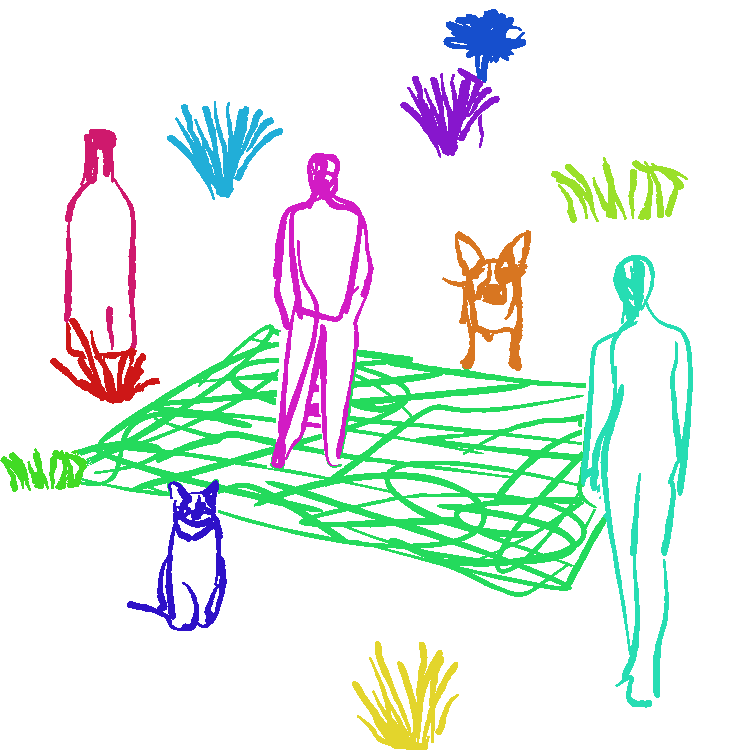}} &
        \frame{\includegraphics[width=0.13\linewidth]{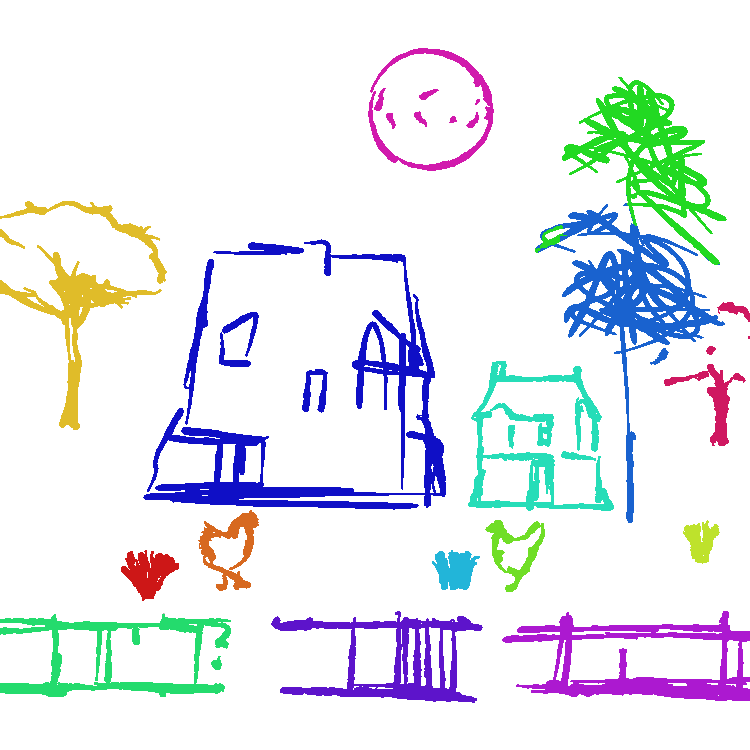}} \\

        \frame{\includegraphics[width=0.13\linewidth]{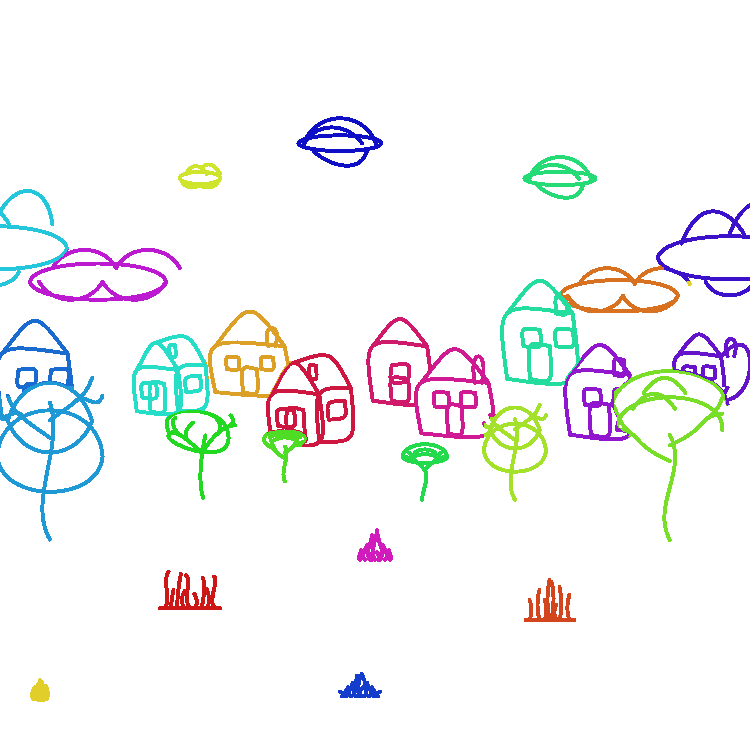}} &
        \frame{\includegraphics[width=0.13\linewidth]{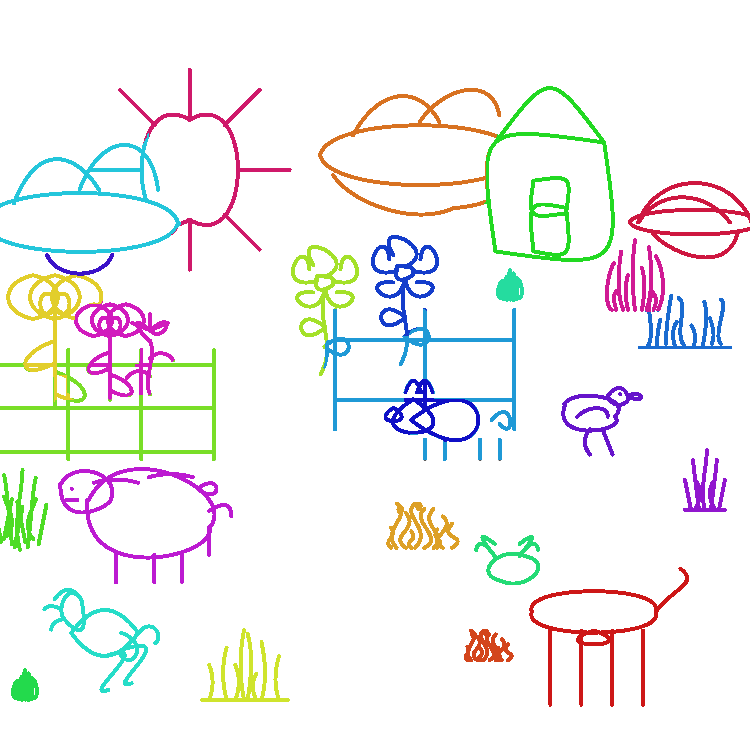}} &
        \frame{\includegraphics[width=0.13\linewidth]{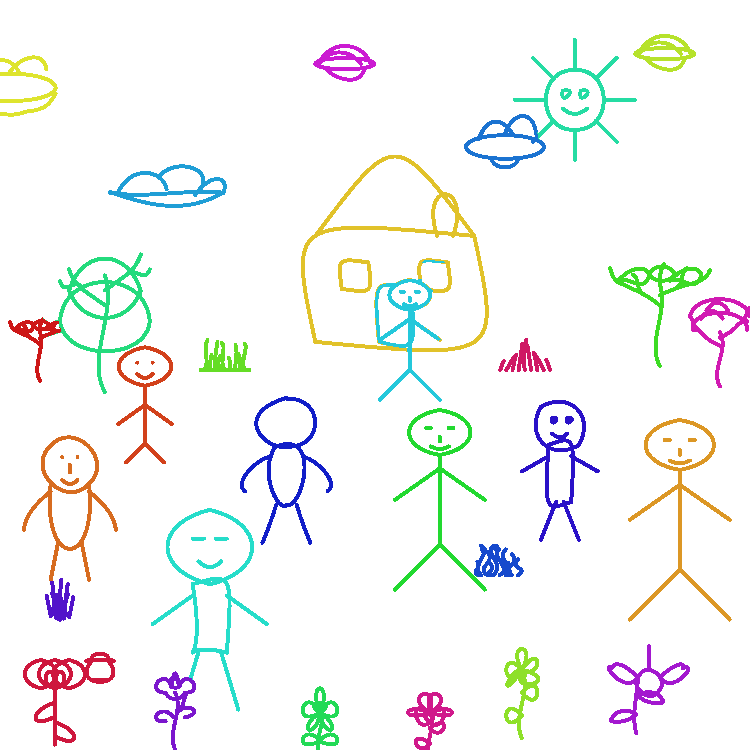}} &
        \frame{\includegraphics[width=0.13\linewidth]{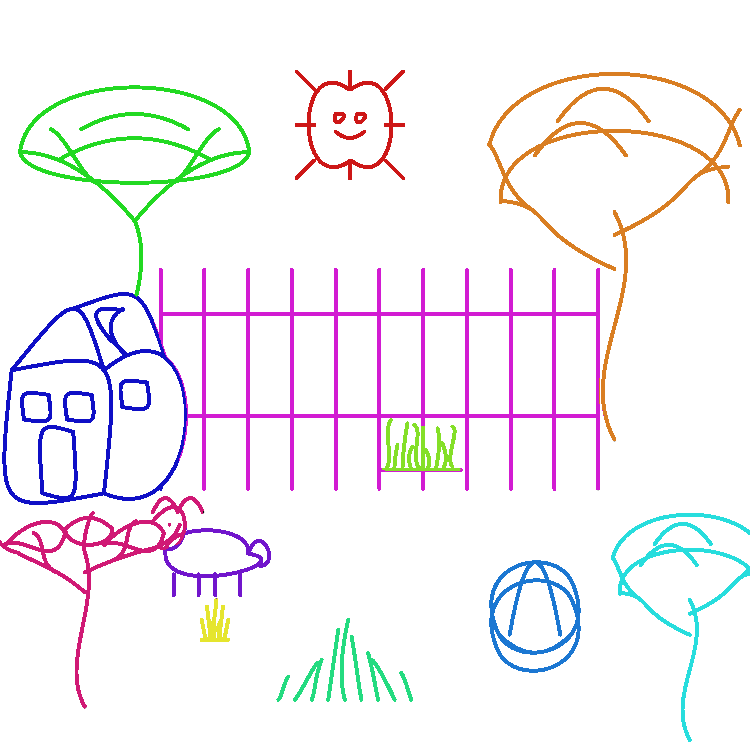}} &
        \frame{\includegraphics[width=0.13\linewidth]{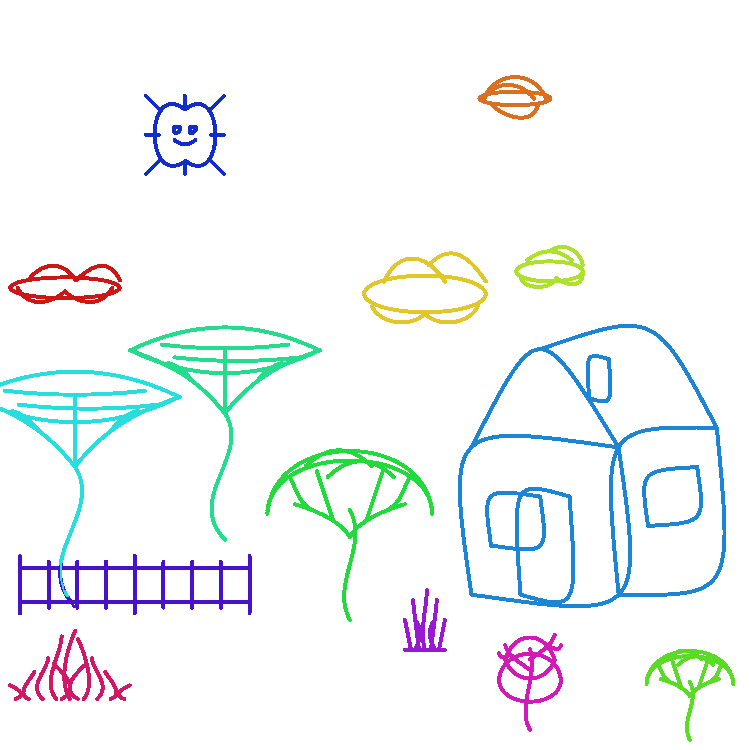}} &
        \frame{\includegraphics[width=0.13\linewidth]{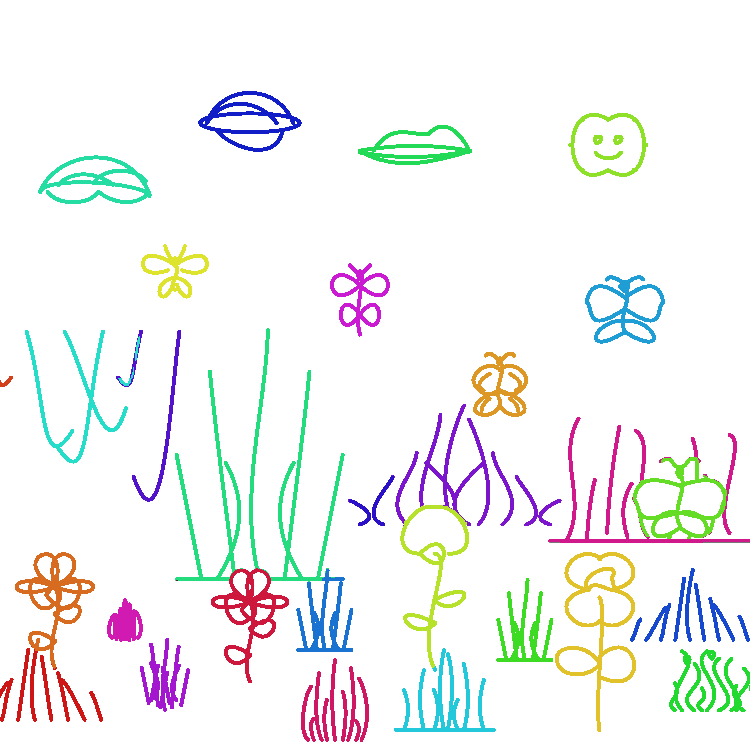}} &
        \frame{\includegraphics[width=0.13\linewidth]{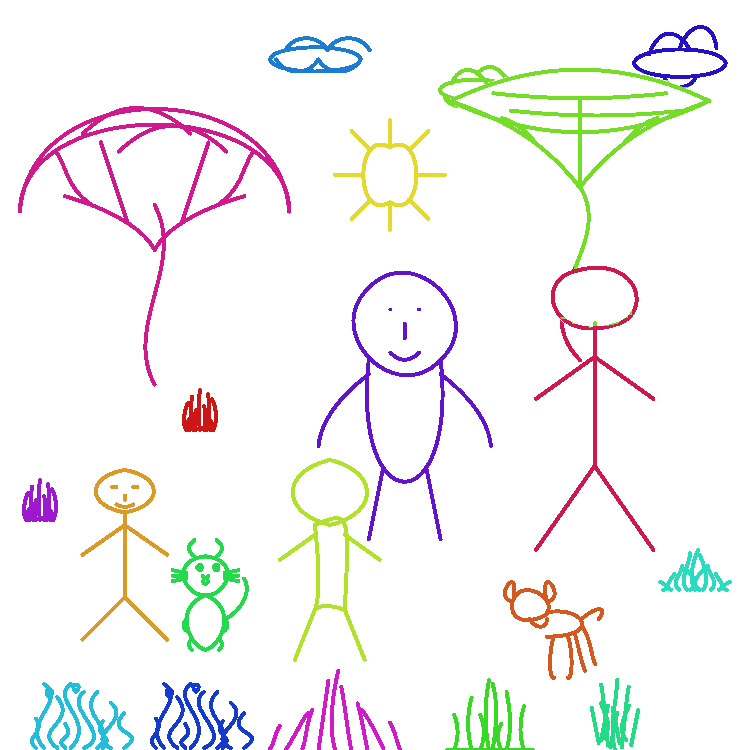}} \\

        \frame{\includegraphics[width=0.13\linewidth]{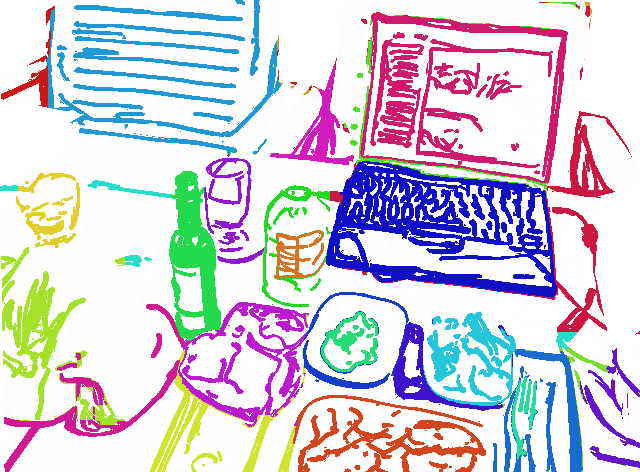}} &
        \frame{\includegraphics[width=0.13\linewidth]{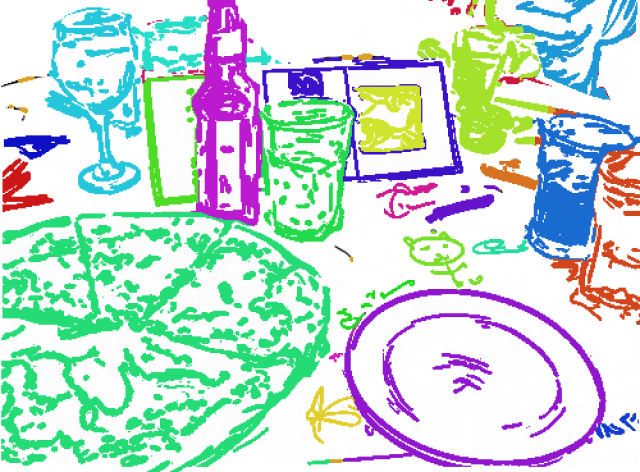}} &
        \frame{\includegraphics[width=0.13\linewidth]{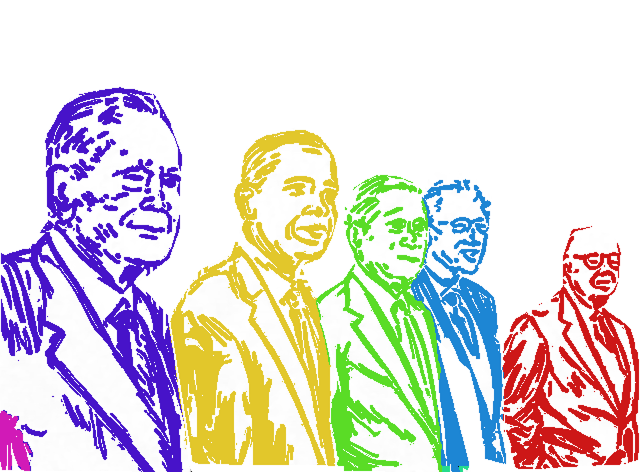}} &
        \frame{\includegraphics[width=0.13\linewidth]{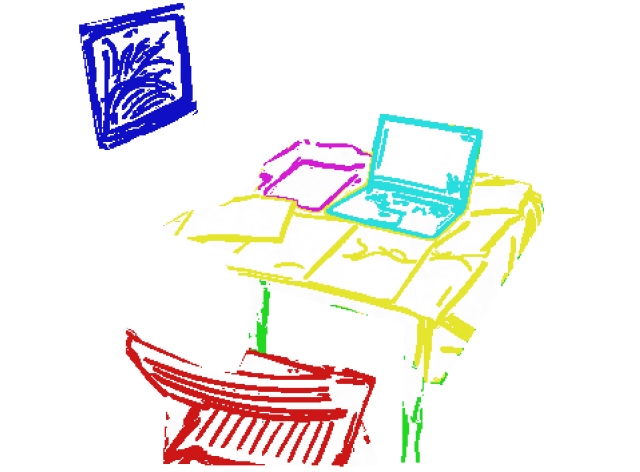}} &
        \frame{\includegraphics[width=0.13\linewidth]{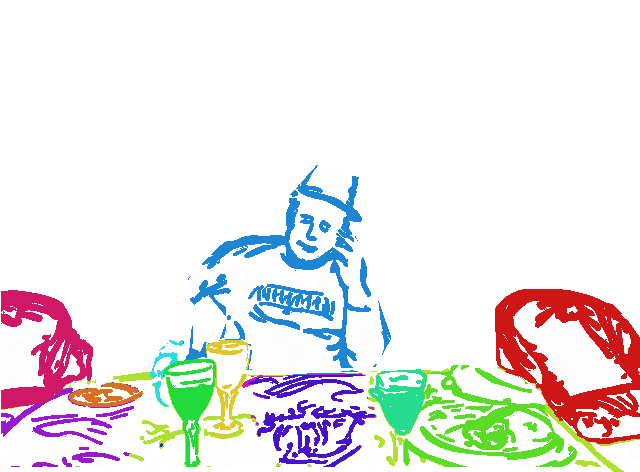}} &
        \frame{\includegraphics[width=0.13\linewidth]{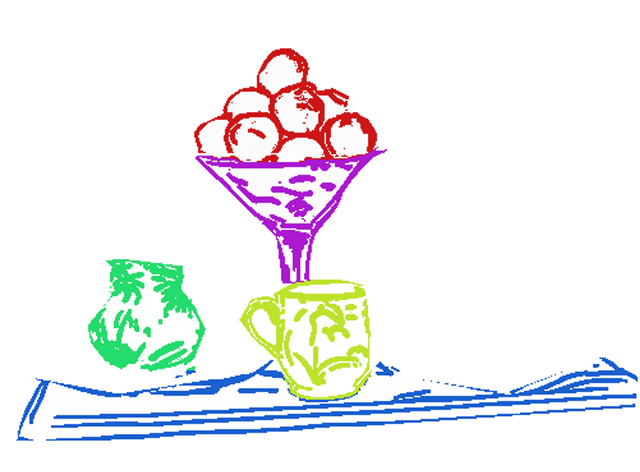}} &
        \frame{\includegraphics[width=0.13\linewidth]{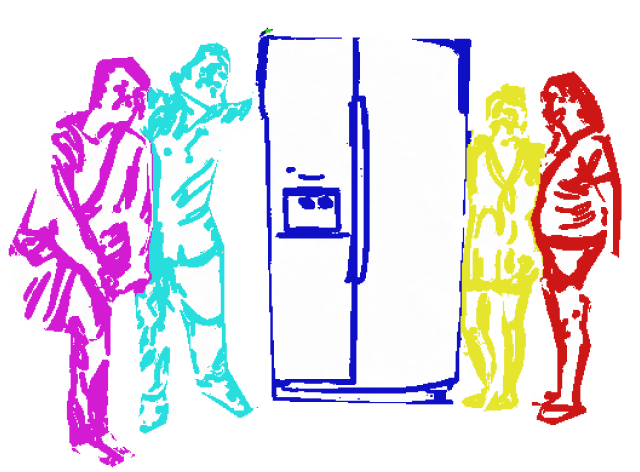}} \\
        
    \\     
    \end{tabular}
    }
    }
    \vspace{-10mm}
    \caption{\textbf{\methodname{} segmentation results on synthetic sketches.}  Our method is able to precisely segment SketchyScene dataset (first row) and our \datasetname{} dataset: CLIPasso variants (second and third rows), abstract SketchAgent sketches (fourth row), and detailed InstantStyle sketches (last row).}
    \label{fig:qualitative3}
\end{figure*}

\section{Limitations and Future Work}
\begin{figure}
    \vspace{-4mm} 
    \includegraphics[width=.94\linewidth]{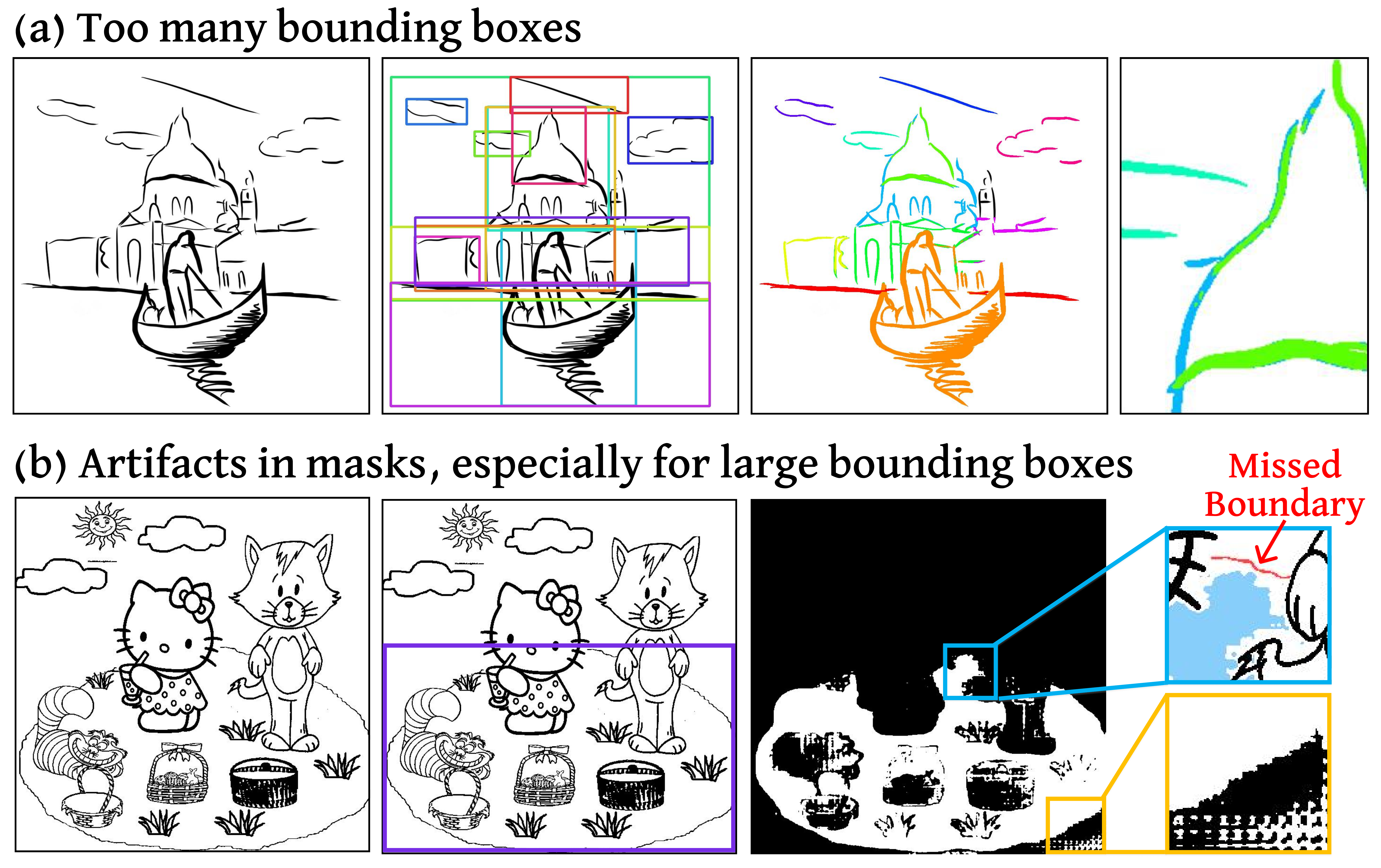}
    \vspace{-3mm} 
    \caption{\textbf{Examples showcasing \methodname{}'s limitations.} (a) The bounding box filtering process can still retain undesired boxes, leading to artifacts in final segmentation. \rev{(b) SAM masks are generally reasonable but can produce artifacts such as missing object boundaries or producing grid-like regions.}}
    \label{fig:limitations}
    \vspace{-4mm}
\end{figure}

While our method successfully segments scene sketches across various styles and challenging cases, it has some limitations.
First, our bounding box filtering technique may still include undesired boxes, potentially introducing artifacts when combining masks into the final segmentation (\Cref{fig:limitations}a).
Second, our mask generation relies on SAM, which generally produces good masks but can occasionally introduce artifacts, particularly for objects occupying large regions in the sketch (\Cref{fig:limitations}b). \rev{In particular, it may miss sketch object boundaries or produce grid-like masks.} Even after applying our refinement stage, some artifacts may persist in the final segmentation. Future work could address this issue by fine-tuning SAM specifically for sketches or incorporating a learned refinement stage. 
% Lastly, oveerlapping masks still pose a challenge, while out depth-based approach aim to choose the, depth cues may not always provide a reliable signal for resolving ambiguous pixels in overlapping masks, leading to artifacts, especially along shared contours between overlapping objects.

% \maneesh{These limitations are largely aspects of the algorithm that could be improved by tweaking. I think it would be stronger to list real limitations -- are their assumptions or types of sketches that we know won't work with our approach?}

\section{Conclusions}
We introduced \methodname{}, a method for instance segmentation of raster scene sketches. Our approach adapts Grounding DINO, an object detection model trained on natural images, to the sketch domain through class-agnostic fine-tuning. We utilized Segment Anything (SAM) for segmentation along with a refinement stage that incorporates depth cues to resolve ambiguous pixels.
Our method significantly improves upon state-of-the-art approaches in this domain, demonstrating the utility of natural image priors for sketch understanding tasks. We additionally provide a synthetic scene-level annotated sketch dataset, \datasetname{}, encompassing a wide range of object categories and significant variations in drawing styles. Our experiments demonstrate that \methodname{} is robust to these variations, achieving consistent performance across diverse datasets.

\begin{acks}
This work was partially supported by the Brown Institute for Media Innovation at Stanford. Yael Vinker was supported in part by IBM Agreement No. W1771646 and Hyundai Motor Company R\&D Center Agmt Dtd 2/22/23.
\end{acks}

% The dataset will be made publicly available to support further research and advancements in sketch understanding.

% \maneesh{Need to add acknowledgements -- This work was partially supported by the Brown Institute for Media Innovation at Stanford. -- Anything else? Yael may have some funding sources that should be thanked.}

\clearpage
\bibliographystyle{ACM-Reference-Format}
\bibliography{main}

% ===== End of main paper =====

% ===== Start TOC only for supp =====
\clearpage
\appendix

% Real TOC begins here
% \clearpage
\twocolumn[{%
  \begin{center}
    % \vspace{2em}
    {\huge \bfseries Instance Segmentation of Scene Sketches Using Natural Image Priors\\[1ex]
    Supplementary Material}
    \vspace{2em}

    {\Large
      Mia Tang$^{1}$ \quad
      Yael Vinker$^{2}$ \quad
      Chuan Yan$^{1}$ \quad
      Lvmin Zhang$^{1}$ \quad
      Maneesh Agrawala$^{1}$\\[0.5em]
      $^1$Stanford University \quad
      $^2$MIT CSAIL\\[0.5em]
      % \texttt{miatang@cs.stanford.edu, yaelvi116@gmail.com, chuanyan@stanford.edu, lvmin@stanford.edu, maneesh@cs.stanford.edu}
    }

    % \vspace{2em}
    {\large Project page: \href{https://inklayer.github.io}{\texttt{\textcolor{magenta}{https://inklayer.github.io}}}}

    \vspace{2em}
  \end{center}
}]

\restoreTOC
\setcounter{tocdepth}{2}
\renewcommand{\contentsname}{Contents}
\tableofcontents
% \clearpage

\section{Sketch Editing Interface}
\begin{figure}[H]
    \centering
    \includegraphics[width=1\linewidth]{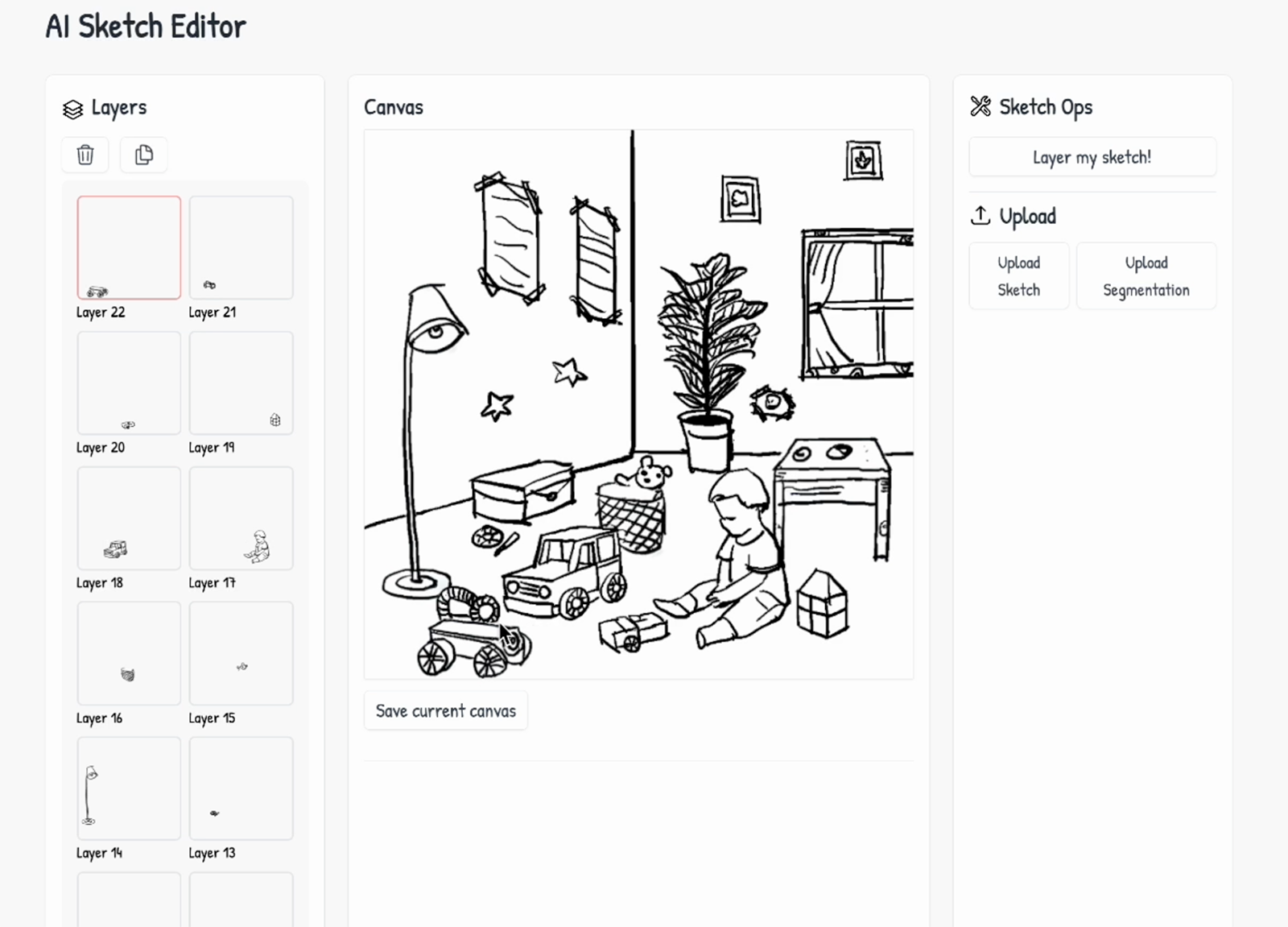}
    \caption{\textbf{InkLayer application. }Interactive interface for sketch editing, powered by our instance segmentation and layer completion algorithm.}
    \vspace{-5mm}
    \label{fig:interface}
\end{figure}

Our sketch segmentation and layering technique facilitates sketch editing, allowing users to drag or manipulate segmented objects without the need to manually sketch the affected regions. We demonstrate this through an interactive sketch editing interface (\Cref{fig:interface}) that enables users to upload a sketch, which is then segmented and transformed into completed, ordered layers as detailed in our paper. This facilitates more efficient sketch editing by allowing artists to easily move, copy, or delete pixels associated with specific object instances, as the sketch is represented as an ordered list of layers.

% We offer an interactive sketch editing interface that enables users to upload a sketch, which is then segmented and transformed into completed, ordered layers as detailed in our paper. This facilitates more efficient sketch editing by allowing artists to easily move, copy, or delete pixels associated with specific object instances, as the sketch is represented as an ordered list of layers.

\section{Refinement Module Ablation}

\begin{figure*}
    \centering
    \setlength{\tabcolsep}{4pt}
    {\small
    \resizebox{0.85\textwidth}{!}{ 
    \begin{tabular}{c c c}
        Sketch & w/o Refinement & \textbf{w/ Refinement (Ours)}  \\
        \frame{\includegraphics[width=0.29\linewidth]{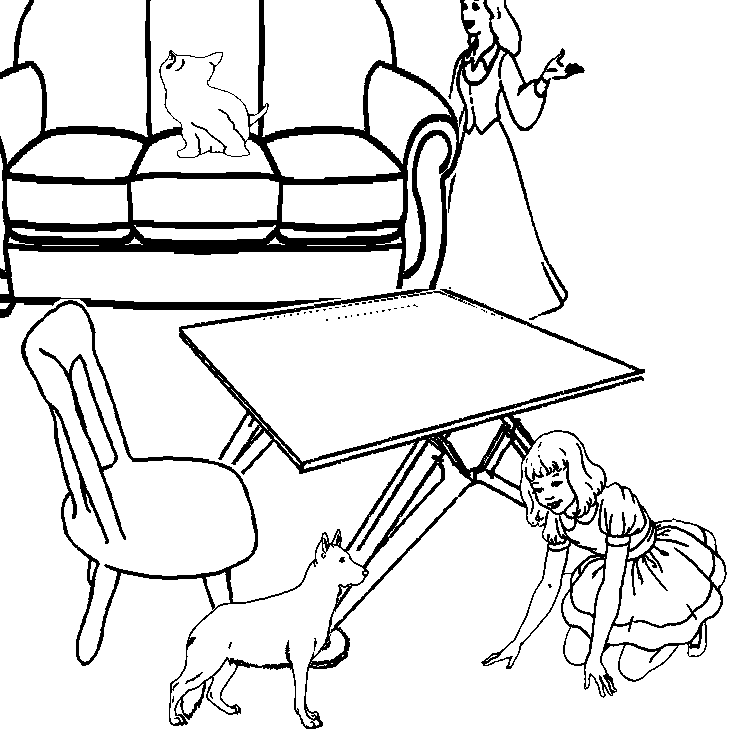}} &
        \frame{\includegraphics[width=0.29\linewidth]{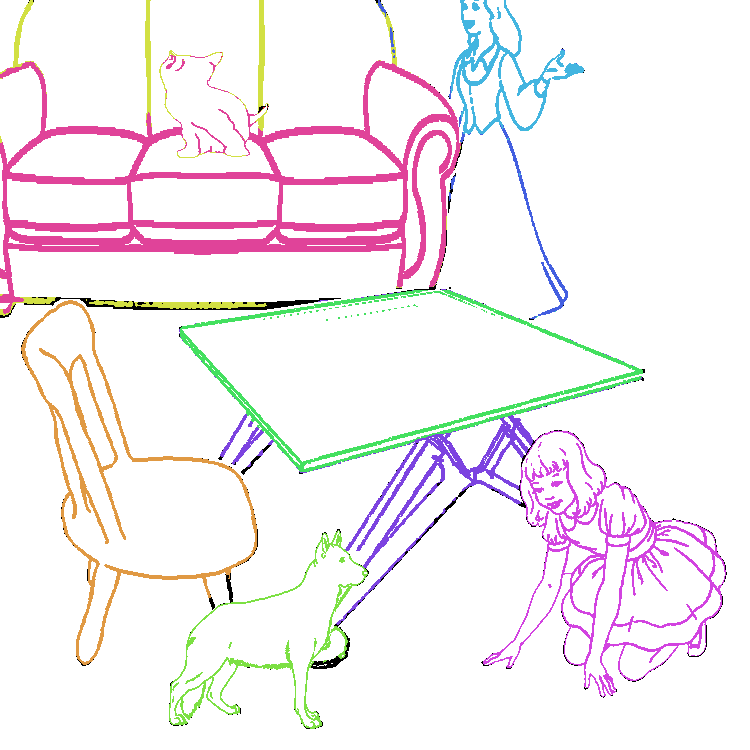}} &
        \frame{\includegraphics[width=0.29\linewidth]{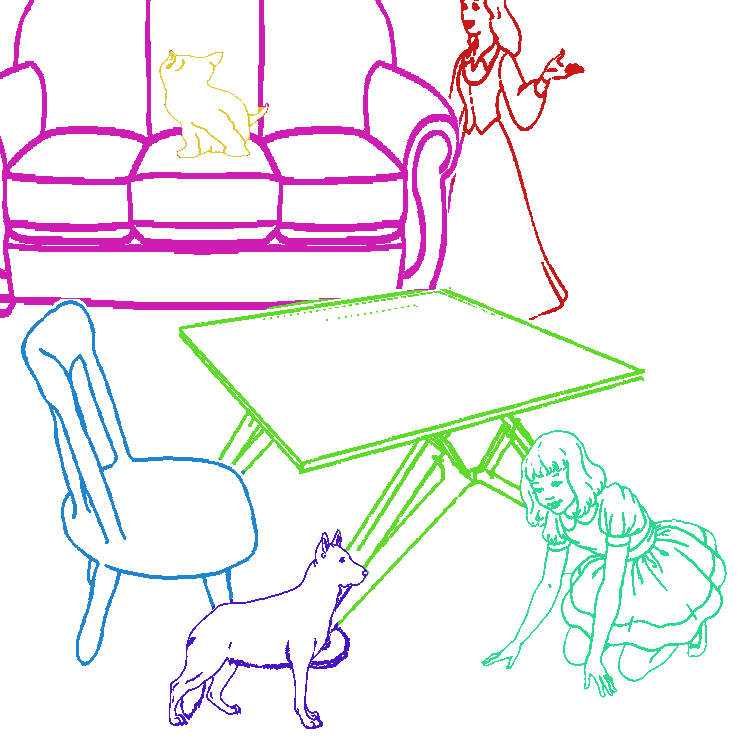}} \\

        \frame{\includegraphics[width=0.29\linewidth]{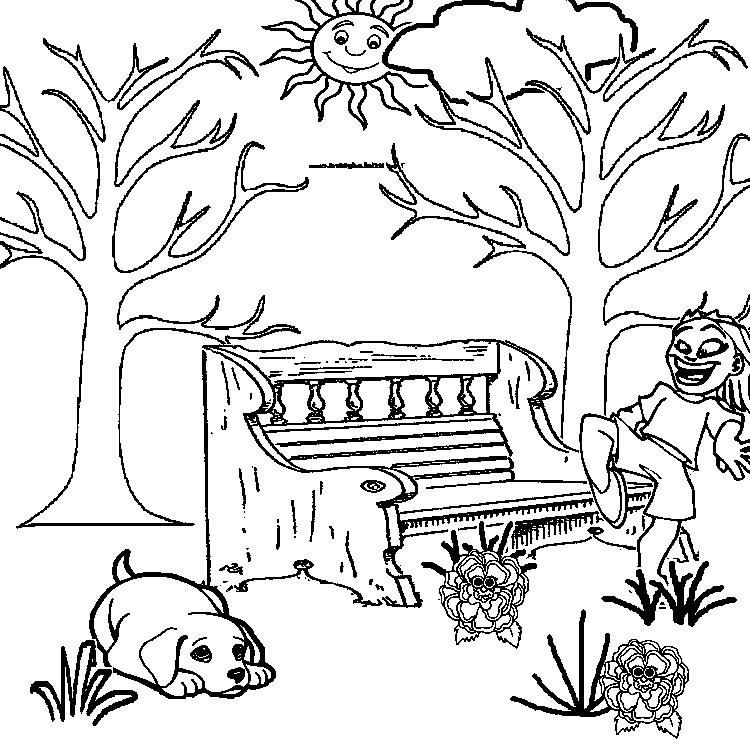}} &
        \frame{\includegraphics[width=0.29\linewidth]{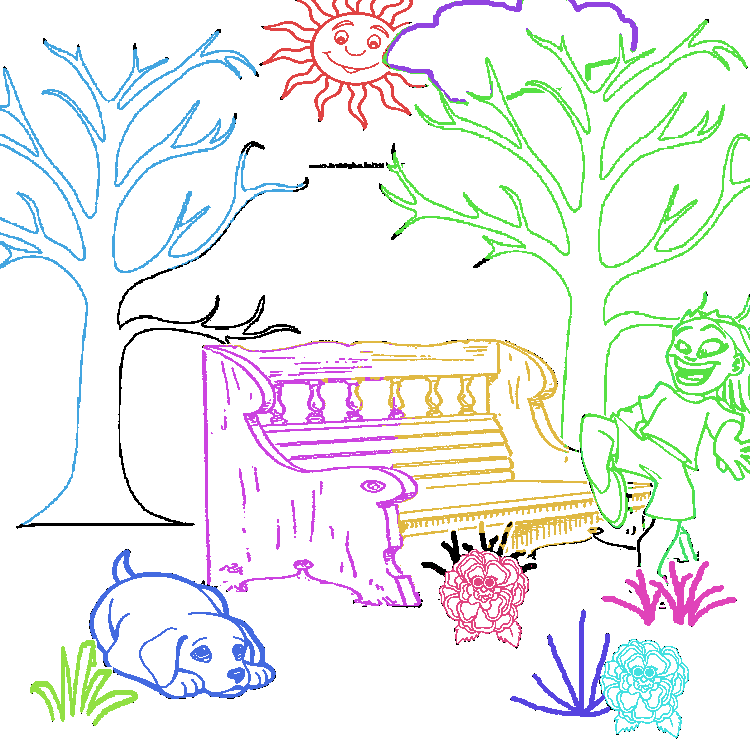}}  &
        \frame{\includegraphics[width=0.29\linewidth]{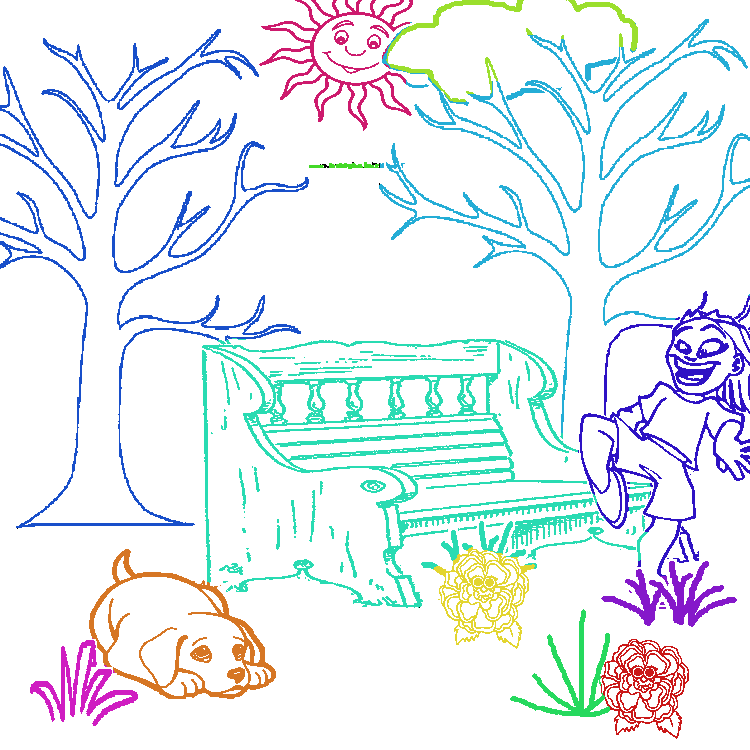}} \\

        \frame{\includegraphics[width=0.29\linewidth]{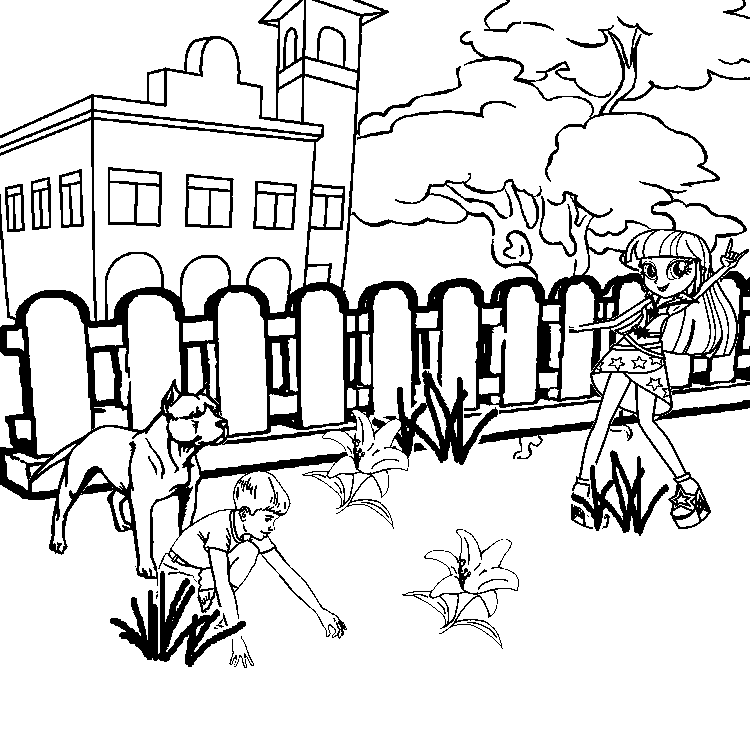}} &
        \frame{\includegraphics[width=0.29\linewidth]{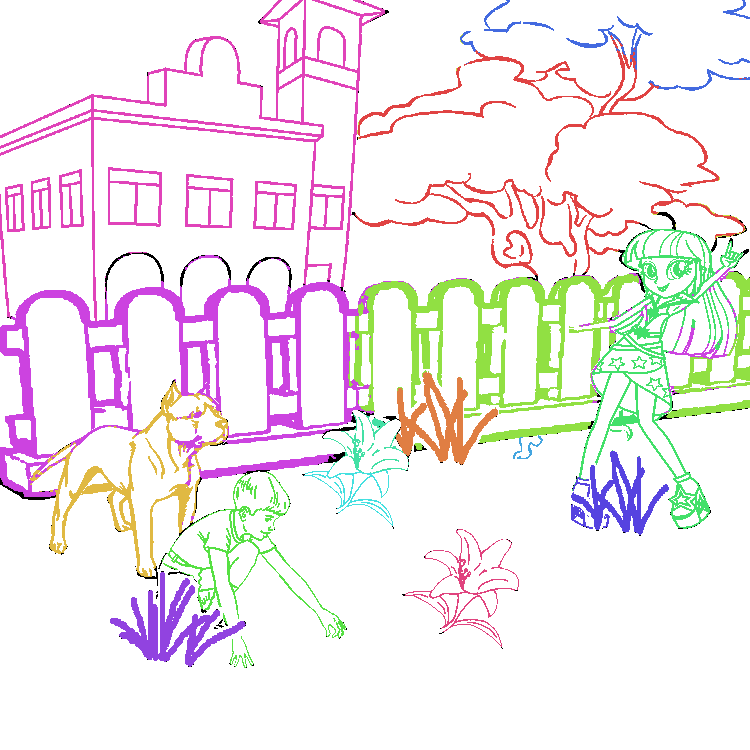}}  &
        \frame{\includegraphics[width=0.29\linewidth]{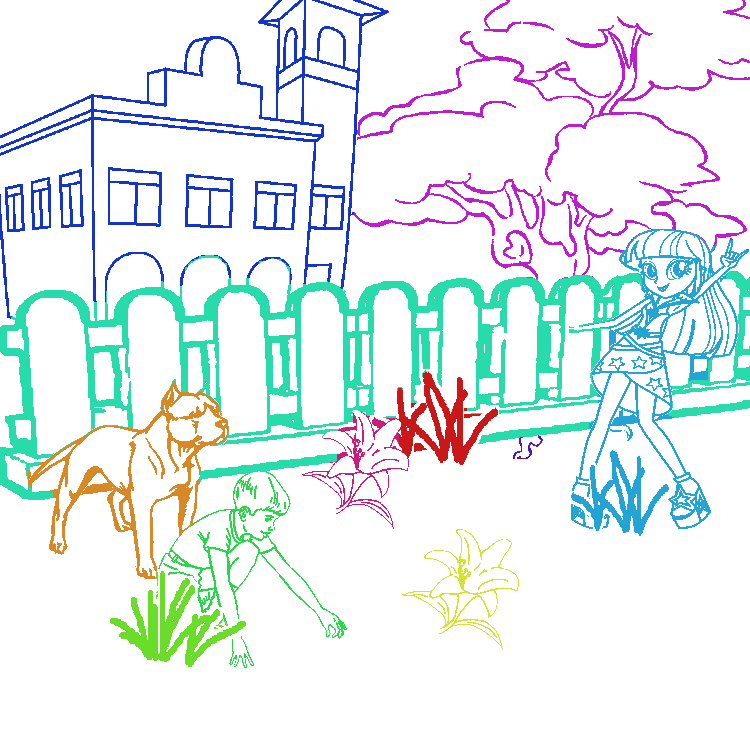}} 
        \\

        \frame{\includegraphics[width=0.29\linewidth]{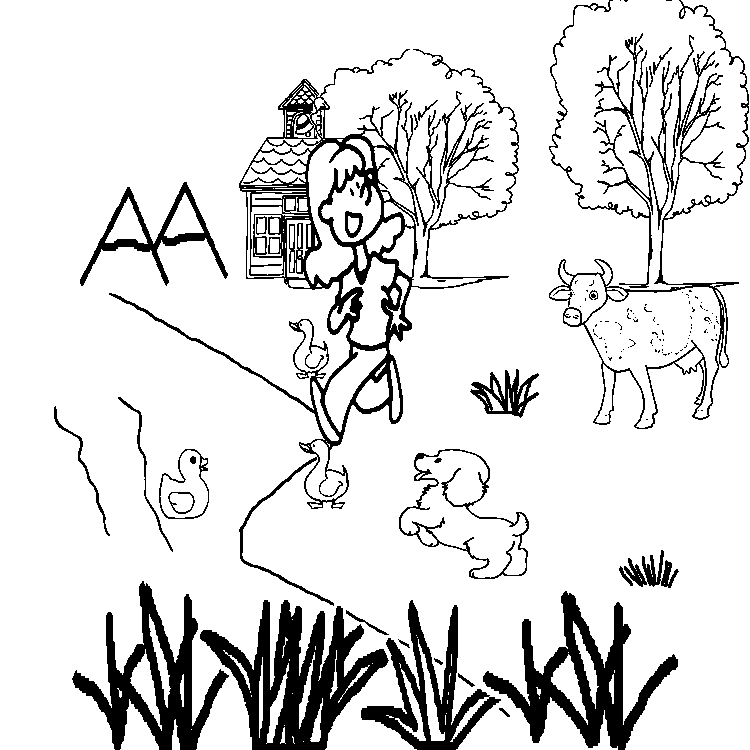}} &
        \frame{\includegraphics[width=0.29\linewidth]{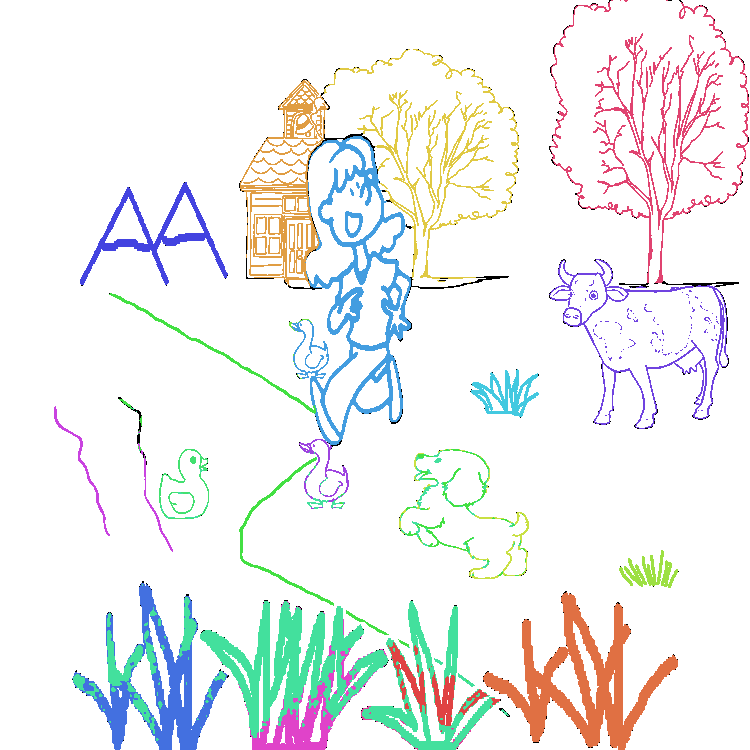}} &
        \frame{\includegraphics[width=0.29\linewidth]{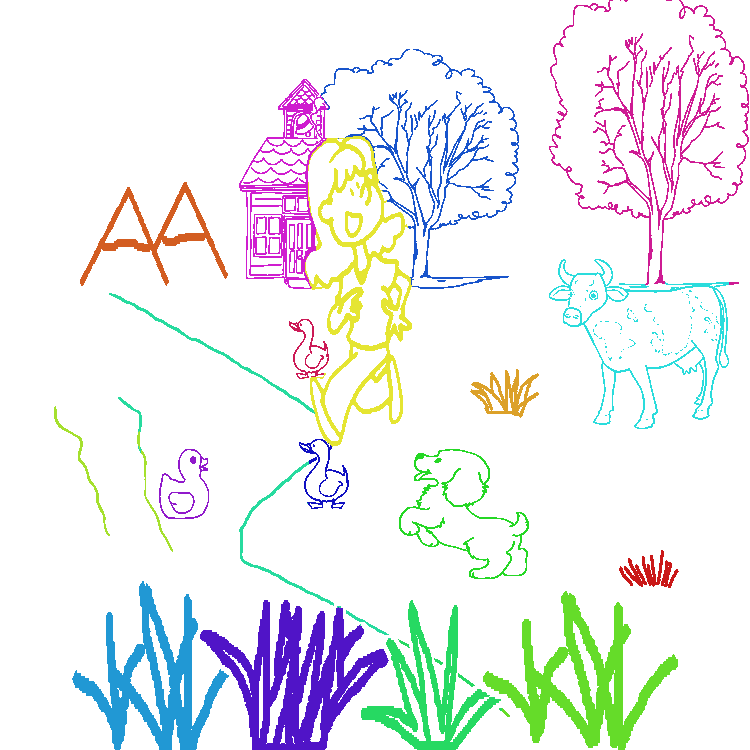}} 
        \\
    \end{tabular}
    }}
    \caption{\rev{\textbf{Refinement module ablation.} We compare segmentation outputs with and without our depth-guided refinement module. The refinement significantly improves mask coherence and reduces visual artifacts, such as duplicated or fragmented object regions. As shown across diverse scene sketches, the refined results exhibit smoother boundaries, better alignment with object structure, and fewer inconsistencies—particularly in complex regions with overlapping or adjacent objects.}}
    \label{fig:refinement_ablation}
\end{figure*}

\begin{table}[]
\centering
\begin{tabular}{l|cc|cc}
\toprule
\multirow{2}{*}{\diagbox{Dataset}{Metric}}  & \multicolumn{2}{c|}{\textbf{Acc} $\uparrow$} & \multicolumn{2}{c}{\textbf{IoU} $\uparrow$} \\
                                  & w/o Ref. & w/ Ref. & w/o Ref. & w/ Ref. \\
\midrule
SketchyScene   & 0.85 & \textbf{0.92} & 0.79 & \textbf{0.88} \\
Zhang \etal   & 0.79 & \textbf{0.86} & 0.67 & \textbf{0.74} \\ 
SketchAgent    & 0.82 & \textbf{0.88} & 0.76 & \textbf{0.84} \\
CLIPasso base  & 0.86 & \textbf{0.91} & 0.81 & \textbf{0.88} \\
CLIPasso 01    & 0.82 & \textbf{0.87} & 0.79 & \textbf{0.86} \\
CLIPasso 04    & 0.85 & \textbf{0.85} & 0.82 & \textbf{0.84} \\
CLIPasso 11    & 0.82 & \textbf{0.89} & 0.77 & \textbf{0.85} \\
InstantStyle  & 0.73 & \textbf{0.78} & 0.60 & \textbf{0.70} \\
\midrule 
All & 0.82 & \textbf{0.87 } & 0.75  & \textbf{0.82} \\
    & $\pm$ 0.04 & $\pm$ 0.04 & $\pm$ 0.08 & $\pm$ 0.07 \\ 
\bottomrule
\end{tabular}
\caption{\textbf{Comparison of segmentation performance with and without refinement module. }We see consistent improvement in performance across all datasets. }
\vspace{-8mm}
\label{tb:ablation}
\end{table}
\rev{
We conduct an ablation study on our depth-guided refinement module across the benchmark datasets, with quantitative results summarized in Table~\ref{tb:ablation}. On average, removing the refinement module results in segmentation performance degradation of a 0.05 decrease in accuracy and a 0.07 decrease in IoU, underscoring its importance in achieving precise instance segmentation.}
\rev{
Figure~\ref{fig:refinement_ablation} presents qualitative comparisons of segmentation results for the same sketch, with and without the refinement module. These examples further illustrate the module’s effectiveness in resolving ambiguities and enhancing segmentation quality.}

\section{Depth Maps Accuracy}
\rev{
The effectiveness of our refinement module depends on accurately sorting object instances, which in turn relies on the quality of the estimated sketch depth maps. To generate these depth maps, we use DepthAnythingV2~\cite{depth_anything_v2}, a model trained on a diverse set of visual inputs including artworks and sketches. However, evaluating depth map accuracy remains challenging due to the lack of ground-truth depth annotations for our synthetic sketches. To address this, we present both qualitative visualizations and a proxy quantitative evaluation. Specifically, we assess depth map quality on InstantStyle sketches, for which we have access to the corresponding natural images. These natural images are assumed to yield more reliable depth estimations, enabling a comparative evaluation of the sketch-derived depth maps.}

\subsection{Qualitative Evaluation of Depth Maps}
\begin{figure*}
    \centering
    \setlength{\tabcolsep}{1.5pt}
    {\small
    \resizebox{0.88\textwidth}{!}{ 
    \begin{tabular}{c c @{\hskip 10pt} c c}
            Sketch & Depth Map & Sketch & Depth Map \\
            \frame{\includegraphics[width=0.22\linewidth]{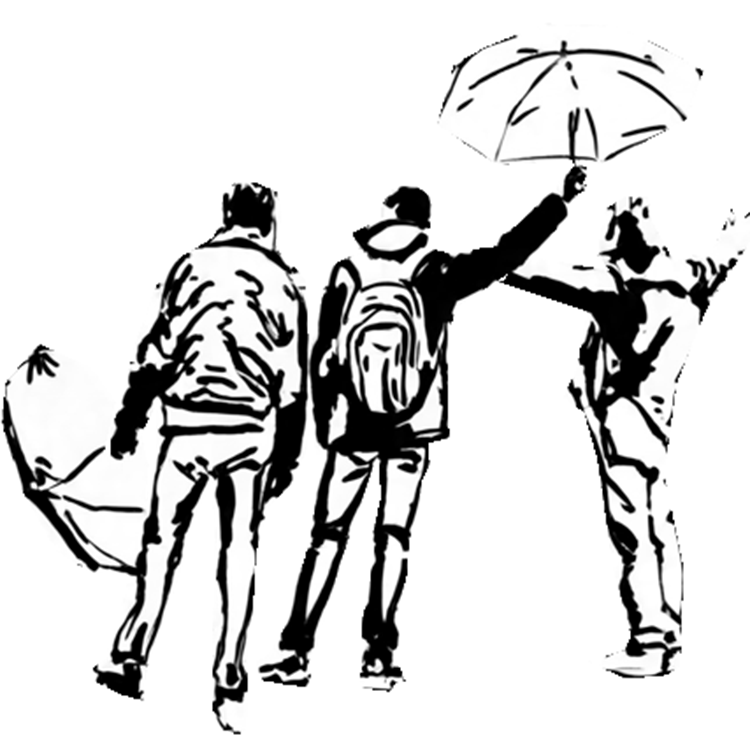}} &
            \frame{\includegraphics[width=0.22\linewidth]{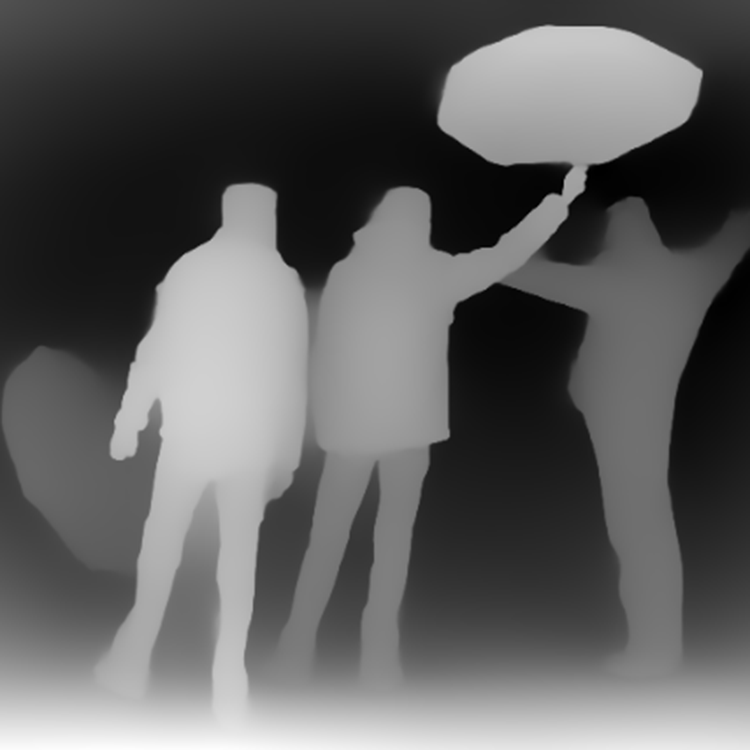}} & 
            \frame{\includegraphics[width=0.22\linewidth]{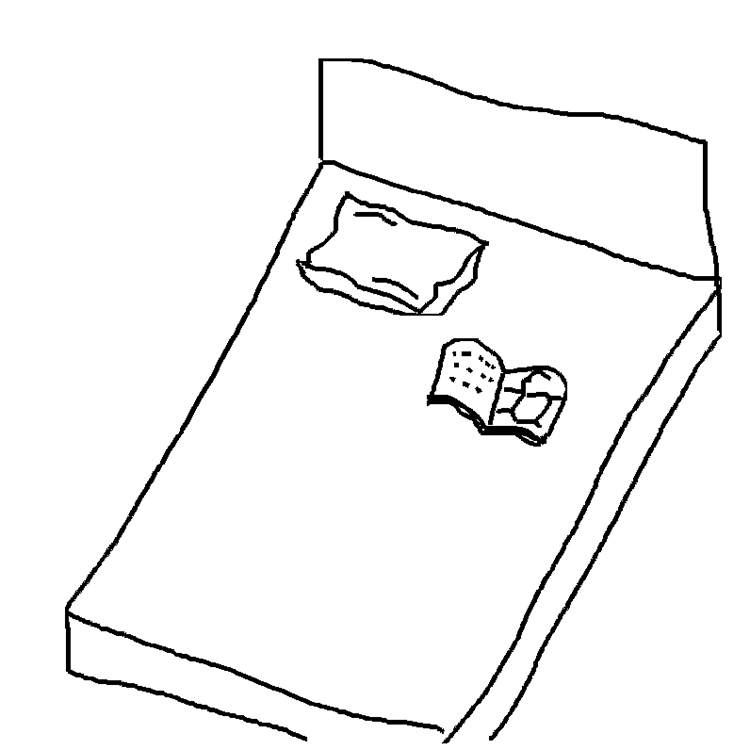}} & 
            \frame{\includegraphics[width=0.22\linewidth]{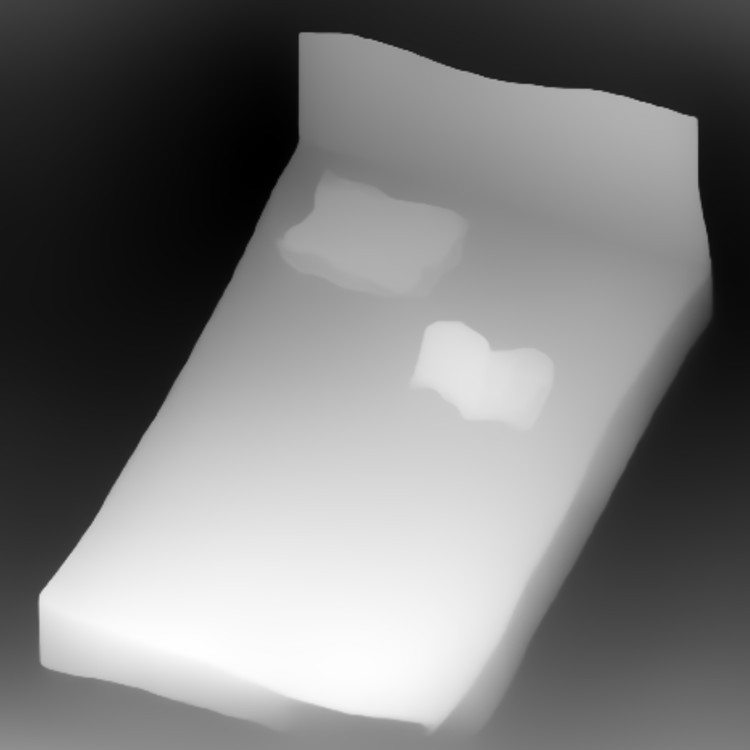}} \\
            
            \frame{\includegraphics[width=0.22\linewidth]{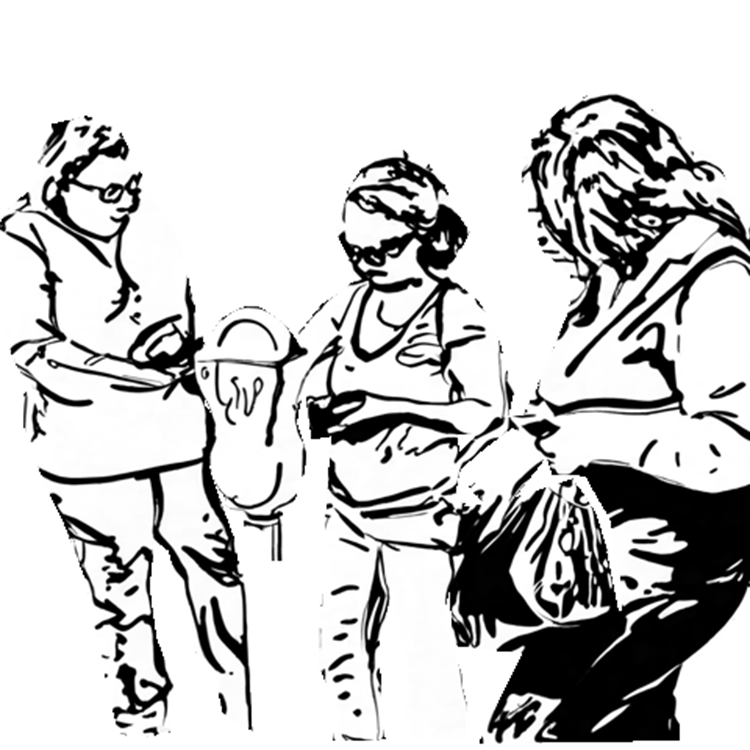}} &
            \frame{\includegraphics[width=0.22\linewidth]{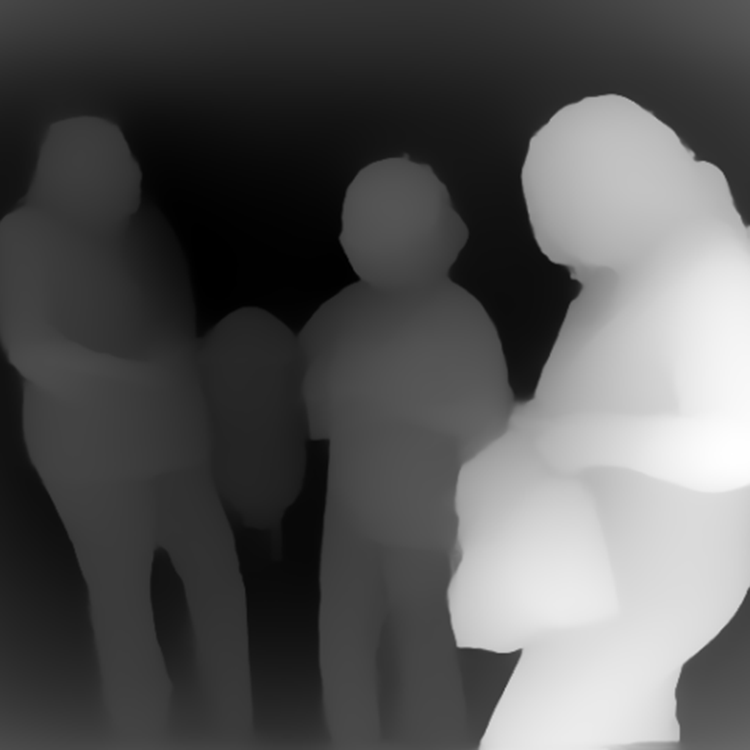}} & 
            \frame{\includegraphics[width=0.22\linewidth]{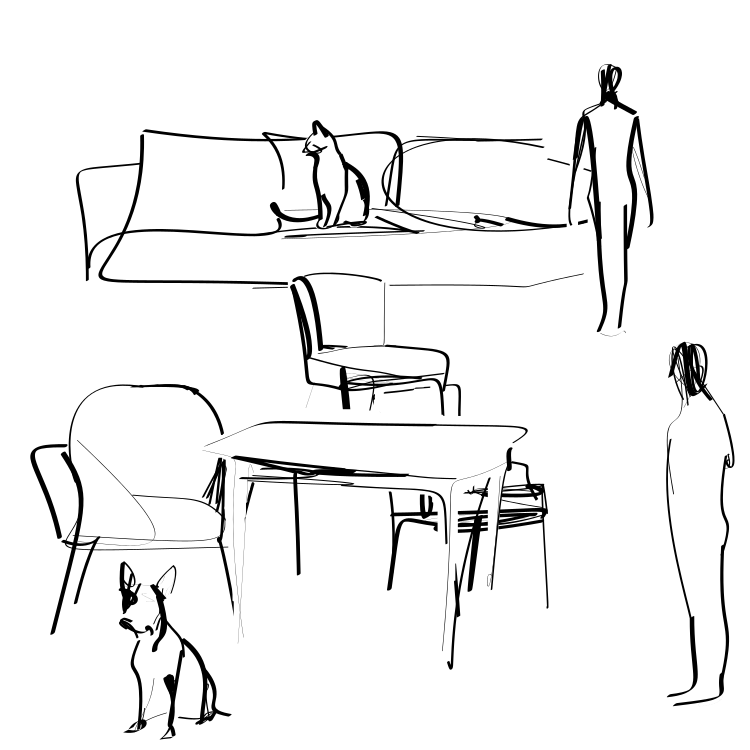}} & 
            \frame{\includegraphics[width=0.22\linewidth]{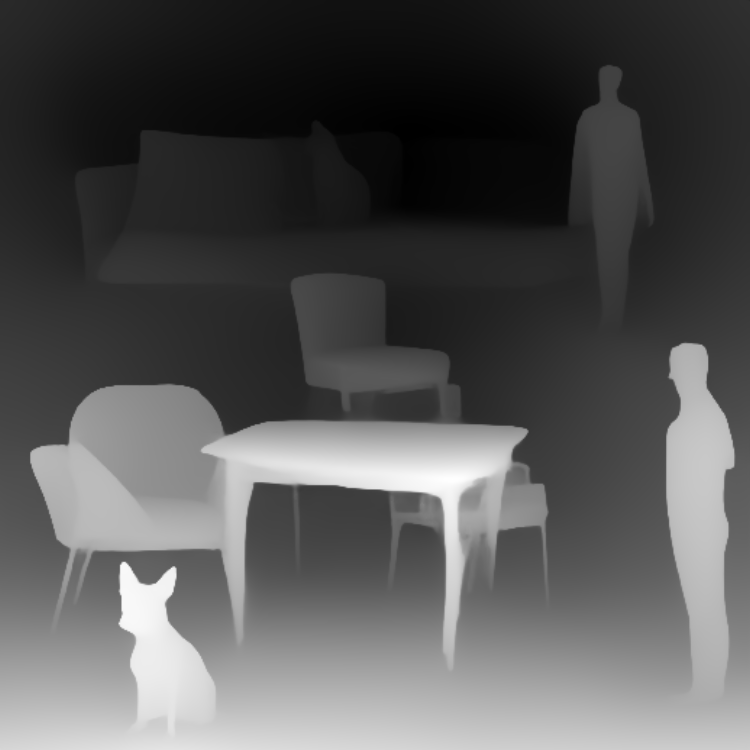}} \\
            
            \frame{\includegraphics[width=0.22\linewidth]{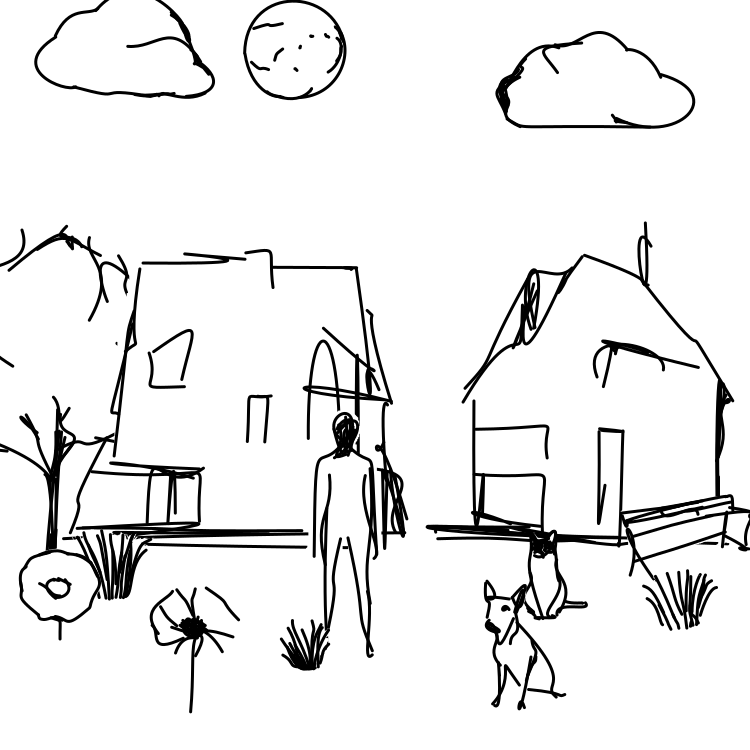}} &
            \frame{\includegraphics[width=0.22\linewidth]{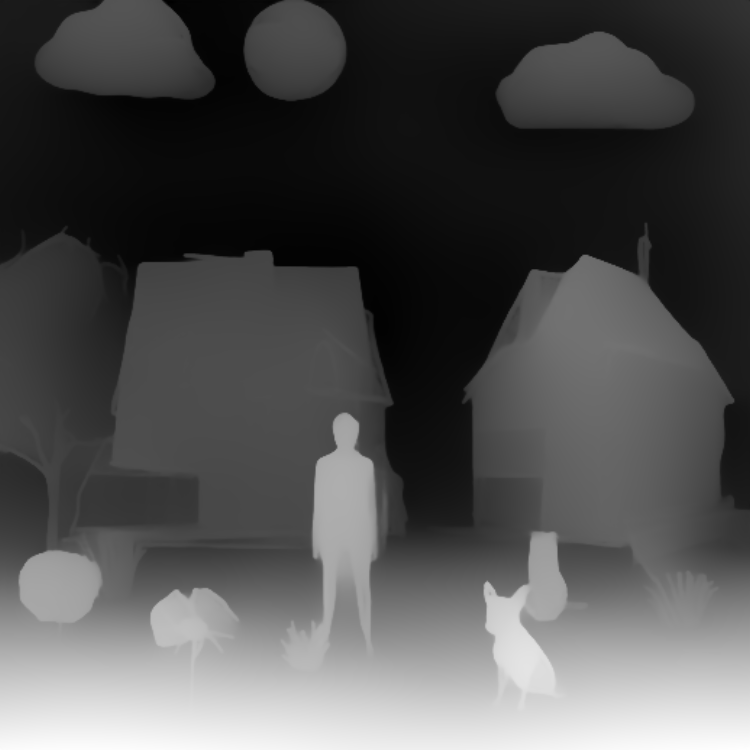}} & 
            \frame{\includegraphics[width=0.22\linewidth]{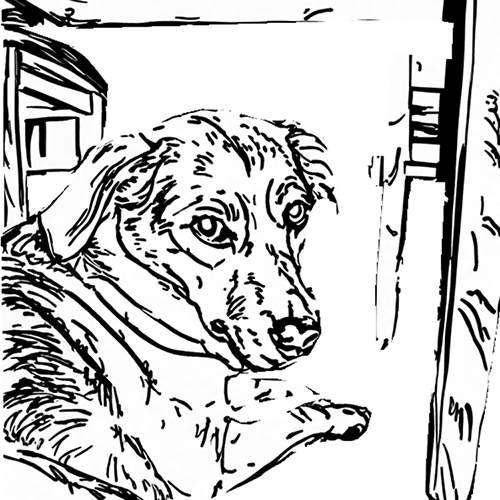}} & 
            \frame{\includegraphics[width=0.22\linewidth]{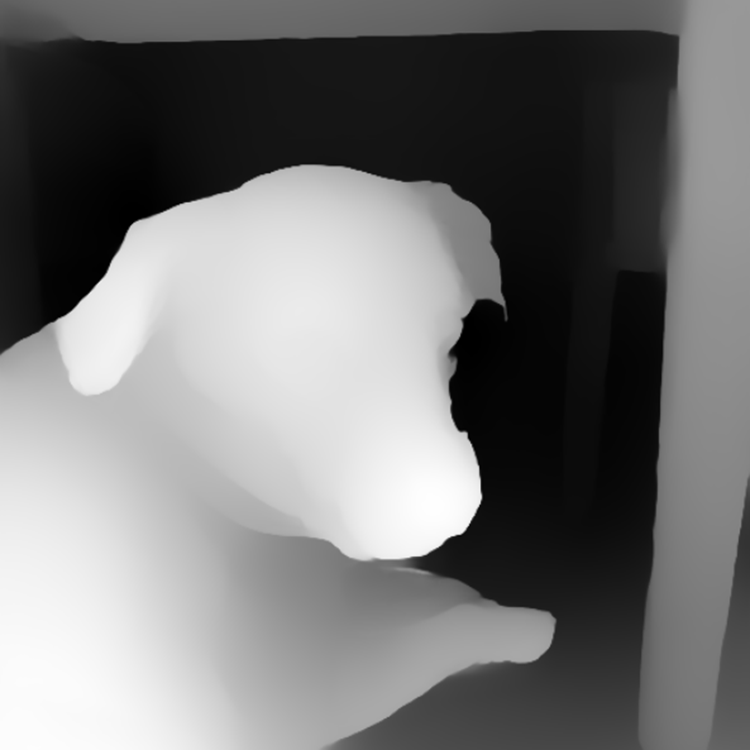}} \\
            
            \frame{\includegraphics[width=0.22\linewidth]{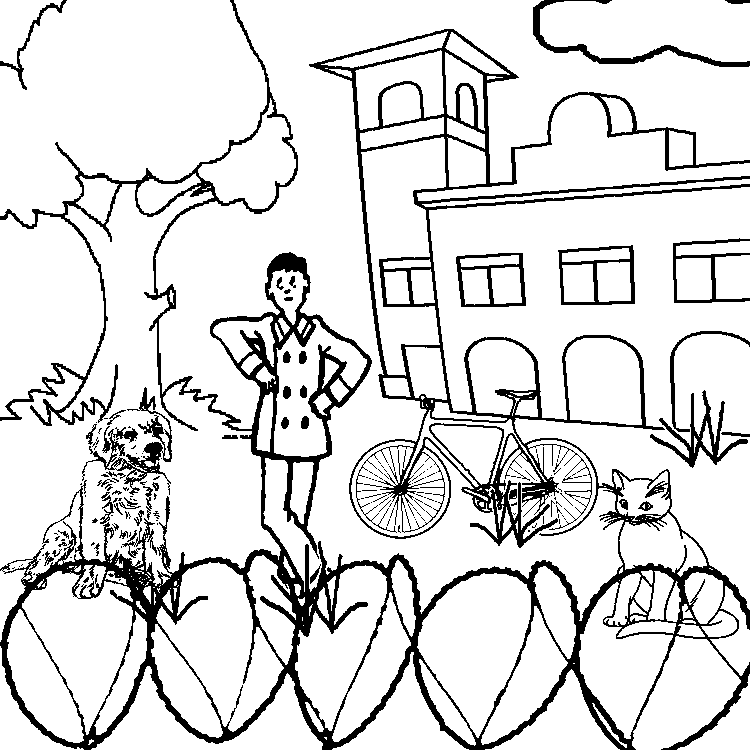}} &
            \frame{\includegraphics[width=0.22\linewidth]{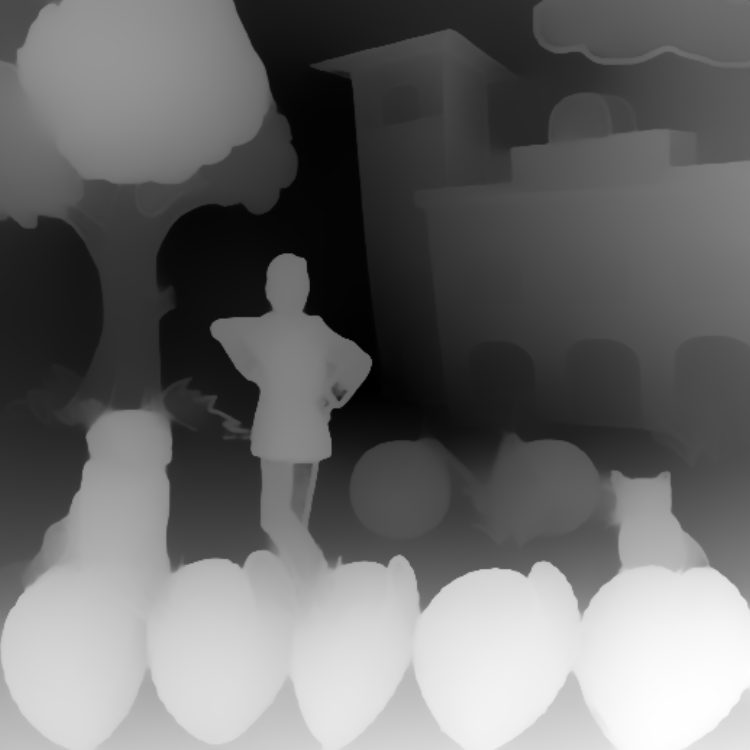}} & 
            \frame{\includegraphics[width=0.22\linewidth]{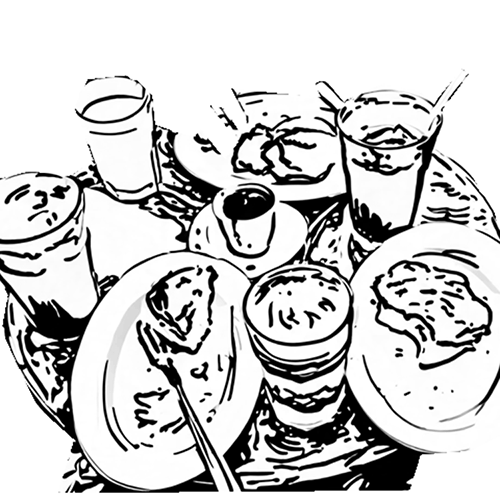}} & 
            \frame{\includegraphics[width=0.22\linewidth]{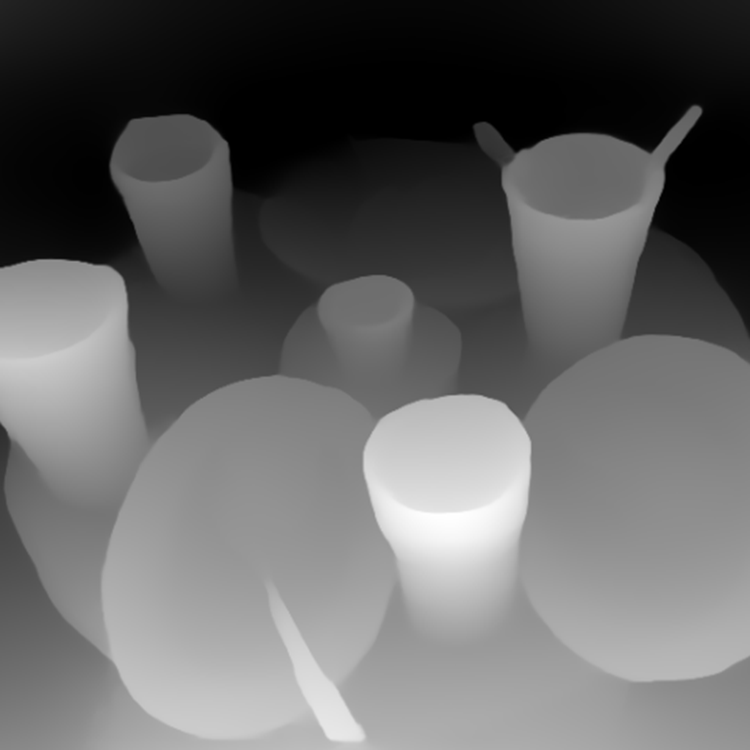}} \\
            
            \frame{\includegraphics[width=0.22\linewidth]{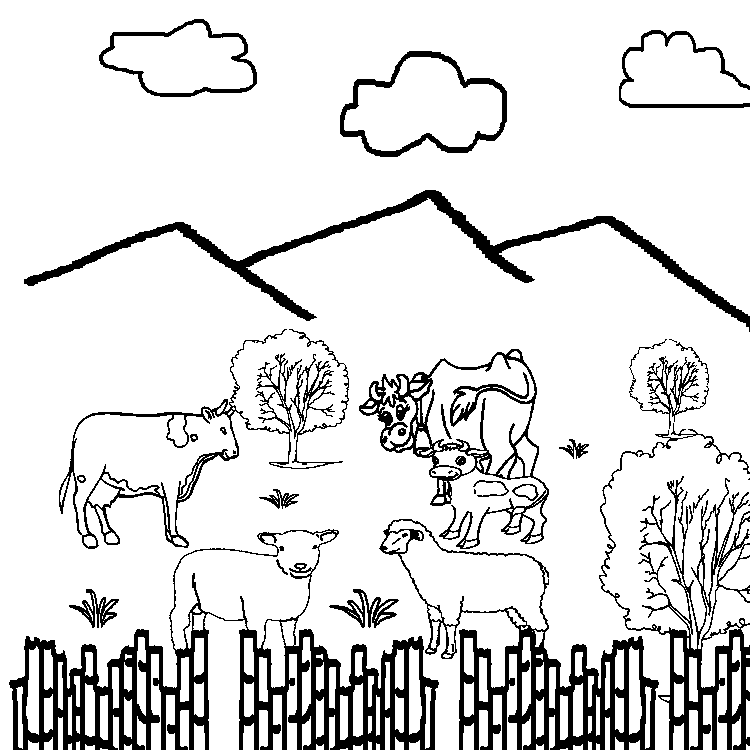}} &
            \frame{\includegraphics[width=0.22\linewidth]{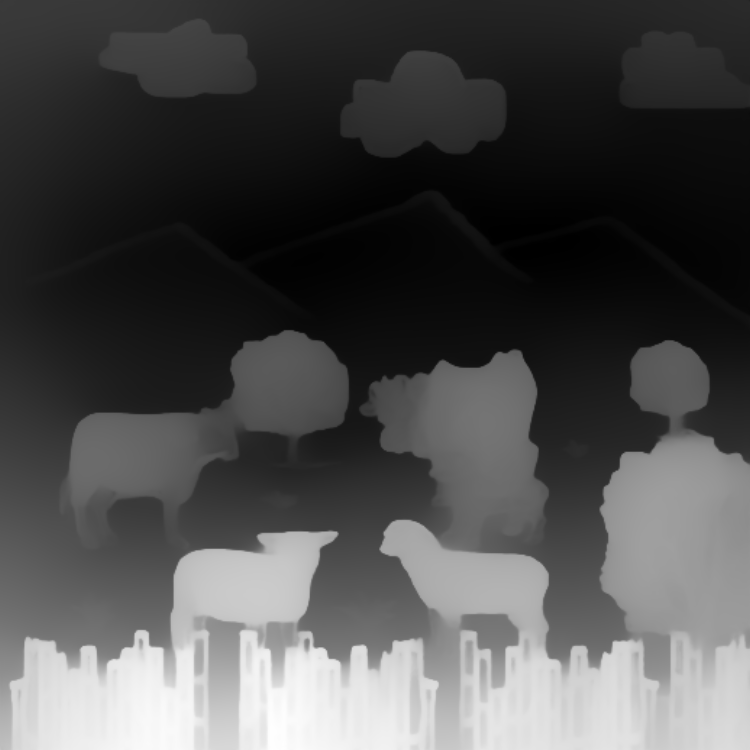}} & 
            \frame{\includegraphics[width=0.22\linewidth]{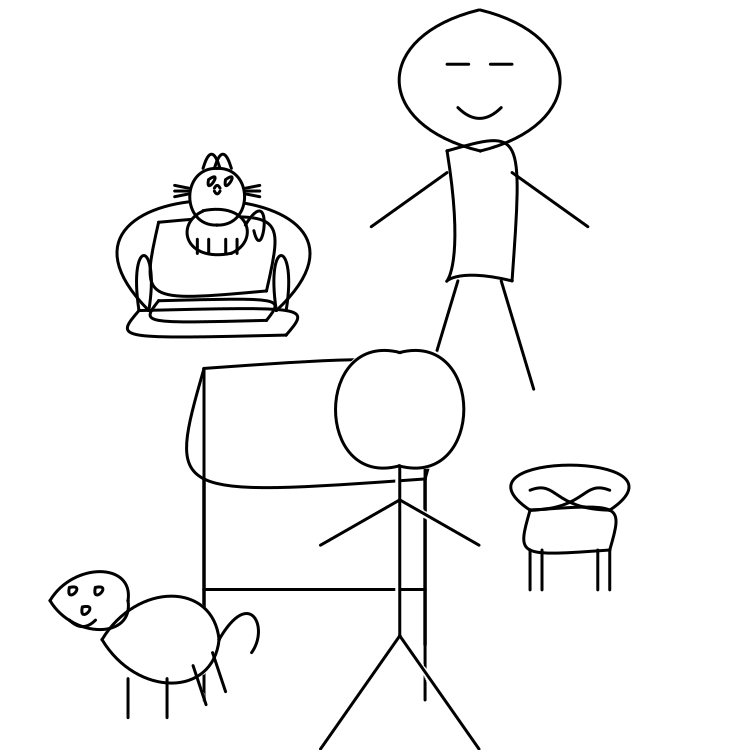}} & 
            \frame{\includegraphics[width=0.22\linewidth]{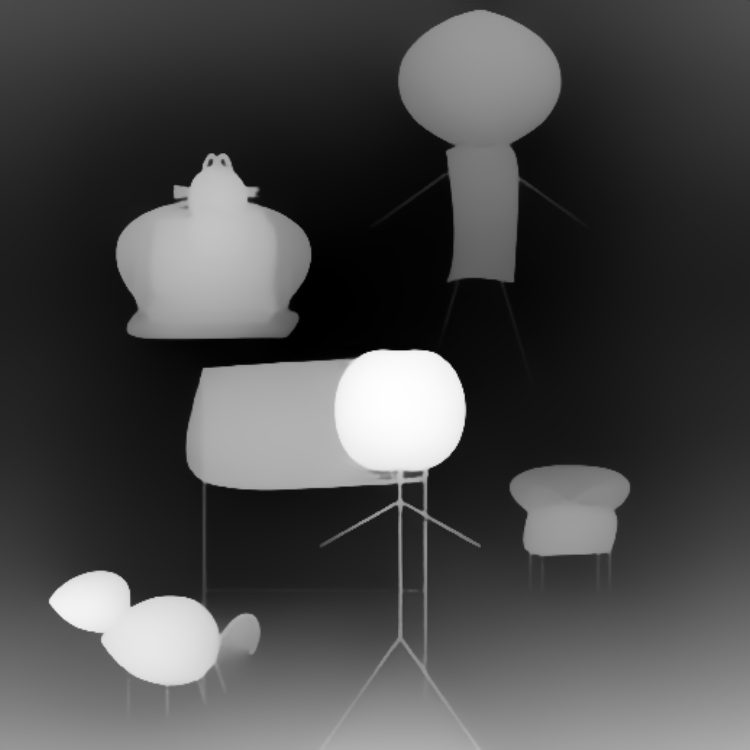}} \\
        \end{tabular}
    }
    \vspace{-0.1cm}
    \caption{\rev{\textbf{Sketch and corresponding DepthAnythingV2~\shortcite{depth_anything_v2} depth map pairs.} We show qualitative results on sketch depth maps of scenes across datasets, demonstrating the reasonable performance of this depth estimator directly on sketches. The model generalizes surprisingly well across a wide spectrum of sketch styles, from highly abstract and symbolic representations to more detailed and realistic ones. Despite the lack of color or shading cues, the predicted depth maps often capture overall scene layout and object ordering effectively.}}
    \label{fig:sketch_depth_qual}
    }
\end{figure*}

\rev{
We present qualitative results of the estimated depth maps in Figure~\ref{fig:sketch_depth_qual}, showcasing a variety of sketches across different datasets and scene types. These examples highlight the ability of DepthAnythingV2 to extract meaningful depth cues from sparse and stylized sketch inputs. Despite the abstract nature of the input sketches—ranging from realistic human figures to symbolic line drawings of animals and buildings—the model produces coherent depth estimates that reflect the relative spatial layout of the scene. These depth cues play a crucial role in our refinement module, helping to disambiguate overlapping regions and enforce depth-aware instance separation.}

\subsection{Quantitative Evaluation of Depth Maps}

%%CHUAN 
% For the quantitative evaluation, rather than computing the average absolute relative error, we assess the consistency of object ordering between the predicted depth maps from sketches and the ground truth depth maps using Kendall’s Tau coefficient. This approach is adopted because our method relies on relative depth for sketch-based object layering, making the absolute depth values less critical.
% The comparative experiments indicate an average of 80\% agreement between the predicted object ordering from sketches and the ground truth. The evaluation methodology is illustrated in Figure~\ref{fig:depth_order_evaluation}.
\rev{For the quantitative evaluation of depth quality, we assess the consistency of object ordering between the predicted depth maps (from sketches) and the ground truth depth maps (from corresponding natural images) using Kendall’s Tau coefficient. Rather than relying on absolute depth metrics such as average relative error, we adopt this rank-based approach to better reflect our method’s reliance on relative depth for instance layering. Since our refinement module operates on object ordering rather than precise depth values, this metric offers a more meaningful evaluation.}
\rev{
Our results show an average of 80\% agreement in object ordering, demonstrating the effectiveness of sketch-based depth estimation for supporting depth-aware segmentation. }

\section{InkScenes Dataset Generation Details}
In this section, we provide additional details on our synthetic dataset \datasetname{}'s creation process. 

\subsection{Generating Vector Scene Sketches}
\begin{figure}[H]
    \centering
    \setlength{\tabcolsep}{2pt}
    {\small
    \resizebox{1\linewidth}{!}{ 
    \begin{tabular}{c c  c c }
        Input Synthetic Image & Object Vector Sketch & Input Synthetic Image & Object Vector Sketch \\
        \frame{\includegraphics[width=0.32\linewidth]{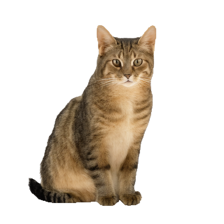}} &
        \frame{\includegraphics[width=0.32\linewidth]{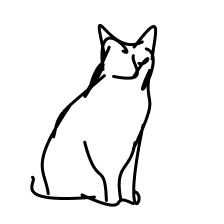}} &
        \frame{\includegraphics[width=0.32\linewidth]{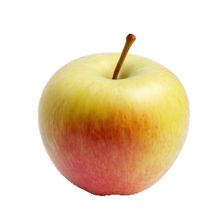}} &
        \frame{\includegraphics[width=0.32\linewidth]{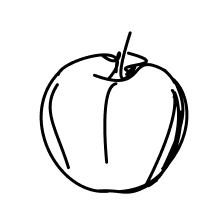}} \\

        \frame{\includegraphics[width=0.32\linewidth]{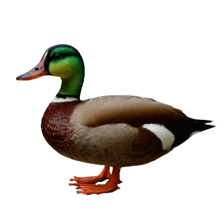}} &
        \frame{\includegraphics[width=0.32\linewidth]{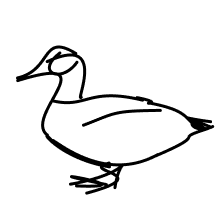}} &
        \frame{\includegraphics[width=0.32\linewidth]{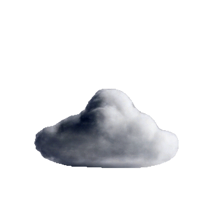}} &
        \frame{\includegraphics[width=0.32\linewidth]{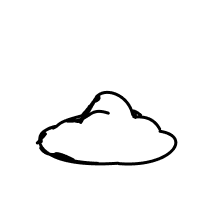}} \\

        \frame{\includegraphics[width=0.32\linewidth]{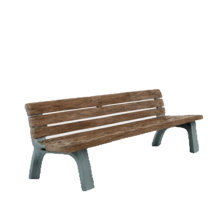}} &
        \frame{\includegraphics[width=0.32\linewidth]{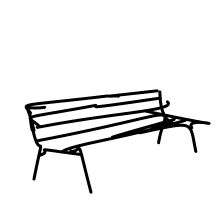}} &
        \frame{\includegraphics[width=0.32\linewidth]{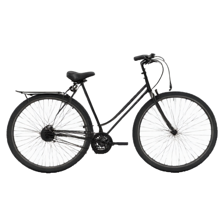}} &
        \frame{\includegraphics[width=0.32\linewidth]{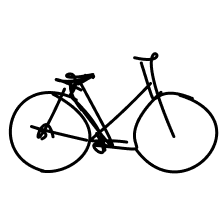}} \\
    \end{tabular}
    }}
    \vspace{-0.3cm}  \caption{Examples pairs of input synthetic image and output generated object vector sketch. }
    \label{fig:clipasso_component}
    \vspace{-5mm}
\end{figure}

\begin{figure}[H]
    \centering
    \setlength{\tabcolsep}{2pt}
    {\small
    \resizebox{1\linewidth}{!}{ 
    \begin{tabular}{c c  c c }
        ``bus" & ``bird" & ``cat" & ``butterfly"\\ 
        \frame{\includegraphics[width=0.32\linewidth]{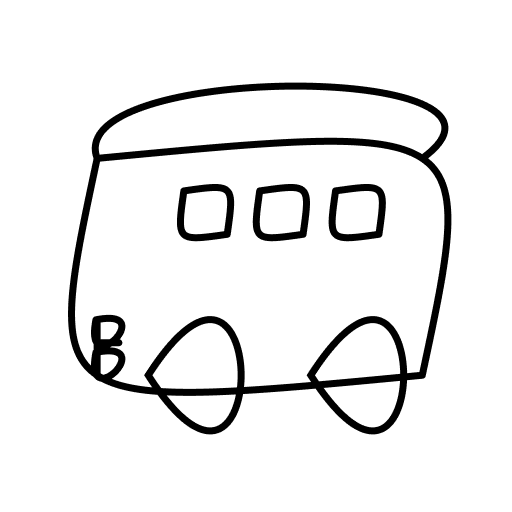}} &
        \frame{\includegraphics[width=0.32\linewidth]{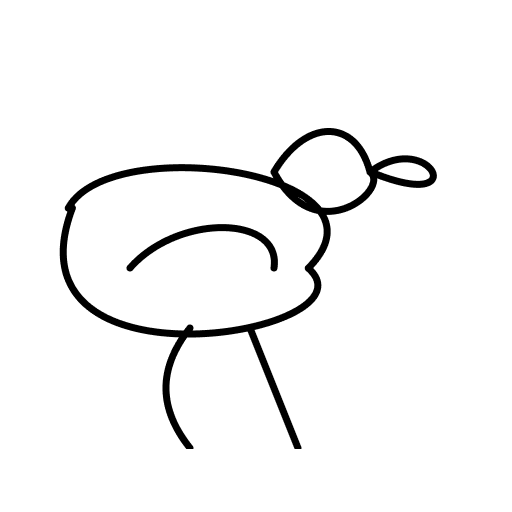}} &
        \frame{\includegraphics[width=0.32\linewidth]{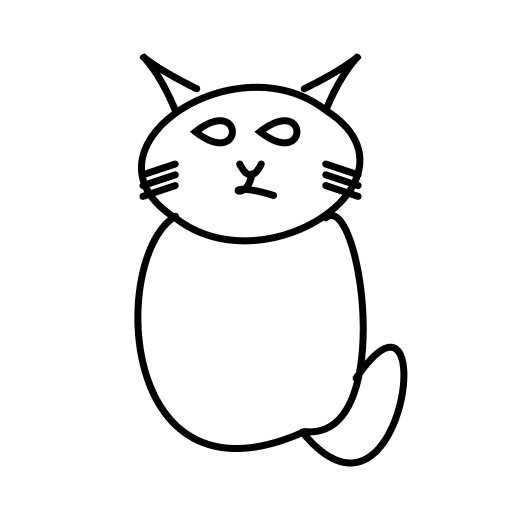}} &
        \frame{\includegraphics[width=0.32\linewidth]{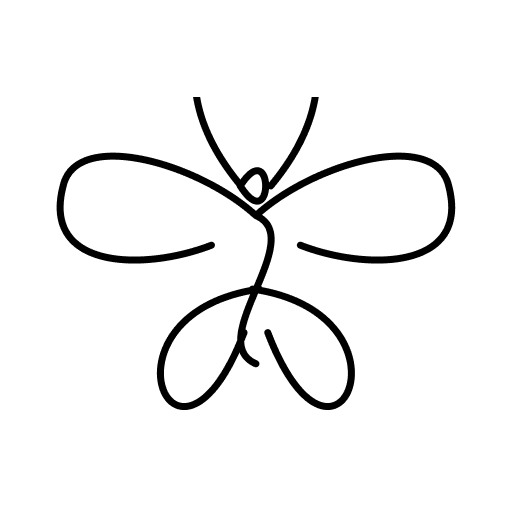}} \\

        ``chair" & ``flower" & ``people" & ``house" \\
        \frame{\includegraphics[width=0.32\linewidth]{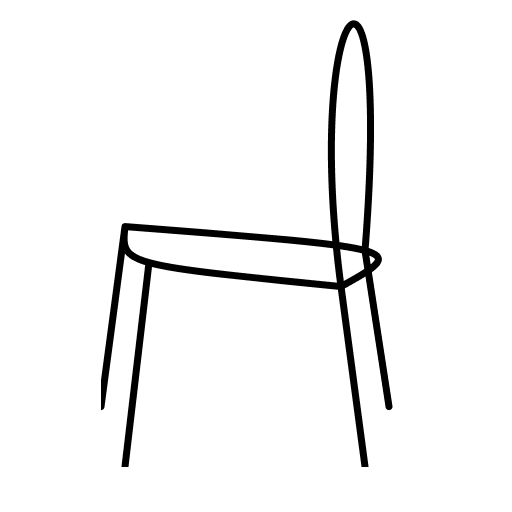}} &
        \frame{\includegraphics[width=0.32\linewidth]{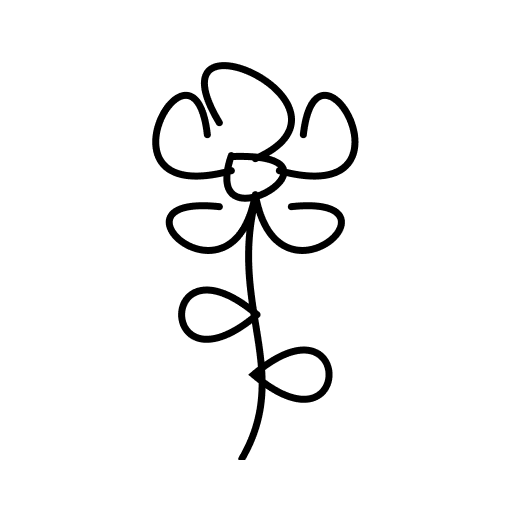}} &
        \frame{\includegraphics[width=0.32\linewidth]{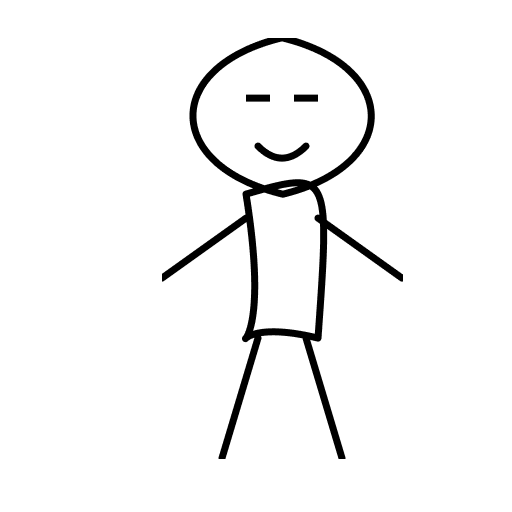}} &
        \frame{\includegraphics[width=0.32\linewidth]{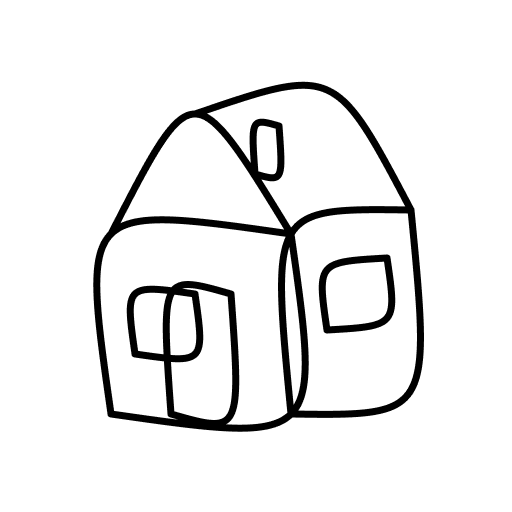}} \\

        ``basket" & ``bench" & ``cup" & ``car" \\
        \frame{\includegraphics[width=0.32\linewidth]{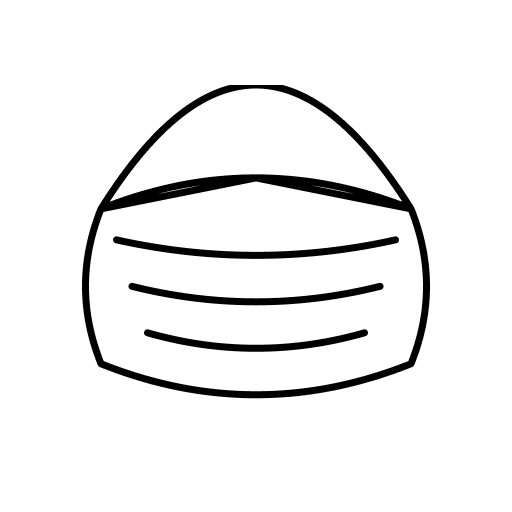}} &
        \frame{\includegraphics[width=0.32\linewidth]{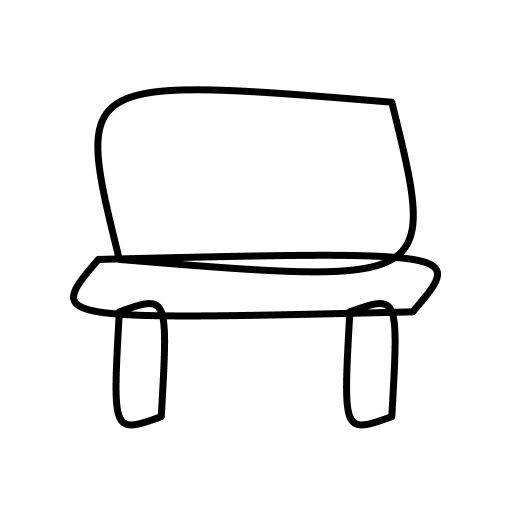}} &
        \frame{\includegraphics[width=0.32\linewidth]{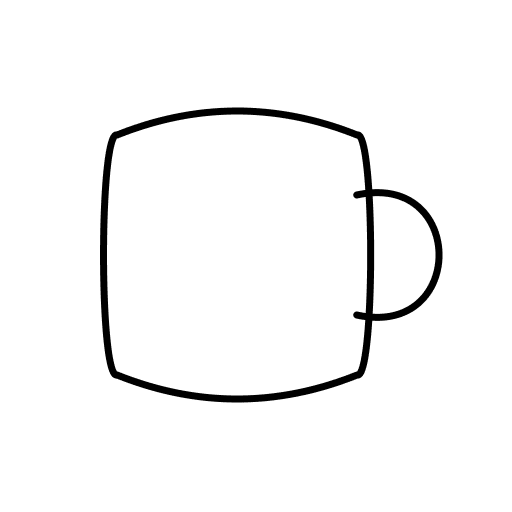}} &
        \frame{\includegraphics[width=0.32\linewidth]{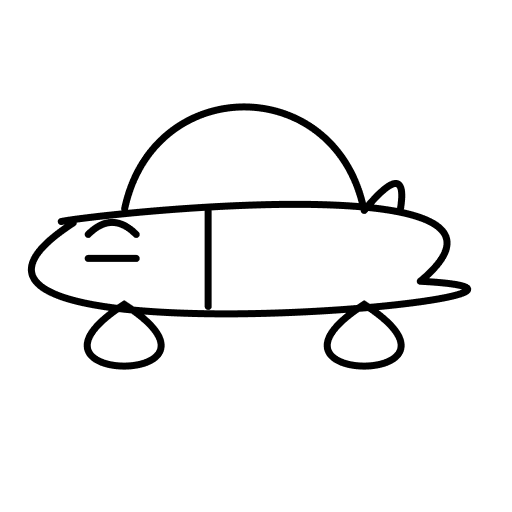}} \\
    \end{tabular}
    }}
    \vspace{-0.3cm}  \caption{Examples of object vector sketches generated by SketchAgent.}
    \label{fig:sketchagent_component}
\end{figure}

We employ CLIPasso \shortcite{vinker2022clipasso} and SketchAgent \shortcite{vinker2024sketchagent} to generate diverse vector sketches of single objects. For each generation method, we create 10 distinct object instances for all 45 classes in the SketchyScene dataset \shortcite{Zou18SketchyScene}, ensuring sufficient variability in our synthetic scenes.
For CLIPasso's image-to-vector conversion, we first generate photorealistic synthetic images using SDXL \shortcite{rombach2021highresolution}. The generation process uses a consistent prompt template: \textit{``A realistic image of a \{class\_name\} with a blank background"}. Figure~\ref{fig:clipasso_component} demonstrates representative pairs of synthetic input images and their corresponding generated vector sketches. 
For SketchAgent, we generate sketches directly from class labels as text prompts, producing 10 samples per class. Figure~\ref{fig:sketchagent_component} illustrates representative examples of the generated object sketches.

\subsection{Generating Sketches from Natural Images}
\begin{figure*}
    \centering
    \setlength{\tabcolsep}{2pt}
    {\small
    \resizebox{0.88\textwidth}{!}{ 
    \begin{tabular}{c c c}
        \frame{\includegraphics[trim=0 1.3cm 0 0.1cm,clip,width=0.32\linewidth]{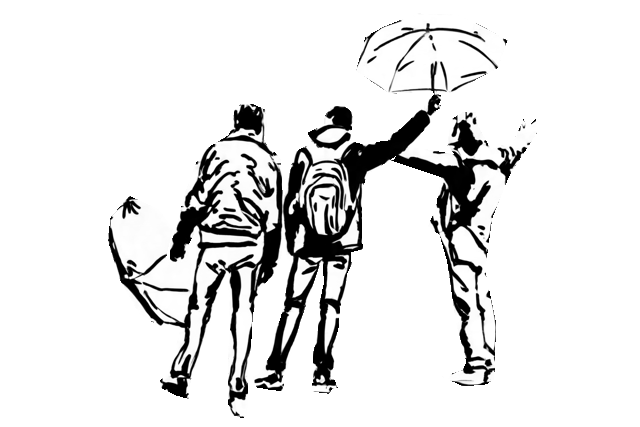}} &
        \frame{\includegraphics[trim=0 1.3cm 0 0,clip,width=0.32\linewidth]{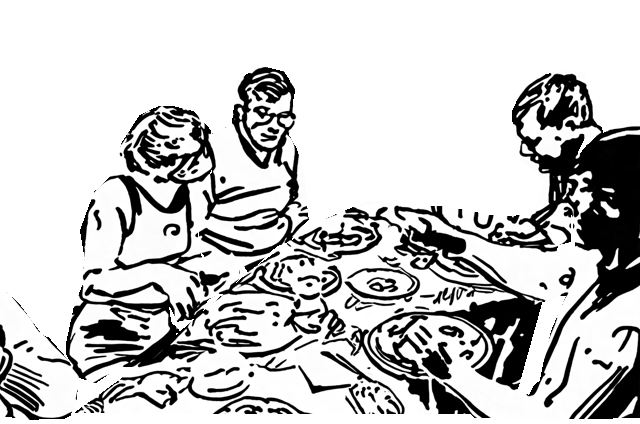}} &
        \frame{\includegraphics[trim=0 1.3cm 0 1.2cm,clip,width=0.32\linewidth]{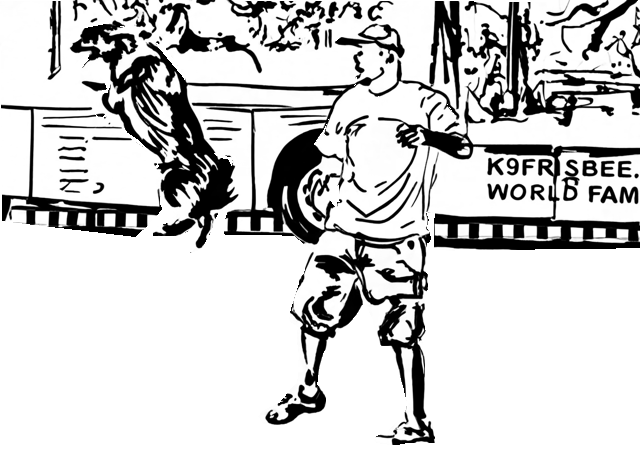}} \\

        \frame{\includegraphics[trim=0 3.2cm 0 0cm,clip,width=0.32\linewidth]{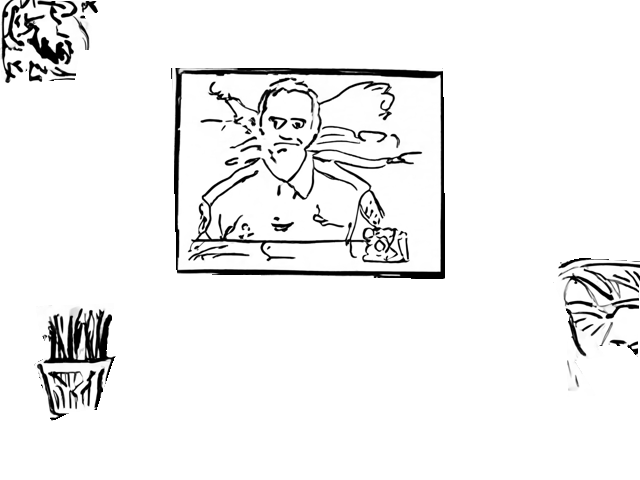}} &
        \frame{\includegraphics[trim=0 1.3cm 0 0,clip,width=0.32\linewidth]{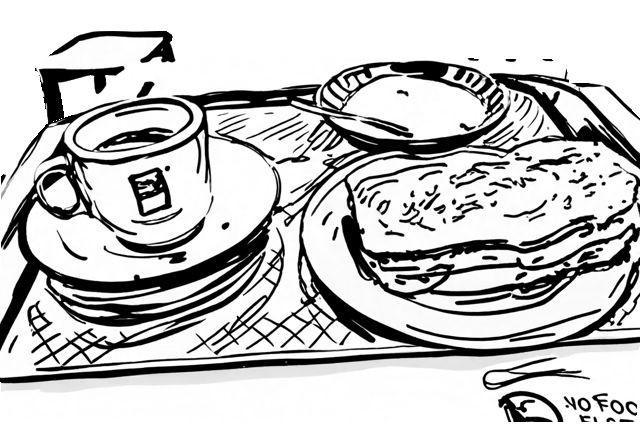}} &
        \frame{\includegraphics[trim=0 1.3cm 0 1.2cm,clip,width=0.32\linewidth]{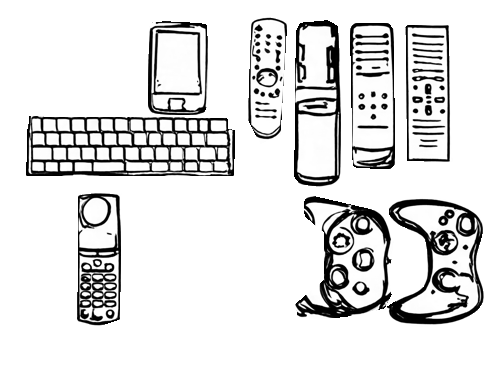}} \\

        \frame{\includegraphics[trim=0 1.3cm 0 1.2cm,clip,width=0.32\linewidth]{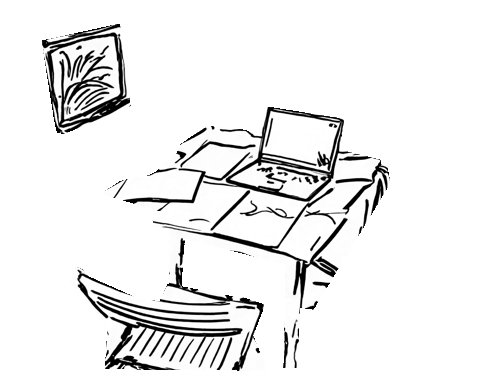}} &
        \frame{\includegraphics[trim=0 1.3cm 0 0,clip,width=0.32\linewidth]{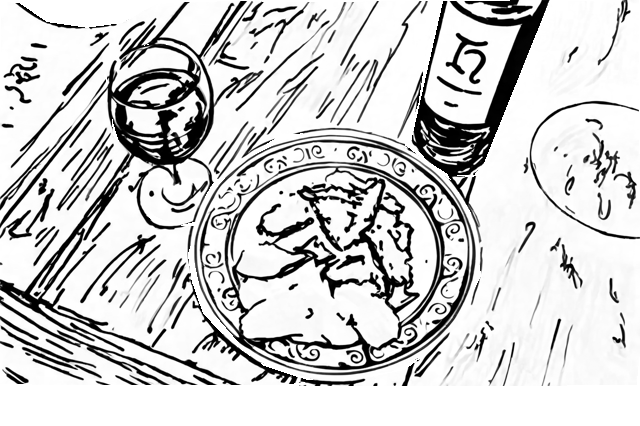}} &
        \frame{\includegraphics[trim=0 1.3cm 0 2cm,clip,width=0.32\linewidth]{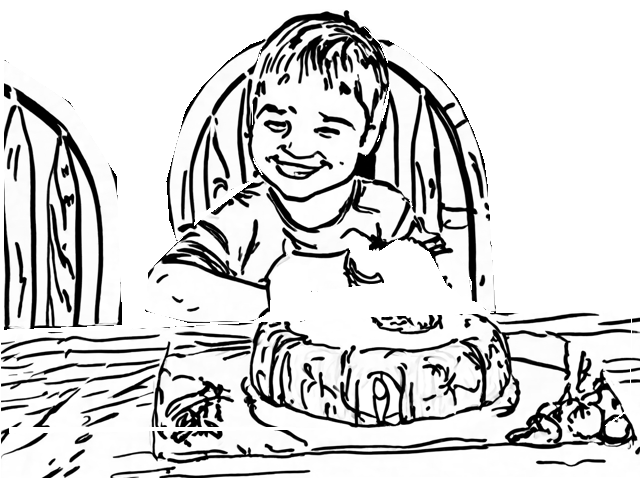}} \\

        \frame{\includegraphics[trim=0 0 0 2.4cm,clip,width=0.32\linewidth]{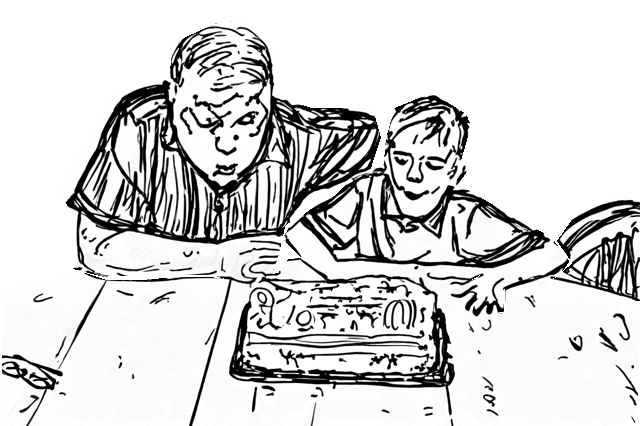}} &
        \frame{\includegraphics[trim=0 0 0 0,clip,width=0.32\linewidth]{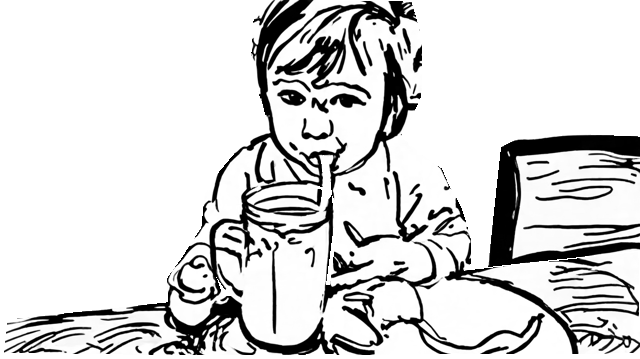}} &
        \frame{\includegraphics[trim=0 1.3cm 0 1cm,clip,width=0.32\linewidth]{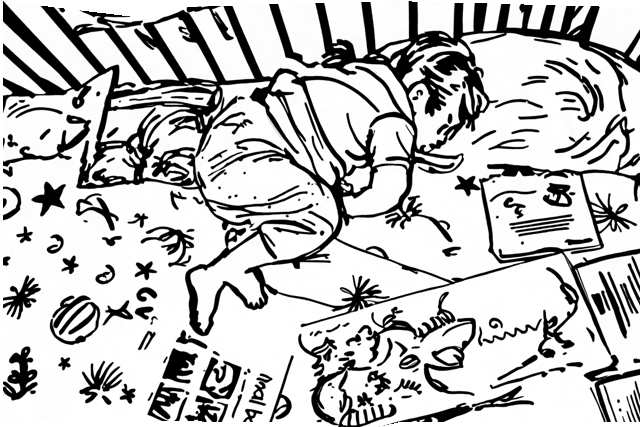}} \\

        \frame{\includegraphics[trim=0 0 0 2.4cm,clip,width=0.32\linewidth]{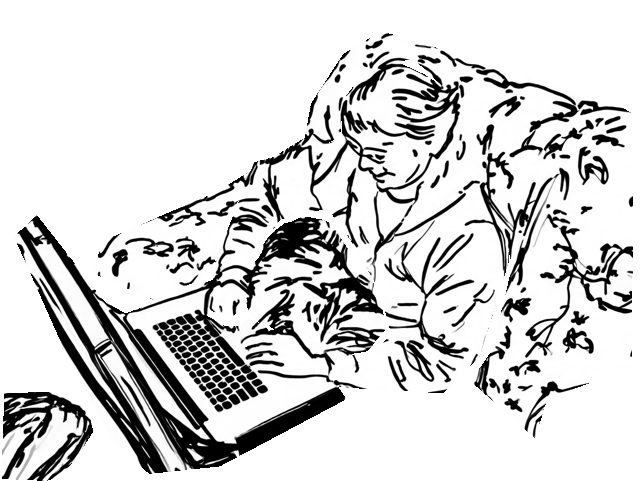}} &
        \frame{\includegraphics[trim=0 0 0 0.5cm,clip,width=0.32\linewidth]{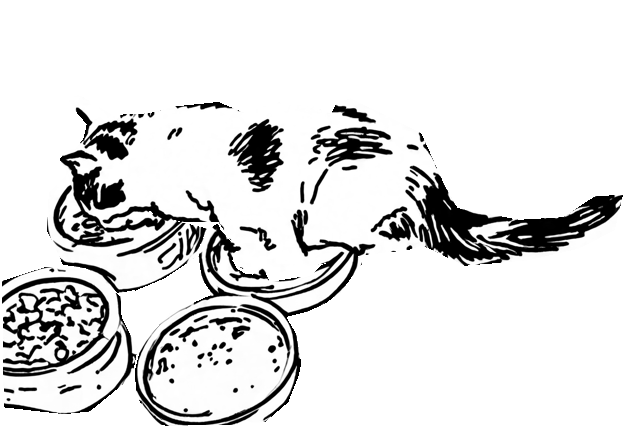}} &
        \frame{\includegraphics[trim=0 0.5cm 0 0,clip,width=0.32\linewidth]{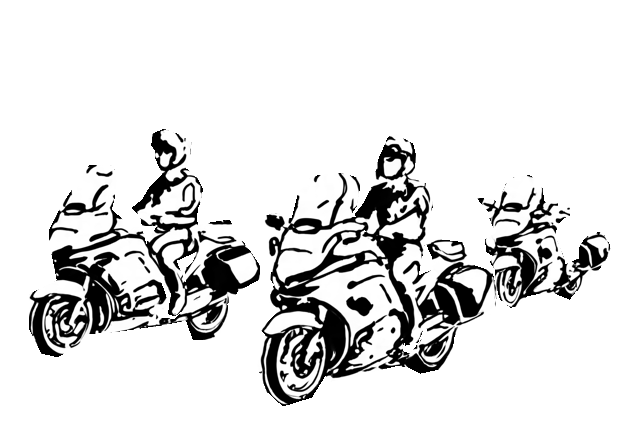}} \\

        \frame{\includegraphics[trim=0 0 0 0,clip,width=0.32\linewidth]{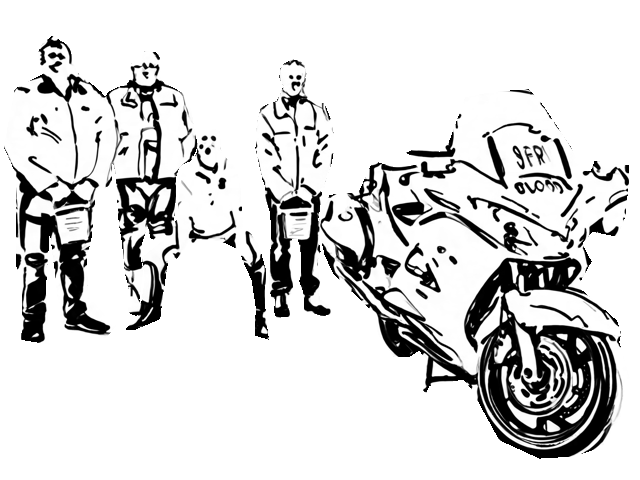}} &
        \frame{\includegraphics[trim=0 0 0 0cm,clip,width=0.32\linewidth]{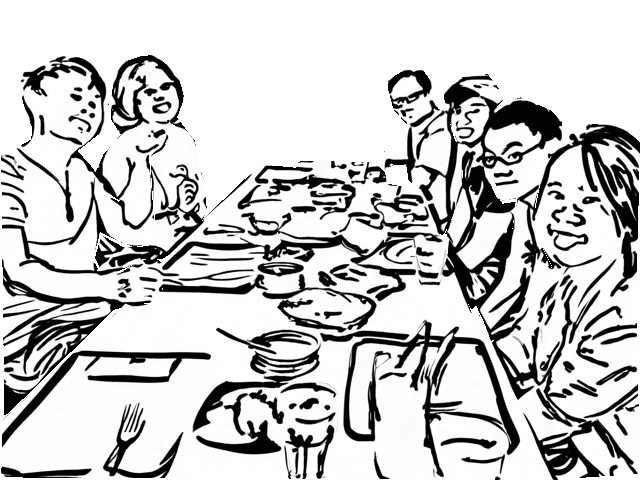}} &
        \frame{\includegraphics[trim=0 0 0 0,clip,width=0.32\linewidth]{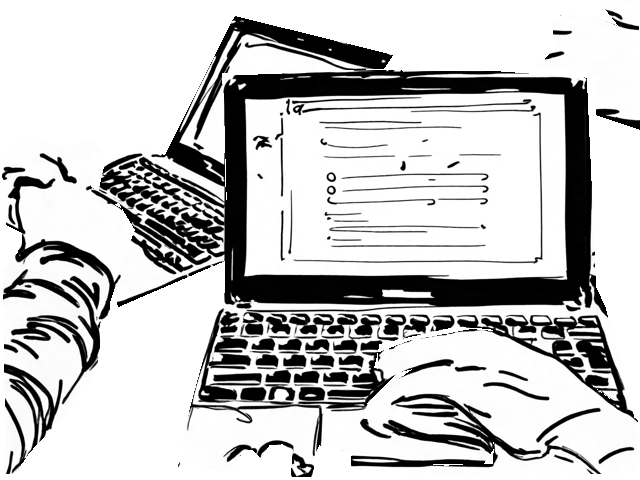}} \\
    \end{tabular}
    }
    \vspace{-0.1cm}
    \caption{\textbf{Example sketches from our InstantStyle dataset}. These sketches are derived from Visual Genome ~\shortcite{VisualGenome2017} containing 5 to 10 annotated objects. For visual clarity, we mask unsegmented regions in the generated sketches. This dataset contains 53 new categories that are not part of SketchyScene's  45 classes, and contain 1068 sketches in total. }
    \label{fig:instant_gallery}
    }
\end{figure*}

To expand beyond SketchyScene's object categories, we employ InstantStyle \shortcite{Wang2024InstantStyleFL} to transform a subset of Visual Genome \shortcite{VisualGenome2017} images into sketches, using a single CLIPasso object sketch as the style reference. Figure~\ref{fig:instant_gallery} showcases a gallery of examples from our InstantStyle-generated scene sketch dataset.

\section{InkScenes Dataset Statistics}
\begin{figure}[t]
    \centering
    \caption*{\textbf{InkScenes Dataset: Semantic Category Distribution}} 
    \includegraphics[width=1\linewidth]{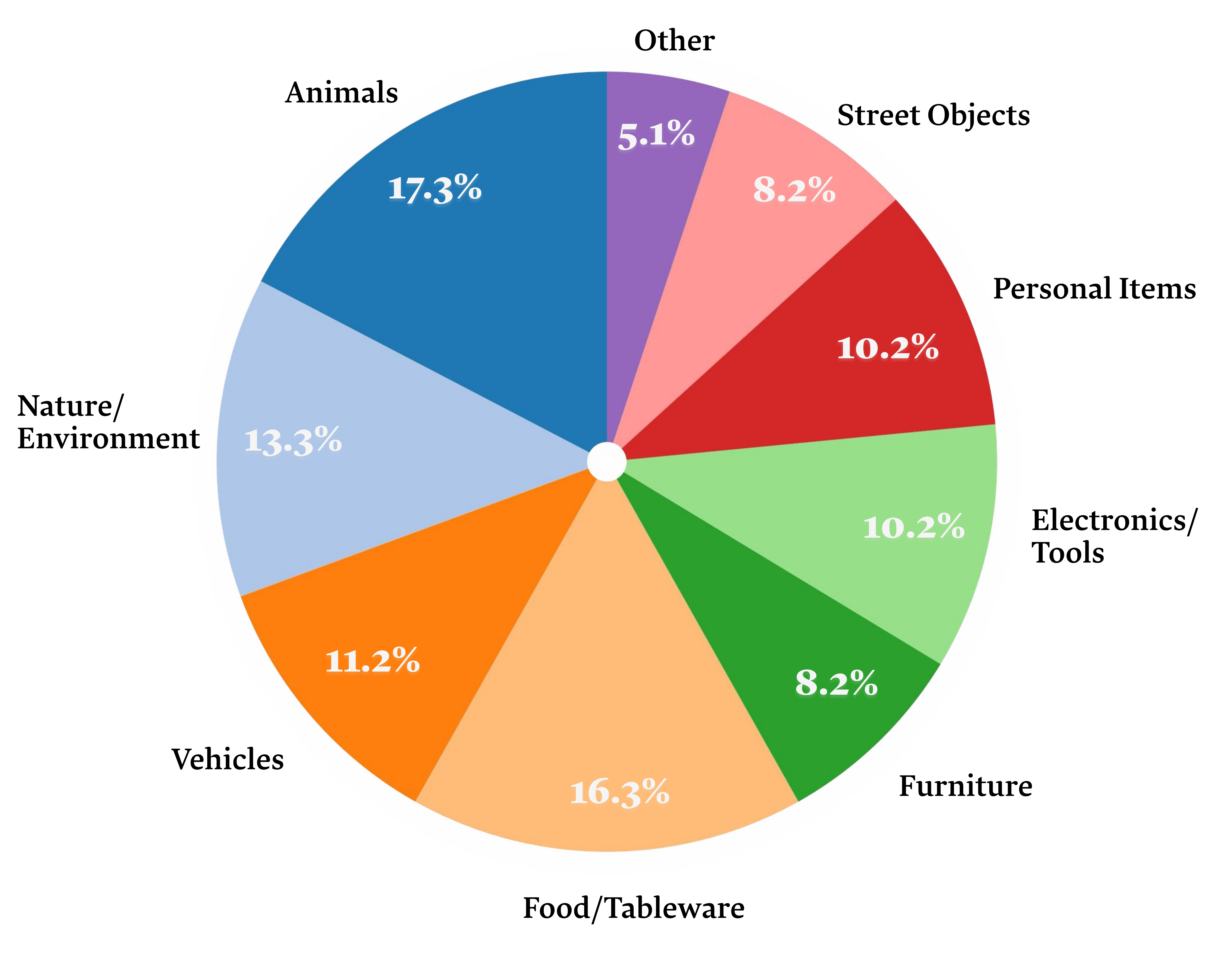}
    \caption{\rev{\textbf{Distribution of object categories in the \datasetname{} dataset, grouped into high-level semantic categories. } We group all 98 unique classes into high-level categories such as animals, nature, vehicles, and household objects. }}
    \label{fig:dataset_distribution}
\end{figure}

\begin{figure*}[t]
    \centering
    \vspace{-5mm}
    \caption*{\textbf{InkScenes InstantStyle Variant: Class Distribution}} 
    \includegraphics[width=1\linewidth]{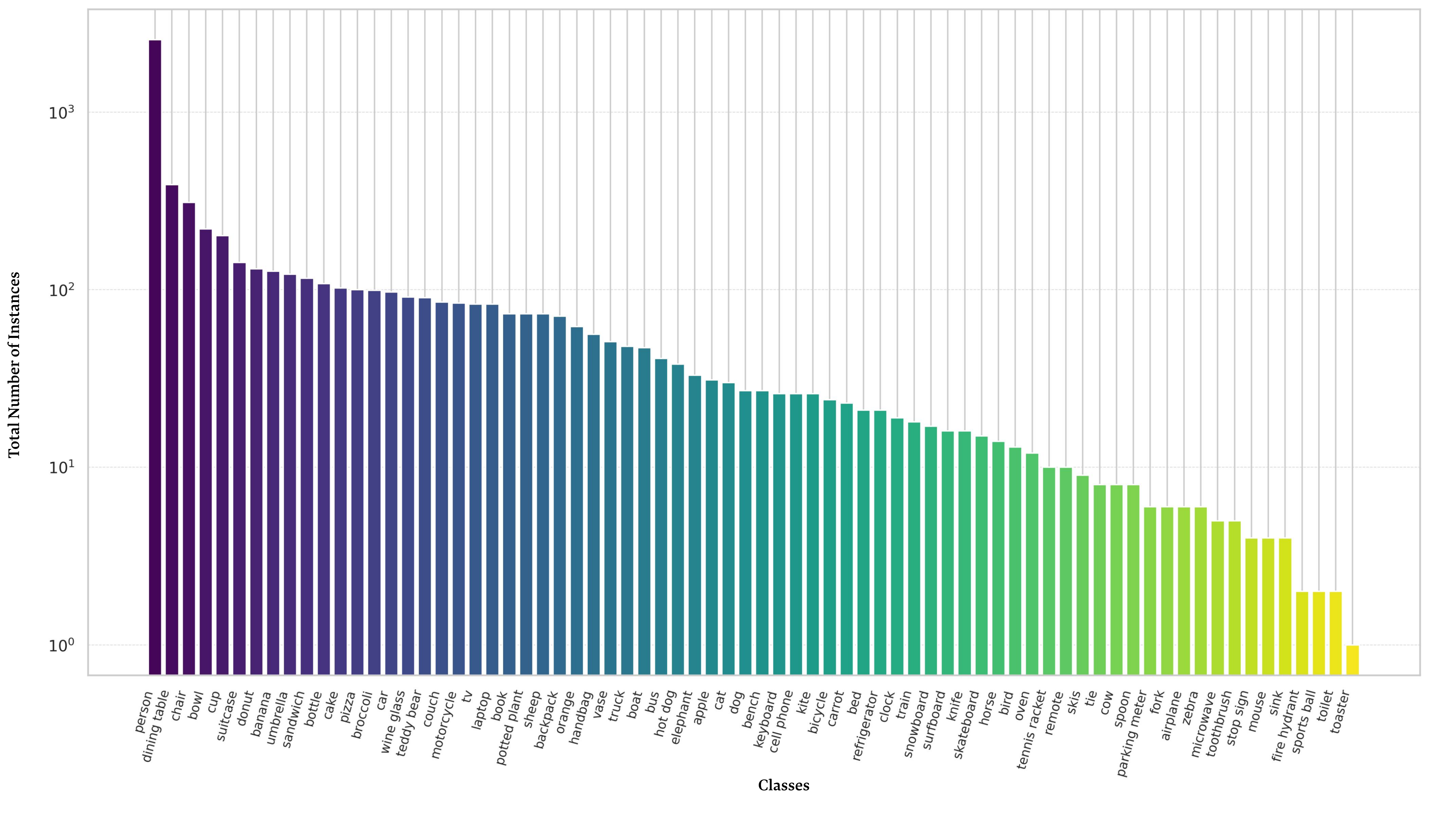}
    \vspace{-8mm}
    \caption{\rev{\textbf{Class distribution of the InstantStyle portion of the \datasetname{} dataset.}  By incorporating 53 novel object categories with varied frequencies, our dataset expands beyond prior scene sketch datasets and introduces realistic long-tailed challenges for instance segmentation. This diverse distribution is essential for evaluating model generalization across both common and rare object classes.}}
    \label{fig:instantstyle_dataset_distribution}
    \vspace{-5mm}
\end{figure*}

\begin{figure}[t]
    \centering
    \caption*{\textbf{InkScenes InstantStyle Variant: Scene Composition Complexity}} 
    \includegraphics[width=1\linewidth]  {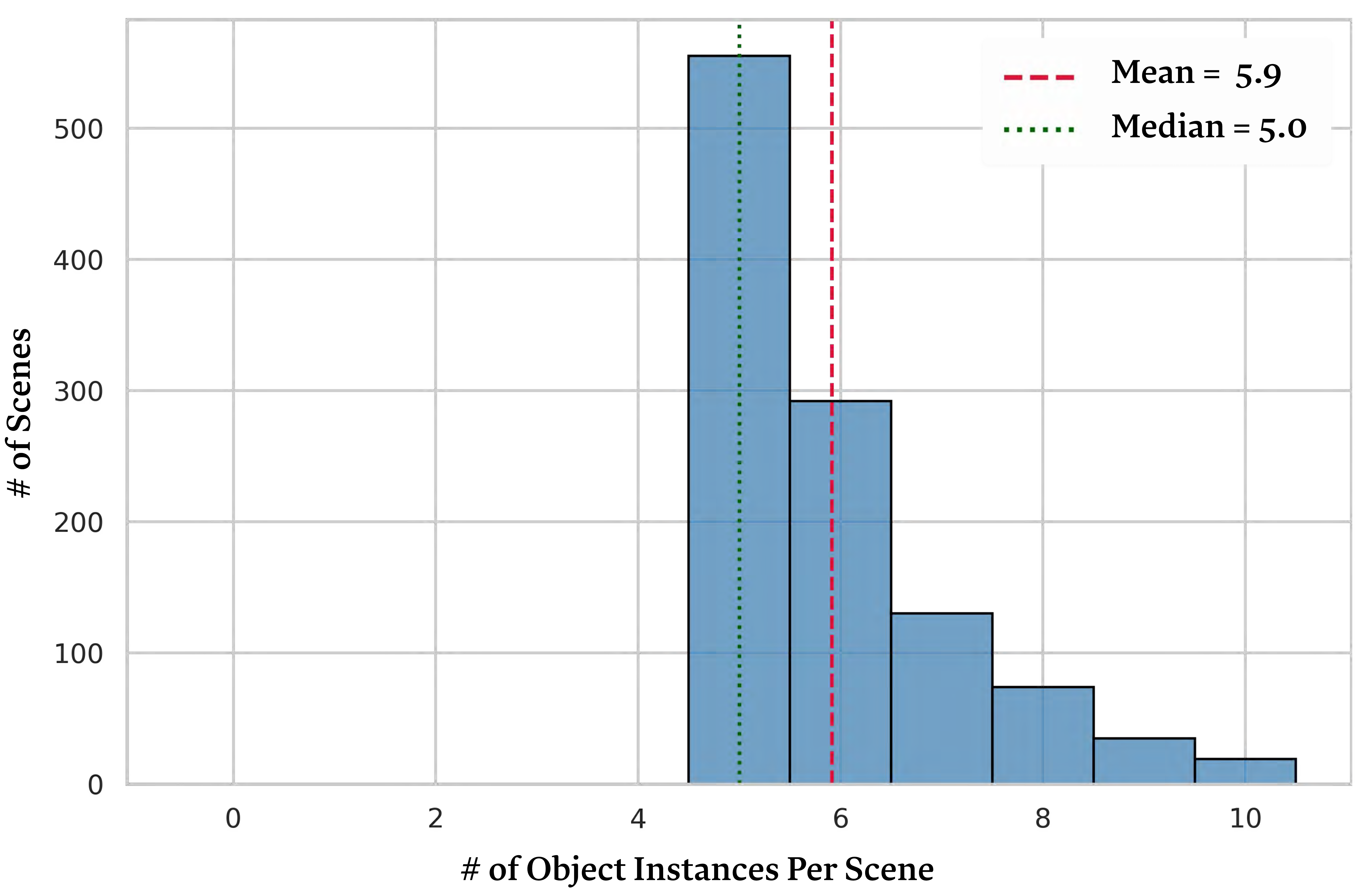}
    \caption{\rev{\textbf{Scene‑level complexity of the InstantStyle portion of \datasetname{}. }
    Each bar counts how many sketches contain a given number of object instances (background excluded). Most scenes cluster around 5–6 objects (median$\thickapprox$5, mean$\thickapprox$6), with a modest long‑tail of denser compositions up to ten objects.  }
    }
    \label{fig:scene_complexity}
\end{figure}

\rev{In this section, we present detailed information about our benchmark dataset, \textit{\datasetname{}}. While the CLIPasso variants and SketchAgent styles of our \datasetname{} dataset contain the same set of classes and scene complexity as original SketchyScene, our InstantStyle dataset contributes new categories and scene layouts. Figure~\ref{fig:dataset_distribution} shows the distribution of categories across the entire dataset, grouped in high-level semantic categories. Additionally, we show the InstantStyle dataset's scene complexity in Figure~\ref{fig:scene_complexity}, and the class distribution in Figure ~\ref{fig:instantstyle_dataset_distribution}.
The full list of categories in our \datasetname{} dataset is provided in Table~\ref{tab:class_split}, grouped into base categories (those included in SketchyScene) and novel categories (unseen and not present in SketchyScene). We omit the "others" category from SketchyScene due to its ambiguity, which makes it unsuitable for generating our new synthetic components.}

\begin{table}[t]
\centering
\small
\vspace{-1mm}
\resizebox{0.95\columnwidth}{!}{%
\begin{tabular}{p{0.45\linewidth} | p{0.45\linewidth}}
\multicolumn{1}{c}{\textbf{Base Classes (45)}} & \multicolumn{1}{c}{\textbf{Novel Classes (53)}} \\
\toprule
airplane\newline
apple\newline
balloon\newline
banana\newline
basket\newline
bee\newline
bench\newline
bicycle\newline
bird\newline
bottle\newline
bucket\newline
bus\newline
butterfly\newline
car\newline
cat\newline
chair\newline
chicken\newline
cloud\newline
cow\newline
cup\newline
dinnerware\newline
dog\newline
duck\newline
fence\newline
flower\newline
grape\newline
grass\newline
horse\newline
house\newline
moon\newline
mountain\newline
people/person\newline
picnic rug\newline
pig\newline
rabbit\newline
road\newline
sheep\newline
sofa\newline
star\newline
street lamp\newline
sun\newline
table\newline
tree\newline
truck\newline
umbrella\newline

&
sandwich\newline
refrigerator\newline
couch\newline
pizza\newline
laptop\newline
bed\newline
mouse\newline
toilet\newline
orange\newline
toaster\newline
kite\newline
cell phone\newline
cake\newline
carrot\newline
parking meter\newline
tv\newline
knife\newline
remote\newline
train\newline
tie\newline
vase\newline
potted plant\newline
clock\newline
sports ball\newline
handbag\newline
fire hydrant\newline
wine glass\newline
elephant\newline
skateboard\newline
keyboard\newline
teddy bear\newline
skis\newline
backpack\newline
spoon\newline
book\newline
stop sign\newline
broccoli\newline
zebra\newline
donut\newline
sink\newline
surfboard\newline
snowboard\newline
motorcycle\newline
suitcase\newline
dining table\newline
boat\newline
fork\newline
microwave\newline
oven\newline
bowl\newline
tennis racket\newline
toothbrush\newline
hot dog
\\
\bottomrule
\end{tabular}
}
% \vspace{-2mm}
\caption{\textbf{Full List of Classes in \datasetname{} Dataset. } The Base Classes are the same as SketchyScene\shortcite{Zou18SketchyScene} dataset. The Novel Classes are  introduced by our InstantStyle dataset. }
\label{tab:class_split}
\vspace{-8mm}
\end{table}

\section{Performance on Human-Drawn Sketches}

\rev{To ensure the practical utility of our method, it is important that it generalizes beyond synthetic datasets to real-world, human-drawn sketches. In the main paper, we evaluate performance on one such dataset introduced by Zhang et al.~\shortcite{zhang2018CBSC}. In this section, we extend our evaluation to two additional datasets containing human-drawn sketches, which required more careful consideration in both setup and analysis.}

\subsection{Performance on FSCOCO-Seg}

\begin{figure*}
    \centering
    \setlength{\tabcolsep}{1.5pt}
    {\small
    \resizebox{0.88\textwidth}{!}{ 
    \begin{tabular}{c c @{\hskip 10pt} c c}
         Sketch & Segmentation &  Sketch & Segmentation \\
        \frame{\includegraphics[width=0.22\linewidth]{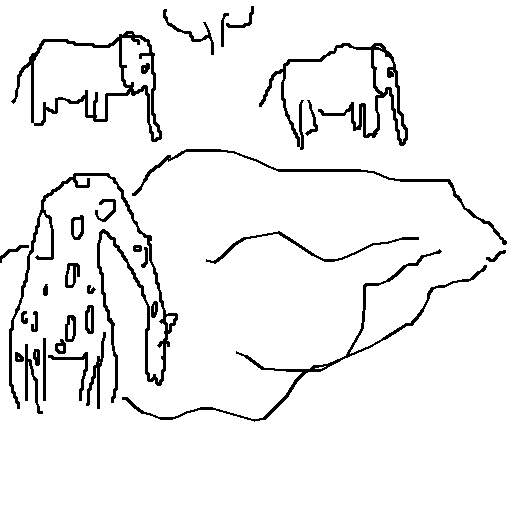}} &
        \frame{\includegraphics[width=0.22\linewidth]{figs_supp/FSCOCO_good_examples/000000019239_final_segment.png}} &
         \frame{\includegraphics[width=0.22\linewidth]{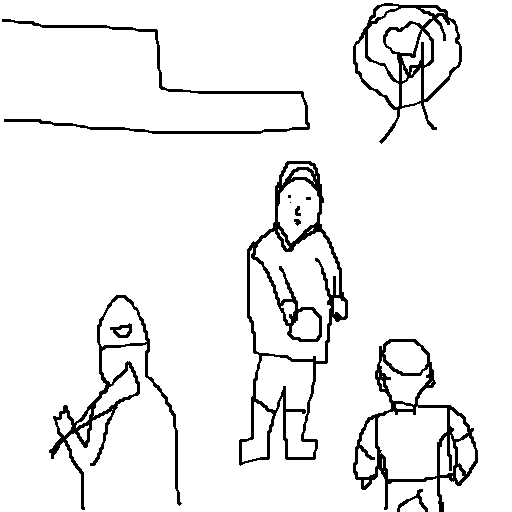}} &
        \frame{\includegraphics[width=0.22\linewidth]{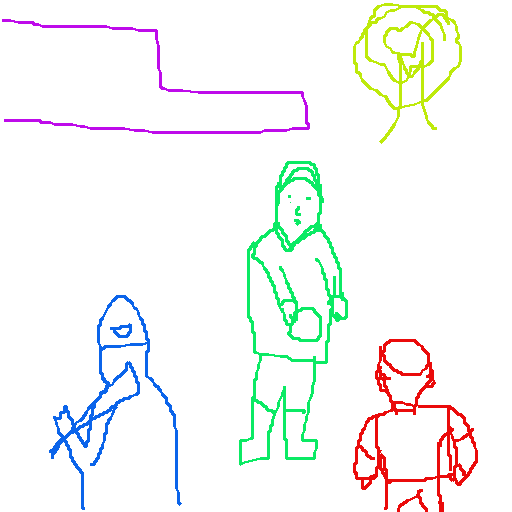}} \\
        
        \frame{\includegraphics[width=0.22\linewidth]{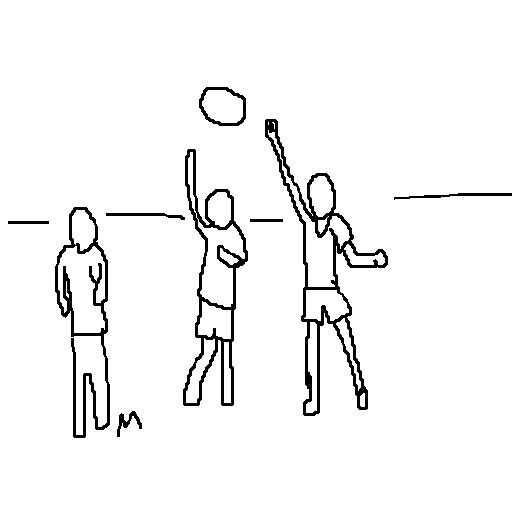}} &
        \frame{\includegraphics[width=0.22\linewidth]{figs_supp/FSCOCO_good_examples/000000111000_final_segment.png}} &
         \frame{\includegraphics[width=0.22\linewidth]{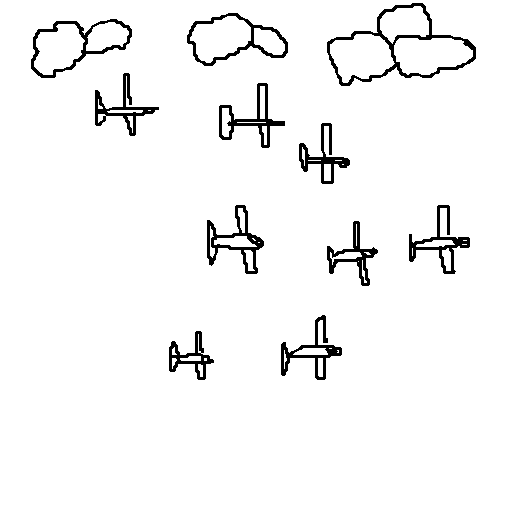}} &
        \frame{\includegraphics[width=0.22\linewidth]{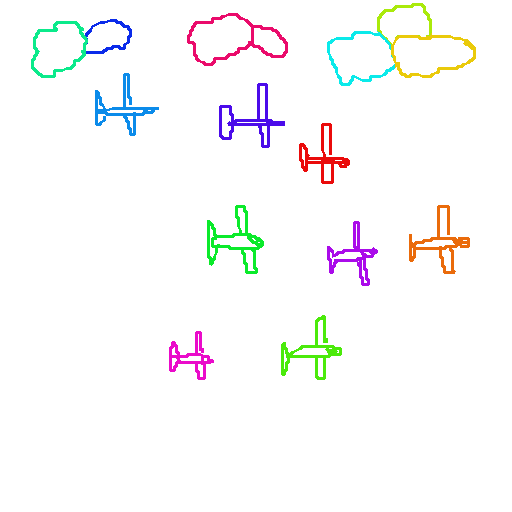}} \\
        
        \frame{\includegraphics[width=0.22\linewidth]{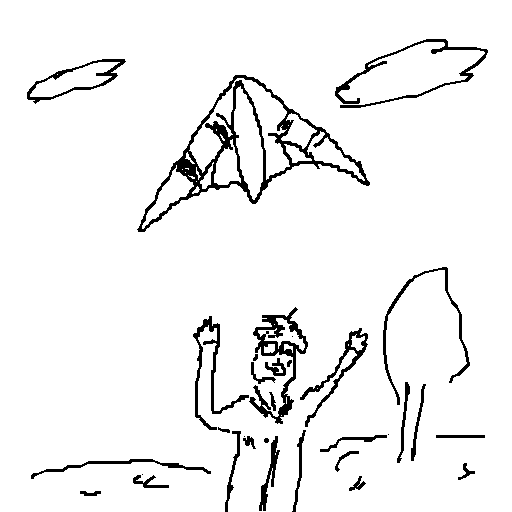}} &
        \frame{\includegraphics[width=0.22\linewidth]{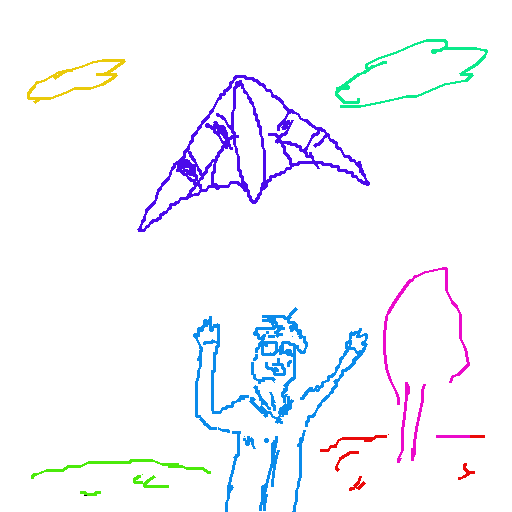}} &
         \frame{\includegraphics[width=0.22\linewidth]{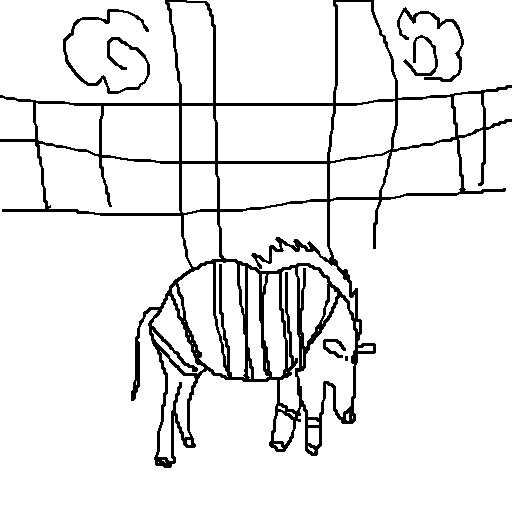}} &
        \frame{\includegraphics[width=0.22\linewidth]{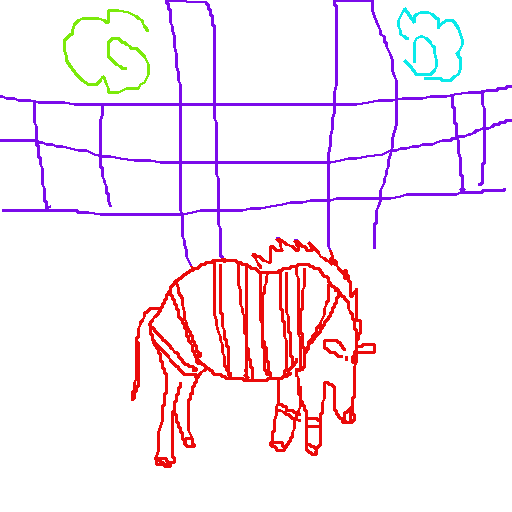}} \\
        
        \frame{\includegraphics[width=0.22\linewidth]{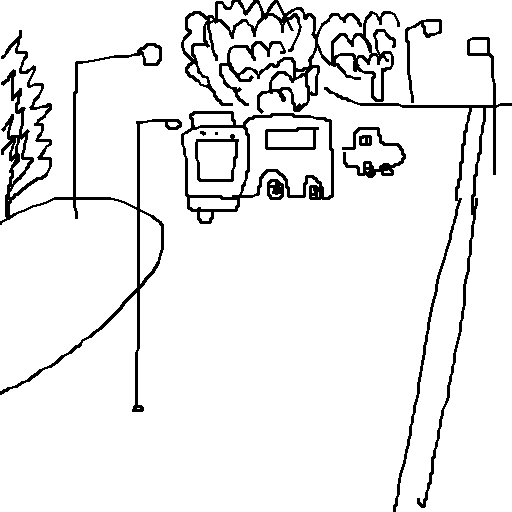}} &
        \frame{\includegraphics[width=0.22\linewidth]{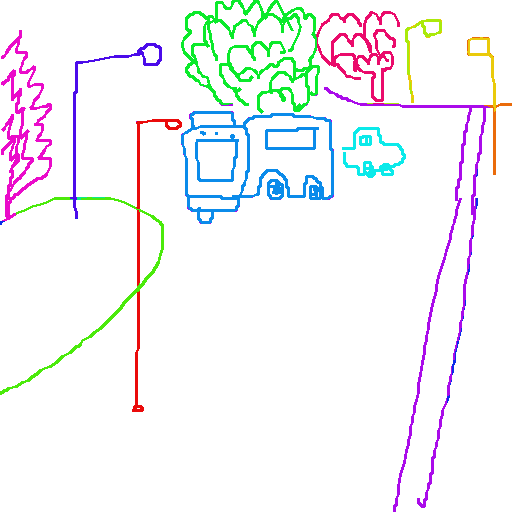}} &
         \frame{\includegraphics[width=0.22\linewidth]{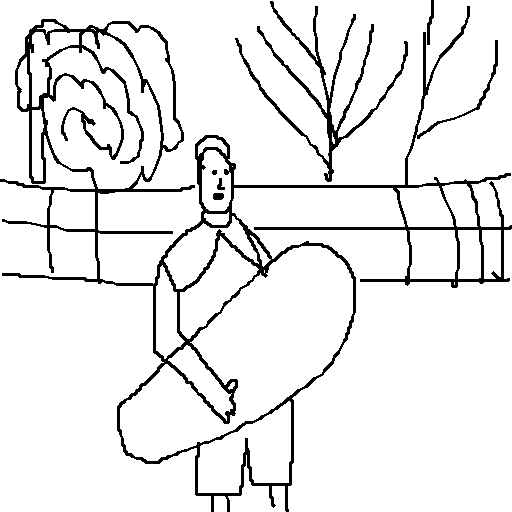}} &
        \frame{\includegraphics[width=0.22\linewidth]{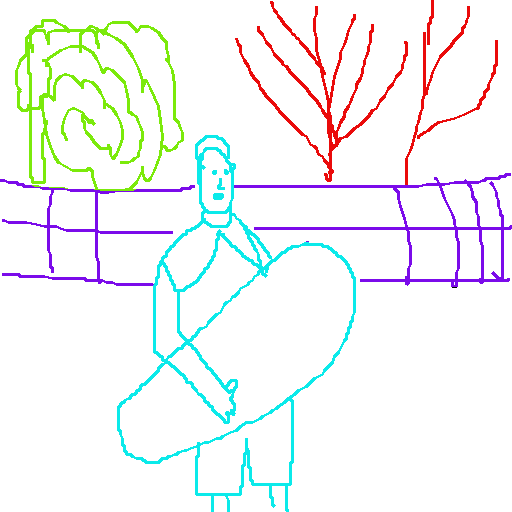}} \\
        
        \frame{\includegraphics[width=0.22\linewidth]{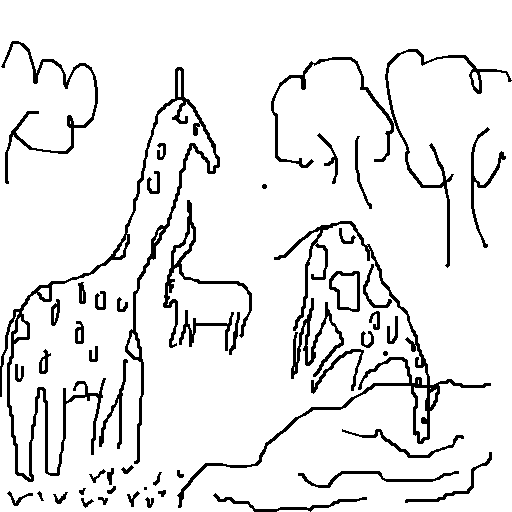}} &
        \frame{\includegraphics[width=0.22\linewidth]{figs_supp/FSCOCO_good_examples/000000477143_final_segment.png}} &
         \frame{\includegraphics[width=0.22\linewidth]{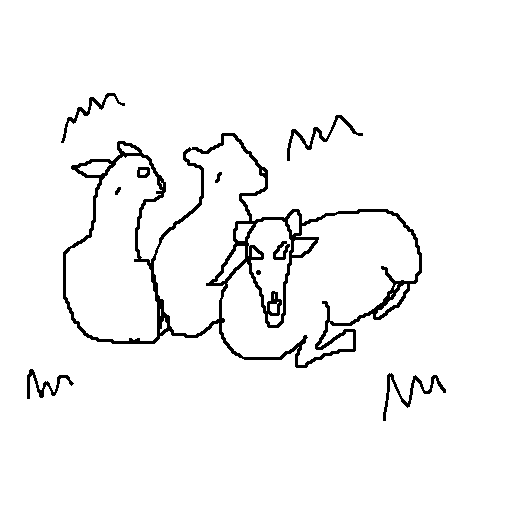}} &
        \frame{\includegraphics[width=0.22\linewidth]{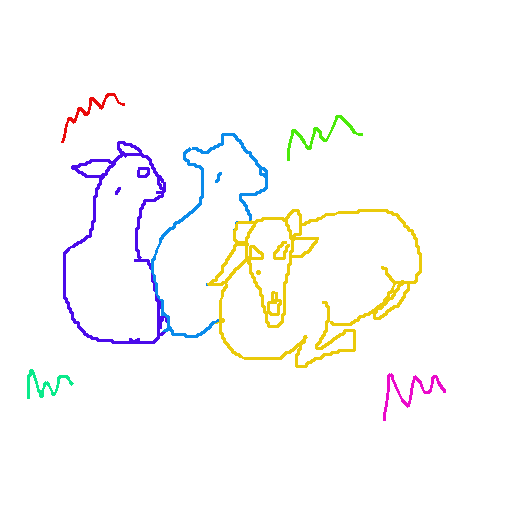}} \\
    \end{tabular}
    }}
    \caption{\rev{\textbf{FSCOCO-Seg Performance. }We qualitatively evaluate our method on scenes from the FSCOCO-Seg dataset. Overall, the segmentations are visually accurate and align well with object boundaries in diverse, hand-drawn scenes. While our method performs well across a variety of object categories and layouts, we observe some artifacts in regions with heavy occlusion, where foreground and background objects may blend or be incompletely separated.}}
    \label{fig:fscoco_results}
\end{figure*}

\rev{FSCOCO-Seg~\cite{bourouis2024open} is a freehand scene sketch dataset comprising 500 sketches (originally from FS-COCO ~\cite{fscoco} dataset) with ground-truth semantic segmentation labels, but no instance-level annotations. To evaluate our method on this dataset, we convert our predicted instance segmentation outputs into semantic segmentation maps using the provided ground-truth labels.}
\rev{
Our method achieves a pixel accuracy of 0.85 and an IoU of 0.75, demonstrating strong generalization to freehand sketches. Figure~\ref{fig:fscoco_results} shows qualitative instance segmentation results on FSCOCO-Seg, highlighting the method’s ability to segment complete object instances even in abstract and loosely drawn scenes.}

\subsection{Performance on Product Design Sketches}
\begin{figure*}[t]
    \centering
    \setlength{\tabcolsep}{1.5pt} % spacing between columns
    {\small
    \resizebox{0.88\textwidth}{!}{ 
    \begin{tabular}{c c @{\hskip 10pt} c c}
        Sketch & Segmentation & Sketch & Segmentation \\
        \frame{\includegraphics[width=0.22\linewidth]{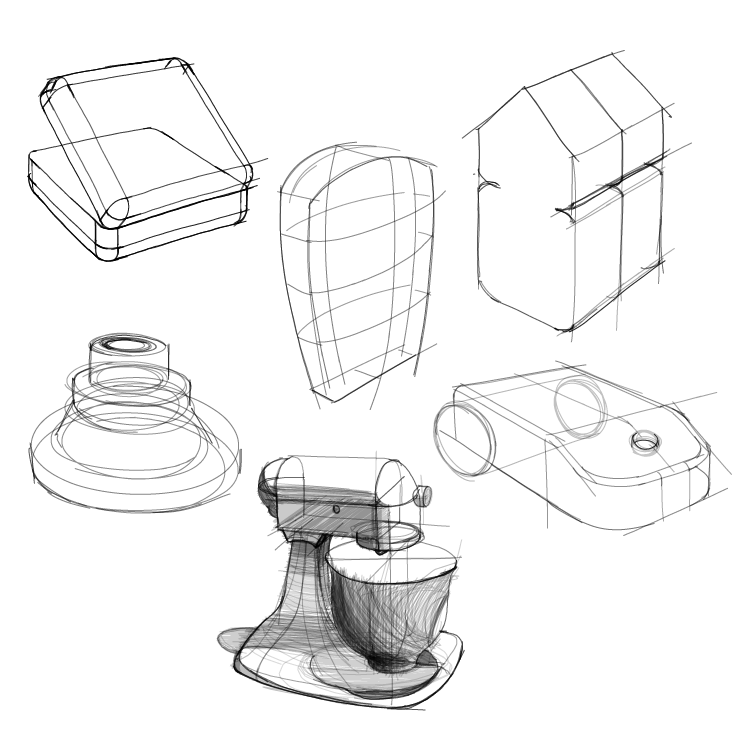}} &
        \frame{\includegraphics[width=0.22\linewidth]{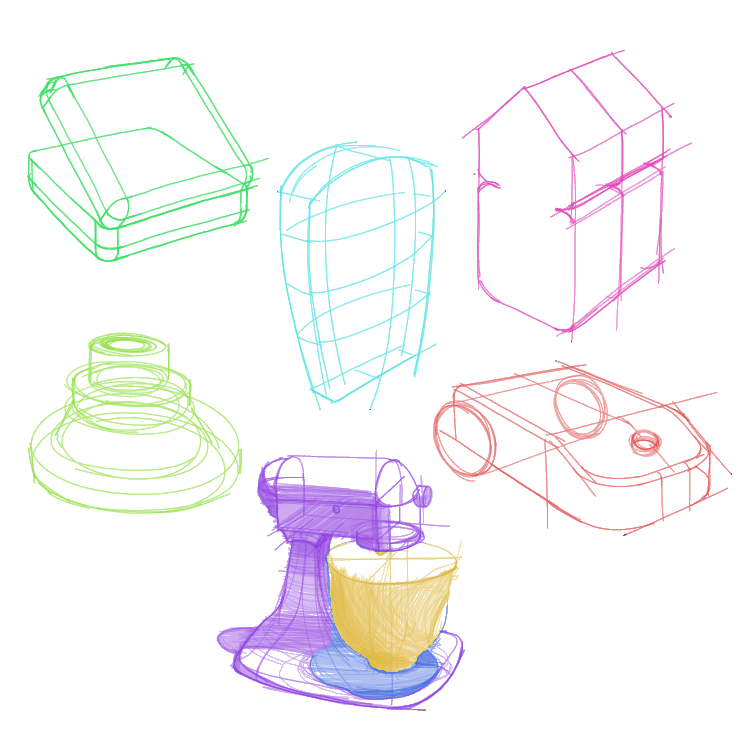}} &
        \frame{\includegraphics[width=0.22\linewidth]{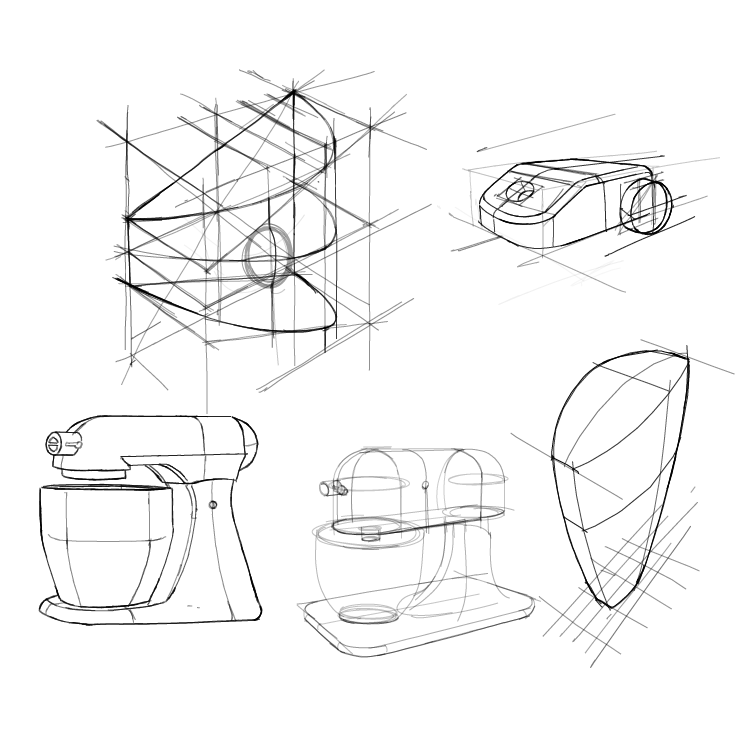}} &
        \frame{\includegraphics[width=0.22\linewidth]{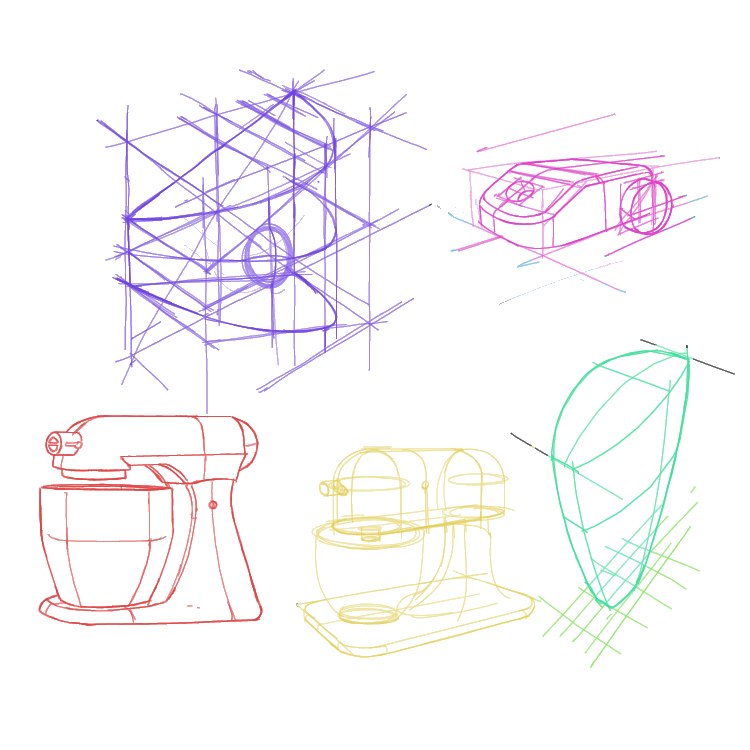}} 
        \\
        
        \frame{\includegraphics[width=0.22\linewidth]{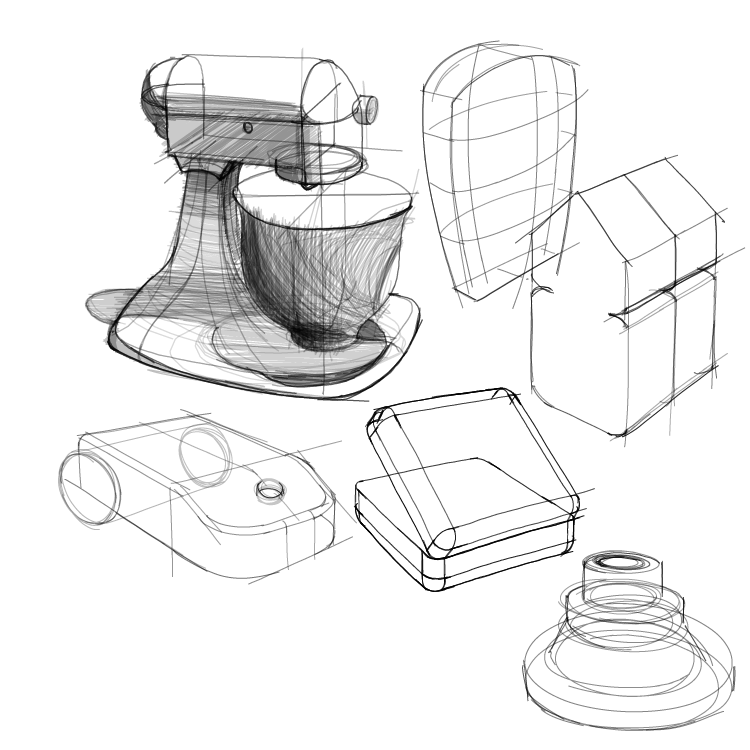}} &
        \frame{\includegraphics[width=0.22\linewidth]{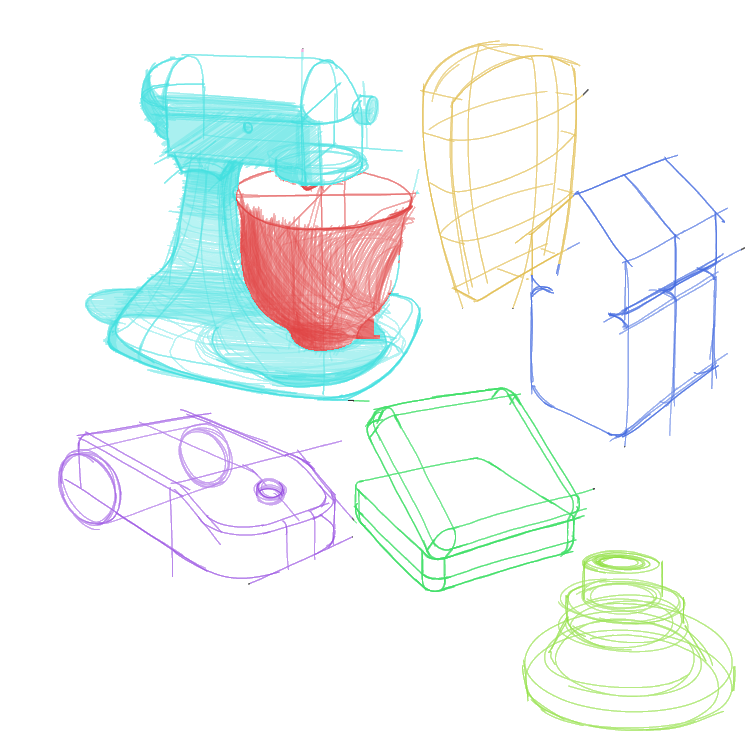}}& \frame{\includegraphics[width=0.22\linewidth]{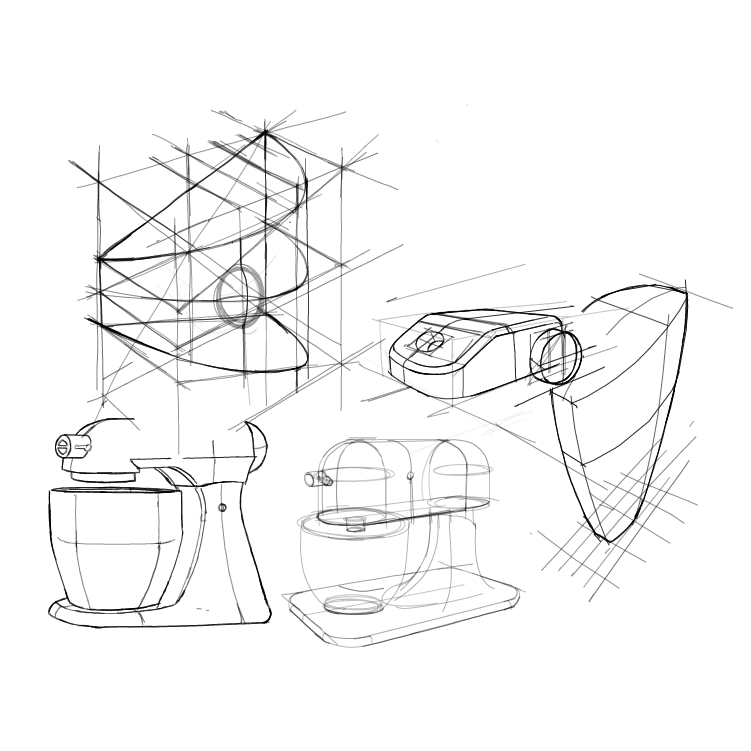}} &
        \frame{\includegraphics[width=0.22\linewidth]{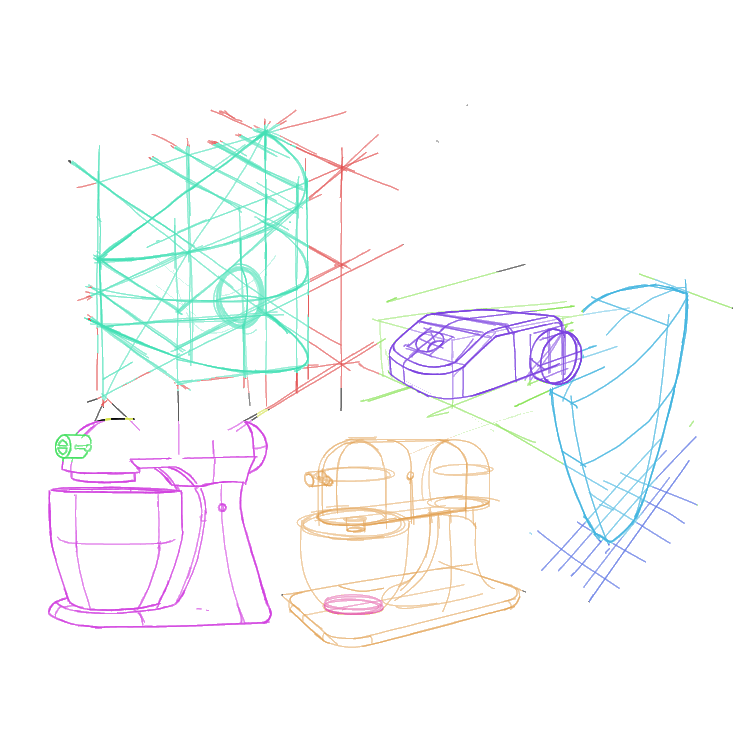}} 
        \\
        
        \frame{\includegraphics[width=0.22\linewidth]{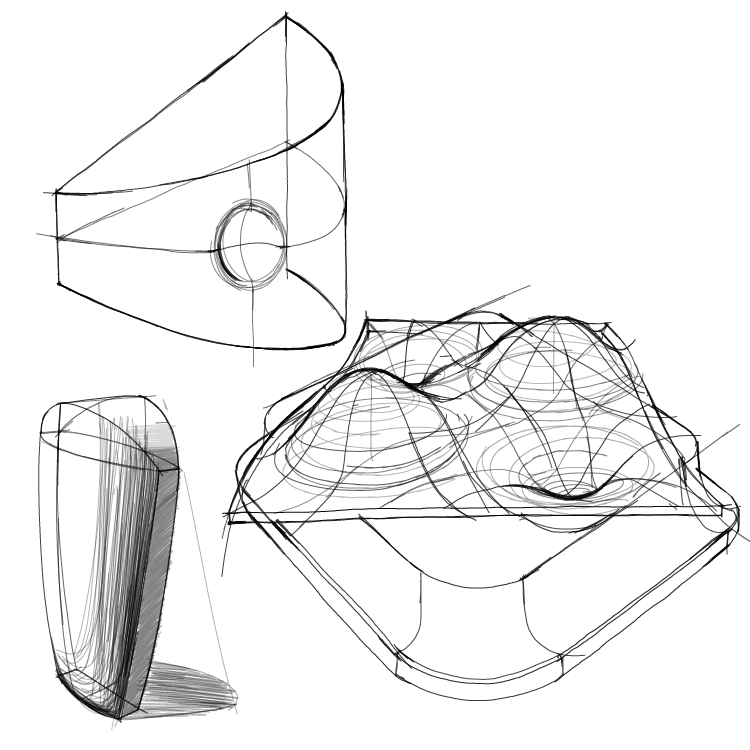}} &
        \frame{\includegraphics[width=0.22\linewidth]{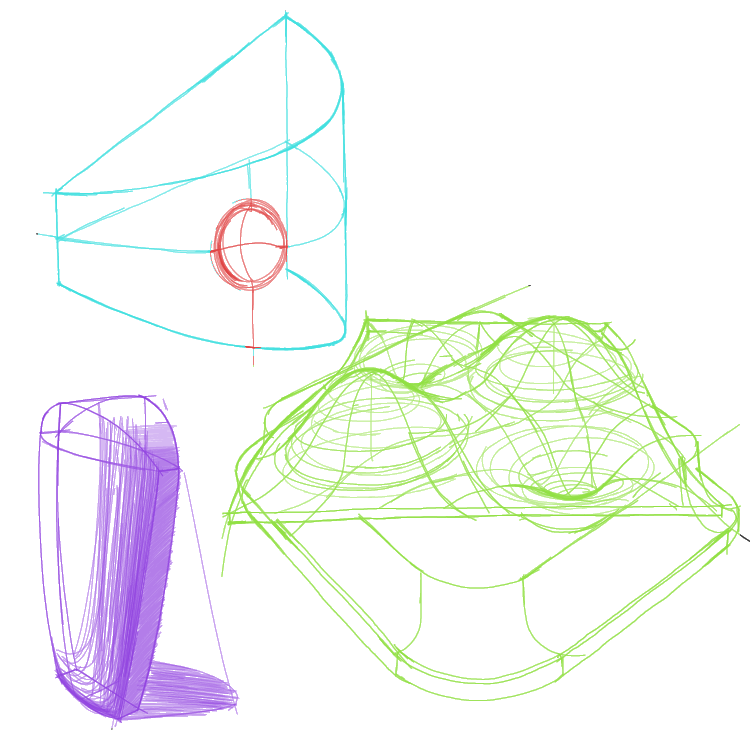}}& \frame{\includegraphics[width=0.22\linewidth]{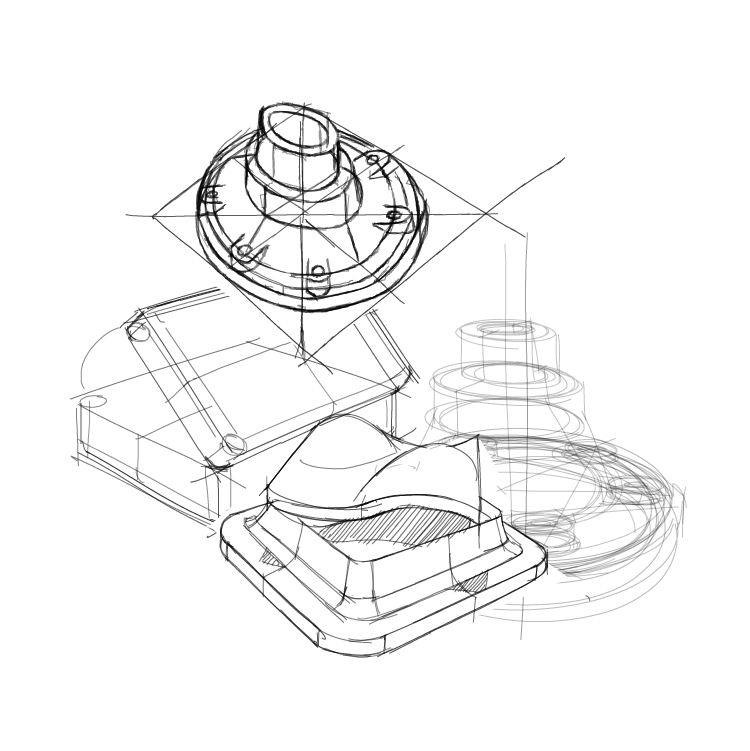}} &
        \frame{\includegraphics[width=0.22\linewidth]{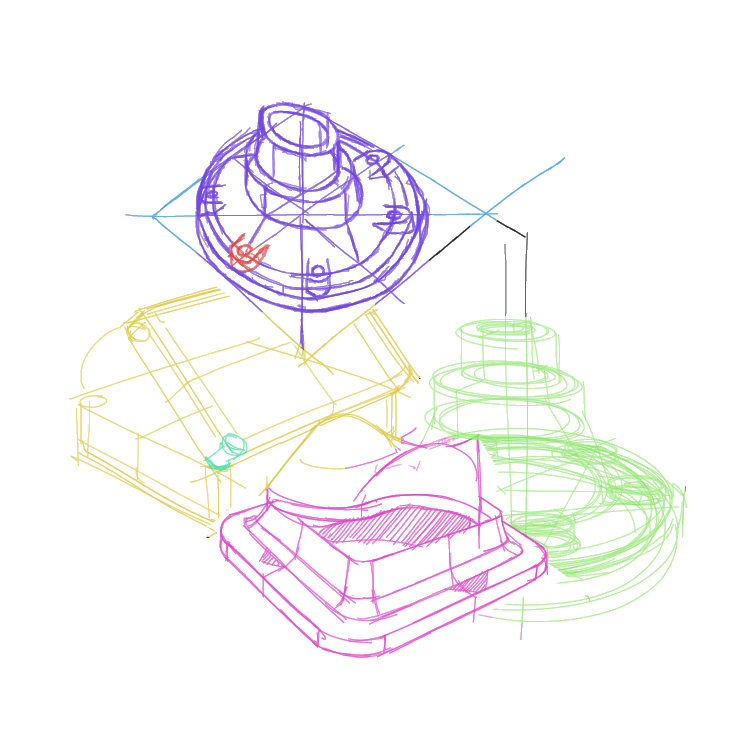}} 
        \\
        
        \frame{\includegraphics[width=0.22\linewidth]{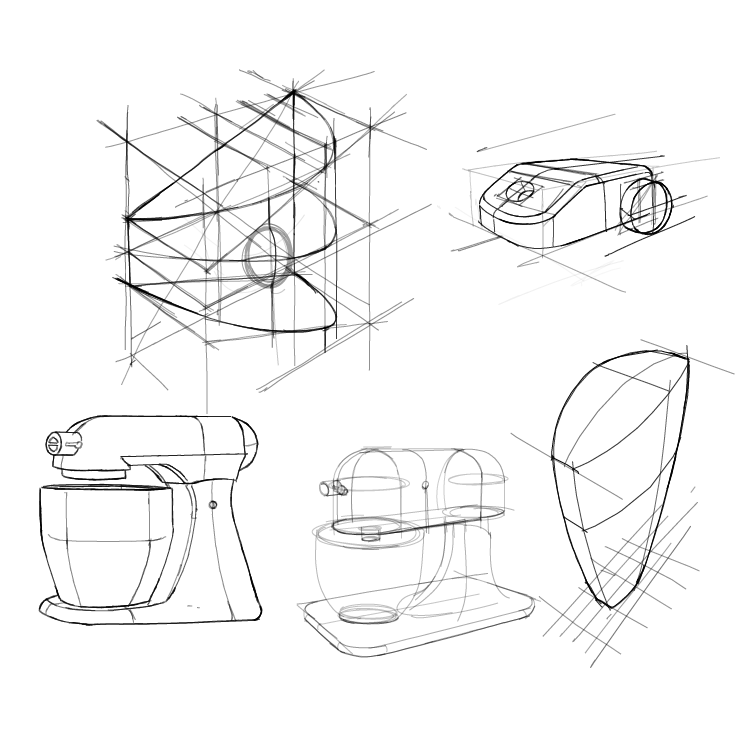}} &
        \frame{\includegraphics[width=0.22\linewidth]{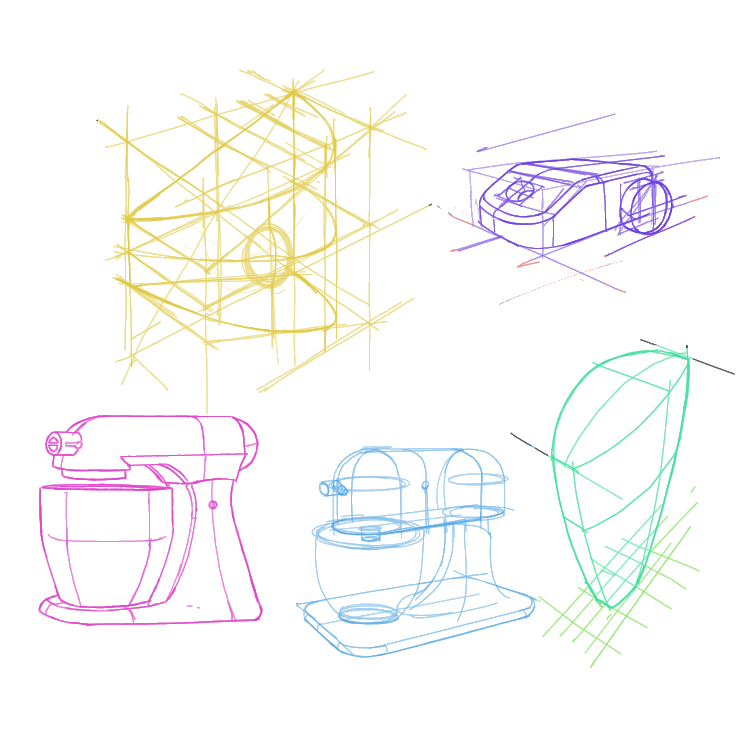}}& \frame{\includegraphics[width=0.22\linewidth]{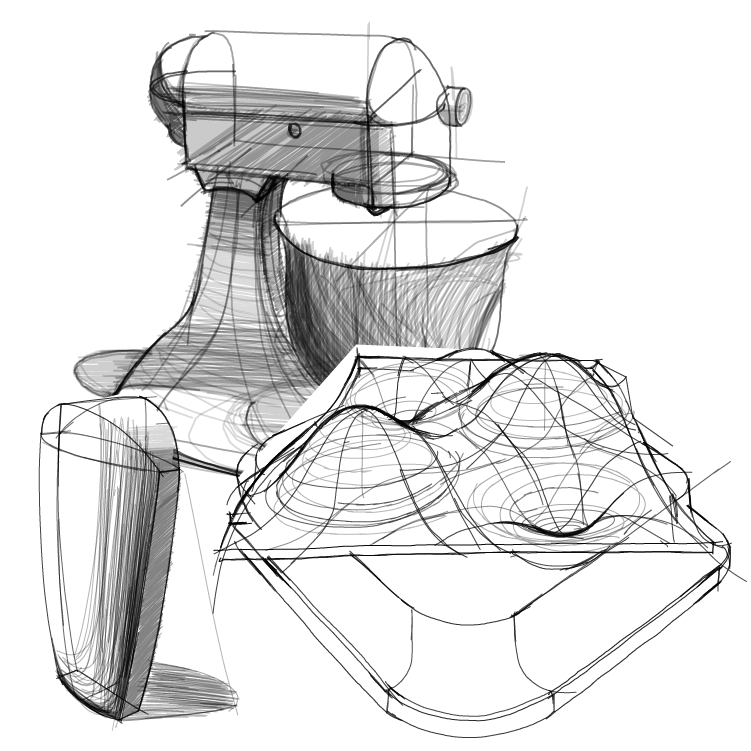}} &
        \frame{\includegraphics[width=0.22\linewidth]{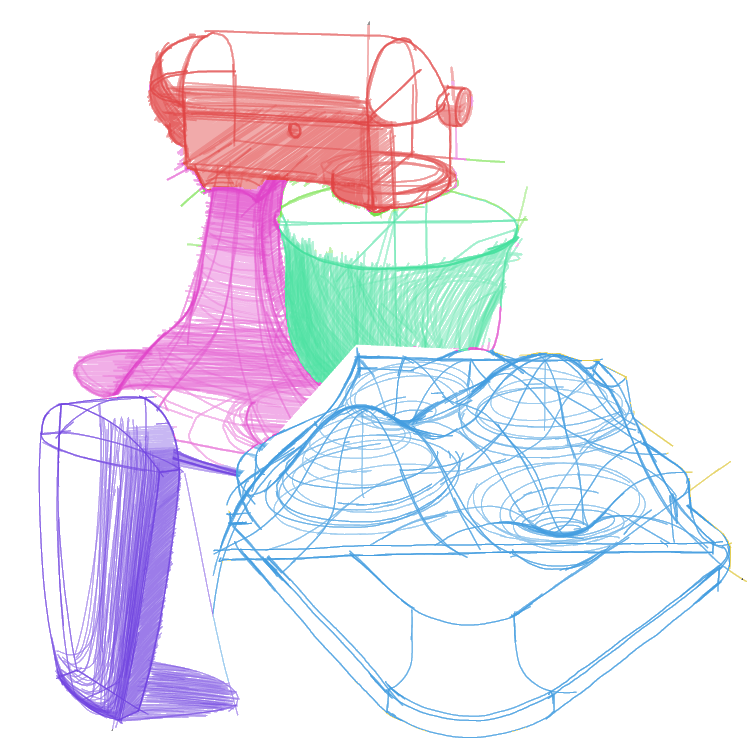}} 
        \\
        
        \frame{\includegraphics[width=0.22\linewidth]{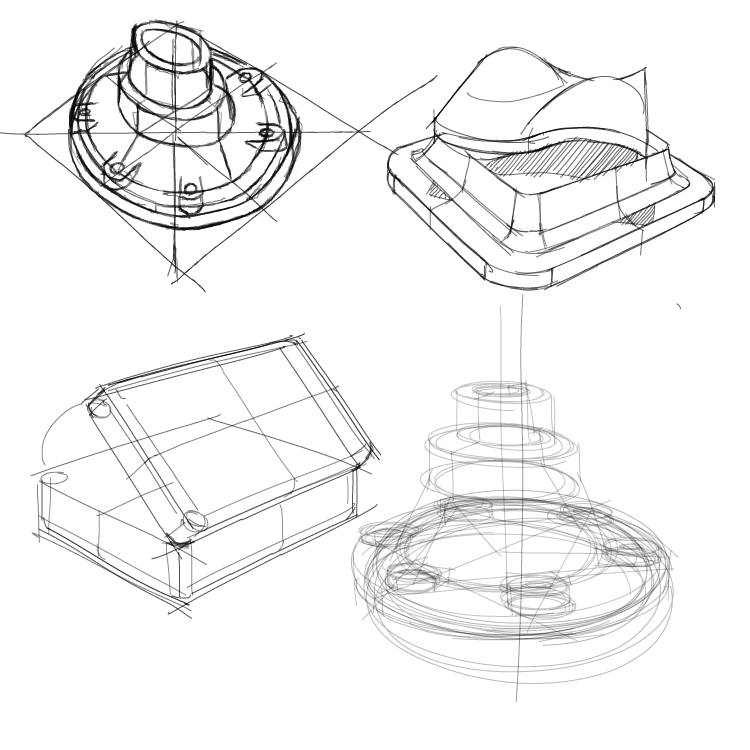}} &
        \frame{\includegraphics[width=0.22\linewidth]{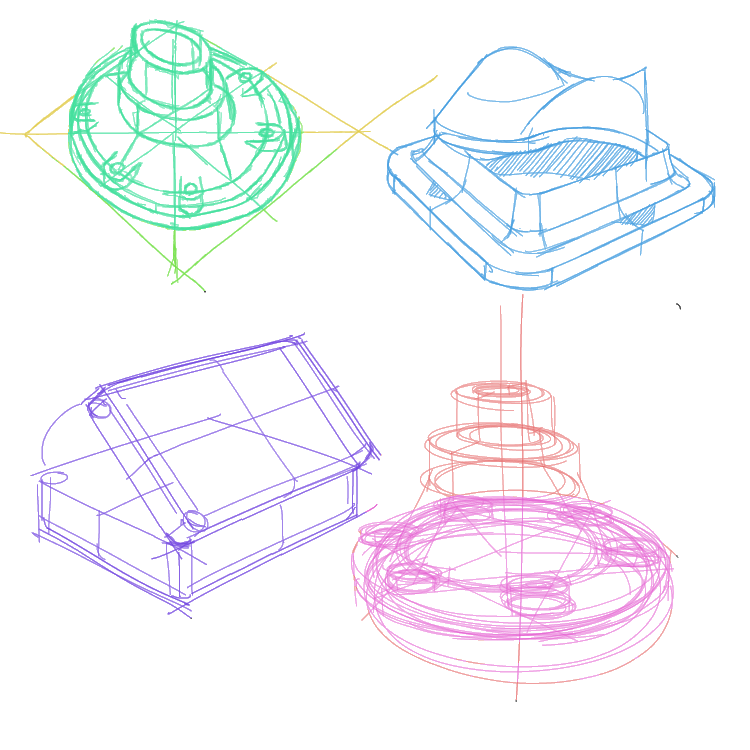}} &
        \frame{\includegraphics[width=0.22\linewidth]{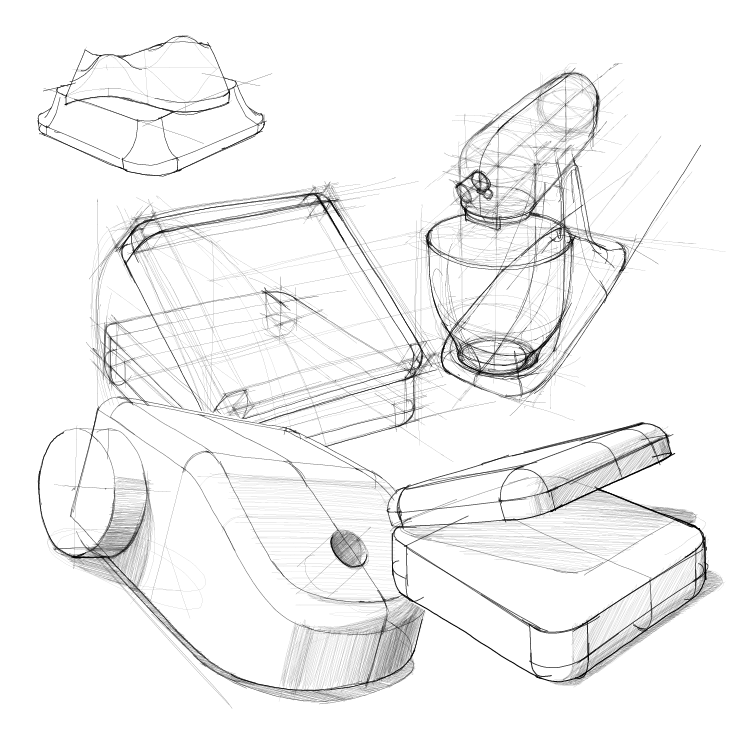}} &
        \frame{\includegraphics[width=0.22\linewidth]{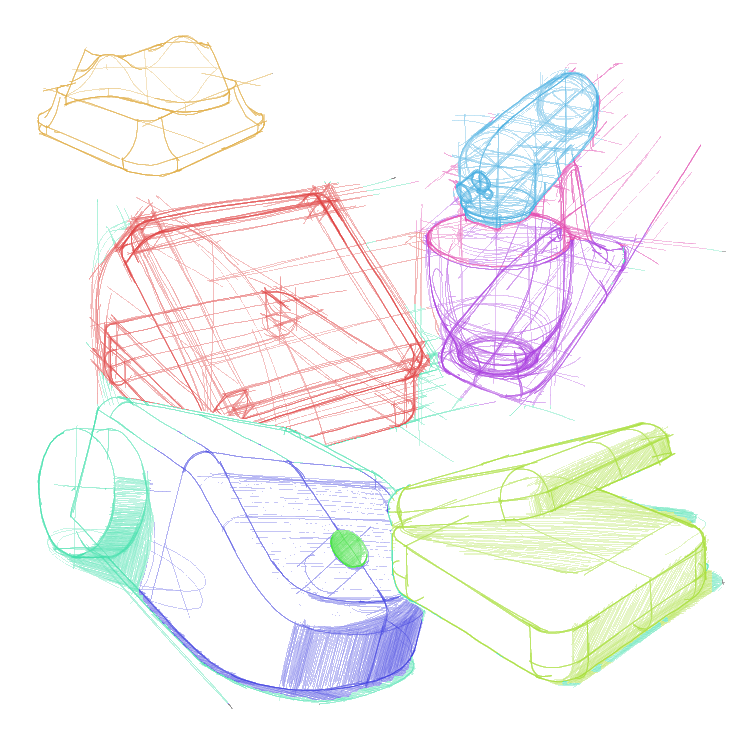}} \\
    \end{tabular}
    }}
    \caption{\rev{\textbf{OpenSketch Performance. }We qualitatively evaluate our method on manually composed scenes from the OpenSketch dataset. For clearer occlusion handling, we manually added coarse white backgrounds behind foreground objects. Our method segments individual objects reasonably well, especially when sketches are not heavily cluttered with construction lines. For example, segmentation struggles with the background object in the second example of row 2, which contains dense construction lines.}}
    \label{fig:opensketch_results}
\end{figure*}

\rev{We do additional qualitative evaluation on product design sketches from OpenSketch dataset ~\cite{OpenSketch19} in Fig. ~\ref{fig:opensketch_results}. Since OpenSketch contains product design sketch of single, individual objects, we manually created scenes by combining multiple object sketches. While our method generally succeeds in segmenting individual objects across scenes, we observe limitations such as artifacts on objects with long, thick construction lines, and segmenting objects apart from their construction lines.}

\subsection{Performance on Zhang \etal}
\begin{figure*}[t]
    \centering
    \setlength{\tabcolsep}{1.5pt} % spacing between columns
    \resizebox{0.88\textwidth}{!}{ 
    {\small
    \begin{tabular}{c c @{\hskip 10pt} c c }
        Sketch & Segmentation & Sketch & Segmentation \\
        \frame{\includegraphics[width=0.22\linewidth]{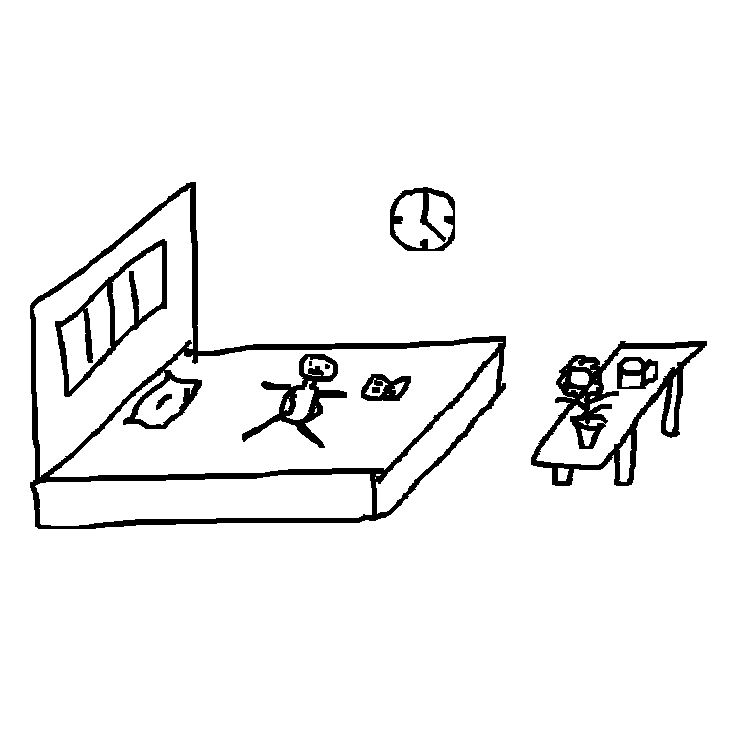}} &
        \frame{\includegraphics[width=0.22\linewidth]{figs_supp/CBSC_good_results/0114153955_final_segment.png}} &
        \frame{\includegraphics[width=0.22\linewidth]{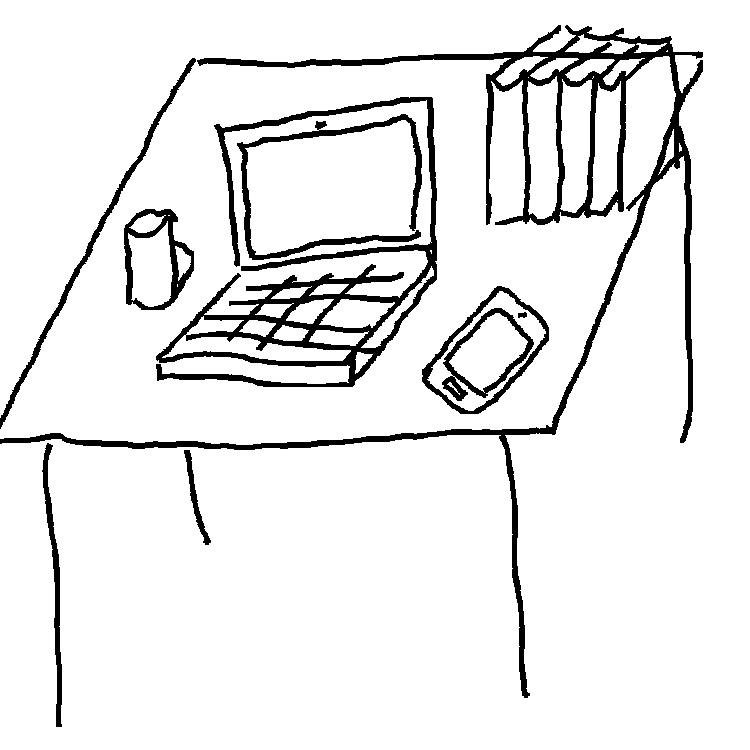}} &
        \frame{\includegraphics[width=0.22\linewidth]{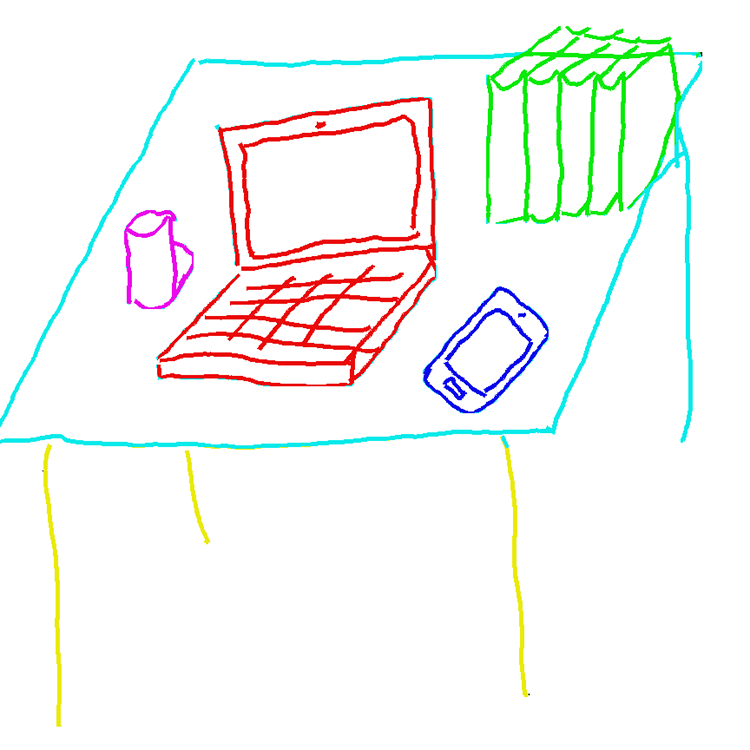}} \\

        \frame{\includegraphics[width=0.22\linewidth]{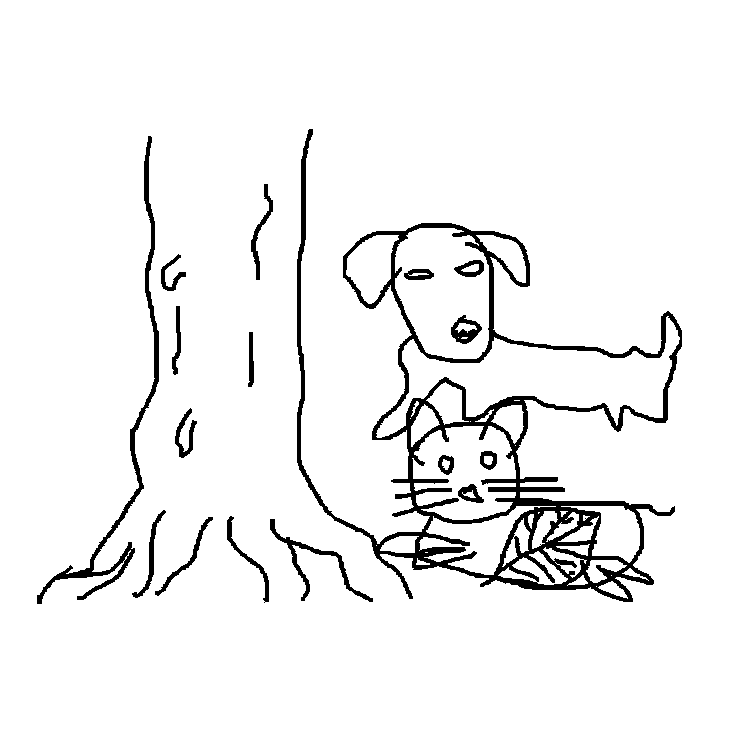}} &
        \frame{\includegraphics[width=0.22\linewidth]{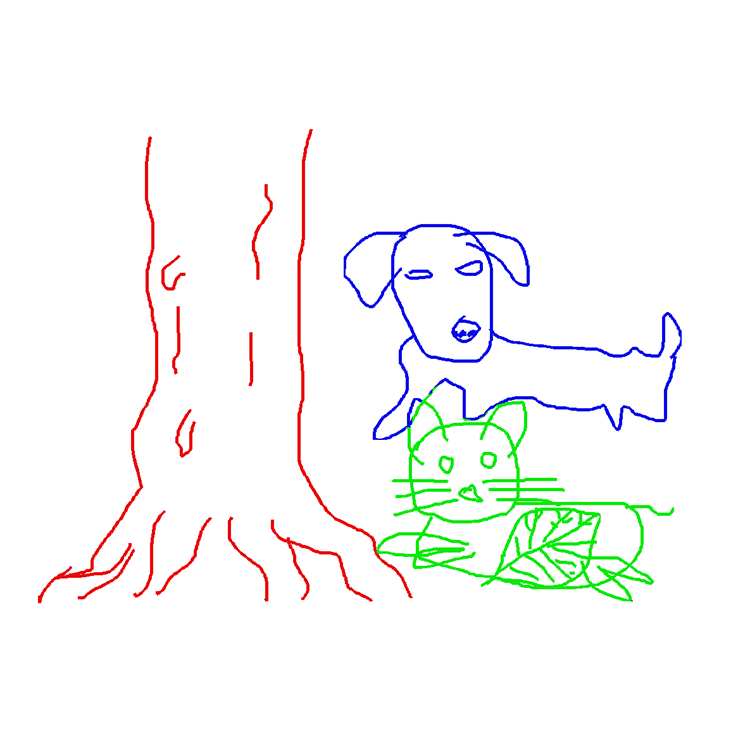}} &
        \frame{\includegraphics[width=0.22\linewidth]{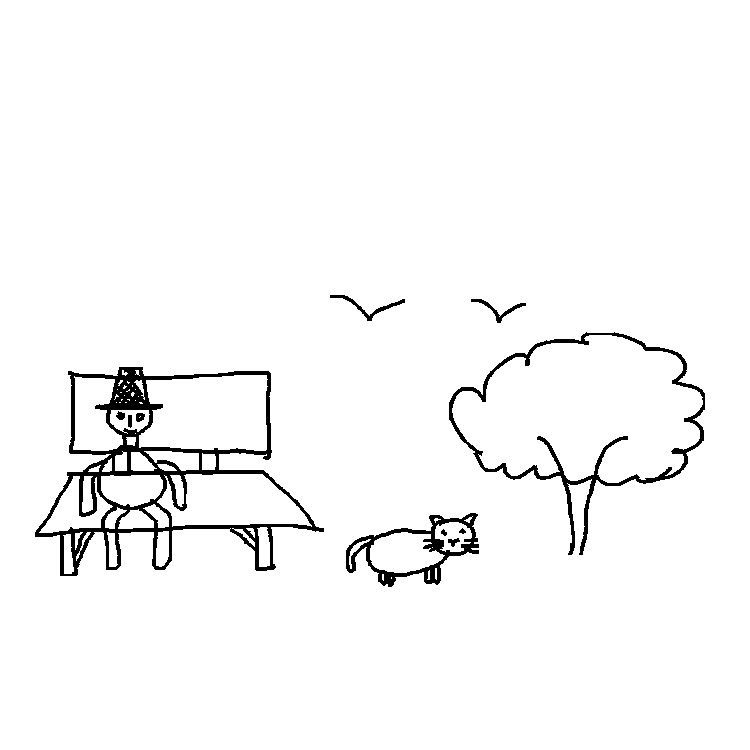}} &
        \frame{\includegraphics[width=0.22\linewidth]{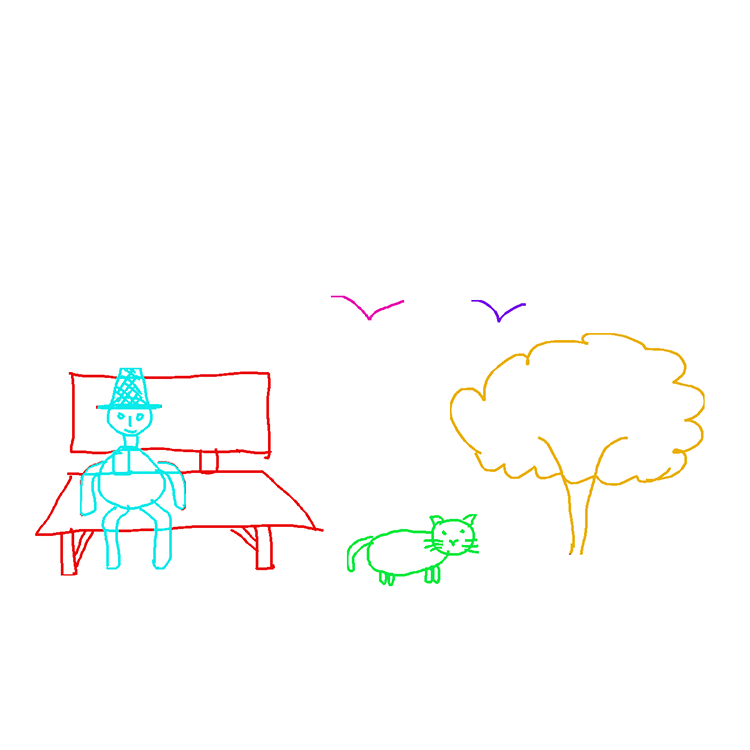}} 
        \\
        
        \frame{\includegraphics[width=0.22\linewidth]{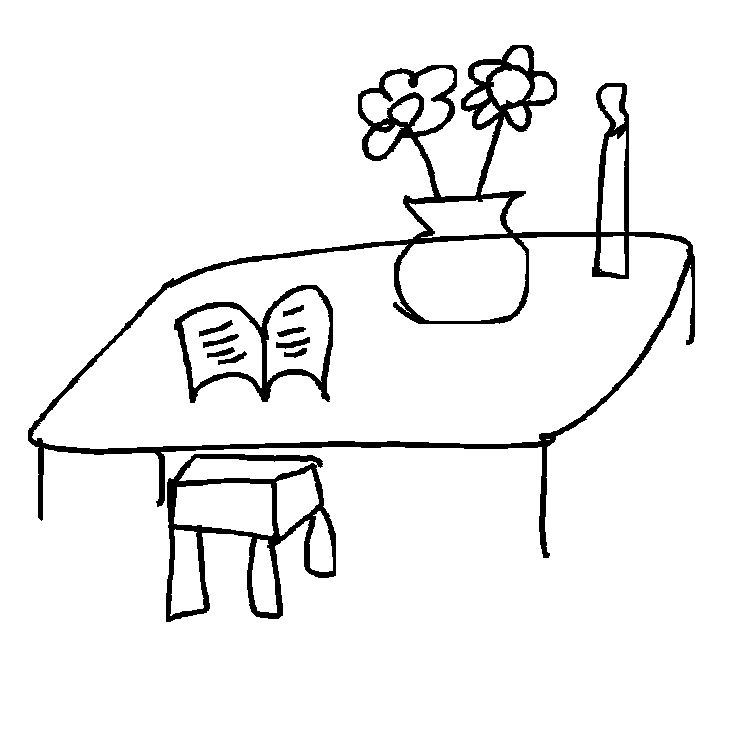}} &
        \frame{\includegraphics[width=0.22\linewidth]{figs_supp/CBSC_good_results/0115124951_final_segment.png}} &
        \frame{\includegraphics[width=0.22\linewidth]{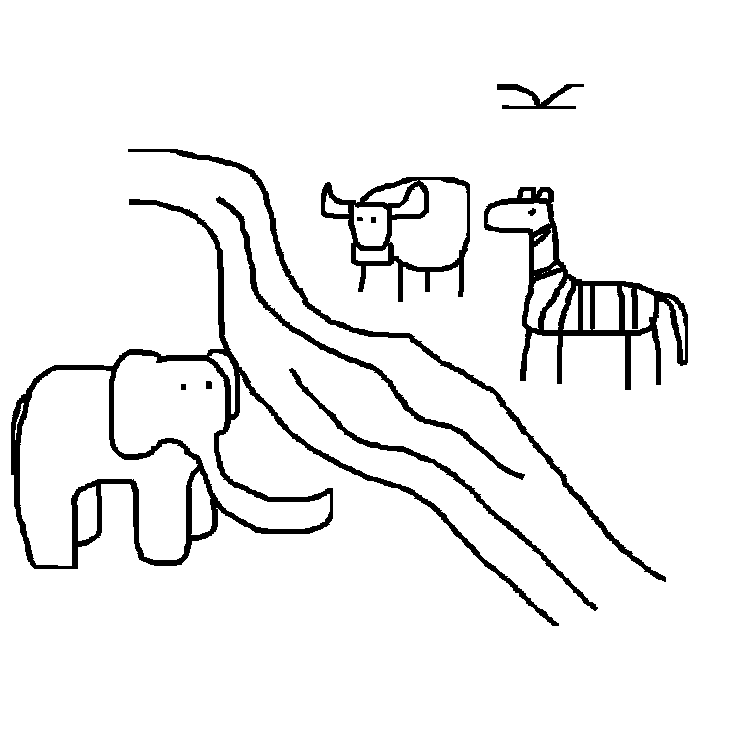}} &
        \frame{\includegraphics[width=0.22\linewidth]{figs_supp/CBSC_good_results/0115173307_final_segment.png}} 
        \\
        
        \frame{\includegraphics[width=0.22\linewidth]{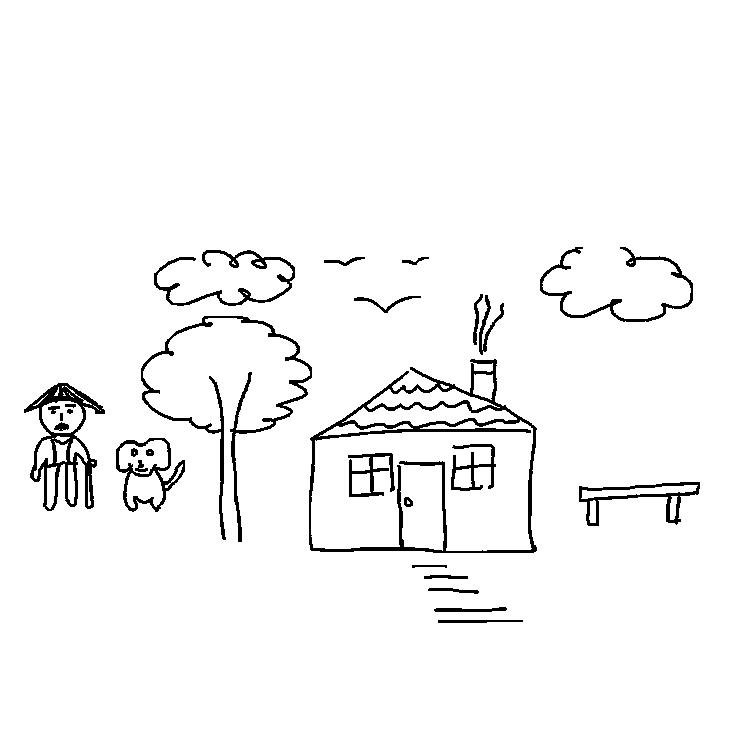}} &
        \frame{\includegraphics[width=0.22\linewidth]{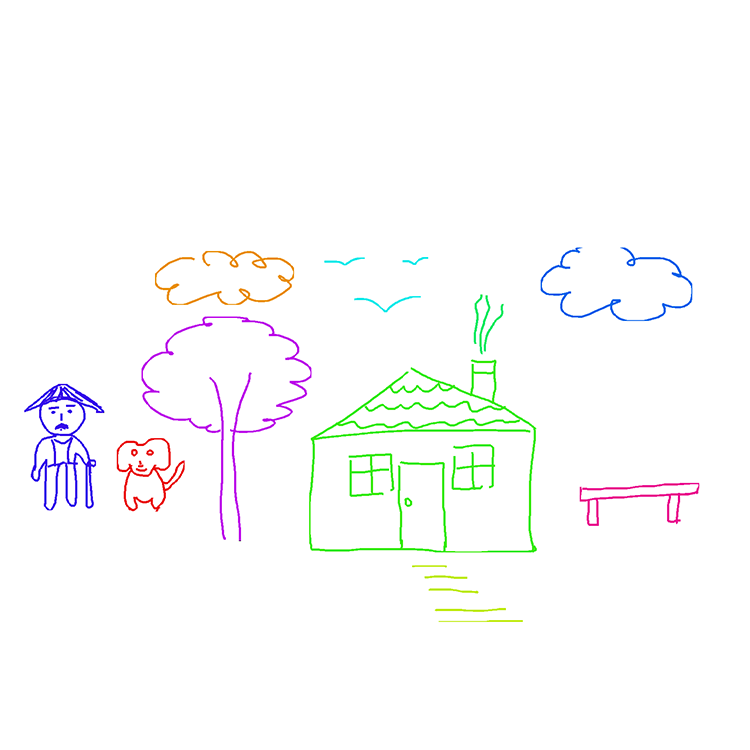}} &
        \frame{\includegraphics[width=0.22\linewidth]{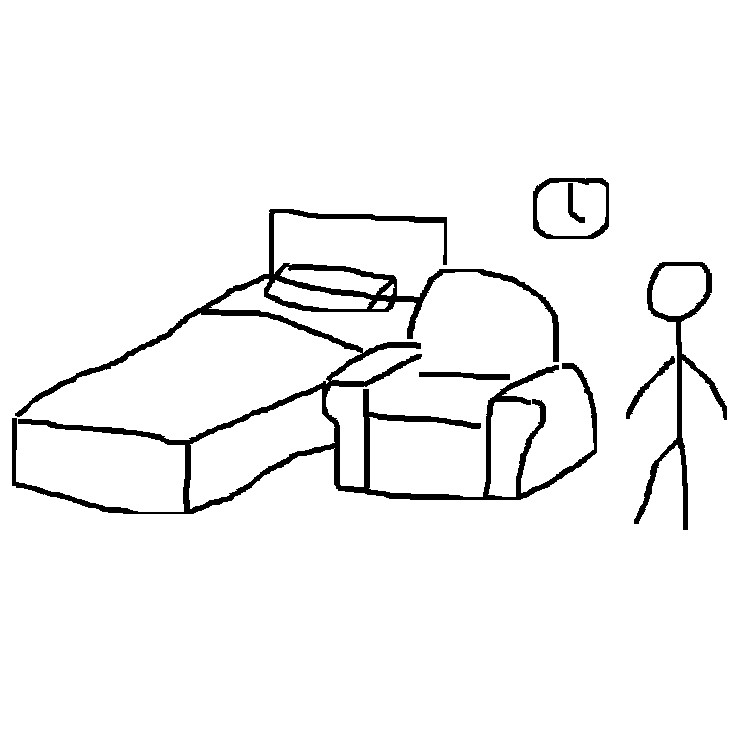}} &
        \frame{\includegraphics[width=0.22\linewidth]{figs_supp/CBSC_good_results/0115170719_final_segment}} 
        \\

        \frame{\includegraphics[width=0.22\linewidth]{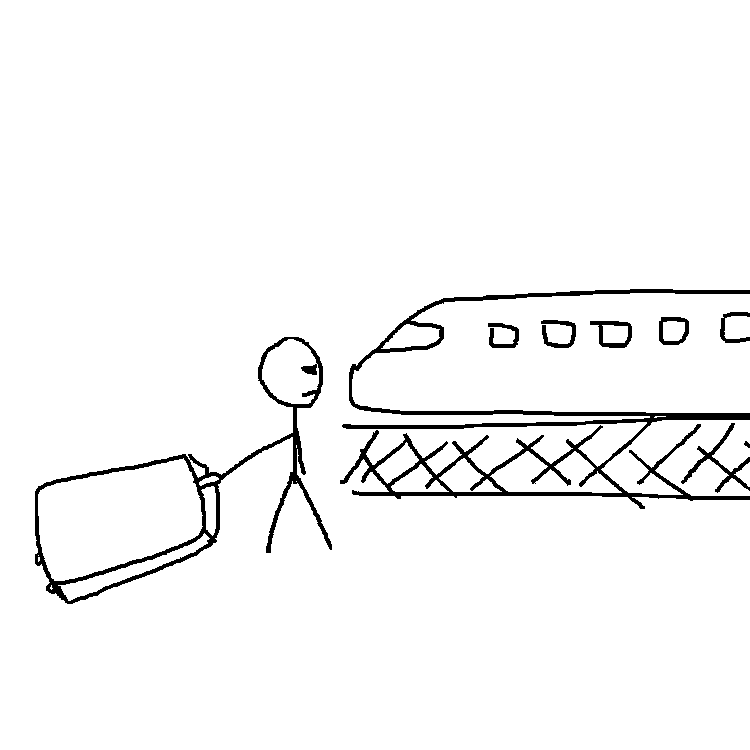}} &
        \frame{\includegraphics[width=0.22\linewidth]{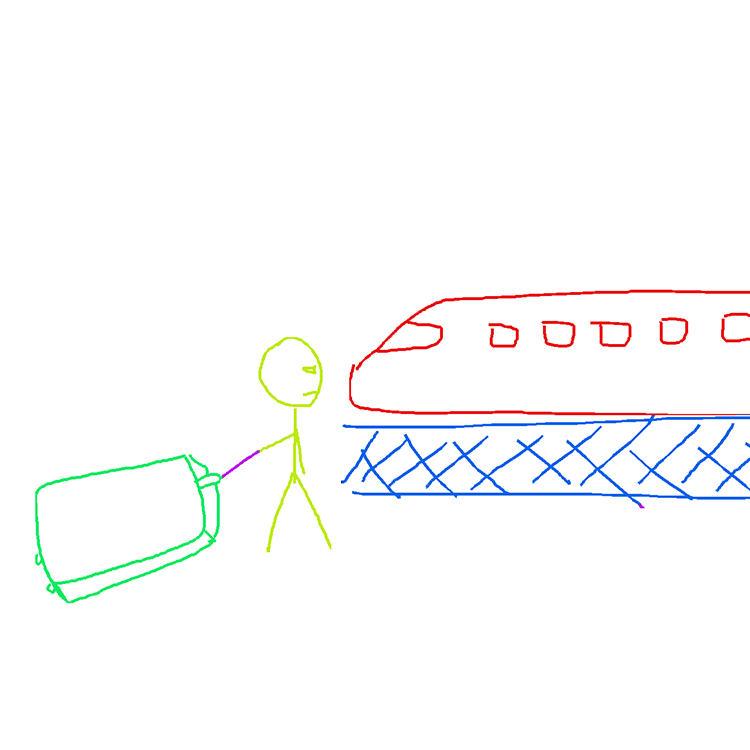}} &
        \frame{\includegraphics[width=0.22\linewidth]{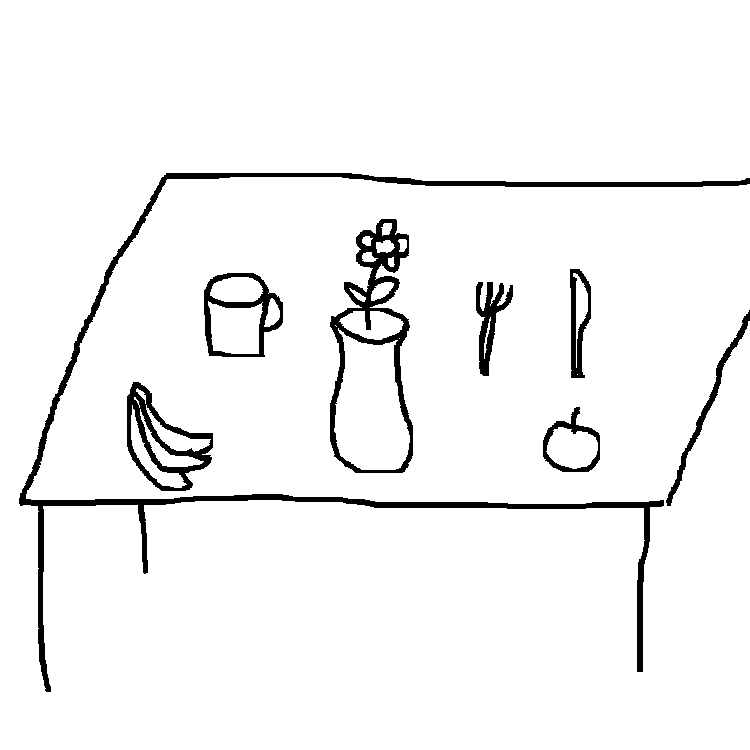}} &
        \frame{\includegraphics[width=0.22\linewidth]{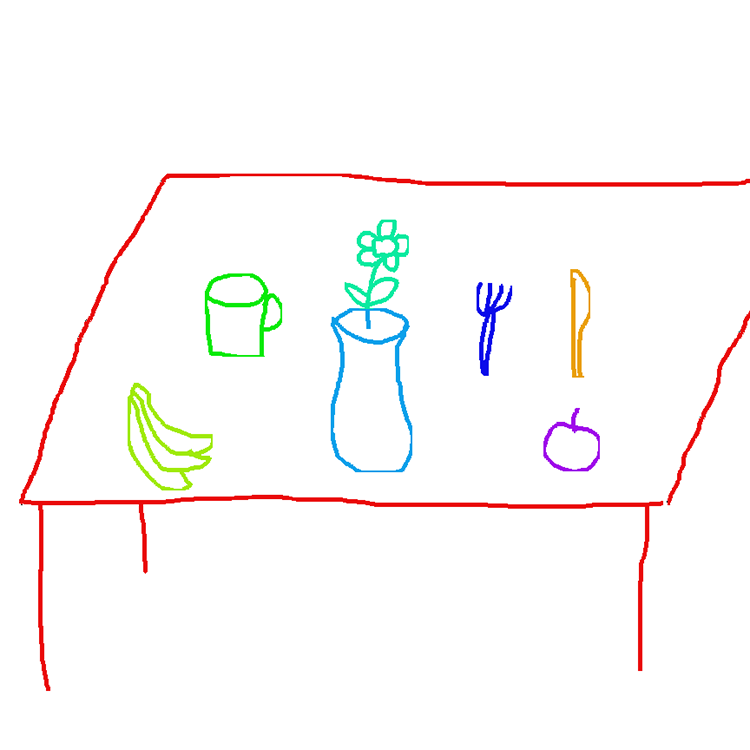}} 
        \\
    \end{tabular}
    }}
    \caption{\rev{\textbf{Zhang \textit{et al.} Performance.} We show our segmentation outputs of Zhang \textit{et al.} across a diverse range of scene sketches. The examples span indoor scenes (e.g., bedrooms, desks), outdoor urban settings (e.g., train station), and natural environments with animals, both in close-up and distant views. Our method demonstrates reasonable segmentation performance across this broad spectrum of semantic and spatial contexts.}}
    \label{fig:zhang_results}
\end{figure*}

\rev{
In addition to the numerical evaluations we include in the main paper, we  present qualitative results on Zhang \etal \shortcite{zhang2018CBSC}'s dataset in Fig ~\ref{fig:zhang_results}. This dataset consists of 330 scene sketches drawn by humans, spanning 74 categories with 24 novel categories not included in our \datasetname{}. Our performance on this dataset further shows our generalization capabilities for unseen categories and scenes.}

\section{Additional Qualitative Comparisons}
\begin{figure*}
    \centering
    \setlength{\tabcolsep}{2pt}
    {\small
    \resizebox{0.92\textwidth}{!}{ 
    \begin{tabular}{c @{\hskip 10pt} c c c c}
        Input & SketchyScene & Grounded SAM & \rev{Automatic SAM} & \textbf{Ours} \\
        
        \frame{\includegraphics[width=0.20\linewidth]{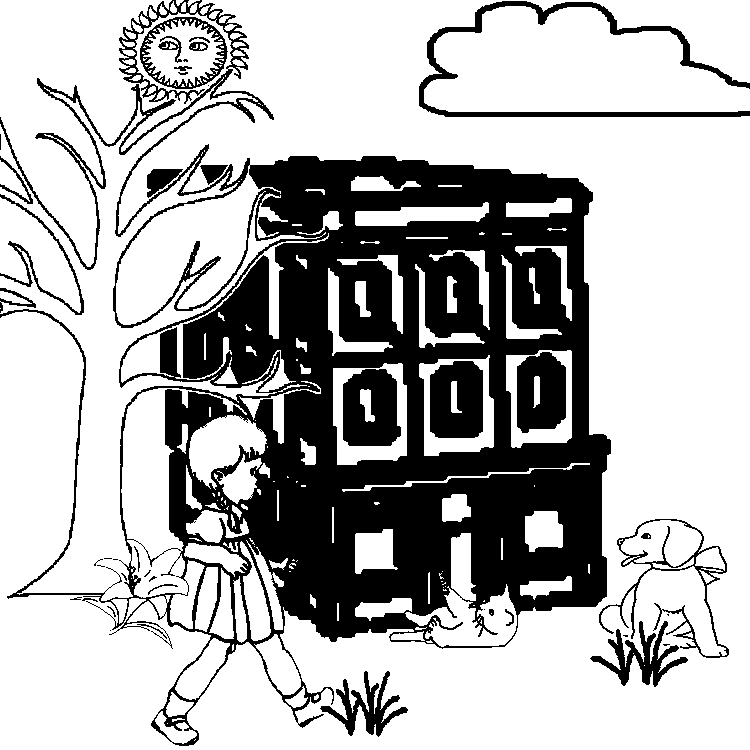}} &
        \frame{\includegraphics[width=0.20\linewidth]{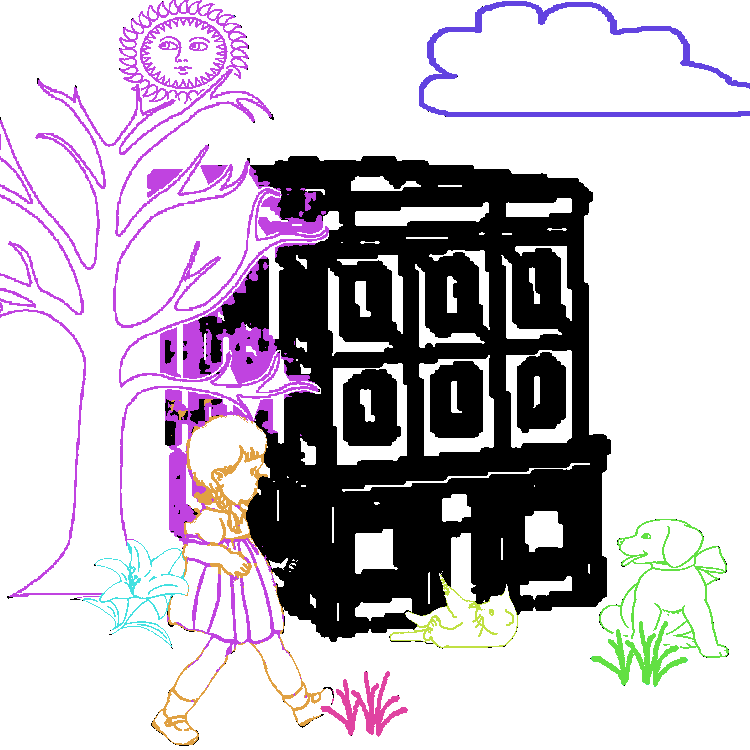}} &
        \frame{\includegraphics[width=0.20\linewidth]{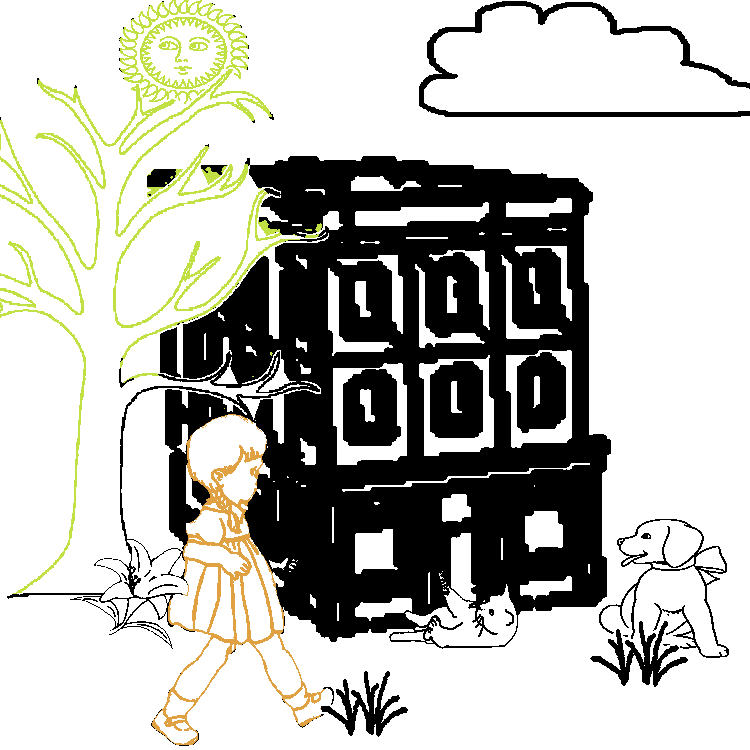}} &
        \frame{\includegraphics[width=0.20\linewidth]{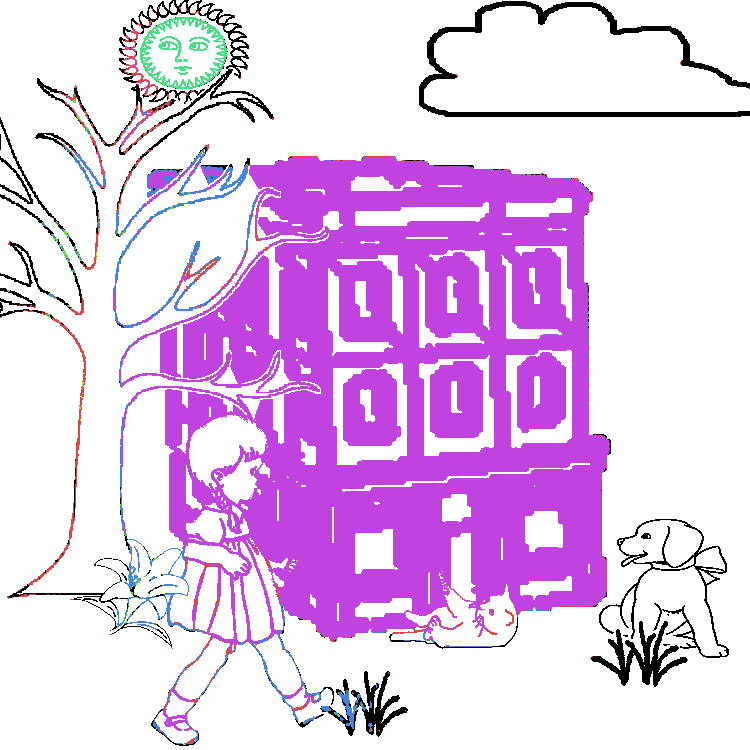}} &
        \frame{\includegraphics[width=0.20\linewidth]{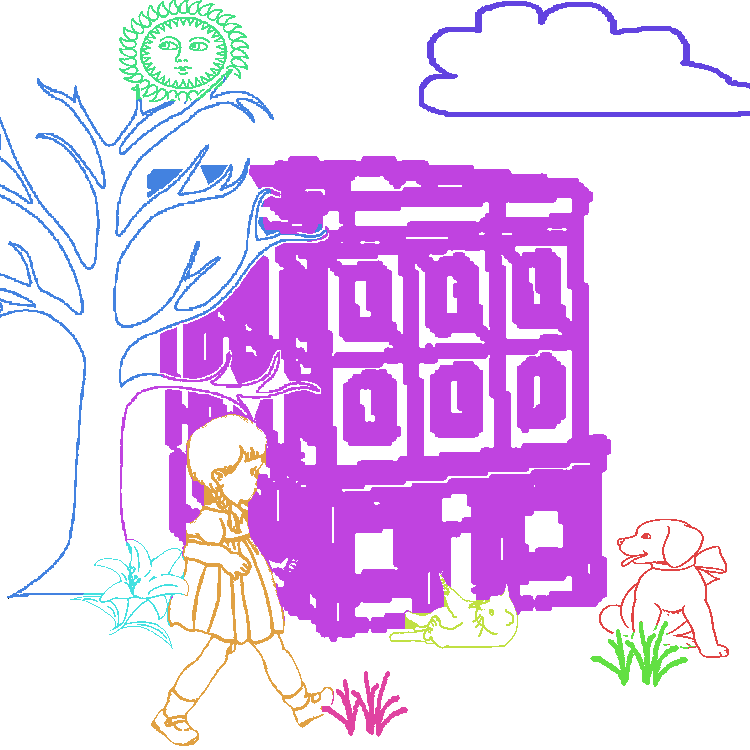}} \\

        \frame{\includegraphics[width=0.20\linewidth]{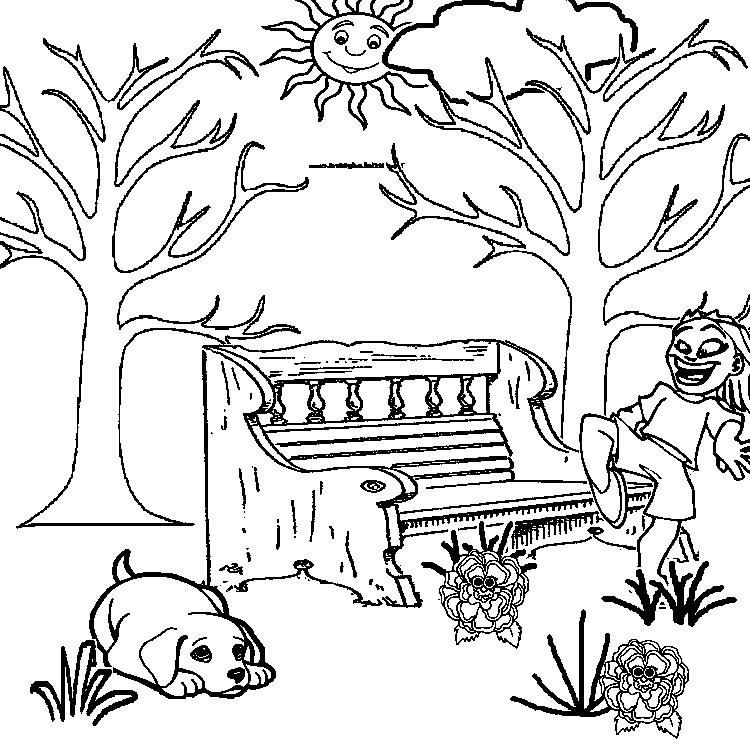}} &
        \frame{\includegraphics[width=0.20\linewidth]{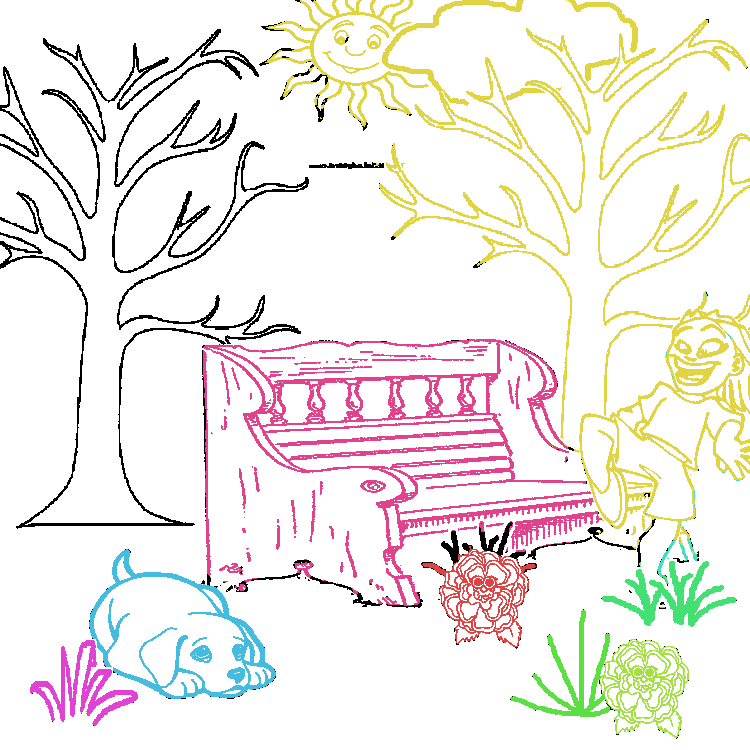}} &
        \frame{\includegraphics[width=0.20\linewidth]{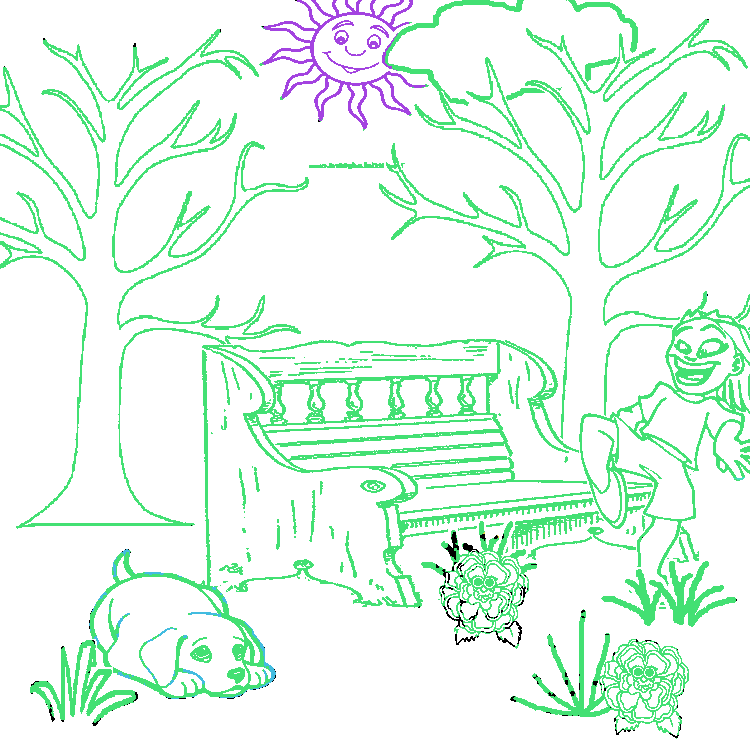}} &
        \frame{\includegraphics[width=0.20\linewidth]{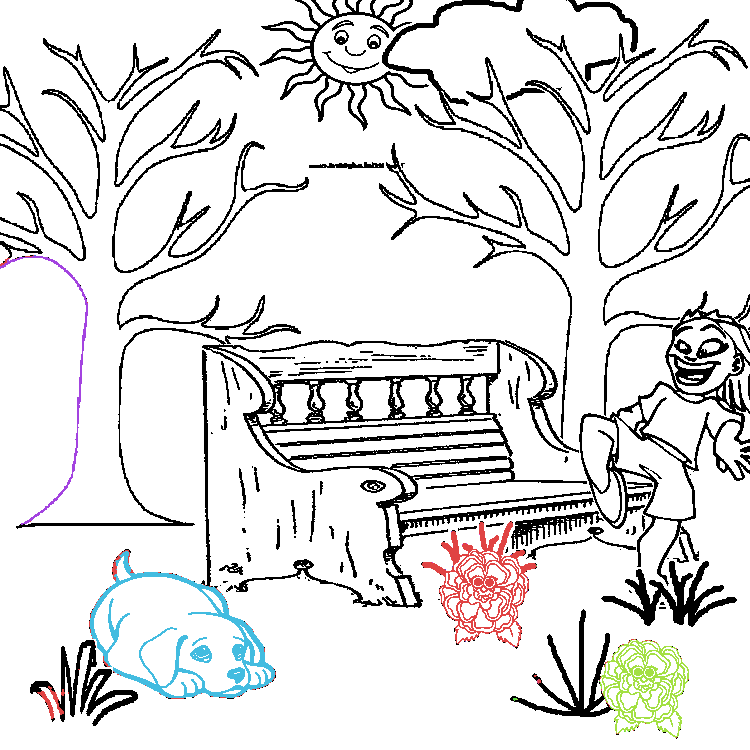}} &
        \frame{\includegraphics[width=0.20\linewidth]{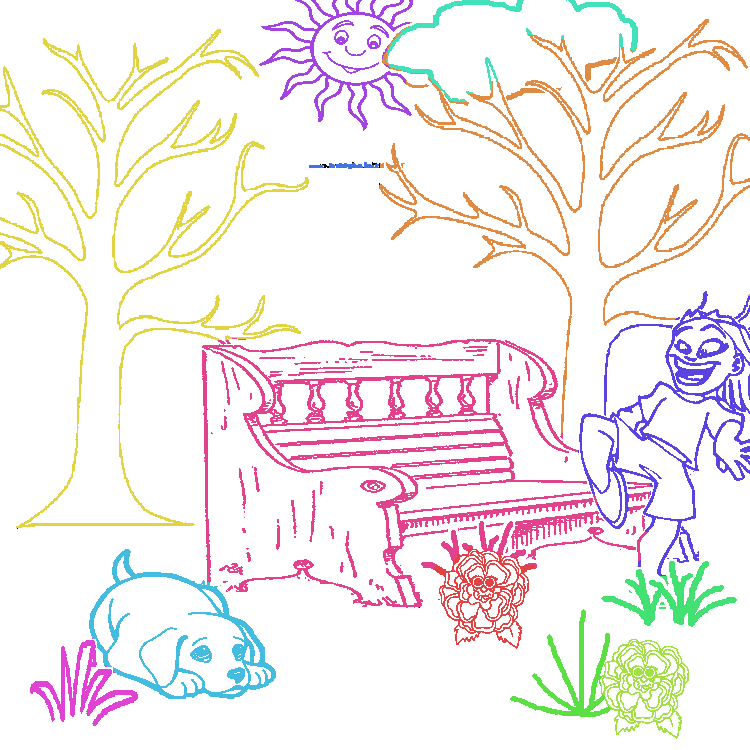}} \\

        \frame{\includegraphics[width=0.20\linewidth]{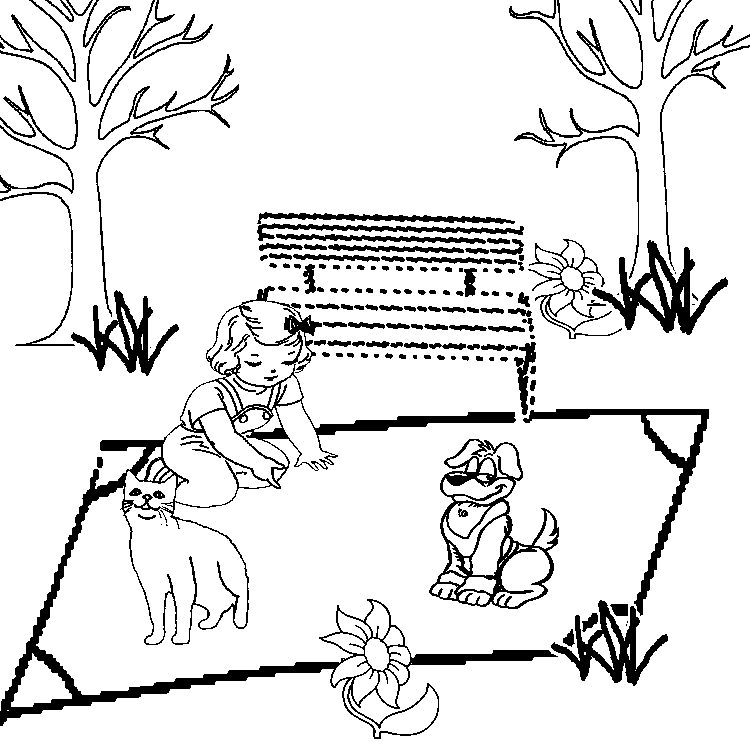}} &
        \frame{\includegraphics[width=0.20\linewidth]{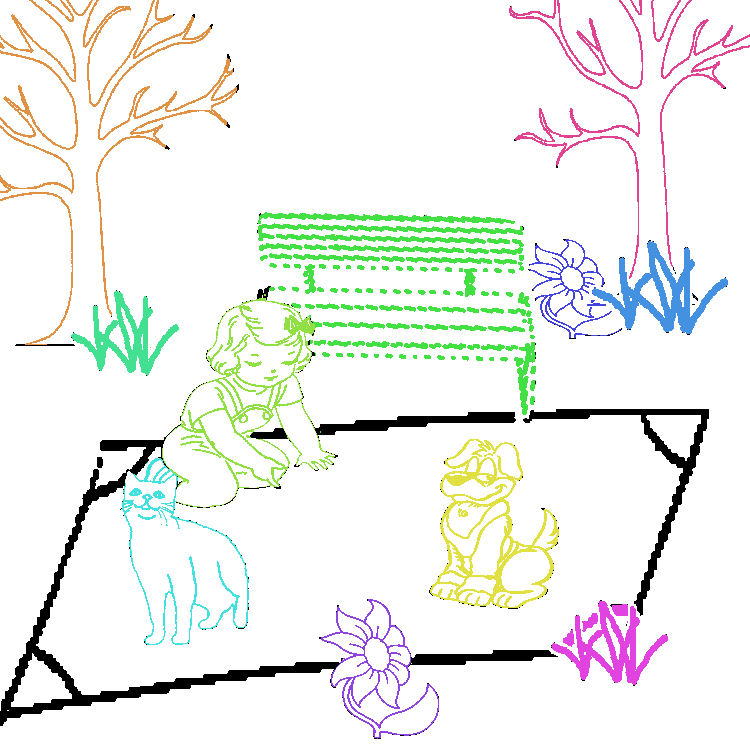}} &
        \frame{\includegraphics[width=0.20\linewidth]{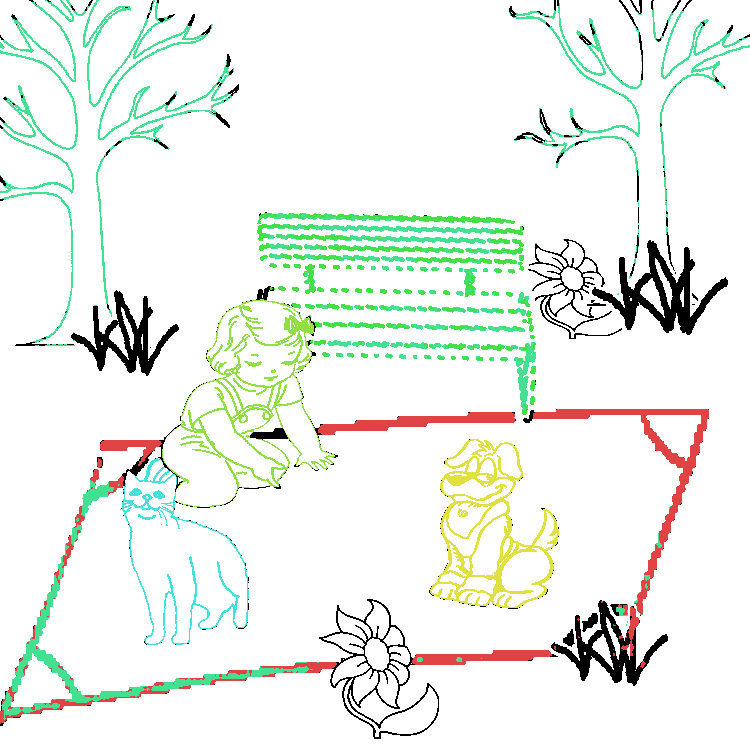}} &
        \frame{\includegraphics[width=0.20\linewidth]{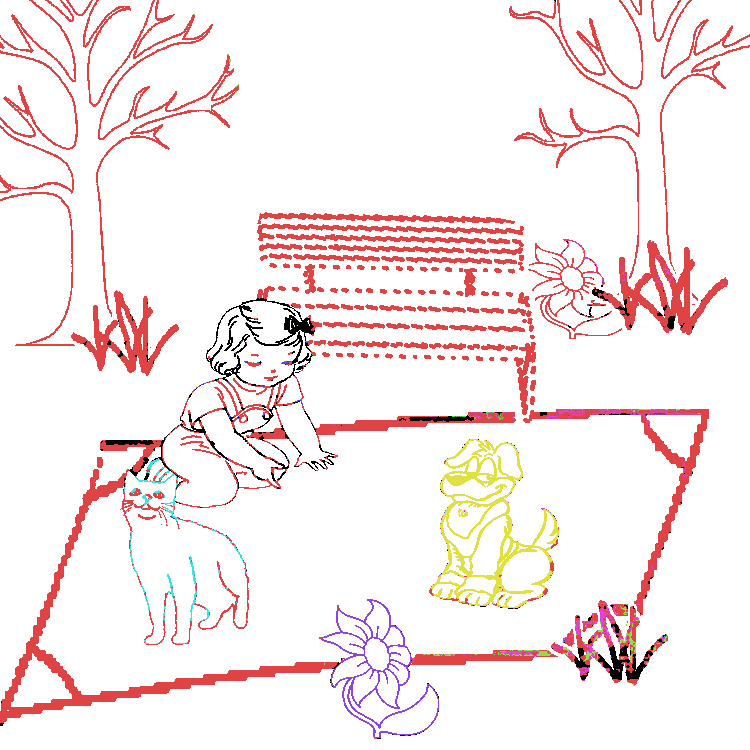}} &
        \frame{\includegraphics[width=0.20\linewidth]{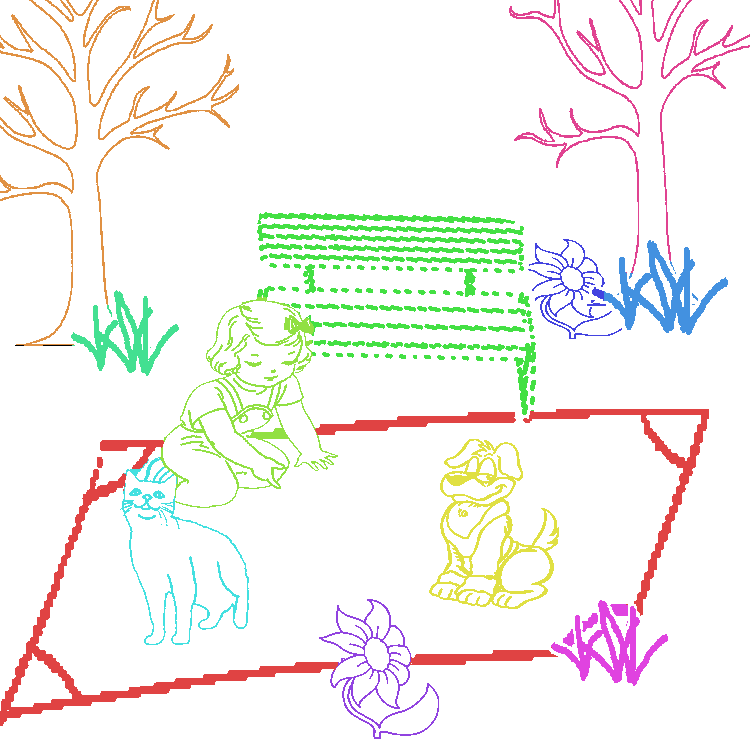}} \\

        \frame{\includegraphics[width=0.20\linewidth]{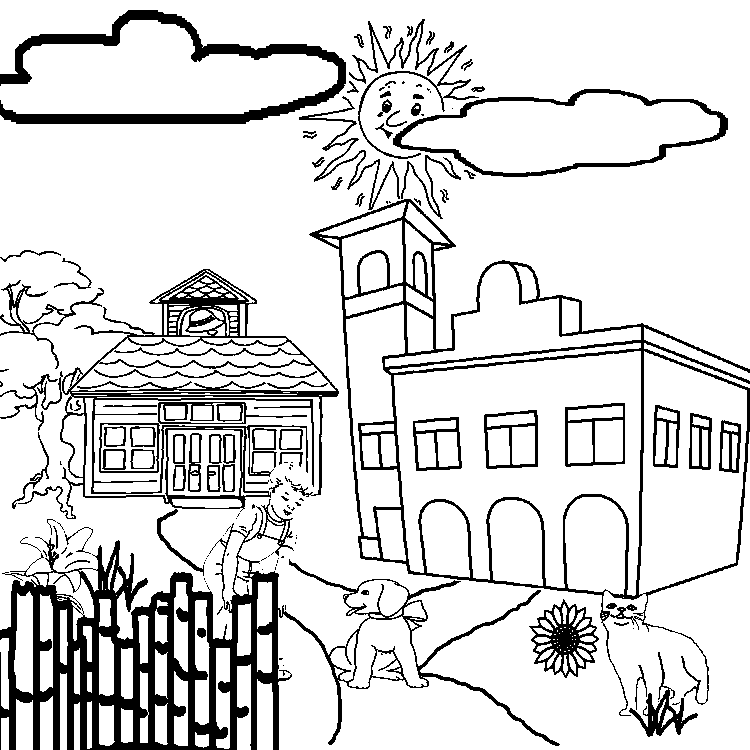}} &
        \frame{\includegraphics[width=0.20\linewidth]{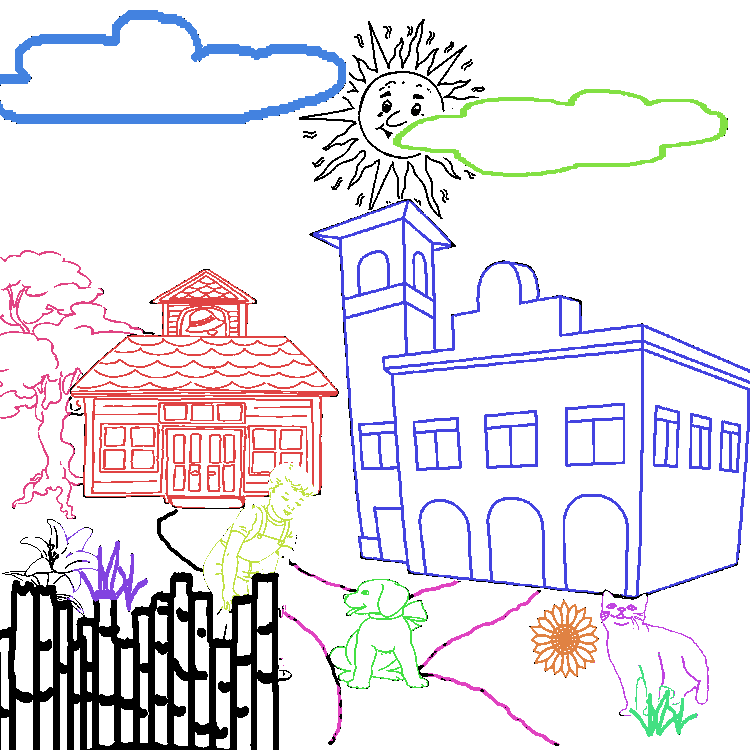}} &
        \frame{\includegraphics[width=0.20\linewidth]{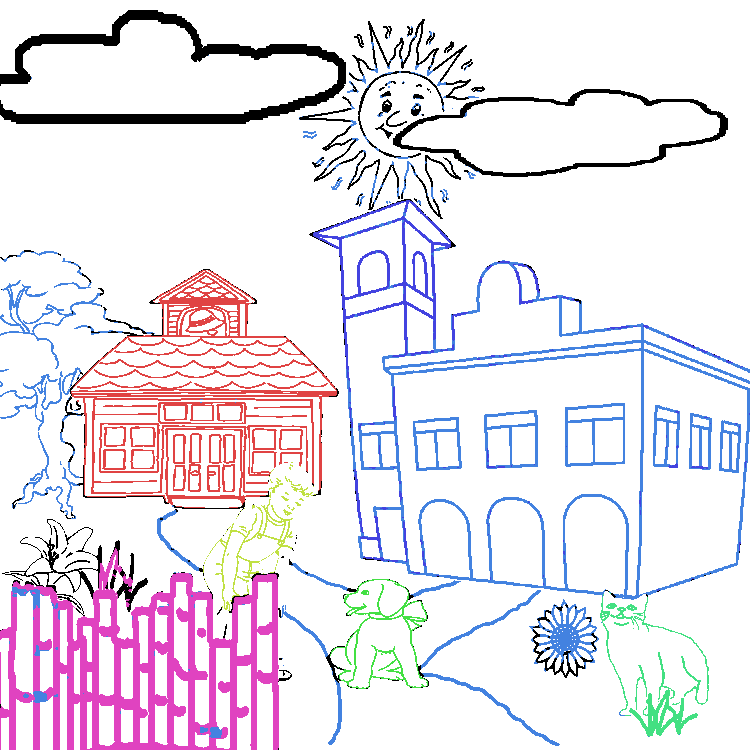}} &
        \frame{\includegraphics[width=0.20\linewidth]{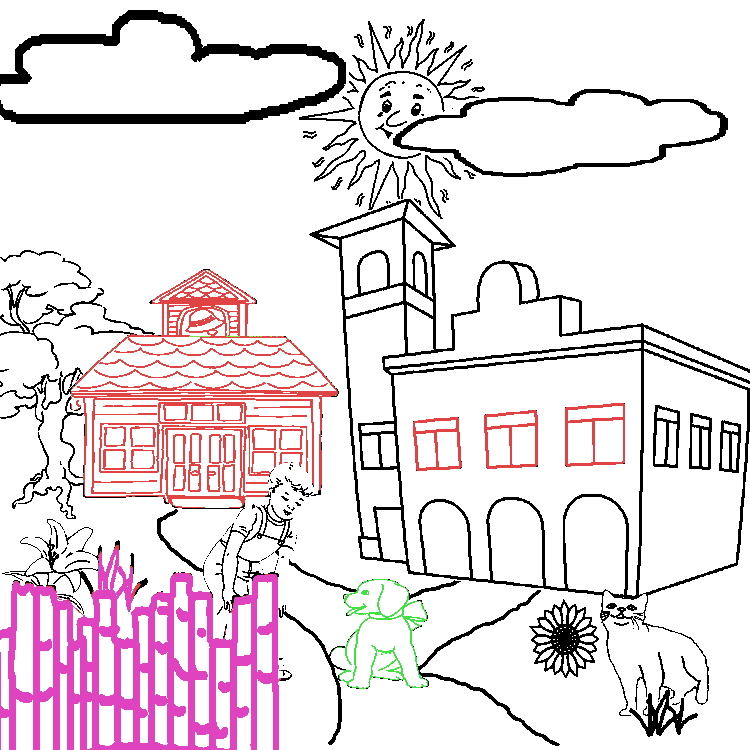}} &
        \frame{\includegraphics[width=0.20\linewidth]{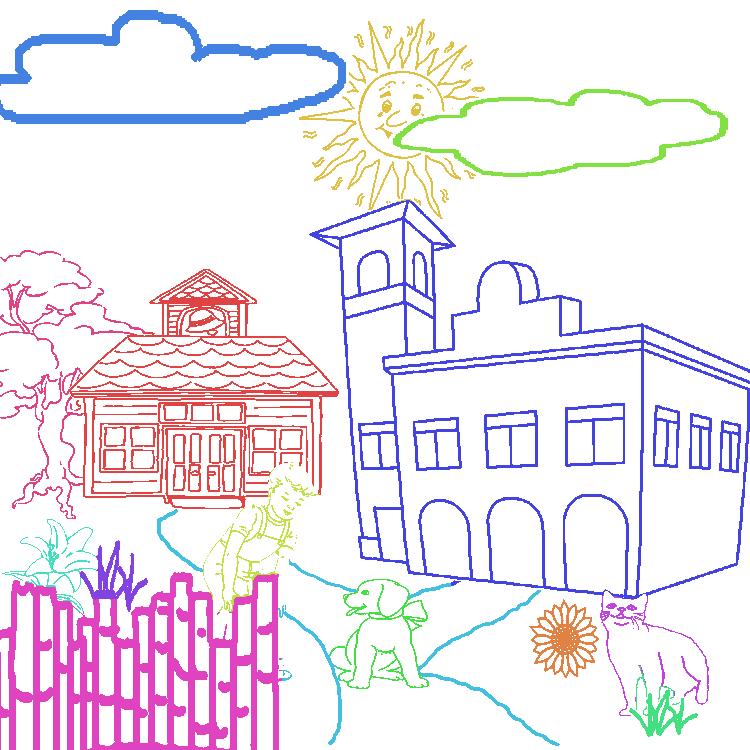}} \\

        \frame{\includegraphics[width=0.20\linewidth]{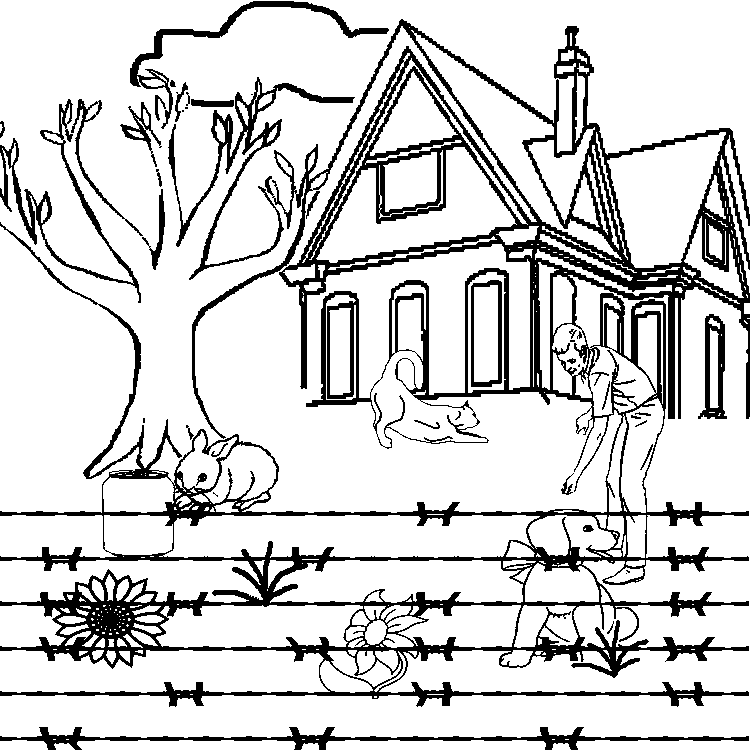}} &
        \frame{\includegraphics[width=0.20\linewidth]{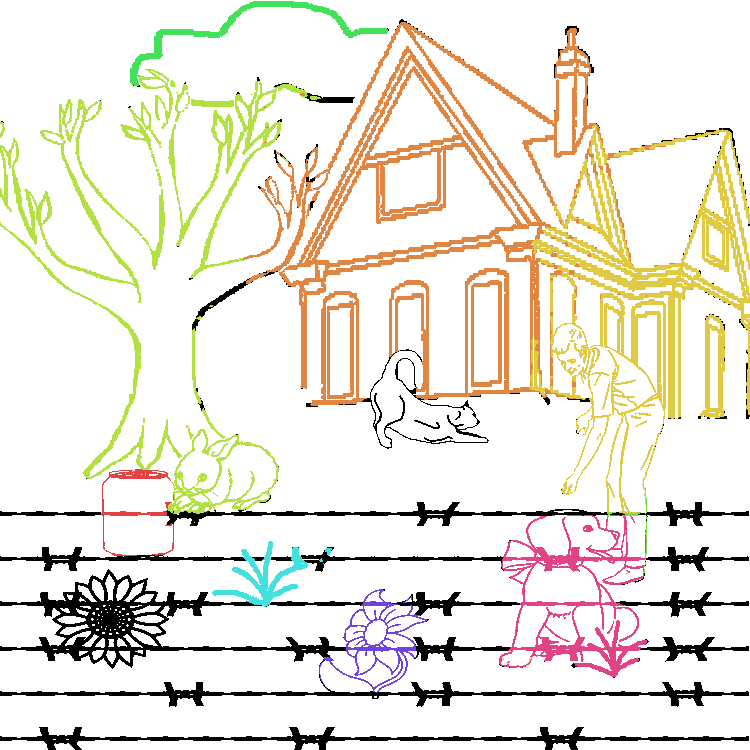}} &
        \frame{\includegraphics[width=0.20\linewidth]{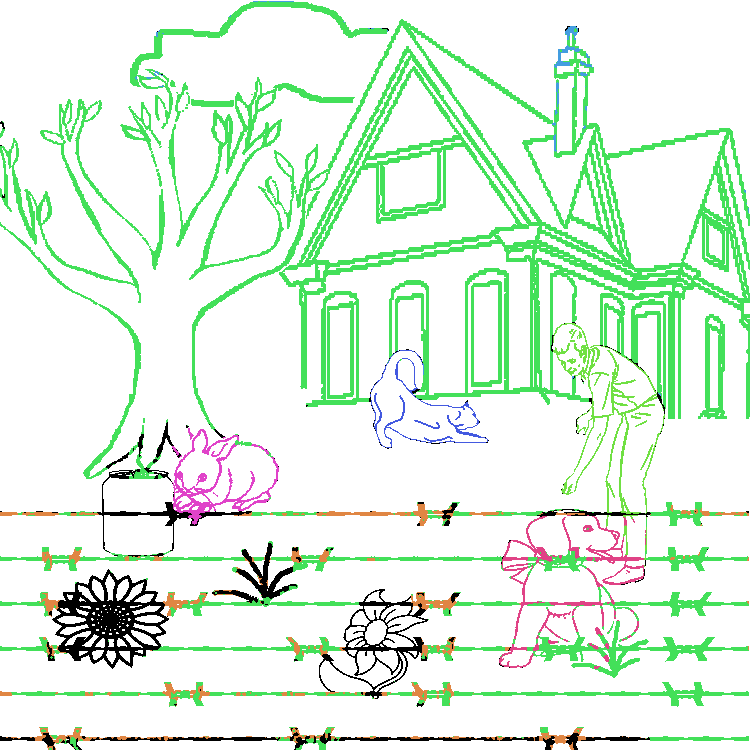}} &
        \frame{\includegraphics[width=0.20\linewidth]{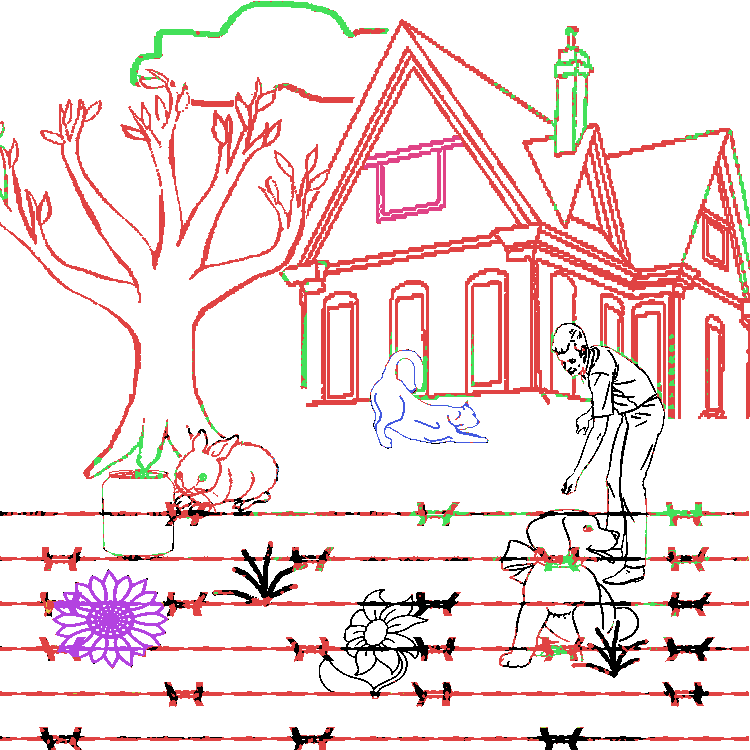}} &
        \frame{\includegraphics[width=0.20\linewidth]{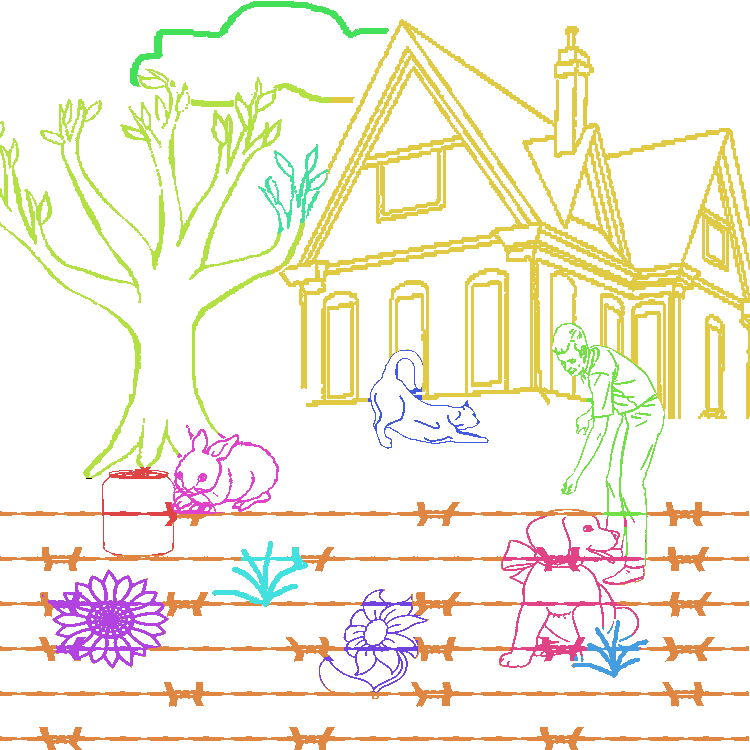}} \\

        \frame{\includegraphics[width=0.20\linewidth]{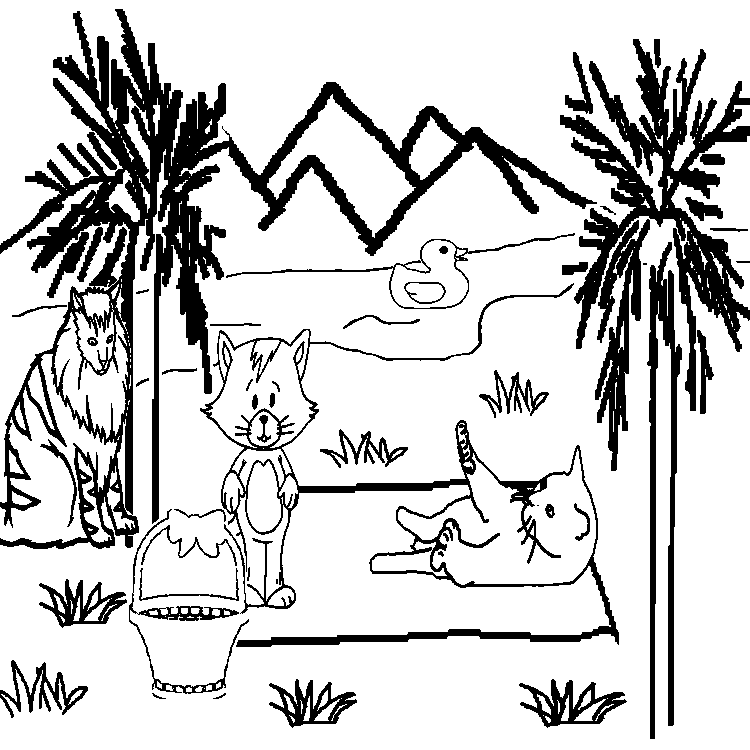}} &
        \frame{\includegraphics[width=0.20\linewidth]{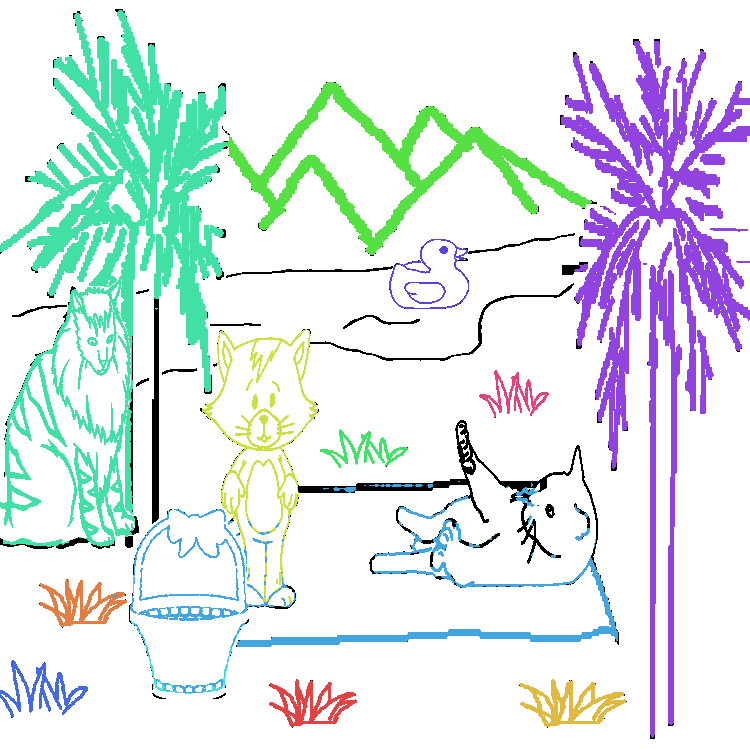}} &
        \frame{\includegraphics[width=0.20\linewidth]{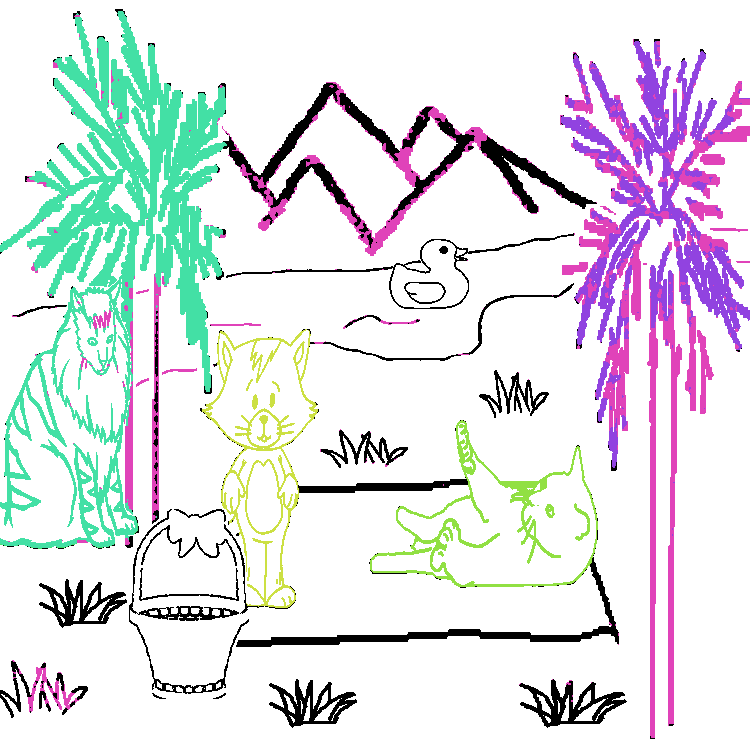}} &
        \frame{\includegraphics[width=0.20\linewidth]{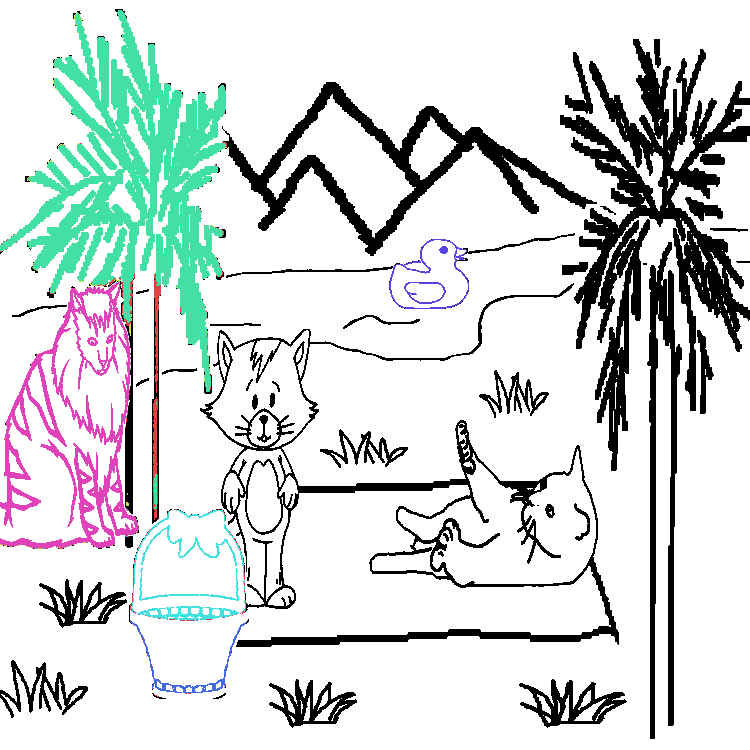}} &
        \frame{\includegraphics[width=0.20\linewidth]{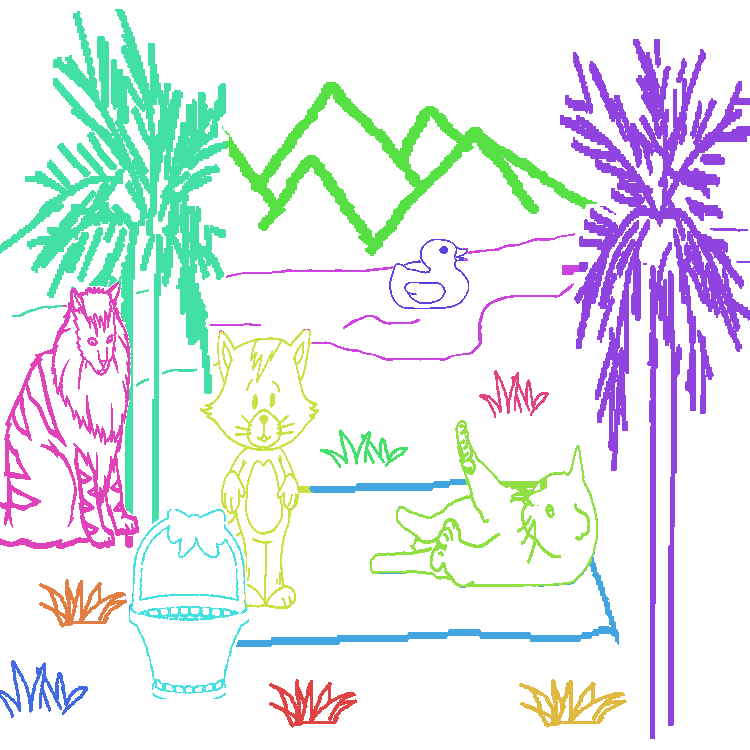}} \\
    \end{tabular}
    }
    \vspace{-0.1cm}
    \caption{\textbf{Qualitative comparison of instance segmentation methods on the SketchyScene dataset.} Our method surpasses SketchyScene, Grounded SAM and Automatic SAM baselines by delivering fine-grained, semantically consistent segmentations with precise boundaries. In the urban scene (row 1), our approach accurately segments all object instances, cleanly separating the sun, tree, building, and character from each other. For the park scenes (rows 2 and 3), it captures intricate details, such as the dog's face and picnic items, which are either missed or over-segmented by other methods. In the residential and cottage scenes (rows 4 and 5), our method effectively delineates repetitive patterns like fences and handles dense objects like trees, preserving structural integrity where other approaches struggle. These results highlight the robustness of our method in managing complex and detailed sketches.}
    \label{fig:comparison_instance_sketchyscene}
    }
\end{figure*}

\begin{figure*}
    \centering
    \setlength{\tabcolsep}{2pt}
    {\small
    \resizebox{0.92\textwidth}{!}{ 
    \begin{tabular}{c @{\hskip 10pt} c c c c}
        Input & SketchyScene & Grounded SAM & \rev{Automatic SAM} & \textbf{Ours} \\
        
        \frame{\includegraphics[width=0.20\linewidth]{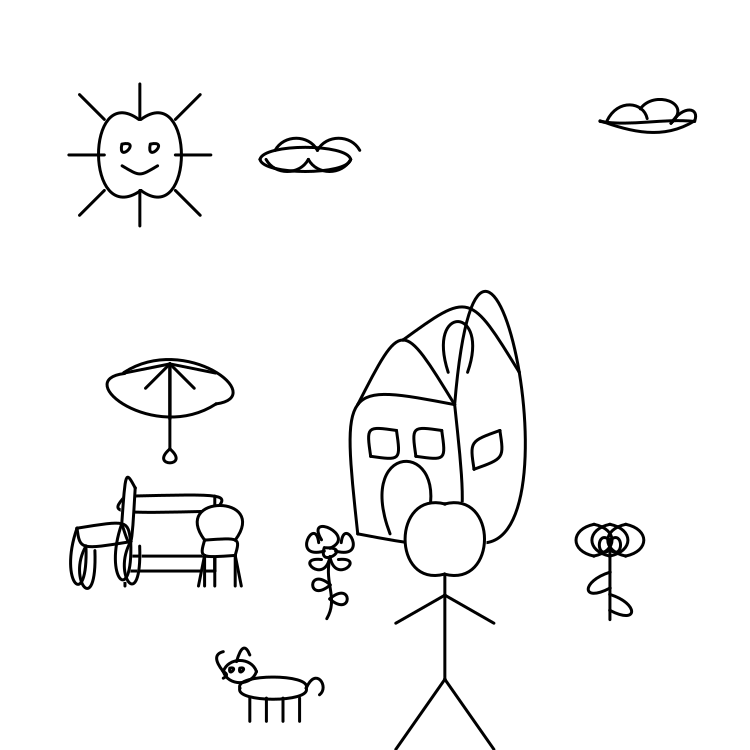}} &
        \frame{\includegraphics[width=0.20\linewidth]{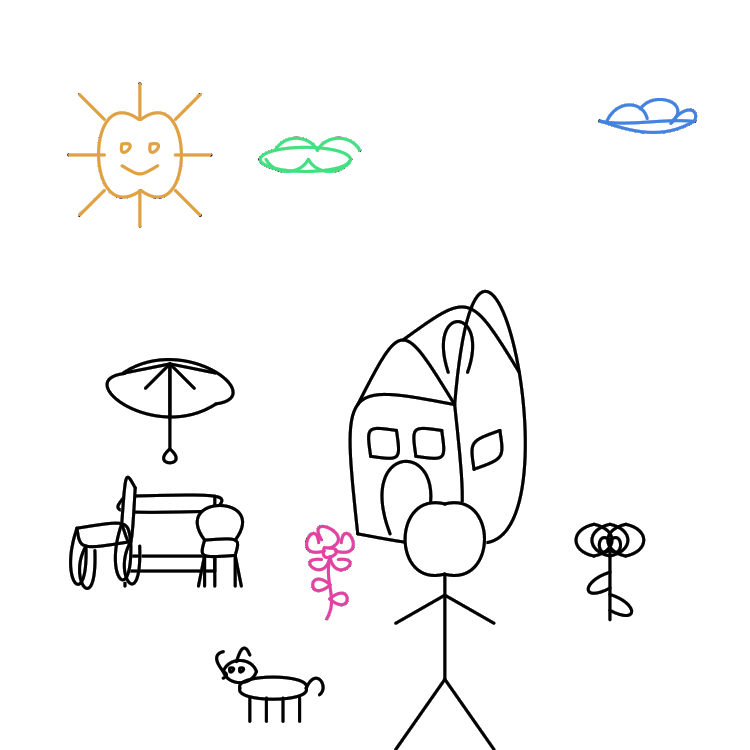}} &
        \frame{\includegraphics[width=0.20\linewidth]{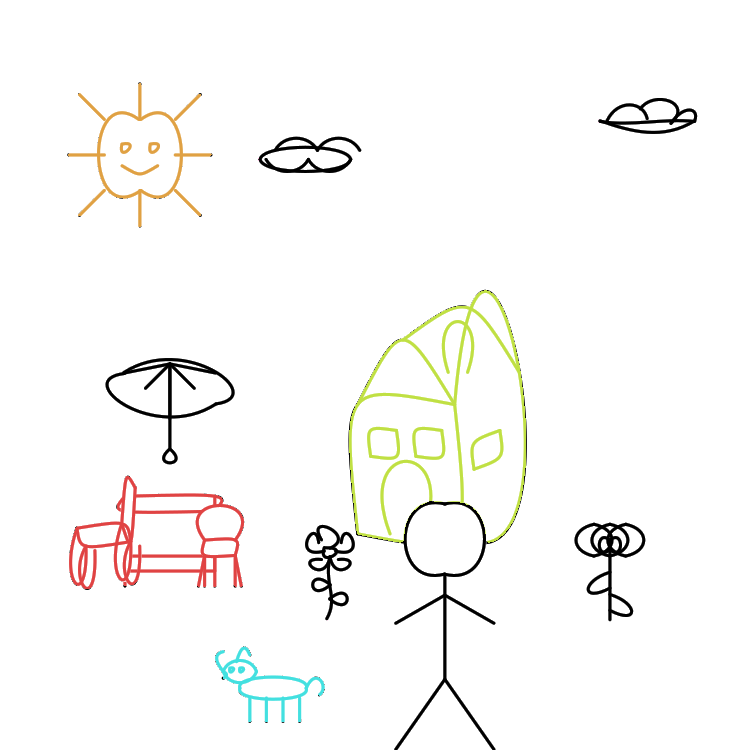}} &
        \frame{\includegraphics[width=0.20\linewidth]{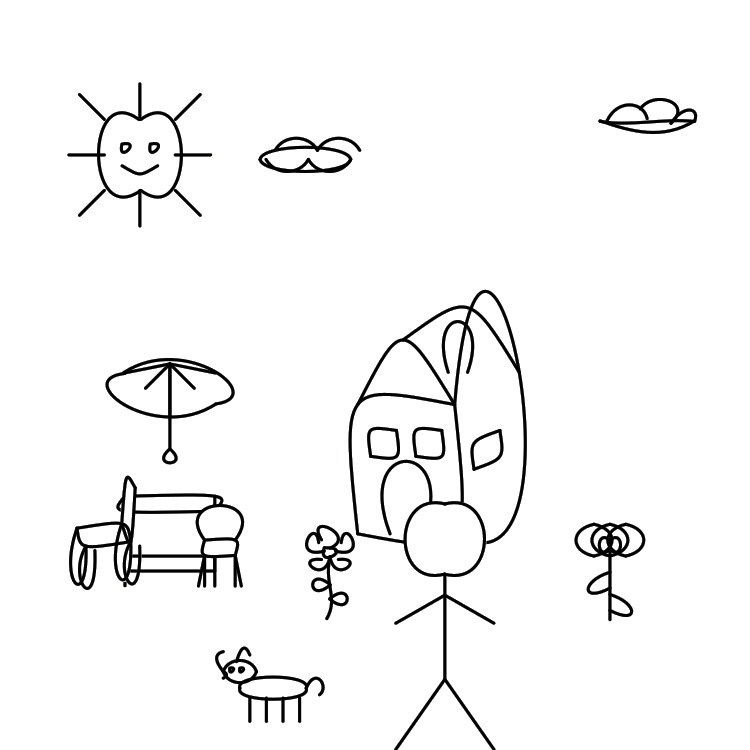}} &
        \frame{\includegraphics[width=0.20\linewidth]{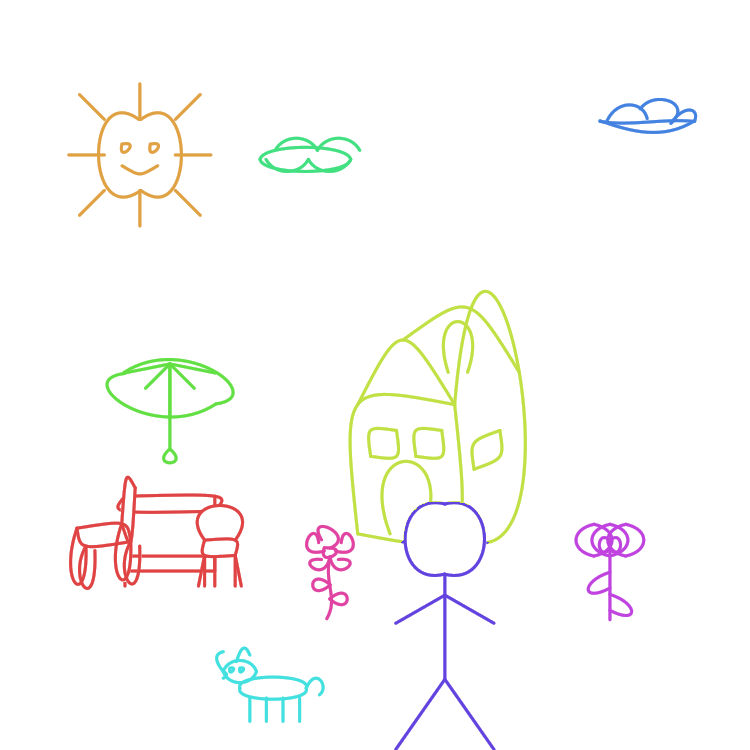}} \\

        \frame{\includegraphics[width=0.20\linewidth]{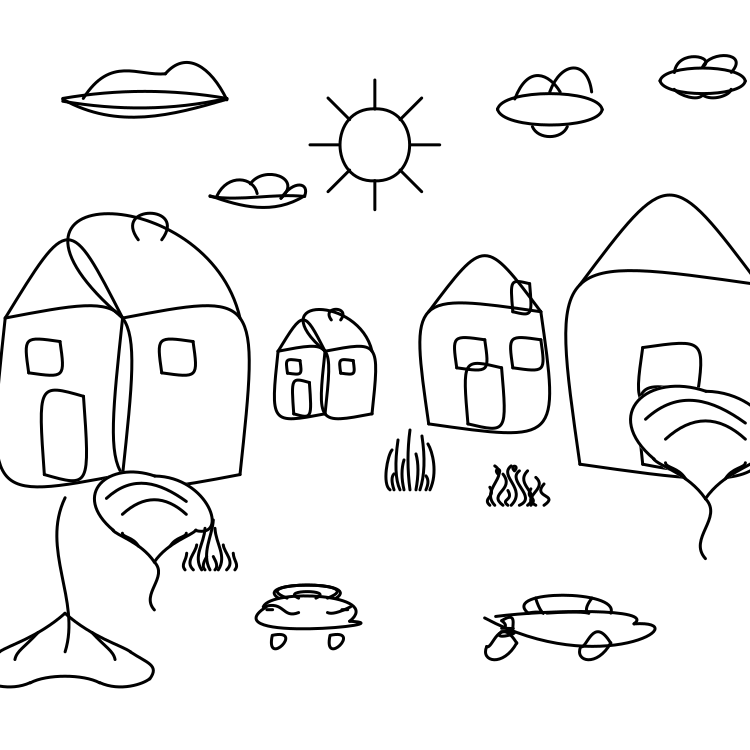}} &
        \frame{\includegraphics[width=0.20\linewidth]{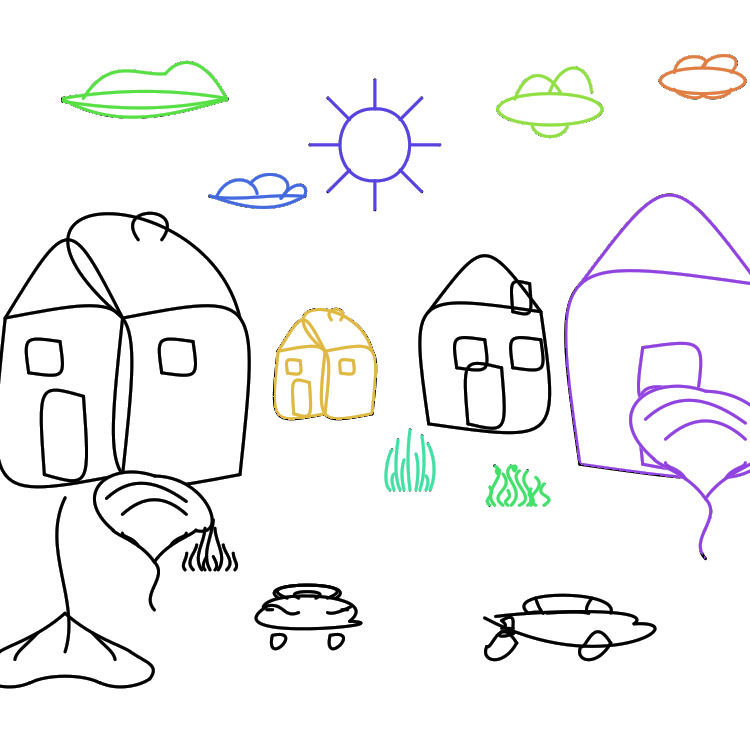}} &
        \frame{\includegraphics[width=0.20\linewidth]{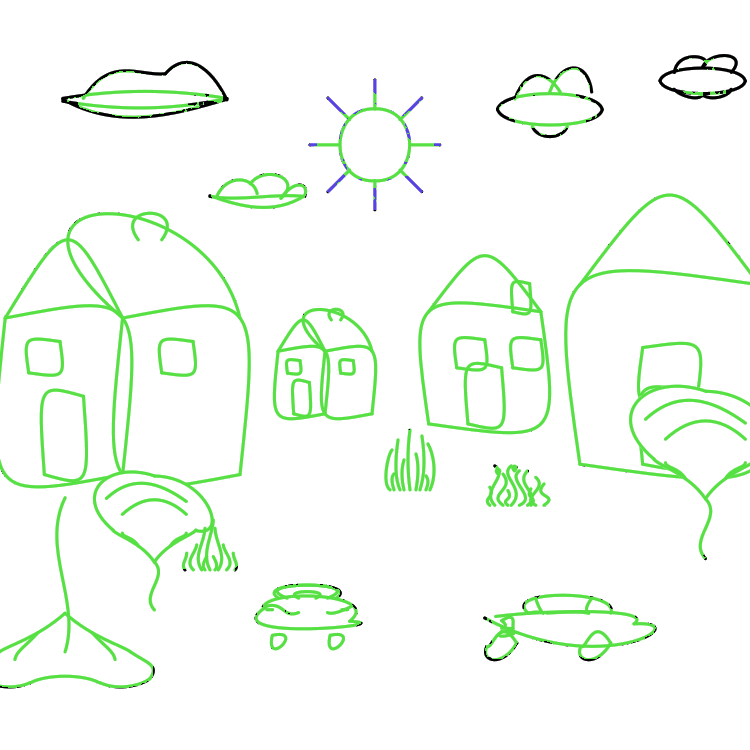}} &
        \frame{\includegraphics[width=0.20\linewidth]{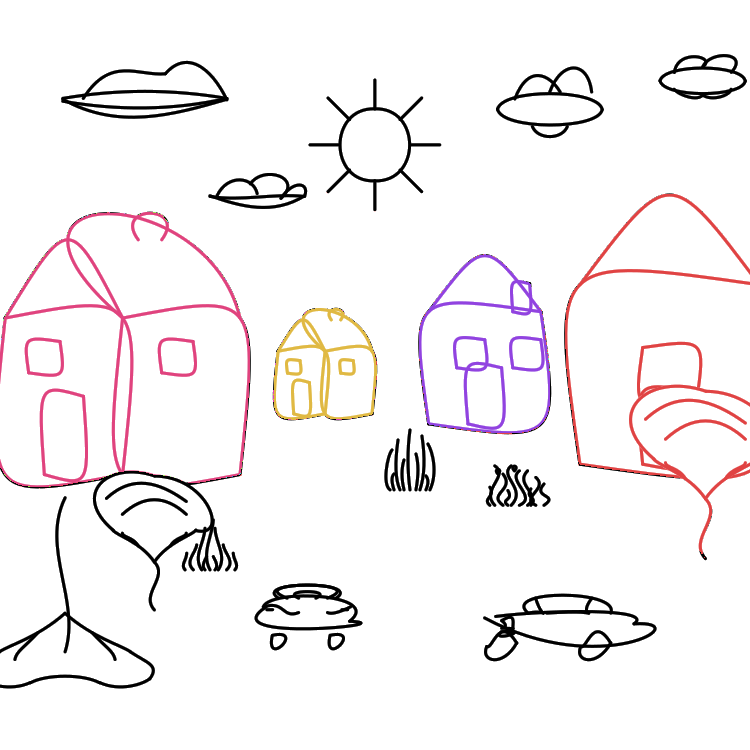}} &
        \frame{\includegraphics[width=0.20\linewidth]{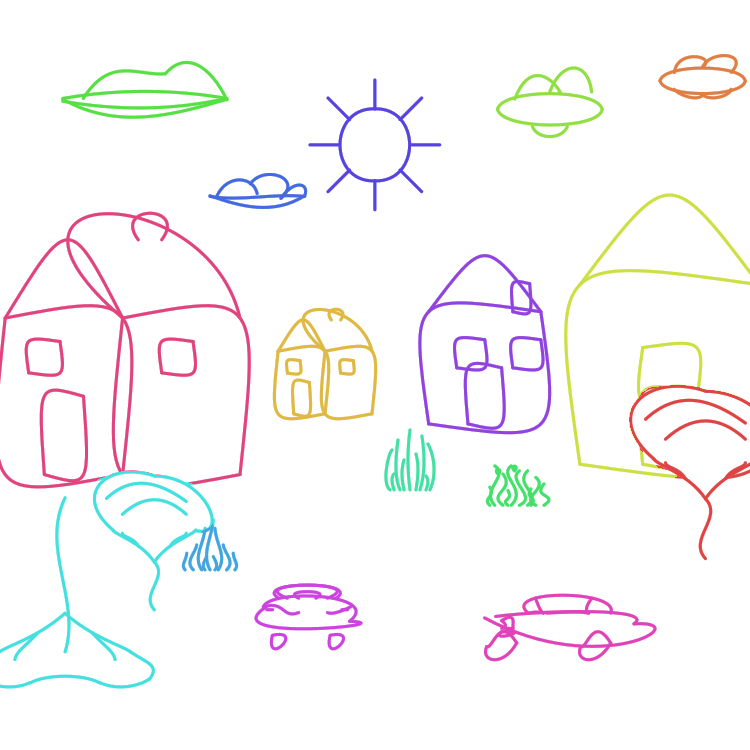}} \\

        \frame{\includegraphics[width=0.20\linewidth]{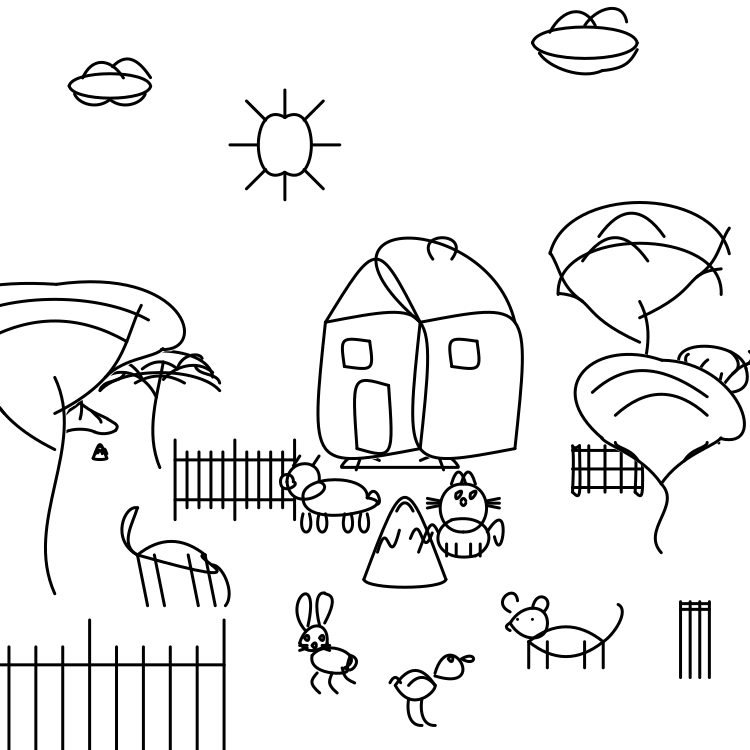}} &
        \frame{\includegraphics[width=0.20\linewidth]{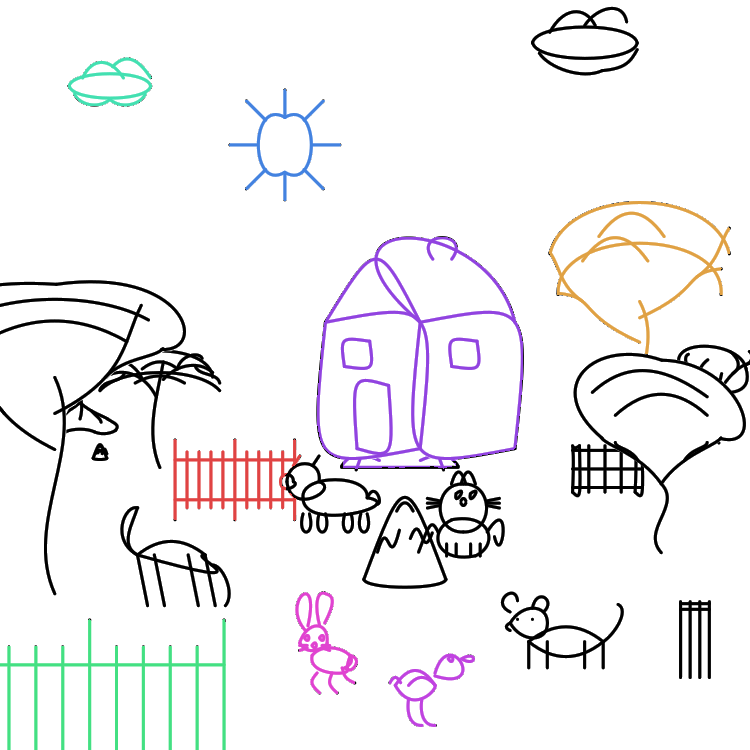}} &
        \frame{\includegraphics[width=0.20\linewidth]{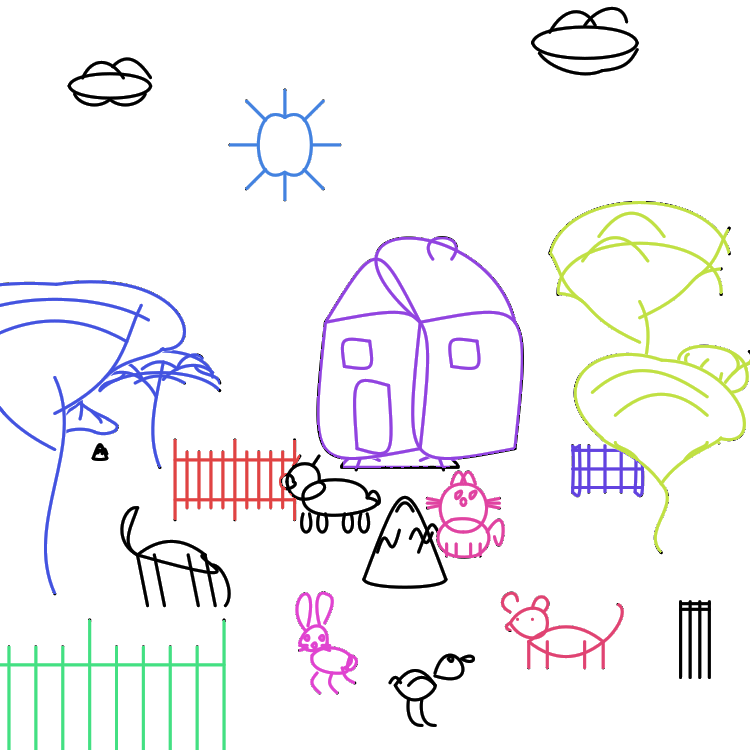}} &
        \frame{\includegraphics[width=0.20\linewidth]{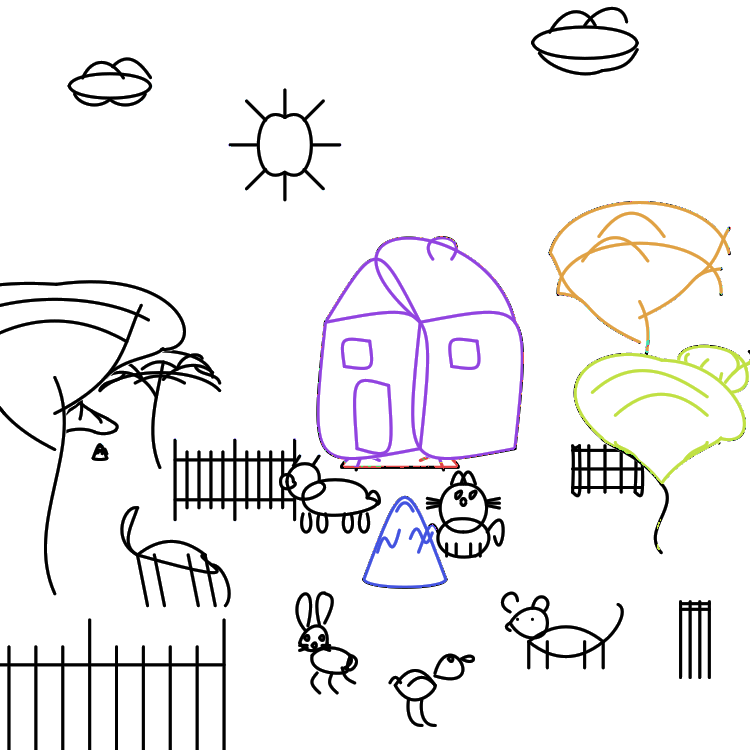}} &
        \frame{\includegraphics[width=0.20\linewidth]{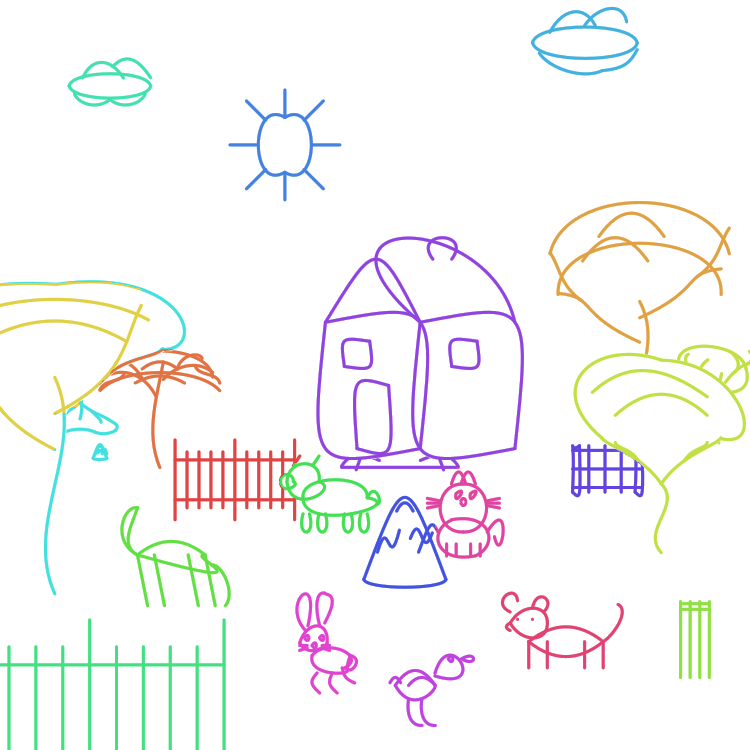}} \\

        \frame{\includegraphics[width=0.20\linewidth]{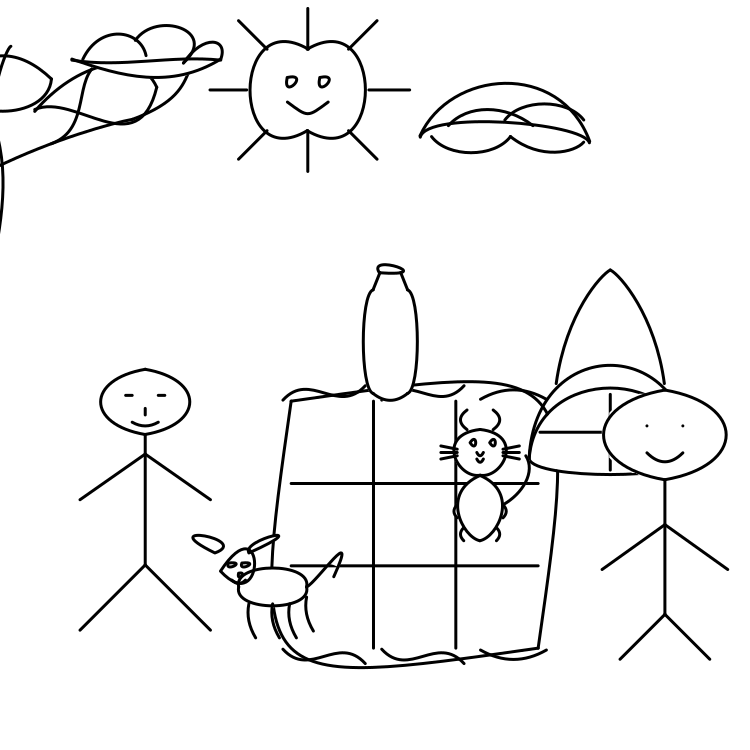}} &
        \frame{\includegraphics[width=0.20\linewidth]{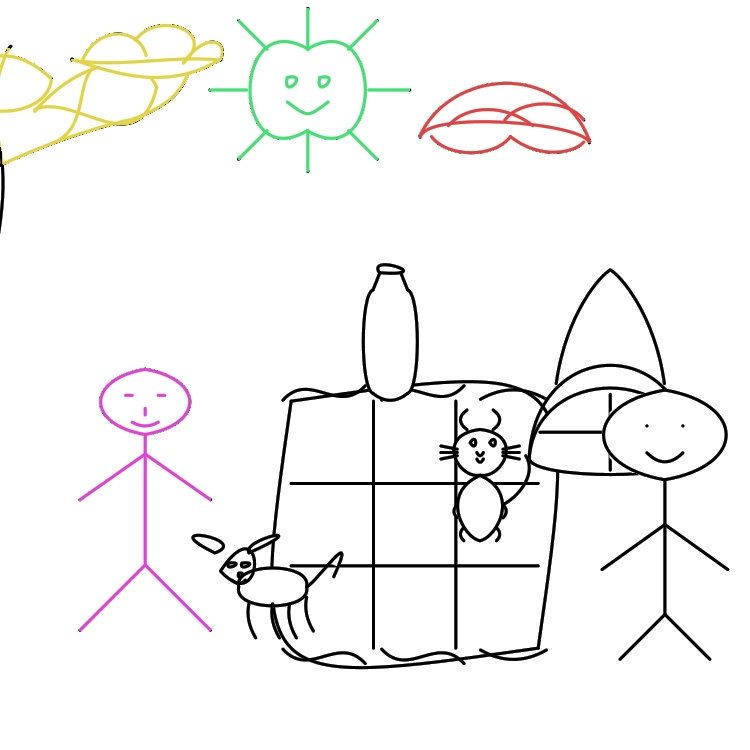}} &
        \frame{\includegraphics[width=0.20\linewidth]{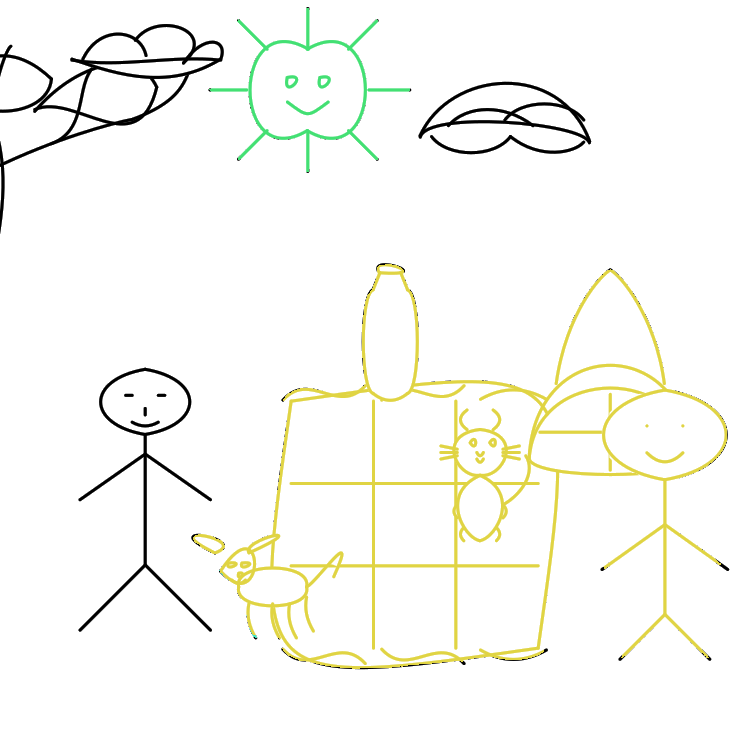}} &
        \frame{\includegraphics[width=0.20\linewidth]{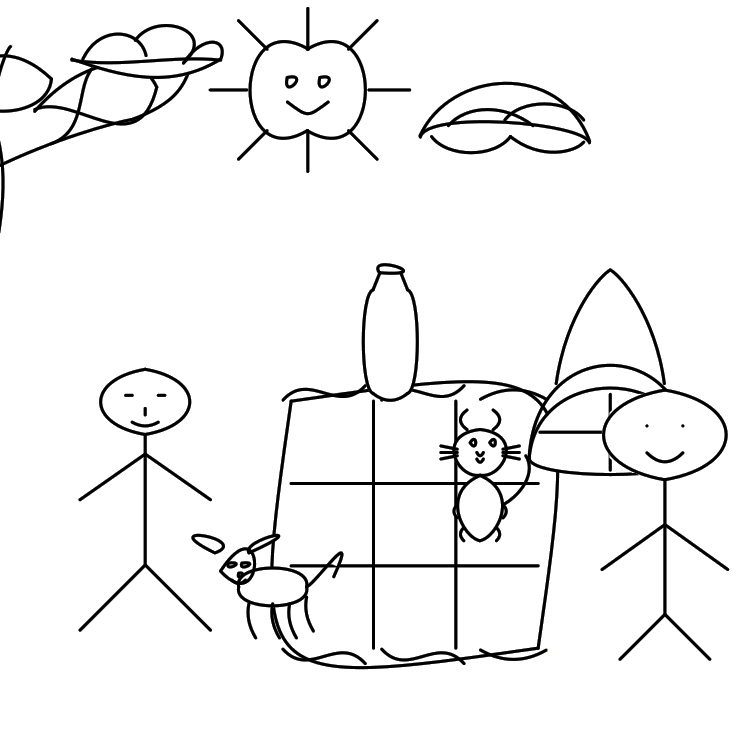}} &
        \frame{\includegraphics[width=0.20\linewidth]{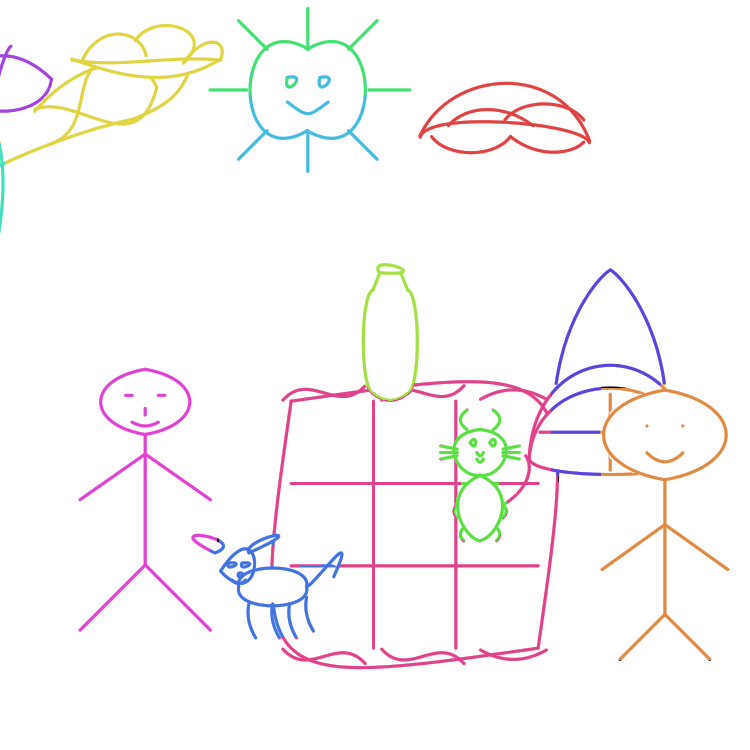}} \\

        \frame{\includegraphics[width=0.20\linewidth]{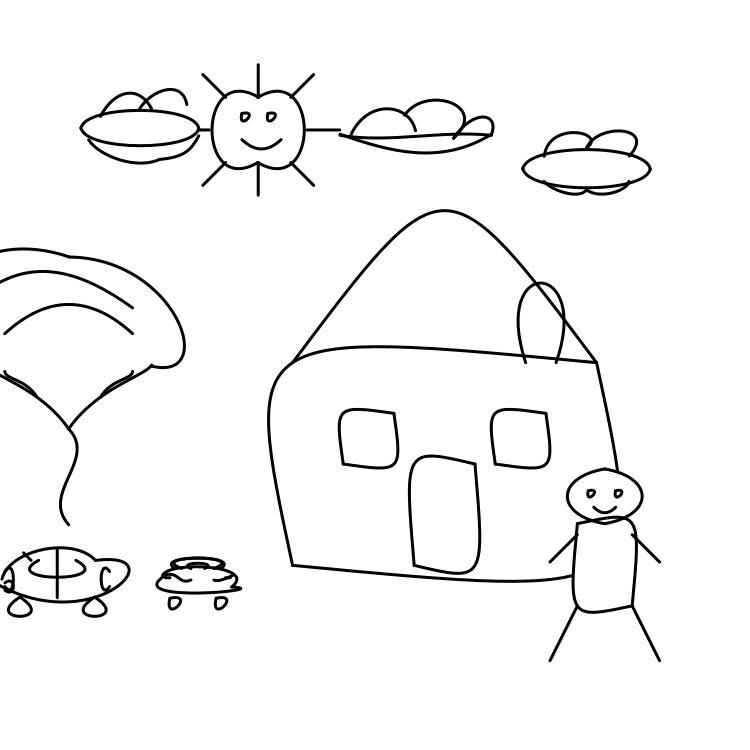}} &
        \frame{\includegraphics[width=0.20\linewidth]{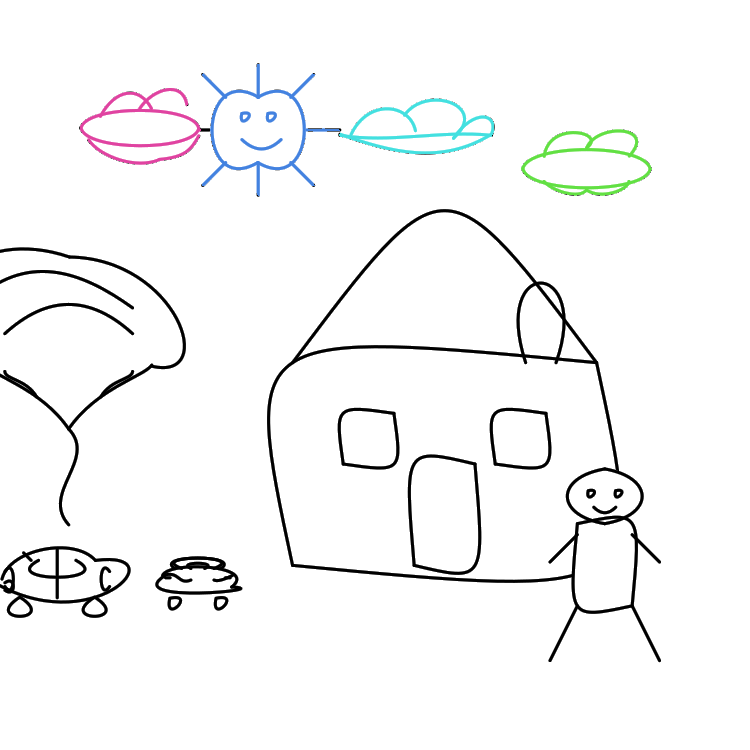}} &
        \frame{\includegraphics[width=0.20\linewidth]{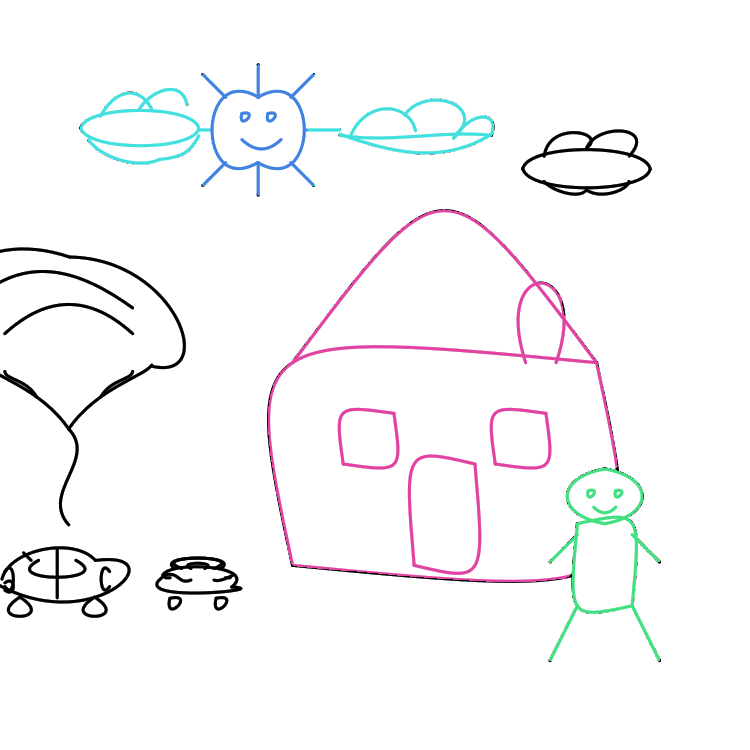}} &
        \frame{\includegraphics[width=0.20\linewidth]{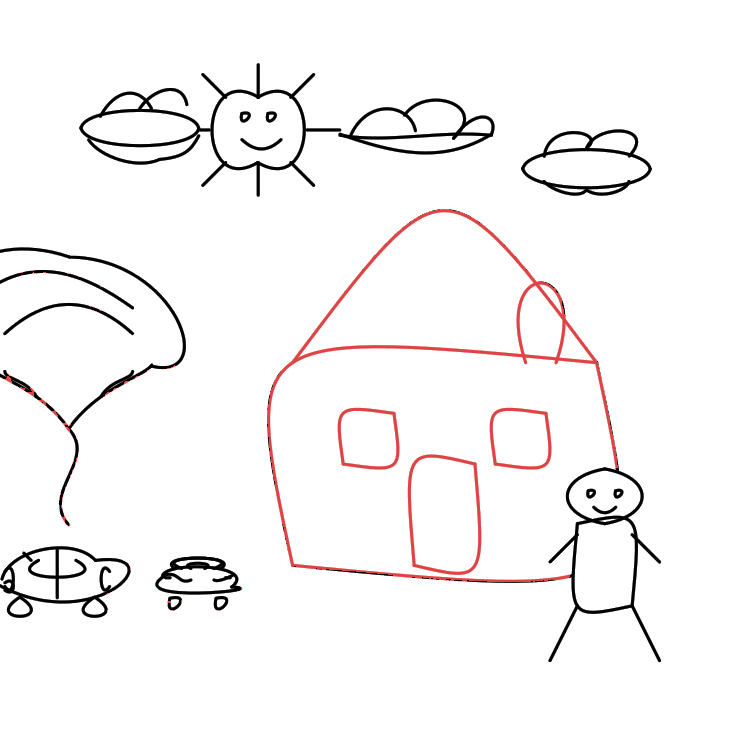}} &
        \frame{\includegraphics[width=0.20\linewidth]{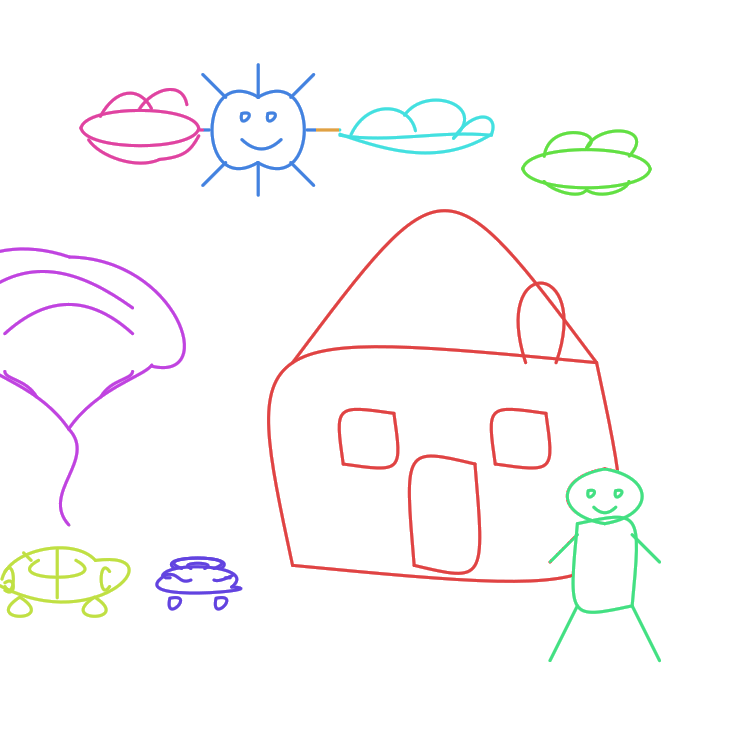}} \\

        \frame{\includegraphics[width=0.20\linewidth]{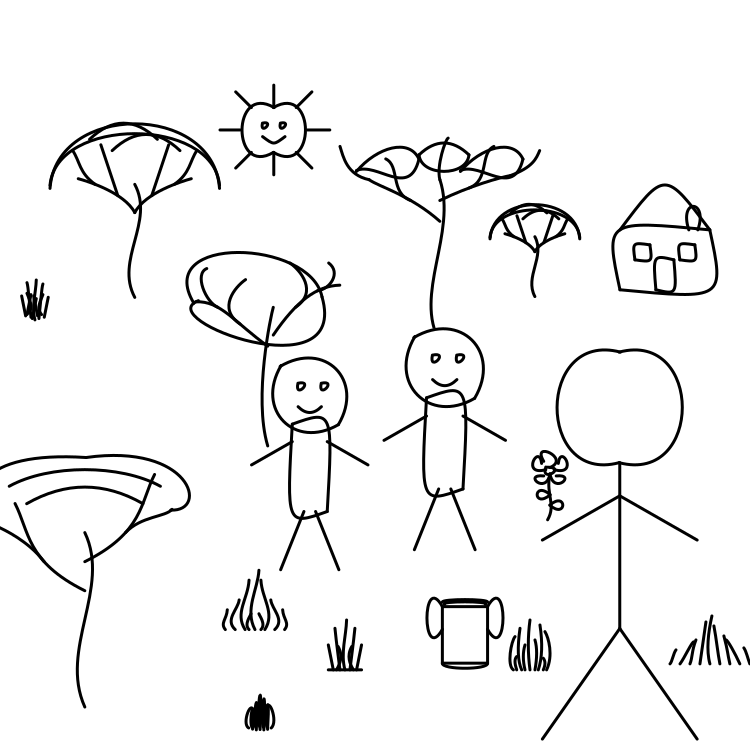}} &
        \frame{\includegraphics[width=0.20\linewidth]{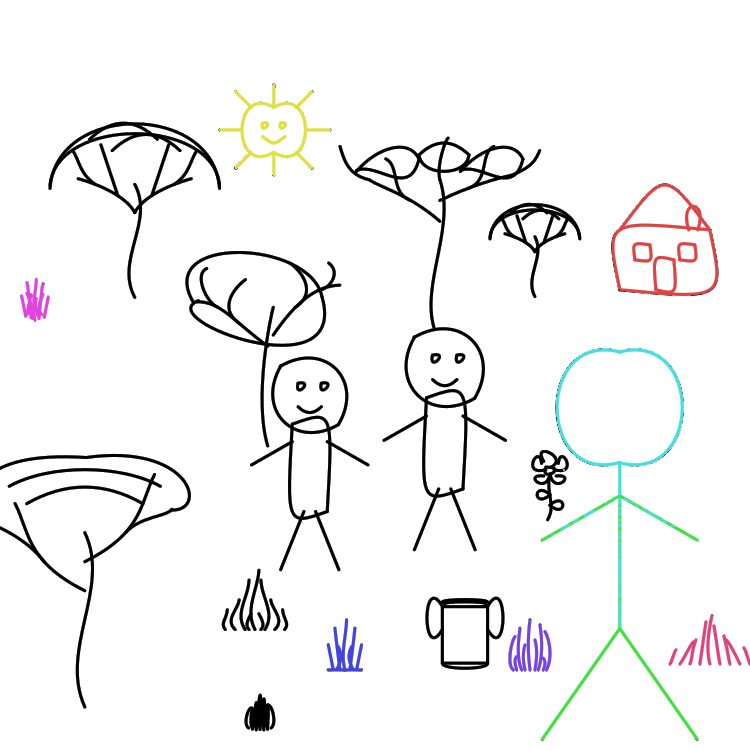}} &
        \frame{\includegraphics[width=0.20\linewidth]{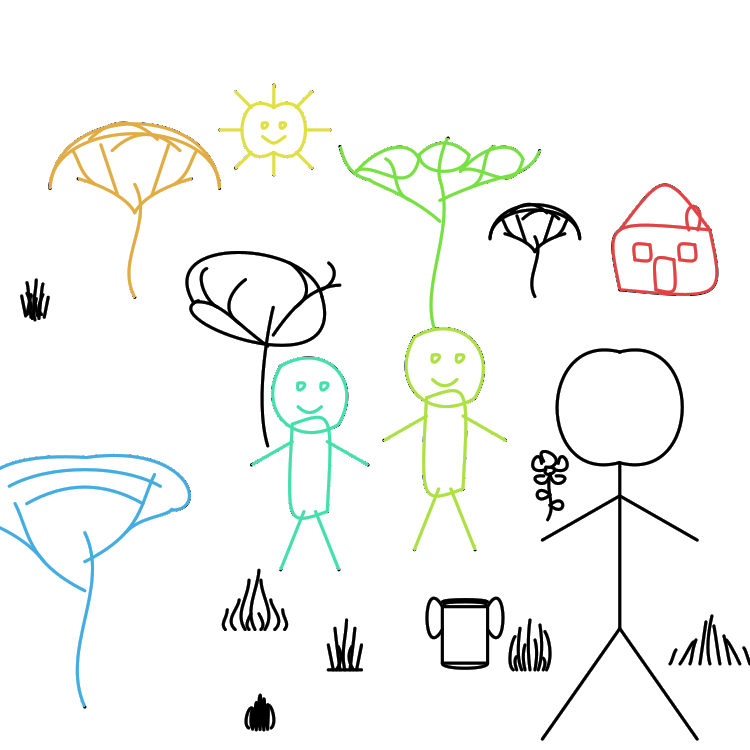}} &
        \frame{\includegraphics[width=0.20\linewidth]{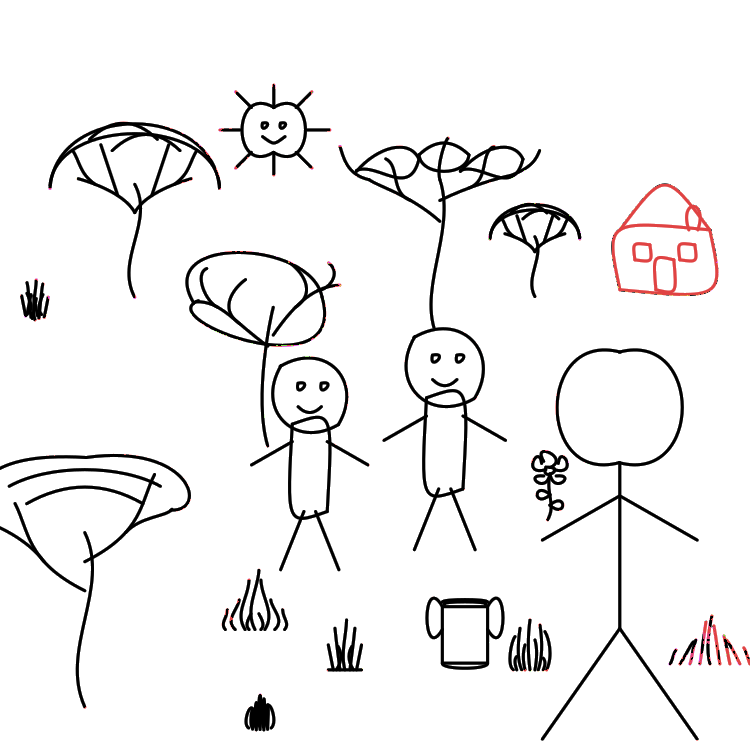}} &
        \frame{\includegraphics[width=0.20\linewidth]{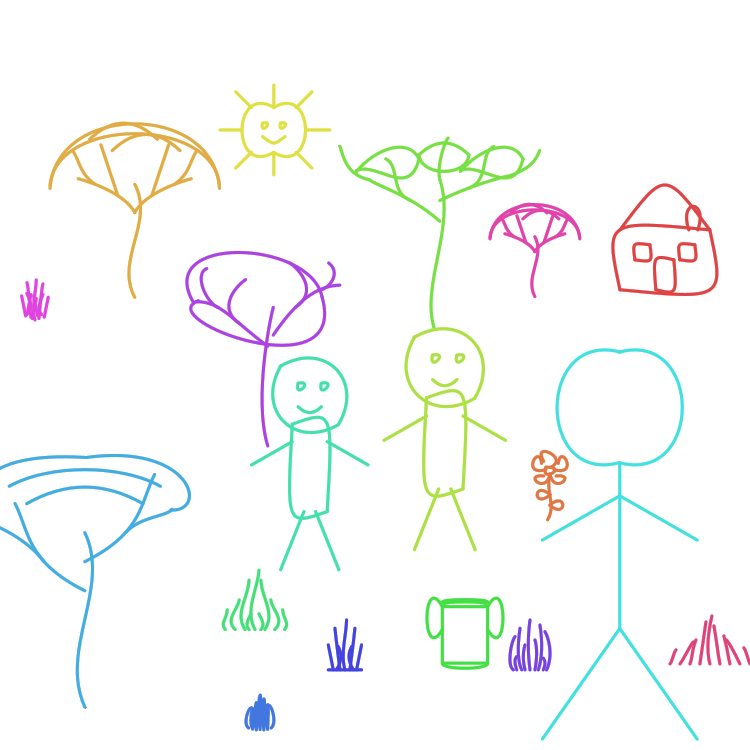}} \\
    \end{tabular}
    }
    \vspace{-0.1cm}
    \caption{\textbf{Qualitative comparison of instance segmentation methods on the SketchAgent dataset.} Both SketchyScene, Grounded SAM and Automatic SAM struggle to segment these abstract sketches, which differ significantly from their training data of clipart-like and real objects. Our method successfully segments individual instances while maintaining object boundaries, even in challenging cases like the last row where objects overlap with a grid-patterned picnic blanket.}
    \label{fig:comparison_instance_sketchagent}
    }
\end{figure*}

\begin{figure*}
    \centering
    \setlength{\tabcolsep}{2pt}
    {\small
    \resizebox{0.92\textwidth}{!}{ 
    \begin{tabular}{c @{\hskip 10pt} c c c c}
        Input & SketchyScene & Grounded SAM & \rev{Automatic SAM} & \textbf{Ours} \\
        
        \frame{\includegraphics[width=0.20\linewidth]{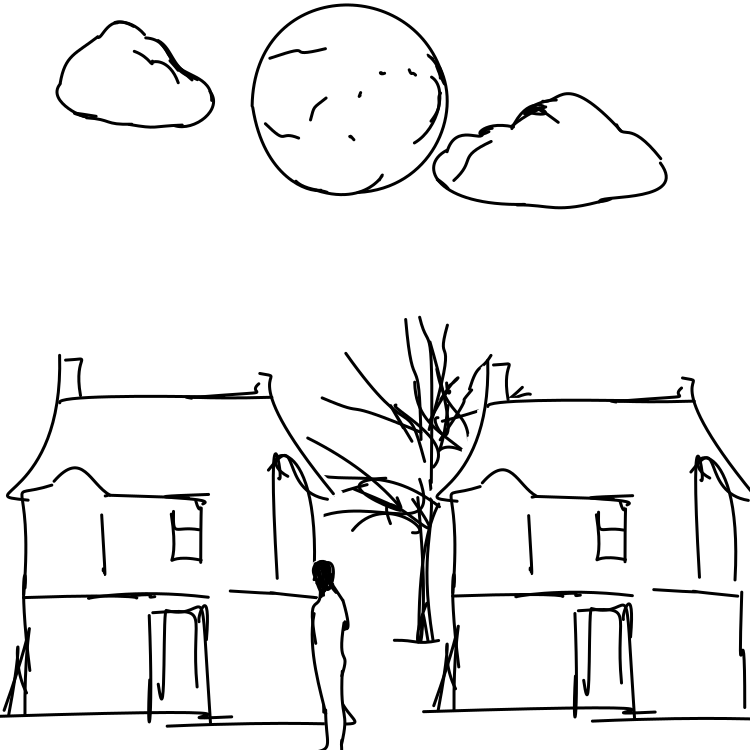}} &
        \frame{\includegraphics[width=0.20\linewidth]{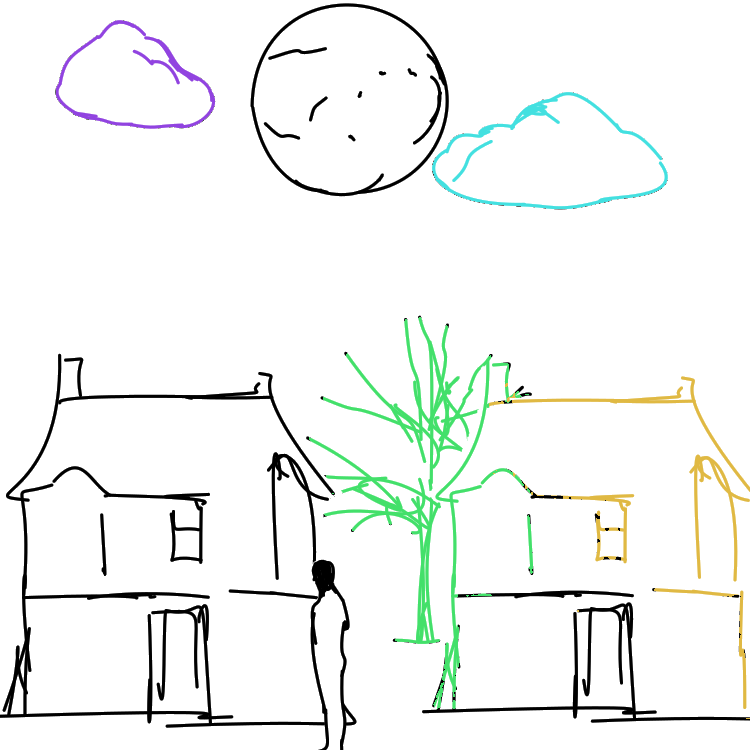}} &
        \frame{\includegraphics[width=0.20\linewidth]{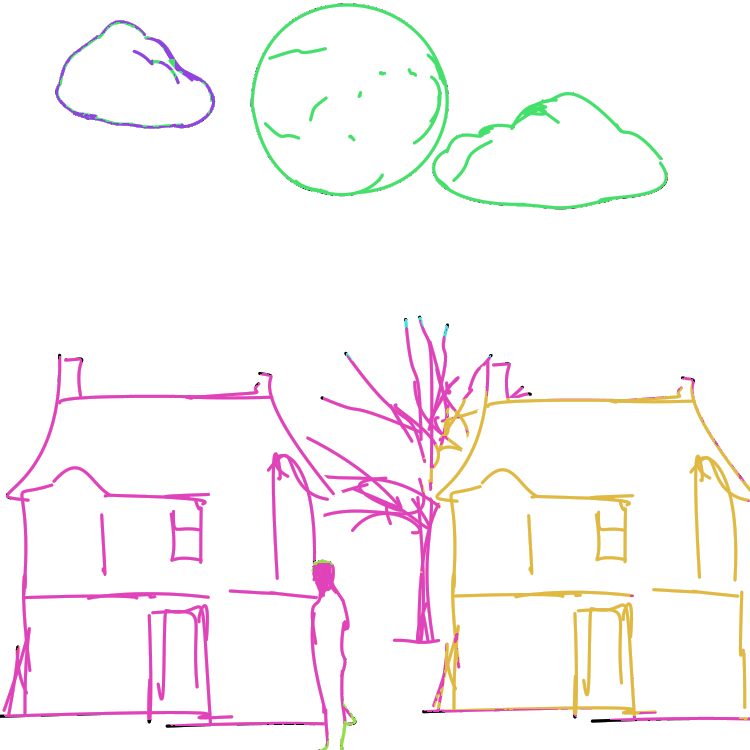}} &
        \frame{\includegraphics[width=0.20\linewidth]{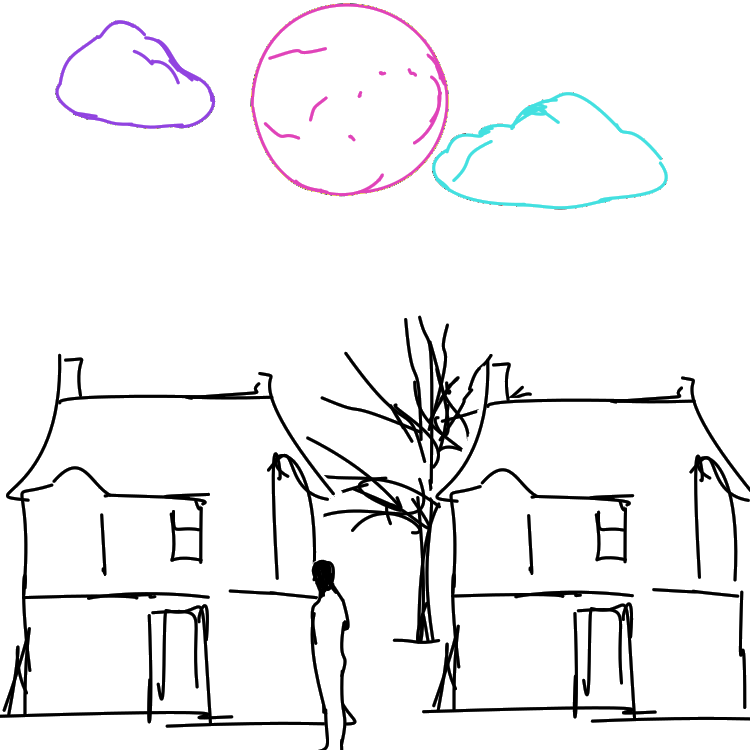}} &
        \frame{\includegraphics[width=0.20\linewidth]{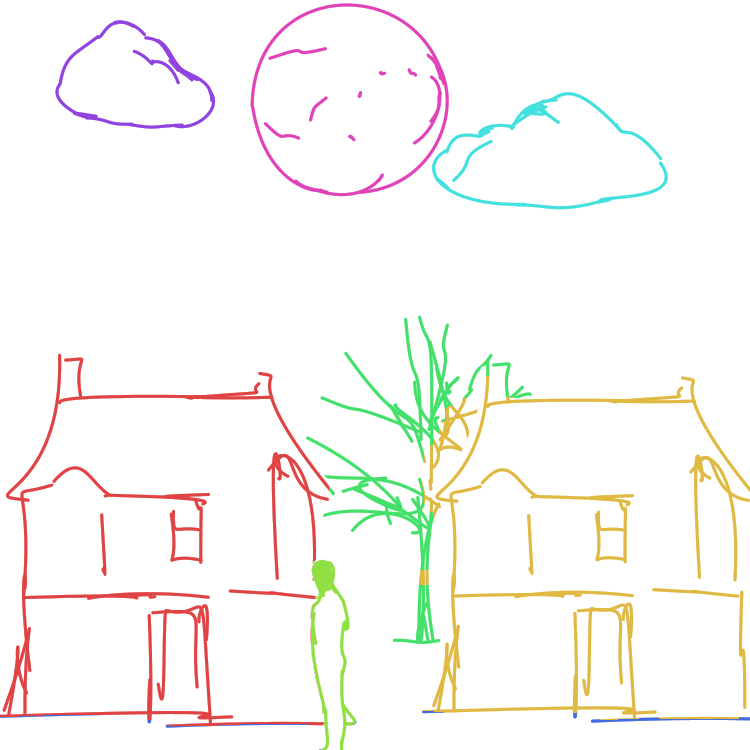}} \\

        \frame{\includegraphics[width=0.20\linewidth]{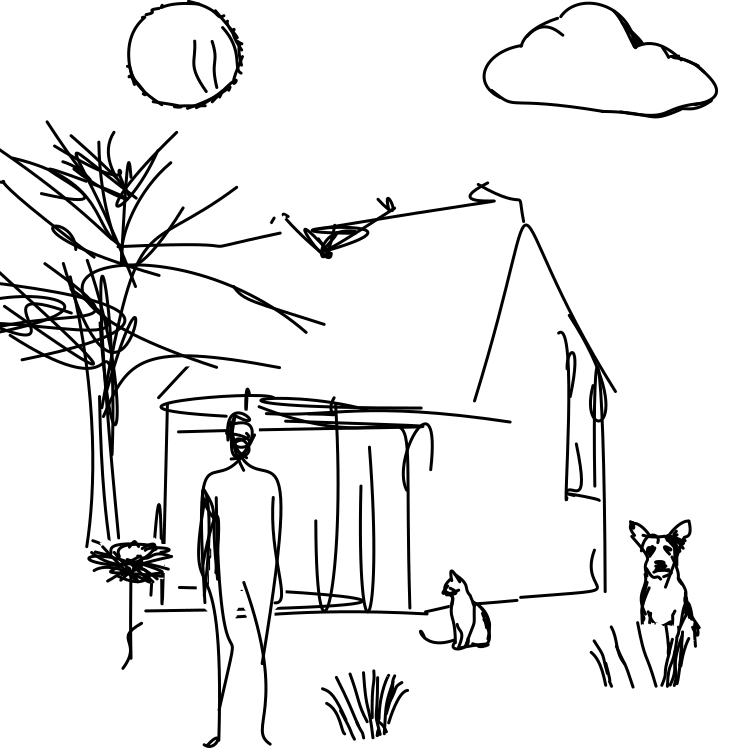}} &
        \frame{\includegraphics[width=0.20\linewidth]{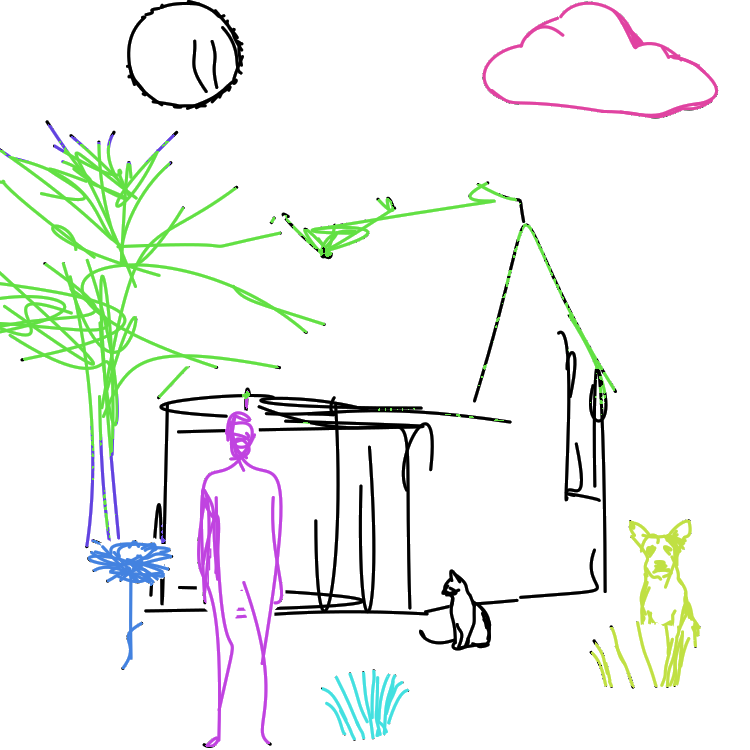}} &
        \frame{\includegraphics[width=0.20\linewidth]{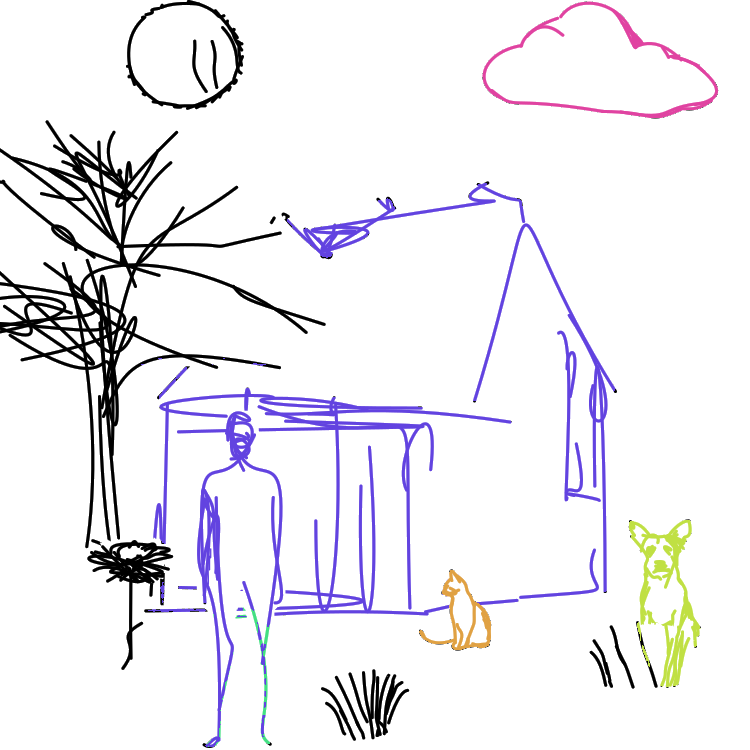}} &
        \frame{\includegraphics[width=0.20\linewidth]{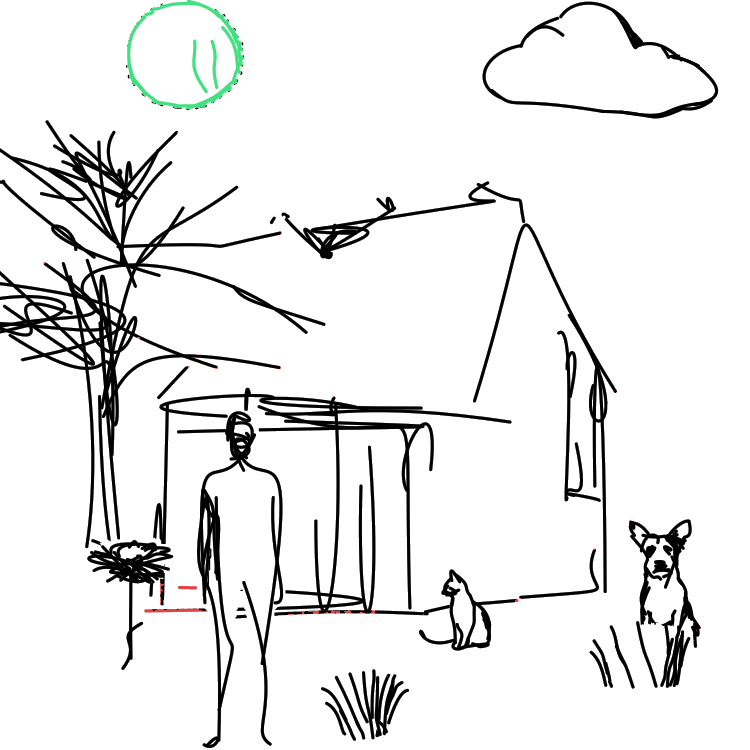}} &
        \frame{\includegraphics[width=0.20\linewidth]{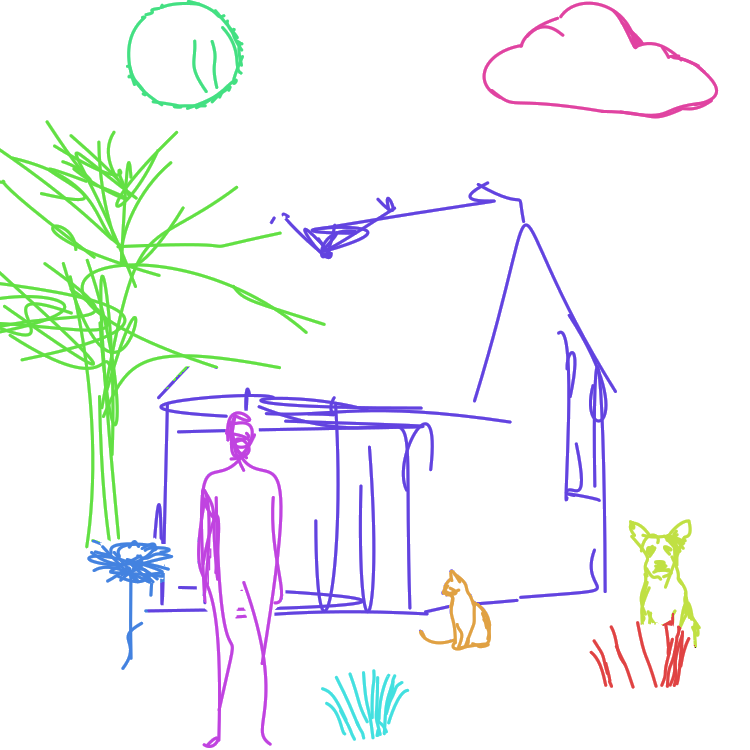}} \\

        \frame{\includegraphics[width=0.20\linewidth]{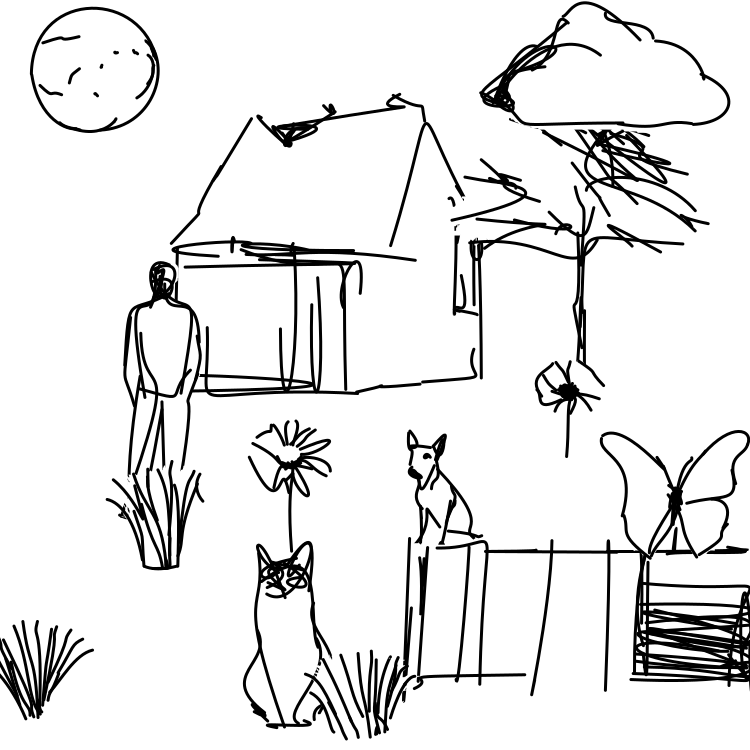}} &
        \frame{\includegraphics[width=0.20\linewidth]{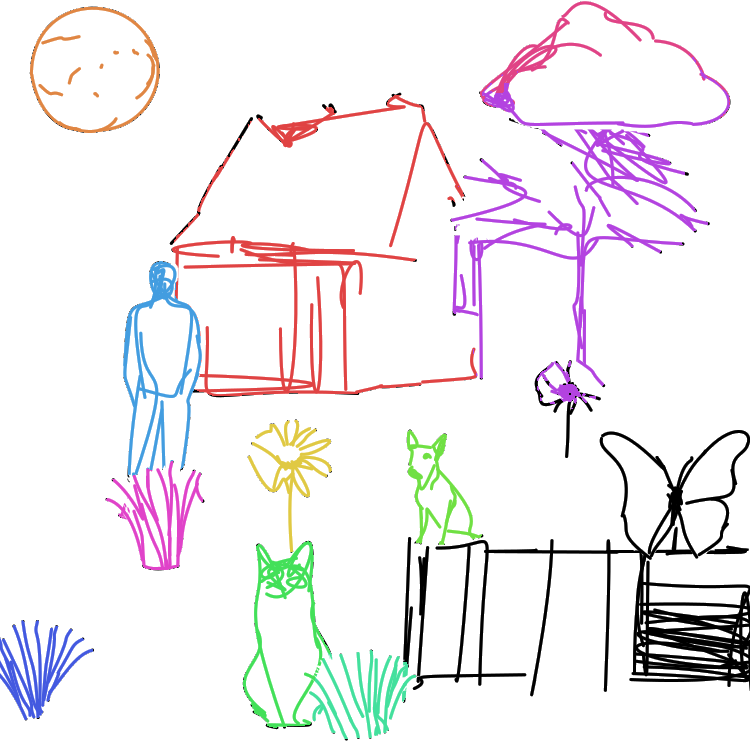}} &
        \frame{\includegraphics[width=0.20\linewidth]{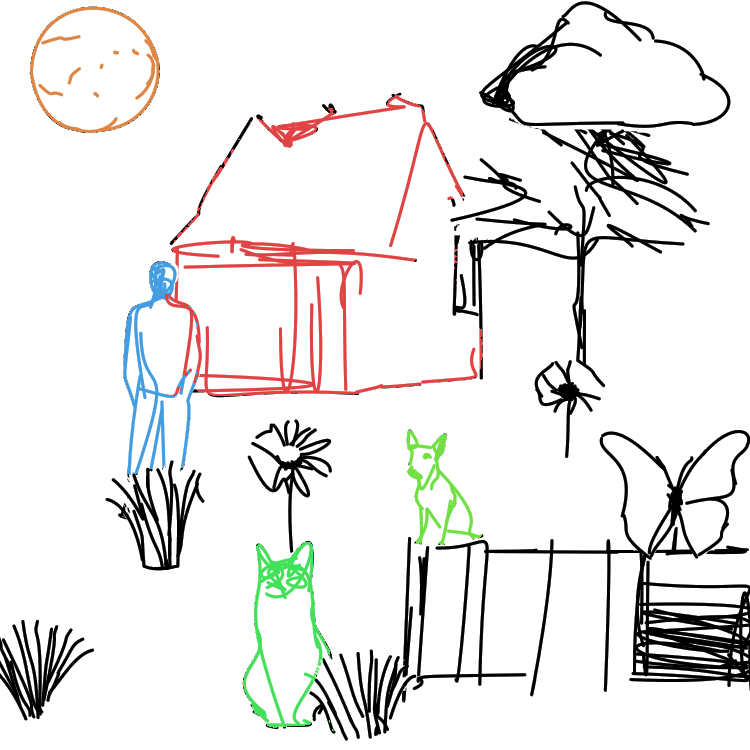}} &
        \frame{\includegraphics[width=0.20\linewidth]{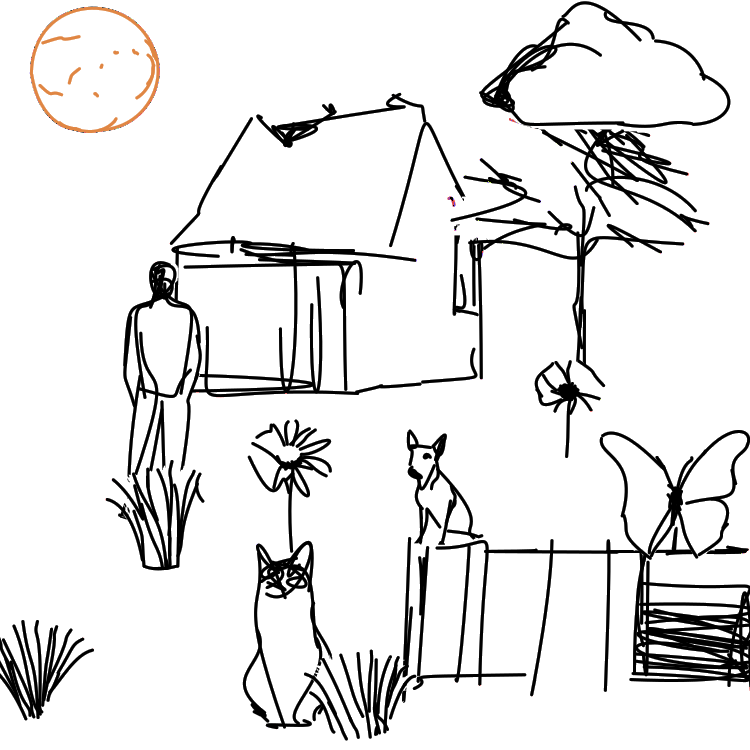}} &
        \frame{\includegraphics[width=0.20\linewidth]{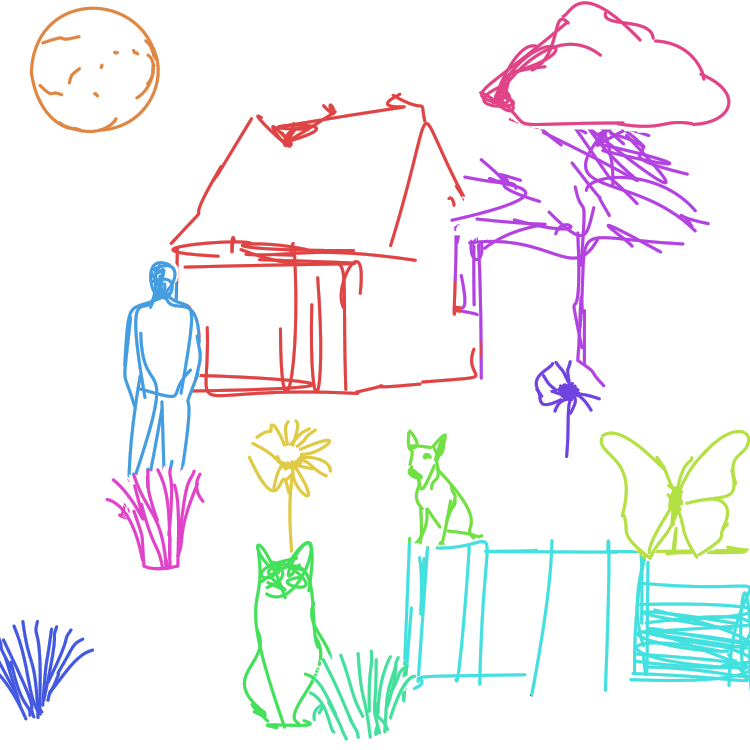}} \\

        \frame{\includegraphics[width=0.20\linewidth]{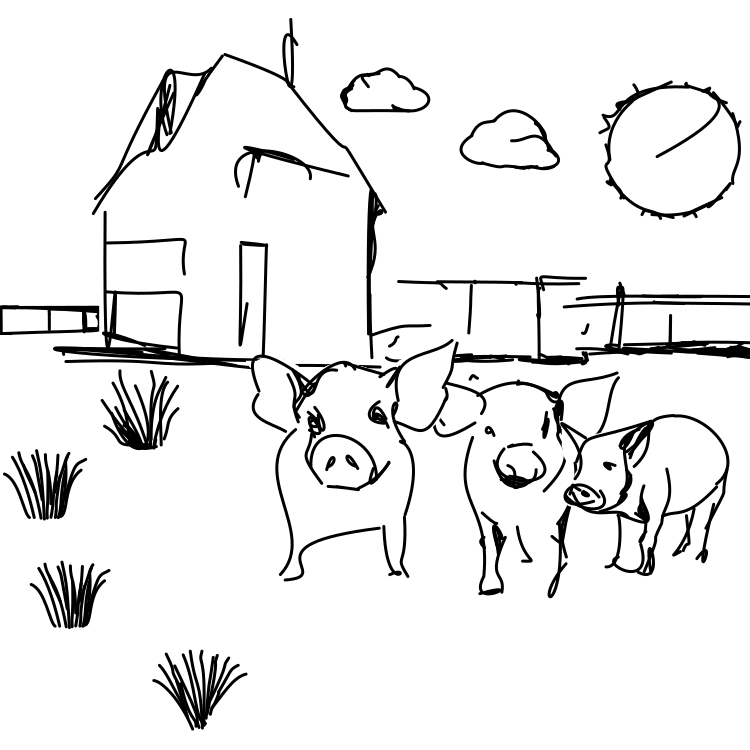}} &
        \frame{\includegraphics[width=0.20\linewidth]{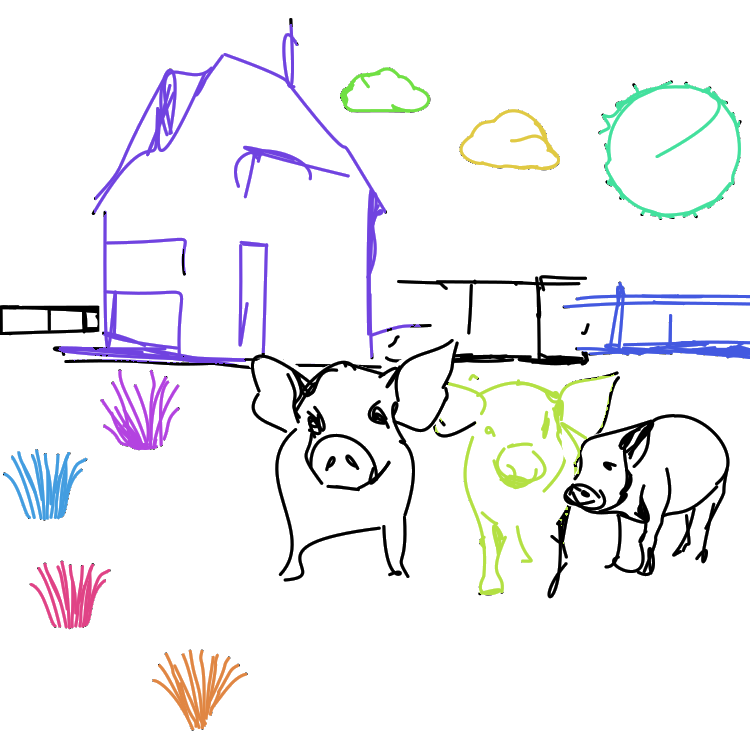}} &
        \frame{\includegraphics[width=0.20\linewidth]{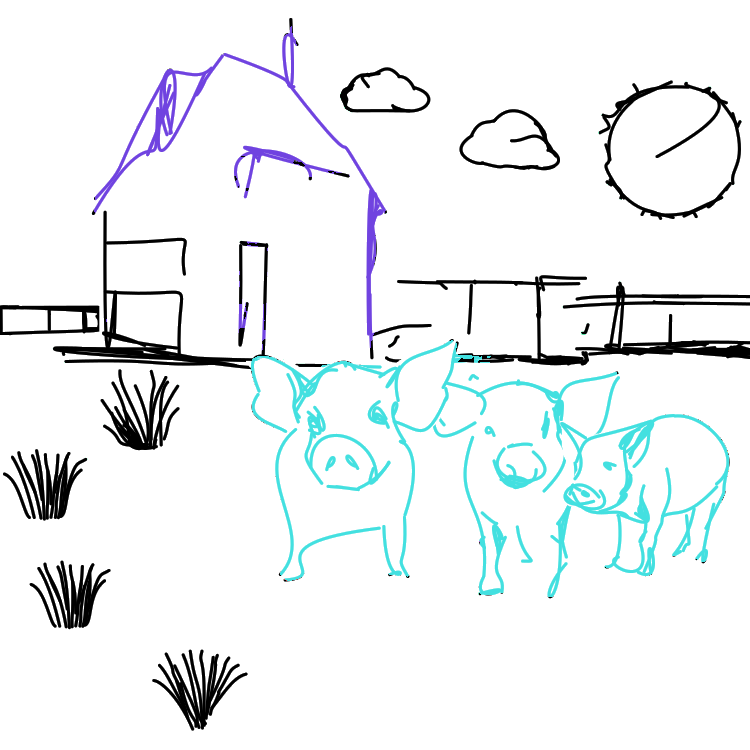}} &
        \frame{\includegraphics[width=0.20\linewidth]{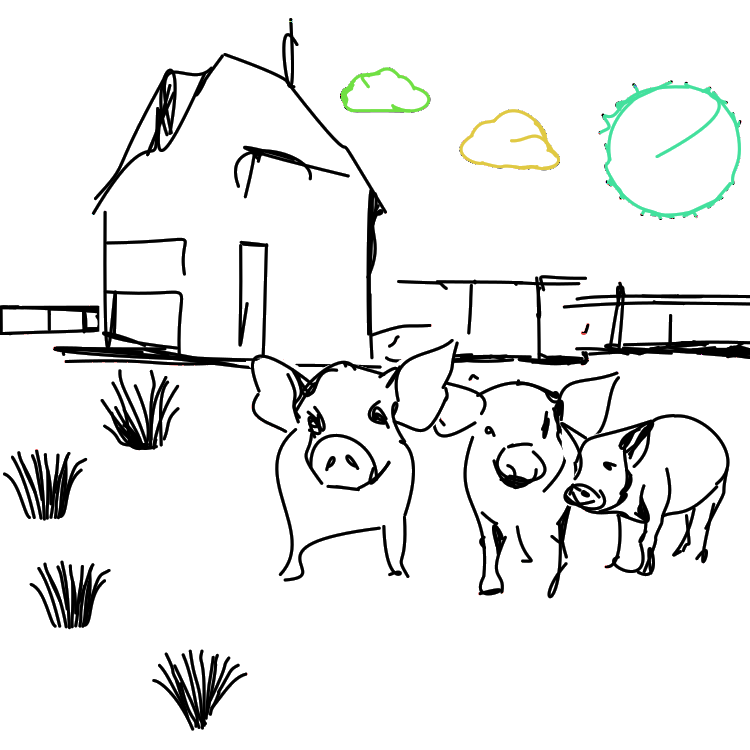}} &
        \frame{\includegraphics[width=0.20\linewidth]{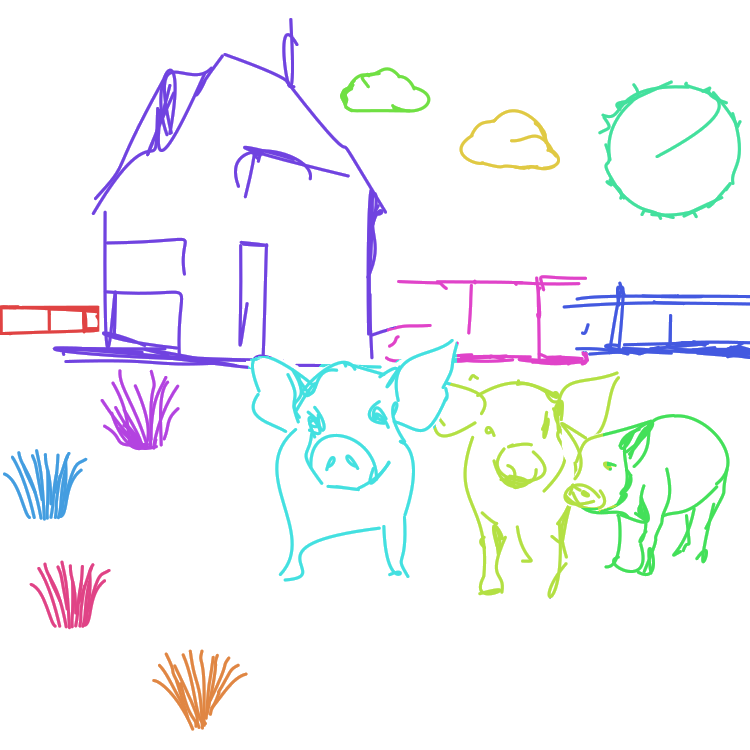}} \\

        \frame{\includegraphics[width=0.20\linewidth]{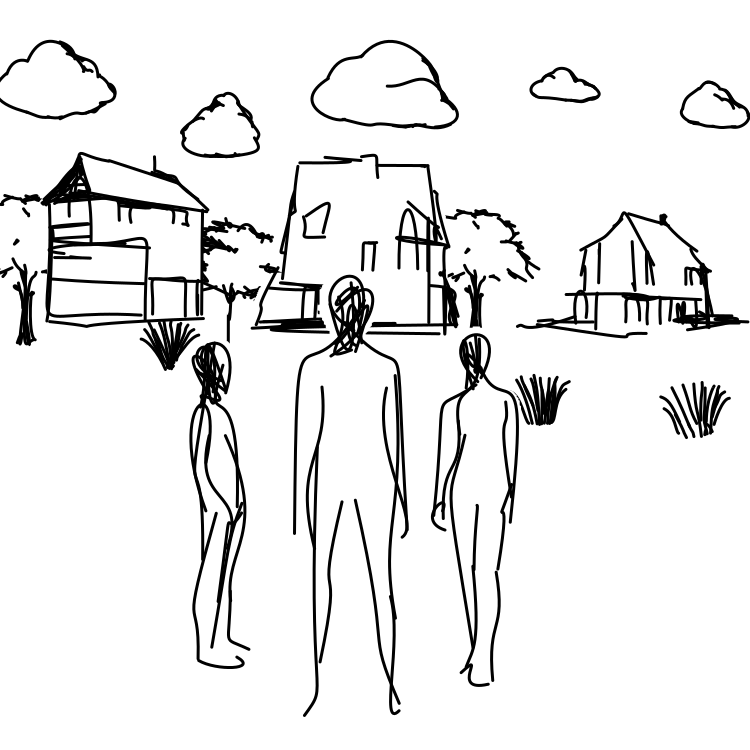}} &
        \frame{\includegraphics[width=0.20\linewidth]{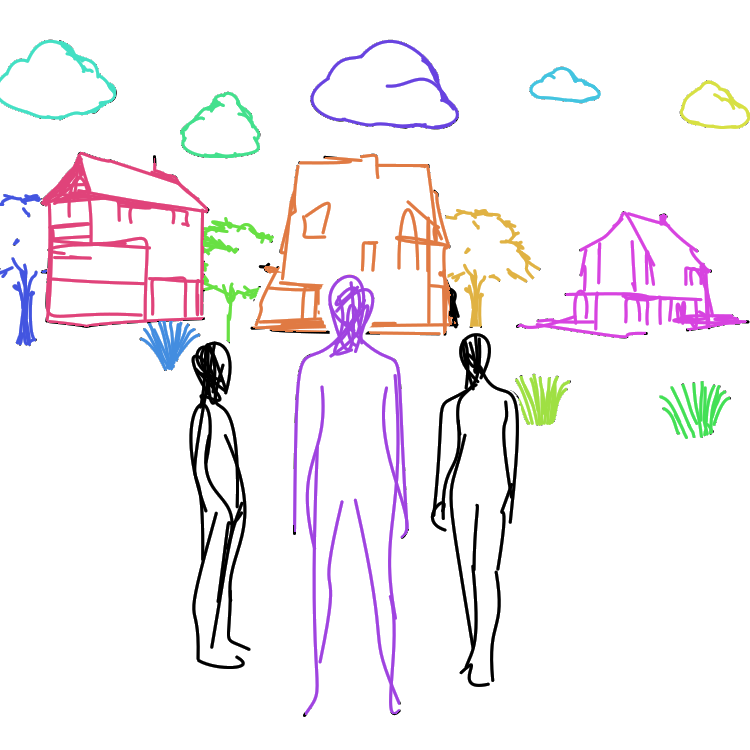}} &
        \frame{\includegraphics[width=0.20\linewidth]{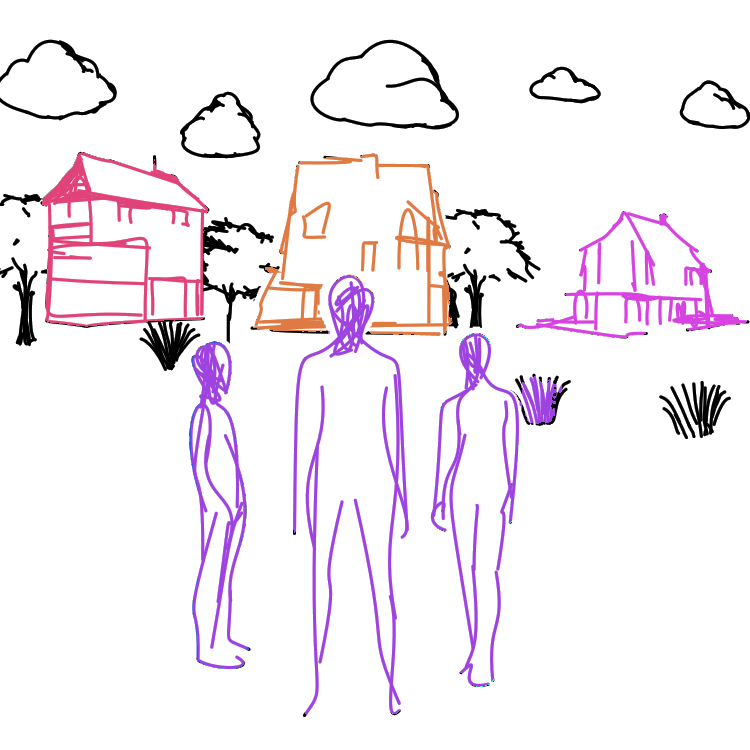}} &
        \frame{\includegraphics[width=0.20\linewidth]{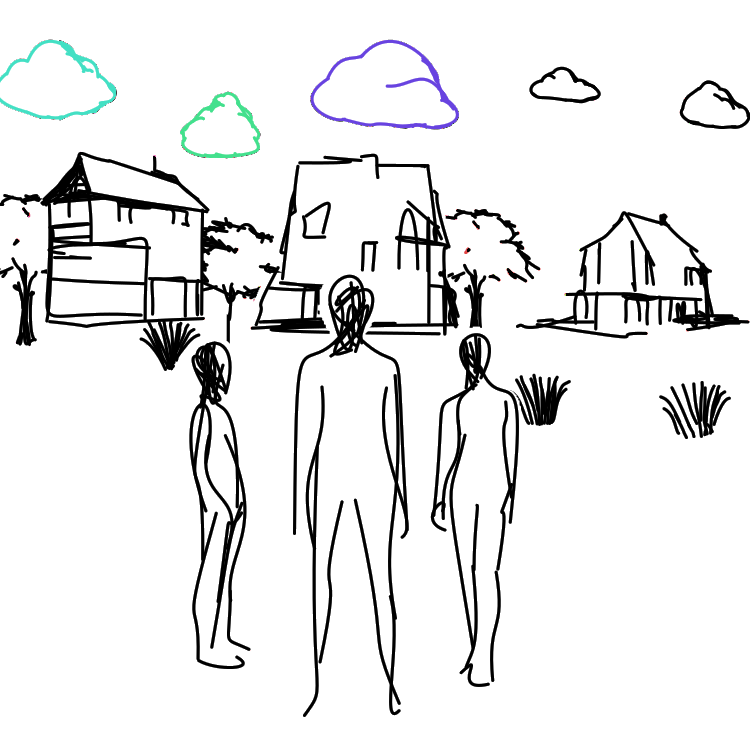}} &
        \frame{\includegraphics[width=0.20\linewidth]{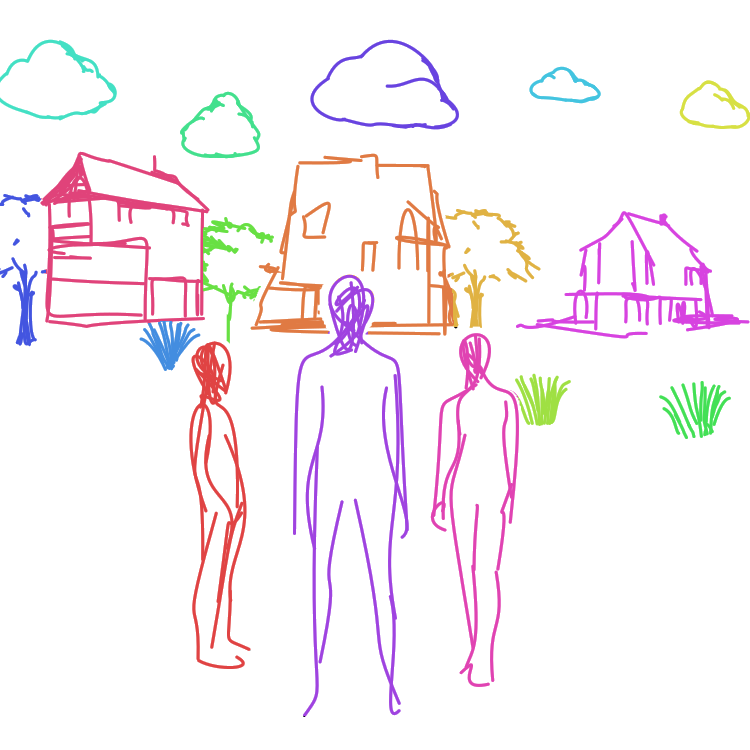}} \\

        \frame{\includegraphics[width=0.20\linewidth]{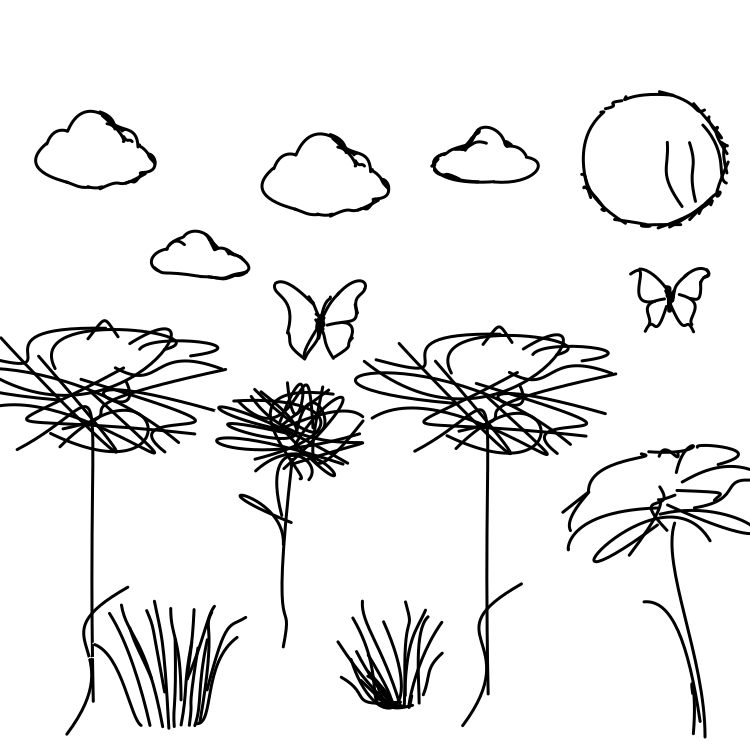}} &
        \frame{\includegraphics[width=0.20\linewidth]{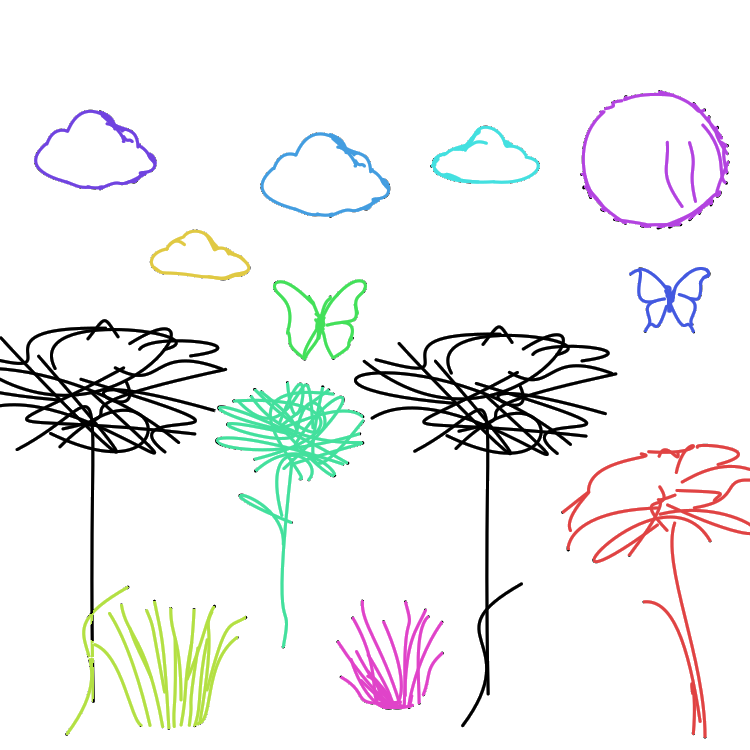}} &
        \frame{\includegraphics[width=0.20\linewidth]{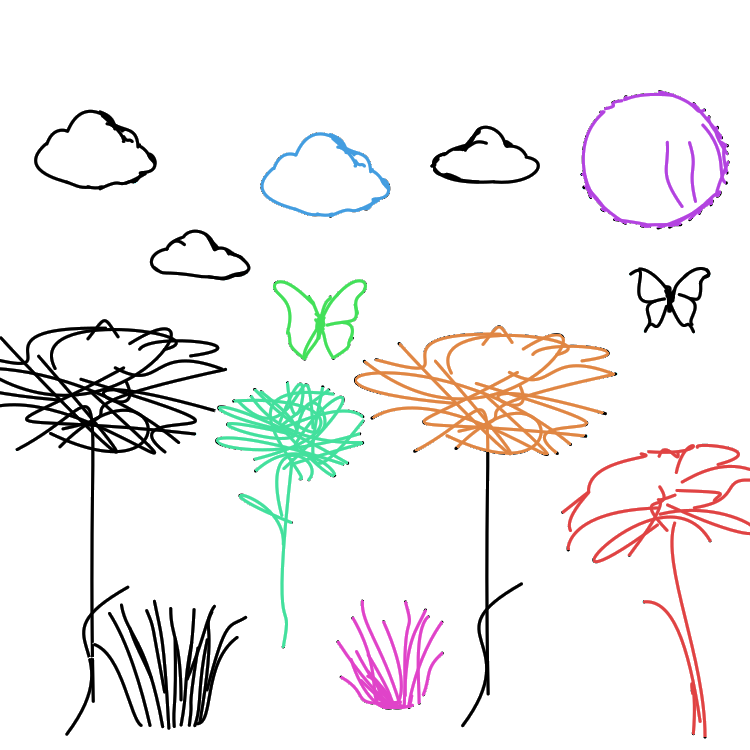}} &
        \frame{\includegraphics[width=0.20\linewidth]{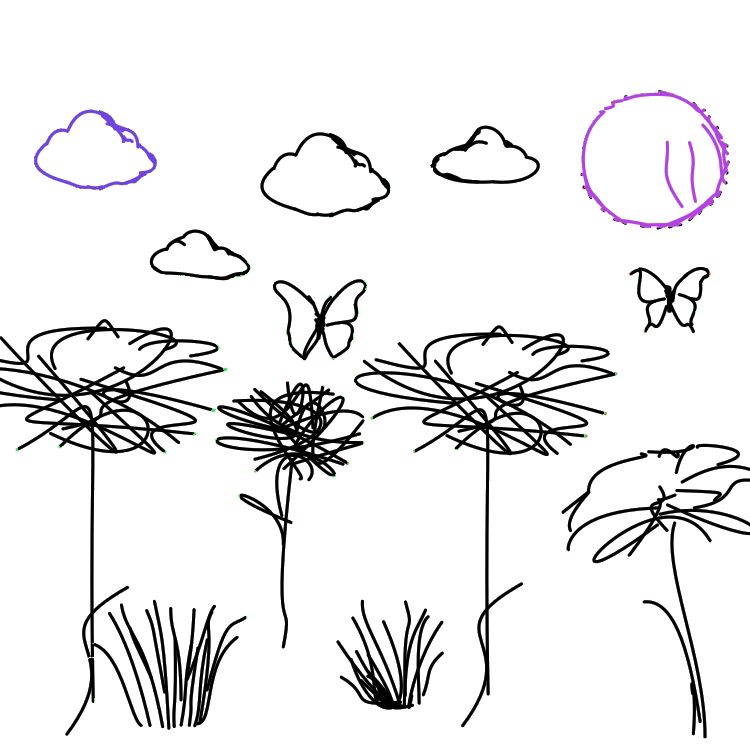}} &
        \frame{\includegraphics[width=0.20\linewidth]{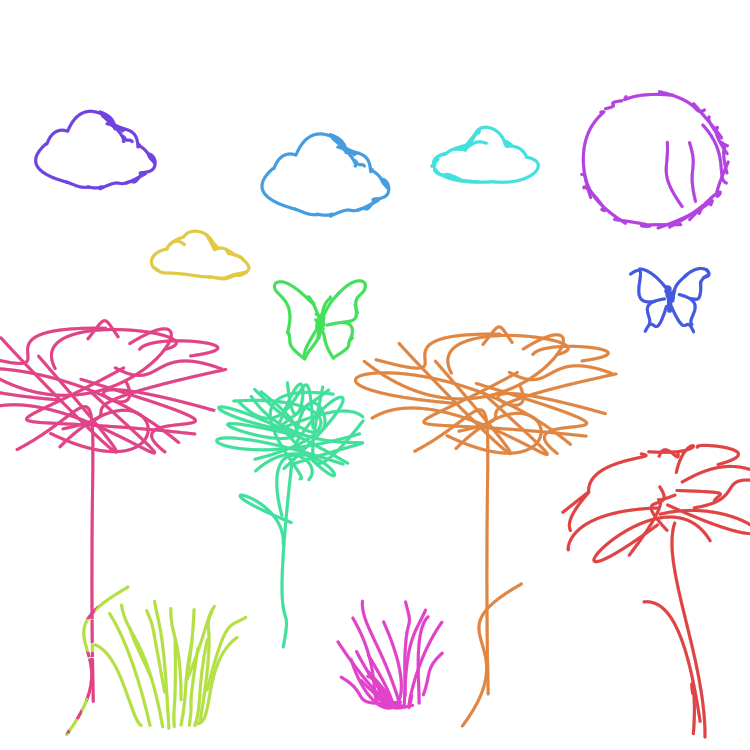}} \\
    \end{tabular}
    }
    \vspace{-0.1cm}
    \caption{\textbf{Qualitative comparison of instance segmentation methods on the CLIPasso base dataset.} Our method successfully detects both large and small objects with ambiguous openings in their silhouettes, whereas baseline methods either merge multiple instances into one or fail to detect them entirely.}
    \label{fig:comparison_instance_clipasso}
    }
\end{figure*}

\begin{figure*}
    \centering
    \setlength{\tabcolsep}{2pt}
    {\small
    \resizebox{0.92\textwidth}{!}{ 
    \begin{tabular}{c @{\hskip 10pt} c c c c}
        Input & SketchyScene & Grounded SAM & \rev{Automatic SAM} & \textbf{Ours} \\
        
        \frame{\includegraphics[width=0.20\linewidth]{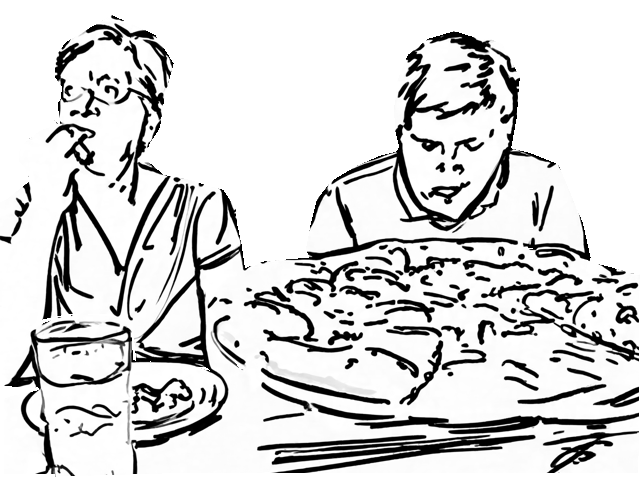}} &
        \frame{\includegraphics[width=0.20\linewidth]{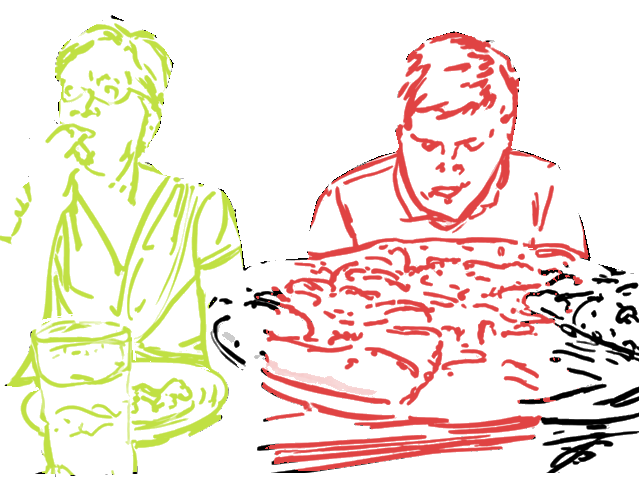}} &
        \frame{\includegraphics[width=0.20\linewidth]{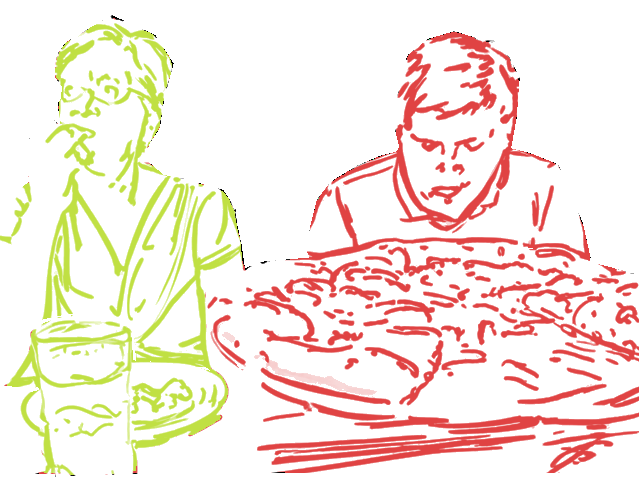}} &
        \frame{\includegraphics[width=0.20\linewidth]{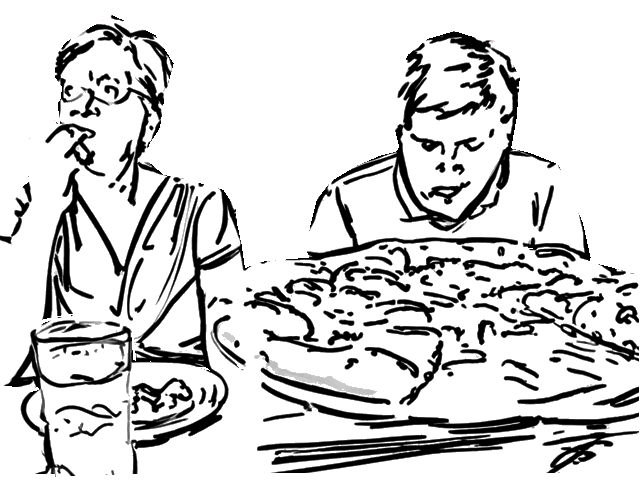}} &
        \frame{\includegraphics[width=0.20\linewidth]{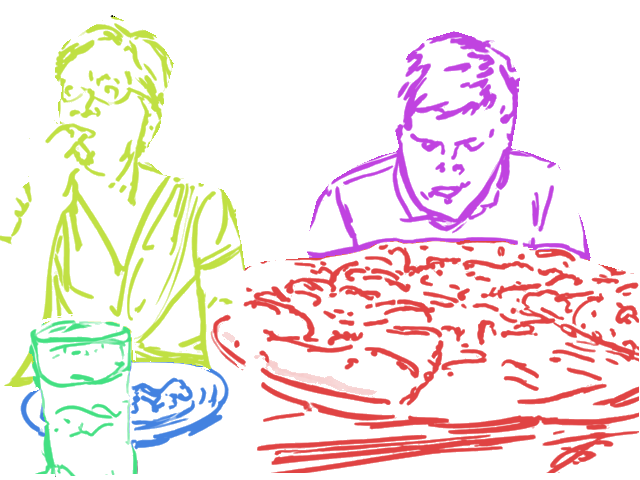}} \\

         \frame{\includegraphics[width=0.20\linewidth]{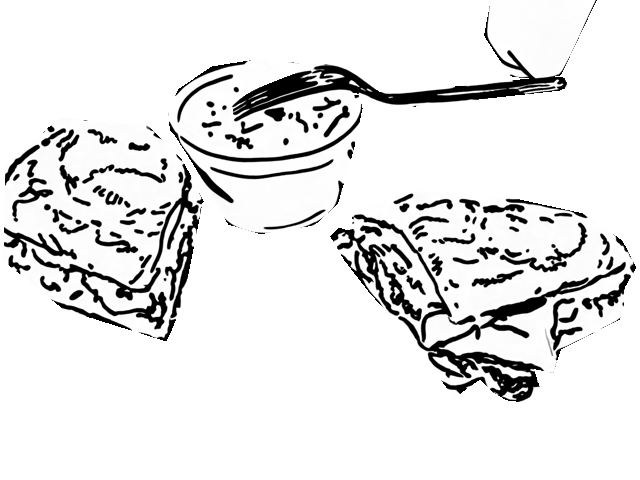}} &
        \frame{\includegraphics[width=0.20\linewidth]{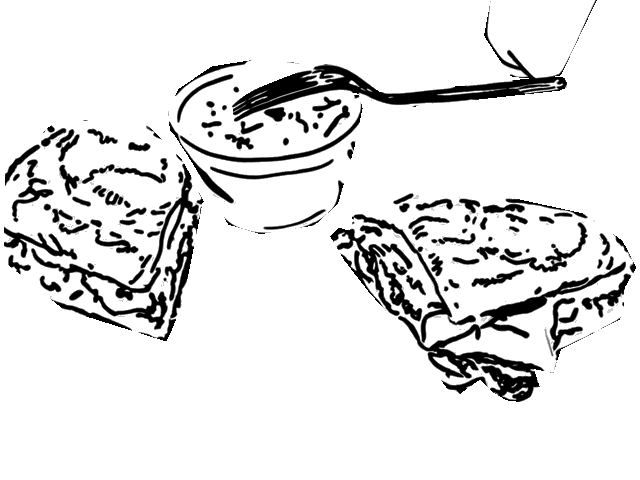}} &
        \frame{\includegraphics[width=0.20\linewidth]{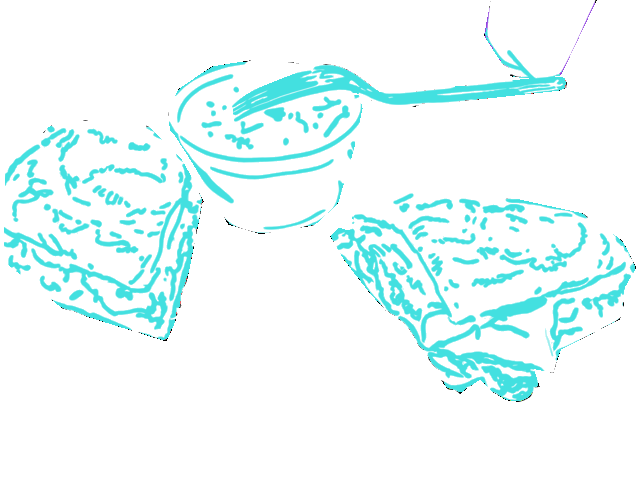}} &
        \frame{\includegraphics[width=0.20\linewidth]{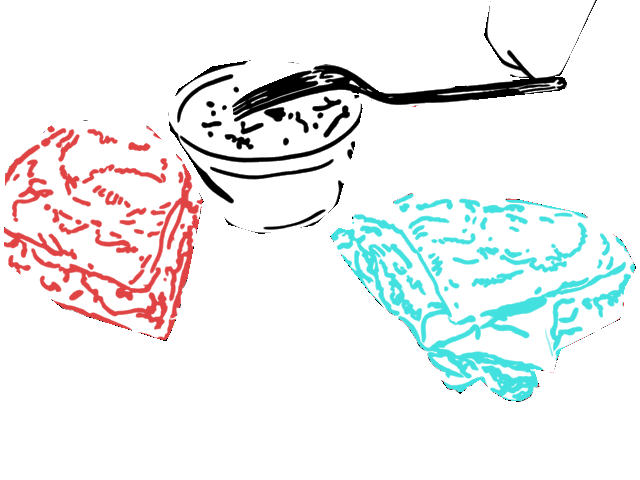}} &
        \frame{\includegraphics[width=0.20\linewidth]{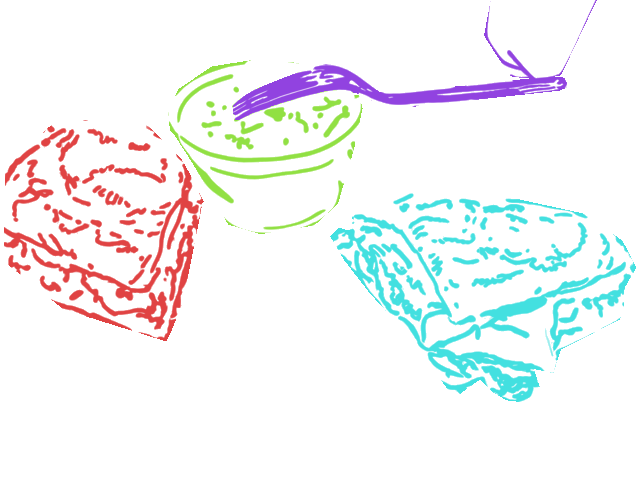}} \\

         \frame{\includegraphics[width=0.20\linewidth]{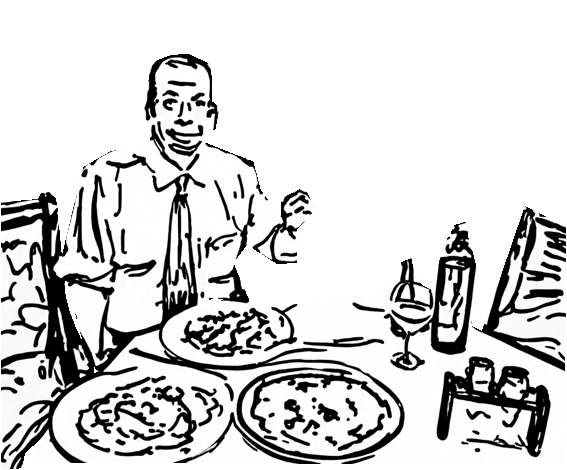}} &
        \frame{\includegraphics[width=0.20\linewidth]{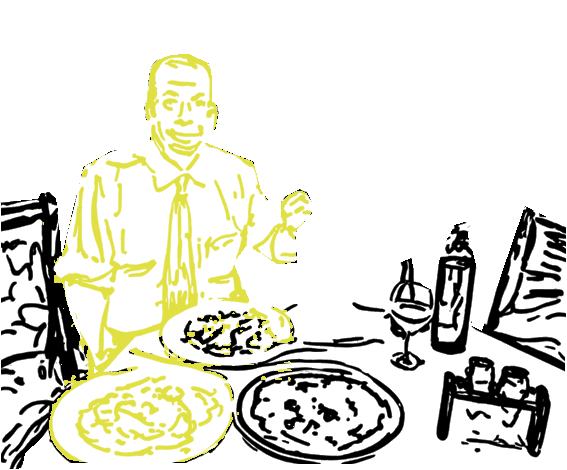}} &
        \frame{\includegraphics[width=0.20\linewidth]{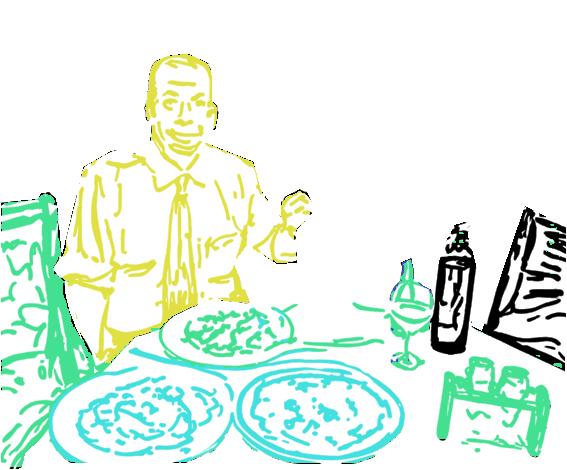}} &
        \frame{\includegraphics[width=0.20\linewidth]{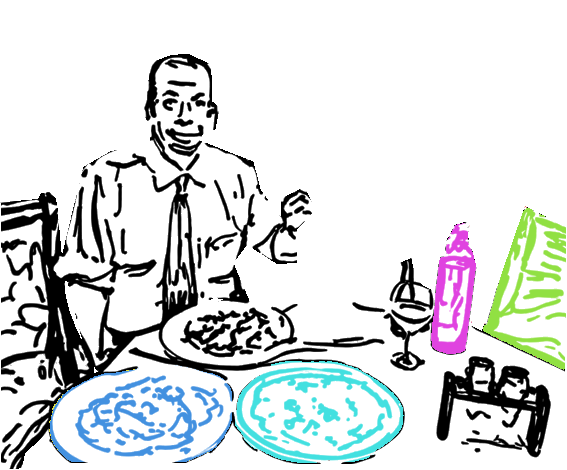}} &
        \frame{\includegraphics[width=0.20\linewidth]{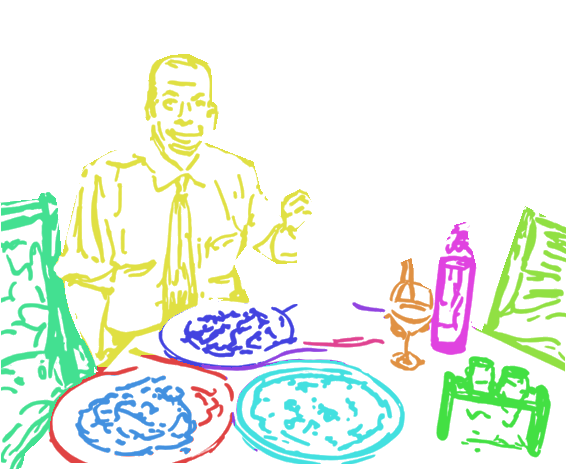}} \\

         \frame{\includegraphics[width=0.20\linewidth]{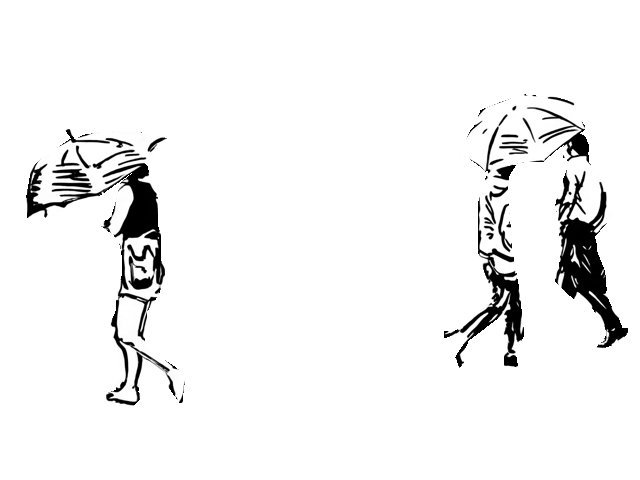}} &
        \frame{\includegraphics[width=0.20\linewidth]{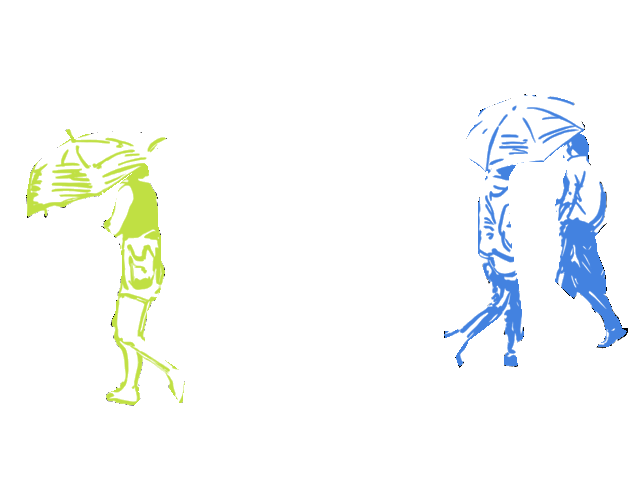}} &
        \frame{\includegraphics[width=0.20\linewidth]{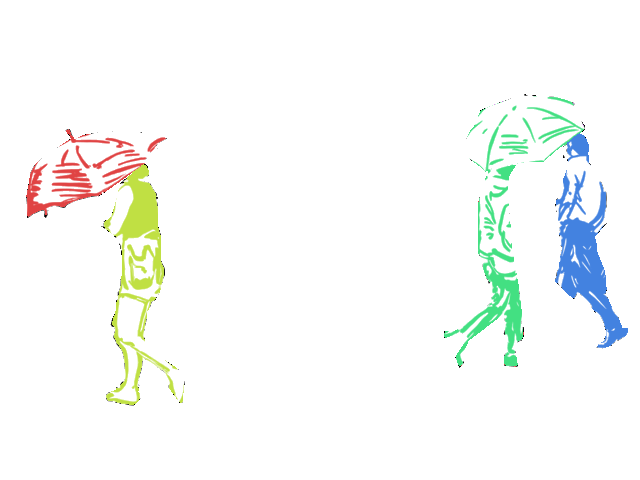}} &
        \frame{\includegraphics[width=0.20\linewidth]{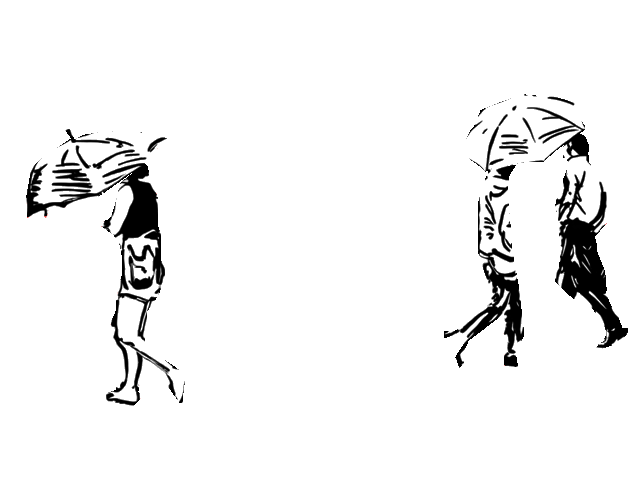}} &
        \frame{\includegraphics[width=0.20\linewidth]{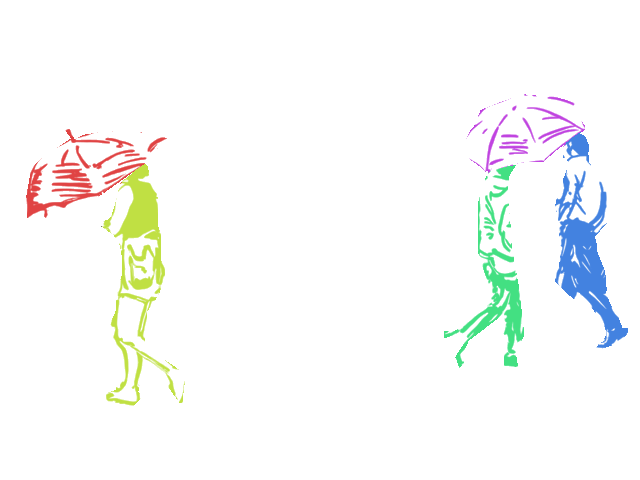}} \\

         \frame{\includegraphics[width=0.20\linewidth]{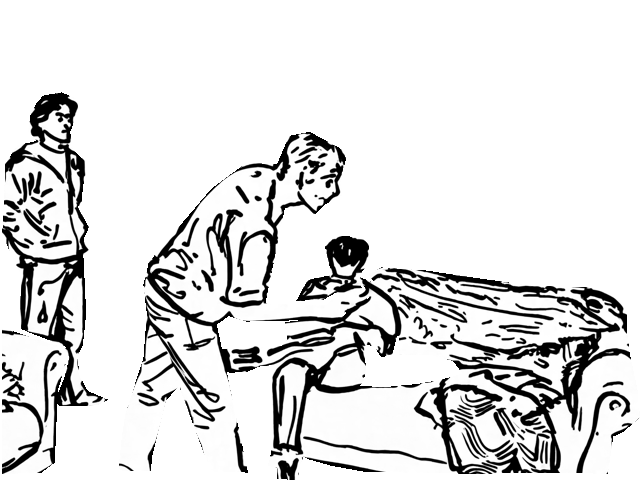}} &
        \frame{\includegraphics[width=0.20\linewidth]{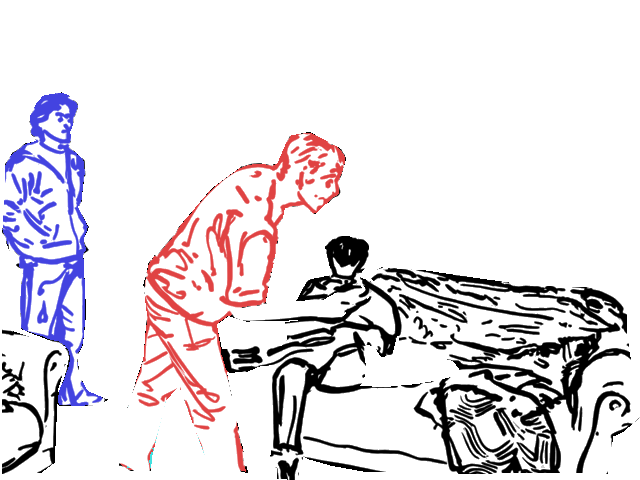}} &
        \frame{\includegraphics[width=0.20\linewidth]{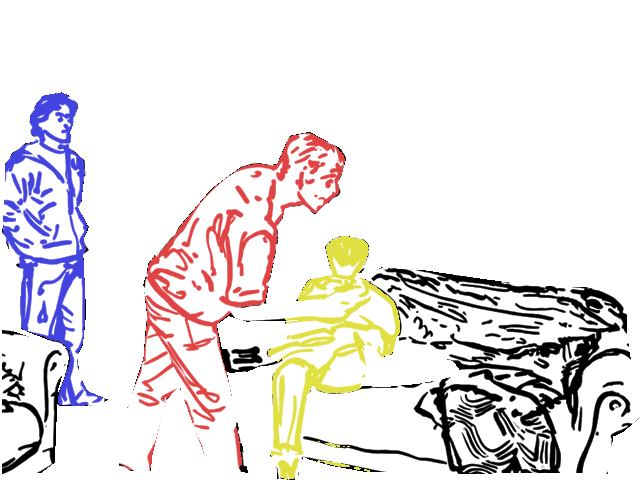}} &
        \frame{\includegraphics[width=0.20\linewidth]{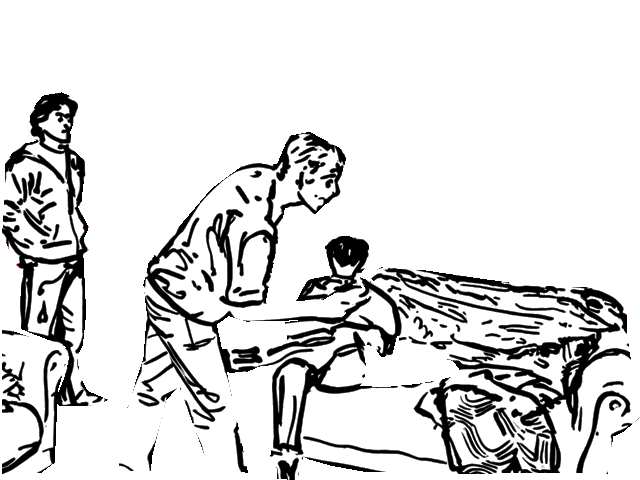}} &
        \frame{\includegraphics[width=0.20\linewidth]{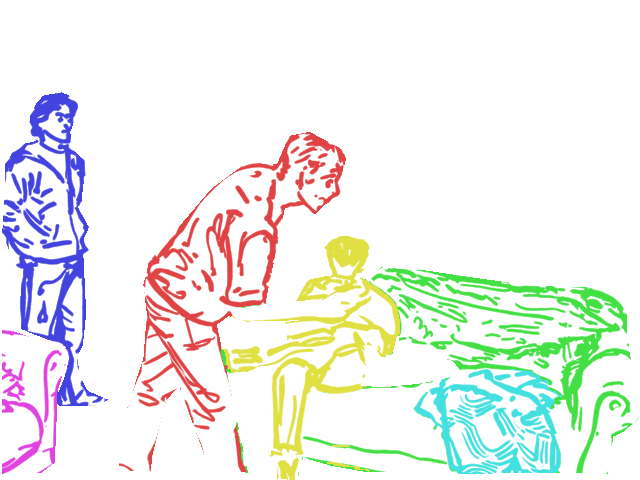}} \\

         \frame{\includegraphics[width=0.20\linewidth]{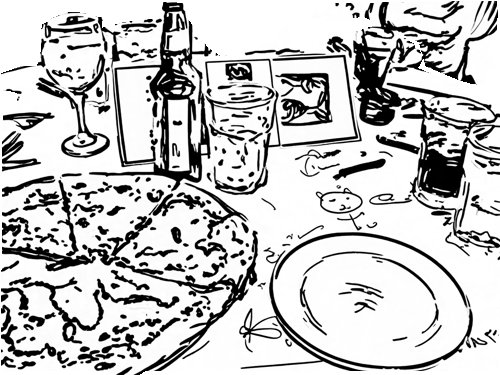}} &
        \frame{\includegraphics[width=0.20\linewidth]{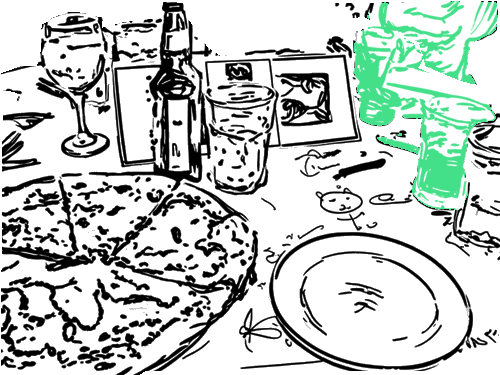}} &
        \frame{\includegraphics[width=0.20\linewidth]{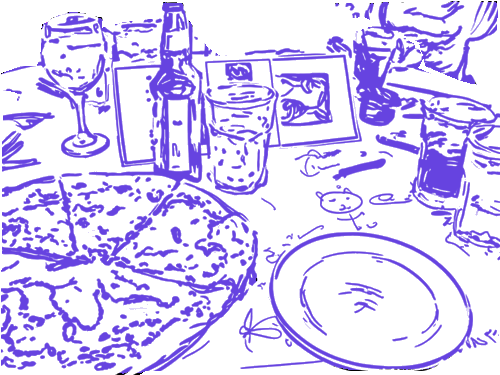}} &
        \frame{\includegraphics[width=0.20\linewidth]{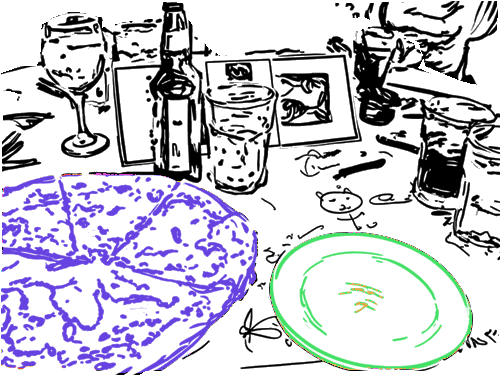}} &
        \frame{\includegraphics[width=0.20\linewidth]{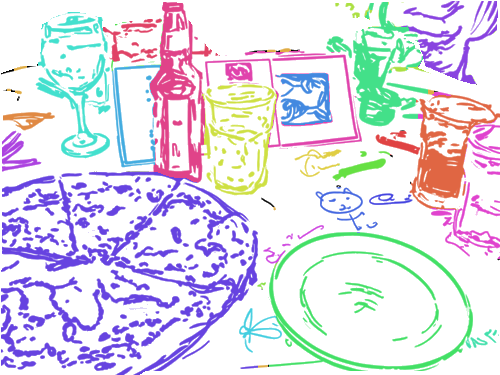}} \\

         \frame{\includegraphics[width=0.20\linewidth]{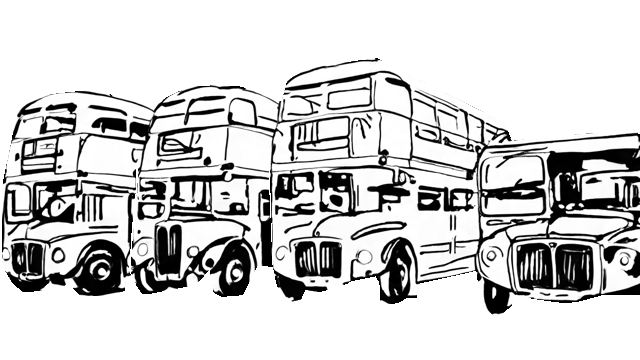}} &
        \frame{\includegraphics[width=0.20\linewidth]{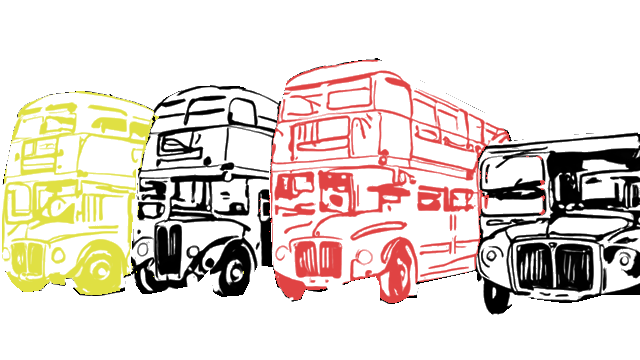}} &
        \frame{\includegraphics[width=0.20\linewidth]{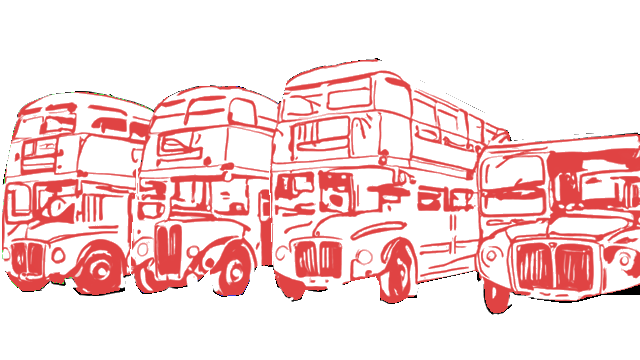}} &
        \frame{\includegraphics[width=0.20\linewidth]{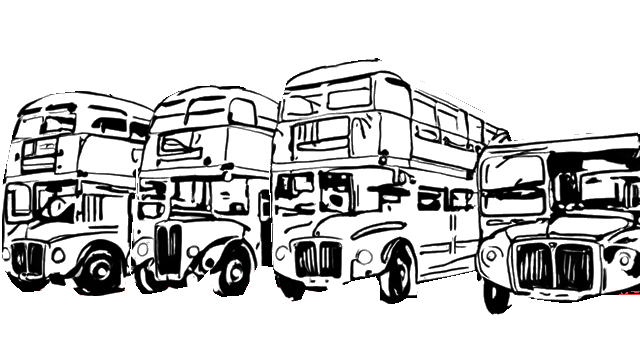}} &
        \frame{\includegraphics[width=0.20\linewidth]{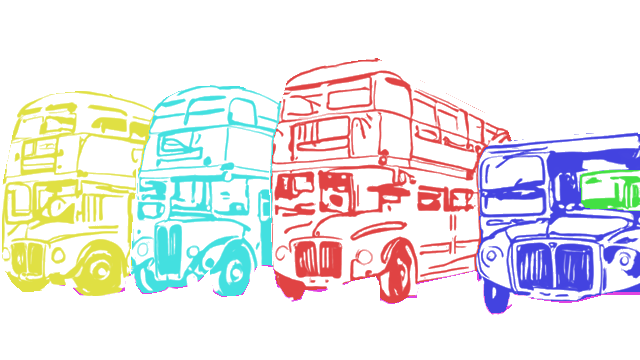}} \\

        \frame{\includegraphics[width=0.20\linewidth]{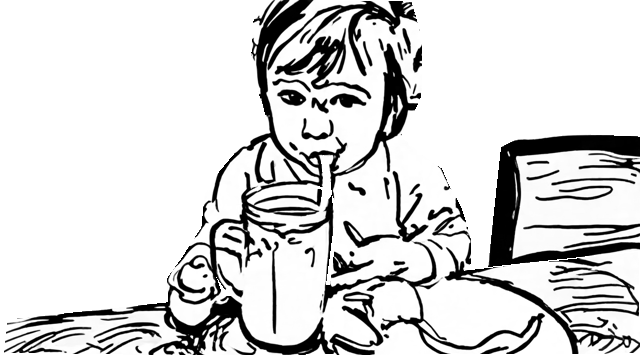}} &
        \frame{\includegraphics[width=0.20\linewidth]{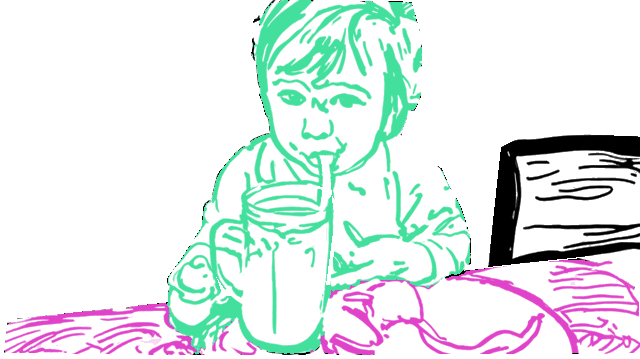}} &
        \frame{\includegraphics[width=0.20\linewidth]{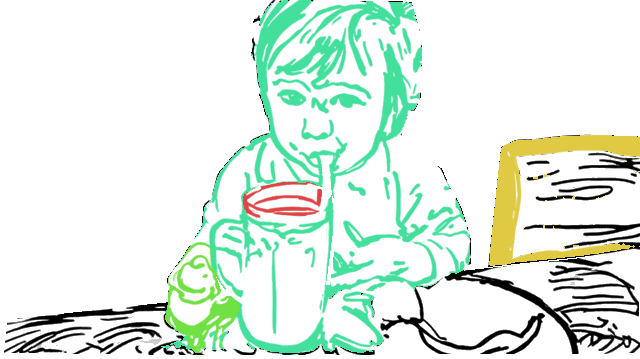}} &
        \frame{\includegraphics[width=0.20\linewidth]{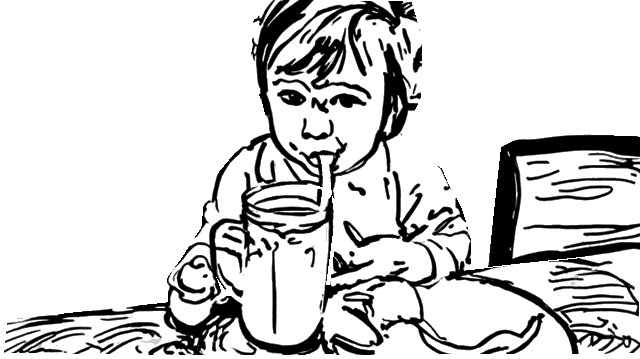}} &
        \frame{\includegraphics[width=0.20\linewidth]{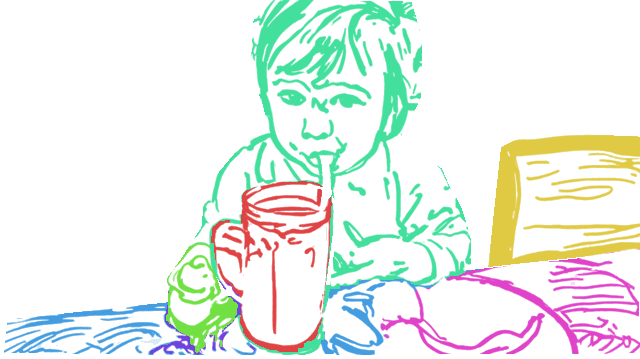}} \\
        
    \end{tabular}
    }
    \vspace{-0.1cm}
    \caption{\textbf{Qualitative comparison of instance segmentation methods on the InstantStyle dataset.} This dataset poses a significantly greater challenge compared to the others due to its increased complexity, including diverse perspectives, intricate textures, and frequent occlusions. Despite these difficulties, our method effectively locates object instances, such as the glass of water in row 2, the fork in row 3, and the food in dishes and bottles in row 4. Even with ambiguous shapes, as shown in row 5, our method outperforms GroundedSAM by successfully segmenting the umbrella on the right separately from the person holding it. In the final row, our approach demonstrates its capability by accurately segmenting both the pile of clothes on the couch and the couch itself.}
    \label{fig:comparison_instance_instantstyle}
    }
\end{figure*}

\begin{figure*}
    \centering
    \setlength{\tabcolsep}{2pt}
    {\small
    \resizebox{0.82\textwidth}{!}{ 
    \begin{tabular}{c @{\hskip 10pt} c c c}
        Input & \rev{Bourouis \etal} & \rev{SketchSeger} & \textbf{Ours}  \\
        \frame{\includegraphics[width=0.23\linewidth]{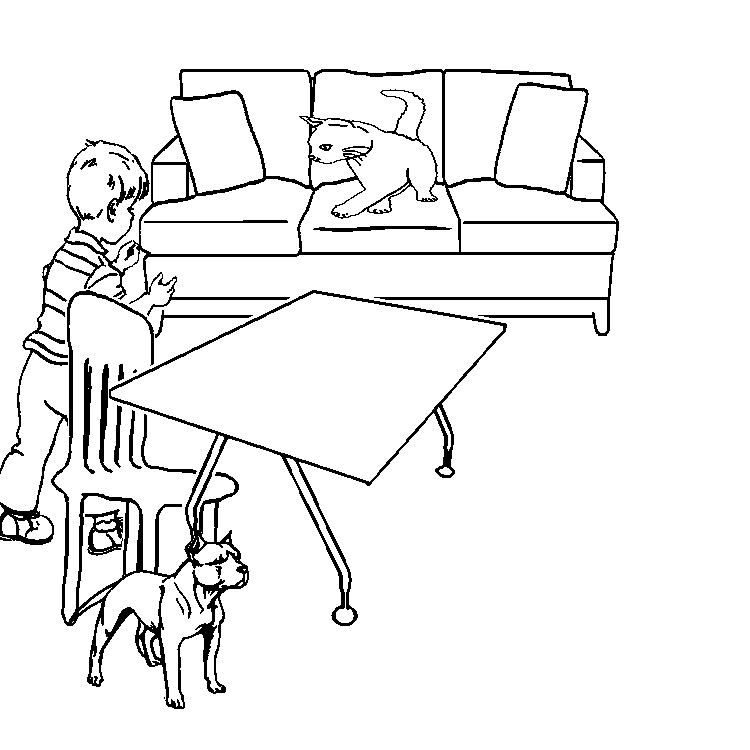}} &
        \frame{\includegraphics[width=0.23\linewidth]{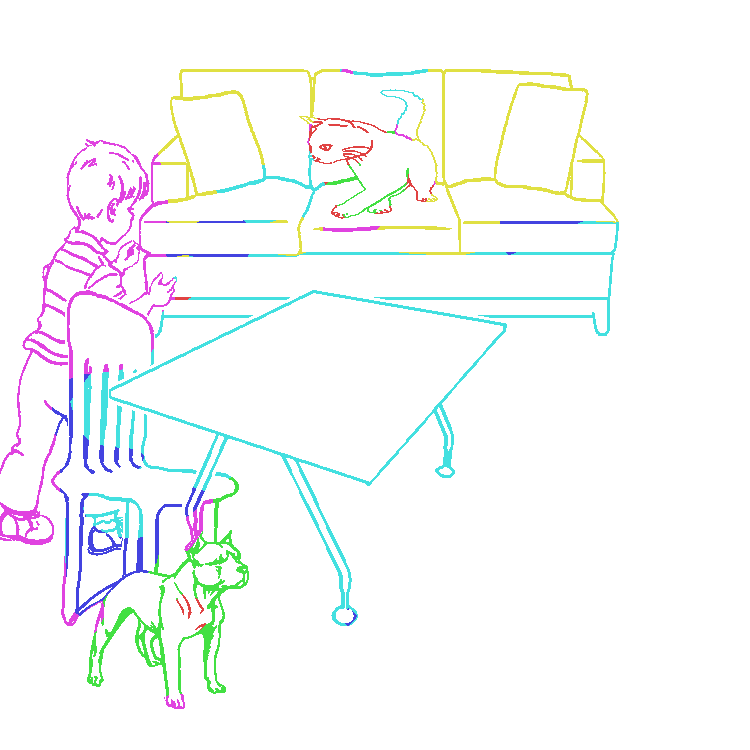}}  &
        \frame{\includegraphics[width=0.23\linewidth]{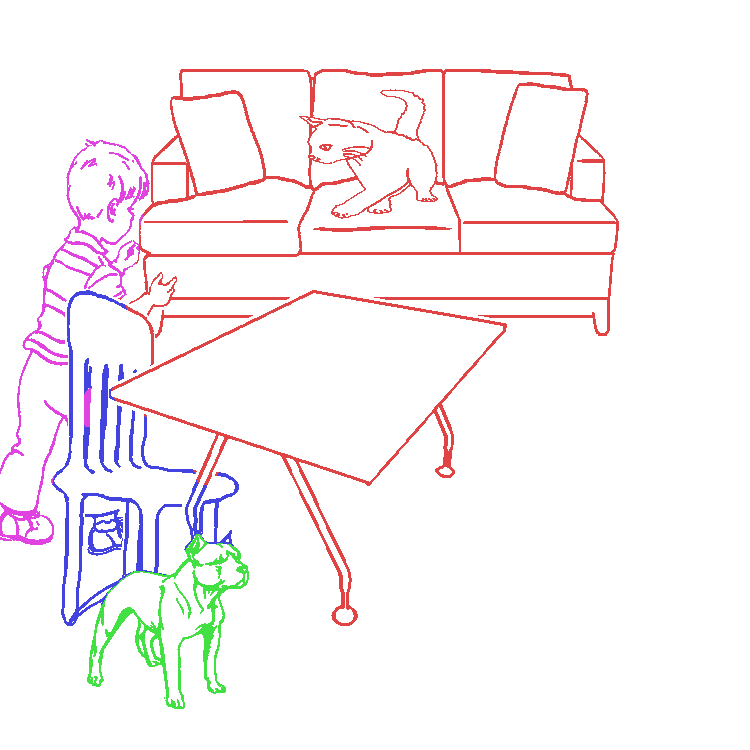}} &
        \frame{\includegraphics[width=0.23\linewidth]{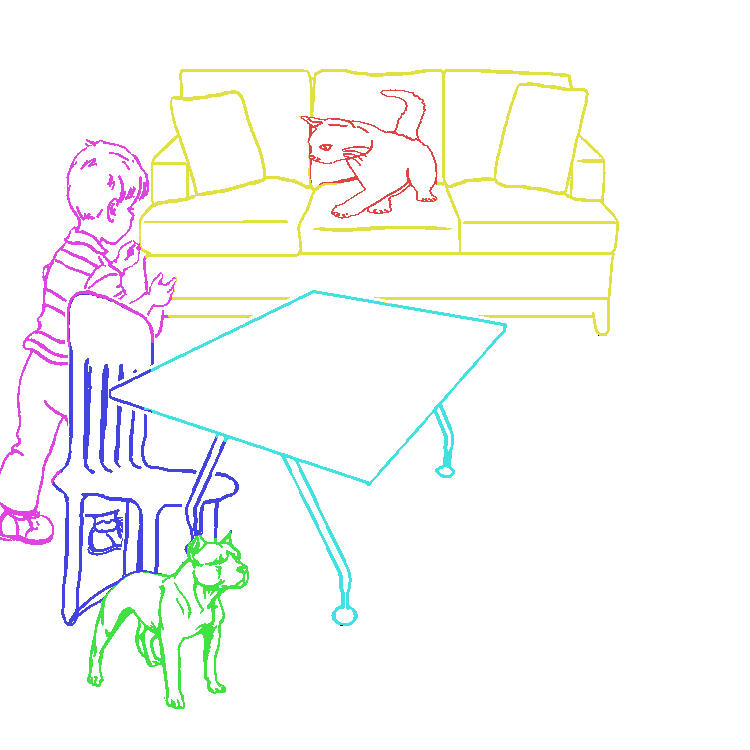}} \\

        & \makecell{cat, chair, dog, people, sofa, table} & & \\

        \frame{\includegraphics[width=0.23\linewidth]{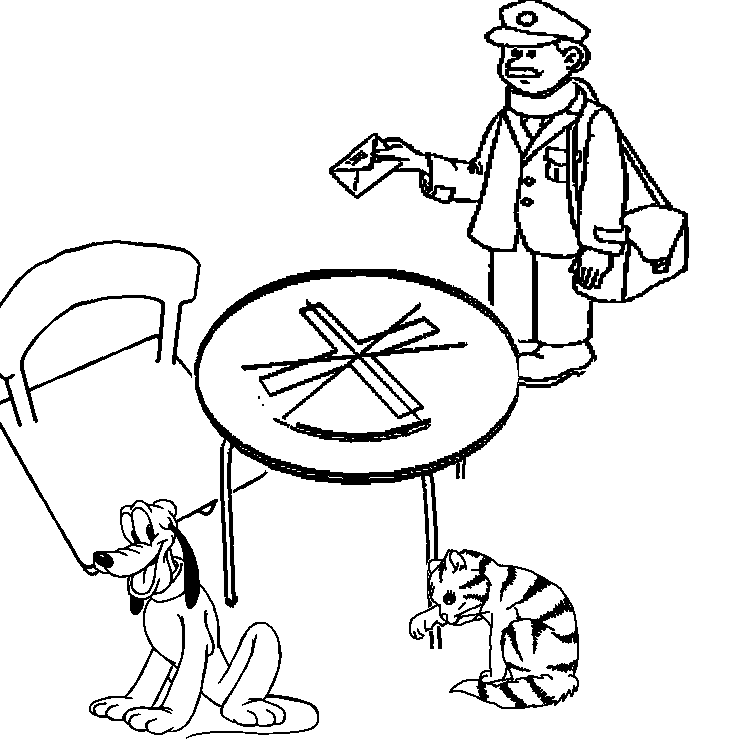}} &
        \frame{\includegraphics[width=0.23\linewidth]{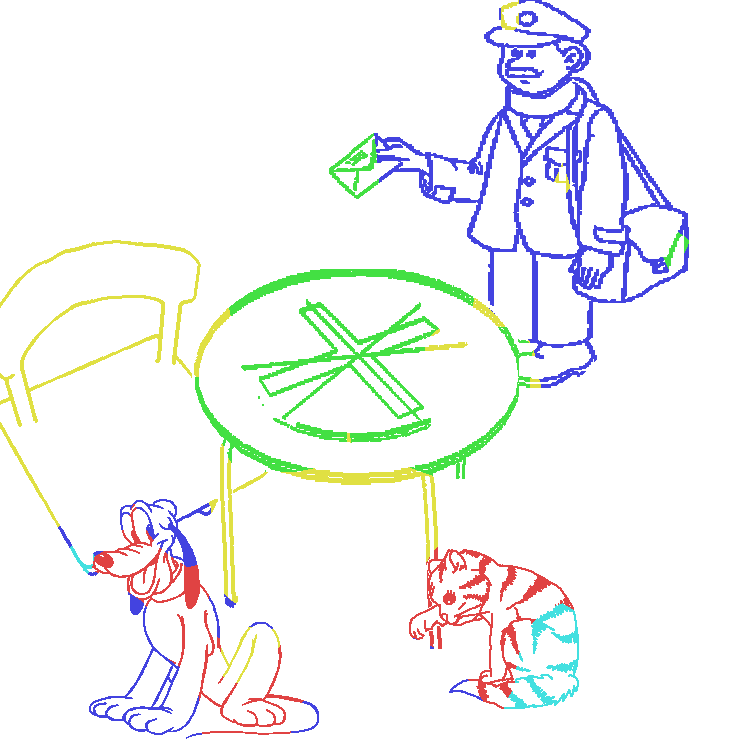}} &
        \frame{\includegraphics[width=0.23\linewidth]{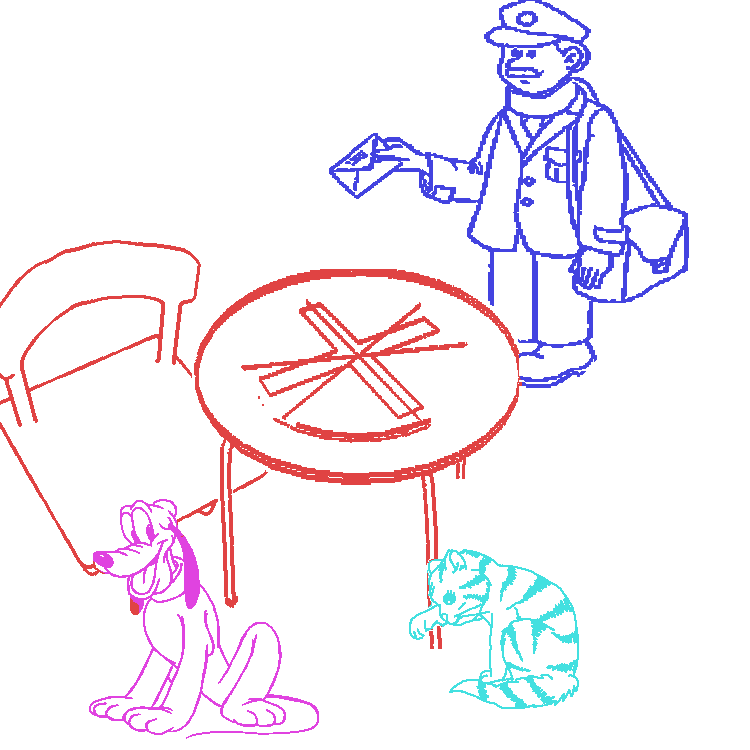}} &
        \frame{\includegraphics[width=0.23\linewidth]{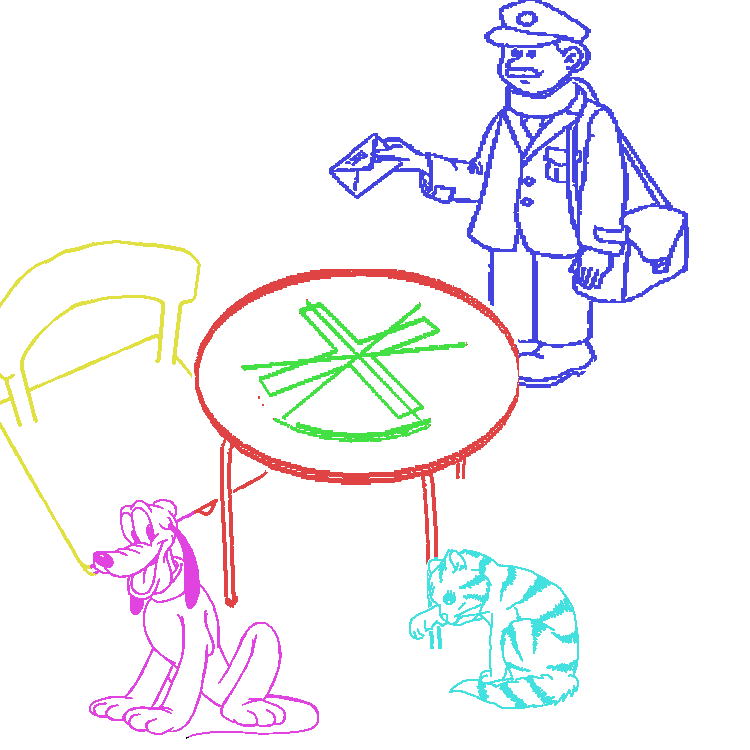}} \\

        & \makecell{cat, chair, dog, people, table} & & \\

        \frame{\includegraphics[width=0.23\linewidth]{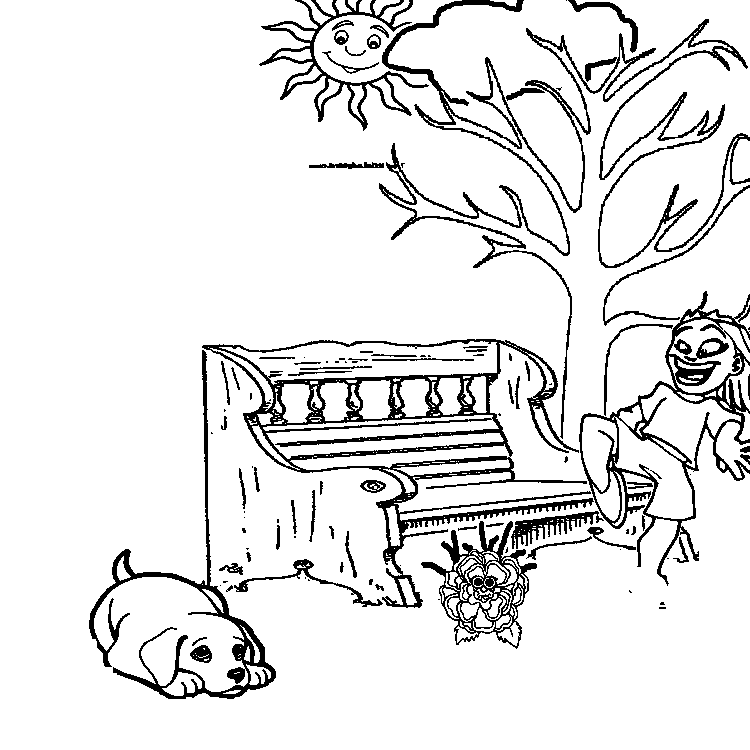}} &
        \frame{\includegraphics[width=0.23\linewidth]{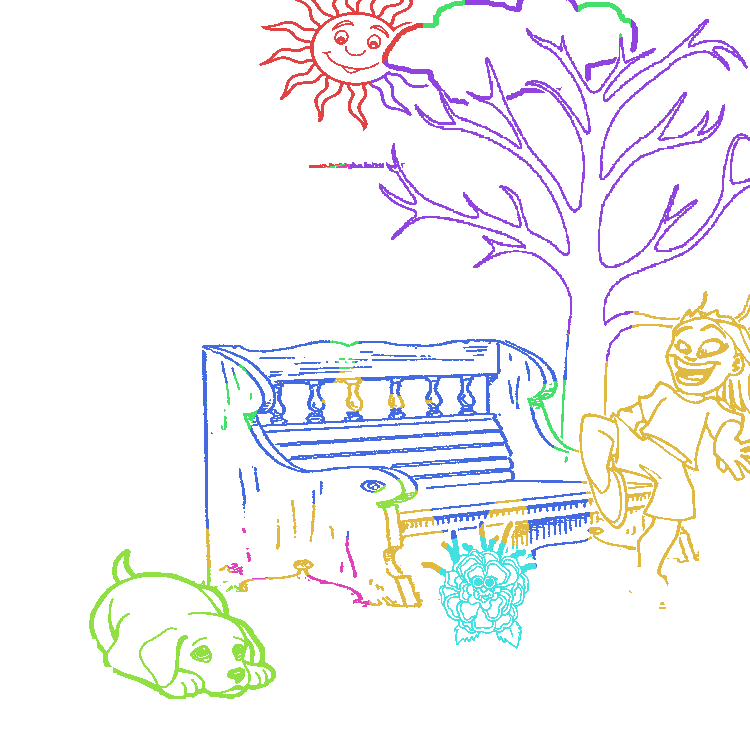}} &
        \frame{\includegraphics[width=0.23\linewidth]{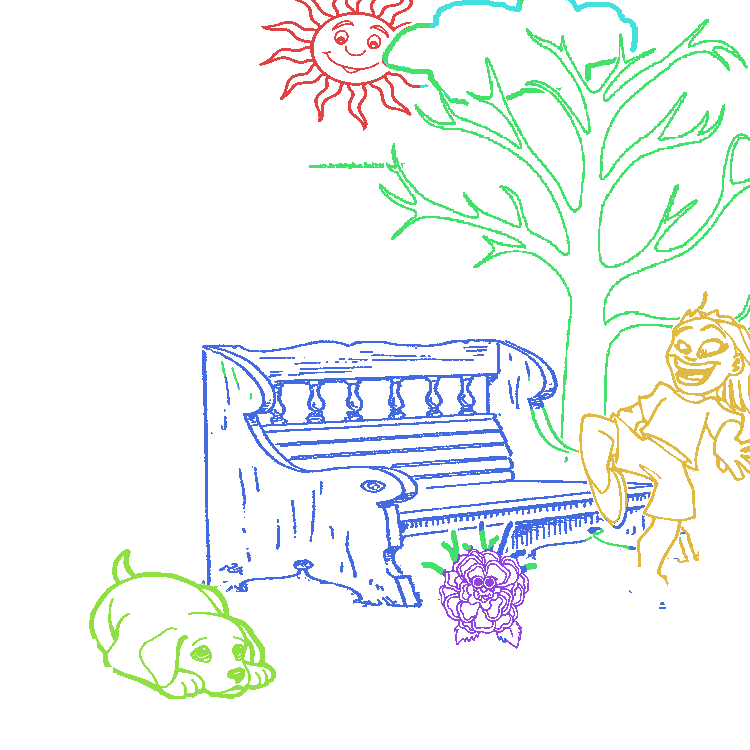}} &
        \frame{\includegraphics[width=0.23\linewidth]{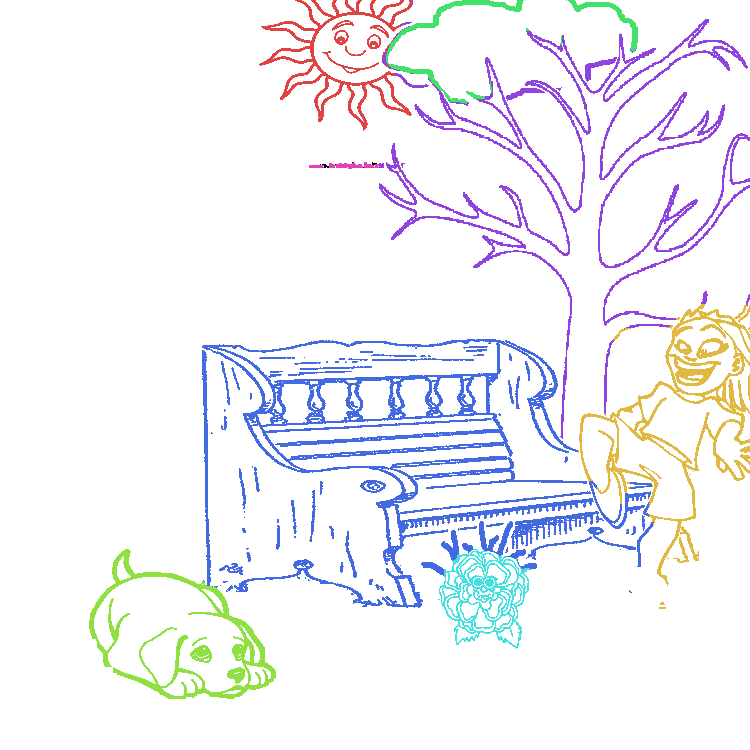}} \\

        & \makecell{bench, cloud, dog, flower, grass,\\ people, sun, tree} & & \\

        \frame{\includegraphics[width=0.23\linewidth]{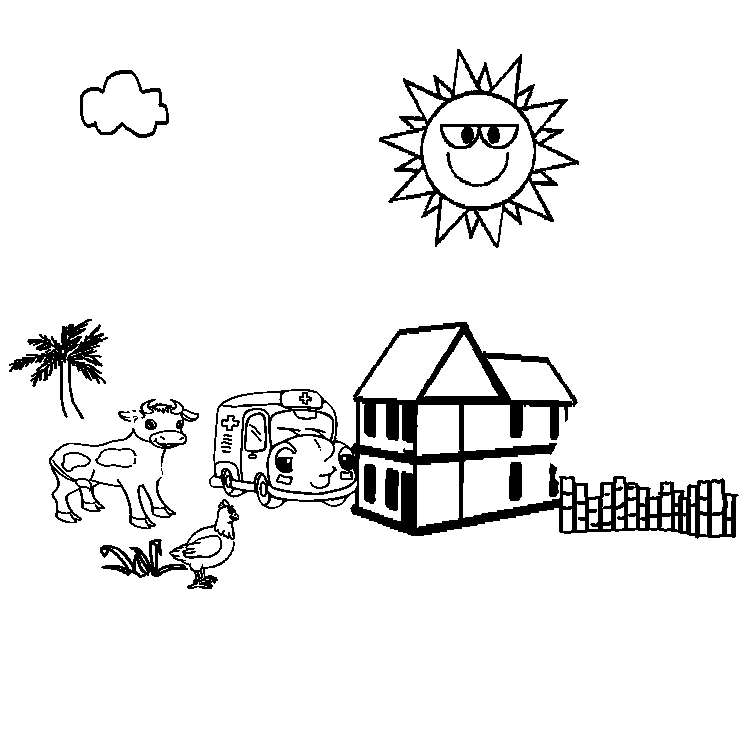}} &
        \frame{\includegraphics[width=0.23\linewidth]{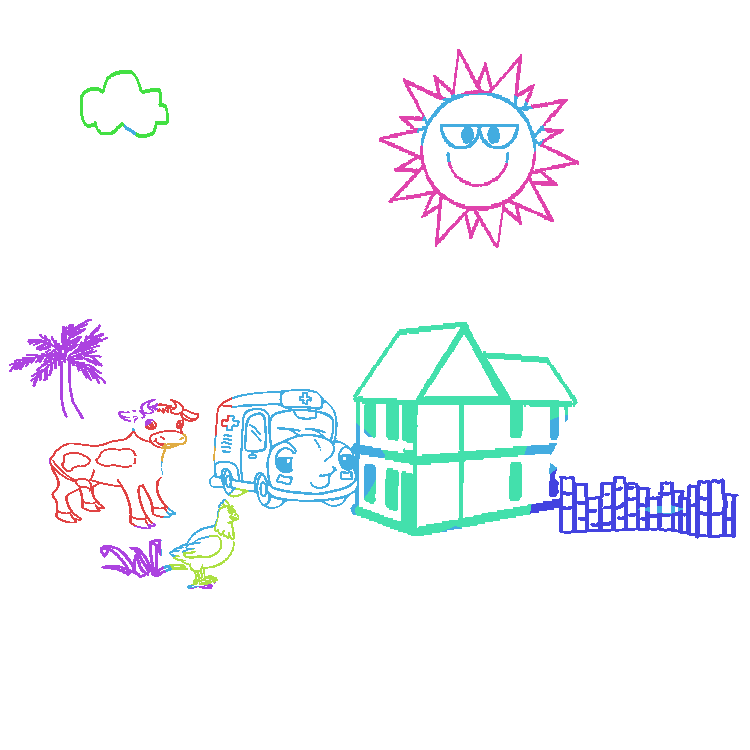}} &
        \frame{\includegraphics[width=0.23\linewidth]{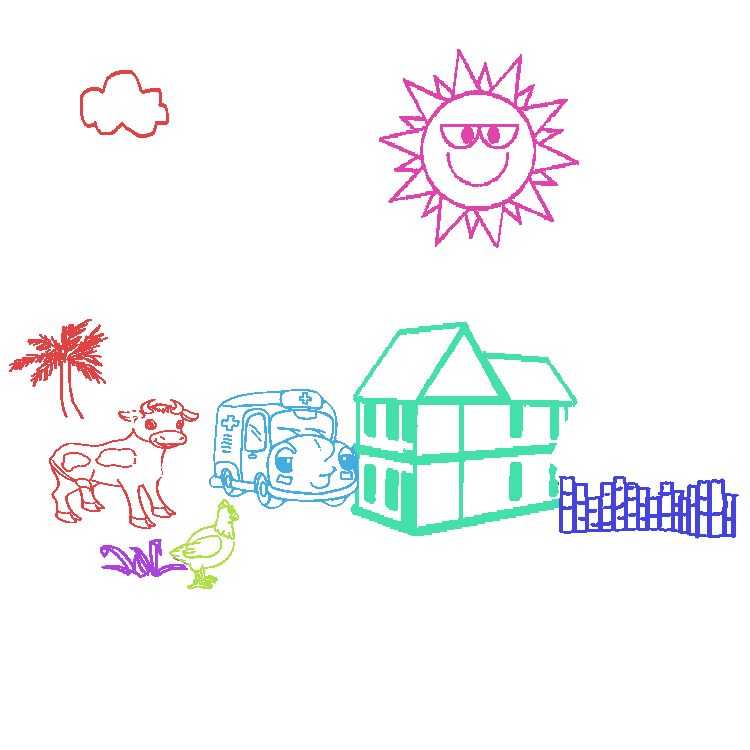}} &
        \frame{\includegraphics[width=0.23\linewidth]{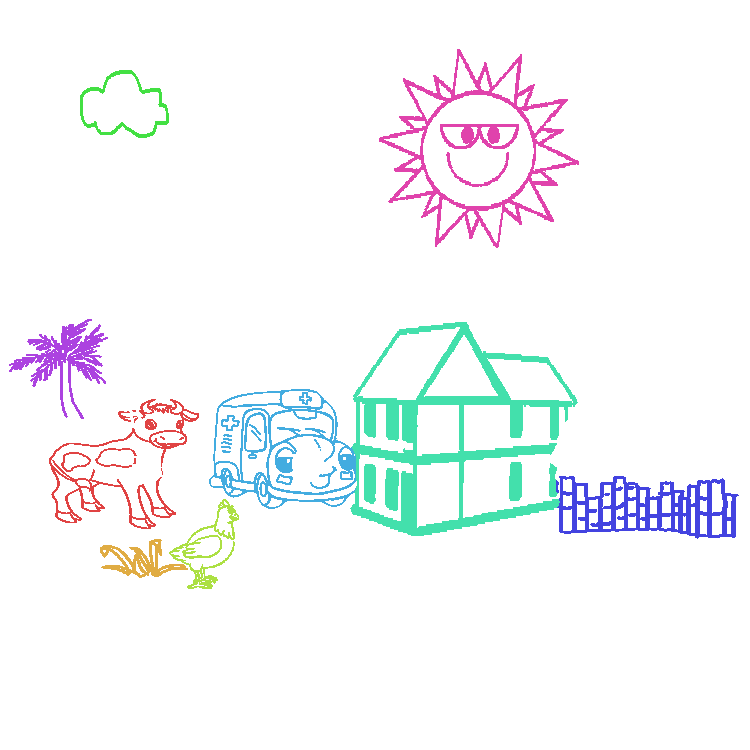}} \\

        & \makecell{car, chicken, cloud, cow, fence,\\ grass, house, sun, tree} & & \\

        \frame{\includegraphics[width=0.23\linewidth]{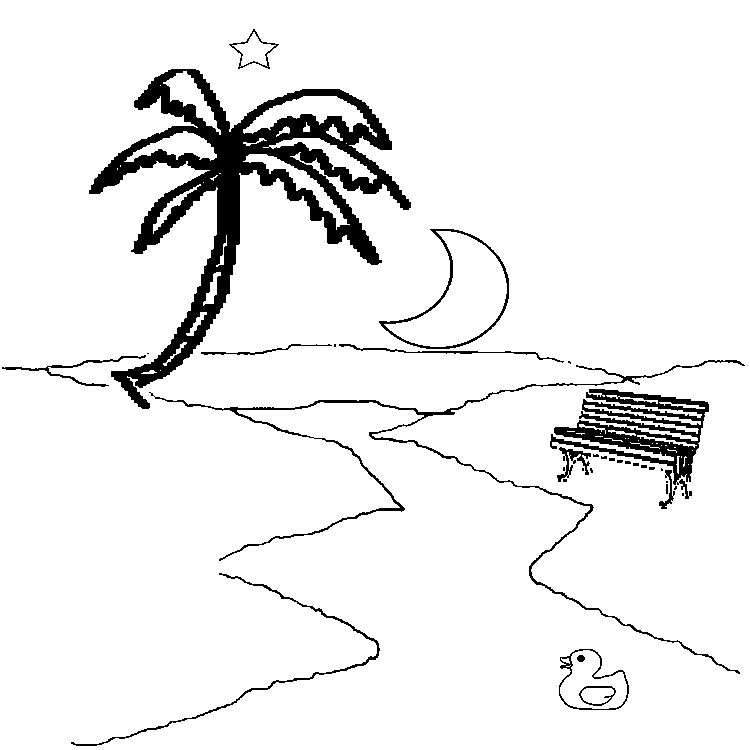}} &
        \frame{\includegraphics[width=0.23\linewidth]{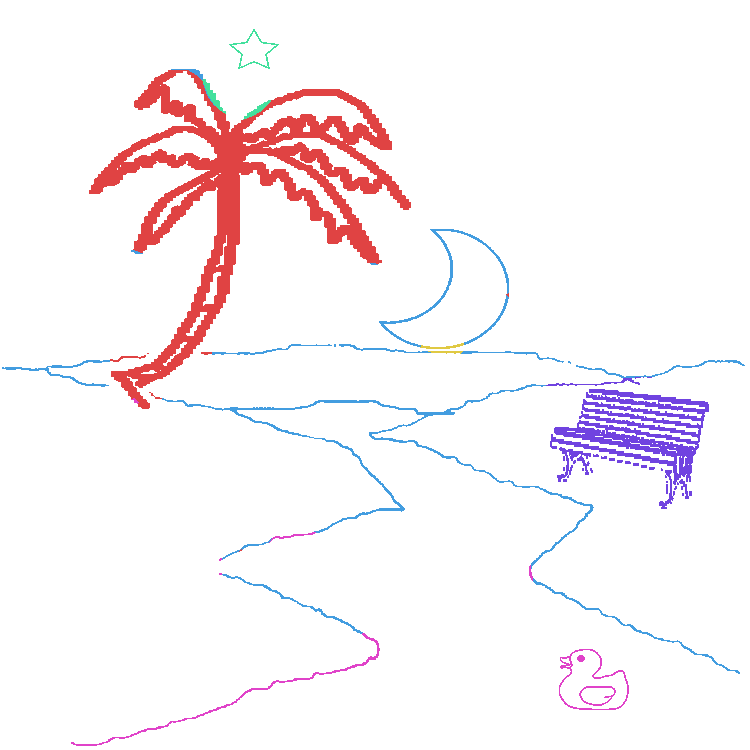}} &
        \frame{\includegraphics[width=0.23\linewidth]{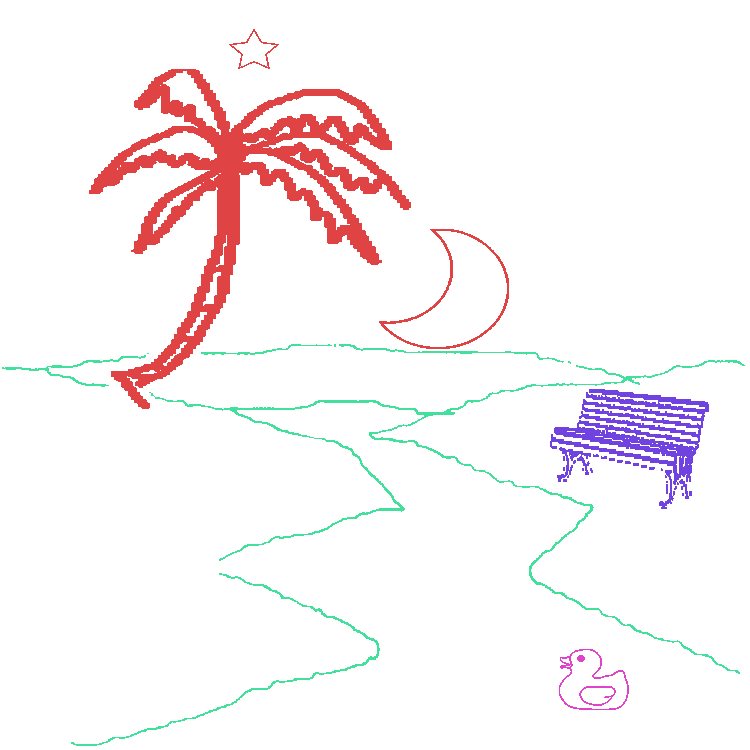}} &
        \frame{\includegraphics[width=0.23\linewidth]{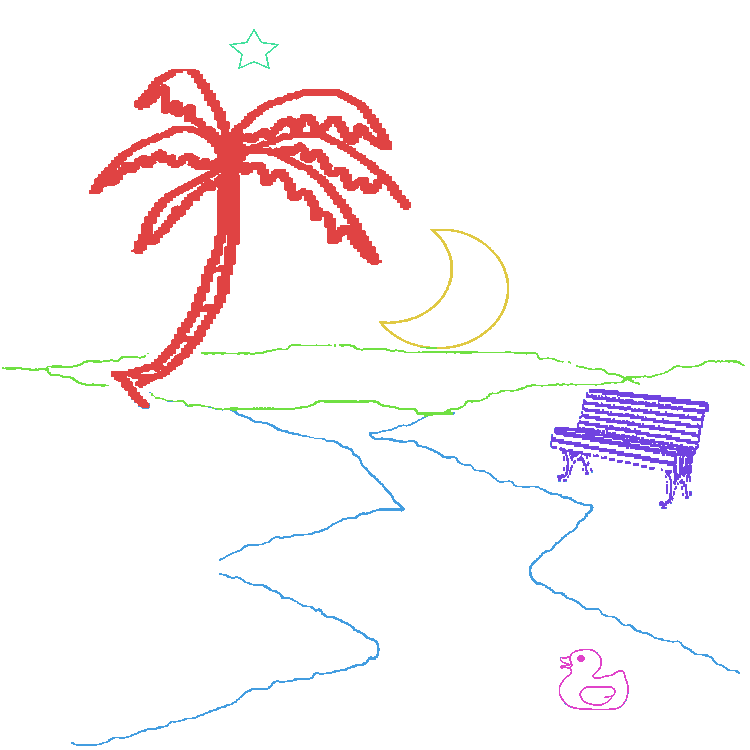}} \\

        & \makecell{bench, duck, moon, star, tree, others}  & & 
    \end{tabular}
    }
    \vspace{-0.3cm}
    \caption{\textbf{Qualitative comparison of segmentation on filtered SketchyScene dataset. } We prompt Bourouis \etal’s method with ground truth labels and use confidence threshold of 0.01 to ensure all sketch pixels are segmented, while using SketchSeger as-is since it does not accept input labels. However, Bourouis \etal struggles to accurately identify instances of the correct class, often assigning multiple class labels to the same object, as indicated by the color gradients. SketchSeger also fails to provide clean segmentations for many scenes. In contrast, our method effectively segments object instances, ensuring clear separation and consistent labeling.}
    \label{fig:comparison_openvocab_sketchyscene}
    }
\end{figure*}

\begin{figure*}
\vspace{-3mm}
    \centering
    \setlength{\tabcolsep}{2pt}
    {\small
    \resizebox{0.82\textwidth}{!}{ 
    \begin{tabular}{c @{\hskip 10pt} c c c}
        Input & \rev{Bourouis \etal} & \rev{SketchSeger} & \textbf{Ours}  \\
        \frame{\includegraphics[width=0.23\linewidth]{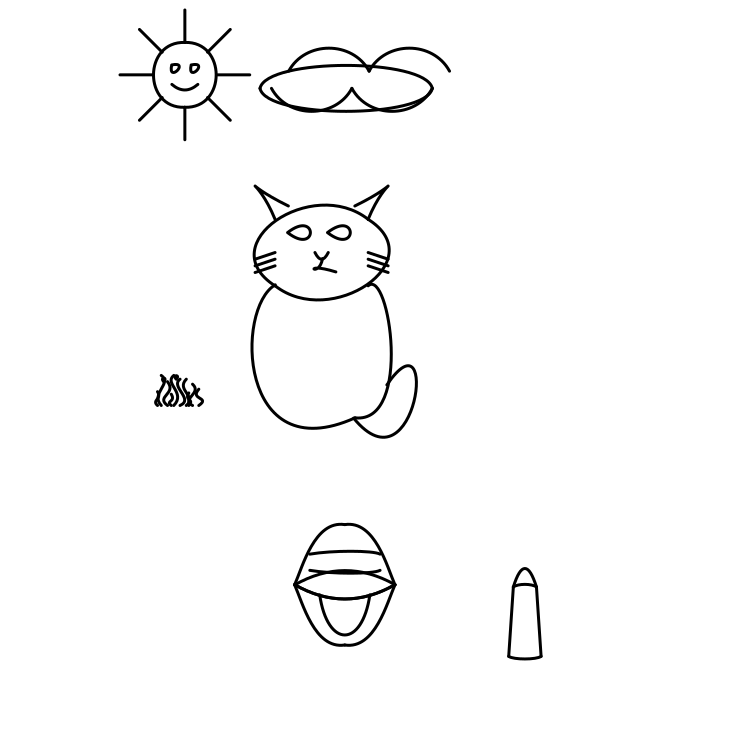}} &
        \frame{\includegraphics[width=0.23\linewidth]{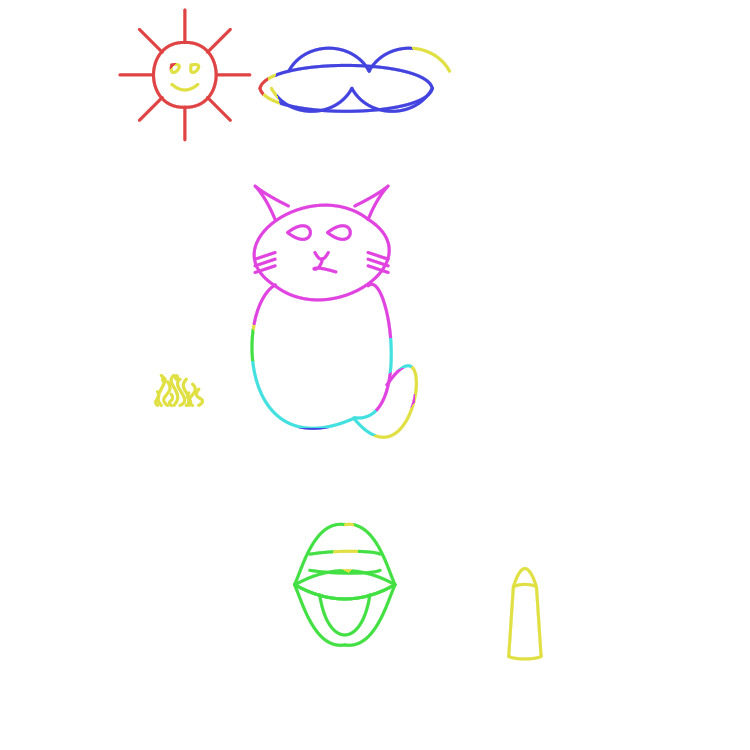}} &
        \frame{\includegraphics[width=0.23\linewidth]{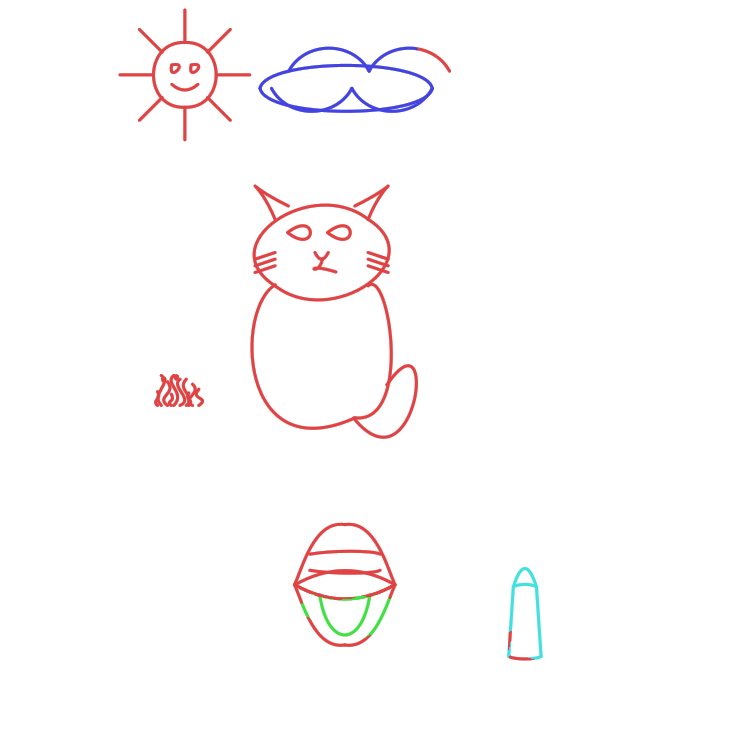}} &
        \frame{\includegraphics[width=0.23\linewidth]{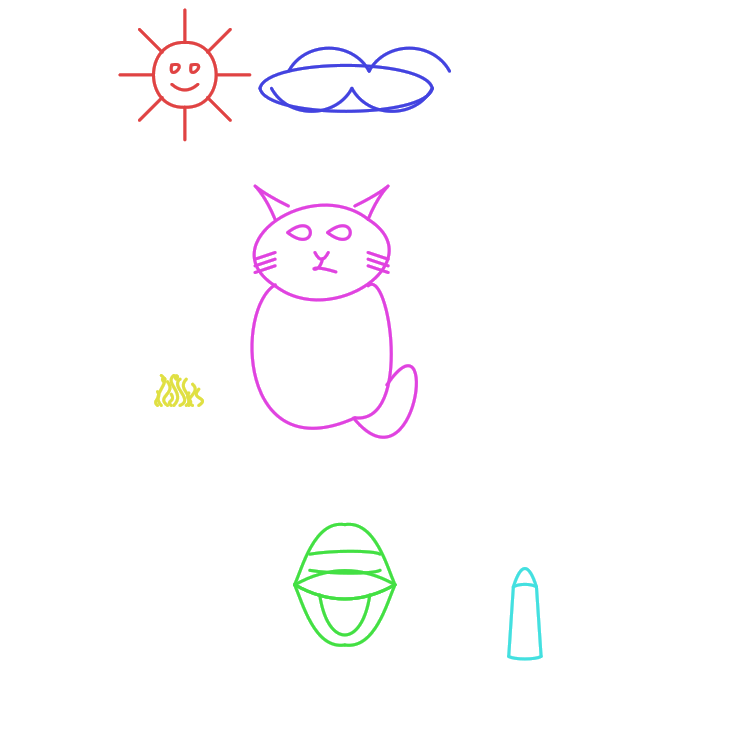}} \\

        & \makecell{basket, bucket, cat, cloud, grass, sun}  & & \\

        \frame{\includegraphics[width=0.23\linewidth]{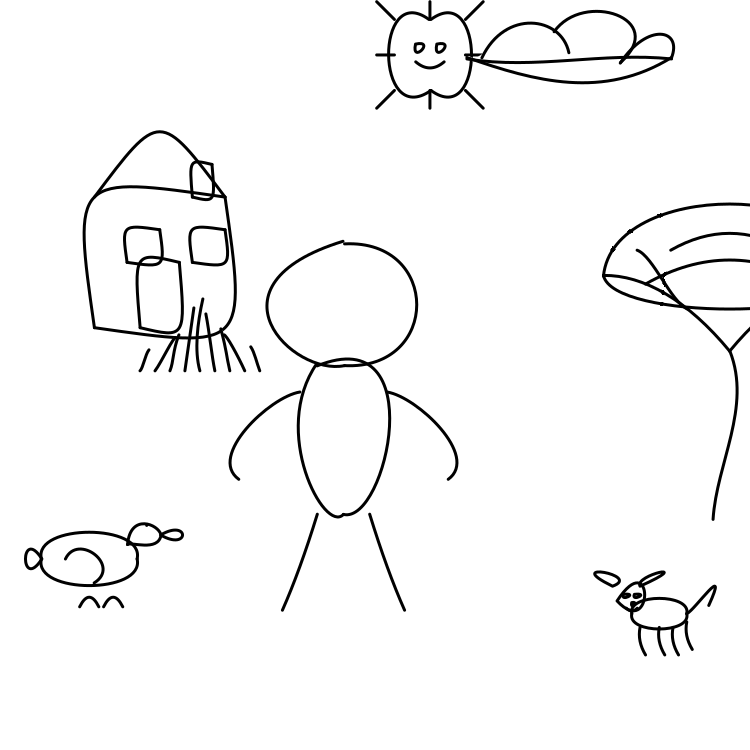}} &
        \frame{\includegraphics[width=0.23\linewidth]{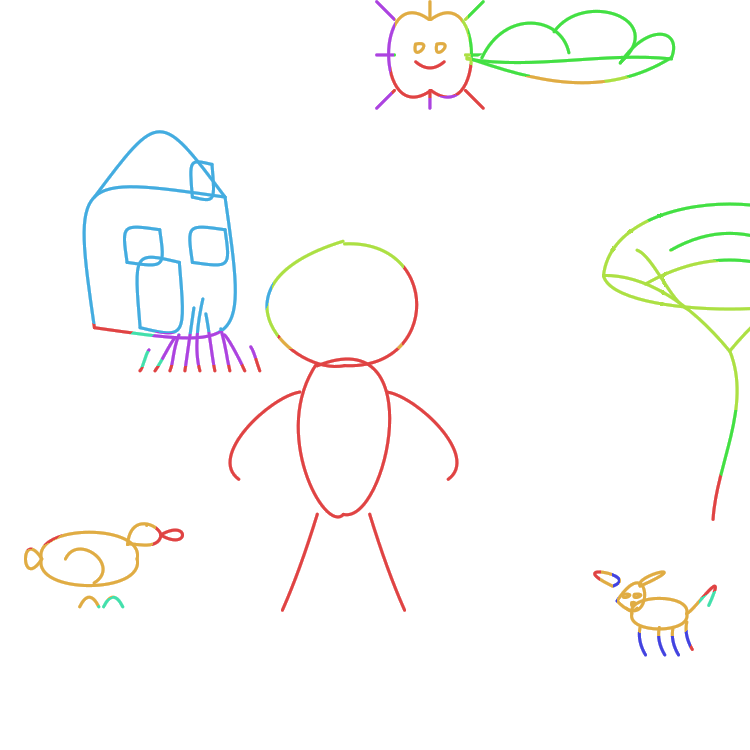}} &
        \frame{\includegraphics[width=0.23\linewidth]{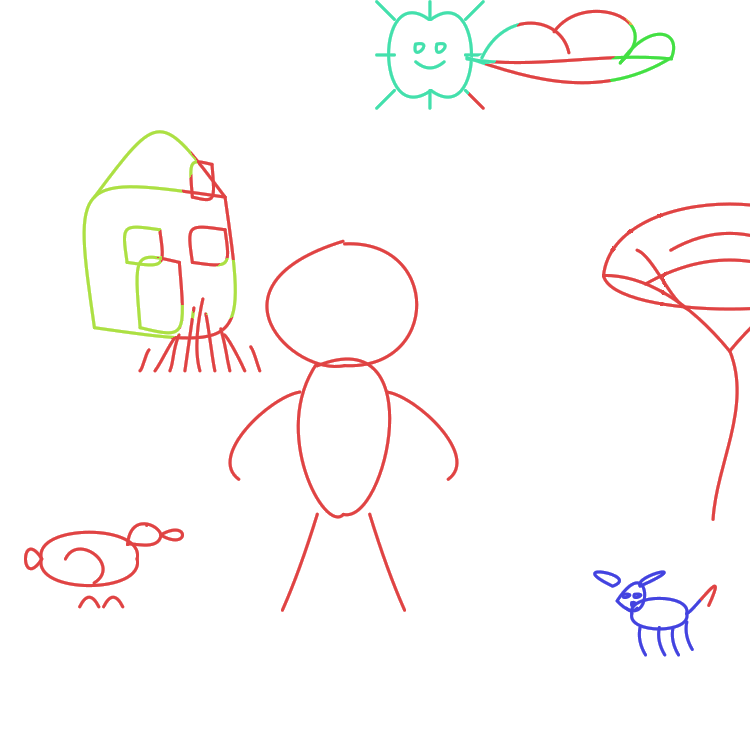}} &
        \frame{\includegraphics[width=0.23\linewidth]{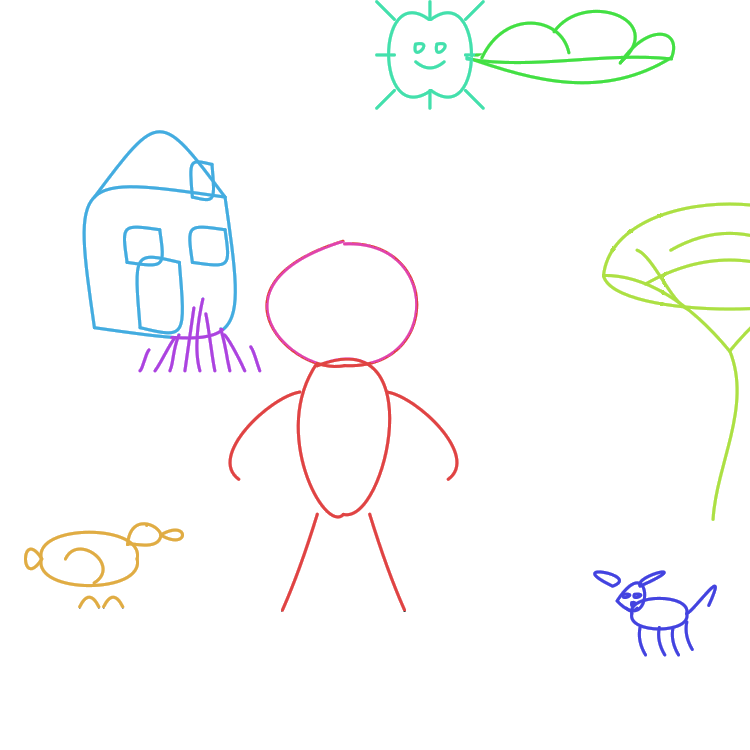}} \\

        & \makecell{cloud, dog, duck, grass, house, \\ people, sun, tree}  & & \\

        \frame{\includegraphics[width=0.23\linewidth]{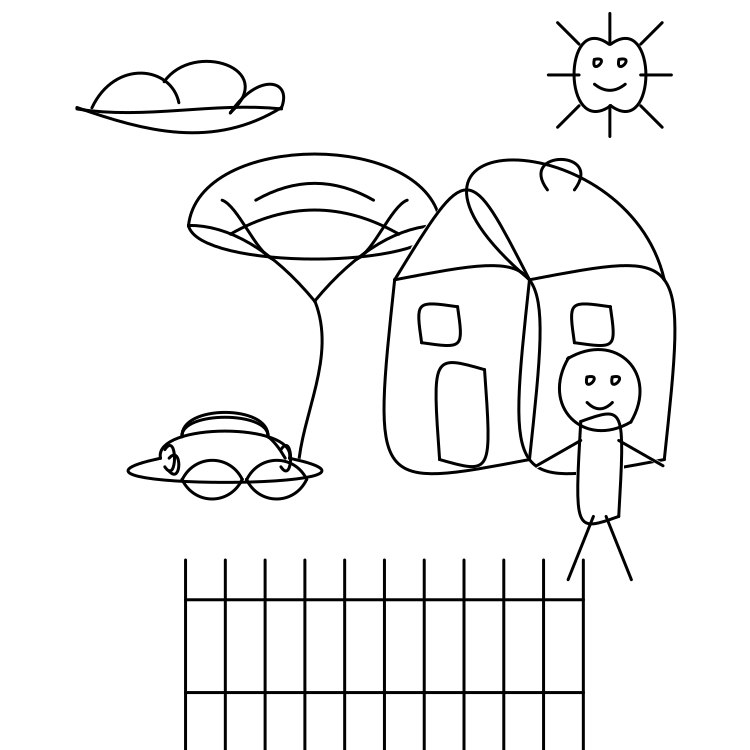}} &
        \frame{\includegraphics[width=0.23\linewidth]{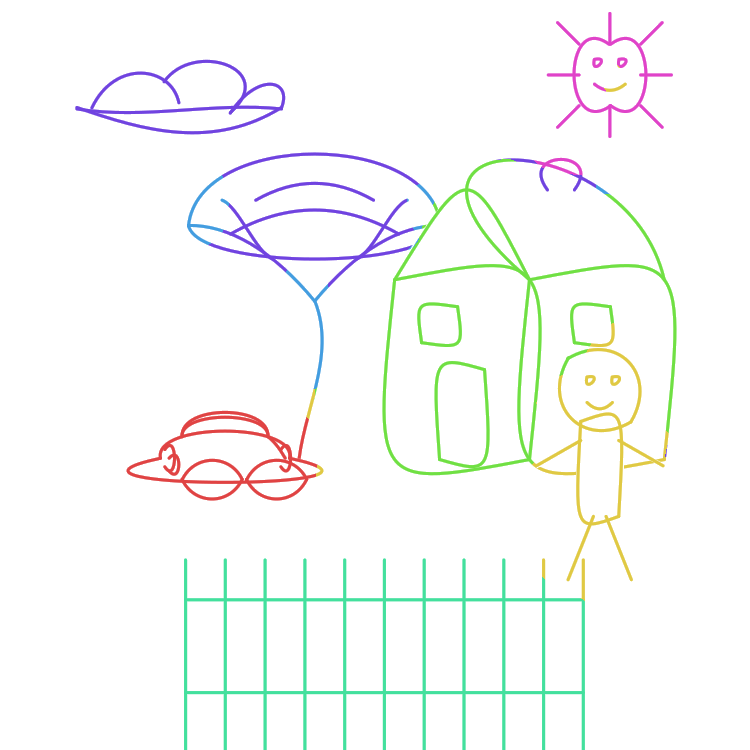}} &
        \frame{\includegraphics[width=0.23\linewidth]{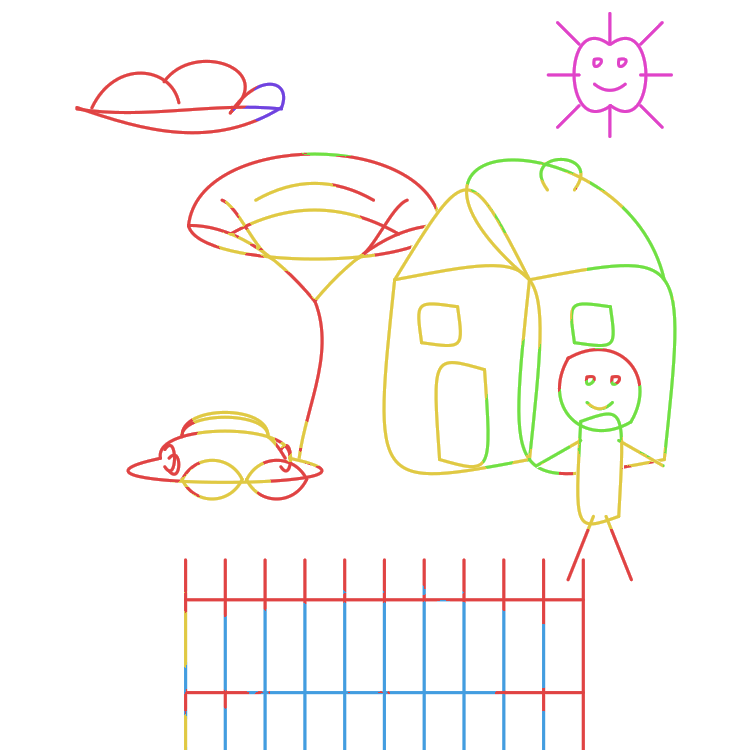}} &
        \frame{\includegraphics[width=0.23\linewidth]{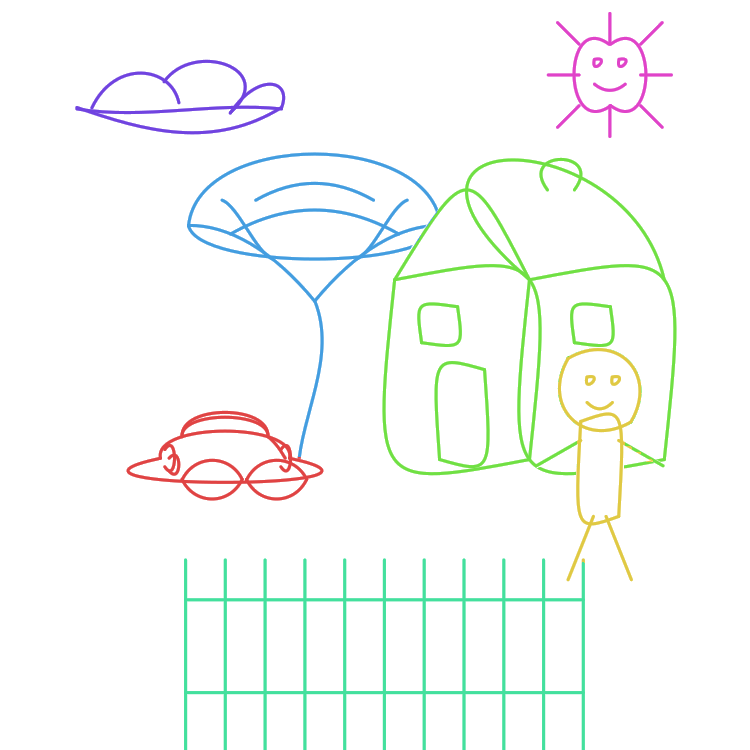}} \\

        & \makecell{car, cloud, fence, house, people,\\ sun, tree }  & & \\

        \frame{\includegraphics[width=0.23\linewidth]{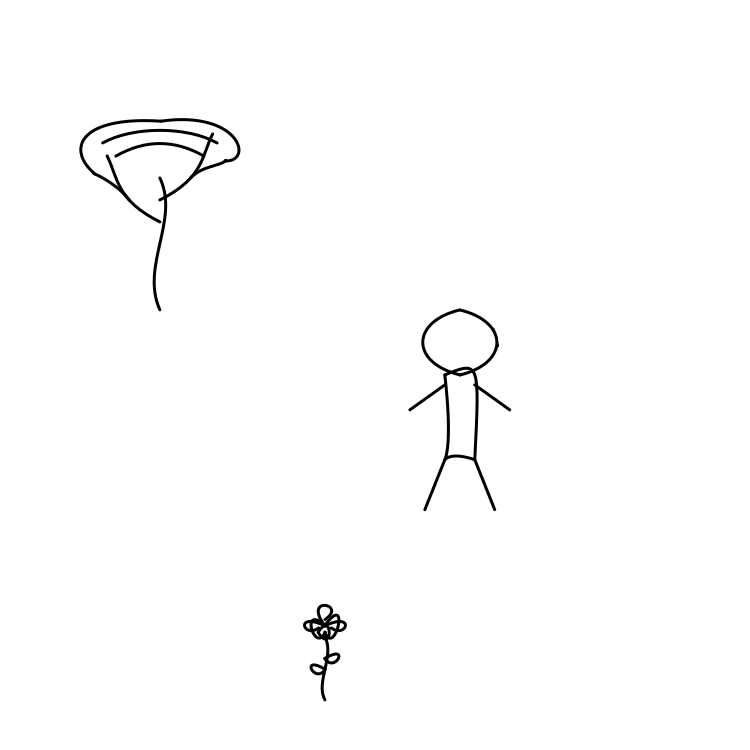}} &
        \frame{\includegraphics[width=0.23\linewidth]{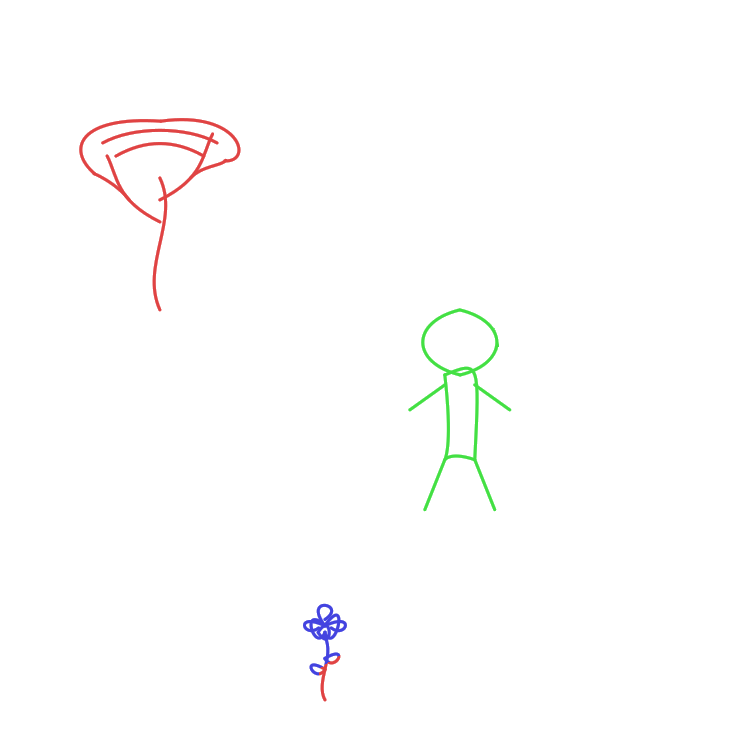}} &
        \frame{\includegraphics[width=0.23\linewidth]{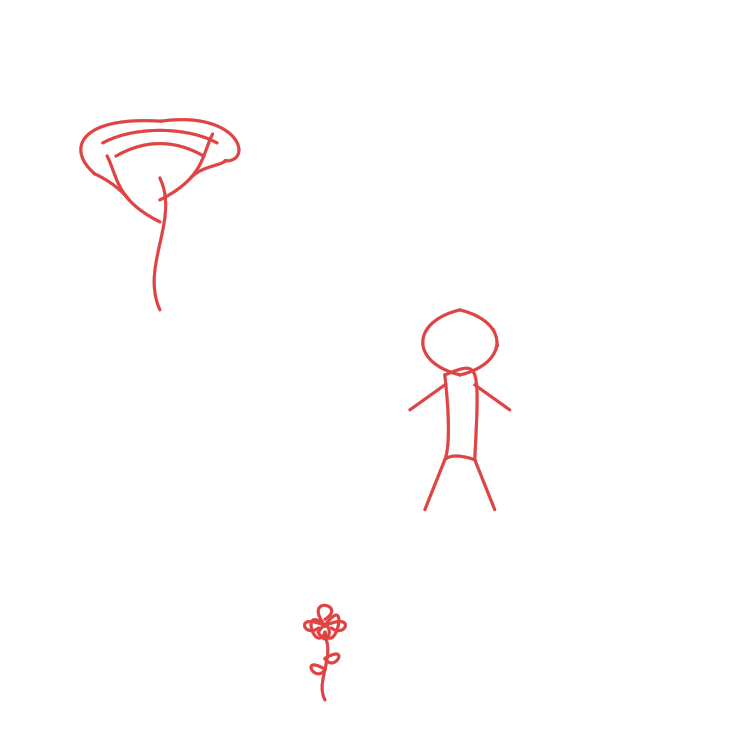}} &
        \frame{\includegraphics[width=0.23\linewidth]{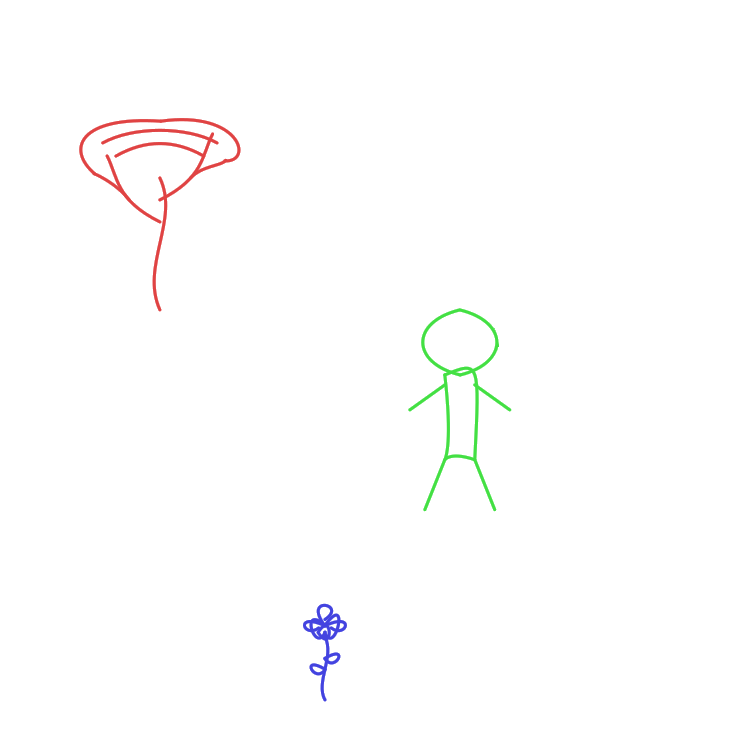}} \\

        & \makecell{flower, people, tree }  & & \\

        \frame{\includegraphics[width=0.23\linewidth]{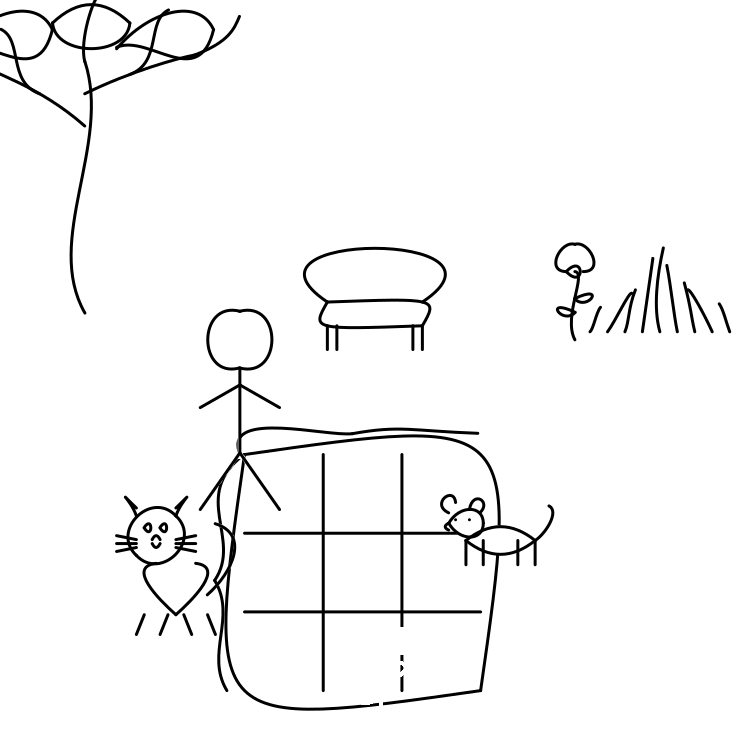}} &
        \frame{\includegraphics[width=0.23\linewidth]{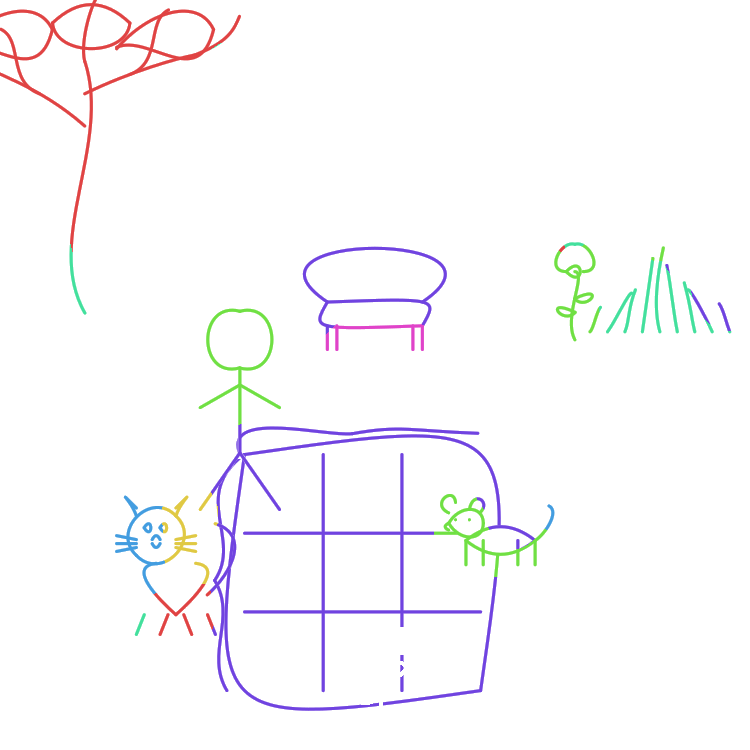}} &
        \frame{\includegraphics[width=0.23\linewidth]{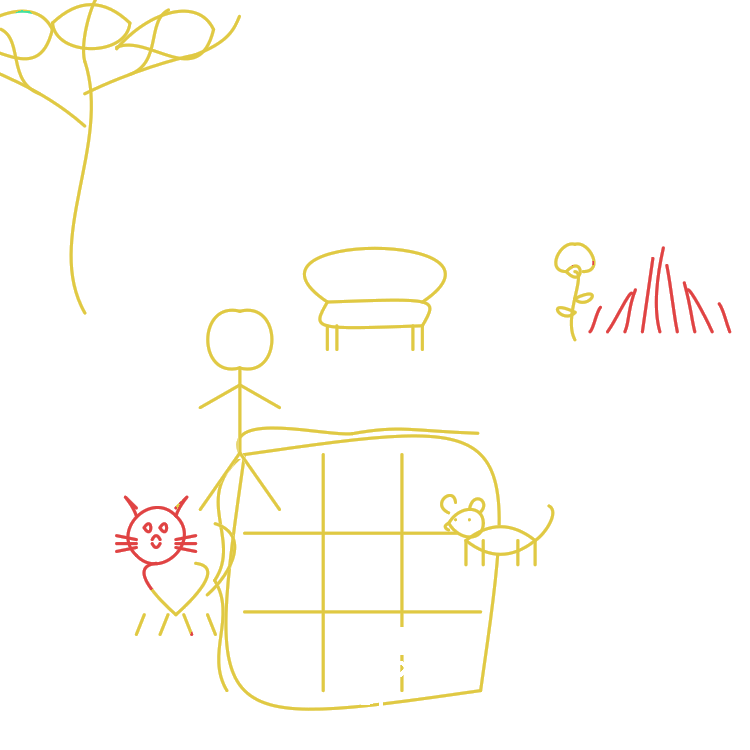}} &
        \frame{\includegraphics[width=0.23\linewidth]{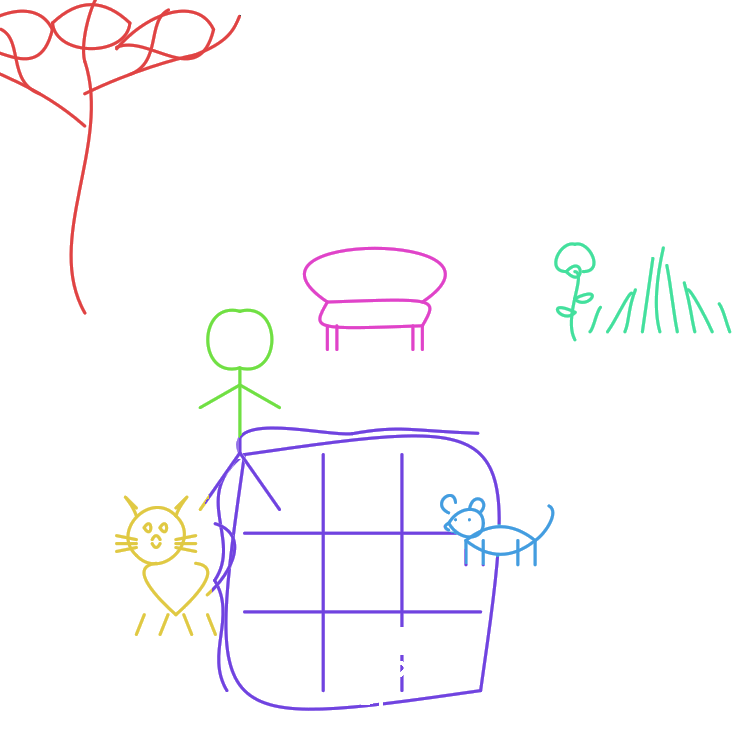}} \\

        & \makecell{bench, cat, dog, flower, grass, people, \\ picnic rug, tree}  & & 
    \end{tabular}
    }
    \vspace{-0.3cm}
    \caption{\textbf{Qualitative comparison of segmentation on filtered SketchAgent dataset. } We prompt Bourouis \etal’s method with ground truth labels and use confidence threshold of 0.01 to ensure all sketch pixels are segmented, while using SketchSeger as-is since it does not accept input labels. However, Bourouis \etal struggles to accurately identify instances of the correct class, often assigning multiple class labels to the same object, as indicated by the color gradients. SketchSeger also fails to provide clean segmentations for many scenes. In contrast, our method effectively segments object instances, ensuring clear separation and consistent labeling.}
    \label{fig:comparison_openvocab_sketchagent}
    }
\end{figure*}

\begin{figure*}
    \centering
    \setlength{\tabcolsep}{2pt}
    {\small
    \resizebox{0.79\textwidth}{!}{ 
    \begin{tabular}{c @{\hskip 10pt} c c c}
        Input & \rev{Bourouis \etal} & \rev{SketchSeger} & \textbf{Ours}  \\
        \frame{\includegraphics[width=0.23\linewidth]{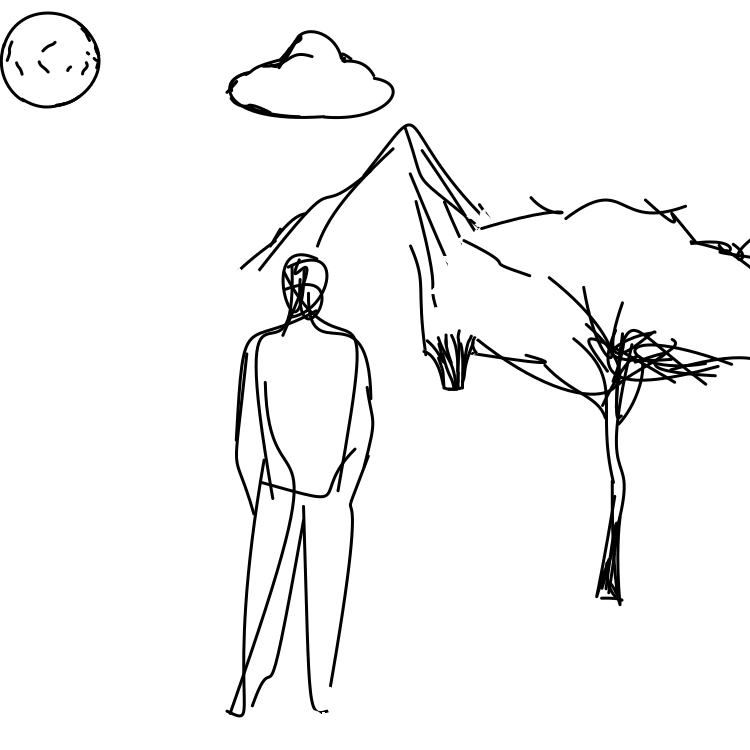}} &
        \frame{\includegraphics[width=0.23\linewidth]{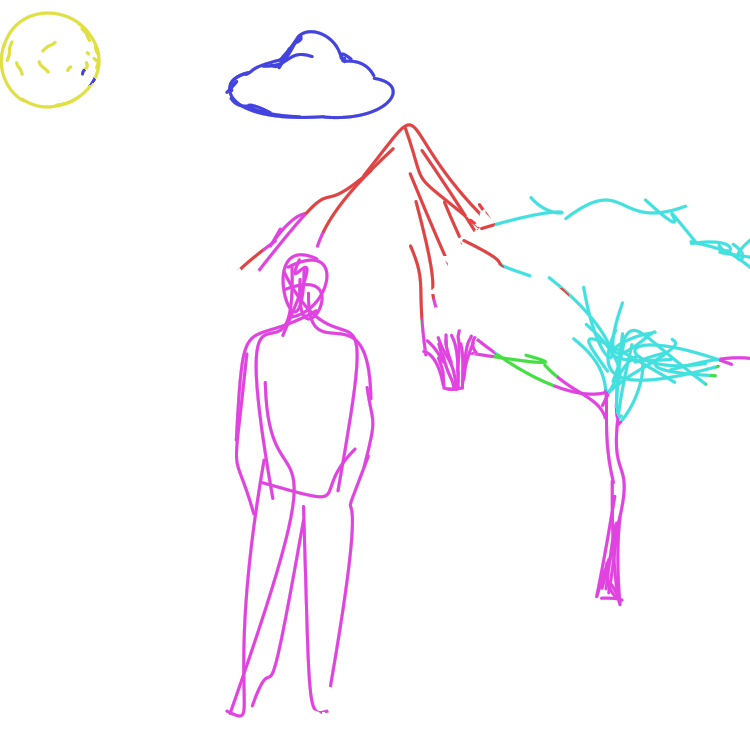}} &
        \frame{\includegraphics[width=0.23\linewidth]{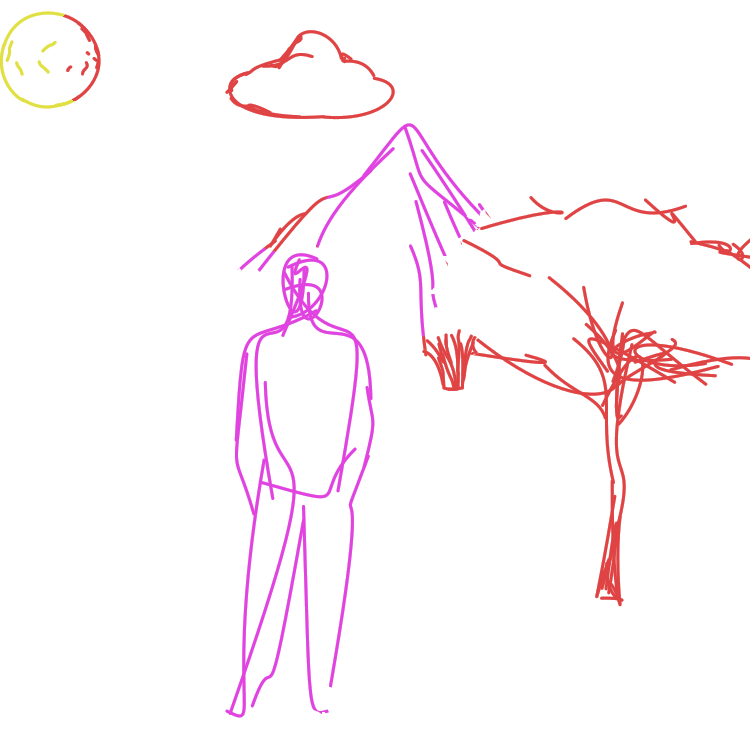}} &
        \frame{\includegraphics[width=0.23\linewidth]{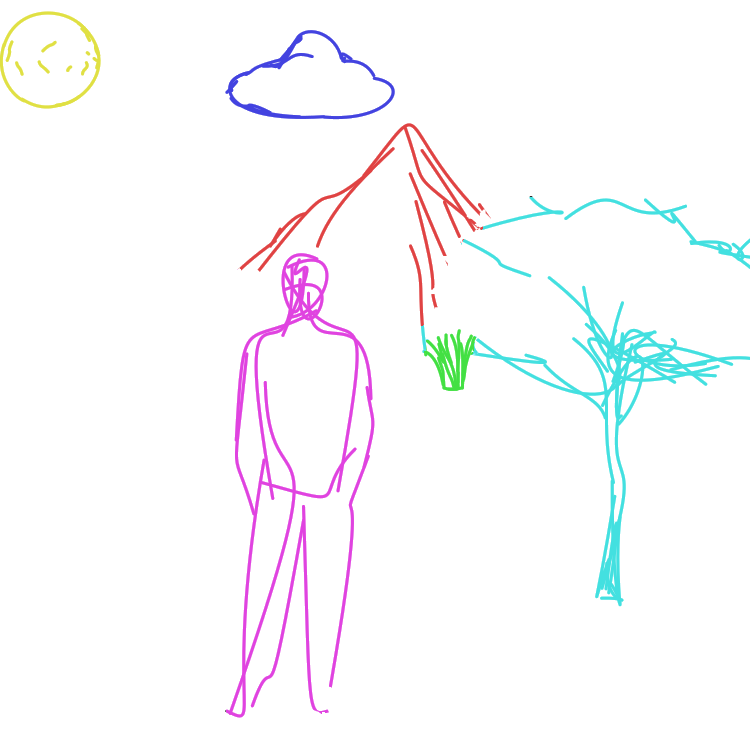}} \\
         & \makecell{cloud, grass, mountain, people, \\ sun, tree}  & & \\
        
        \frame{\includegraphics[width=0.23\linewidth]{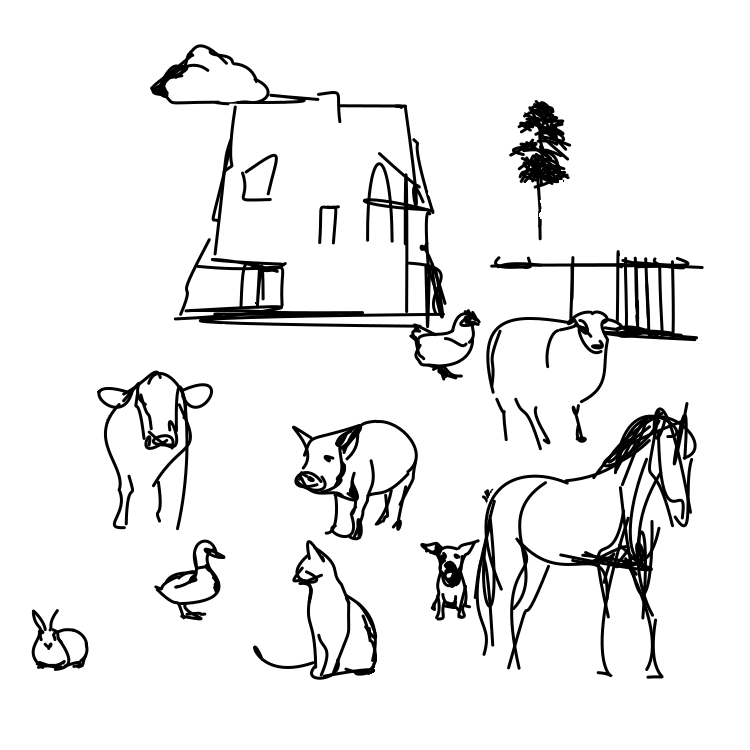}} &
        \frame{\includegraphics[width=0.23\linewidth]{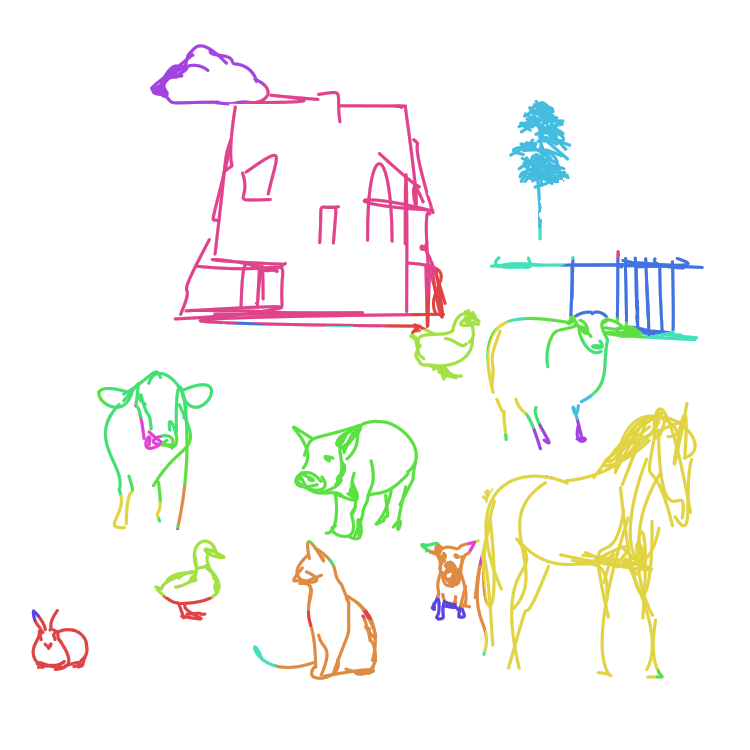}} &
        \frame{\includegraphics[width=0.23\linewidth]{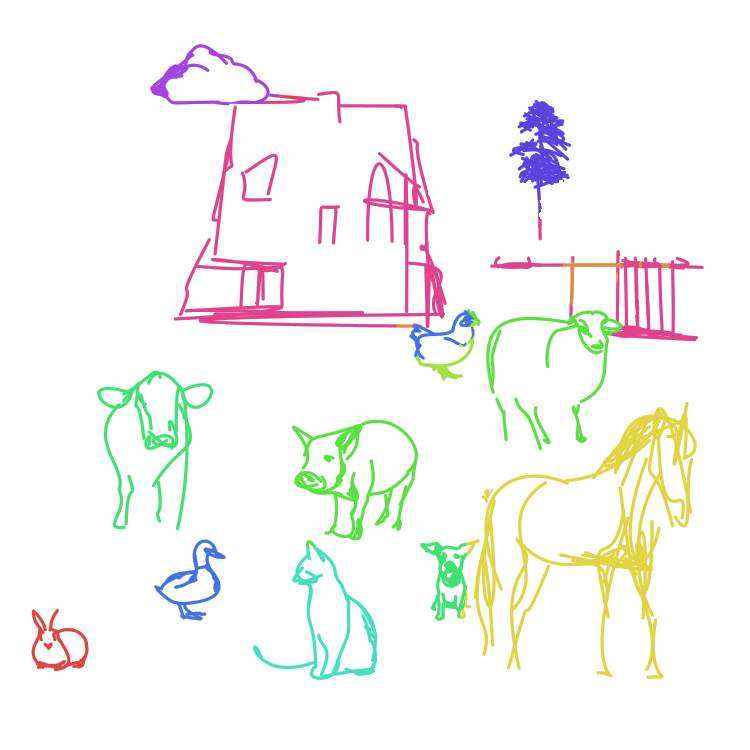}} &
        \frame{\includegraphics[width=0.23\linewidth]{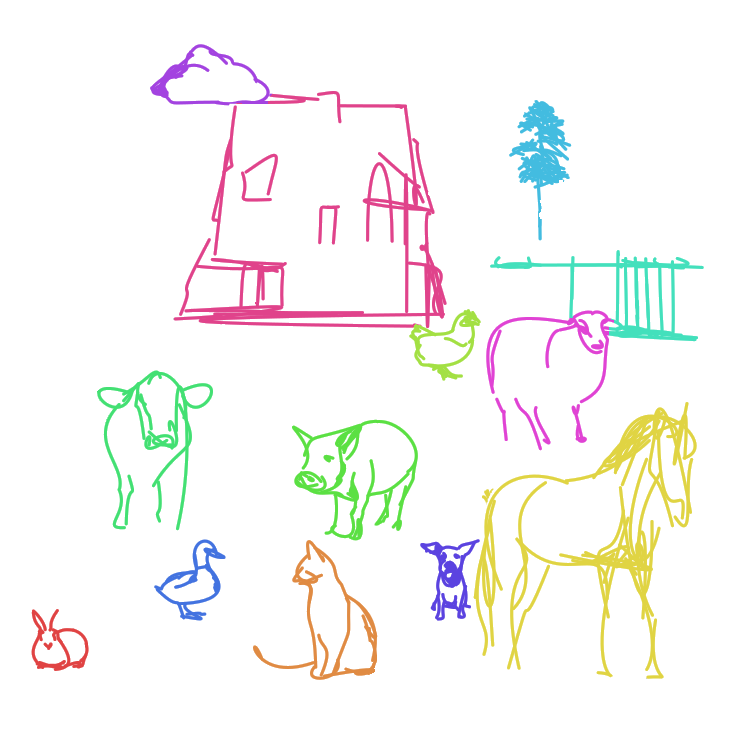}} \\
         & \makecell{cat, chicken, cloud, cow, dog,\\ duck, fence, grass, horse, house, \\pig, rabbit, sheep, tree }  & & \\

        \frame{\includegraphics[width=0.23\linewidth]{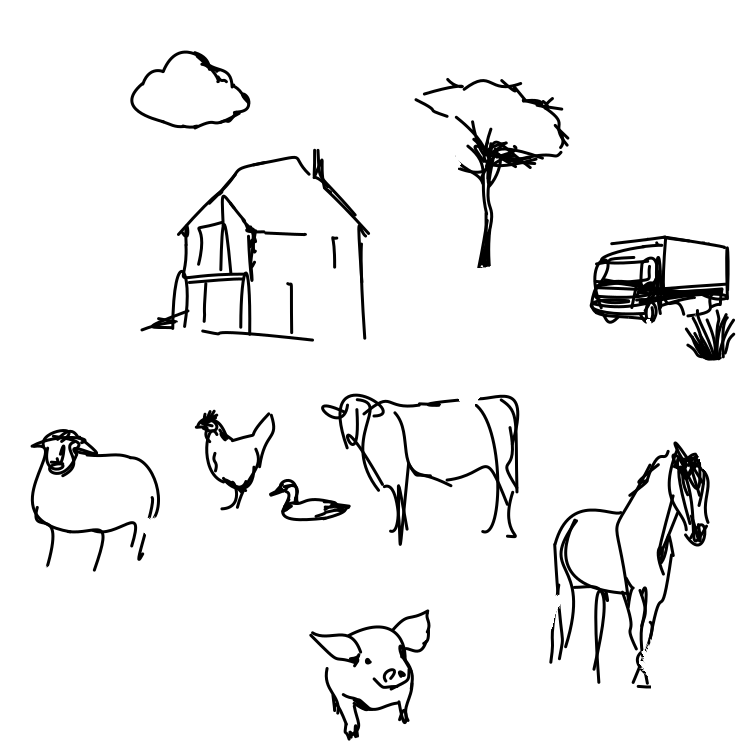}} &
        \frame{\includegraphics[width=0.23\linewidth]{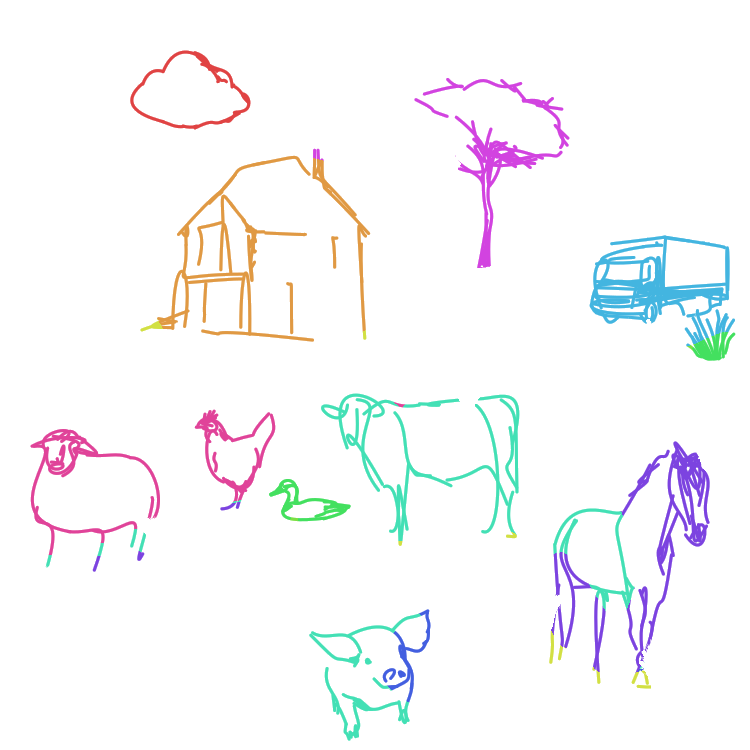}} &
        \frame{\includegraphics[width=0.23\linewidth]{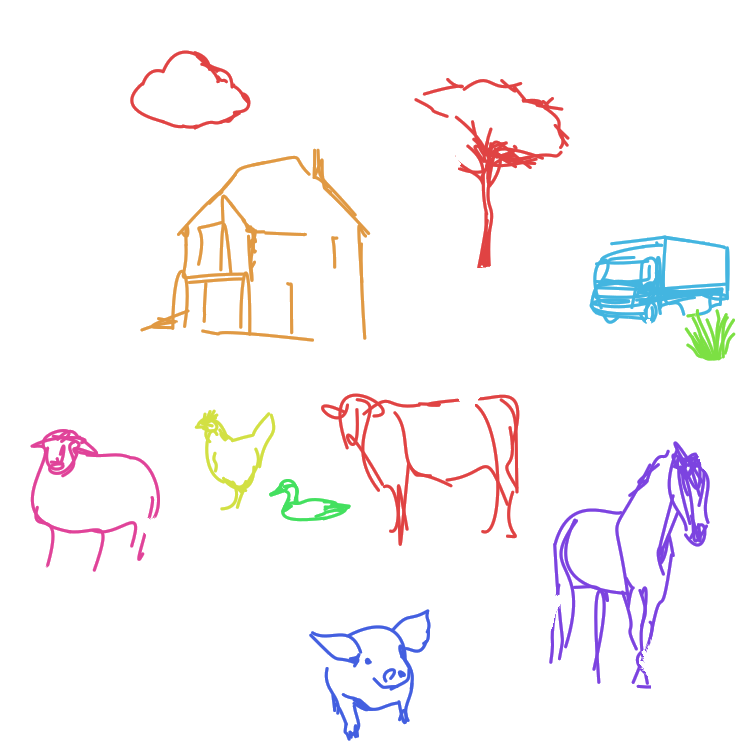}} &
        \frame{\includegraphics[width=0.23\linewidth]{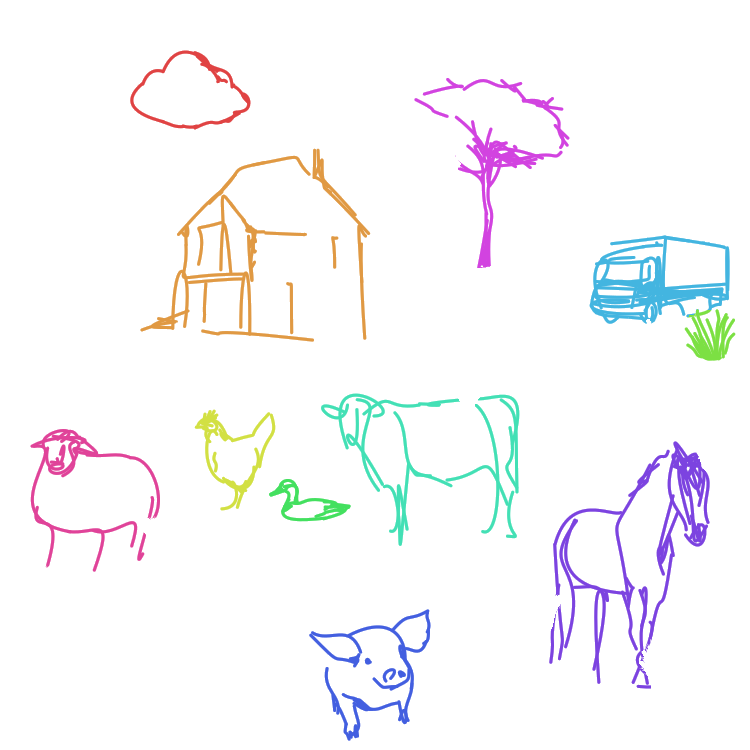}} \\
         & \makecell{chicken, cloud, cow, duck, grass,\\ horse, house, pig, sheep, tree, truck}  & & \\

        \frame{\includegraphics[width=0.23\linewidth]{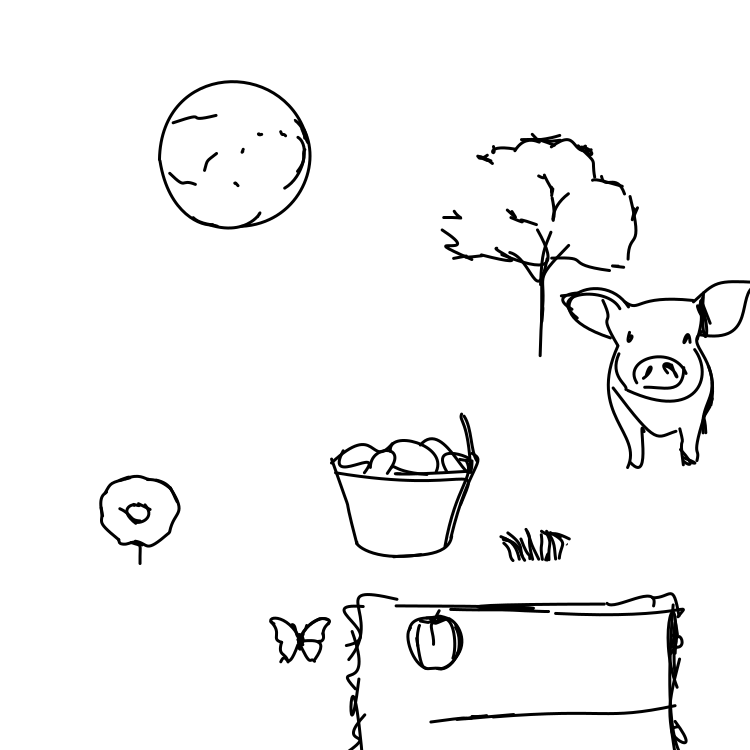}} &
        \frame{\includegraphics[width=0.23\linewidth]{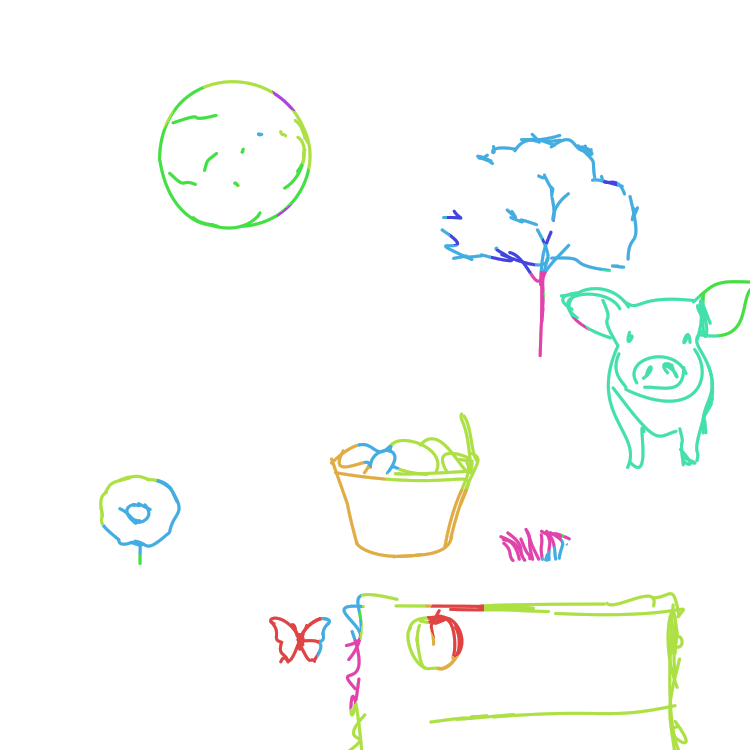}} &
        \frame{\includegraphics[width=0.23\linewidth]{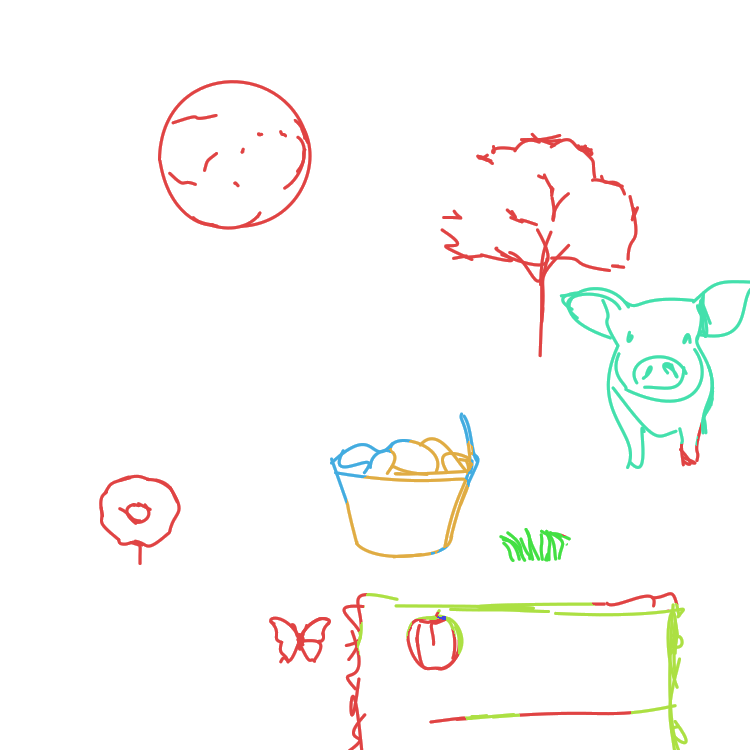}} &
        \frame{\includegraphics[width=0.23\linewidth]{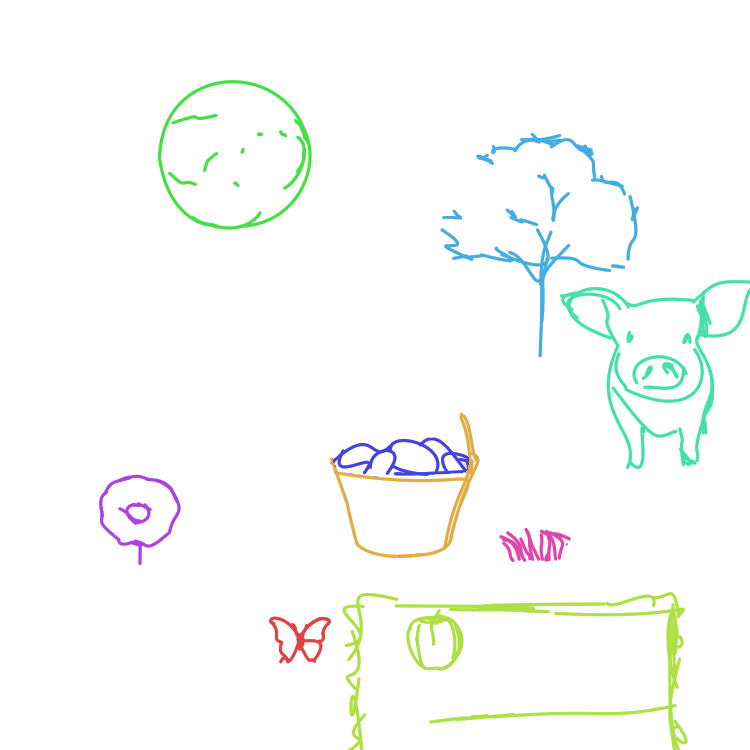}} \\
         & \makecell{apple, basket, butterfly, flower, grass,\\ picnic rug, pig, sun, tree}  & & \\

        \frame{\includegraphics[width=0.23\linewidth]{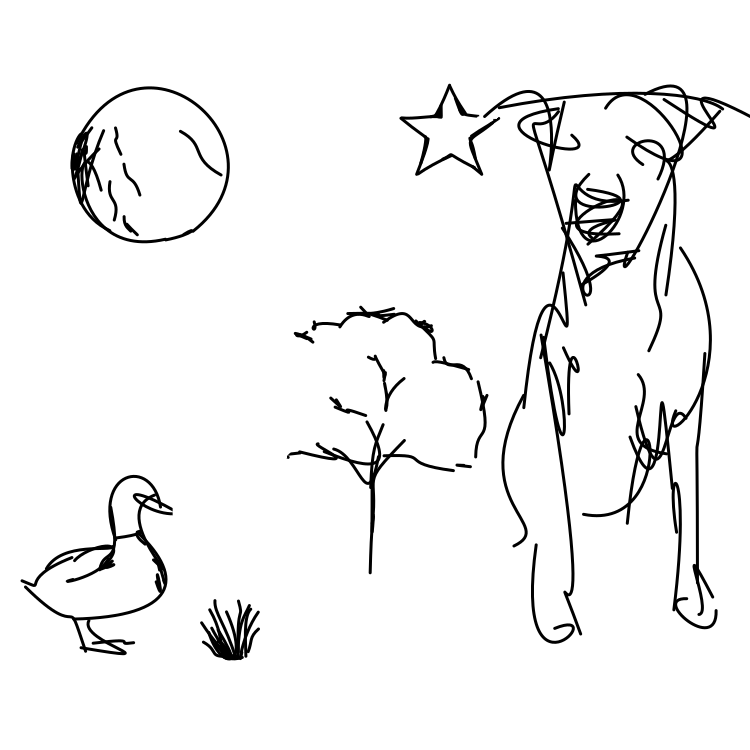}} &
        \frame{\includegraphics[width=0.23\linewidth]{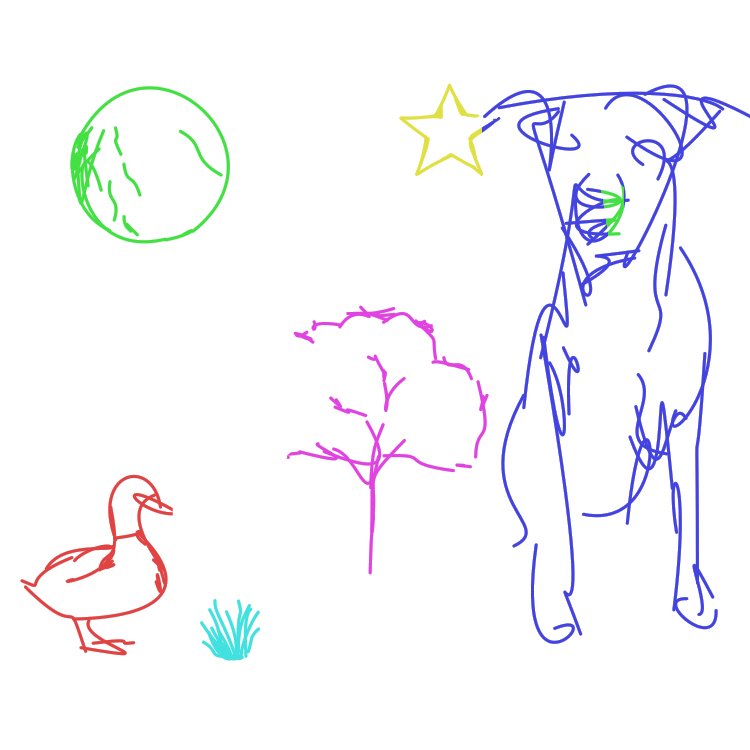}} &
        \frame{\includegraphics[width=0.23\linewidth]{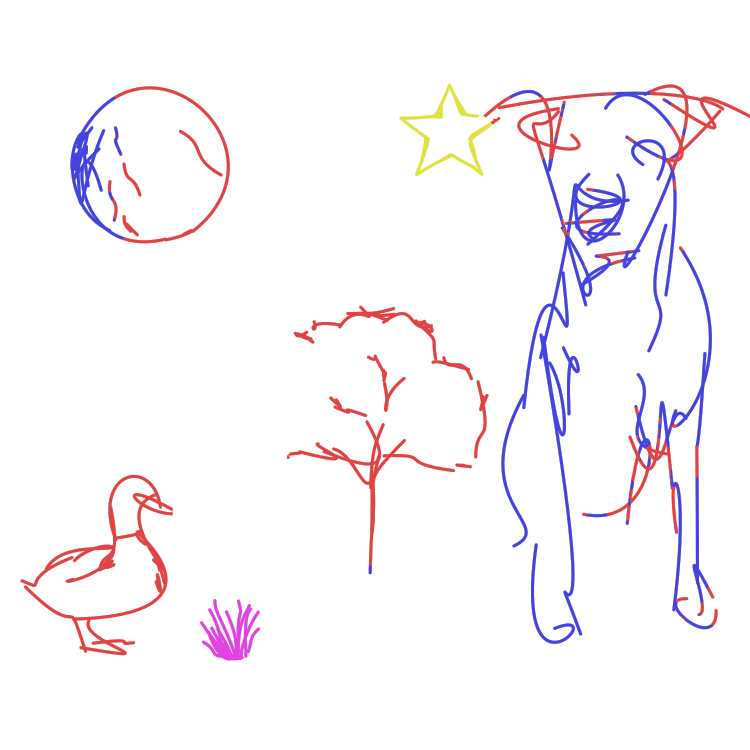}} &
        \frame{\includegraphics[width=0.23\linewidth]{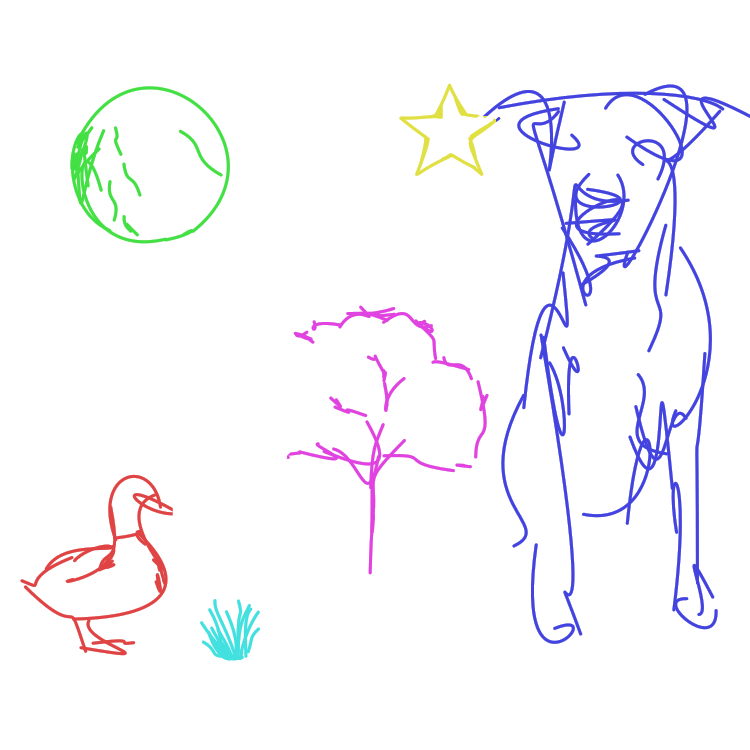}} \\
         & \makecell{dog, duck, grass, moon, star, tree}  & & 
    \end{tabular}
    }
    \vspace{-0.1cm}
    \caption{\textbf{Qualitative comparison of segmentation on filtered CLIPasso dataset. } We prompt Bourouis \etal’s method with ground truth labels and use confidence threshold of 0.01 to ensure all sketch pixels are segmented, while using SketchSeger as-is since it does not accept input labels. However, Bourouis \etal struggles to accurately identify instances of the correct class, often assigning multiple class labels to the same object, as indicated by the color gradients. SketchSeger also fails to provide clean segmentations for many scenes. In contrast, our method effectively segments object instances, ensuring clear separation and consistent labeling.}
    \label{fig:comparison_openvocab_clipasso}
    }
\end{figure*}

\begin{figure*}
    \centering
    \setlength{\tabcolsep}{2pt}
    {\small
    \resizebox{0.95\textwidth}{!}{ 
    \begin{tabular}{c @{\hskip 10pt} c c c}
        Input & Bourouis \etal & SketchSeger & \textbf{Ours}  \\
        \frame{\includegraphics[trim=200 50 0 100, clip, width=0.23\linewidth]{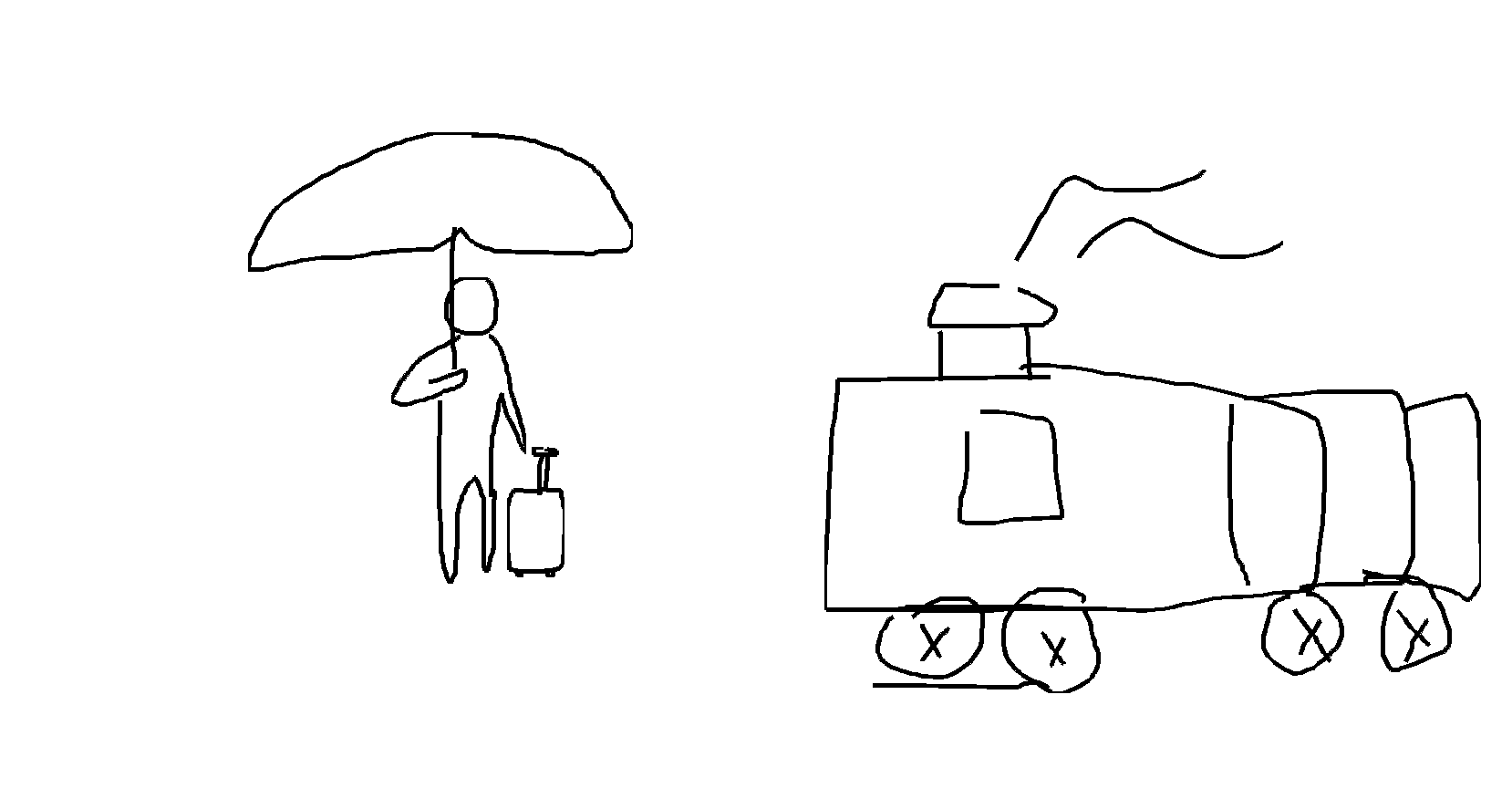}} &
        \frame{\includegraphics[trim=200 50 0 100, clip, width=0.23\linewidth]{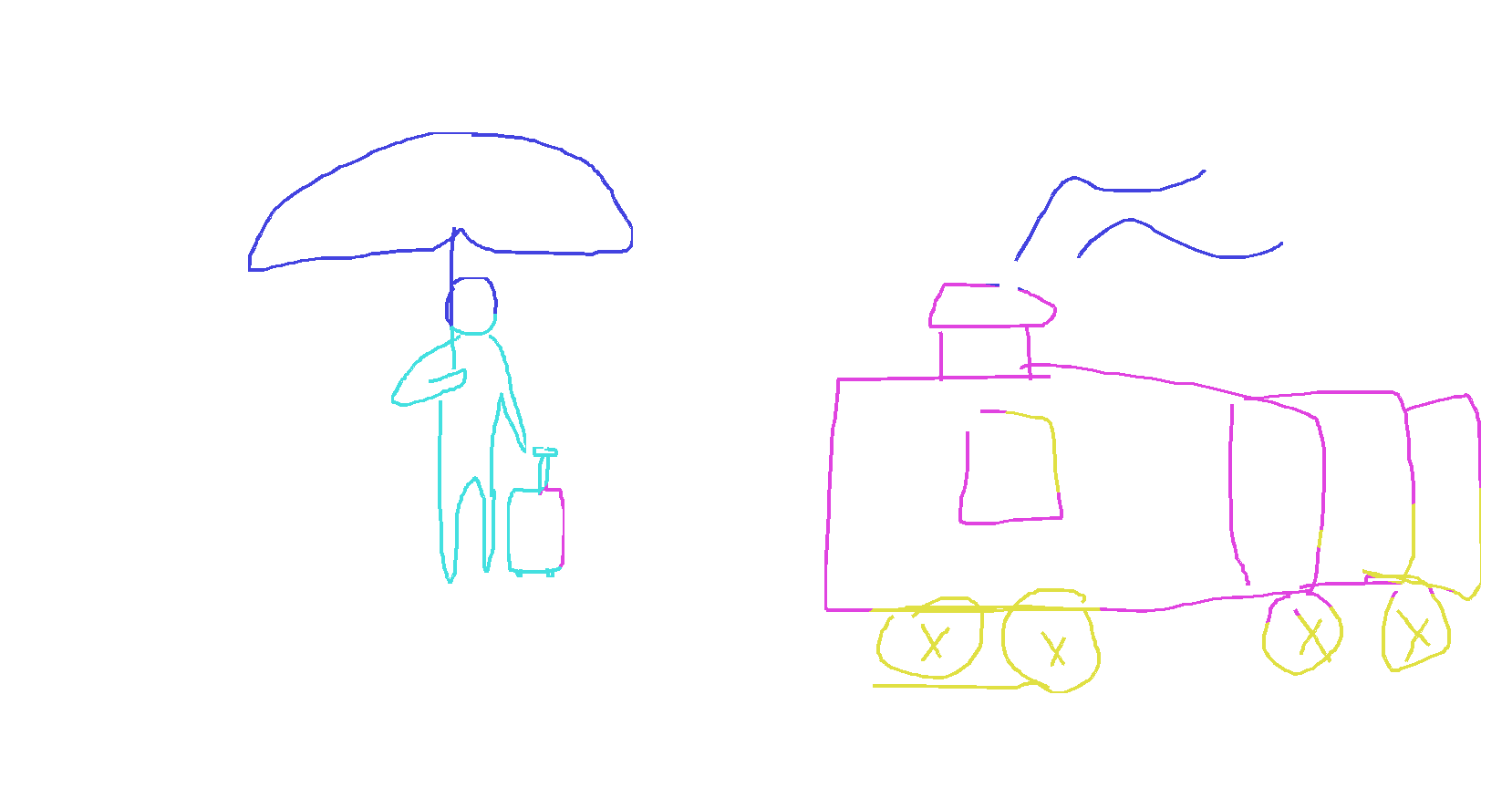}} &
        \frame{\includegraphics[trim=200 50 0 100, clip, width=0.23\linewidth]{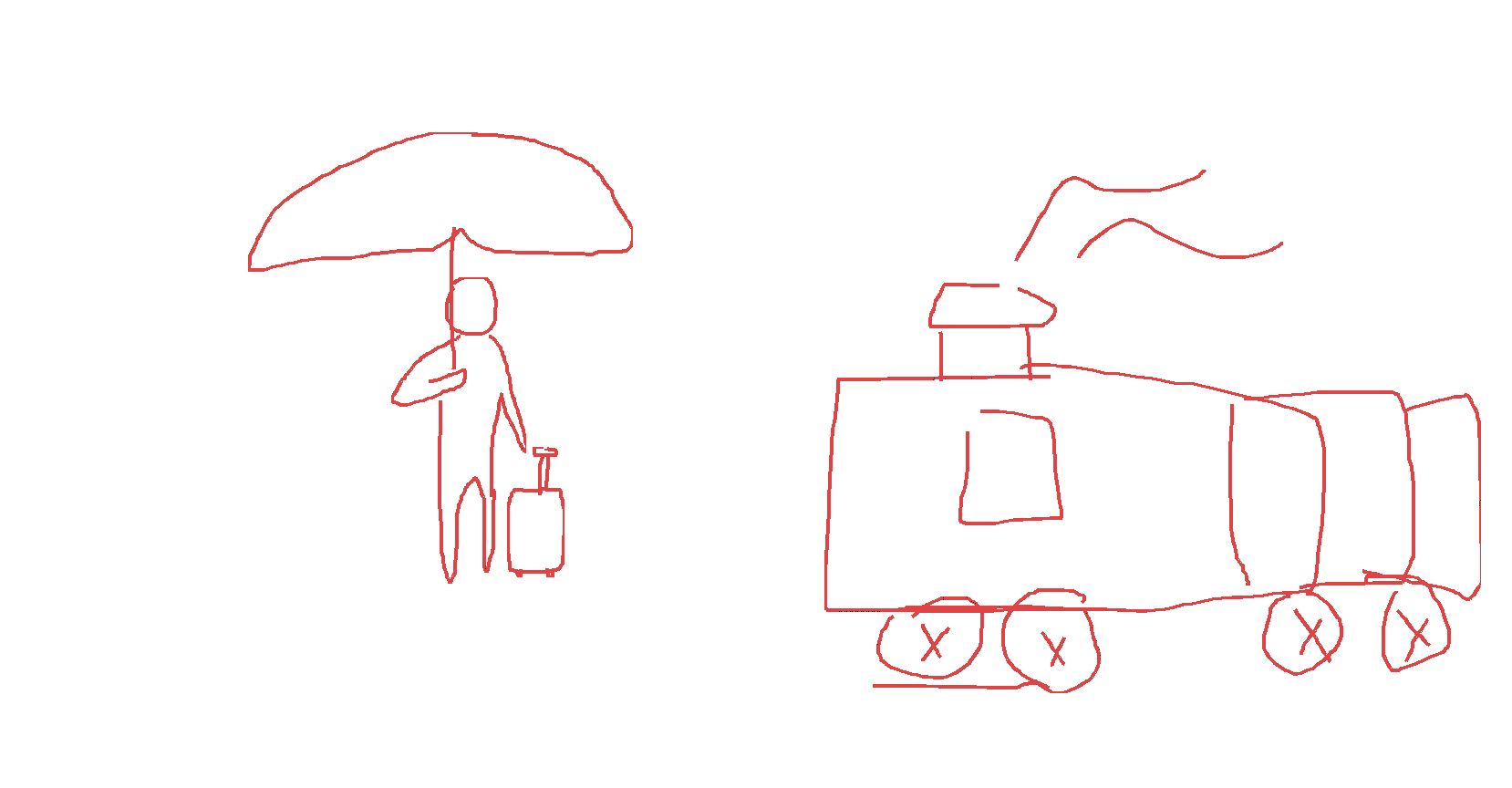}} &
        \frame{\includegraphics[trim=200 50 0 100, clip, width=0.23\linewidth]{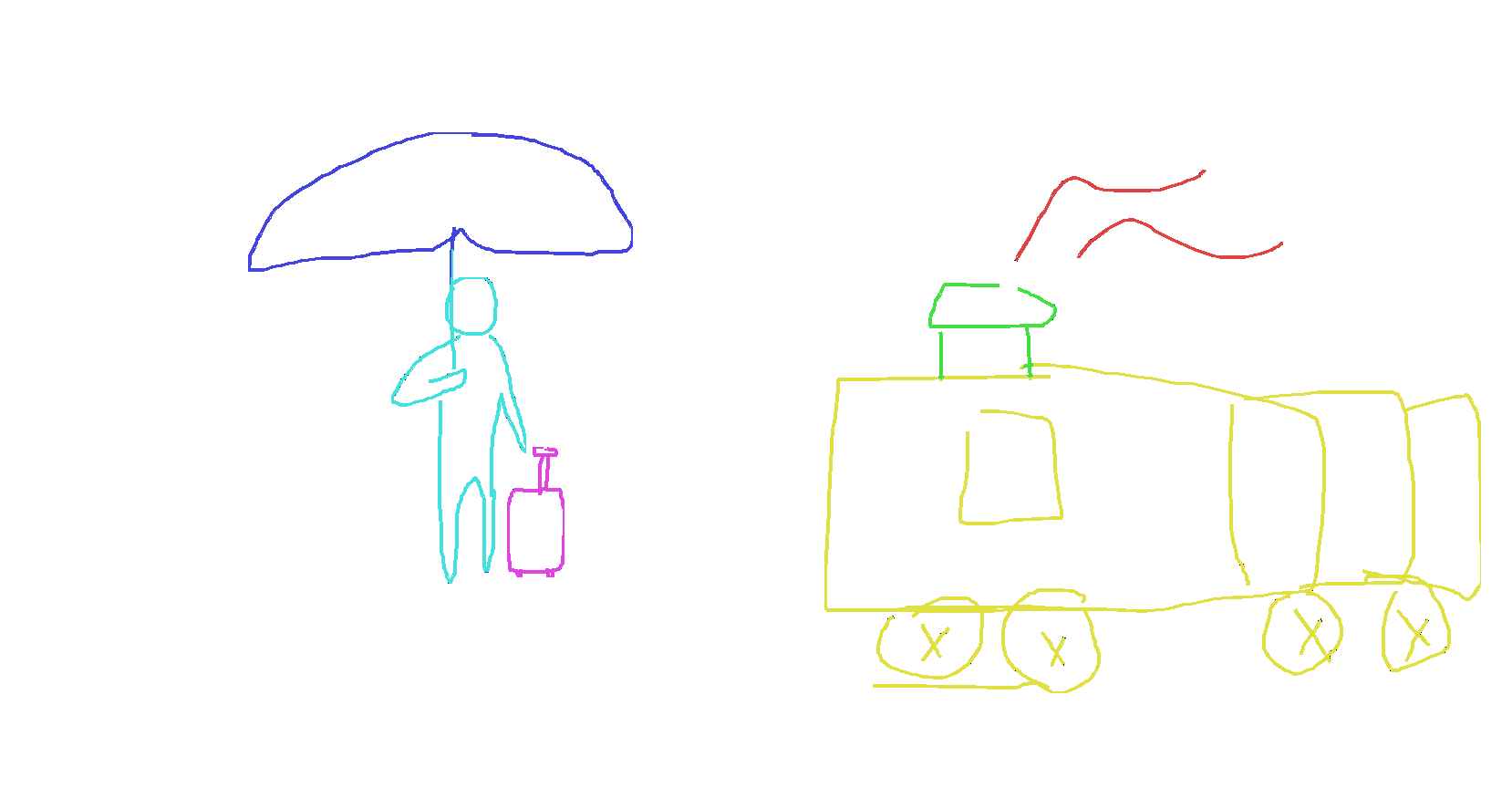}} \\
        & \makecell{person, suitcase, train, umbrella}  & & \\

        \frame{\includegraphics[trim=200 200 100 0, clip, width=0.23\linewidth]{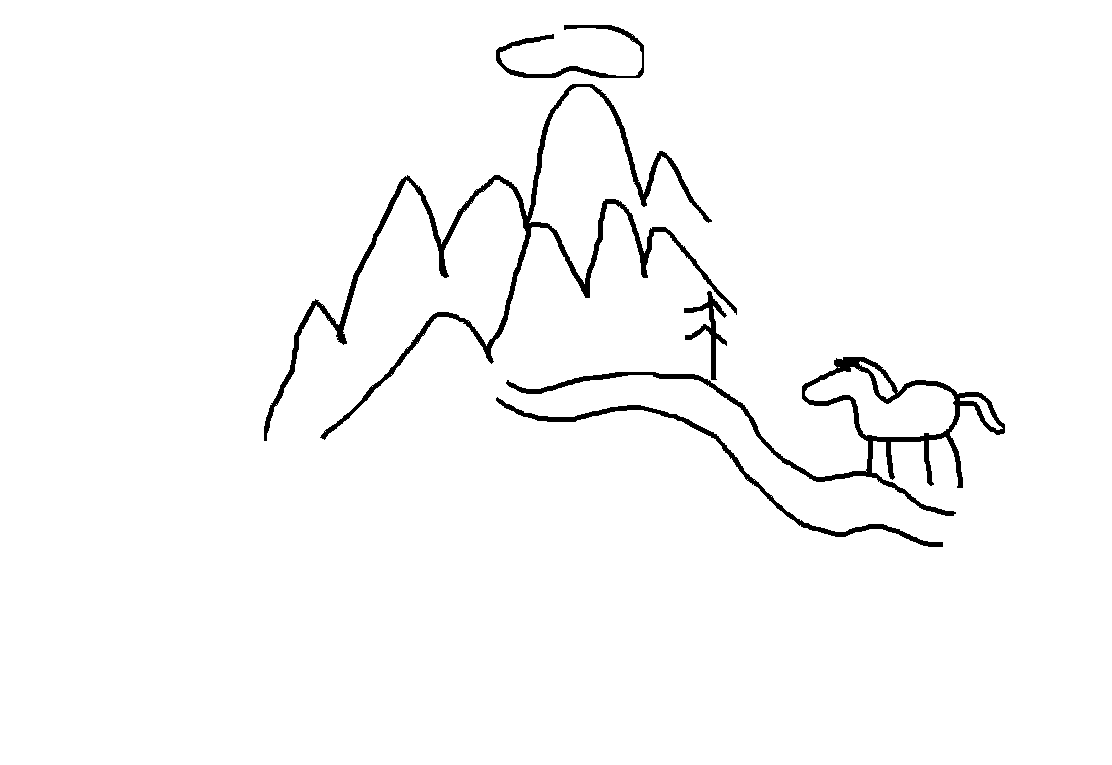}} &
        \frame{\includegraphics[trim=200 200 100 0, clip, width=0.23\linewidth]{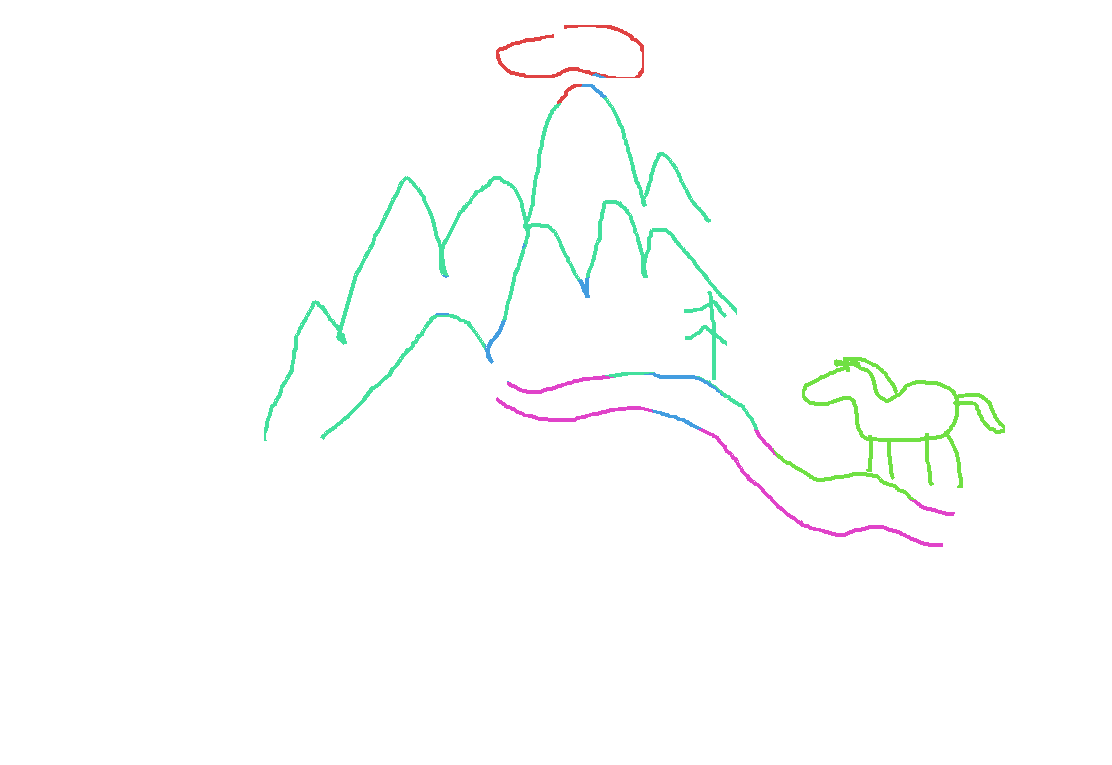}} &
        \frame{\includegraphics[trim=200 200 100 0, clip, width=0.23\linewidth]{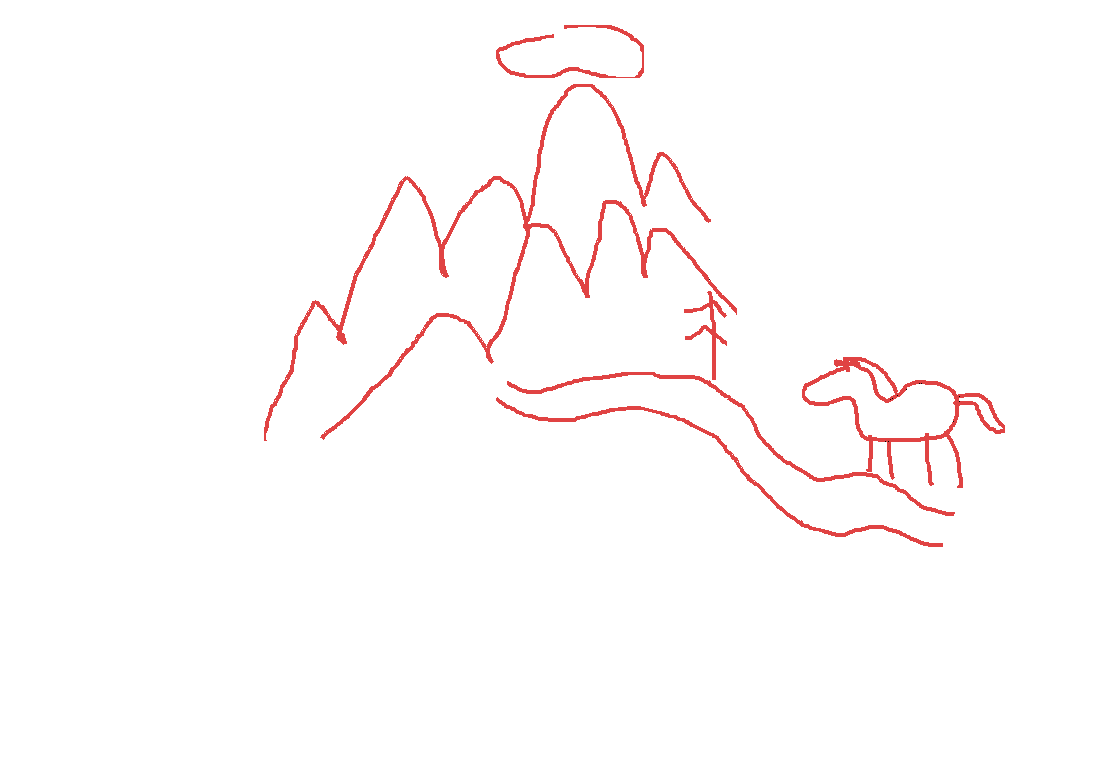}} &
        \frame{\includegraphics[trim=200 200 100 0, clip, width=0.23\linewidth]{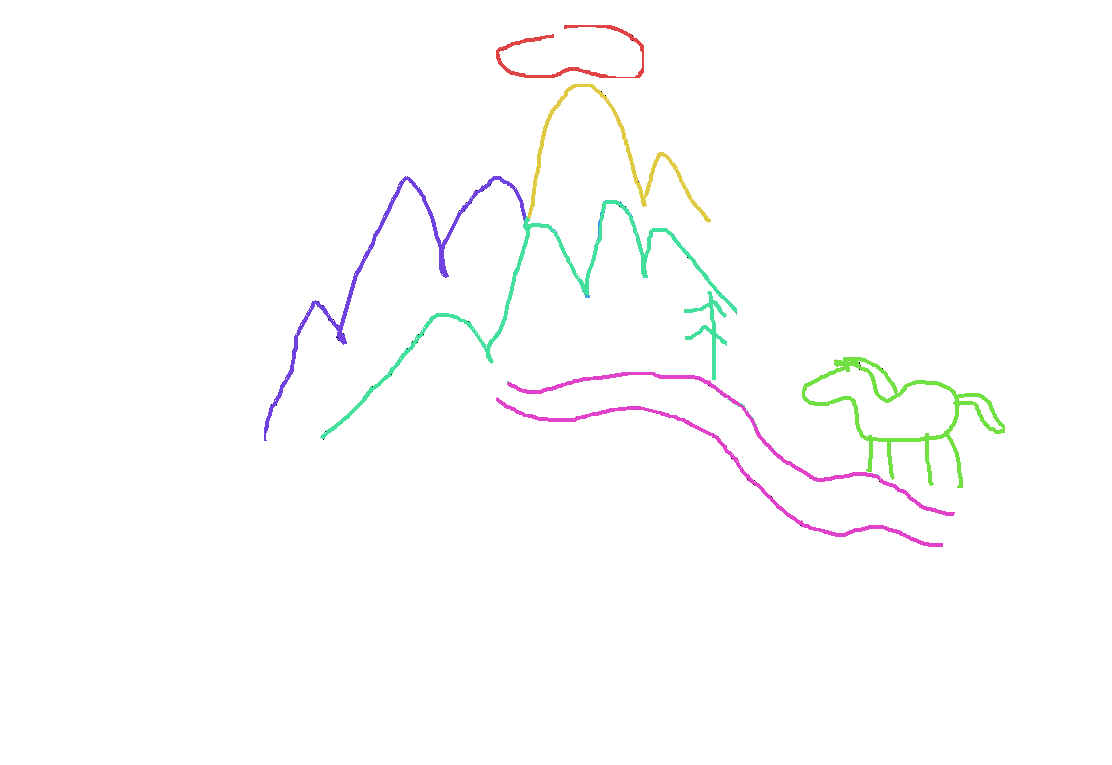}} \\
        & \makecell{cloud, horse, mountain, river, tree}  & & \\

        \frame{\includegraphics[trim=450 200 400 100, clip, width=0.23\linewidth]{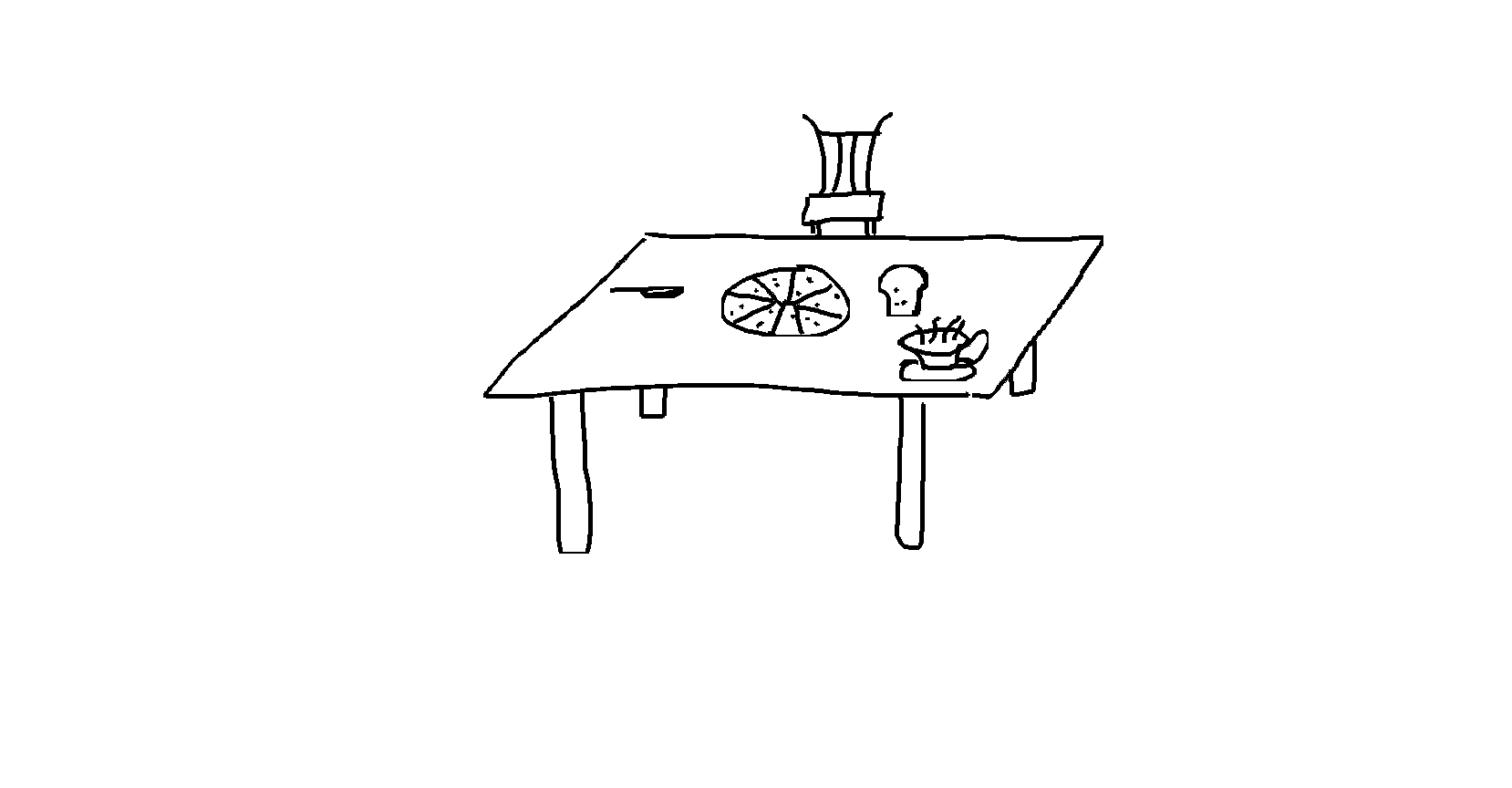}} &
        \frame{\includegraphics[trim=450 200 400 100, clip, width=0.23\linewidth]{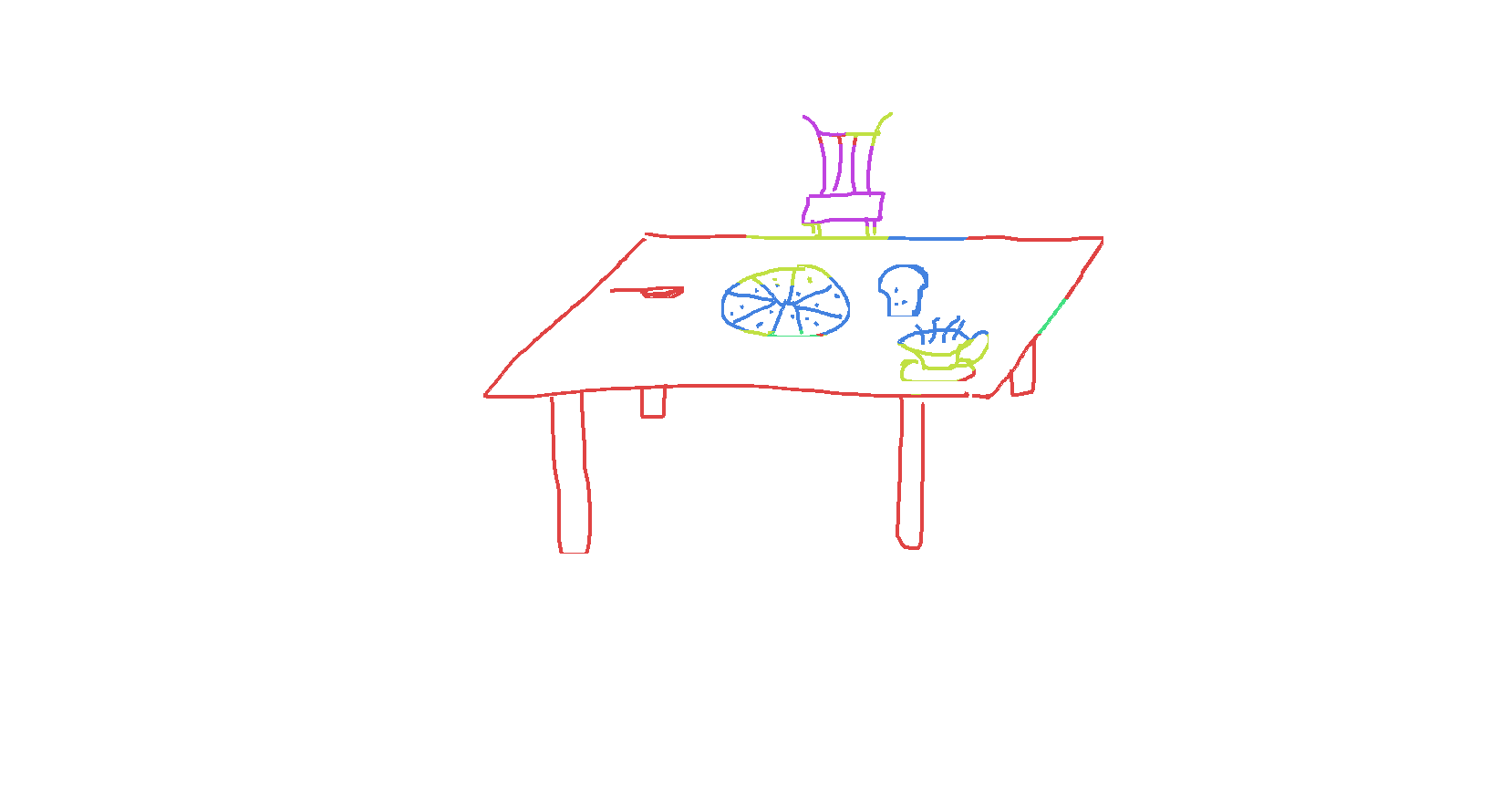}} &
        \frame{\includegraphics[trim=450 200 400 100, clip, width=0.23\linewidth]{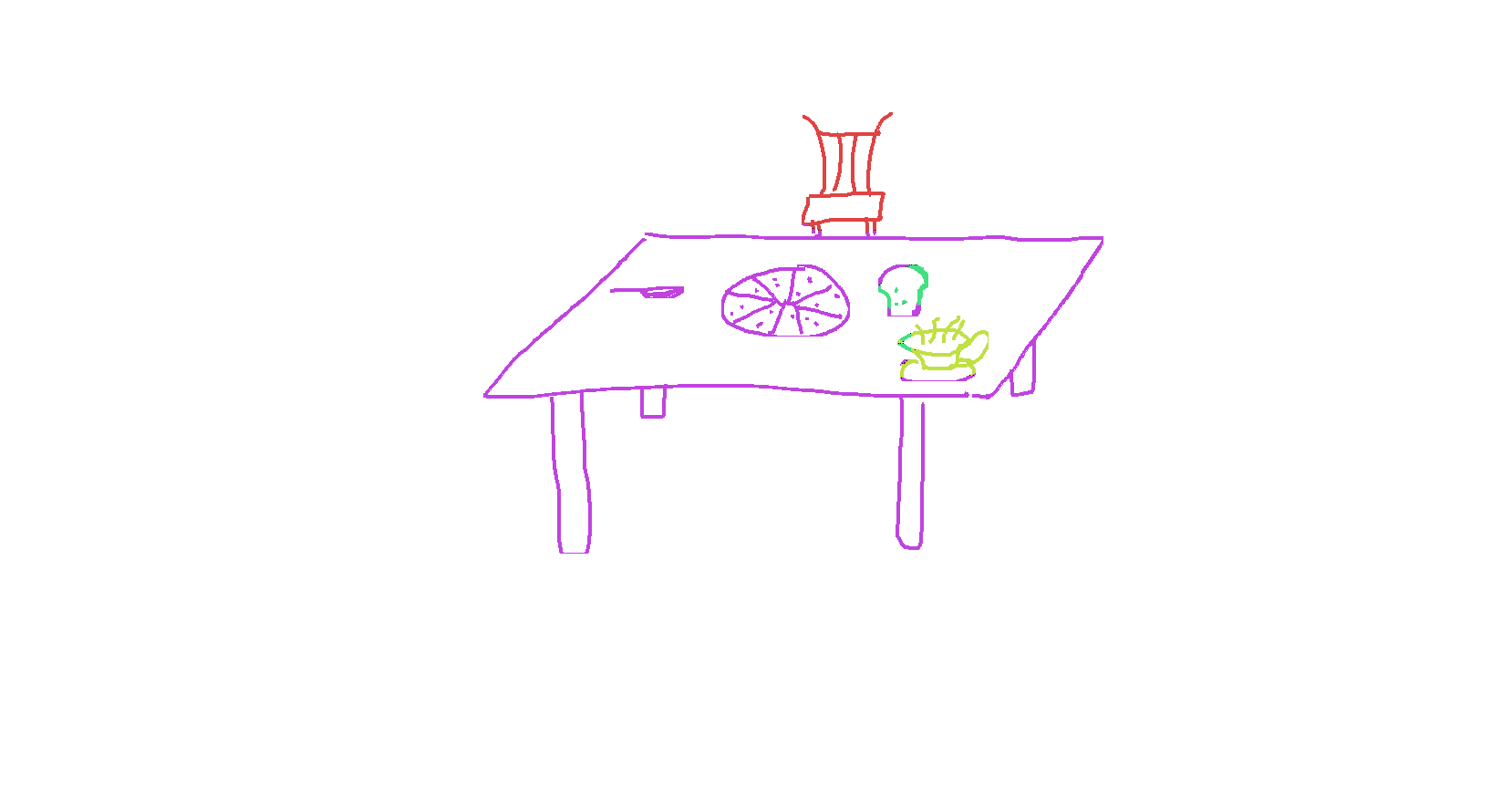}} &
        \frame{\includegraphics[trim=450 200 400 100, clip, width=0.23\linewidth]{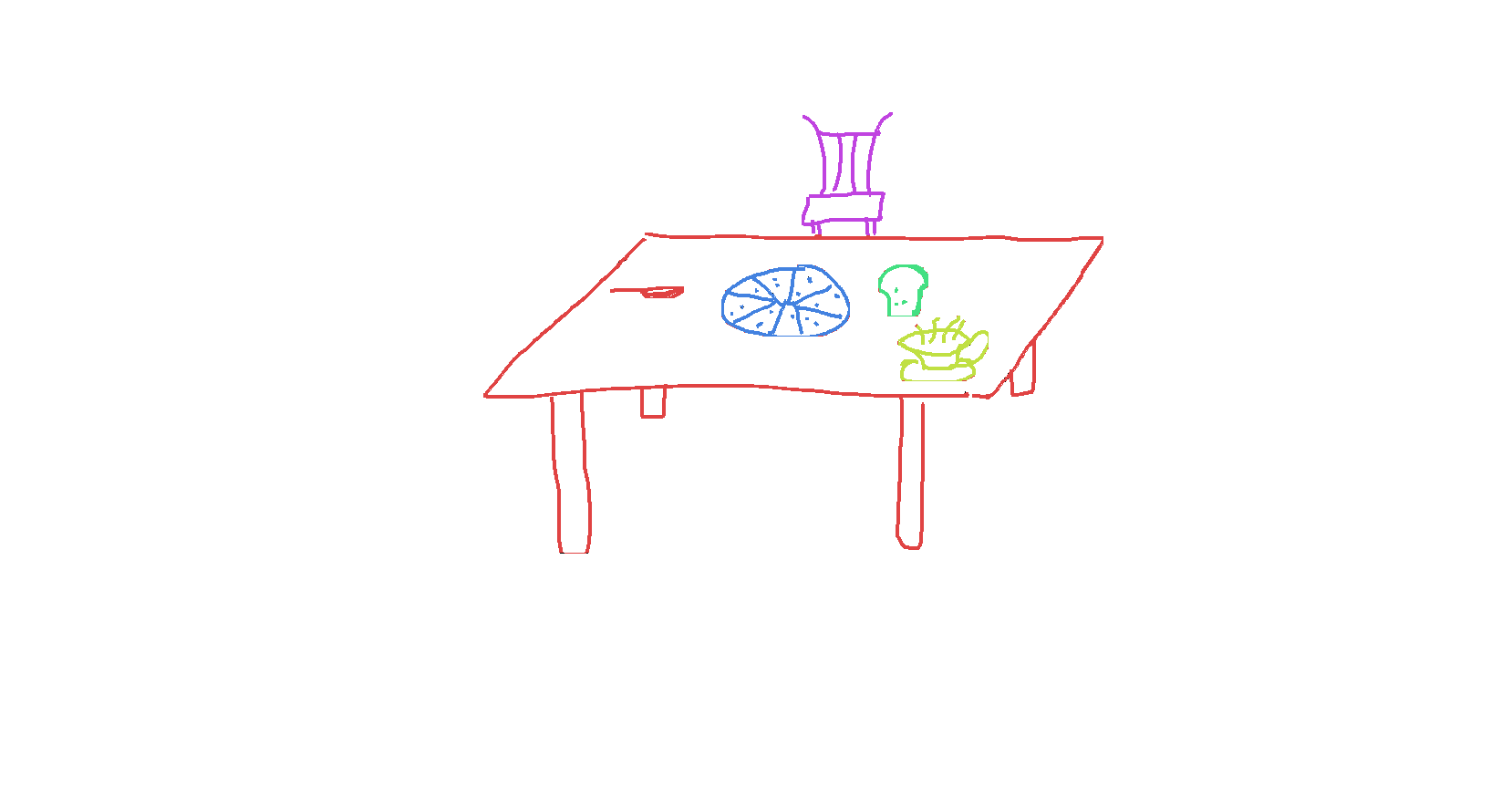}} \\
        & \makecell{bread, chair, cup, knife, pizza, table}  & & \\

        \frame{\includegraphics[trim=0 0 300 200, clip, width=0.23\linewidth]{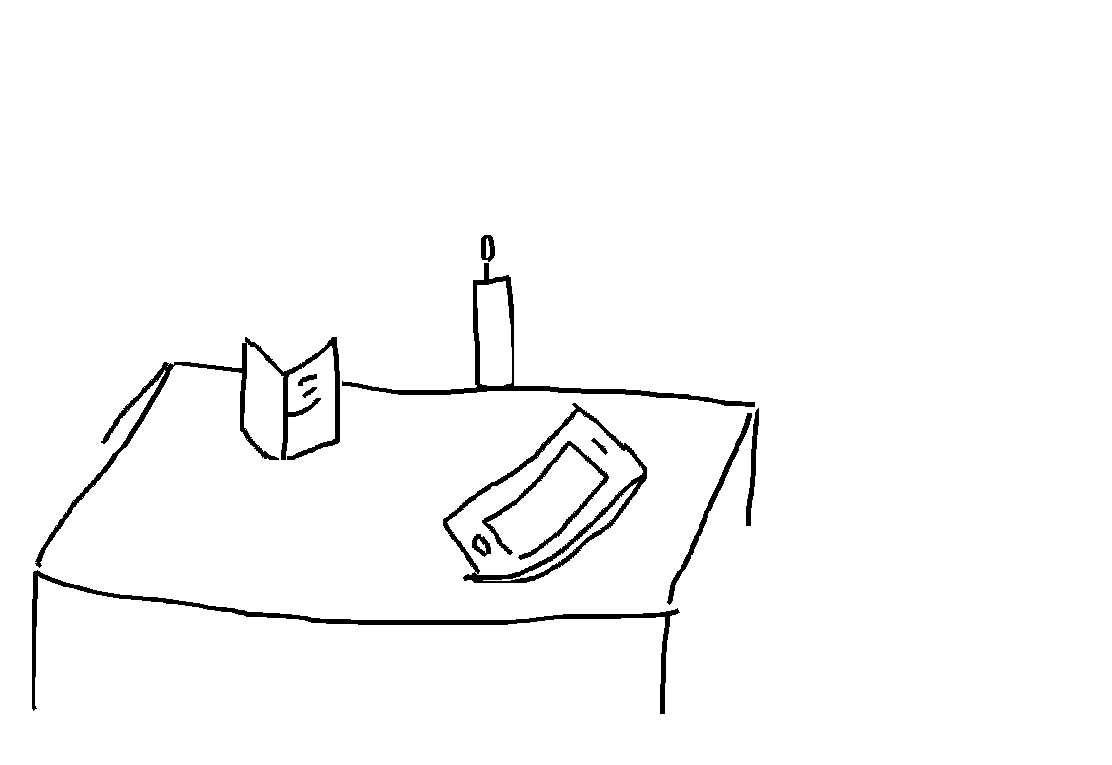}} &
        \frame{\includegraphics[trim=0 0 300 200, clip,width=0.23\linewidth]{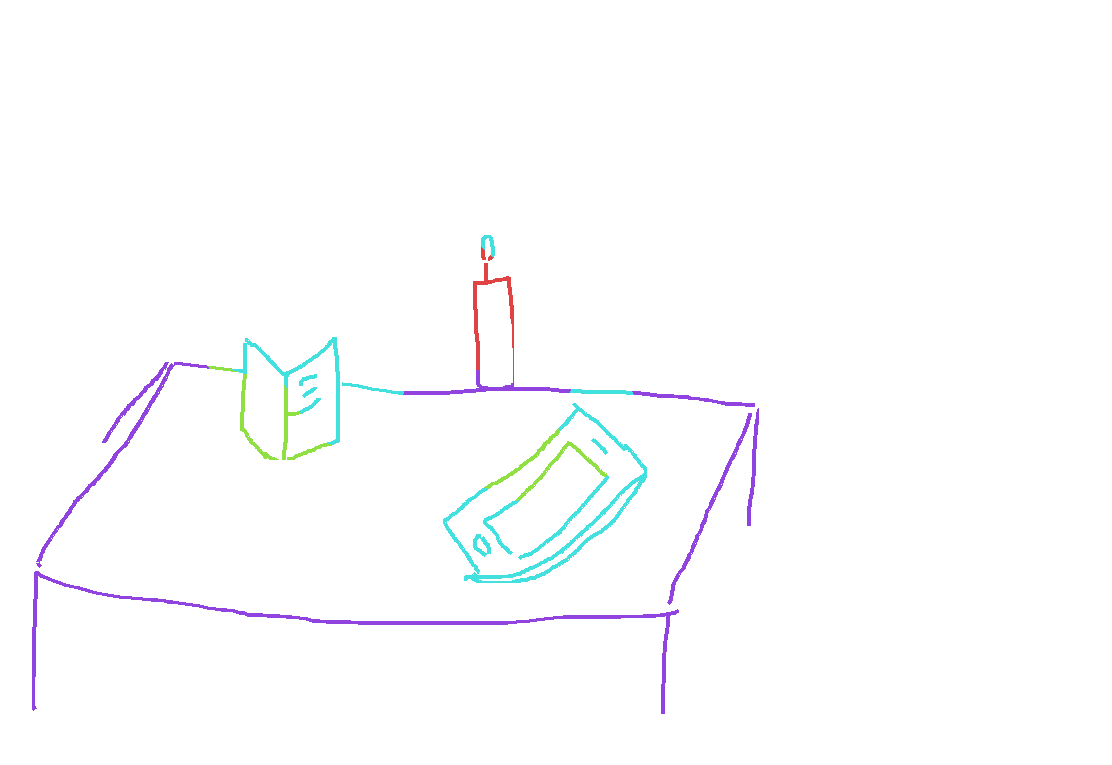}} &
        \frame{\includegraphics[trim=0 0 300 200, clip,width=0.23\linewidth]{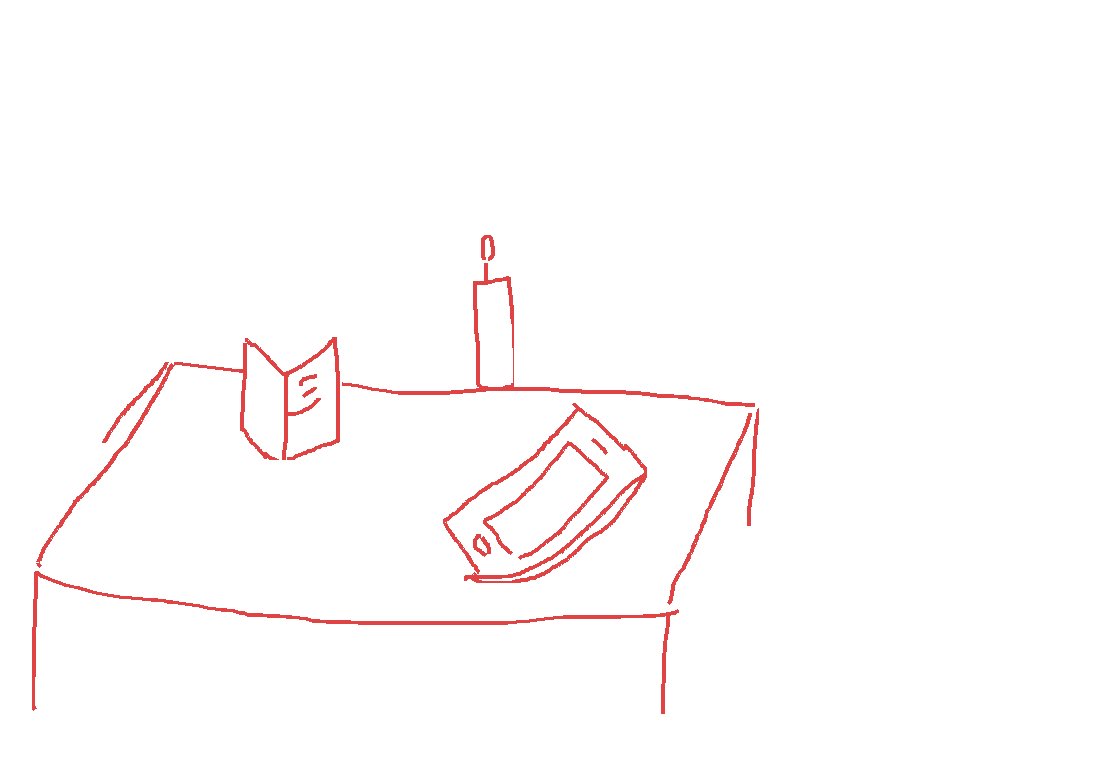}} &
        \frame{\includegraphics[trim=0 0 300 200, clip,width=0.23\linewidth]{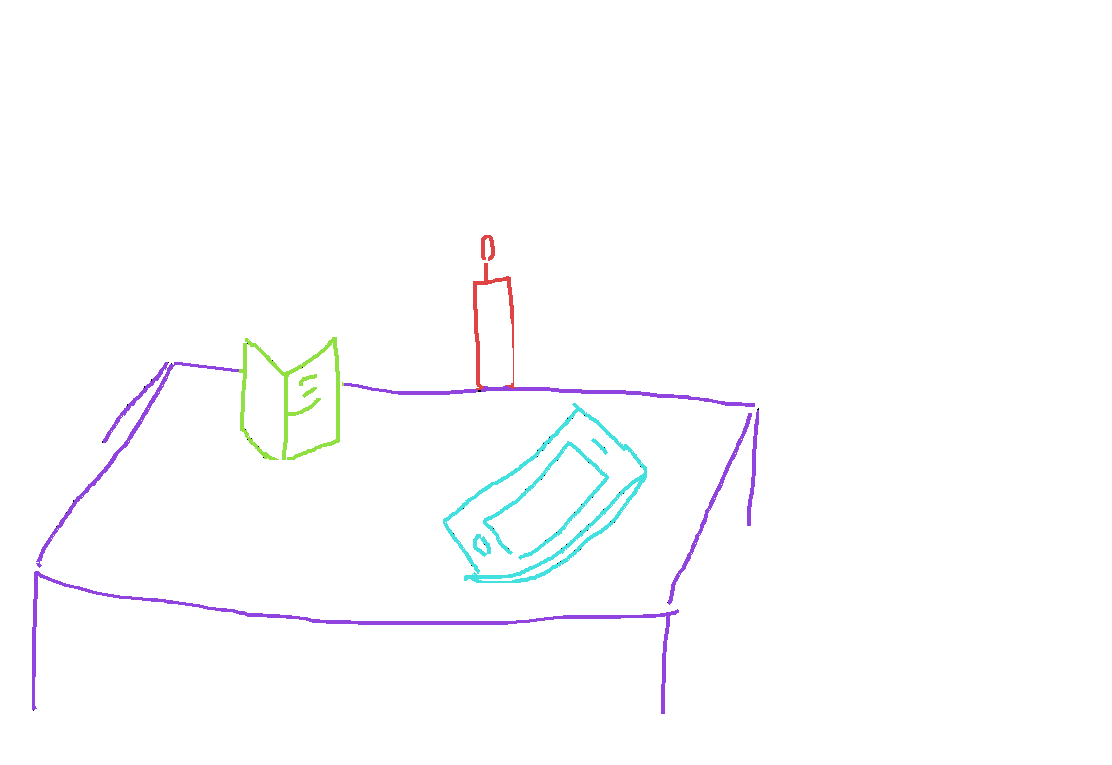}} \\
        & \makecell{book, candle, cell phone, table}  & & \\

        \frame{\includegraphics[trim=0 400 700 0, clip,width=0.23\linewidth]{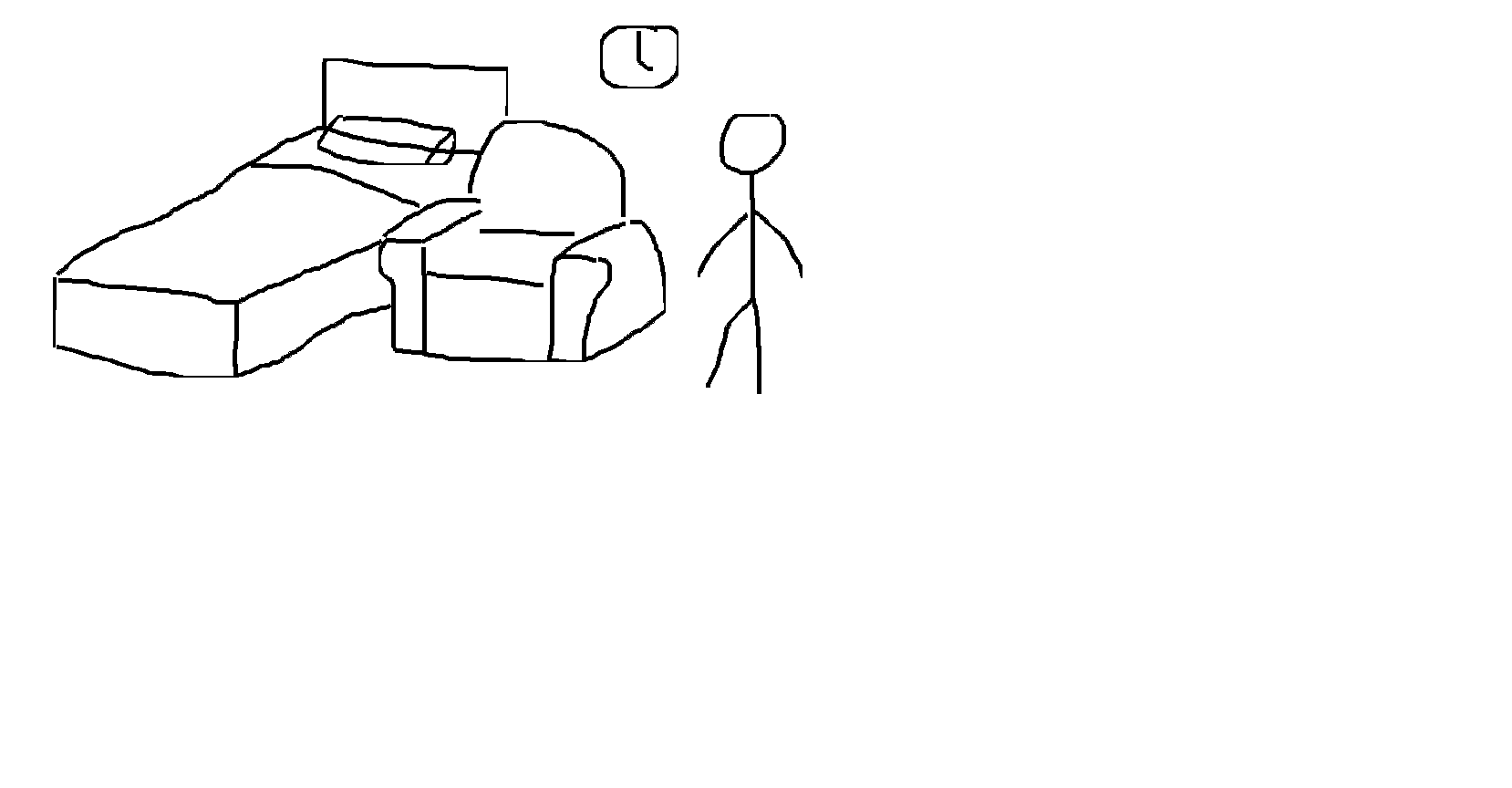}} &
        \frame{\includegraphics[trim=0 400 700 0, clip,width=0.23\linewidth]{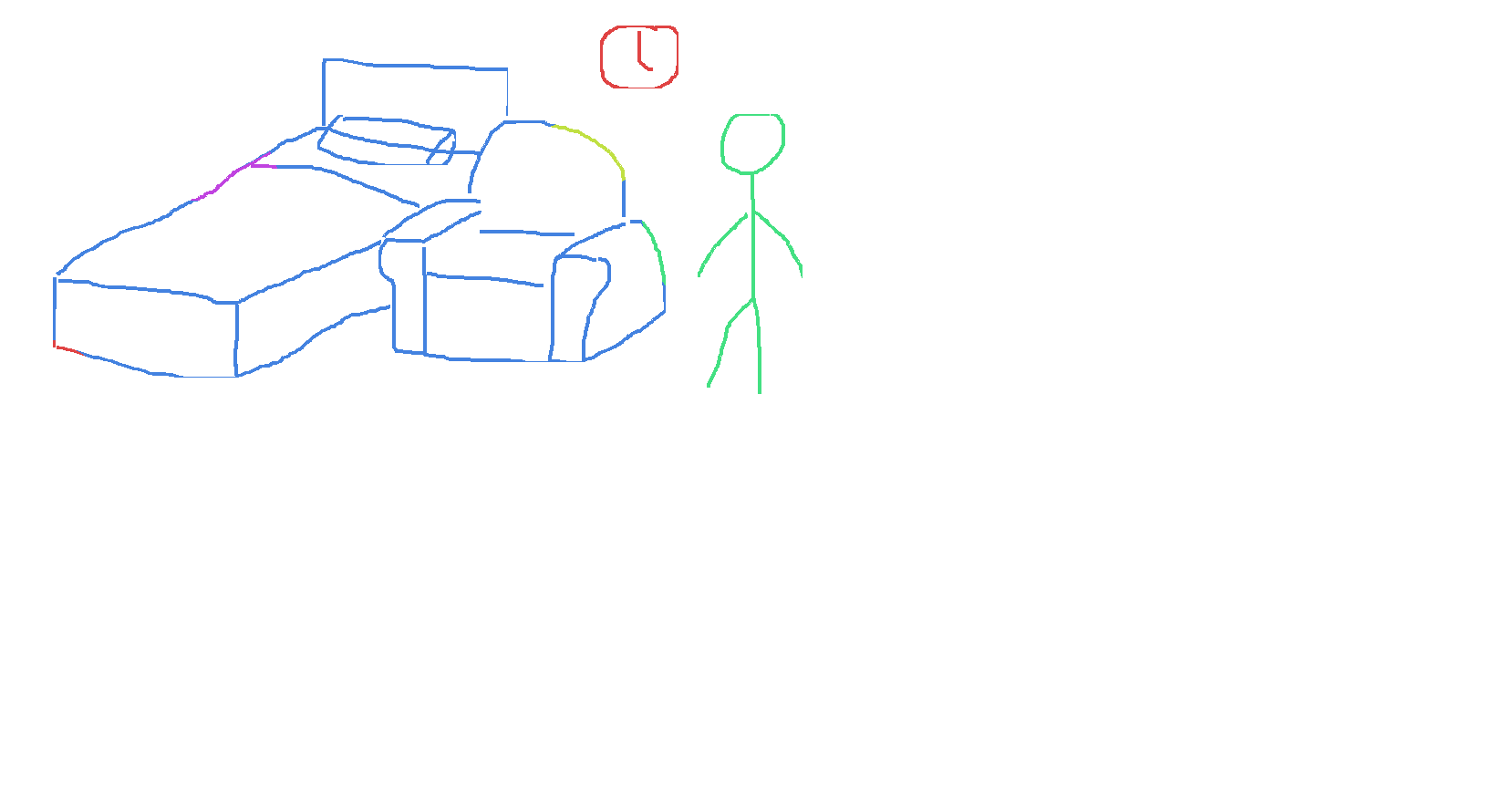}} &
        \frame{\includegraphics[trim=0 400 700 0, clip,width=0.23\linewidth]{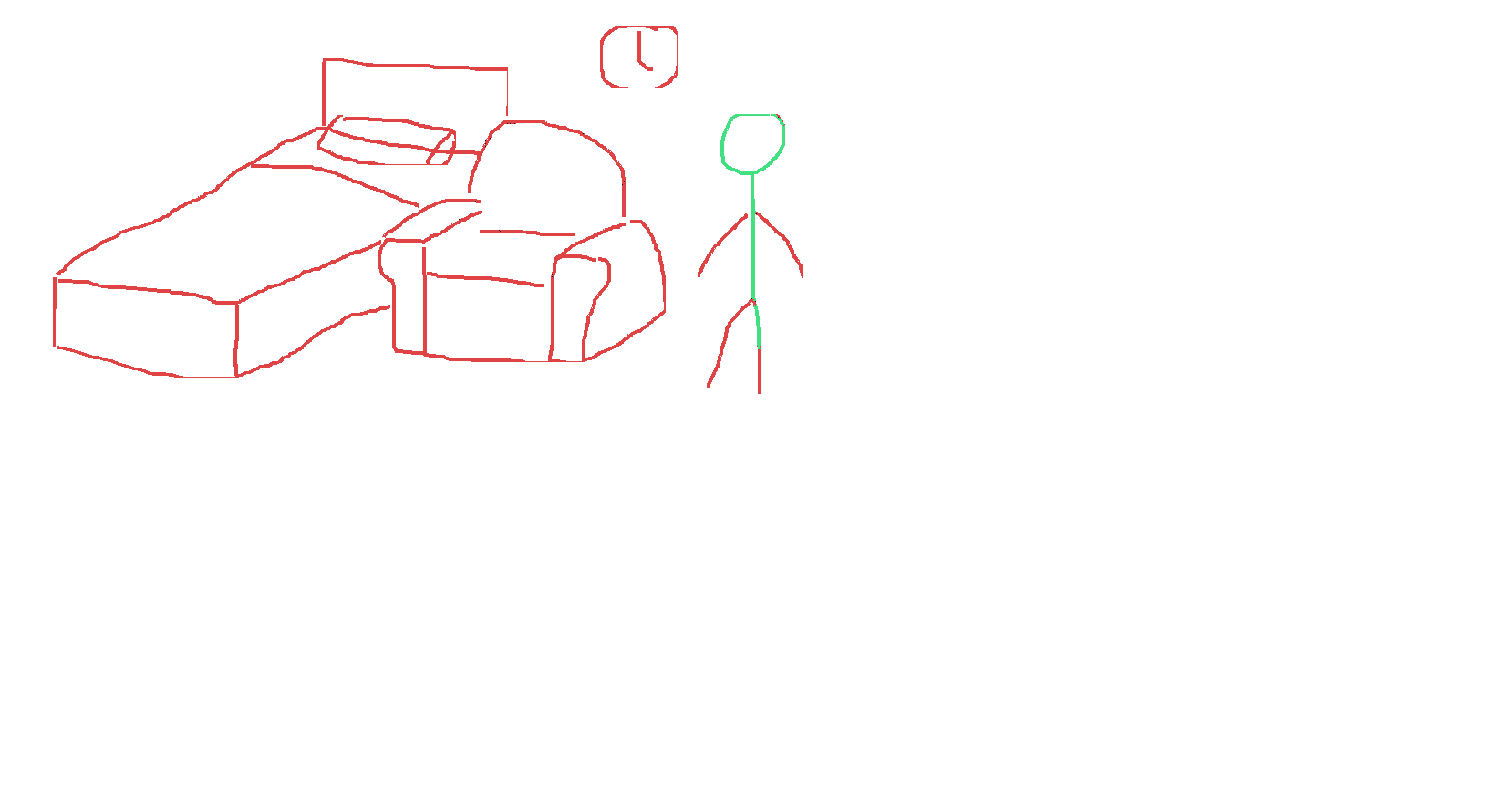}} &
        \frame{\includegraphics[trim=0 400 700 0, clip,width=0.23\linewidth]{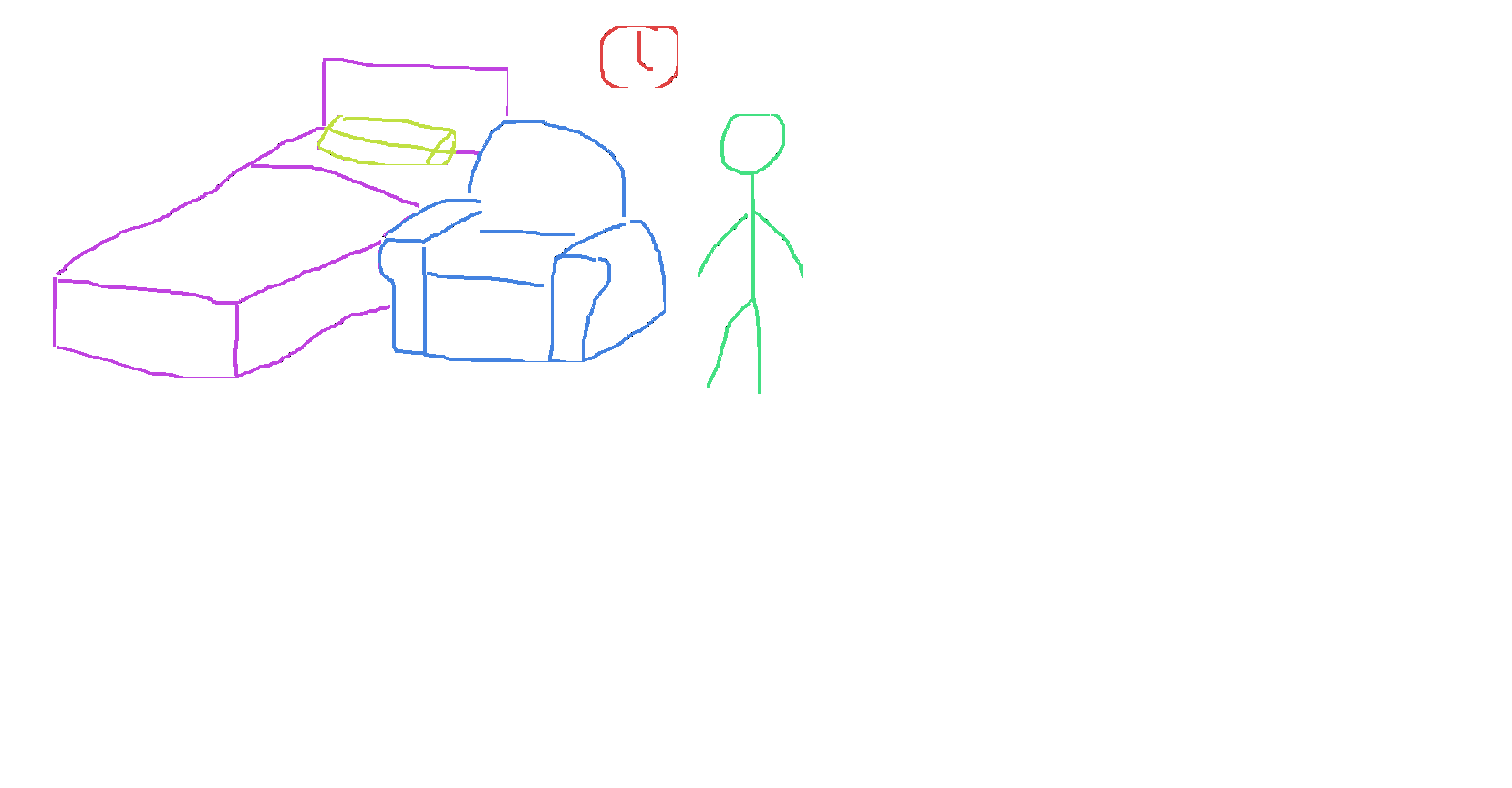}} \\
        & \makecell{bed, clock, couch, person, pillow}  & & \\

        \frame{\includegraphics[trim=250 250 550 200, clip,width=0.23\linewidth]{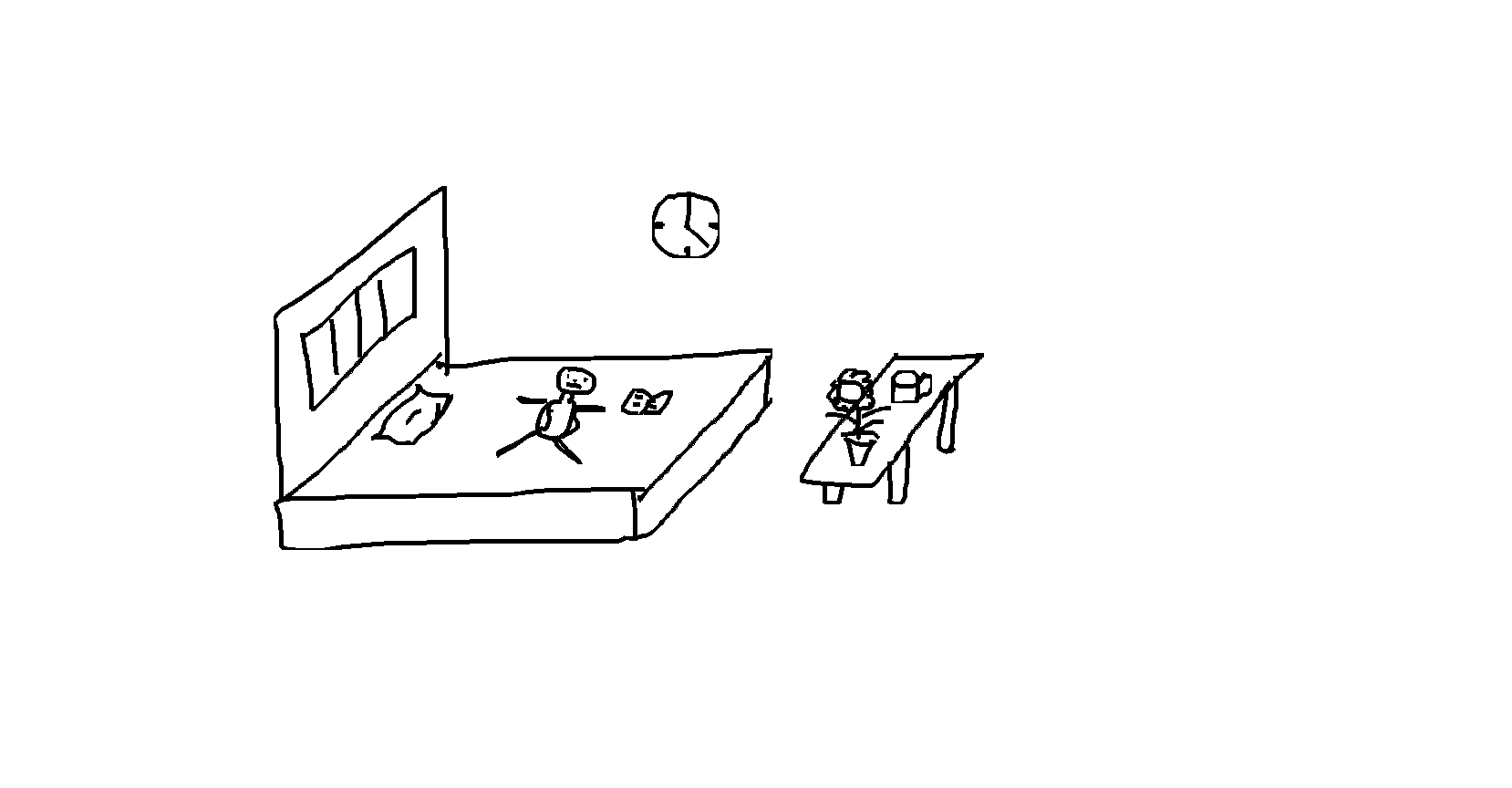}} &
        \frame{\includegraphics[trim=250 250 550 200, clip,width=0.23\linewidth]{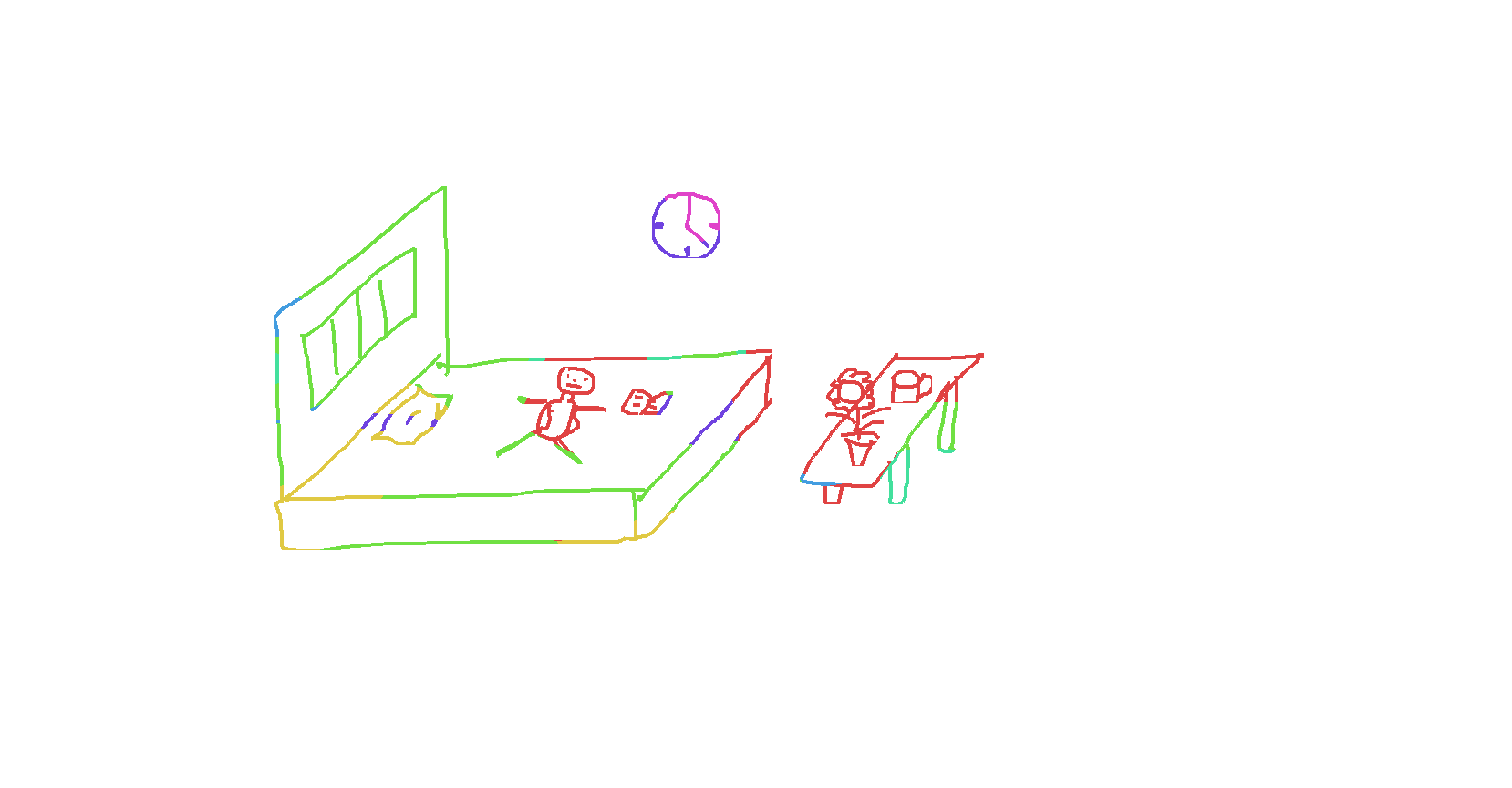}} &
        \frame{\includegraphics[trim=250 250 550 200, clip,width=0.23\linewidth]{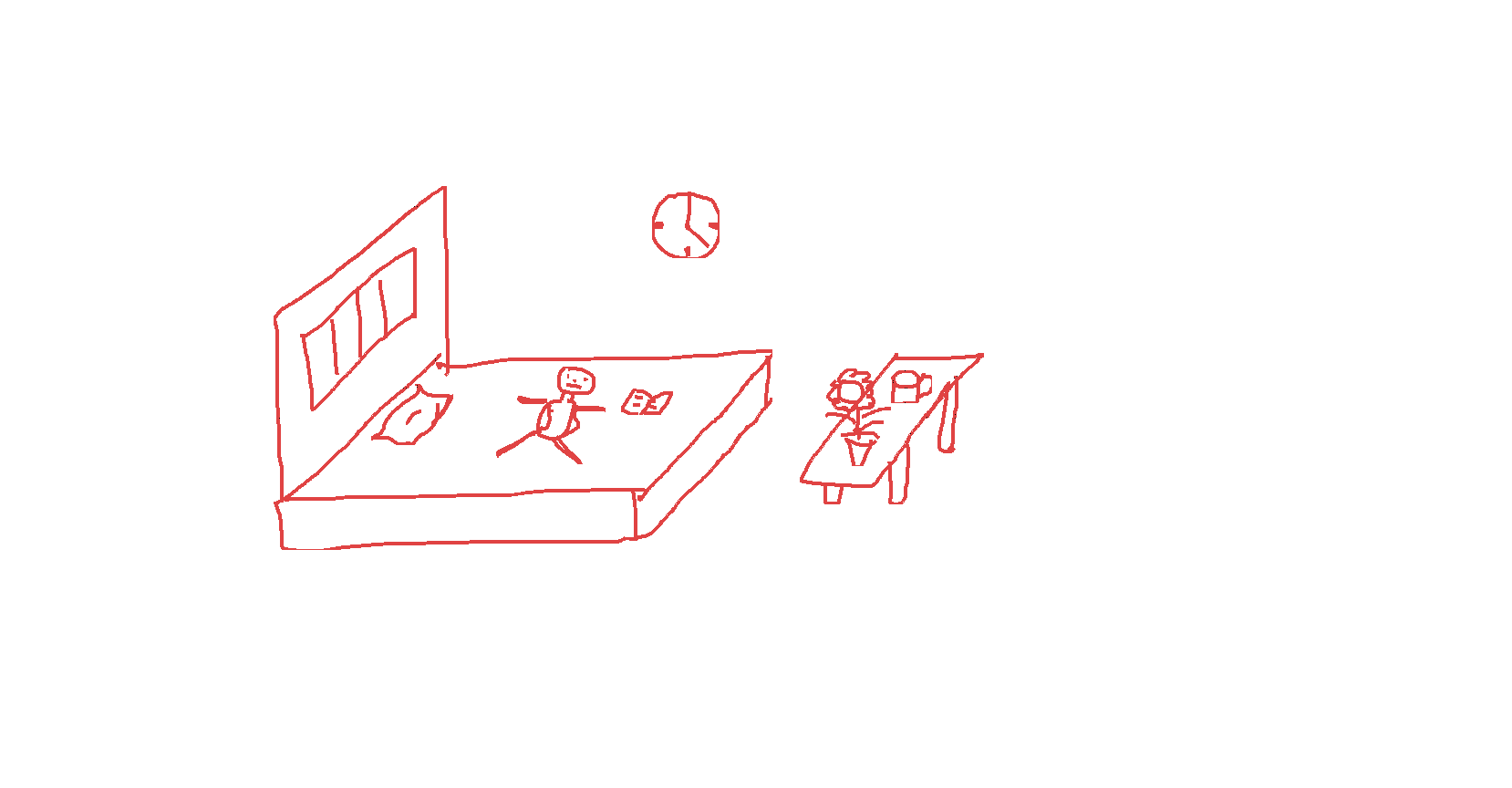}} &
        \frame{\includegraphics[trim=250 250 550 200, clip,width=0.23\linewidth]{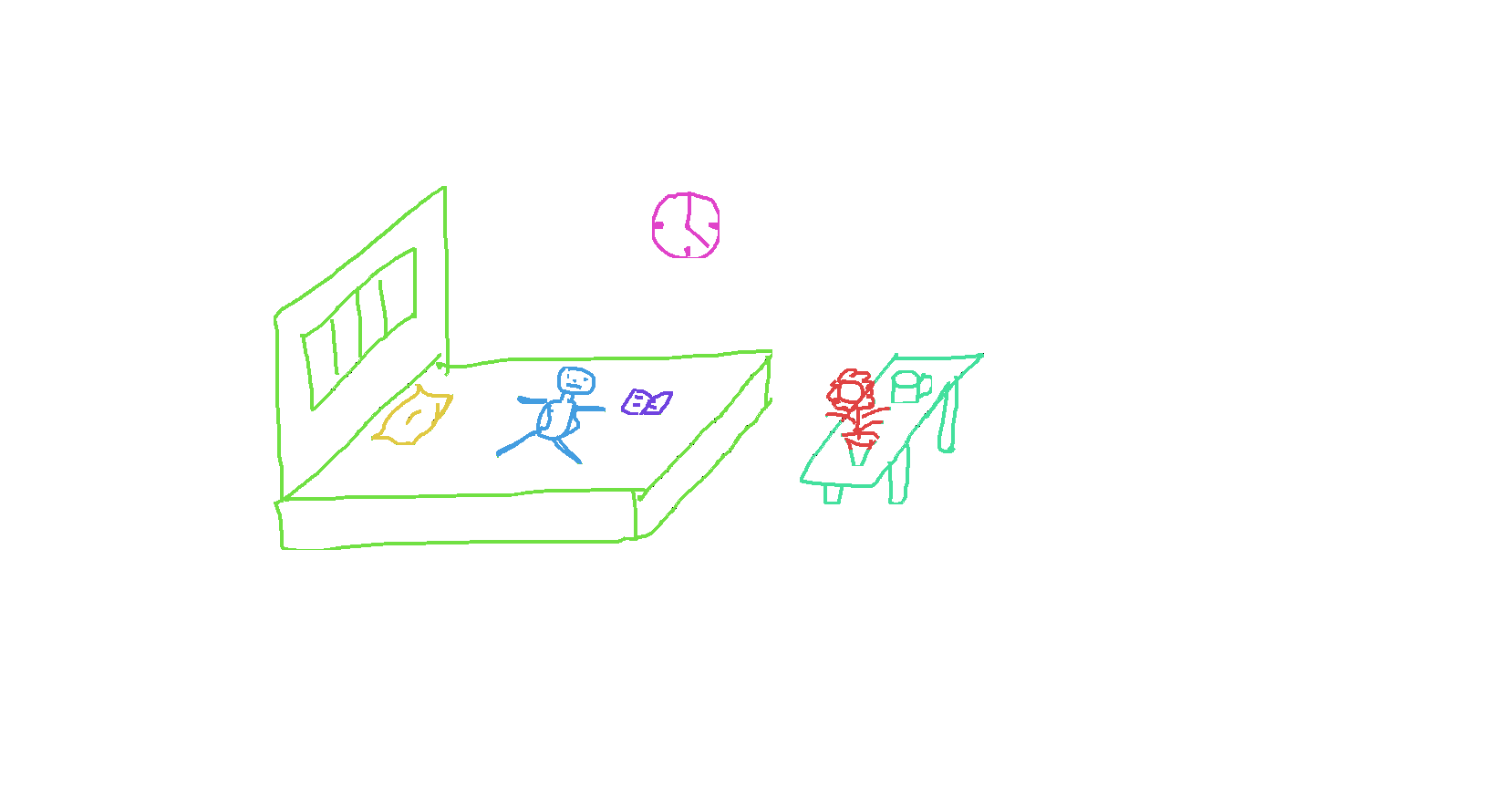}} \\
        & \makecell{bed, book, clock, mug, person,\\ pillow, table, vase}  & & \\

        \frame{\includegraphics[trim=25 250 100 0, clip, width=0.23\linewidth]{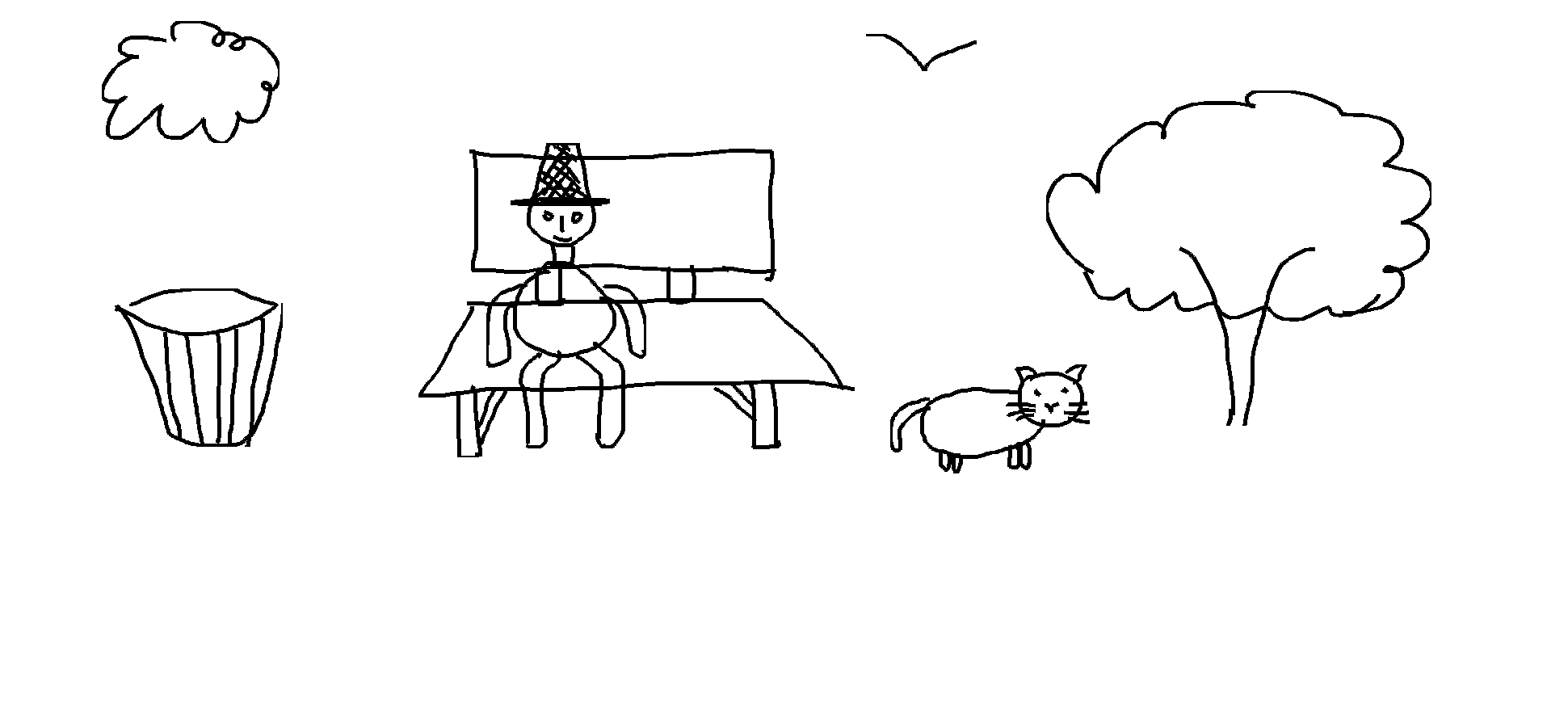}} &
        \frame{\includegraphics[trim=25 250 100 0, clip, width=0.23\linewidth]{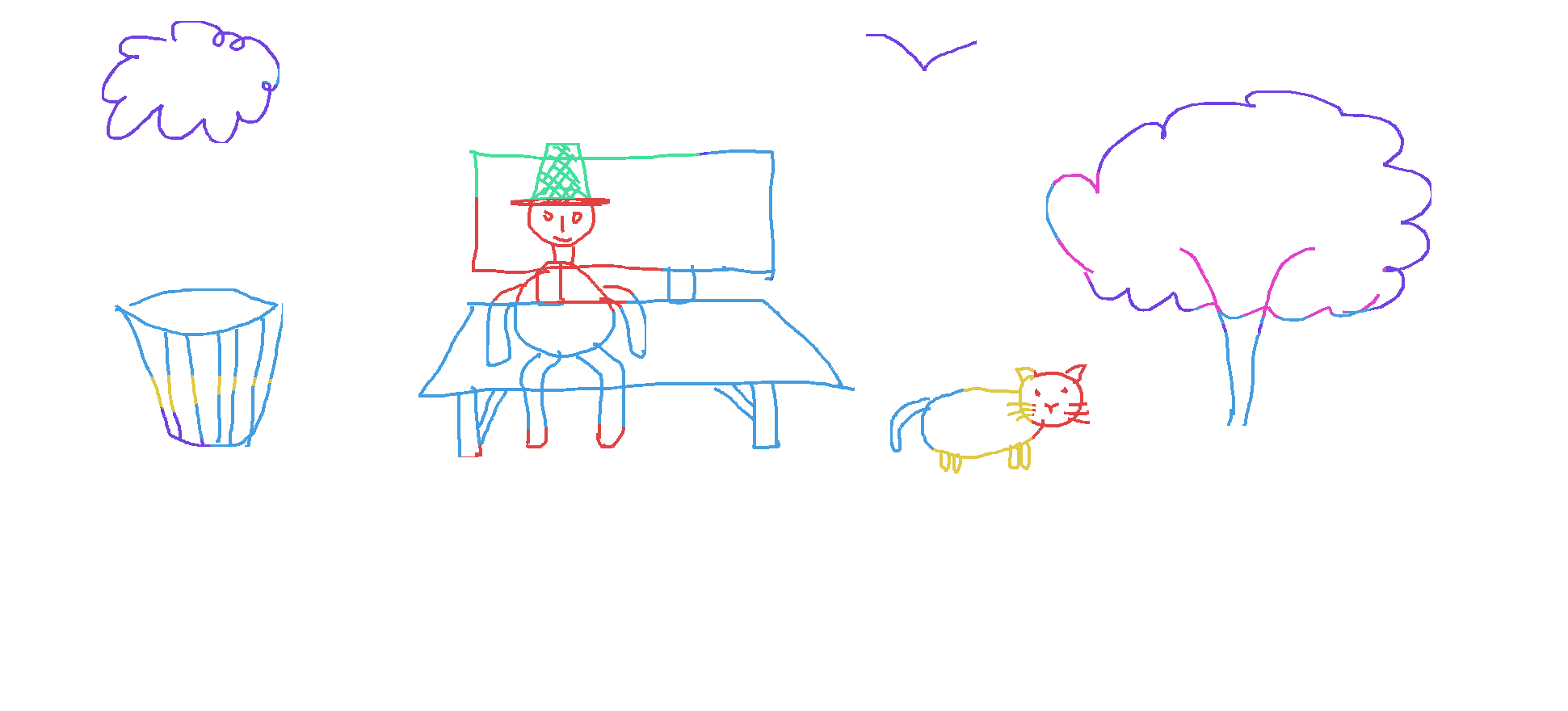}} &
        \frame{\includegraphics[trim=25 250 100 0, clip, width=0.23\linewidth]{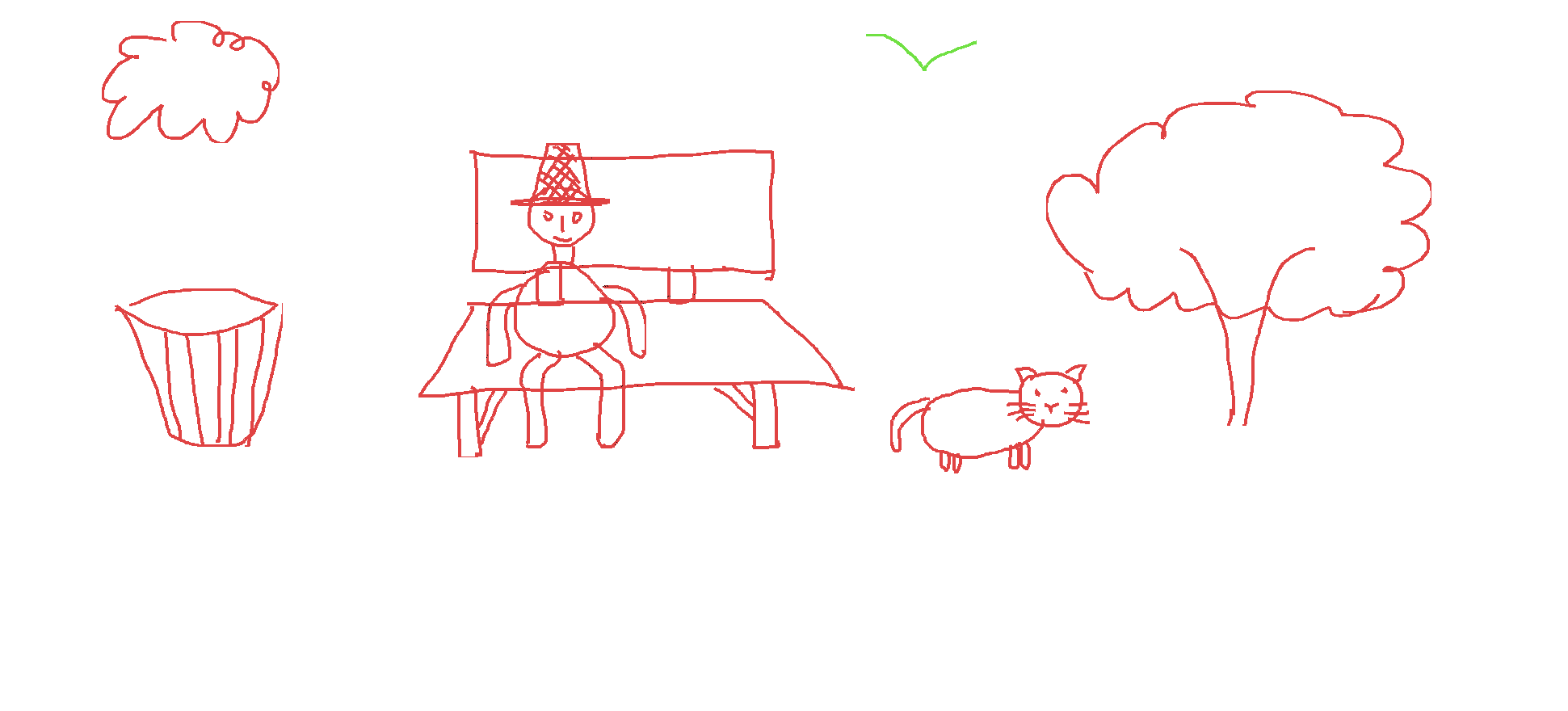}} &
        \frame{\includegraphics[trim=25 250 100 0, clip, width=0.23\linewidth]{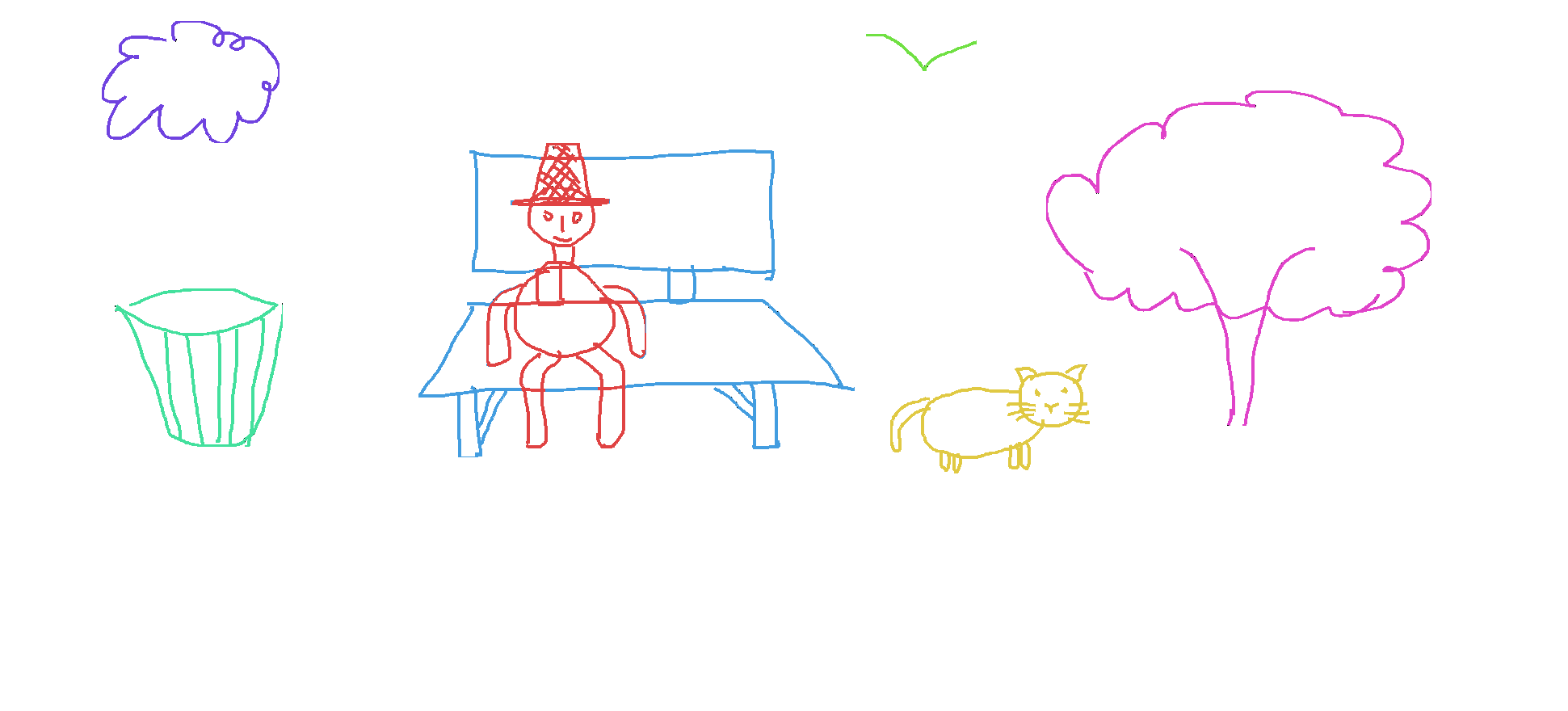}} \\
        & \makecell{bench,  bird, bush,  cat, cloud,\\ hat, person, tree}  & & \\
    \end{tabular}
    }
    \vspace{-0.1cm}
    \caption{\rev{\textbf{Qualitative comparison of segmentation on filtered Zhang \etal dataset. } We prompt Bourouis \etal’s method with ground truth labels and use confidence threshold of 0.01 to ensure all sketch pixels are segmented, while using SketchSeger as-is since it does not accept input labels. Our method accurately segments object instances across diverse scenes, including city life (first and last rows), natural environments (second row), and indoor settings (third to sixth rows). In contrast, baseline methods often struggle to produce clean segmentations for individual objects.}}
    \label{fig:comparison_openvocab_zhangetal}
    }
\end{figure*}

We present additional qualitative comparisons between our approach and baseline methods across  benchmark scene sketch datasets in Fig.~\ref{fig:comparison_instance_sketchyscene} for SketchyScene, Fig.~\ref{fig:comparison_instance_sketchagent} for SketchAgent, Fig.~\ref{fig:comparison_instance_clipasso} for CLIPasso, Fig.~\ref{fig:comparison_instance_instantstyle} for InstantStyle, to accompany our numerical evaluations included in the paper. 
To compare with semantic segmentation method, Bourouis \etal \shortcite{bourouis2024open} and \rev{SketchSeger ~\cite{yang2023sceneHierTransformer}}, we created a filtered version of all eight datasets, where each scene contains at most one instance per object class. These filtered scenes remain challenging for existing methods despite their reduced complexity. We show qualitative results in Fig. ~\ref{fig:comparison_openvocab_sketchyscene} for filtered SketchyScene dataset, Fig. ~\ref{fig:comparison_openvocab_sketchagent} for filtered SketchAgent dataset, and Fig. ~\ref{fig:comparison_openvocab_clipasso} for filtered CLIPasso dataset, Fig. ~\ref{fig:comparison_openvocab_zhangetal} for filtered Zhang \etal dataset,  to accompany our qualitative results shown in the paper. 

\newpage

\newpage

\newpage

\newpage

%%%% FIGS
%%%% dataset: InstantStyle

\newpage
\newpage
\newpage

\newpage

\newpage

\newpage

\newpage

\newpage

\newpage

\begin{figure*}
    \centering
    \setlength{\tabcolsep}{2pt}
    {\small
    \resizebox{0.95\textwidth}{!}{ 
    \begin{tabular}{c @{\hskip 10pt} c c c}
        Input & Bourouis \etal & SketchSeger & \textbf{Ours}  \\
        \frame{\includegraphics[trim=0 0 0 100, clip, width=0.23\linewidth]{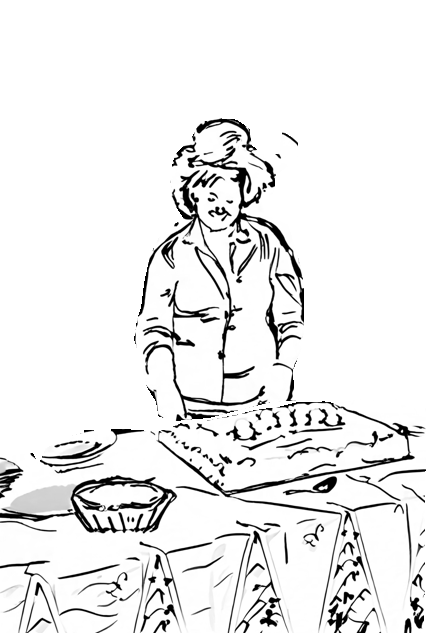}} &
        \frame{\includegraphics[trim=0 0 0 100, clip, width=0.23\linewidth]{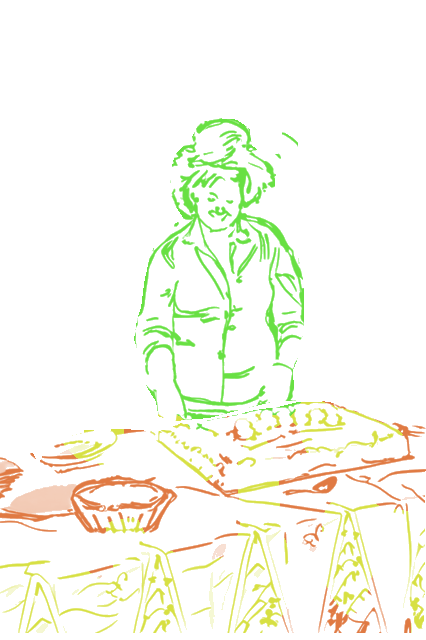}} &
        \frame{\includegraphics[trim=0 0 0 100, clip, width=0.23\linewidth]{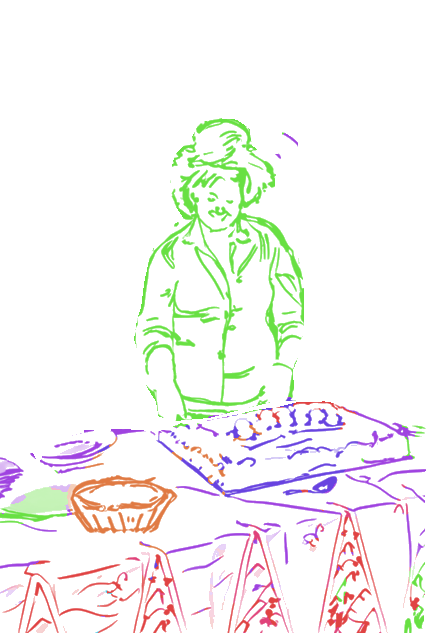}} &
        \frame{\includegraphics[trim=0 0 0 100, clip, width=0.23\linewidth]{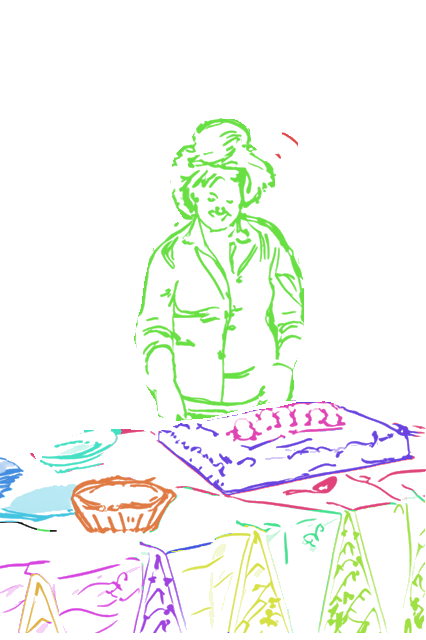}} \\
        & \makecell{person, dining table,  cake}  & & \\

        \frame{\includegraphics[width=0.23\linewidth]{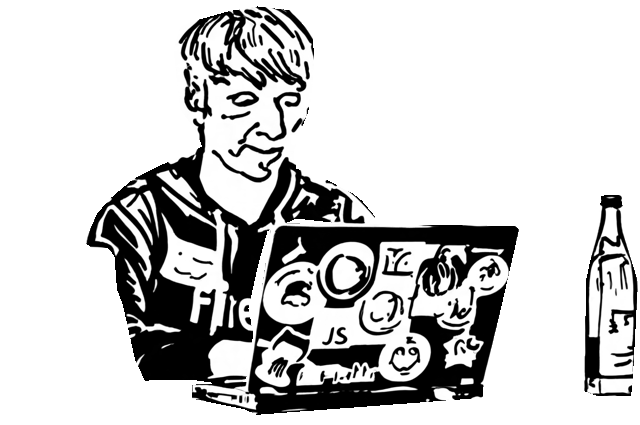}} &
        \frame{\includegraphics[width=0.23\linewidth]{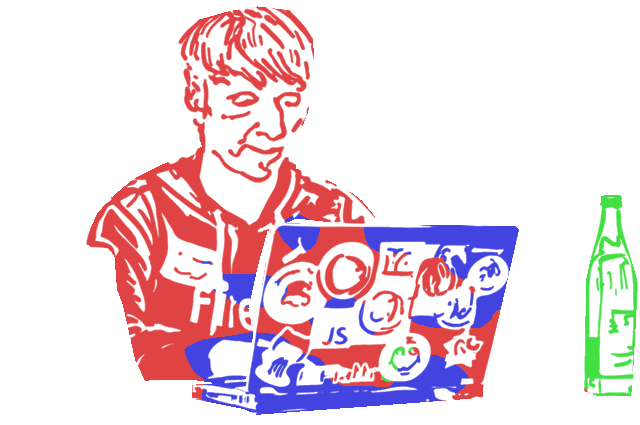}} &
        \frame{\includegraphics[width=0.23\linewidth]{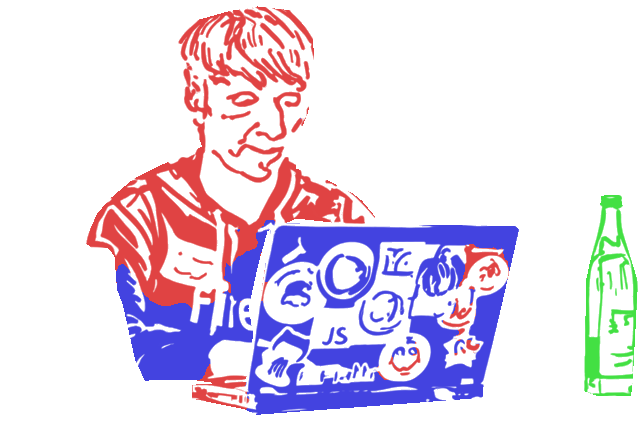}} &
        \frame{\includegraphics[width=0.23\linewidth]{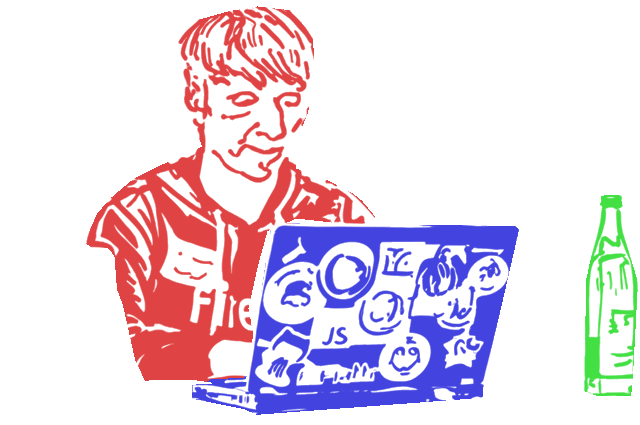}} \\
        & \makecell{person, bottle, laptop}  & & \\

        \frame{\includegraphics[width=0.23\linewidth]{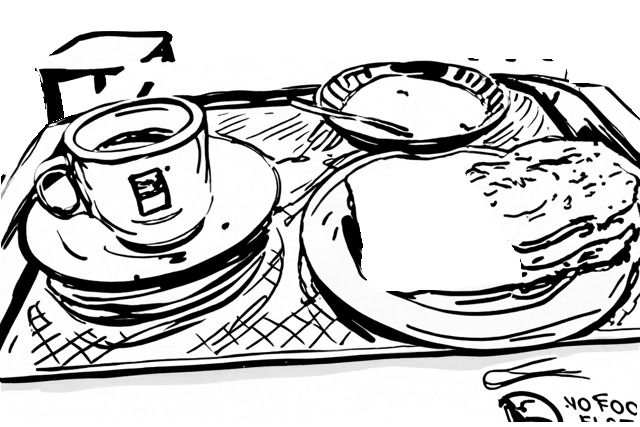}} &
        \frame{\includegraphics[width=0.23\linewidth]{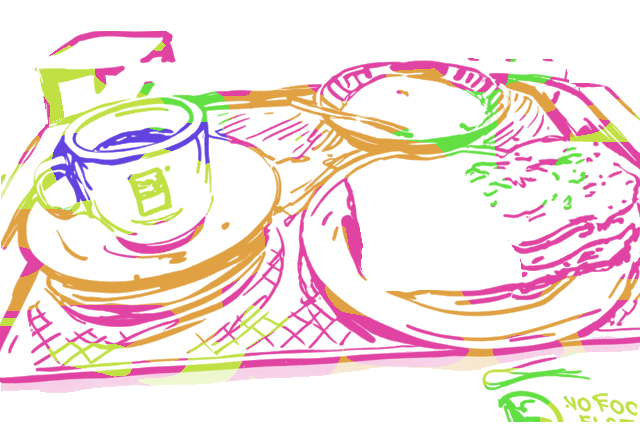}} &
        \frame{\includegraphics[width=0.23\linewidth]{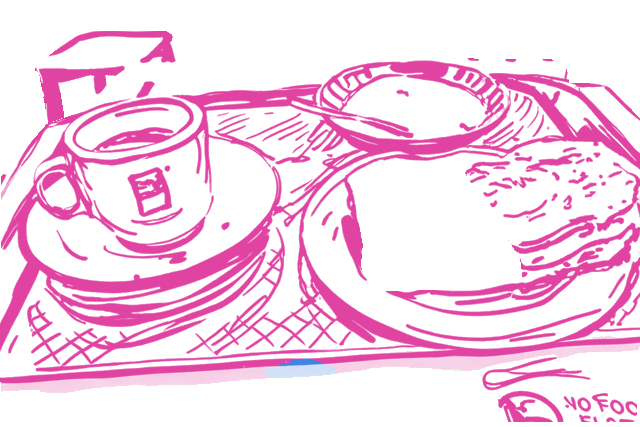}} &
        \frame{\includegraphics[width=0.23\linewidth]{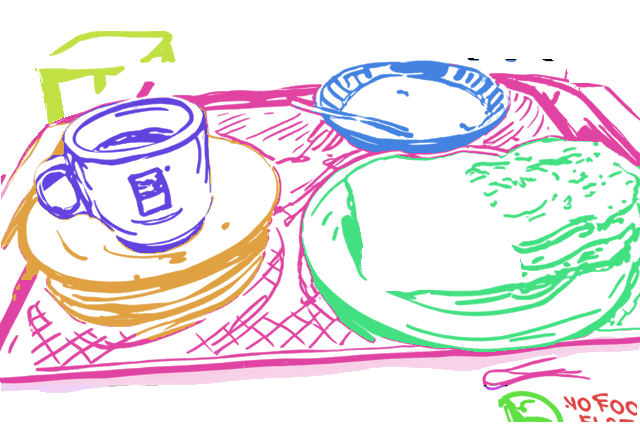}} \\
        & \makecell{bowl, chair, cup, dining table, sandwich}  & & \\

        \frame{\includegraphics[width=0.23\linewidth]{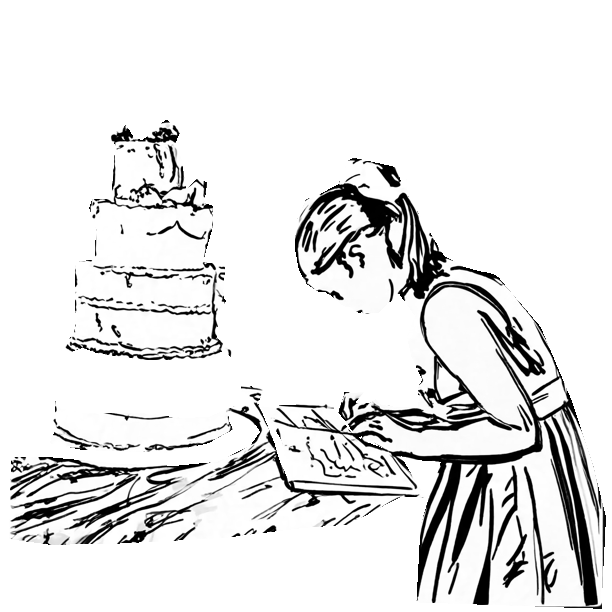}} &
        \frame{\includegraphics[width=0.23\linewidth]{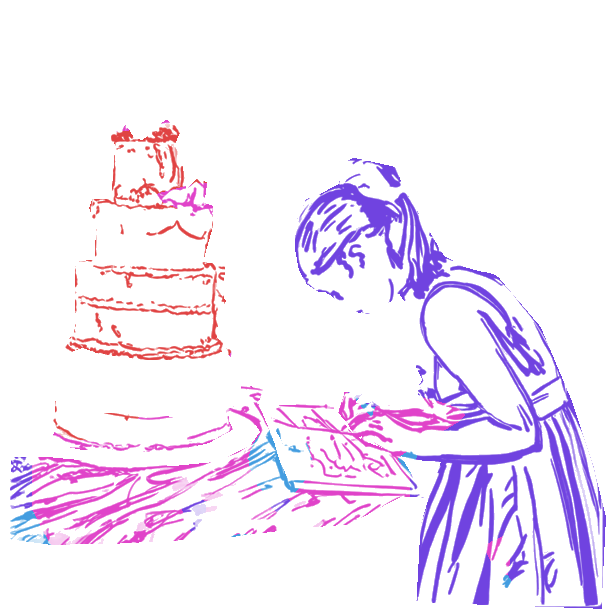}} &
        \frame{\includegraphics[width=0.23\linewidth]{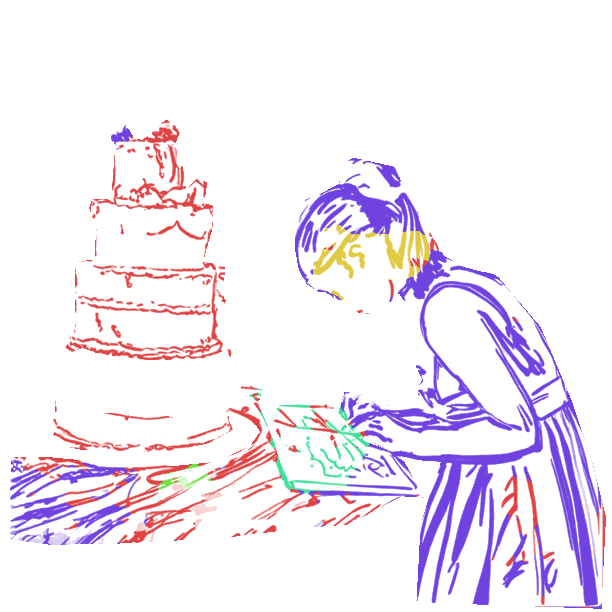}} &
        \frame{\includegraphics[width=0.23\linewidth]{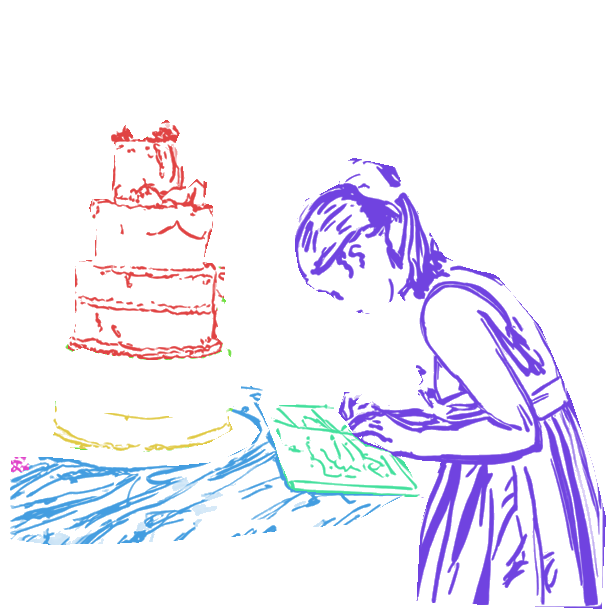}} \\
        & \makecell{book, dining table, cake, person}  & & \\

        \frame{\includegraphics[width=0.23\linewidth]{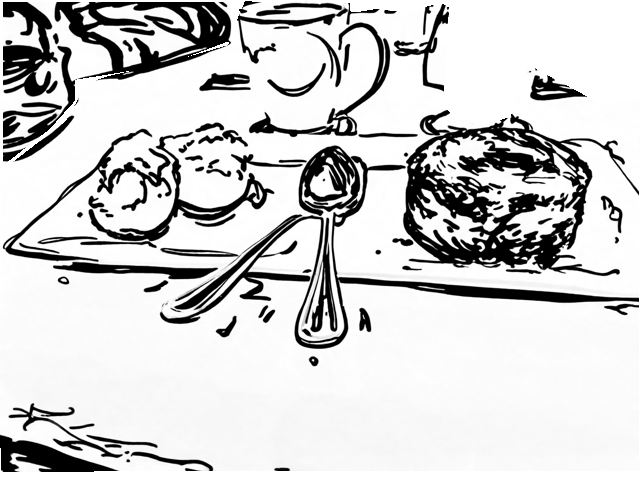}} &
        \frame{\includegraphics[width=0.23\linewidth]{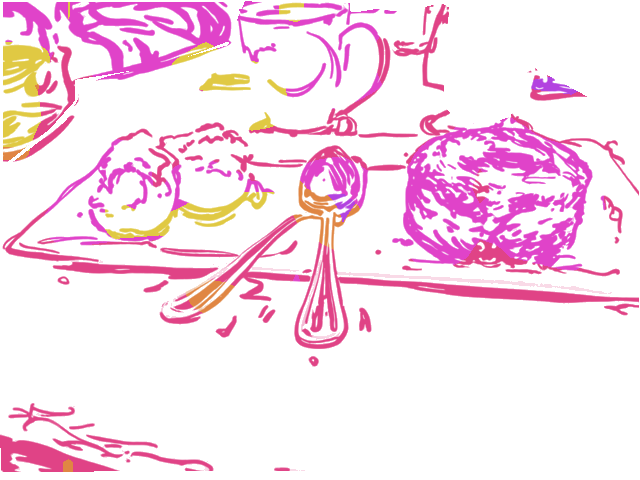}} &
        \frame{\includegraphics[width=0.23\linewidth]{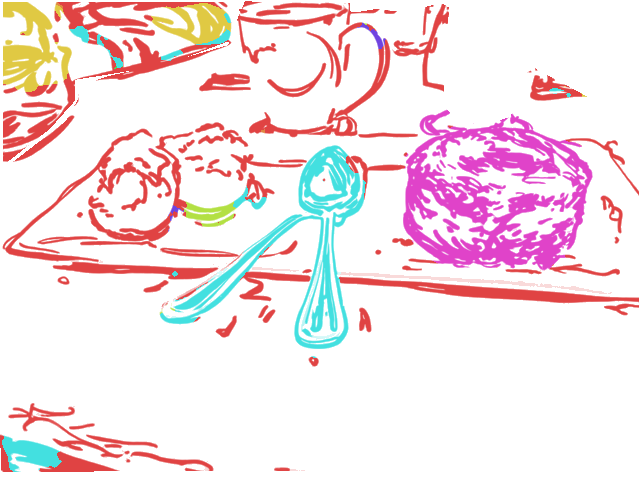}} &
        \frame{\includegraphics[width=0.23\linewidth]{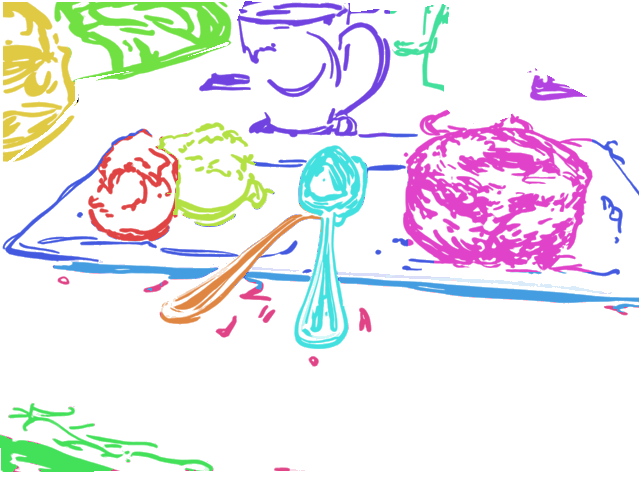}} \\
        & \makecell{cake, cup, dining table, person, spoon}  & & \\

    \end{tabular}
    }
    \vspace{-0.1cm}
    \caption{\rev{\textbf{Qualitative comparison of segmentation on filtered InstantStyle dataset. } We prompt Bourouis \etal’s method with ground truth labels and use confidence threshold of 0.01 to ensure all sketch pixels are segmented, while using SketchSeger as-is since it does not accept input labels. Compared to Bourouis \etal, which sometimes fails to delineate smaller or overlapping objects, and SketchSeger, which often merges semantically distinct regions or misses foreground elements, our method produces cleaner, more precise instance boundaries and better preserves object semantics—particularly in scenes with clutter, occlusion, or fine-grained details.}}
    \label{fig:comparison_openvocab_instanstyle}
    }
\end{figure*}

\end{document}